\documentclass[11pt]{article}

\usepackage{fullpage}
\usepackage{amsmath,amsfonts,amssymb,amsopn,amsthm}
\usepackage{subfig}
\usepackage{url}
\usepackage{graphicx}
\usepackage{cite}

\usepackage{enumerate}
\usepackage{prettyref}

\usepackage{xcolor}
\usepackage{adjustbox}
\usepackage{colortbl}
\usepackage{tabularx}
\usepackage{booktabs} 
\usepackage[linesnumbered, vlined, ruled]{algorithm2e}
\SetArgSty{textup}
\usepackage{color, colortbl}
\usepackage{lscape}
\usepackage{pdflscape}
\usepackage[hidelinks]{hyperref} 
\usepackage{tcolorbox}

 \usepackage{authblk}
\usepackage{geometry}
\newcommand{\argmax}{\mathop{\rm argmax}\limits}
\newcommand{\argmin}{\mathop{\rm argmin}\limits}
\def\vec#1{\mathbf{#1}}
\def\set#1{\mathcal{#1}}

\newcommand\tabblue[1]{{\color[HTML]{1f77b4}{#1}}}
\newcommand\taborange[1]{{\color[HTML]{ff7f0e}{#1}}}
\newcommand\tabgreen[1]{{\color[HTML]{2ca02c}{#1}}}
\newcommand\tabred[1]{{\color[HTML]{d62728}{#1}}}

\newcommand\tabolive[1]{{\color[HTML]{bcbd22}{#1}}}

\newcommand{\pref}{\prettyref}
\newrefformat{fig}{Figure~\ref{#1}}
\newrefformat{tab}{Table~\ref{#1}}
\newrefformat{sec}{Section~\ref{#1}}
\newrefformat{app}{Appendix~\ref{#1}}
\newrefformat{alg}{Algorithm~\ref{#1}}
\newrefformat{property}{Property~\ref{#1}}
\newrefformat{theorem}{Theorem~\ref{#1}}
\newrefformat{corollary}{Corollary~\ref{#1}}
\newrefformat{proposition}{Proposition~\ref{#1}}
\newrefformat{def}{Definition~\ref{#1}}
\newrefformat{eq}{equation~(\ref{#1})}

\usepackage{lscape}

\begin{document}


\title{\LARGE Investigating Normalization in Preference-based Evolutionary Multi-objective Optimization Using a Reference Point}

\author[1]{\normalsize Ryoji Tanabe}
\affil[1]{\normalsize Faculty of Environment and Information Sciences, Yokohama National University, Yokohama, Japan}
\affil[$\ast$]{\normalsize Email: \texttt{rt.ryoji.tanabe@gmail.com}}

\date{}
\maketitle

\vspace{-3ex}
{\normalsize\textbf{Abstract: } Normalization of objectives plays a crucial role in evolutionary multi-objective optimization (EMO) to handle objective functions with different scales, which can be found in real-world problems.
Although the effect of normalization methods on the performance of EMO algorithms has been investigated in the literature, that of preference-based EMO (PBEMO) algorithms is poorly understood.
Since PBEMO aims to approximate a region of interest, its population generally does not cover the Pareto front in the objective space.
This property may make normalization of objectives in PBEMO difficult.
This paper investigates the effectiveness of three normalization methods in three representative PBEMO algorithms.
We present a bounded archive-based method for approximating the nadir point.
First, we demonstrate that the normalization methods in PBEMO perform significantly worse than that in conventional EMO in terms of approximating the ideal point, nadir point, and range of the PF.
Then, we show that PBEMO requires normalization of objectives on problems with differently scaled objectives.
Our results show that there is no clear ``best normalization method'' in PBEMO, but an external archive-based method performs relatively well.

\section{Introduction}
\label{sec:introduction}


Multi-objective optimization generally aims to find a Pareto optimal solution preferred by a decision maker (DM)~\cite{Miettinen98}.
In an \textit{a posteriori} decision-making, the DM selects a single solution from a solution set according to her/his preference.
It is desirable that the solution set approximates the Pareto front (PF) in the objective space well.
Evolutionary multi-objective optimization (EMO)~\cite{Deb01} is generally used to find the solution set for the \textit{a posteriori} decision-making.
Representative EMO algorithms include NSGA-II~\cite{DebAPM02}, IBEA~\cite{ZitzlerK04}, and MOEA/D~\cite{ZhangL07}.

The \textit{a posteriori} decision-making assumes that the DM does not have any specific preference \textit{a priori}.
However, the DM's preference information is available in some cases~\cite{PurshouseDMMW14}.
A preference-based EMO (PBEMO) algorithm~\cite{CoelloCoello00,BechikhKSG15} aims to find a solution set that approximates the region of interest (ROI) in the objective space. 
Here, the ROI is a subregion of the PF defined by the DM's preference information~\cite{RuizSL15}.
PBEMO has at least two advantages compared to conventional EMO.
The one is that PBEMO does not need to approximate the whole PF.
In general, approximating the ROI is easier than approximating the PF.
This advantage is especially noticeable when the number of objectives is large~\cite{LiLDMY20}.
The other is that the PBEMO approach can reduce the DM's cognitive load.
In the \textit{a posteriori} decision-making, the DM examines a solution set to find her/his preferred solution.
Since conventional EMO algorithms (e.g., NSGA-II) aim to approximate the PF, solution sets found by them are likely to include solutions in which the DM is not interested.
Examining many irrelevant solutions increases the burden on decision-making.
In contrast, solution sets found by PBEMO algorithms are likely to include only solutions preferred by the DM.

In the context of PBEMO, the DM's preference information is frequently expressed by a reference point that consists of her/his desired objective values~\cite{Wierzbicki80,LiLDMY20,AfsarMR21}.
In this reference point-based approach, a PBEMO algorithm aims to find a non-dominated solution set that approximates the ROI defined by the reference point in the objective space.
Representative PBEMO algorithms using the reference point include R-NSGA-II~\cite{DebSBC06}, PBEA~\cite{ThieleMKL09}, and MOEA/D-NUMS~\cite{LiCMY18}.

Since objective functions are defined independently from each other (e.g., time vs. price), their scales are different in most real-world applications~\cite{TanabeI20asoc}.
For this reason, EMO algorithms require a normalization method to handle objective functions with different scales~\cite{HeITWNS21}.
Normalization of objectives is an operation that rescales the range of each objective value to $[0,1]$ based on approximated ideal and nadir points.\footnote{Some normalization methods (e.g., \cite{DebJ14}) do not translate each objective value into the range $[0,1]$.}
Since the true ideal and nadir points are generally unavailable before the search, a normalization method needs to approximate them.
Thus, two methods for approximating the ideal and nadir points play an important role in normalization of objectives.
As surveyed in~\cite{HeITWNS21}, a number of normalization methods have been considered in the EMO community.
Some previous studies also investigated the influence of normalization methods on the performance of EMO algorithms.
For example, Ishibuchi et al.~\cite{IshibuchiDN17} investigated the performance of MOEA/D with/without a simple normalization method.
Their results showed that the effectiveness of the normalization method is inconsistent, e.g., MOEA/D with the normalization method outperforms MOEA/D without the normalization method on some test problems and vice versa.
Blank et al.~\cite{BlankDR19} analyzed the performance of NSGA-III~\cite{DebJ14} with three methods for estimating the nadir point on test problems with regular PFs (DTLZ1--4~\cite{DebTLZ05} and SDTLZ1--2~\cite{DebJ14}).
Their results showed the effectiveness of the hyperplane-based estimation method in NSGA-III.
In contrast, He et al.~\cite{HeIS21} demonstrated that the hyperplane-based estimation method performs poorly on a problem with a disconnected PF (DTLZ7~\cite{DebTLZ05}).

\subsubsection*{Motivation}

Unlike EMO, normalization of objectives has received little attention in the field of PBEMO.
Let us briefly review the following six representative PBEMO algorithms: R-NSGA-II~\cite{DebSBC06}, r-NSGA-II~\cite{SaidBG10}, g-NSGA-II~\cite{MolinaSHCC09}, PBEA~\cite{ThieleMKL09}, R-MEAD2~\cite{MohammadiOLD14}, and MOEA/D-NUMS~\cite{LiCMY18}, where their behavior was investigated in~\cite{LiLDMY20}.
The two scalarizing function-based PBEMO algorithms (R-MEAD2 and MOEA/D-NUMS) do not perform normalization of objectives.
In addition, g-NSGA-II does not require any normalization method, where the crowding distance measure in g-NSGA-II implicitly performs normalization of objectives as in NSGA-II.
PBEA is a preference-incorporated version of IBEA.
PBEA uses a preference-based indicator that combines the binary additive $\epsilon$-indicator $I_{\epsilon}$ \cite{ZitzlerK04} and the augmented achievement scalarizing function (AASF) \cite{MiettinenM02}.
Although normalization of objectives is performed in $I_{\epsilon}$, that is not performed in the AASF.
In the six PBEMO algorithms, only R-NSGA-II and r-NSGA-II explicitly use a normalization method.
However, R-NSGA-II and r-NSGA-II estimate the ideal and nadir points based on the minimum and maximum objective values of the union of the current population and offspring for each iteration.
Although this simple normalization method has been commonly used in the EMO community~\cite{HeITWNS21}, its effectiveness has not been investigated in PBEMO.
The population of a PBEMO algorithm generally does not cover the whole PF in the objective space.
This property may influence the performance of the population-based normalization method in R-NSGA-II and r-NSGA-II.

\subsubsection*{Contributions}

Motivated by the above discussion, this paper investigates the influence of normalization methods on the performance of PBEMO algorithms.
We analyze the behavior of three PBEMO algorithms (R-NSGA-II, r-NSGA-II, and MOEA/D-NUMS)\footnote{Here, g-NSGA-II does not require a normalization method. PBEA requires a fine-tuning of a parameter $\delta$ for each problem. Previous studies \cite{LiLDMY20,Tanabe23} showed that R-MEAD2 is unlikely to find a good approximation of the ROI. For these reasons, this paper does not consider g-NSGA-II, PBEA, and R-MEAD2.} with three normalization methods.
We present a computationally cheap archiving method for approximating the nadir point (Section \ref{sec:nm_apd}), where it is not a main contribution of this paper.
This archiving method can behave an unbounded archive-based method.
Through analysis, we address the following two research questions:

\begin{enumerate}[RQ1:]
\item How different is the performance of normalization methods between PBEMO and EMO? Which normalization method is the best in terms of approximating the ideal point, nadir point, and range of the PF?
\item Which normalization method is suitable for PBEMO algorithms?
\end{enumerate}

\subsubsection*{Outline}

Section \ref{sec:preliminaries} gives some preliminaries.
Section \ref{sec:review} introduces the three normalization methods investigated in this paper.
Section \ref{sec:setting} describes our experimental setup.
Section \ref{sec:results} shows the analysis results. 
Section \ref{sec:conclusion} concludes this paper.

\subsubsection*{Supplementary file}

Figure S.$*$ and Table S.$*$ indicate a figure and a table in the supplementary file, respectively.

\section{Preliminaries}
\label{sec:preliminaries}

\subsection{Multi-objective optimization}
\label{sec:def_MOPs}

Let $n$ be the dimension of the solution space $\mathbb{X} \subseteq \mathbb{R}^n$.
Let also $m$ be the dimension of the objective space $\mathbb{R}^m$.
Multi-objective optimization aims to find a solution $\vec{x} \in \mathbb{X}$ that minimizes an $m$-dimensional objective function vector $\vec{f}: \mathbb{X} \rightarrow \mathbb{R}^m$.
There generally does not exist the absolute optimal solution that minimizes all $m$ objective functions.

A solution $\vec{x}_1$ is said to dominate another solution $\vec{x}_2$ if $f_i (\vec{x}_1) \leq f_i (\vec{x}_2)$ for all $i \in \{1, \ldots, m\}$ and $f_i (\vec{x}_1) < f_i (\vec{x}_2)$ for at least one index $i$.
We denote $\vec{x}_1 \prec \vec{x}_2$ when $\vec{x}_1$ dominates $\vec{x}_2$.
In addition, $\vec{x}_1$ is said to weakly dominate $\vec{x}_2$ if $f_i (\vec{x}_1) \leq f_i (\vec{x}_2)$ for all $i \in \{1, \ldots, m\}$.
If a solution $\vec{x}^\ast$ is not dominated by any solution in the solution space $\mathbb{X}$,  $\vec{x}^\ast$ is said to be a Pareto optimal solution.
The set of all Pareto optimal solutions in $\mathbb{X}$ is said to be the Pareto optimal solution set $\set{X}^{*} = \{\vec{x}^* \in  \mathbb{X} \,|\, \nexists \vec{x} \in  \mathbb{X} \: \text{s.t.} \: \vec{x} \prec \vec{x}^* \}$.
The image of the Pareto optimal solution set $\set{X}^{*}$ in the objective space $\mathbb{R}^m$ is also said to be the PF $\vec{f}(\set{X}^{*})$. 
The ideal point $\vec{p}^{\mathrm{ideal}} \in \mathbb{R}^m$ and nadir point $\vec{p}^{\mathrm{nadir}} \in \mathbb{R}^m$ consist of the minimum and maximum values of the PF for $m$ objective functions, respectively.
Thus, $p^{\mathrm{ideal}}_i=\min_{\vec{x}^*\in\set{X}^\ast}\{f_i(\vec{x}^*)\}$ and $p^{\mathrm{nadir}}_i=\max_{\vec{x}^*\in\set{X}^\ast}\{f_i(\vec{x}^*)\}$ for each $i\in\{1,\ldots,m\}$.

\subsection{Preference-based multi-objective optimization using the reference point}
\label{sec:def_PBMOPs}


Conventional EMO aims to find a solution set $\set{X}$ that approximates the PF in the objective space.
In contrast, PBEMO using the reference point $\vec{z}  \in \mathbb{R}^m$ aims to find a solution set that approximates the ROI in the objective space.
Here, the ROI is a subset of the PF defined by the reference point $\vec{z}$.


As pointed out in \cite{TanabeL23}, the term ``ROI'' has been loosely defined in the literature. 
To address this issue, the previous study \cite{TanabeL23} mathematically defined three types of ROI.
This paper considers one out of the three ROIs called the ROI-C in \cite{TanabeL23}.
The ROI-C is the most intuitive ROI.
Let $\vec{x}^{\mathrm{c}*}$ be a Pareto optimal solution that is closest to the reference point $\vec{z}$ in the objective space, i.e., $\vec{x}^{\mathrm{c}*}=\argmin_{\vec{x}^* \in \set{X}^*} \left\{\texttt{distance}\left(\vec{f}(\vec{x}^*),\vec{z}\right) \right\}$.
Mathematically, this ROI is defined as follows:
\begin{align}
  \label{eq:roi}
  \text{ROI}=\left\{\vec{f}(\vec{x}^*) \in \vec{f}(\set{X}^*) \, | \, \texttt{distance}\bigl(\vec{f}(\vec{x}^*), \vec{f}(\vec{x}^{\mathrm{c}*})\bigr)<r\right\},
\end{align}
where $\texttt{distance}(\cdot,\cdot)$ returns the Euclidean distance between two vectors.
In \pref{eq:roi}, a parameter $r$ is supplied by the DM and determines the radius of the ROI.
\pref{fig:roi} shows an example of this ROI on the bi-objective DTLZ2 problem, where the shape of the PF is nonconvex.
As shown in \pref{fig:roi}, the ROI is a set of objective vectors on the PF in a hypersphere of a radius $r$ centered at $\vec{f}(\vec{x}^{\mathrm{c}*})$.

\begin{figure}[t]
  \centering
  \includegraphics[width=0.3\textwidth]{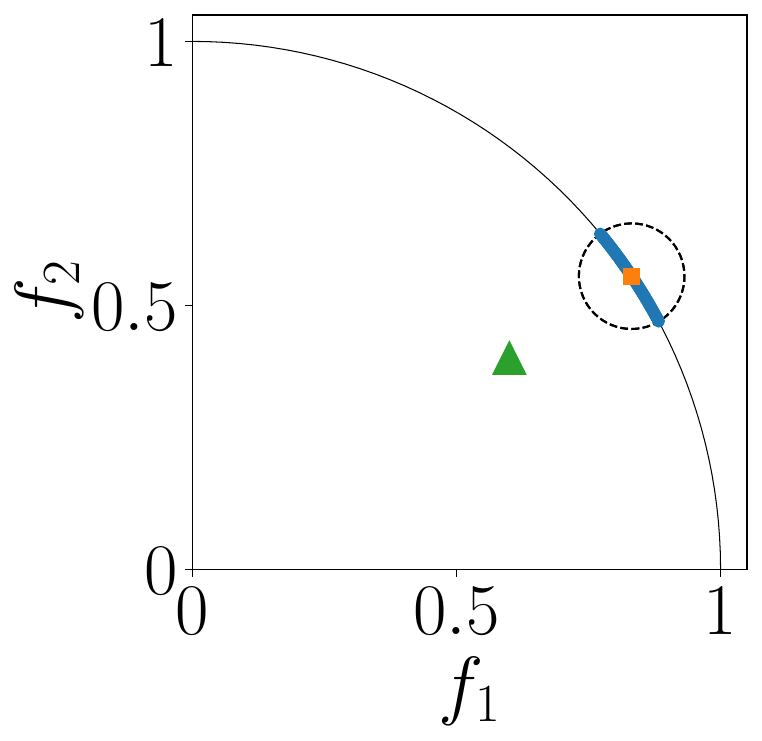}
  \caption{Example of the ROI on the bi-objective DTLZ2 problem, where \tabgreen{$\blacktriangle$} is the reference point $\vec{z}=(0.6,0.4)^\top$, \taborange{$\blacksquare$} is the objective vector of the closet Pareto optimal solution $\vec{f}(\vec{x}^{\mathrm{c}*})$ to $\vec{z}$, and $r=0.1$. A set of objective vectors \tabblue{$\bullet$} in the dotted circle is the ROI.
  }
  \label{fig:roi}
\end{figure}

\subsection{PBEMO algorithms}
\label{sec:pemo_ref}

Here, we briefly describe environmental selection in R-NSGA-II \cite{DebSBC06}, r-NSGA-II \cite{SaidBG10}, and MOEA/D-NUMS \cite{LiCMY18}.
As in \cite{LiLDMY20}, we focus on PBEMO using a single reference point as the first step to analyze normalization methods.
While R-NSGA-II and r-NSGA-II are dominance-based PBEMO algorithms, MOEA/D-NUMS is a scalarizing function-based one.
R-NSGA-II and r-NSGA-II are preference-incorporated versions of NSGA-II.
MOEA/D-NUMS is a straightforward extension of MOEA/D.

Below, the population of size $\mu$ is represented as $\set{P}=\{\vec{x}_i\}_{i=1}^\mu$.
A set of $\lambda$ children is also denoted as $\set{Q}=\{\vec{x}_i\}_{i=1}^\lambda$.
As in \cite{HeITWNS21}, we use the symbols $\vec{z}^{\mathrm{lb}} \in \mathbb{R}^m$ and $\vec{z}^{\mathrm{ub}} \in \mathbb{R}^m$ to represent the estimated ideal and nadir points, respectively.
Here, ``lb'' and ``ub'' in $\vec{z}^{\mathrm{lb}}$ and $\vec{z}^{\mathrm{ub}}$ stand for the lower bounds and upper bounds, respectively.

\subsubsection{R-NSGA-II}
\label{sec:Rnsga2}

Here, we describe environmental selection in R-NSGA-II that selects $\mu$ individuals from the $\mu+\lambda$ individuals in the union of $\set{P}$ and $\set{Q}$.
As in NSGA-II, the primary criterion in R-NSGA-II is based on the non-domination level.
While the secondary criterion in NSGA-II is based on the crowding distance, that in R-NSGA-II is based on the following distance from a solution $\vec{x}$ to $\vec{z}$ in the objective space:
\begin{align}
    \label{eq:wdist}
    d^{\mathrm{R}}(\vec{x}) &=\sqrt{\sum^{m}_{i=1} w_i \left(\frac{f_i(\vec{x}) - z^{\mathrm{lb}}_i}{z^{\mathrm{ub}}_i - z^{\mathrm{lb}}_i} - \frac{z_i - z^{\mathrm{lb}}_i}{z^{\mathrm{ub}}_i - z^{\mathrm{lb}}_i}\right)^2},\notag \\
    &=\sqrt{\sum^{m}_{i=1} w_i\left(\frac{f_i(\vec{x}) - z_i}{z^{\mathrm{ub}}_i - z^{\mathrm{lb}}_i}\right)^2},
\end{align}
where $z^{\mathrm{lb}}_i$ and $z^{\mathrm{ub}}_i$ are the minimum and maximum values of the $i$-th objective function $f_i$ in the union of $\set{P}$ and $\set{Q}$.\footnote{The original work \cite{DebSBC06} describes that $z^{\mathrm{ub}}_i$  and $z^{\mathrm{lb}}_i$ are ``\textit{the population maximum and minimum function values of $i$-th objective}''. However, there is no reasonable reason not to consider  $\set{Q}$ to define $z^{\mathrm{ub}}_i$  and $z^{\mathrm{lb}}_i$.}
The weight vector $\vec{w} \in \mathbb{R}^m$ in \pref{eq:wdist} represents the relative importance of each objective function, where $\sum^m_{i=1} w_i = 1$ and $w_i \geq 0$ for any $i$.
Usually, $\vec{w} = (1/m,\ldots,1/m)^\top$.
%


After non-dominated sorting, multiple individuals may belong to the same non-domination level.
In this case, ties are broken by their $d^{\mathrm{R}}$ values, where a smaller $d^{\mathrm{R}}$ value is preferred.
Thus, a non-dominated individual close to $\vec{z}$ has a priority to survive to the next iteration.
In addition, R-NSGA-II uses $\epsilon$-clearing to maintain the diversity of the population.
If the distance between two individuals in the normalized objective space based on $\vec{z}^{\mathrm{lb}}$ and $\vec{z}^{\mathrm{ub}}$ is less than $\epsilon$, a randomly selected one is removed from the population. 

\subsubsection{r-NSGA-II}
\label{sec:rnsga2}

r-NSGA-II is a preference-incorporated version of NSGA-II by replacing the conventional Pareto dominance relation with an r-dominance relation.
Here, a solution $\vec{x}_1$ is said to r-dominate another solution $\vec{x}_2$ if one of the following two criteria is met: 1) $\vec{x}_1 \prec \vec{x}_2$; 2) $\vec{x}_1 \nprec \vec{x}_2$, $\vec{x}_1 \nsucc \vec{x}_2$, and $d^{\mathrm{r}}(\vec{x}_1, \vec{x}_2) < -\delta$.
Here, $d^{\mathrm{r}}(\vec{x}_1, \vec{x}_2)$ is defined as follows:
\begin{equation}
	\label{eq:rdominance_d}
  	d^{\mathrm{r}} (\vec{x}_1, \vec{x}_2) = \frac{d^{\mathrm{R}}(\vec{x}_1) - d^{\mathrm{R}}(\vec{x}_2)}{\max_{\vec{x} \in \set{P}} \{d^{\mathrm{R}}(\vec{x})\} - \min_{\vec{x} \in \set{P}} \{d^{\mathrm{R}}(\vec{x})\}},
\end{equation}
%
where $d^{\mathrm{R}}$ is defined in \pref{eq:wdist}.
Thus, the normalization methods in R-NSGA-II and r-NSGA-II are the same.
The r-dominance relation requires the threshold $\delta \in [0, 1]$ that determines the spread of individuals in the objective space.
The r-dominance relation with a small $\delta$ value prefers a non-dominated individual close to the reference point $\vec{z}$ in the objective space.
When $\delta = 0$, the r-dominance relation between two individuals is simply based on their $d^{\mathrm{R}}$ values.
In contrast, when $\delta = 1$, the r-dominance relation is identical to the conventional Pareto dominance relation.

\subsubsection{MOEA/D-NUMS}
\label{sec:nums}

Let $\set{W}$ be a set of $\mu$ uniformly distributed weight vectors $\vec{w} \in \mathbb{R}^m$.
In general, scalarizing function-based EMO algorithms (e.g., MOEA/D) require $\set{W}$ before the search.
A nonuniform mapping scheme (NUMS)~\cite{LiCMY18} shifts $\set{W}$ toward the reference point $\vec{z}$.
Let $\set{W}^\prime$ be the shifted version of $\set{W}$ by NUMS.
It is expected that the distribution of the $\mu$ weighted vector in $\set{W}^\prime$ is biased toward $\vec{z}$.
The extent of $\set{W}^\prime$ is determined by a parameter $\tau$ in NUMS.

MOEA/D-NUMS is a preference-incorporated version of MOEA/D by replacing $\set{W}$ with $\set{W}^\prime$.
MOEA/D-NUMS also uses the following augmented achievement scalarizing function (AASF)~\cite{MiettinenM02} as a scalarizing function: 
\begin{equation}
    \label{eq:aasf}
    \mathrm{AASF}(\vec{x})= \max_{i\in\{1,\ldots,m\}} w_i \left(f_i(\vec{x})-z_i\right) + \rho\sum_{i=1}^m\left(f_i(\vec{x})-z_i \right), 
\end{equation}
where $\rho$ is a small positive value, e.g., $\rho=10^{-6}$.
Similar to \pref{eq:wdist}, the weight vector $\vec{w}$ determines the relative importance of each objective function.
Note that a solution with the minimum AASF value is Pareto-optimal with respect to $\mathbf{z}$ and $\mathbf{w}$~\cite{MiettinenM02}.

\subsection{Normalization in EMO}
\label{sec:nm}

Here, we describe general normalization methods in the EMO literature.
Below, we consider normalization of objectives in environmental selection for each iteration.
For each $i \in \{1, \ldots, m\}$, the \textit{perfect} normalization method translates the $i$-th objective value of a solution $\vec{x}$ as follows:
\begin{align}
  \label{eq:ideal_norm}
    f^{\prime}_i(\vec{x}) = \frac{f_i(\vec{x}) - z^{\mathrm{ideal}}_i}{z^{\mathrm{nadir}}_i - z^{\mathrm{ideal}}_i}.
\end{align}
where $\vec{z}^{\mathrm{ideal}} \in \mathbb{R}^m$ and $\vec{z}^{\mathrm{nadir}} \in \mathbb{R}^m$
are the true ideal point and nadir point, respectively.
After normalization by \pref{eq:ideal_norm}, it is expected that $m$ normalized objective values $f'_1(\vec{x}), \ldots, f'_m(\vec{x})$ have the same magnitude.

Unfortunately, the perfect normalization method in \pref{eq:ideal_norm} is unavailable in real-world applications.
This is because the true ideal and nadir points are unknown in most real-world problems.
For this reason, approximated ideal $\vec{z}^{\mathrm{lb}} \in \mathbb{R}^m$ and nadir points $\vec{z}^{\mathrm{ub}} \in \mathbb{R}^m$ are used instead of $\vec{z}^{\mathrm{ideal}}$ and $\vec{z}^{\mathrm{nadir}}$ as follows:
\begin{align}
  \label{eq:practical_norm}
    f^{\prime}_i(\vec{x}) = \frac{f_i(\vec{x}) - z^{\mathrm{lb}}_i}{z^{\mathrm{ub}}_i - z^{\mathrm{lb}}_i}.
\end{align}
The remaining question is how to find $\vec{z}^{\mathrm{lb}}$ and $\vec{z}^{\mathrm{ub}}$.
It is desired that $\vec{z}^{\mathrm{lb}}$ and $\vec{z}^{\mathrm{ub}}$ approximate the ideal and nadir points well, respectively.
Since the normalization result in \pref{eq:practical_norm} depends on the choice of $\vec{z}^{\mathrm{lb}}$ and $\vec{z}^{\mathrm{ub}}$, it can be considered that a normalization method consists of two methods for obtaining $\vec{z}^{\mathrm{lb}}$ and $\vec{z}^{\mathrm{ub}}$.
As surveyed in~\cite{HeITWNS21}, a number of normalization methods (i.e., approximation methods of the ideal and nadir points) have been proposed in the EMO community.
In general, a normalization method adaptively adjusts $\vec{z}^{\mathrm{lb}}$ and $\vec{z}^{\mathrm{ub}}$ based on solutions found by an EMO algorithm.

According to \cite{HeITWNS21}, most previous studies used the following two simple methods for approximating the ideal point for each $i \in \{1, \ldots, m\}$:
\begin{align}
  \label{eq:est_ide_pop}
  z^{\mathrm{lb}}_i &= \min_{\vec{x} \in \set{P} \cup \set{Q}} \left\{f_i(\vec{x})\right\},\\
  \label{eq:est_ide_all}
  z^{\mathrm{lb}}_i &= \min_{\vec{x} \in \set{A}} \left\{ f_i(\vec{x})\right\},
\end{align}
where $\set{A}$ in \pref{eq:est_ide_all} is a set of non-dominated solutions found so far.\footnote{Technically, as in MOEA/D, \pref{eq:est_ide_all} can be implemented by simply maintaining the minimum value for each objective function. In this case, $z^{\mathrm{lb}}_i$ is updated for each iteration as follows~\cite{HeITWNS21}: $z^{\mathrm{lb}}_i := \min \left\{z^{\mathrm{lb}}_i, \min_{\vec{x} \in \set{P} \cup \set{Q}} \{ f_i(\vec{x})\} \right\}.$}
In \pref{eq:est_ide_pop}, $z^{\mathrm{lb}}_i$ is the minimum objective value of the union of $\set{P}$ and $\set{Q}$ for the $i$-th objective function.
In \pref{eq:est_ide_all}, $z^{\mathrm{lb}}_i$ is the minimum objective value of all non-dominated solutions found so far for the $i$-th objective function.
While $z^{\mathrm{lb}}_i$ in \pref{eq:est_ide_pop} can be changed for every iteration, $z^{\mathrm{lb}}_i$ in \pref{eq:est_ide_all} monotonically decreases.

In general, approximating the nadir point is more difficult than approximating the ideal point~\cite{DebMC10}.
For each $i \in \{1, \ldots, m\}$, the simplest method estimates the $i$-th value of the nadir point as follows:
\begin{align}
  \label{eq:est_nad_pop}
  z^{\mathrm{ub}}_i = \max_{\vec{x} \in \set{P} \cup \set{Q}} \left\{f_i(\vec{x})\right\}.
\end{align}
Thus, \pref{eq:est_nad_pop} simply sets $z_i^{\mathrm{ub}}$ to the maximum value of the union of $\set{P}$ and $\set{Q}$ for the $i$-th objective function.
The previous study \cite{HeITWNS21} describes that some previous studies used only non-dominated solutions in $\set{P} \cup \set{Q}$ to obtain $\vec{z}^{\mathrm{ub}}$.
%
According to \cite{HeITWNS21}, a few previous studies (e.g., \cite{WangZILZ18}) approximate the nadir point based on a set of all non-dominated solutions found so far $\set{A}$ as follows:
\begin{align}
  \label{eq:est_nad_all}
  z^{\mathrm{ub}}_i = \min_{\vec{x} \in \set{A}} \left\{f_i(\vec{x})\right\}.
\end{align}
In equations \eqref{eq:est_nad_pop} and \eqref{eq:est_nad_all}, if $\set{P} \cup \set{Q}$ and $\set{A}$ are the Pareto optimal solution set, $\vec{z}^{\mathrm{ub}}$ is the same as the true nadir point $\vec{z}^{\mathrm{nadir}}$.

NSGA-II and IBEA approximate the ideal and nadir points using equations \eqref{eq:est_ide_pop} and \eqref{eq:est_nad_pop}, respectively.
However, as pointed out in \cite{HeITWNS21}, equations \eqref{eq:est_ide_pop} and \eqref{eq:est_nad_pop} in NSGA-II are essentially the same as equations \eqref{eq:est_ide_all} and \eqref{eq:est_nad_all}, respectively.
This is because the population $\set{P}$ in NSGA-II always maintains non-dominated individuals with minimum and maximum objective values for each $i \in \{1, \ldots, m\}$ found so far, where it is assumed that $\mu \geq 2 \times m$.
Unlike NSGA-II and IBEA, the most general version of MOEA/D uses equations \eqref{eq:est_ide_all} and \eqref{eq:est_nad_pop} to approximate the ideal and nadir points, respectively.
In summary, different EMO algorithms use different normalization methods.


\section{Three normalization methods investigated in this work}
\label{sec:review}

This section describes three normalization methods in PBEMO investigated in this work.
Since it is not realistic to investigate all normalization methods presented so far, we consider three simple methods as the first step.
Due to the issue found in \cite{HeIS21}, we do not consider the hyperplane-based normalization method~\cite{DebJ14}.

Sections \ref{sec:nm_pq}--\ref{sec:nm_apd} describe the \underline{p}opulation-based normalization method (PP), the \underline{b}est-so-far objective values- and \underline{p}opulation-based normalization method (BP), and the \underline{b}est-so-far objective values- and external \underline{a}rchive-based normalization method (BA), respectively.
\pref{tab:nm} shows the properties of the three normalization methods, which use different methods for approximating the ideal and nadir points.
Note that the terms PP, BP, and BA have not commonly used in the literature, but this paper uses them for the sake of clarity.
After obtaining $\vec{z}^{\mathrm{lb}}$ and $\vec{z}^{\mathrm{ub}}$, the three methods perform normalization of objectives using \pref{eq:practical_norm}.
The three normalization methods (PP, BP, and BA) can be incorporated into any PBEMO algorithm in a plug-in manner.
\pref{sec:results} analyzes the effectiveness of the three normalization methods by incorporating them into R-NSGA-II, r-NSGA-II, and MOEA/D-NUMS.


\begin{table}[t]  
\setlength{\tabcolsep}{4.5pt} 
\renewcommand{\arraystretch}{0.65}
\centering
\caption{Components of the three normalization methods investigated in this paper.}
{\footnotesize
  \label{tab:nm}
\scalebox{1}[1]{ 
\begin{tabular}{cccc}
\toprule 
Method & Section & Approx. of the ideal point & Approx. of the nadir point\\
\midrule
PP & \pref{sec:nm_pq} &  \pref{eq:est_ide_pop} & \pref{eq:est_nad_pop}\\
BP & \pref{sec:nm_bp} &  \pref{eq:est_ide_all} & \pref{eq:est_nad_pop}\\
BA & \pref{sec:nm_apd} &  \pref{eq:est_ide_all} & \pref{eq:est_nad_all}\\
\bottomrule 
\end{tabular}
}
}
\end{table}





\subsection{PP}
\label{sec:nm_pq}

PP is one of the most basic normalization methods for EMO. 
PP uses equations \eqref{eq:est_ide_pop} and \eqref{eq:est_nad_pop} to approximate the ideal and nadir points, respectively.
Thus, PP uses only information from the current population for normalization of objectives.
As described in \pref{sec:pemo_ref}, R-NSGA-II and r-NSGA-II use PP when calculating the weighted distance in \pref{eq:wdist}.
In addition, R-NSGA-II uses PP when performing $\epsilon$-clearing.
%

\subsection{BP}
\label{sec:nm_bp}

BP is a combination of equations \eqref{eq:est_ide_all} and \eqref{eq:est_nad_pop}, which has been used in variants of MOEA/D.
While BP approximates the ideal point based on the best-so-far objective values, it approximates the nadir point based on the maximum values of the population and offspring population.
Thus, the difference between PP and BP is the approximation method of the ideal point.

Although BP is one of the general normalization methods in variants of MOEA/D, it has not been used in PBEMO as far as we know.
The population of a PBEMO algorithm aims to converge to the ROI, which is a subregion of the PF.
Thus, the spread of the population on the whole PF is generally poor in PBEMO.
For this reason, PP may fail to approximate the ideal and nadir points.
In contrast, BP approximates the ideal point based on the best-so-far objective values, which do not depend on the distribution of the current population and offspring population.
Thus, it is expected that BP can approximate the ideal point more accurately than PP.

One may think that PBEMO algorithms generate only solutions close to the ROI in the objective space.
In this case, BP and PP is likely to behave similarly.
However, a previous study \cite{Tanabe23} reported that a PBEMO algorithm can unintentionally generate non-dominated solutions far from the ROI in the objective space.
Although such outlier solutions are usually removed from the population by environmental selection in PBEMO, their objective values can be maintained in BP and BA (described in \pref{sec:nm_apd} later). 

\subsection{BA}
\label{sec:nm_apd}

PP and BP approximate the nadir point based on the maximum values of the population $\set{P}$ and offspring population $\set{Q}$.
Since the population $\set{P}$ of a PBEMO algorithm does not maintain non-dominated solutions far from the ROI, the approximation quality of the nadir point based on $\set{P}$ and $\set{Q}$ can deteriorate during the search. 
This issue can possibly be addressed by using a set of all non-dominated solutions found so far $\set{A}$ as in \pref{eq:est_nad_all}.
Motivated by the above discussion, this paper investigates the performance of BA, which is a combination of equations \eqref{eq:est_ide_all} and \eqref{eq:est_nad_all}.
The difference between BP and BA is the approximation method of the nadir point.

However, \pref{eq:est_nad_all} is time-consuming.
Let $\set{A}$ be a set of all non-dominated solutions found so far by a PBEMO algorithm.
The maintenance of $\set{A}$ requires high computational cost \cite{LiLY23}.
For each iteration, updating $\set{A}$ requires $\mathcal{O}(m|\set{A}|^2)$ in the worst case.
Finding the $m$ maximum objective values of $\set{A}$ also requires $\mathcal{O}(m|\set{A}|)$.
The size of $\set{A}$ can also be very large as the search progresses, especially for many objectives.

To address this issue, we here present a computationally cheap archiving method that uses a \textit{bounded} external archive $\set{B}$ of size $m$.
The update of \pref{eq:est_nad_all} requires only $m$ non-dominated solutions with the maximum objective values.
Thus, it is not necessary for \pref{eq:est_nad_all} to use $\set{A}$.
Simply, $\set{B}$ can be used in \pref{eq:est_nad_all} instead of $\set{A}$.
Note that the same results can be obtained even if $\set{B}$ is used in \pref{eq:est_nad_all} instead of $\set{A}$.
$\set{B}$ is a subset of $\set{A}$.

\IncMargin{0.5em}
\begin{algorithm}[t]
\SetSideCommentRight
\SetKwInOut{Input}{input}
\SetKwInOut{Output}{output}
\Input{A bounded archive $\set{B}$ and a set of new solutions $\set{X}$}
%
$\set{Y} \leftarrow \mathtt{select\_nondom\_solutions}(\set{B} \cup \set{X})$\;
$\set{B} \leftarrow \emptyset$\;
\For{$i \in \{1, \ldots, m\}$}{
  $\vec{x}^{\mathrm{max}} \leftarrow \argmax_{\vec{x} \in \set{Y}} \left\{ f_i(\vec{x}) \right\}$\;
  $\set{B} \leftarrow \set{B} \cup \{\vec{x}^{\mathrm{max}} \}$\;
}
\Return $\mathcal{B}$\;
\caption{The update method of the bounded external archive}
\label{alg:update_bea}
\end{algorithm}
\DecMargin{0.5em}

\pref{alg:update_bea} shows the method for updating the bounded external archive $\set{B}$ using a new solution set $\set{X}$.
Here, $\set{X}$ is either the initial population $\set{P}$ or the offspring population $\set{Q}$.
In any case, it is assumed that $|\set{X}| \geq 1$.
First, all non-dominated solutions in the union of $\set{B}$ and $\set{X}$ are added to a temporary solution set $\set{Y}$ (line 1).
After that, all solutions in $\set{B}$ are discarded (line 2).
For each $i \in \{1, \ldots, m\}$, a solution $\vec{x}^{\mathrm{max}}$ with the maximum value of the $i$-th objective function is added to $\set{B}$ (lines 4--5).
Here, $\set{B}$ can contain duplicate non-dominated solutions.
After these operations, $\set{B}$ include $m$ non-dominated solutions with the maximum values of the $m$ objective functions.

\section{Experimental setup}
\label{sec:setting}

This section describes the experimental setup in this work.
As mentioned in \pref{sec:introduction}, we focused on R-NSGA-II, r-NSGA-II, and MOEA/D-NUMS.
We used the source code of the three PBEMO algorithms provided by the authors of~\cite{LiLDMY20}.
We incorporated the three normalization methods (PP, BP, and BA) into R-NSGA-II and r-NSGA-II by replacing the definition of $\vec{z}^{\mathrm{lb}}$ and $\vec{z}^{\mathrm{ub}}$.
For MOEA/D-NUMS, we normalize the reference point $\vec{z}$ and the objective vector $\vec{f}(\vec{x})$ in \pref{eq:aasf} by $\vec{z}^{\mathrm{lb}}$ and $\vec{z}^{\mathrm{ub}}$ found by each normalization method.
We performed 31 independent runs for each PBEMO algorithm.
We set the population size $\mu$ to $100$.
We set other parameters of the three PBEMO algorithms according to~\cite{LiLDMY20}.
Note that R-NSGA-II and r-NSGA-II use the most general genetic algorithm (GA) operator (i.e., a combination of simulated binary crossover and polynomial mutation~\cite{DebA95}).
In contrast, MOEA/D-NUMS uses the differential evolution (DE) operator~\cite{StornP97} and polynomial mutation.
We generated the weight vector set in MOEA/D-NUMS using the source code provided by the authors of \cite{LiCMY18}.
We also generated the original weight vector set using the method proposed in \cite{BlankDDBS21}, which can generate the weight vector set of any size.
Here, we used the \texttt{pymoo} implementation \cite{BlankD20}.


\begin{table}[t]
\begin{center}
  \caption{Properties of the DTLZ test problems.
  }
{\scriptsize
  \label{tab:dtlz}
\begin{tabular}{ccccc}
\toprule
Problem & Scale & Shape of the PF & Multimodality & Others\\
\midrule
DTLZ1 & None & Linear & $\checkmark$ & \\
DTLZ2 & None & Nonconvex &  &\\
DTLZ3 & None & Nonconvex & $\checkmark$& \\
DTLZ4 & None & Nonconvex & & Biased\\
DTLZ5 & Weak & Partially degenerate & &\\
DTLZ6 & Weak & Partially degenerate & &\\
DTLZ7 & Weak & Disconnected & $\checkmark$ &\\
\midrule
SDTLZ1 & Strong & Linear & $\checkmark$ & \\
SDTLZ2 & Strong & Nonconvex &  &\\
SDTLZ3 & Strong & Nonconvex & $\checkmark$& \\
SDTLZ4 & Strong & Nonconvex & & Biased\\
\midrule
IDTLZ1 & None & Inverted, linear & $\checkmark$ & \\
IDTLZ2 & None & Inverted, convex &  &\\
IDTLZ3 & None & Inverted, convex & $\checkmark$& \\
IDTLZ4 & None & Inverted, convex & & Biased\\
\bottomrule
\end{tabular}
}
\end{center}
\end{table}

\pref{tab:dtlz} shows the properties of the test problems used in this work.
We used the DTLZ1--DTLZ7~\cite{DebTLZ05} problems.
We set the number of objectives $m$ as follows: $m \in \{2, 3, 4, 5, 6\}$.
We set the number of decision variables as in \cite{DebTLZ05}.
While the scales of the objective functions are the same in the DTLZ1--DTLZ4 problems, those are slightly different in the DTLZ5--DTLZ7 problems.
In addition, we used SDTLZ1--SDTLZ4~\cite{DebJ14} and IDTLZ1--IDTLZ4~\cite{JainD14} problems.
SDTLZ$*$ is a modified version of DTLZ$*$ so that it has differently scaled objective values by multiplying $10^{i-1}$ to the $i$-th objective function $f_i$ for $i \in \{1, \ldots, m\}$.
For example, the nadir point of the three-objective SDTLZ2 is $(1, 10, 100)^{\top}$.
We investigate effects of normalization methods by comparing the results on DTLZ$*$ and SDTLZ$*$.
IDTLZ$*$ is a modified version of DTLZ$*$ so that it has an inverted triangular PF.
A previous study \cite{HeIS21} showed that some estimation methods of the nadir point do not work well for IDTLZ$*$.

We used two settings of the reference point as in~\cite{LiLDMY20}.
The one is called a ``balanced'' setting, where the reference point is close to the center of the PF.
The other is called an ``extreme'' setting, where the reference point is close to an extreme region of the PF.
\pref{tab:ref_points_dtlz} shows the two settings of the reference point.
However, the results for the ``extreme'' setting were not notable in our preliminary study.
For this reason, this paper focuses only on the results for the ``balanced'' setting.

We used the IGD$^+$-C \cite{Tanabe23} indicator, which was specially designed for evaluating the performance of PBEMO algorithms.
IGD$^+$-C is a Pareto-compliant version of IGD-C \cite{MohammadiOLD14,TanabeL23} by replacing IGD \cite{CoelloS04} with IGD$^+$ \cite{IshibuchiMTN15}.
%
Let $\mathcal{S}$ be a set of IGD-reference points $\mathbf{s} \in \mathbb{R}^m$ that are uniformly distributed on the PF.
While IGD$^+$ uses $\mathcal{S}$, IGD$^+$-C uses its subset $\mathcal{S}^{\prime} \subseteq \mathcal{S}$.
Let $\mathbf{s}^{\mathrm{c}}$ be the closest point in $\mathcal{S}$ to $\mathbf{z}$, i.e., $\mathbf{s}^{\mathrm{c}}=\argmin_{\mathbf{s} \in \mathcal{S}} \left\{\texttt{distance}\left(\mathbf{s},\mathbf{z}\right) \right\}$.
$\mathcal{S}^{\prime}$ is a set of all IGD-reference points in the region of a hypersphere of radius $r$ centered at $\mathbf{s}^{\mathrm{c}}$, i.e., $\mathcal{S}^{\prime} = \{\mathbf{s} \in \mathcal{S} \, | \, \texttt{distance}(\mathbf{s}, \mathbf{s}^{\mathrm{c}}) < r \}$.
In the example in \pref{fig:roi}, all IGD-reference points in $\mathcal{S}^{\prime}$ should be in the dotted circle.
%
The IGD$^+$-C value of a solution set $\mathcal{X}$ is calculated as follows:
\begin{align}
      \mathrm{IGD}^*{\mathrm{\mathchar`-C}} (\mathcal{X}) &= \frac{1}{|\mathcal{S}'|} \left(\sum_{\mathbf{s} \in \mathcal{S}'} \min_{\mathbf{x} \in \mathcal{X}} \Bigl\{ \mathtt{dist\_plus}\bigl(\mathbf{x}, \mathbf{s}\bigr) \Bigr\} \right), \notag\\
\mathtt{dist\_plus}(\mathbf{x}, \mathbf{s}) &= \sqrt{\sum^m_{i=1} \bigl(\max\{f_i (\mathbf{x})- s_i, 0\}\bigr)^2}. \notag      
\end{align}
A small IGD$^+$-C value indicates that $\mathcal{X}$ is a good approximation of the ROI in the objective space in terms of the convergence to the PF, the convergence to the reference point $\mathbf{z}$, and the diversity in the ROI.
Before calculating the IGD$^+$-C value, we normalized all objective vectors by using the true ideal and nadir points for each problem.
Then, we calculated the IGD$^+$-C value in the normalized objective space $[0,1]^m$.
We set $r$ to $0.1$ in the normalized objective space.



In addition to the IGD$^+$-C indicator, we used the following three error indicators: the squared ideal point and nadir point estimation errors ($e^{\mathrm{ideal}}$ and $e^{\mathrm{nadir}}$)~\cite{BlankDR19} and the objective range error (ORE)~\cite{HeIS21}.
While the IGD$^+$-C indicator is to evaluate the performance of PBEMO algorithms, the three error indicators are to evaluate the performance of normalization methods.
The following $e^{\mathrm{ideal}}$ and $e^{\mathrm{nadir}}$ indicators measure how well $\vec{z}^{\mathrm{lb}}$ and $\vec{z}^{\mathrm{ub}}$ approximate the ideal point $\vec{z}^{\mathrm{ideal}}$ and nadir point $\vec{z}^{\mathrm{nadir}}$, respectively:
\begin{align}
e^{\mathrm{ideal}}(\vec{z}^{\mathrm{lb}})  &= \sum^m_{i=1} \left(\frac{z^{\mathrm{lb}}_i - z^{\mathrm{ideal}}_i}{z^{\mathrm{nadir}}_i - z^{\mathrm{ideal}}_i} \right)^2,\notag \\
e^{\mathrm{nadir}}(\vec{z}^{\mathrm{ub}})  &= \sum^m_{i=1} \left(\frac{z^{\mathrm{ub}}_i - z^{\mathrm{nadir}}_i}{z^{\mathrm{nadir}}_i - z^{\mathrm{ideal}}_i} \right)^2. \notag
\end{align}
Small $e^{\mathrm{ideal}}$ and $e^{\mathrm{nadir}}$ values indicate that the corresponding $\vec{z}^{\mathrm{lb}}$ and $\vec{z}^{\mathrm{ub}}$ are close to the true ideal and nadir points, respectively.


Unlike $e^{\mathrm{ideal}}$ and $e^{\mathrm{nadir}}$, ORE measures how well $\vec{z}^{\mathrm{lb}}$ and $\vec{z}^{\mathrm{ub}}$ approximate the range of the PF as follows:
\begin{align}
  \mathrm{ORE}(\vec{z}^{\mathrm{lb}}, \vec{z}^{\mathrm{ub}}) = \mathtt{std}\left(\frac{z^{\mathrm{ub}}_1 - z^{\mathrm{lb}}_1}{z^{\mathrm{nadir}}_1 - z^{\mathrm{ideal}}_1}, \ldots, \frac{z^{\mathrm{ub}}_m - z^{\mathrm{lb}}_m}{z^{\mathrm{nadir}}_m - z^{\mathrm{ideal}}_m} \right), \notag
\end{align}
where the $\mathtt{std}$ function returns the standard deviation of inputs.
A small ORE value indicates that normalization of $m$ objectives is well performed by using the corresponding pair of $\vec{z}^{\mathrm{lb}}$ and $\vec{z}^{\mathrm{ub}}$.

\definecolor{c1}{RGB}{150,150,150}
\definecolor{c2}{RGB}{220,220,220}

\section{Results}
\label{sec:results}

This section describes our analysis results.
Through experiments, Sections \ref{sec:est_pbemo} and \ref{sec:comparison_nm} aim to address the two research questions (\textbf{RQ1}--\textbf{RQ2}) described in \pref{sec:introduction}, respectively.

\subsection{On the performance of normalization methods in PBEMO in terms of $e^{\mathrm{ideal}}$, $e^{\mathrm{nadir}}$, and ORE}
\label{sec:est_pbemo}

As in \cite{BlankDR19,HeIS21}, this section investigates the performance of the three normalization methods (PP, BP, and BA) almost independently from the performance of the three PBEMO algorithms (R-NSGA-II, r-NSGA-II, and MOEA/D-NUMS).
Figures \ref{fig:3error_SDTLZ1_m4} and \ref{fig:3error_IDTLZ1_m4} show the results of the three normalization methods in the three PBEMO algorithms on the SDTLZ1 and IDTLZ1 problems with $m = 4$.
Figures \ref{fig:3error_SDTLZ1_m4} and \ref{fig:3error_IDTLZ1_m4} show $e^{\mathrm{ideal}}$, $e^{\mathrm{nadir}}$, and ORE values.
Note that the x-axis in Figures \ref{fig:3error_SDTLZ1_m4} and \ref{fig:3error_IDTLZ1_m4} starts at $200$ function evaluations.
This is simply because R-NSGA-II and r-NSGA-II with $\mu=100$ perform the first normalization operation at $200$ function evaluations.
For the sake of reference, Figures \ref{fig:3error_SDTLZ1_m4} and \ref{fig:3error_IDTLZ1_m4} also show the results of NSGA-II with $\mu=100$, which is a representative conventional EMO algorithm.
Figures \ref{supfig:3error_RNSGA2_DTLZ1}--\ref{supfig:3error_MOEADNUMS_IDTLZ4} show the results on all test problems with  $m \in \{2, 4, 6\}$.


\begin{figure*}[t]
\newcommand{\wvar}{0.3}  
\centering
\includegraphics[width=0.7\textwidth]{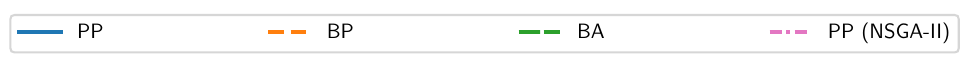}
\vspace{-3.9mm}
   \\
   \subfloat[$e^{\mathrm{ideal}}$ (R-NSGA-II)]{\includegraphics[width=\wvar\textwidth]{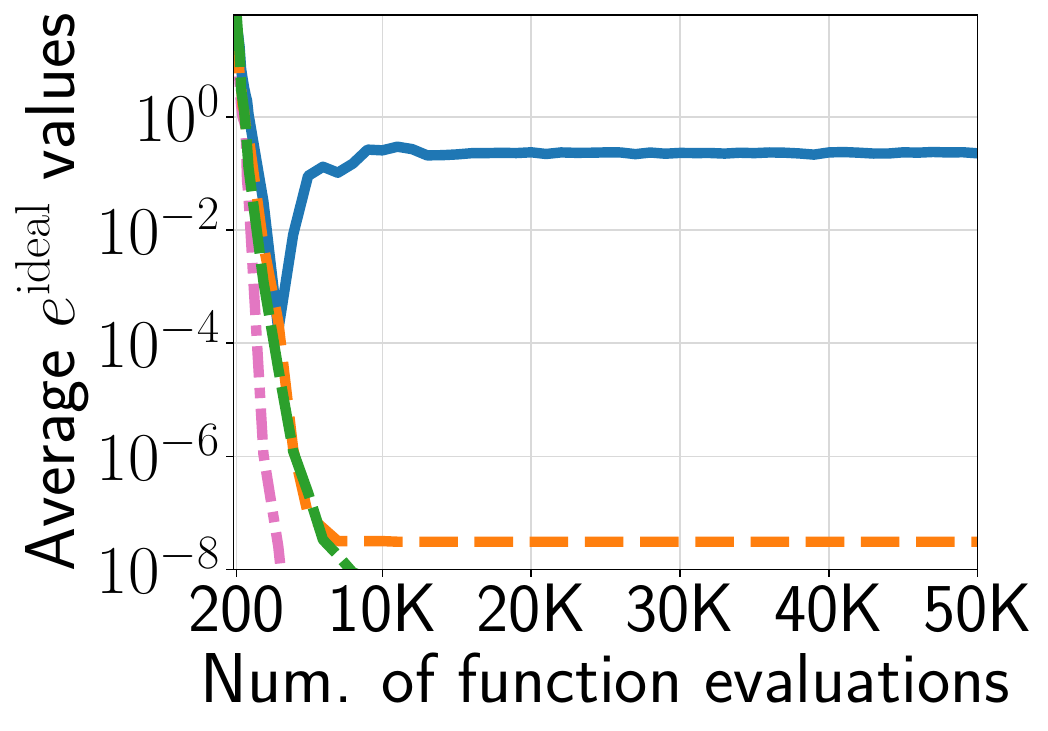}}
   \subfloat[$e^{\mathrm{ideal}}$ (r-NSGA-II)]{\includegraphics[width=\wvar\textwidth]{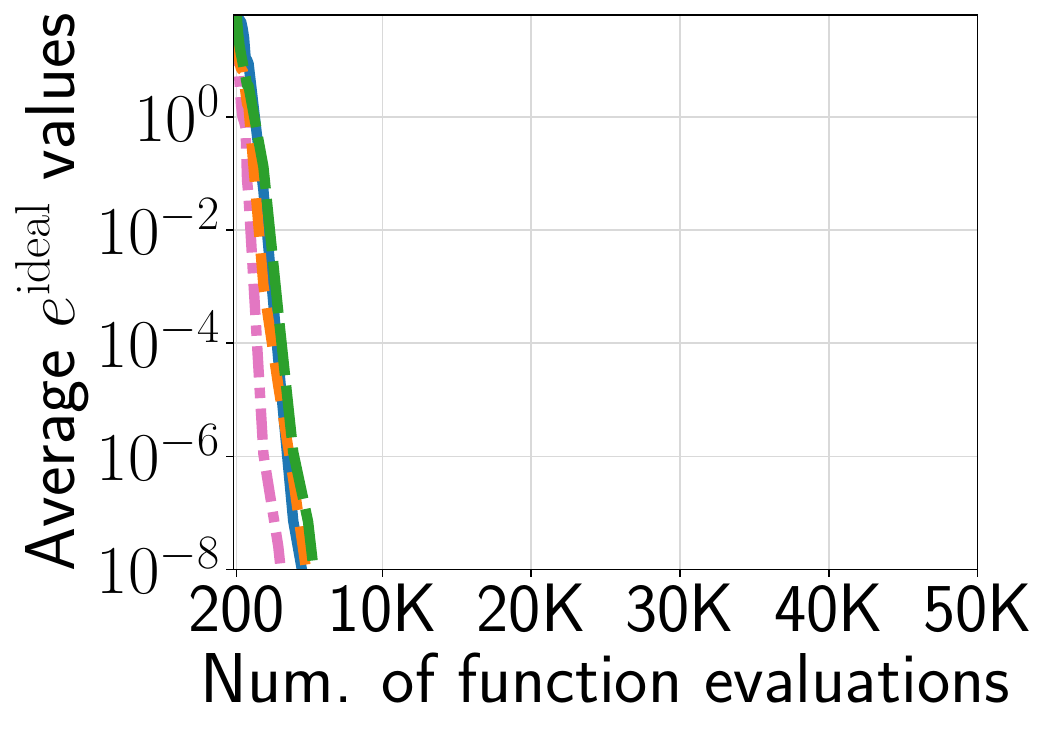}}
      \subfloat[$e^{\mathrm{ideal}}$ (NUMS)]{\includegraphics[width=\wvar\textwidth]{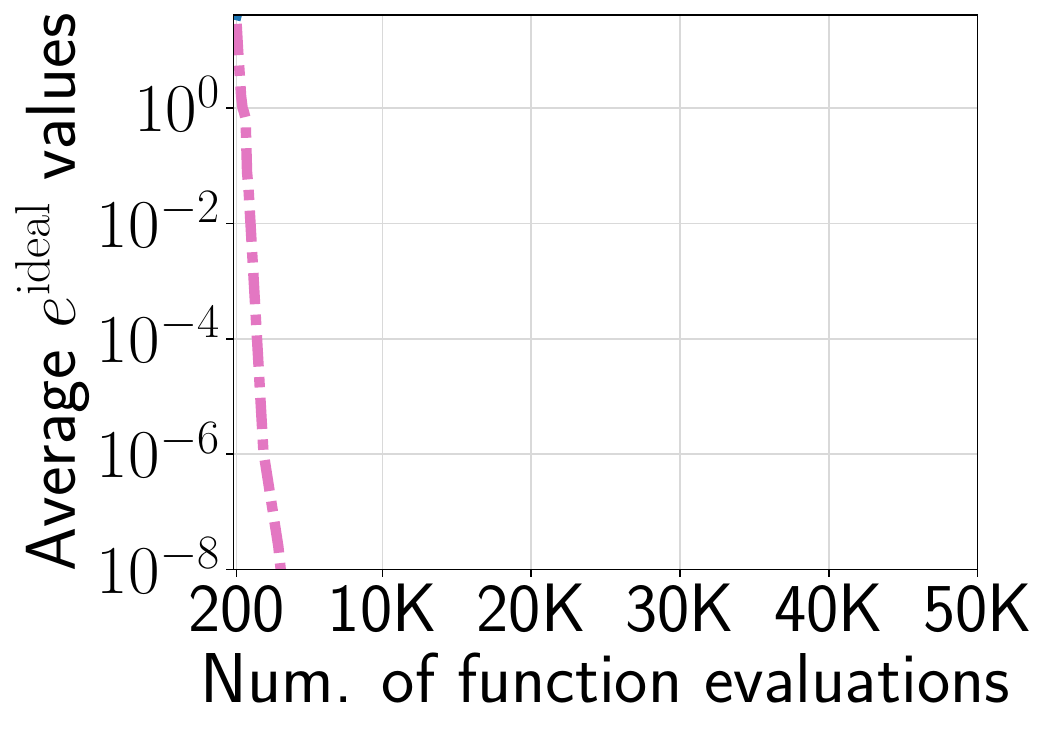}}        
\\
   \subfloat[$e^{\mathrm{nadir}}$ (R-NSGA-II)]{\includegraphics[width=\wvar\textwidth]{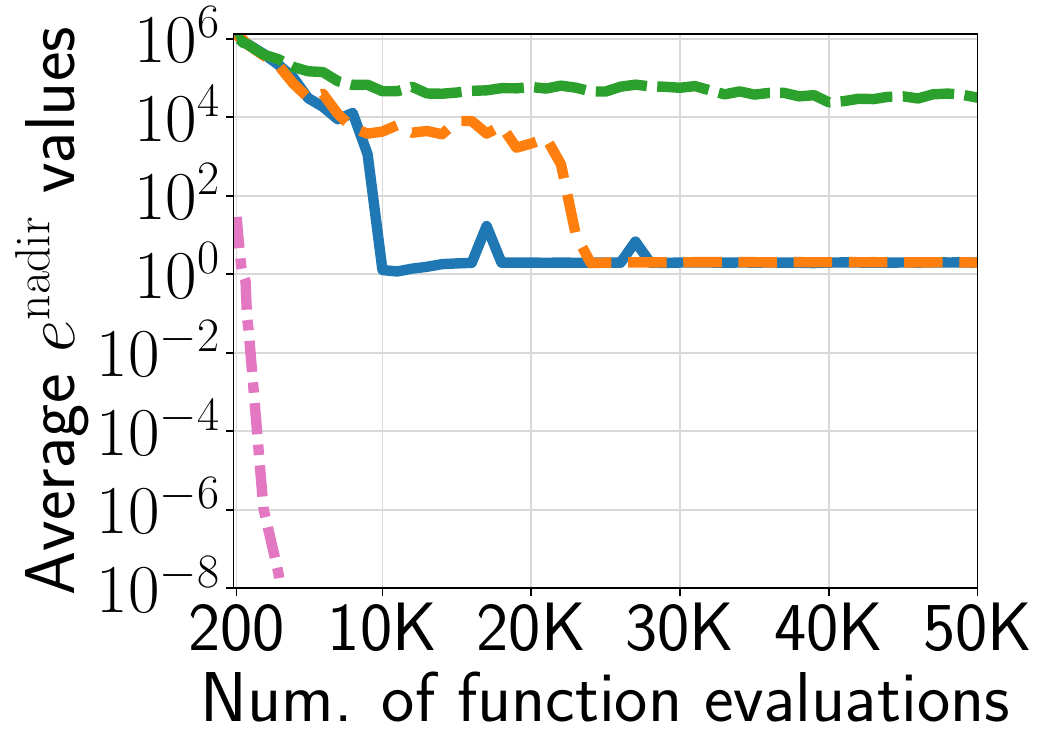}}
   \subfloat[$e^{\mathrm{nadir}}$ (r-NSGA-II)]{\includegraphics[width=\wvar\textwidth]{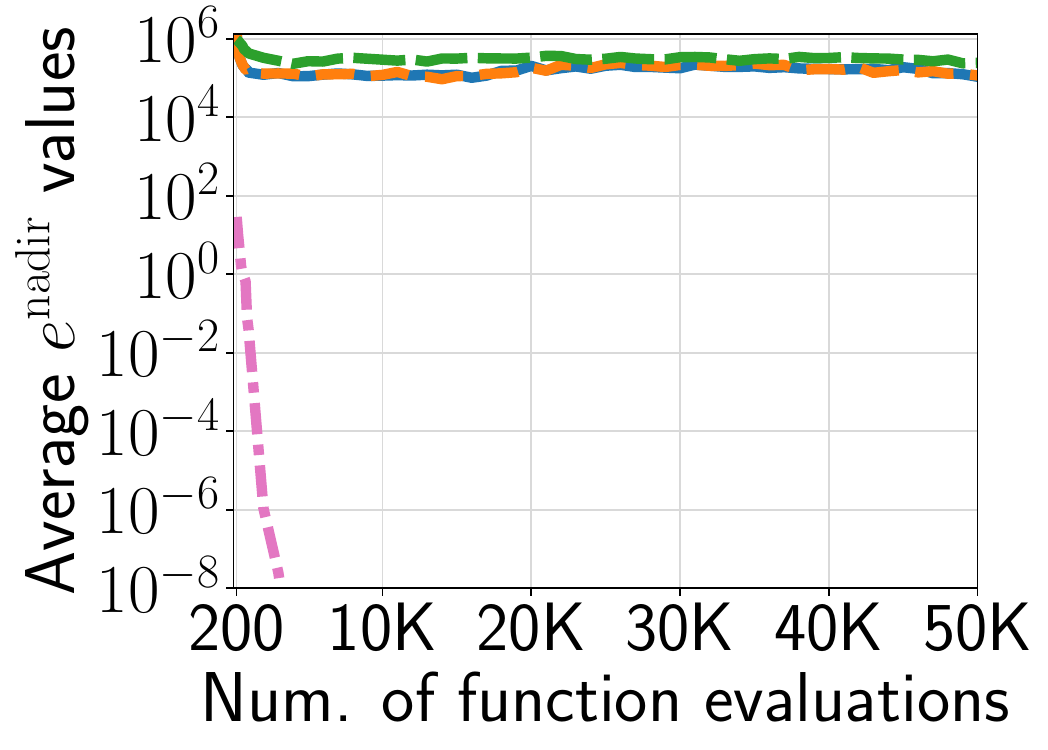}}
      \subfloat[$e^{\mathrm{nadir}}$ (NUMS)]{\includegraphics[width=\wvar\textwidth]{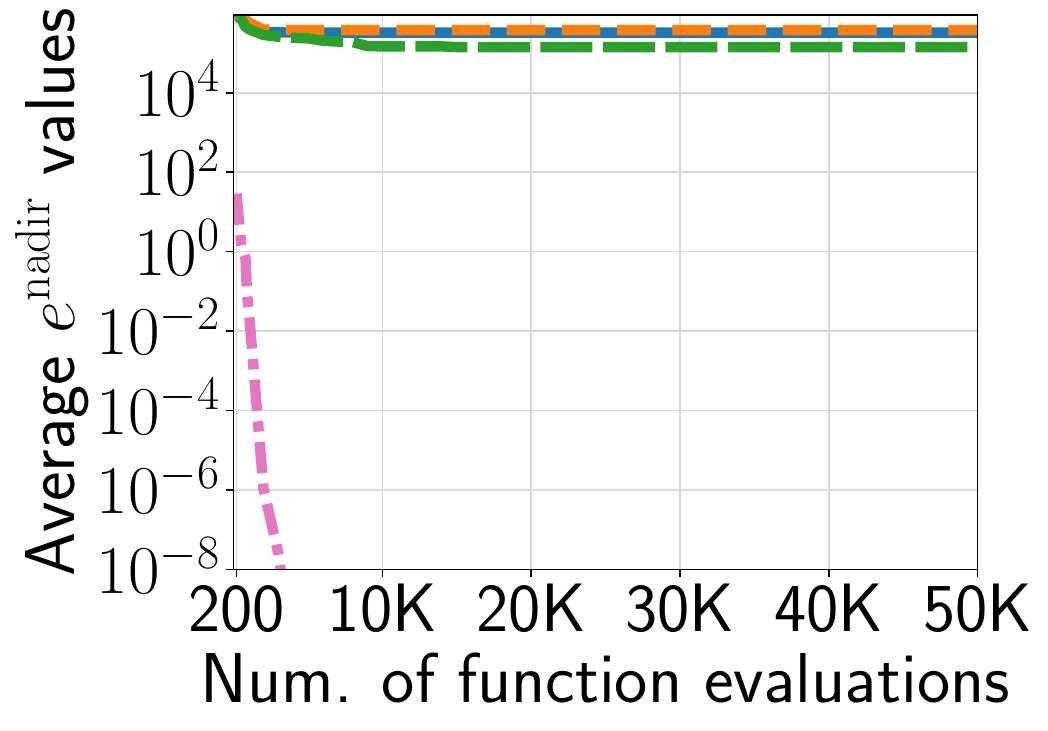}}        
\\
   \subfloat[ORE (R-NSGA-II)]{\includegraphics[width=\wvar\textwidth]{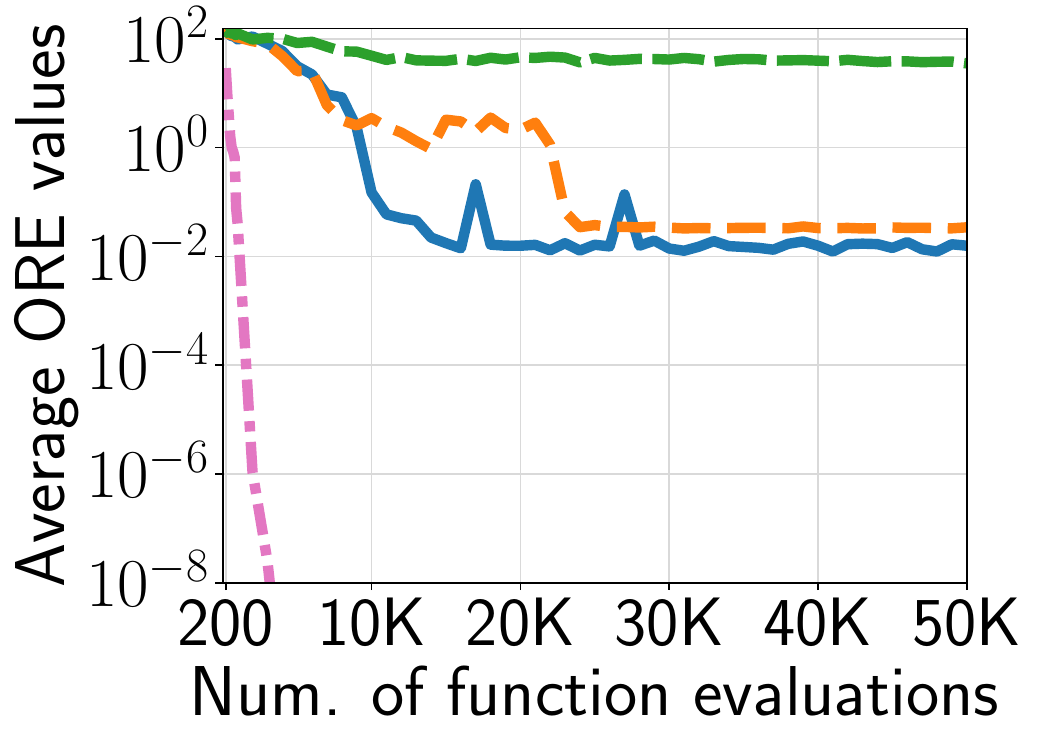}}
   \subfloat[ORE (r-NSGA-II)]{\includegraphics[width=\wvar\textwidth]{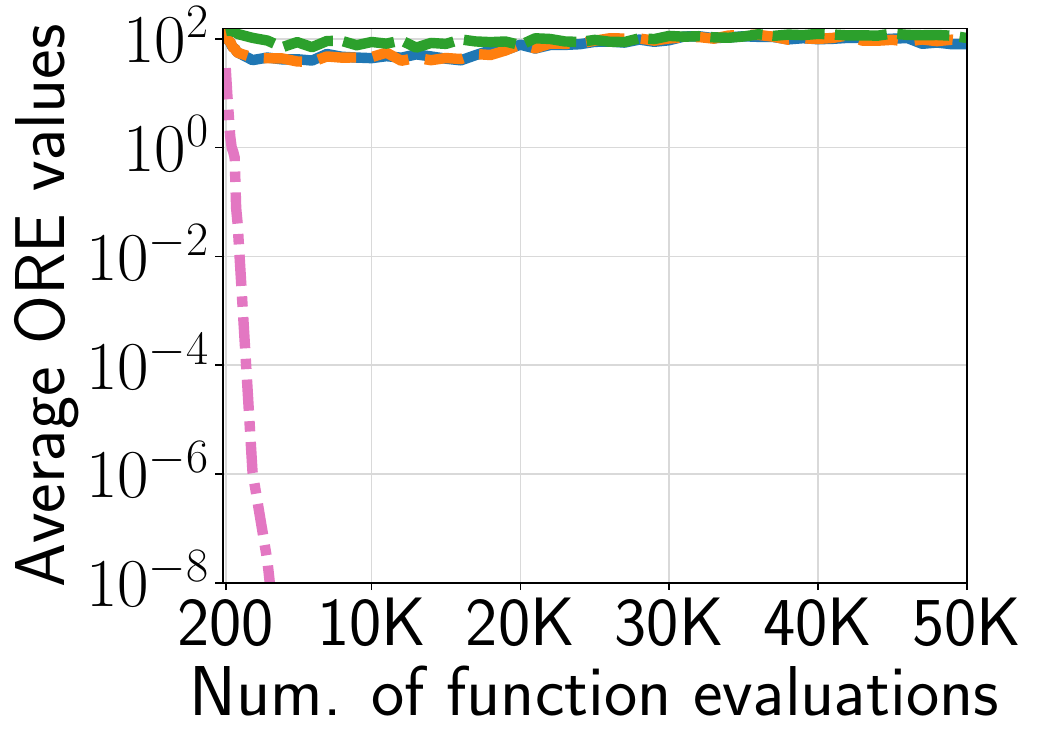}}
      \subfloat[ORE (NUMS)]{\includegraphics[width=\wvar\textwidth]{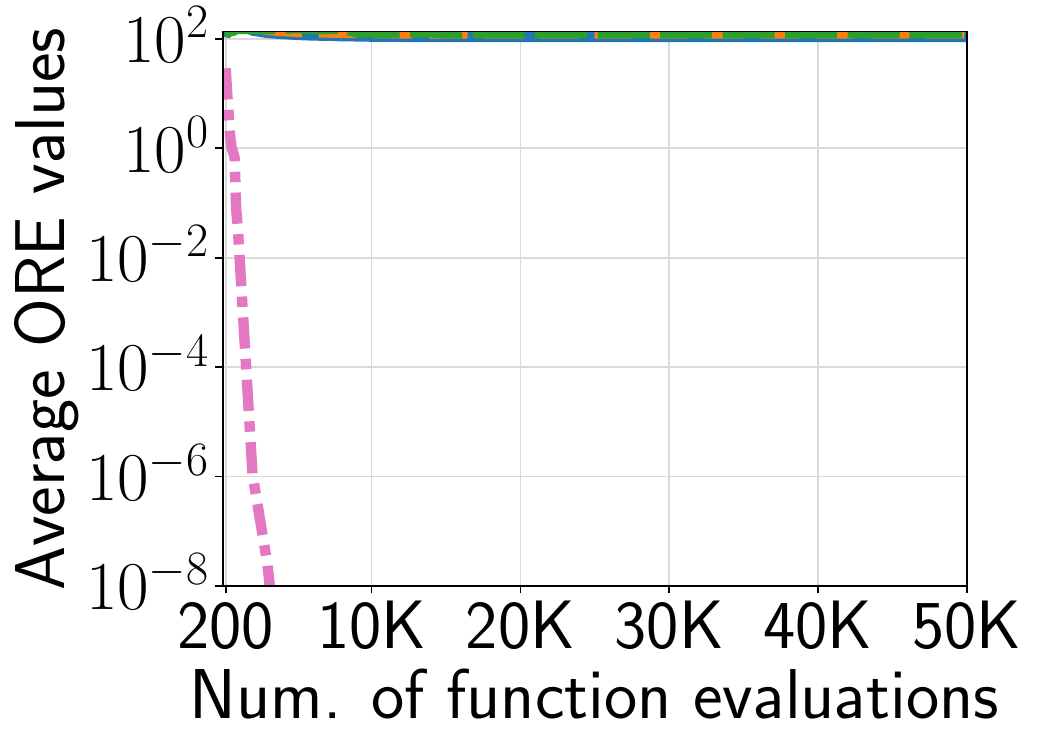}}        
\caption{Average $e^{\mathrm{ideal}}$, $e^{\mathrm{nadir}}$, and ORE values of the three normalization methods in the three PBEMO algorithms on SDTLZ1 with $m=4$.}
   \label{fig:3error_SDTLZ1_m4}
\end{figure*}

\begin{figure*}[t]
\newcommand{\wvar}{0.3}         
\centering
\includegraphics[width=0.7\textwidth]{./figs/legend/legend_3.pdf}
\vspace{-3.9mm}
   \\
   \subfloat[$e^{\mathrm{ideal}}$ (R-NSGA-II)]{\includegraphics[width=\wvar\textwidth]{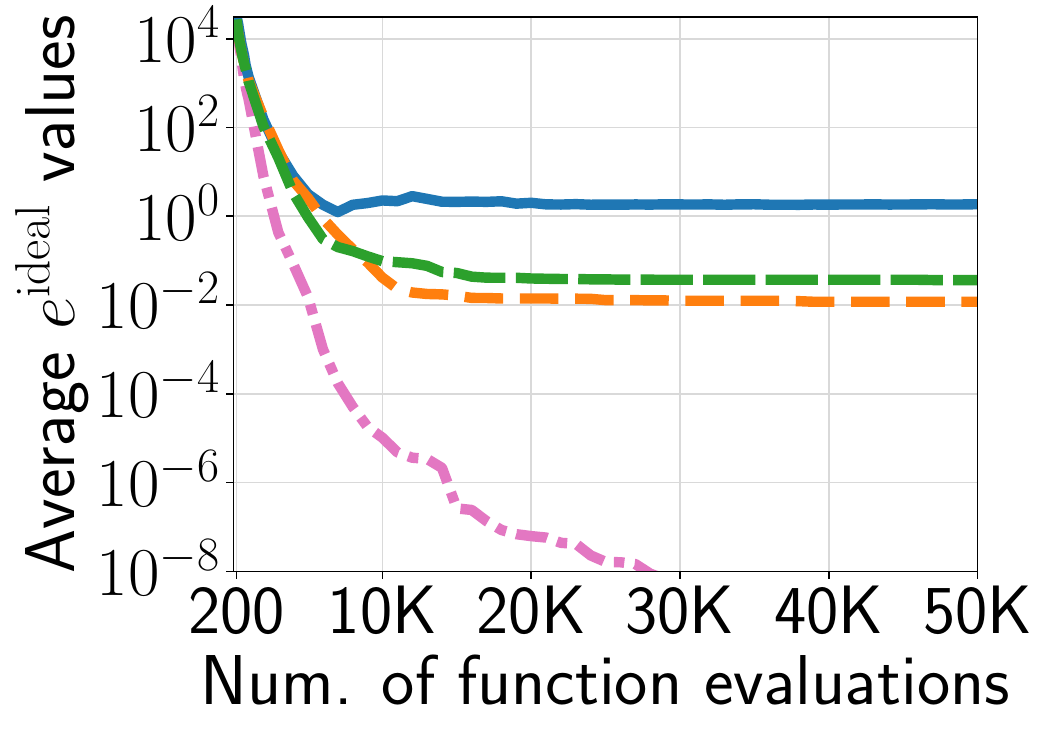}}
   \subfloat[$e^{\mathrm{ideal}}$ (r-NSGA-II)]{\includegraphics[width=\wvar\textwidth]{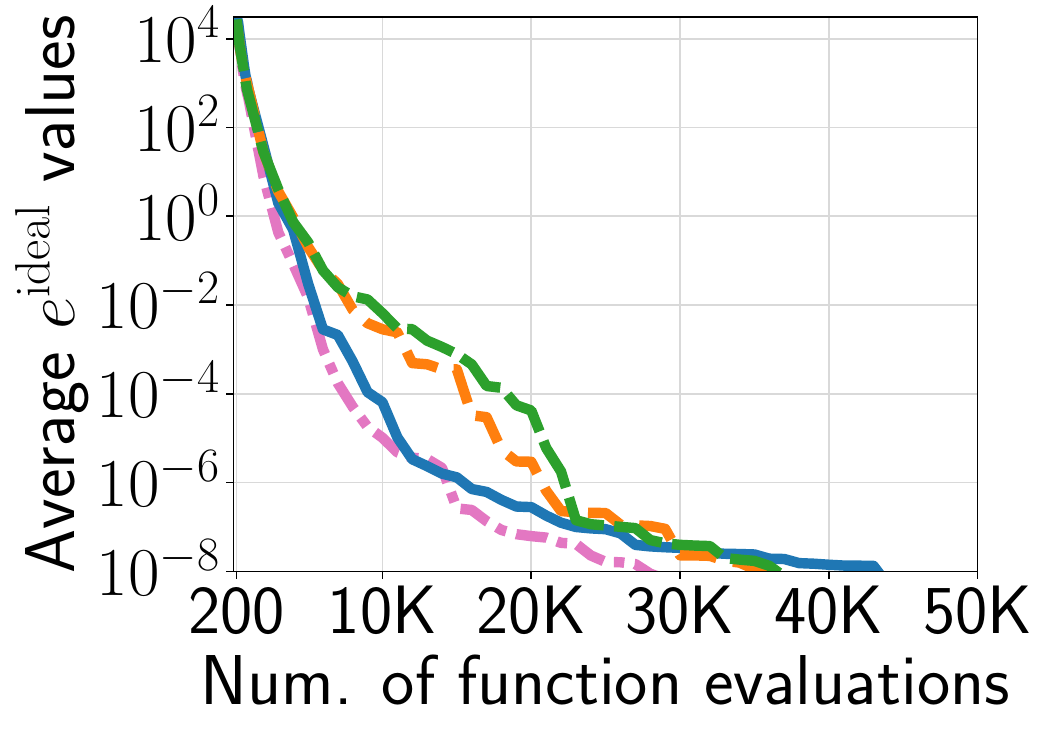}}
      \subfloat[$e^{\mathrm{ideal}}$ (NUMS)]{\includegraphics[width=\wvar\textwidth]{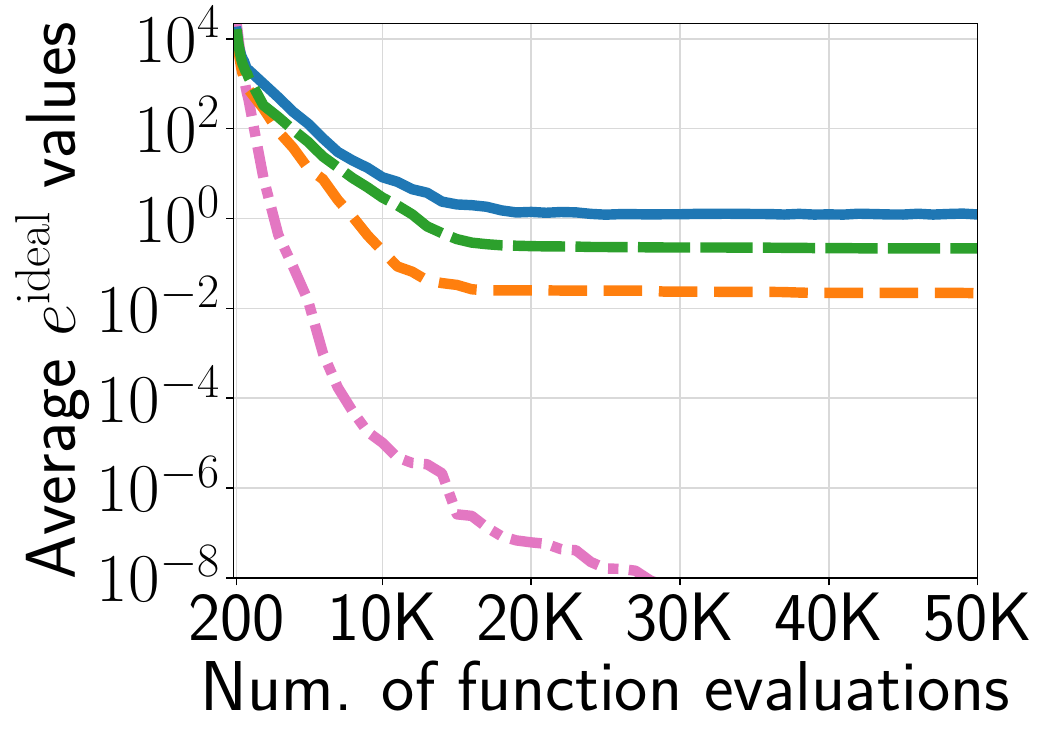}}        
\\
   \subfloat[$e^{\mathrm{nadir}}$ (R-NSGA-II)]{\includegraphics[width=\wvar\textwidth]{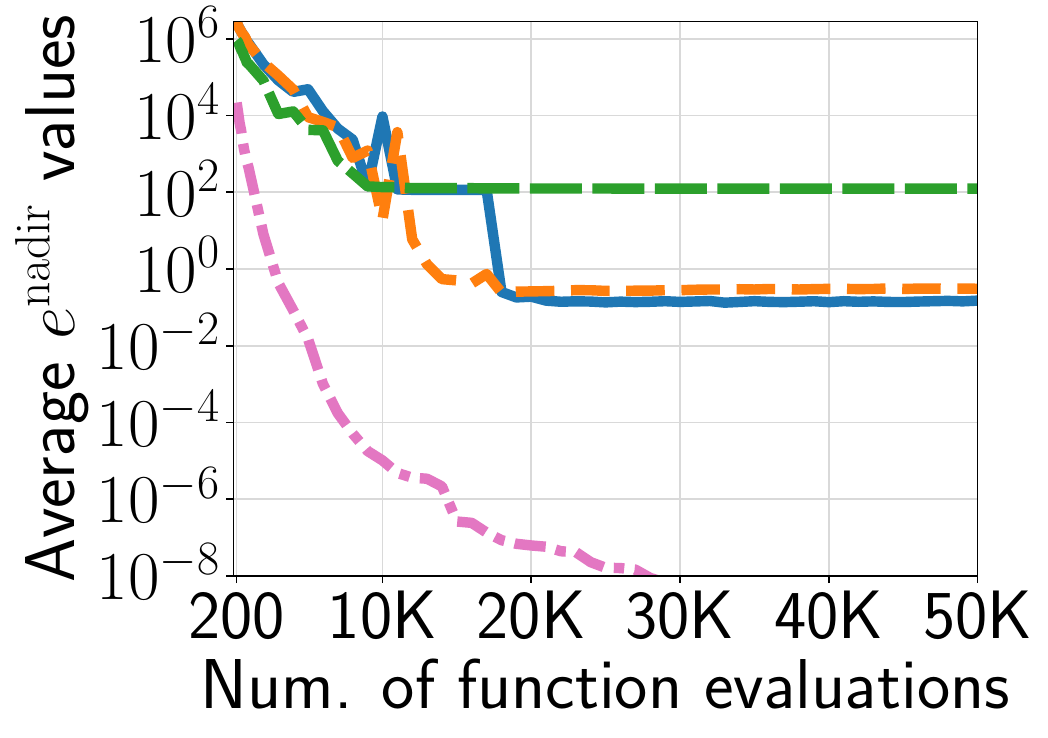}}
   \subfloat[$e^{\mathrm{nadir}}$ (r-NSGA-II)]{\includegraphics[width=\wvar\textwidth]{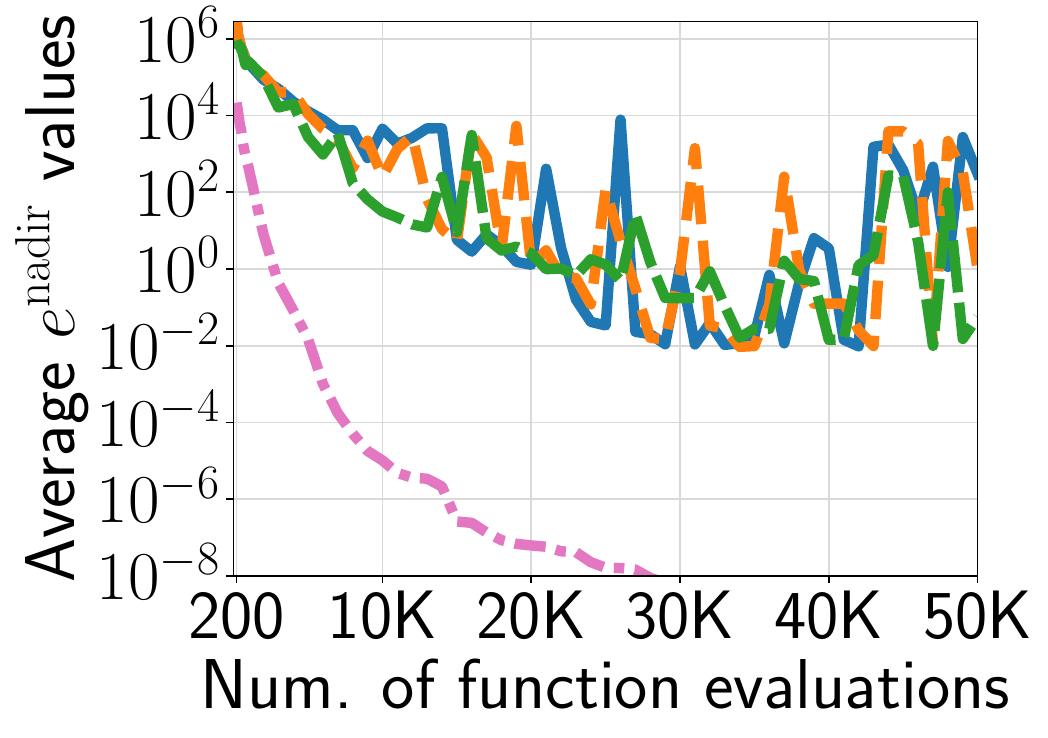}}
      \subfloat[$e^{\mathrm{nadir}}$ (NUMS)]{\includegraphics[width=\wvar\textwidth]{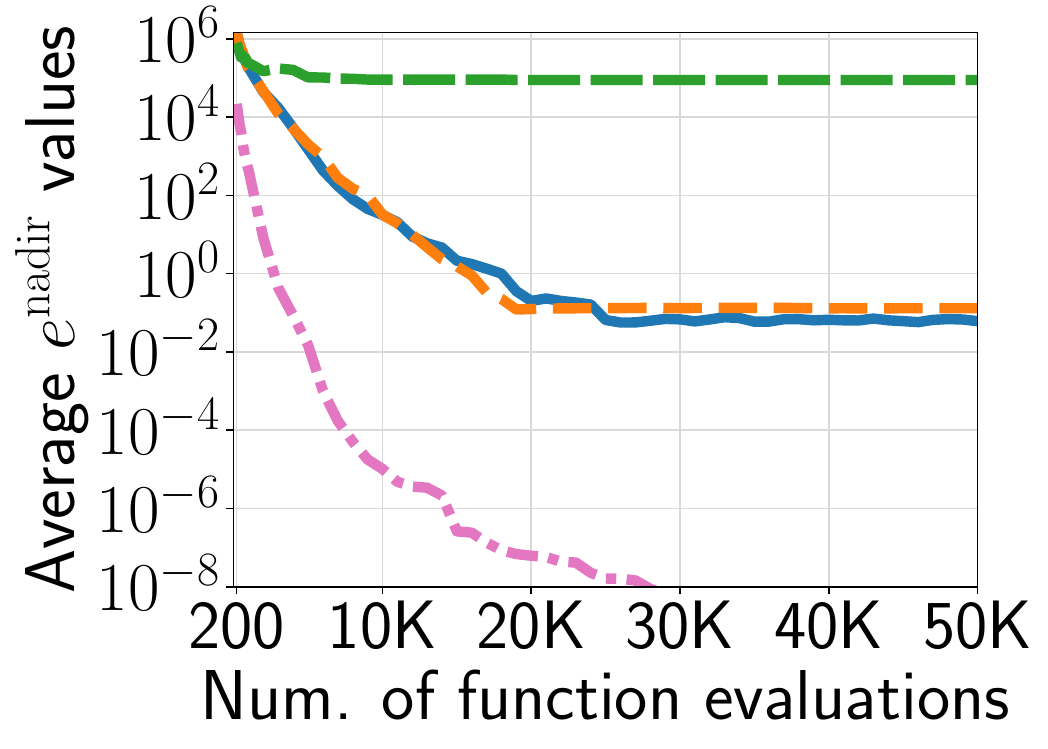}}        
\\
   \subfloat[ORE (R-NSGA-II)]{\includegraphics[width=\wvar\textwidth]{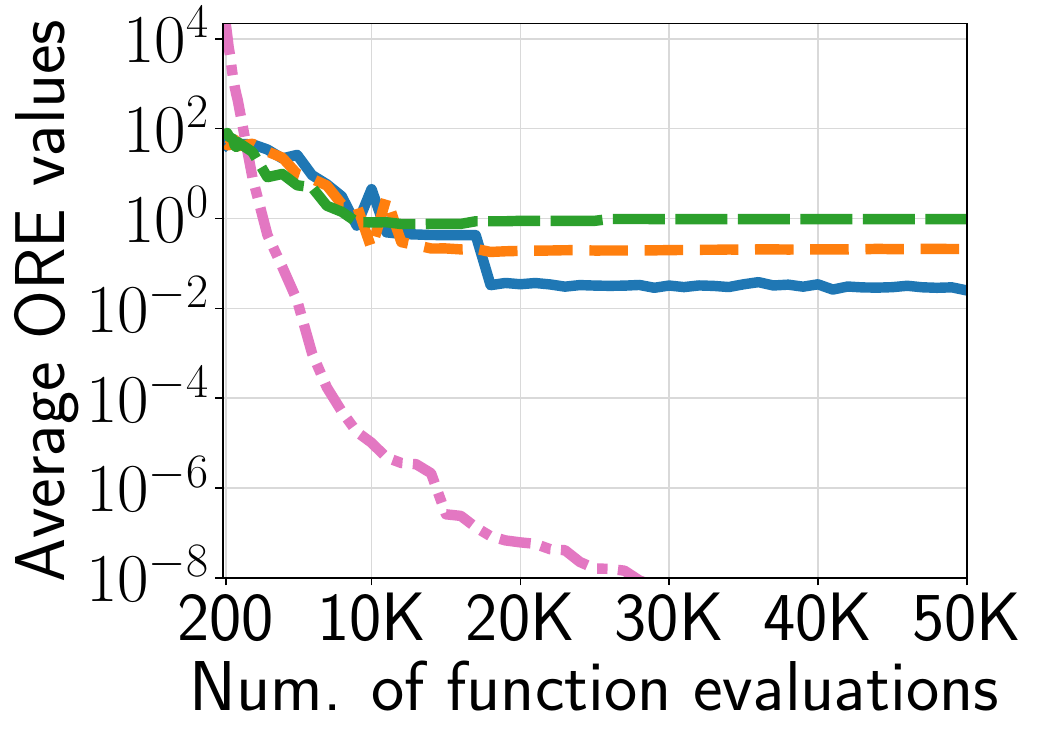}}
   \subfloat[ORE (r-NSGA-II)]{\includegraphics[width=\wvar\textwidth]{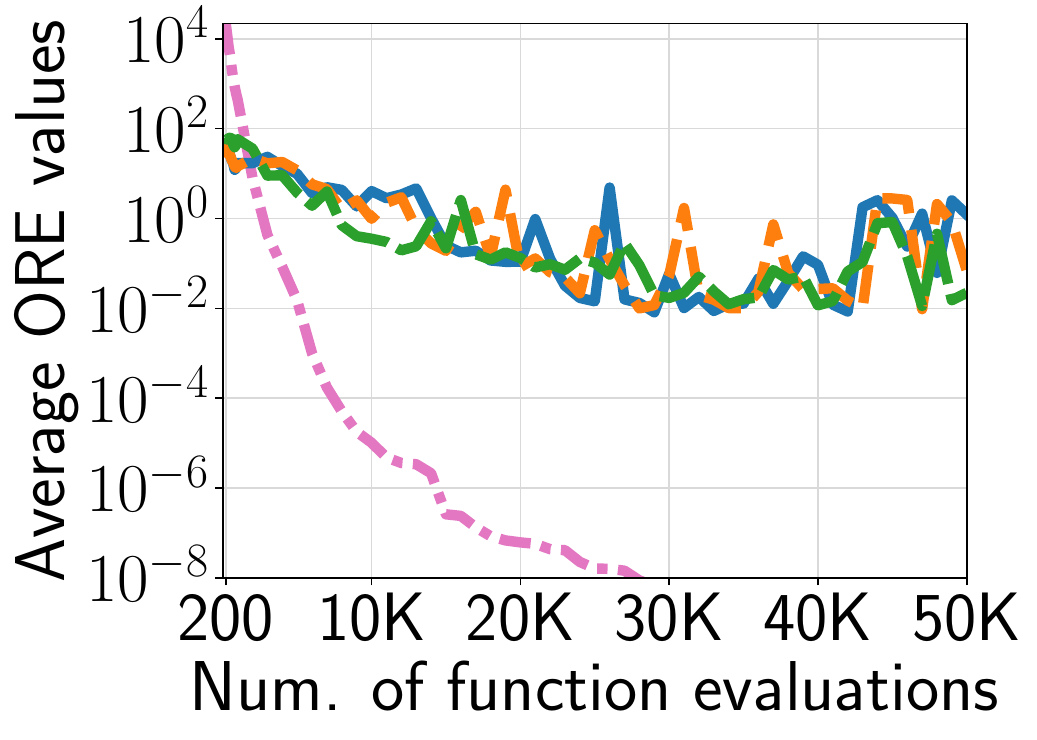}}
      \subfloat[ORE (NUMS)]{\includegraphics[width=\wvar\textwidth]{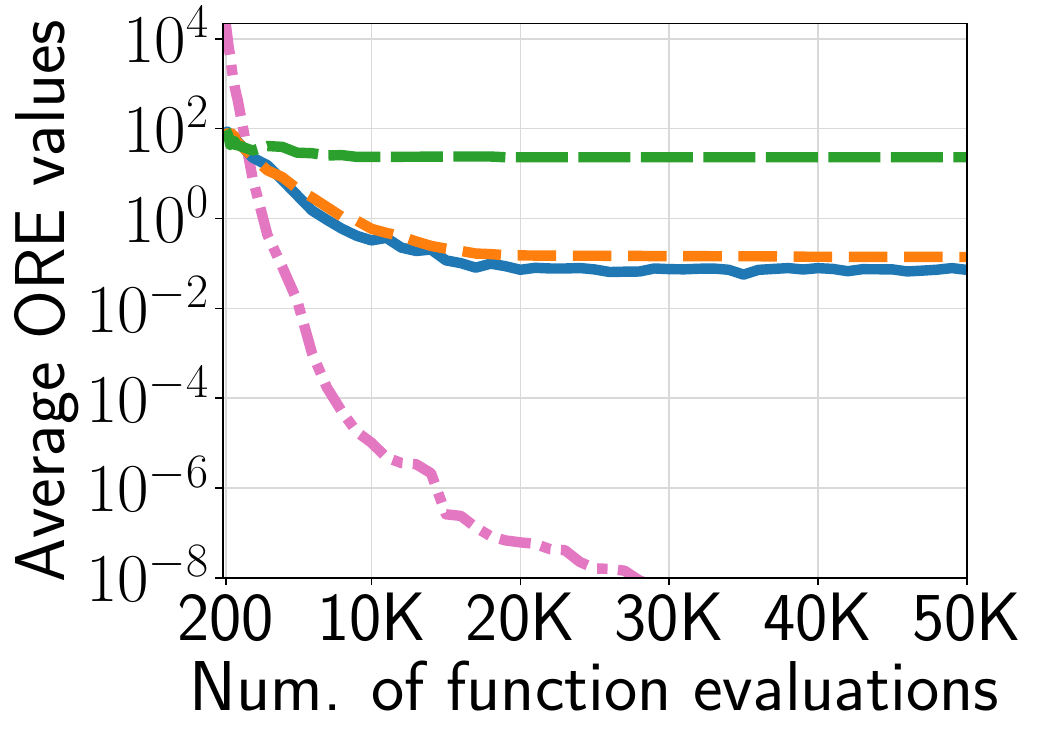}}        
\caption{Average $e^{\mathrm{ideal}}$, $e^{\mathrm{nadir}}$, and ORE values of the three normalization methods in the three PBEMO algorithms on IDTLZ1 with $m=4$.}
   \label{fig:3error_IDTLZ1_m4}   
\end{figure*}

\subsubsection{Comparison with the general normalization method in NSGA-II}

As seen from Figures \ref{fig:3error_SDTLZ1_m4} and \ref{fig:3error_IDTLZ1_m4}, the normalization method in NSGA-II performs the best in terms of $e^{\mathrm{ideal}}$, $e^{\mathrm{nadir}}$, and ORE.\footnote{A clear exception is the results of MOEA/D-NUMS. 
As seen from \pref{fig:3error_SDTLZ1_m4}(c), PP, BP, and BA achieve $e^{\mathrm{ideal}} = 0$ at 200 function evaluations.
However, this amazing performance is simply due to the synergy of the properties of MOEA/D-NUMS and some DTLZ problems.
Interested readers can find a further discussion in Section \ref{supsec:nums_dtlz}.
}
Similar results can be found in most of Figures \ref{supfig:3error_RNSGA2_DTLZ1}--\ref{supfig:3error_MOEADNUMS_IDTLZ4}.
This observation suggests that normalization methods in PBEMO algorithms perform significantly worse than those in conventional EMO algorithms in terms of approximating the ideal and nadir points. 

The poor performance of the normalization methods in PBEMO is mainly because PBEMO algorithms do not approximate the whole PF.
\pref{fig:points_nsga2_rnsga2} shows distributions of objective vectors of solutions in the final population in NSGA-II and R-NSGA-II with PP on the SDTLZ1 problem with $m=2$.
\pref{fig:points_nsga2_rnsga2} shows the results of a single run with the median IGD$^+$-C value among 31 runs.
As shown in \pref{fig:points_nsga2_rnsga2}, the population in NSGA-II covers the whole PF while that in R-NSGA-II converges to the ROI.
Since R-NSGA-II does not intensively exploit the $m$ extreme regions of the PF, R-NSGA-II is likely to fail to approximate the ideal and nadir points.

\begin{figure*}[t]
\newcommand{\wvara}{0.29}         
\centering
   \subfloat[R-NSGA-II]{\includegraphics[width=\wvara\textwidth]{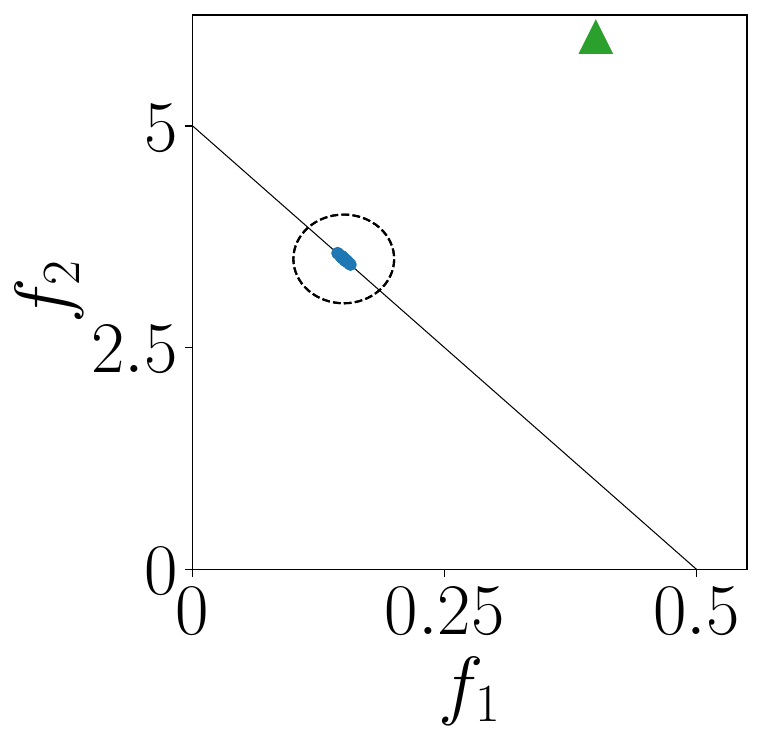}}
   \subfloat[NSGA-II]{\includegraphics[width=\wvara\textwidth]{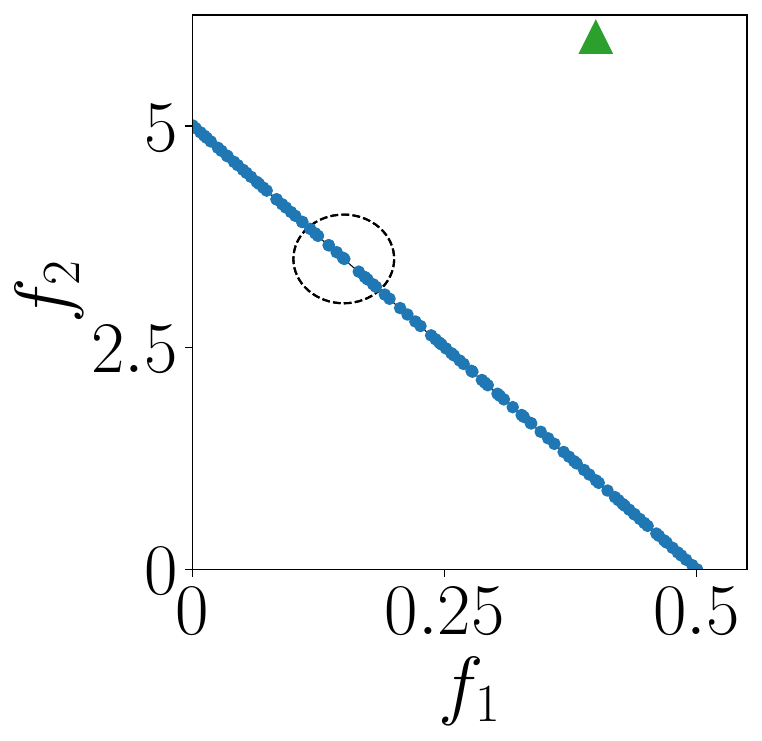}}
\caption{Distributions of the objective vectors of the solutions found by R-NSGA-II (with PP) and NSGA-II on the SDTLZ1 problem with $m=2$, where \tabgreen{$\blacktriangle$} is the reference point $\mathbf{z}$. The dotted circle represents the true ROI.}
   \label{fig:points_nsga2_rnsga2}   
\newcommand{\wvar}{0.29}         
\centering
   \subfloat[$10\,000$ FEs]{\includegraphics[width=\wvar\textwidth]{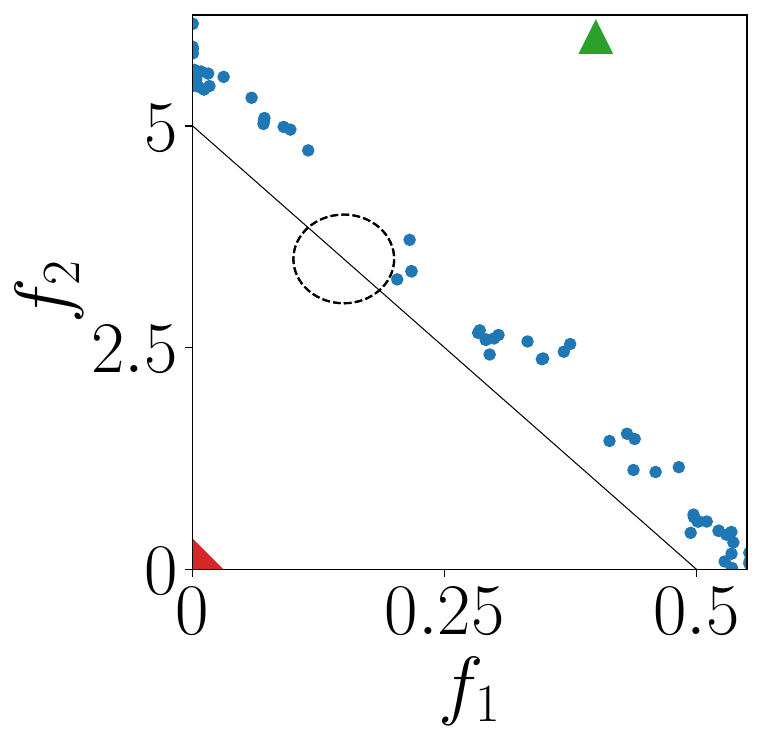}}
      \subfloat[$13\,000$ FEs]{\includegraphics[width=\wvar\textwidth]{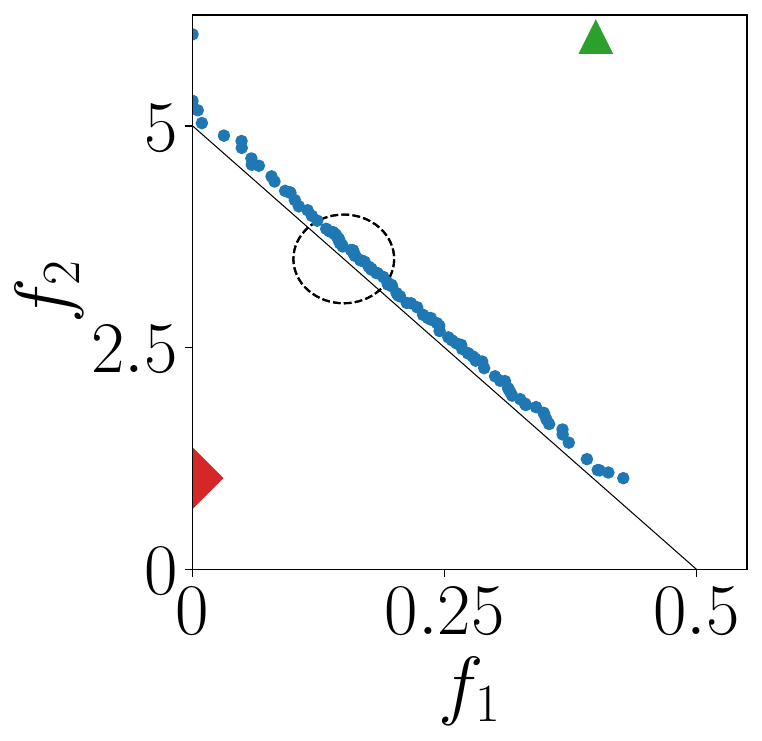}}
\subfloat[$15\,000$ FEs]{\includegraphics[width=\wvar\textwidth]{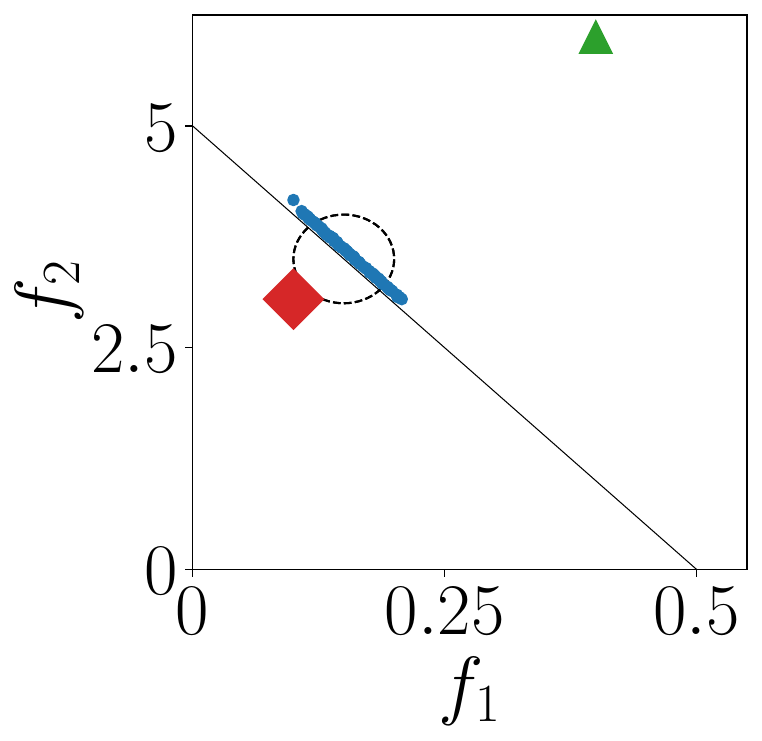}}
\caption{Distributions of the objective vectors of the solutions found by R-NSGA-II with PP on the SDTLZ1 problem with $m=2$, where \tabgreen{$\blacktriangle$} is the reference point $\mathbf{z}$, and \tabred{$\blacklozenge$} is an approximated ideal point by PP. The dotted circle represents the true ROI. The results at $1\,000$, $5\,000$, and $10\,000$ function evaluations (FEs) are shown.}
   \label{fig:points_rnsga2_progress}   
\end{figure*}

\begin{figure*}[t]
\centering
      \subfloat[SDTLZ1]{\includegraphics[width=0.29\textwidth]{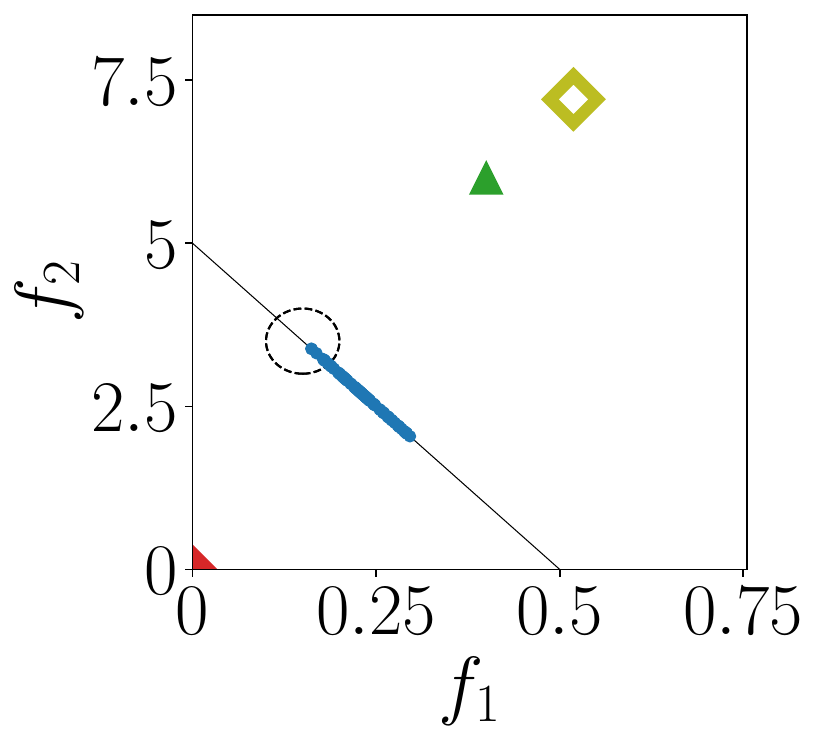}}
      \subfloat[SDTLZ2]{\includegraphics[width=0.28\textwidth]{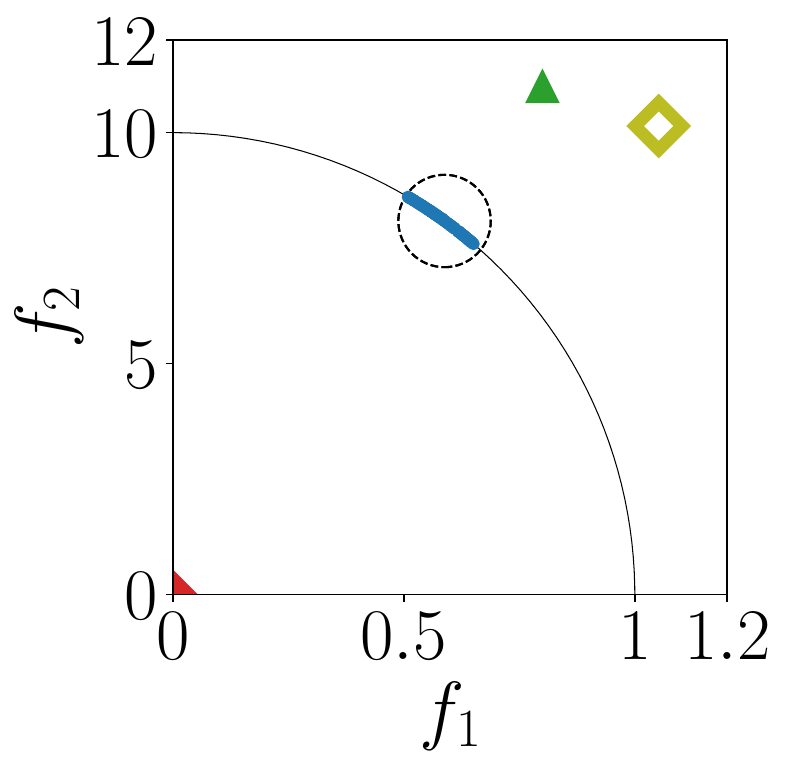}}      
\caption{Approximated ideal (\tabred{$\blacklozenge$}) and nadir (\tabolive{$\Diamond$}) points in R-NSGA-II with BA on the SDTLZ1 problem with $m=2$, where \tabgreen{$\blacktriangle$} is the reference point $\mathbf{z}$. The dotted circle represents the true ROI.}
   \label{fig:points_rnsga2_ba}   
\end{figure*}

\subsubsection{Comparison among the three normalization method in PBEMO}

As shown in Figures \ref{fig:3error_SDTLZ1_m4}(a)--(c) and \ref{fig:3error_IDTLZ1_m4}(a)--(c), BP and BA achieve good $e^{\mathrm{ideal}}$ values.
The $e^{\mathrm{ideal}}$ value decreases as the search progresses.
This is because BP and BA maintain the minimum value for each objective function found so far.

However, as seen from Figures \ref{fig:3error_SDTLZ1_m4}(a) and \ref{fig:3error_IDTLZ1_m4}(a), for R-NSGA-II, the $e^{\mathrm{ideal}}$ value of PP increases at the early stage.
This means that R-NSGA-II with PP moves away from the ideal point as the search progresses.

This is because the population in R-NSGA-II converges to the ROI as the search progresses.
\pref{fig:points_rnsga2_progress} shows the population in R-NSGA-II in the objective space on the SDTLZ1 problem with $m=2$.
\pref{fig:points_rnsga2_progress} shows the results at $10\,000$, $13\,000$, and $15\,000$ function evaluations.
As shown in \pref{fig:points_rnsga2_progress}(a), individuals in the population are widely distributed in the objective space at $10\,000$ function evaluations.
The approximated ideal point $\vec{z}^{\mathrm{lb}}$ is also the same as the true ideal point $\vec{z}^{\mathrm{ideal}} = (0, 0)^{\top}$, where $\vec{z}^{\mathrm{lb}}$ (\tabred{$\blacklozenge$}) is on the bottom left of the figure.
However, the population converges to the ROI as the search progresses.
As a result, as shown in Figures \ref{fig:points_rnsga2_progress}(b) and (c), the approximated ideal point moves away from the ideal point.\footnote{We observed that the population in r-NSGA-II covers the whole PF and does not converge to the ROI only on the (S, I)DTLZ1 problem.
For this reason, the deterioration of $e^{\mathrm{ideal}}$ is not observed in the results of r-NSGA-II.
Determining the cause of this unexpected behavior of r-NSGA-II is beyond the scope of this paper.
}


In contrast to the results for $e^{\mathrm{ideal}}$, BA performs the worst in terms of $e^{\mathrm{nadir}}$ in many cases.
\pref{fig:points_rnsga2_ba} shows approximated ideal and nadir points in R-NSGA-II with BA on the SDTLZ1 and SDTLZ2 problems.
The results at $50\,000$ function evaluations are shown.
As seen from \pref{fig:points_rnsga2_ba}, the approximated ideal point is almost the same as the true ideal point.
In contrast, the approximated nadir point is far from the true nadir point.
Since most PBEMO algorithms (including R-NSGA-II) do not focus on the $m$ extreme regions of the PF, it is difficult for new solutions to dominate the $m$ extreme solutions in the bounded archive $\set{B}$.
As a result, the approximated nadir point is unchanged even at the late of the search.

Although PP performs the worst in terms of $e^{\mathrm{ideal}}$ as discussed above, PP performs the best in terms of ORE in many cases.
Thus, we can say that PP generally fails to locate the ideal point but successfully approximate the range of the objective values.
Similar results are observed for the three PBEMO algorithms on other test problems in Figures \ref{supfig:3error_RNSGA2_DTLZ1}--\ref{supfig:3error_MOEADNUMS_IDTLZ4}.

\begin{tcolorbox}[title=Answers to RQ1, sharpish corners, top=2pt, bottom=2pt, left=4pt, right=4pt, boxrule=0.5pt]
Our results show that the estimation performance of a normalization method in PBEMO is significantly worse than that in conventional EMO in most cases. 
Since PBEMO does not approximate the whole PF, its normalization methods fail to locate the ideal and nadir points in most cases.
We observed that PP cannot approximate the ideal point well, but PP approximates the range of the PF better than BP and BA.
In contrast, our results show that BP and BA can approximate the ideal point well, but they perform worse than PP in many cases in terms of approximating the range of the PF for normalization.

\end{tcolorbox}

\begin{figure*}[t]
  \centering
\newcommand{\mywidth}{0.3}
\includegraphics[width=0.8\textwidth]{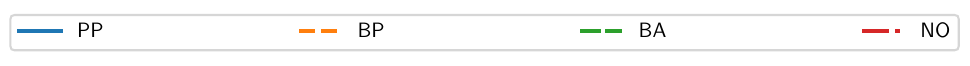}
\vspace{-3.9mm}
  \\
   \subfloat[DTLZ ($m=2$)]{\includegraphics[width=\mywidth\textwidth]{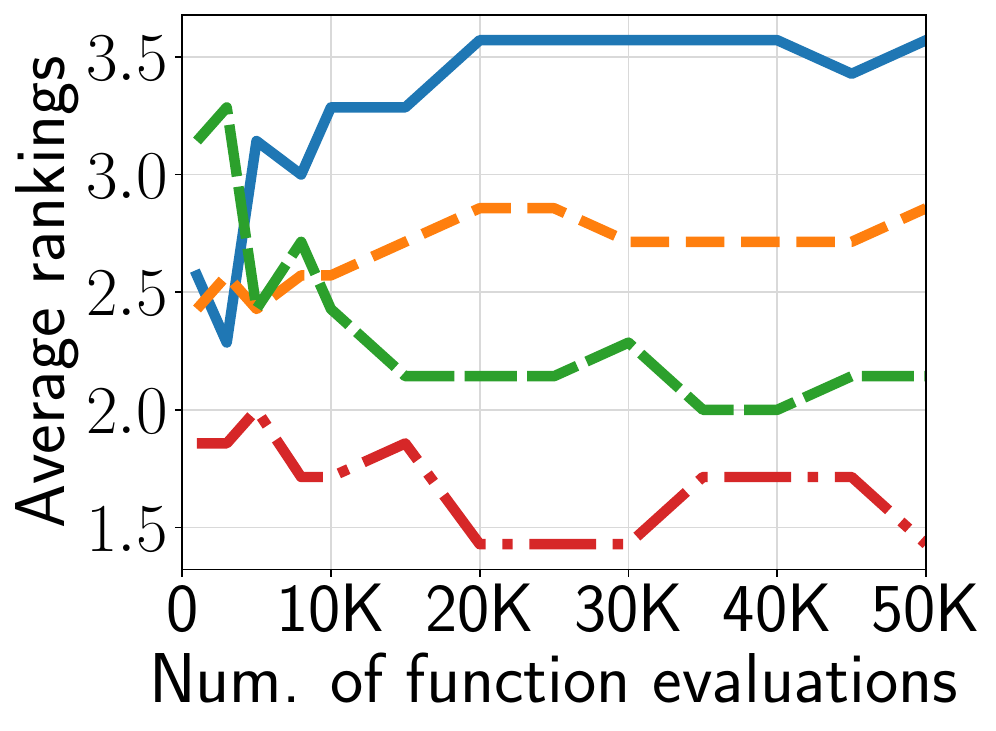}}
   \subfloat[DTLZ ($m=4$)]{\includegraphics[width=\mywidth\textwidth]{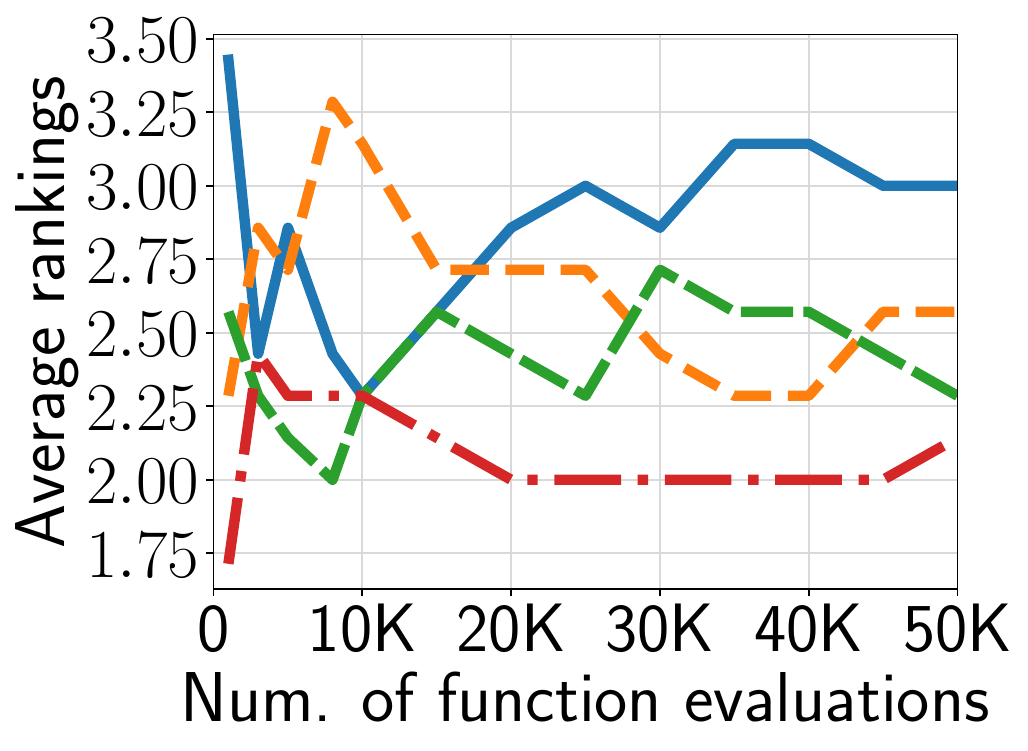}}
   \subfloat[DTLZ ($m=6$)]{\includegraphics[width=\mywidth\textwidth]{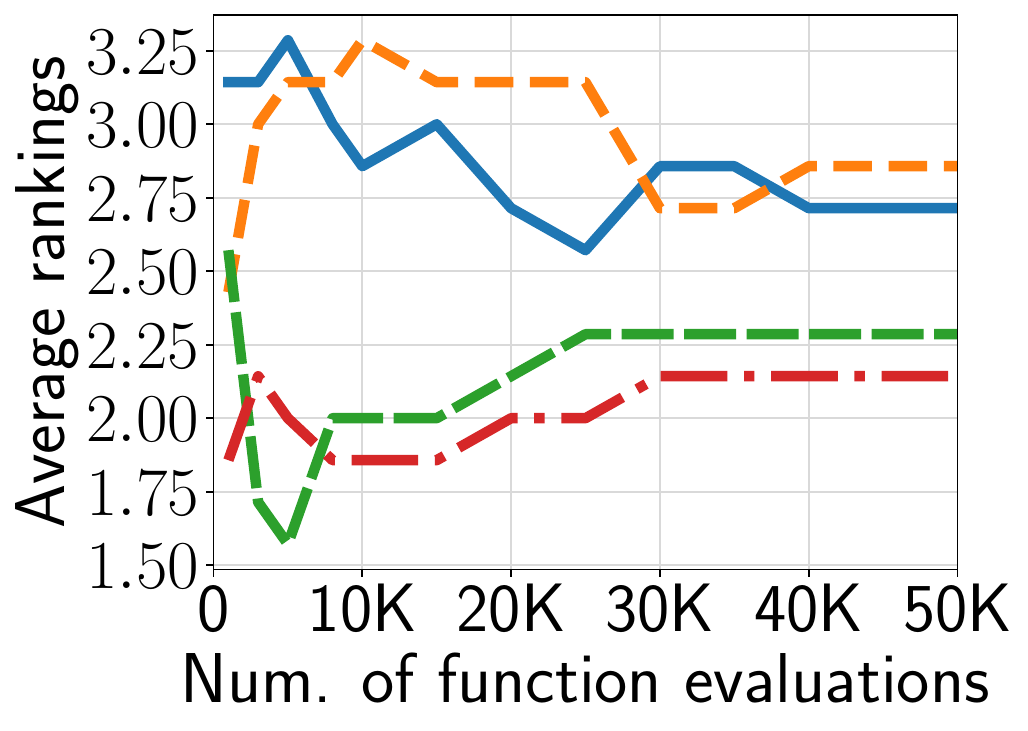}}
\\
   \subfloat[SDTLZ ($m=2$)]{\includegraphics[width=\mywidth\textwidth]{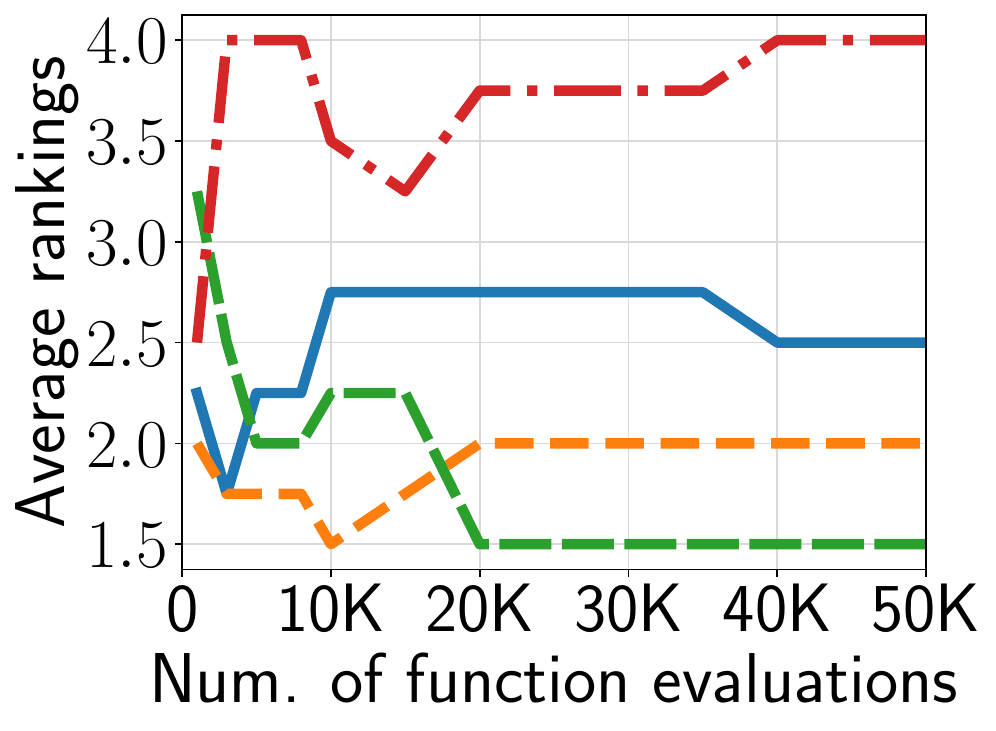}}
   \subfloat[SDTLZ ($m=4$)]{\includegraphics[width=\mywidth\textwidth]{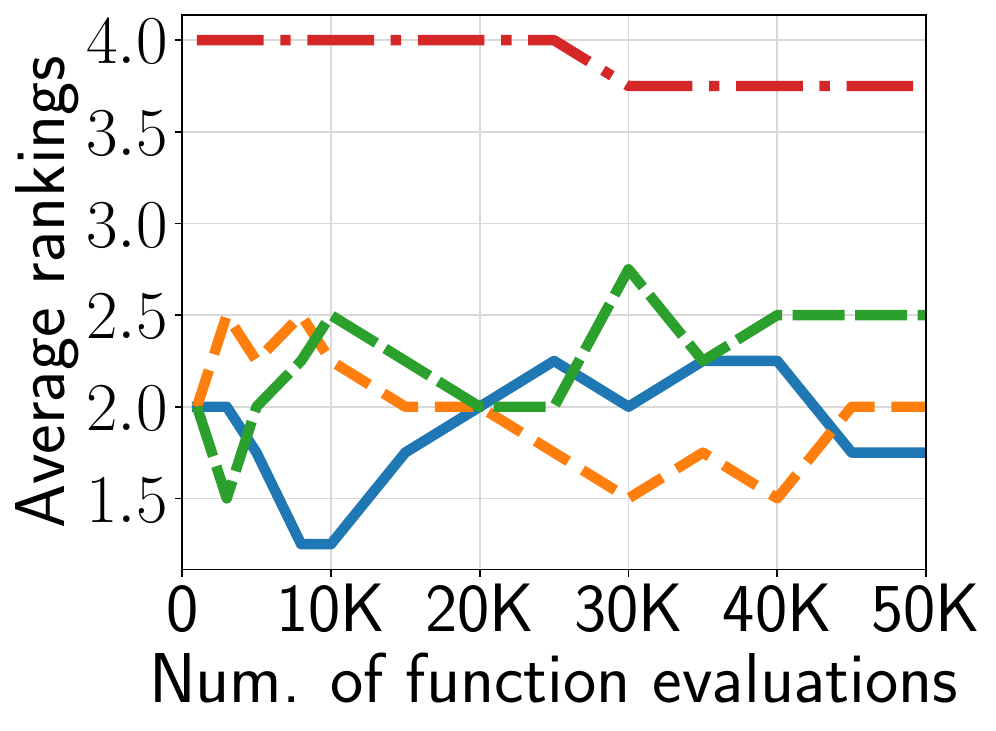}}
   \subfloat[SDTLZ ($m=6$)]{\includegraphics[width=\mywidth\textwidth]{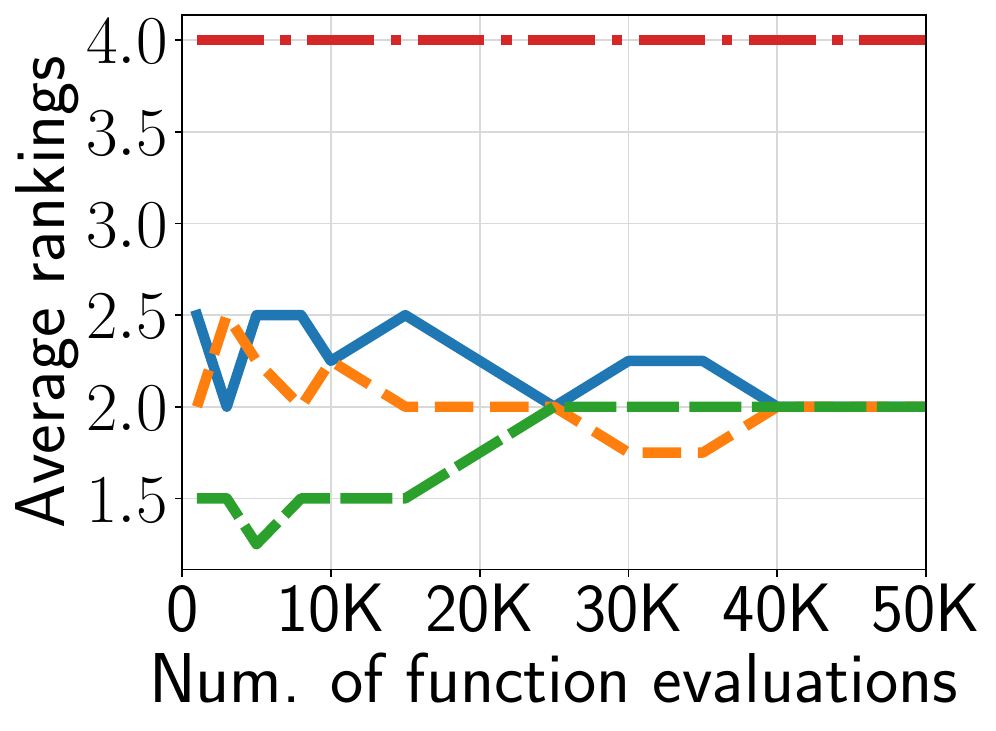}}
\\
   \subfloat[IDTLZ ($m=2$)]{\includegraphics[width=\mywidth\textwidth]{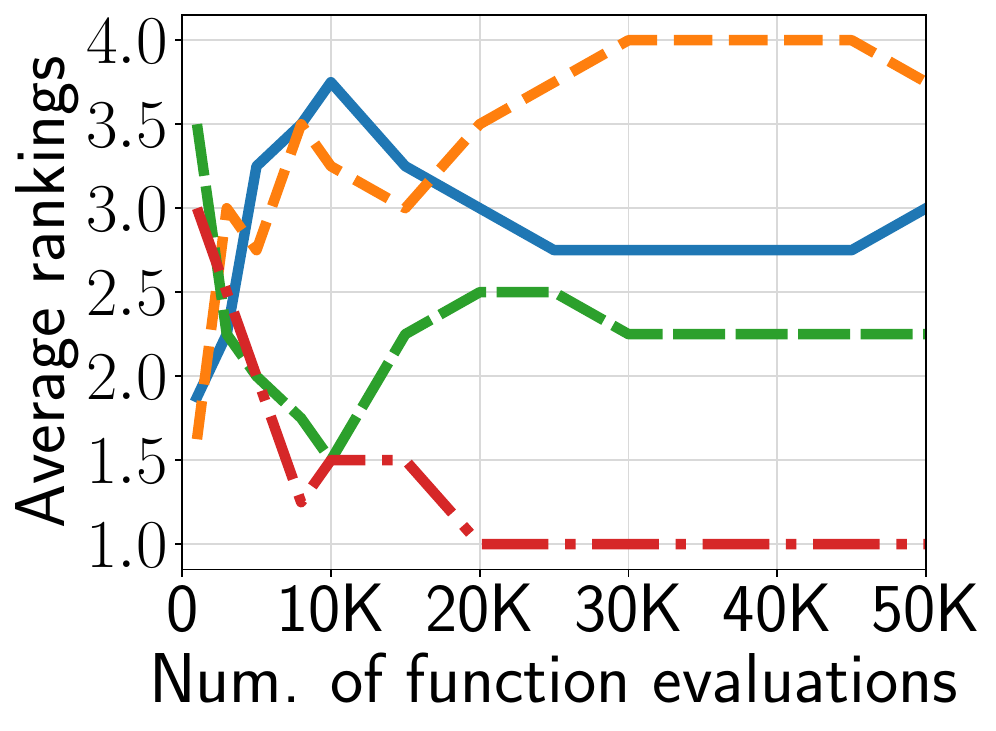}}
   \subfloat[IDTLZ ($m=4$)]{\includegraphics[width=\mywidth\textwidth]{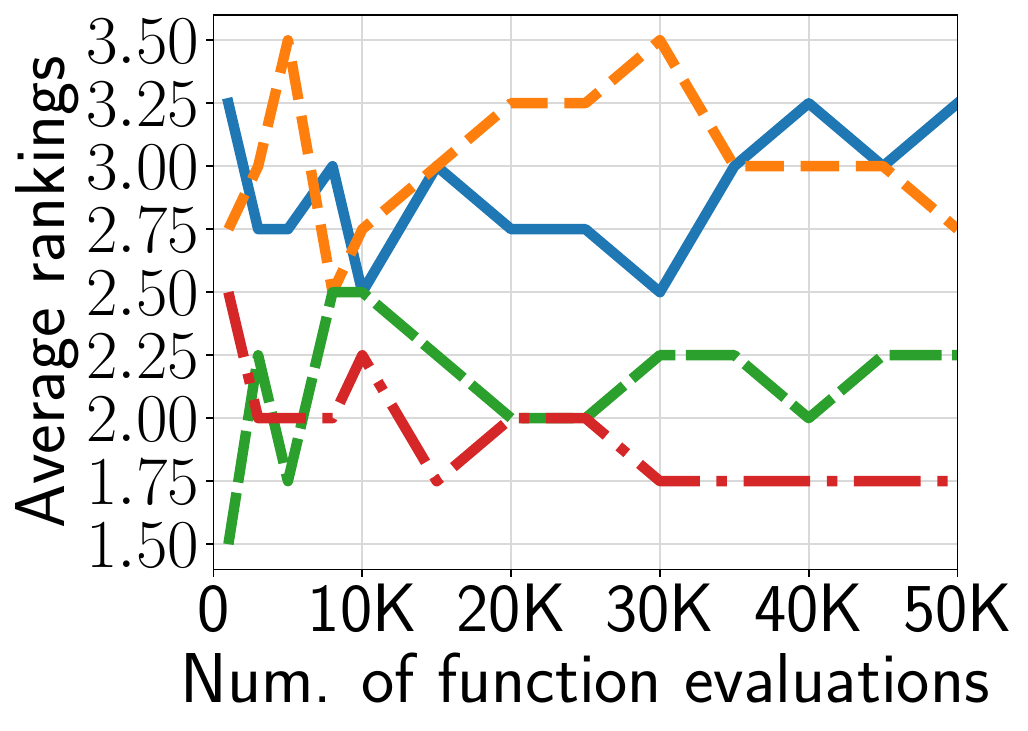}}
   \subfloat[IDTLZ ($m=6$)]{\includegraphics[width=\mywidth\textwidth]{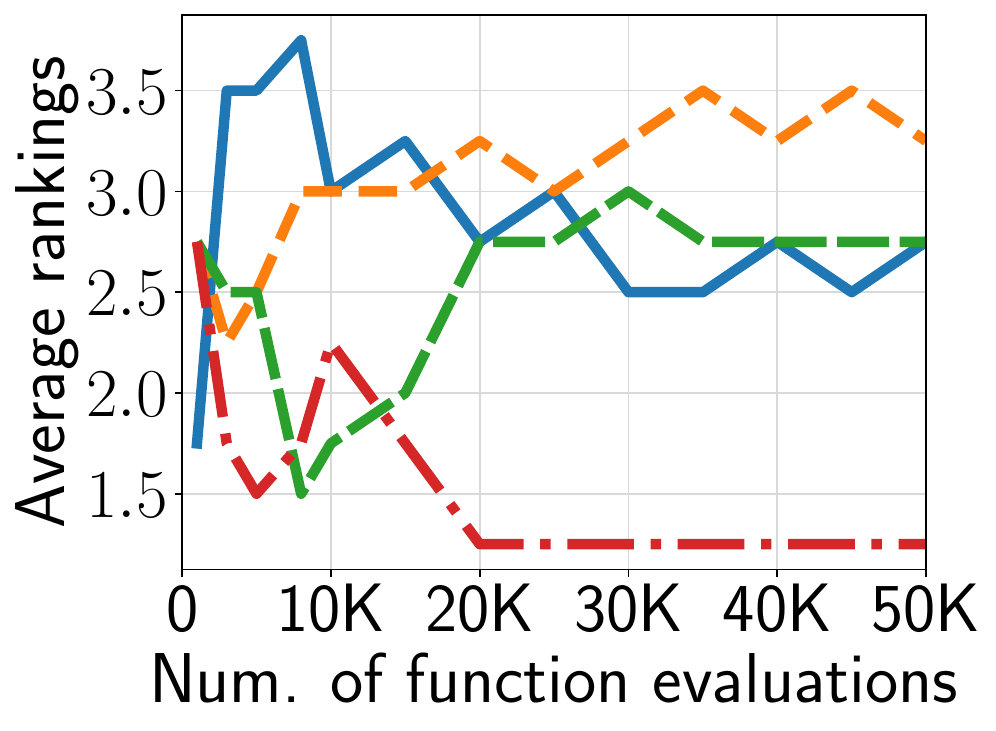}}
         \caption{Average rankings of R-NSGA-II with the three normalization methods (PP, BP, and BA) and with no normalization method (NO) on the three problem sets.}
   \label{fig:rnsga2_franking}
\end{figure*}

\begin{figure*}[t]
  \centering
\newcommand{\mywidth}{0.3}
\includegraphics[width=0.8\textwidth]{./figs/friedman_rank/legend.pdf}
\vspace{-3.9mm}
  \\
   \subfloat[DTLZ ($m=2$)]{\includegraphics[width=\mywidth\textwidth]{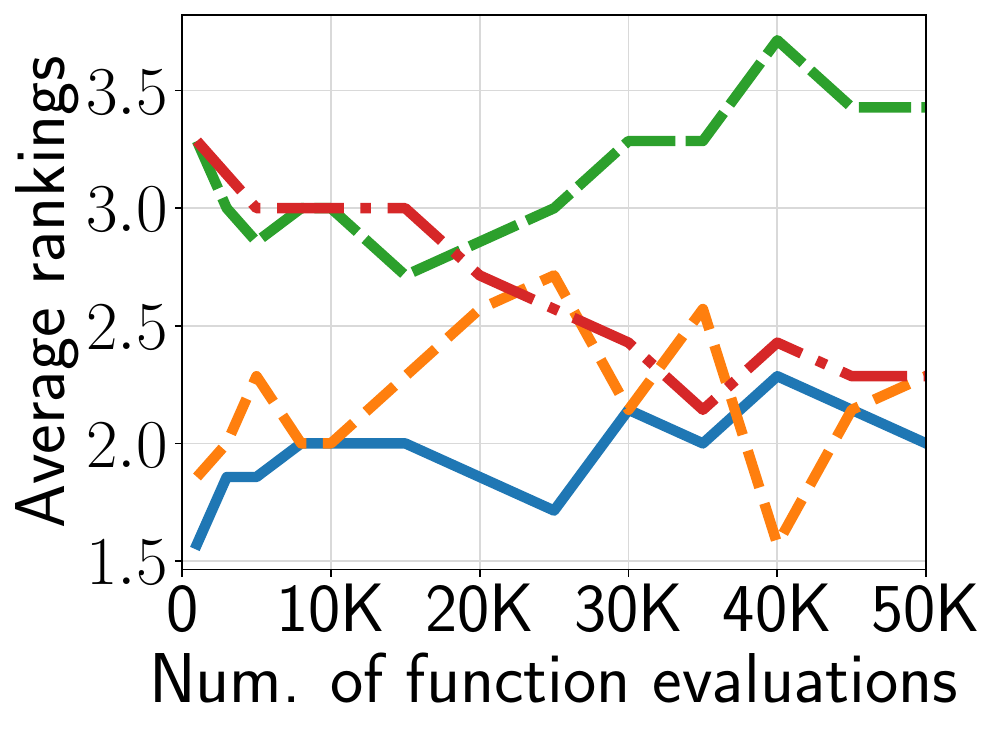}}
   \subfloat[DTLZ ($m=4$)]{\includegraphics[width=\mywidth\textwidth]{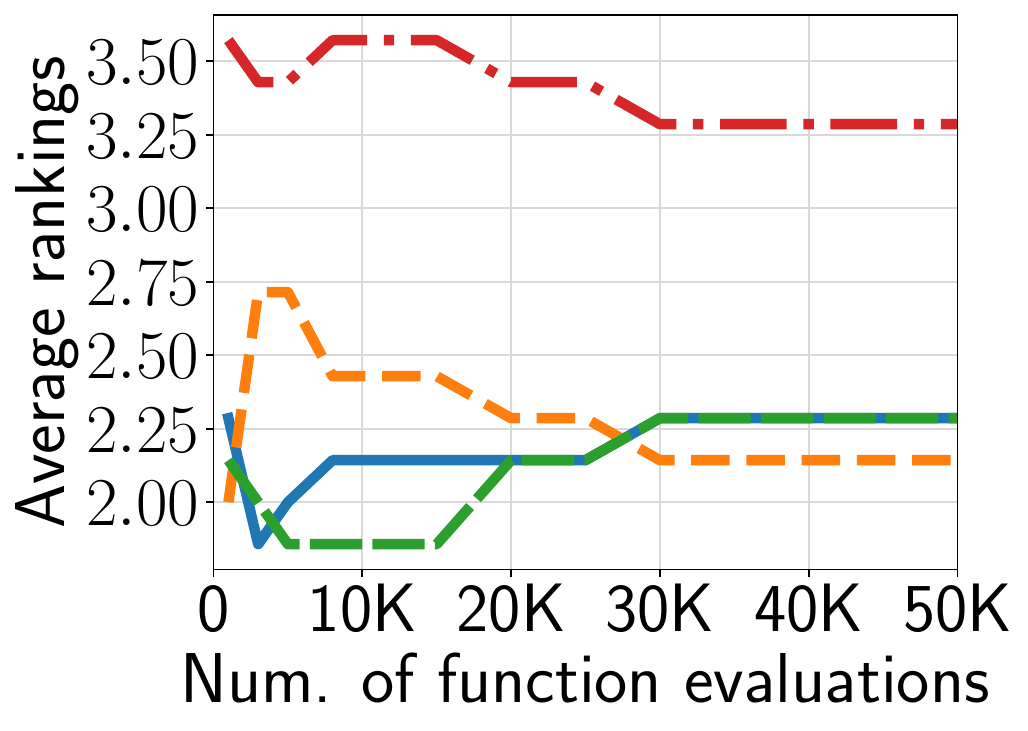}}
   \subfloat[DTLZ ($m=6$)]{\includegraphics[width=\mywidth\textwidth]{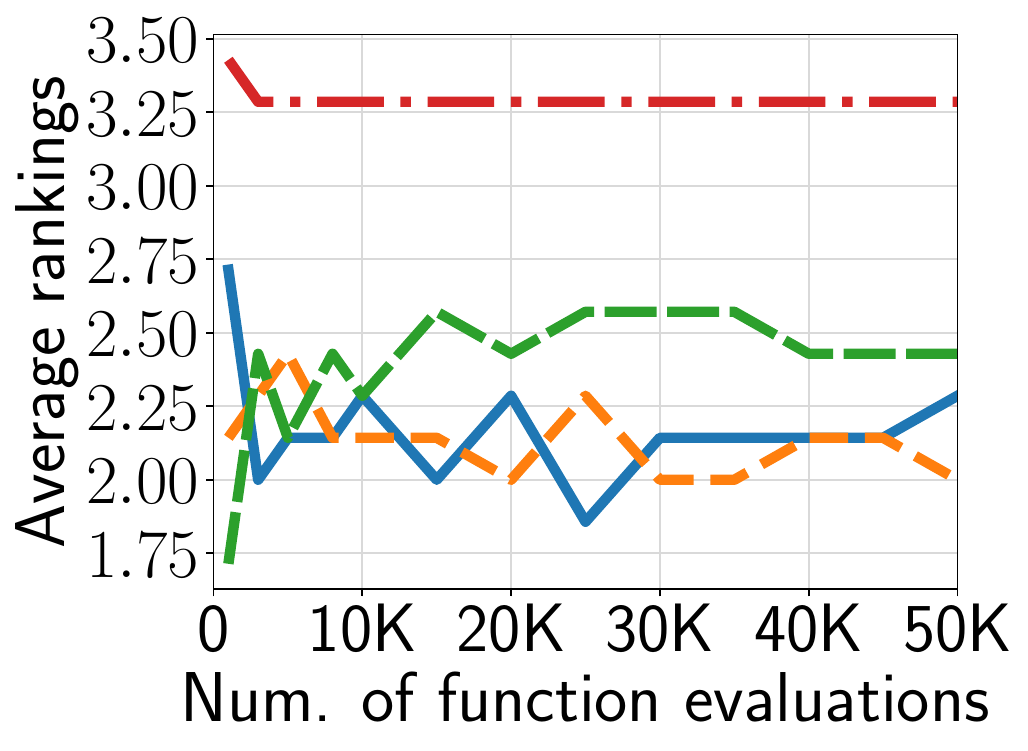}}
\\
   \subfloat[SDTLZ ($m=2$)]{\includegraphics[width=\mywidth\textwidth]{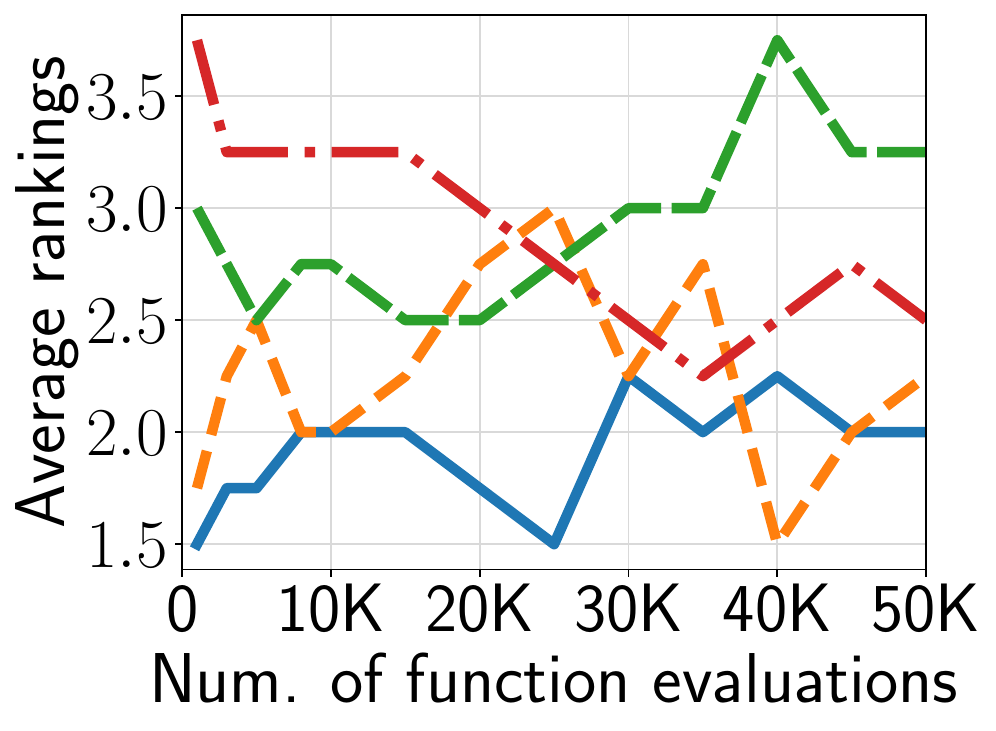}}
   \subfloat[SDTLZ ($m=4$)]{\includegraphics[width=\mywidth\textwidth]{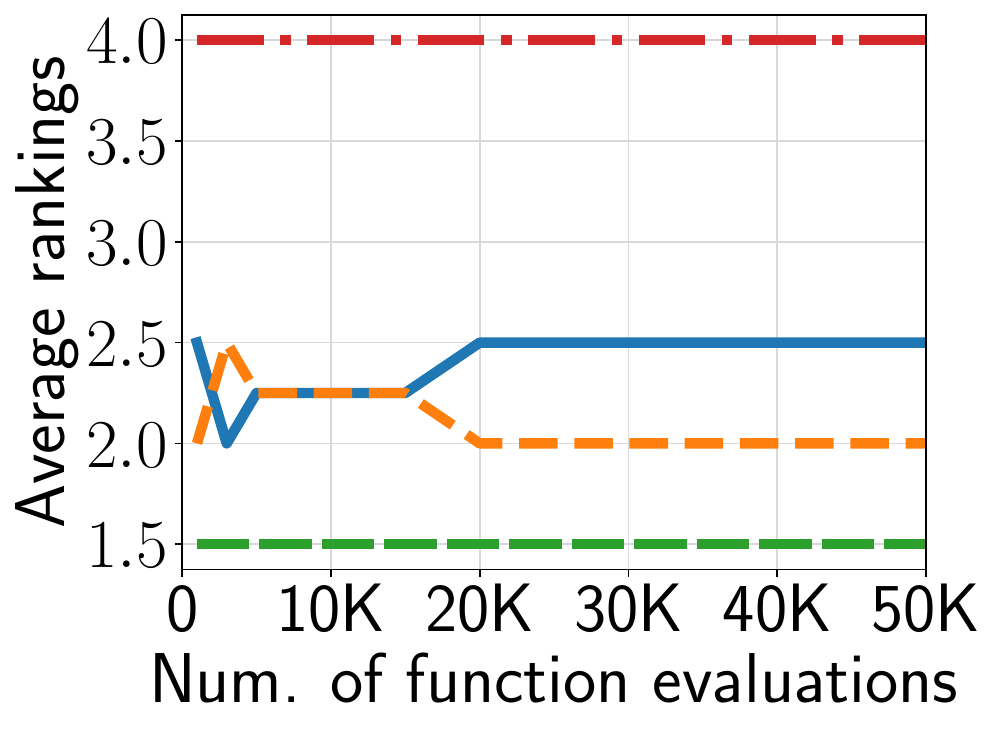}}
   \subfloat[SDTLZ ($m=6$)]{\includegraphics[width=\mywidth\textwidth]{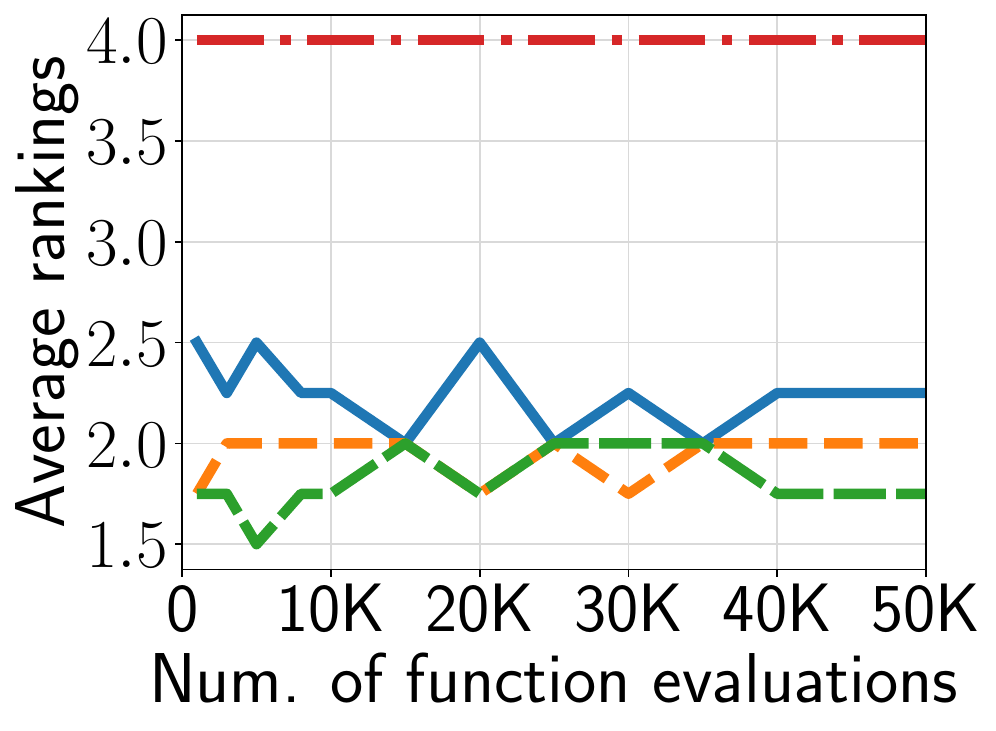}}
\\
   \subfloat[IDTLZ ($m=2$)]{\includegraphics[width=\mywidth\textwidth]{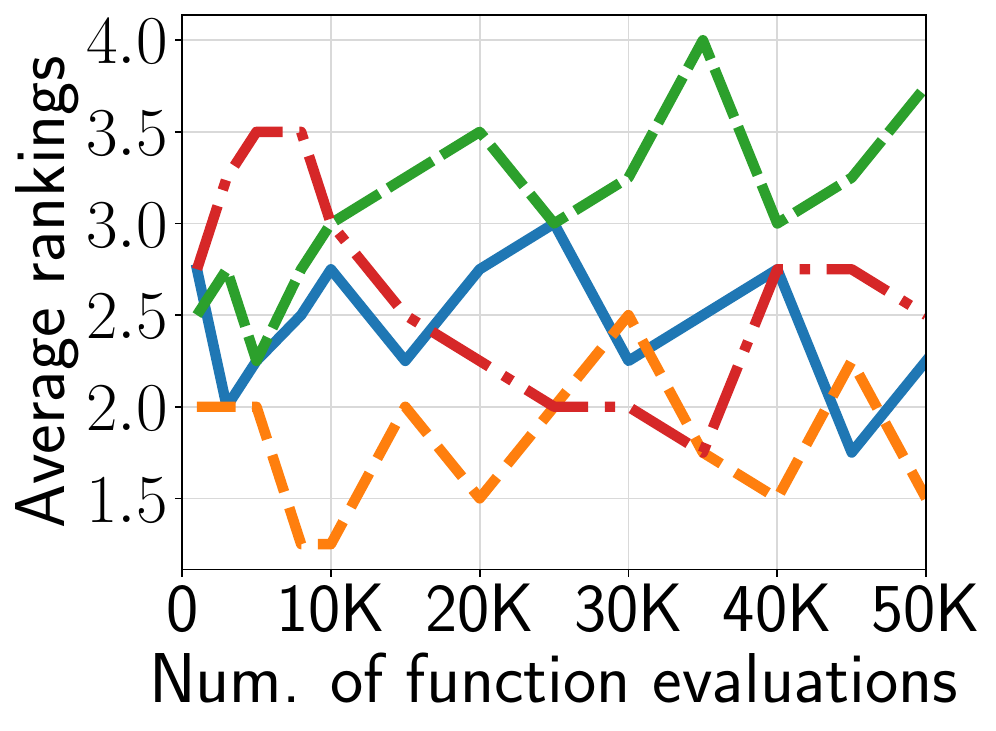}}
   \subfloat[IDTLZ ($m=4$)]{\includegraphics[width=\mywidth\textwidth]{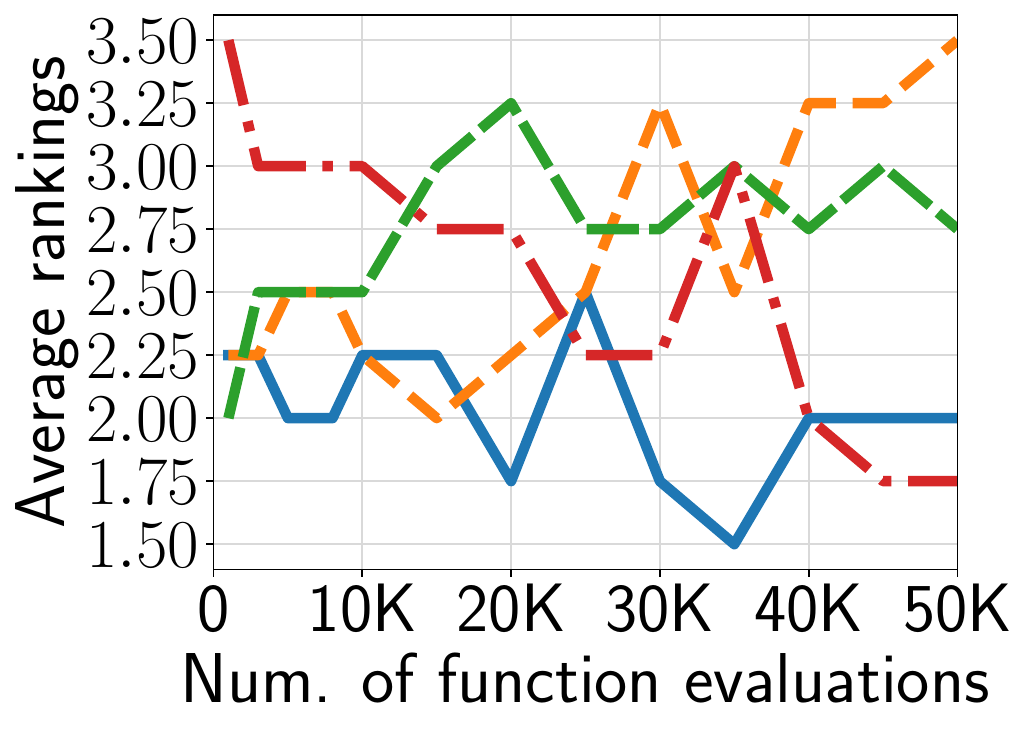}}
   \subfloat[IDTLZ ($m=6$)]{\includegraphics[width=\mywidth\textwidth]{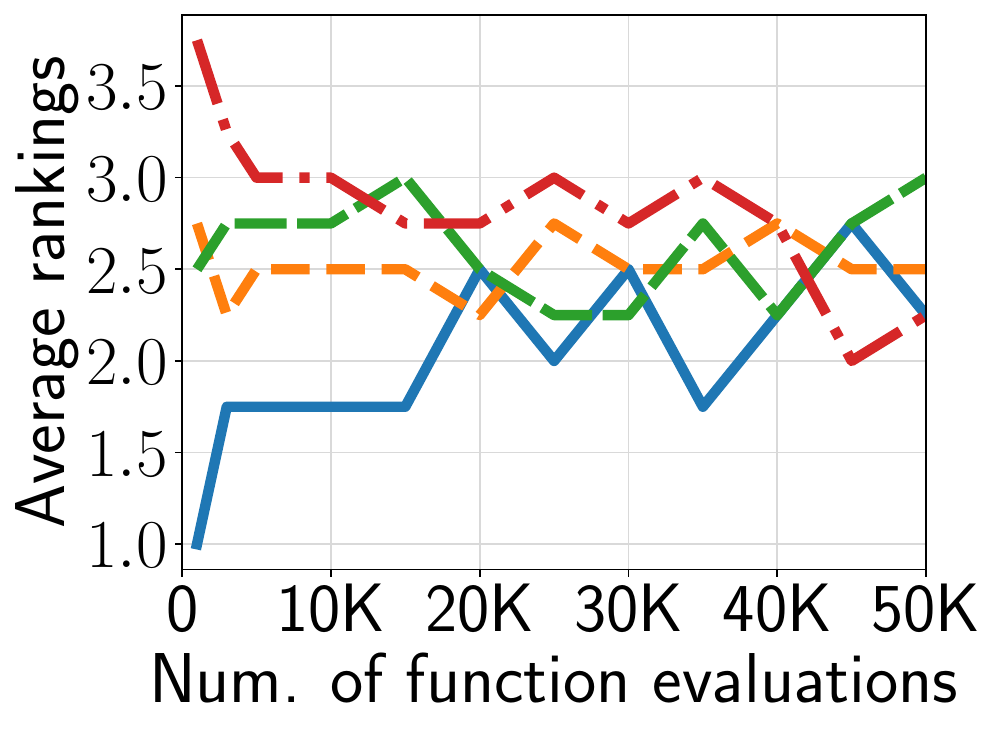}}
         \caption{Average rankings of r-NSGA-II with the three normalization methods (PP, BP, and BA) and with no normalization method (NO) on the three problem sets.}
   \label{fig:r2nsga2_franking}
\end{figure*}

\begin{figure*}[t]
  \centering
\newcommand{\mywidth}{0.3}
\includegraphics[width=0.8\textwidth]{./figs/friedman_rank/legend.pdf}
\vspace{-3.9mm}
  \\
   \subfloat[DTLZ ($m=2$)]{\includegraphics[width=\mywidth\textwidth]{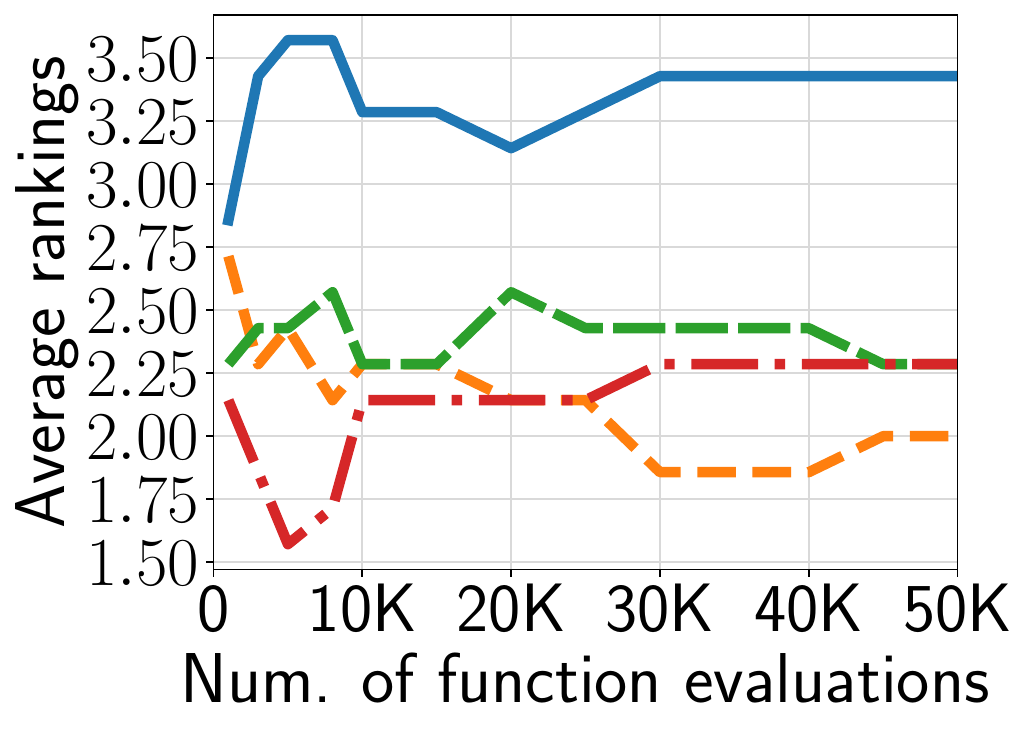}}
   \subfloat[DTLZ ($m=4$)]{\includegraphics[width=\mywidth\textwidth]{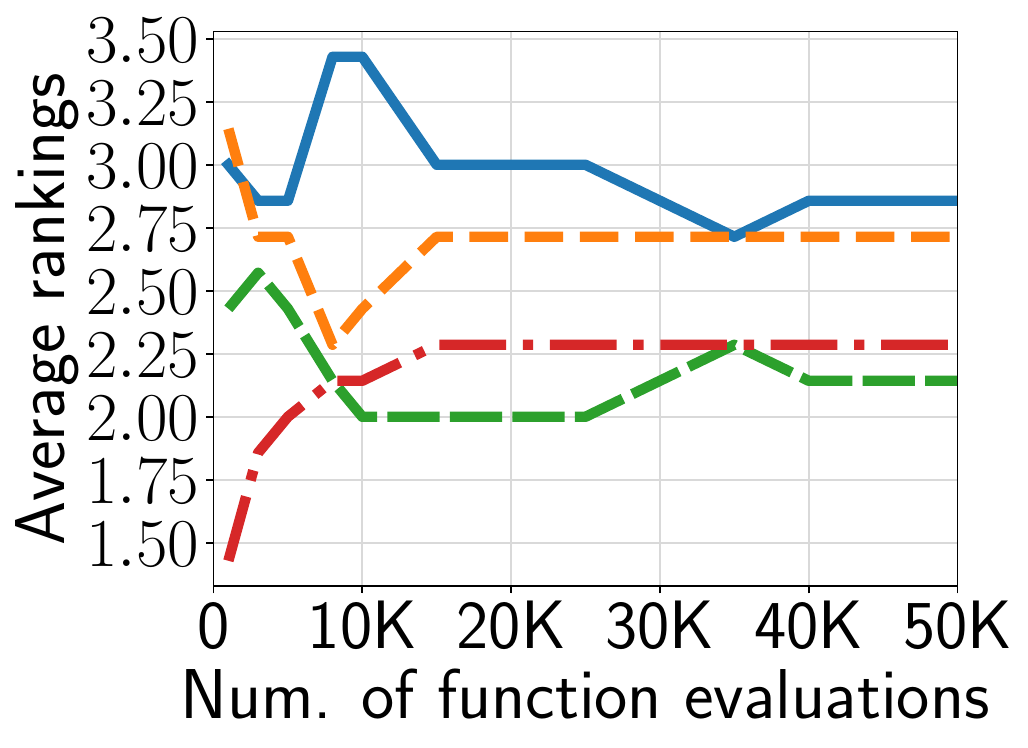}}
   \subfloat[DTLZ ($m=6$)]{\includegraphics[width=\mywidth\textwidth]{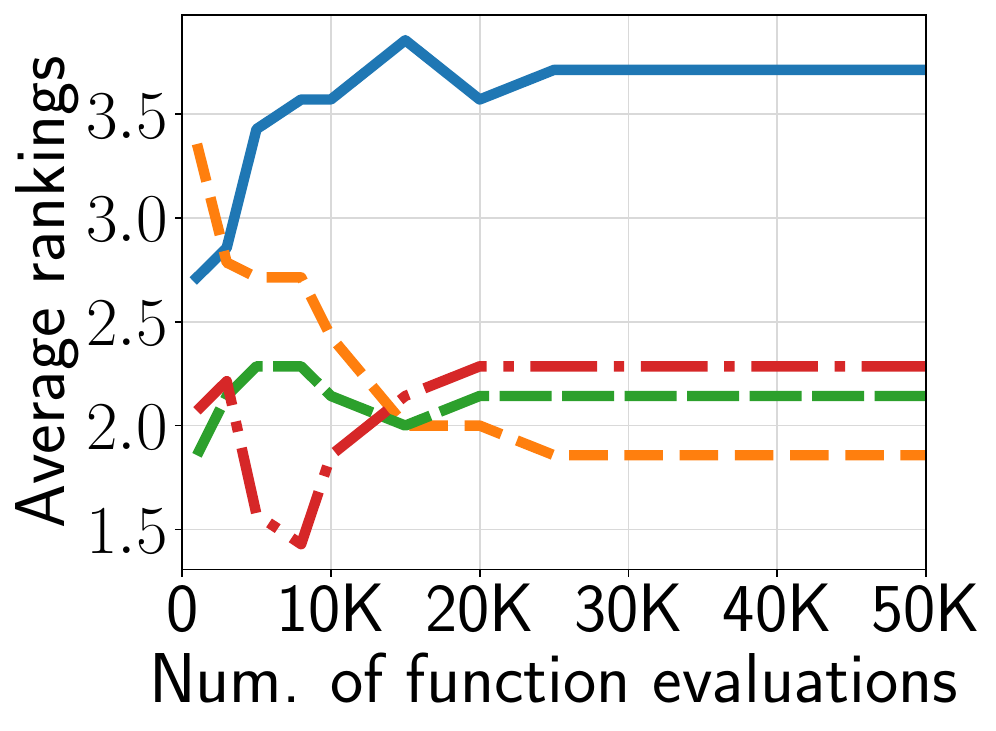}}
\\
   \subfloat[SDTLZ ($m=2$)]{\includegraphics[width=\mywidth\textwidth]{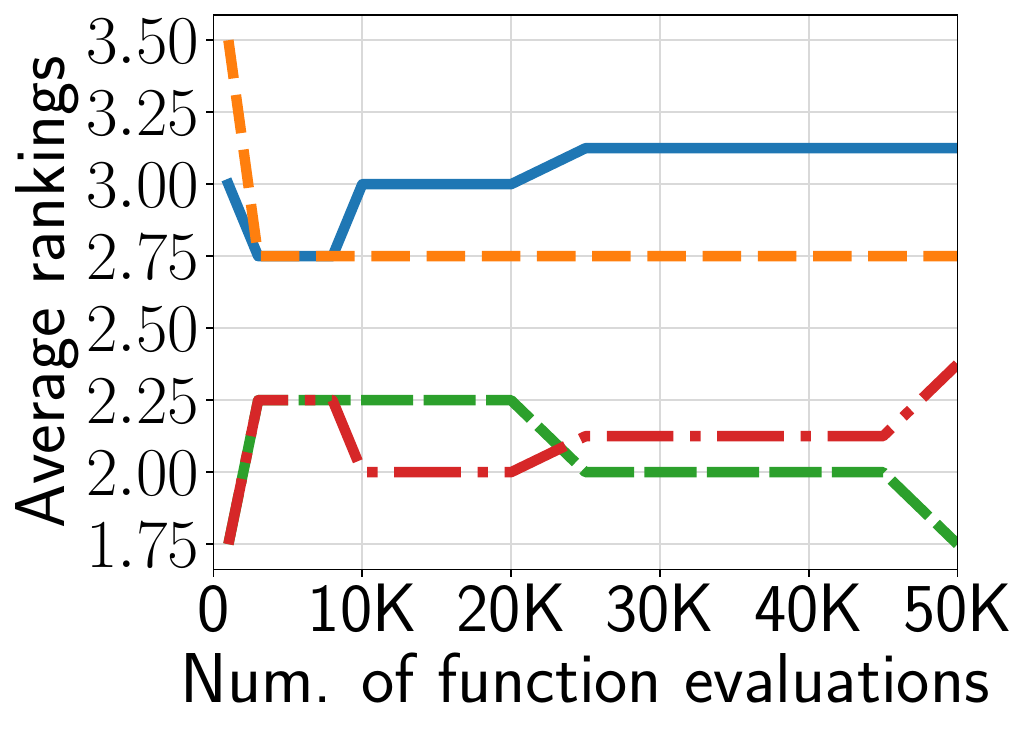}}
   \subfloat[SDTLZ ($m=4$)]{\includegraphics[width=\mywidth\textwidth]{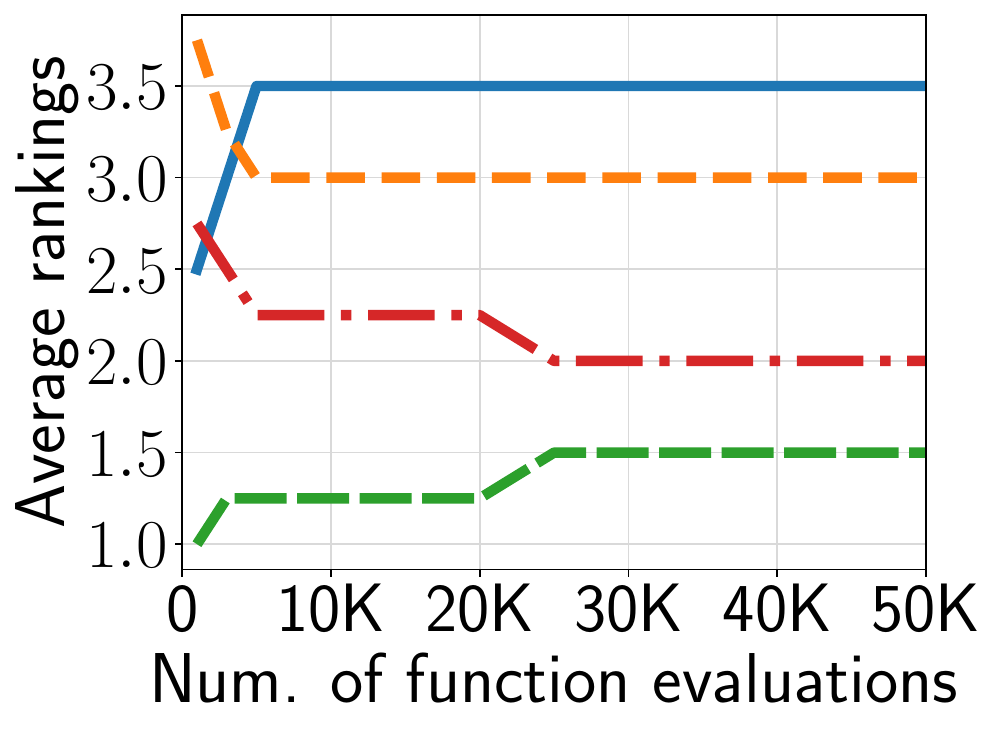}}
   \subfloat[SDTLZ ($m=6$)]{\includegraphics[width=\mywidth\textwidth]{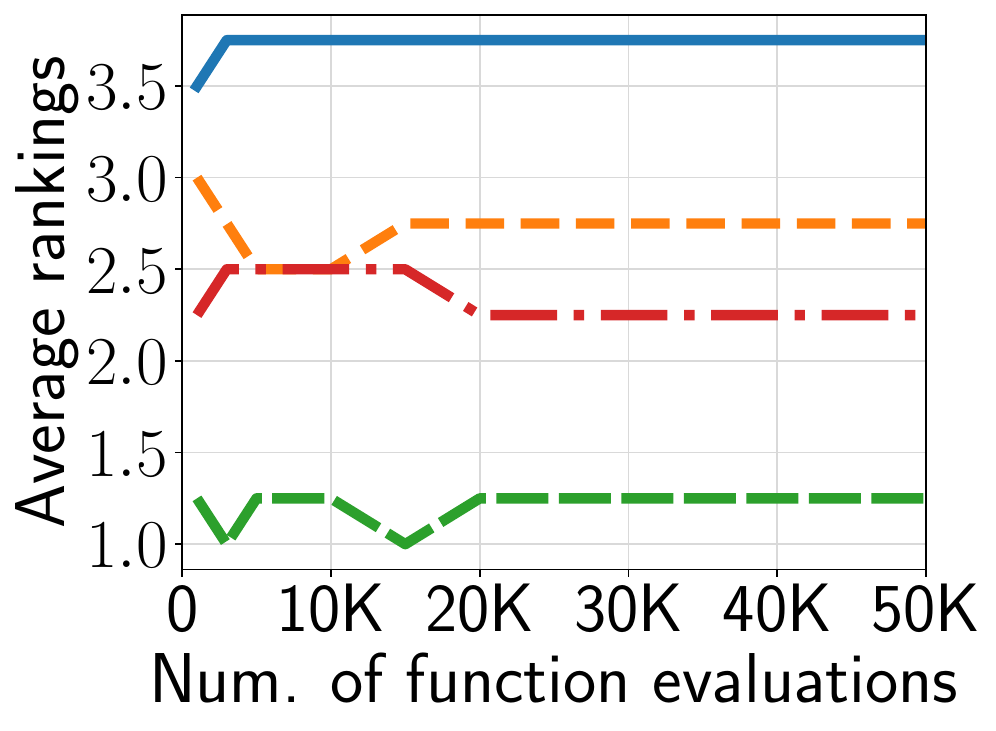}}
\\
   \subfloat[IDTLZ ($m=2$)]{\includegraphics[width=\mywidth\textwidth]{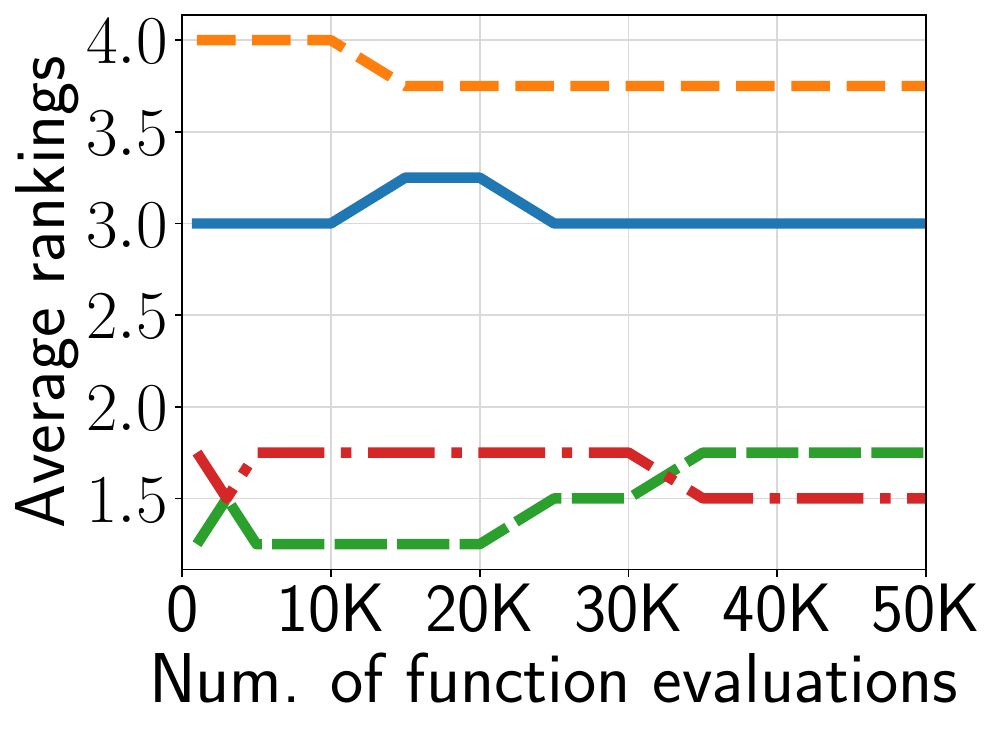}}
   \subfloat[IDTLZ ($m=4$)]{\includegraphics[width=\mywidth\textwidth]{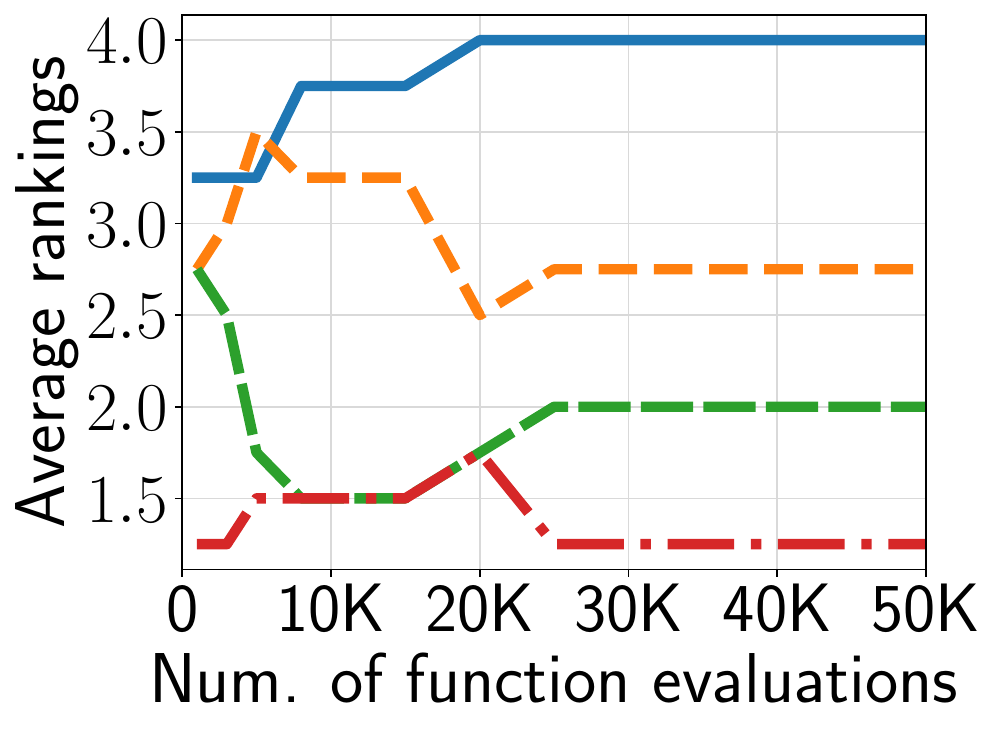}}
   \subfloat[IDTLZ ($m=6$)]{\includegraphics[width=\mywidth\textwidth]{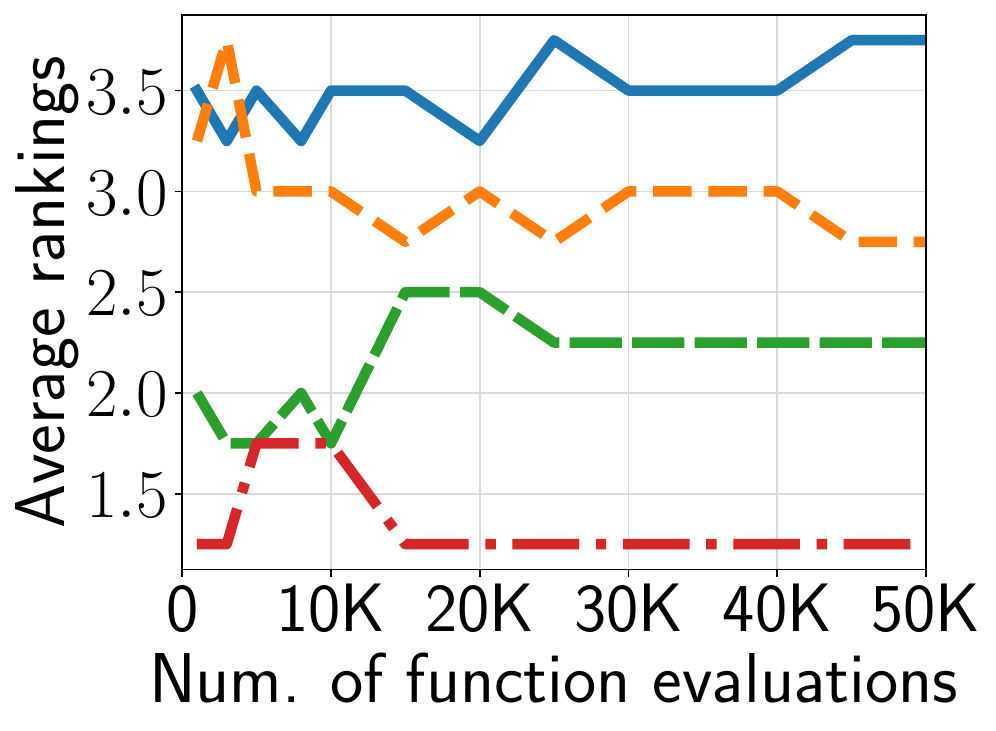}}
         \caption{Average rankings of MOEA/D-NUMS with the three normalization methods (PP, BP, and BA) and with no normalization method (NO) on the three problem sets.}        
   \label{fig:nums_franking}
\end{figure*}

\subsection{Comparison of normalization methods in PBEMO algorithms by means of IGD$^+$-C}
\label{sec:comparison_nm}

Figures \ref{fig:rnsga2_franking}--\ref{fig:nums_franking} show the average rankings of the three PBEMO algorithms (R-NSGA-II, r-NSGA-II, and MOEA/D-NUMS) on the three problem suites (DTLZ, SDTLZ, and IDTLZ) with $m \in \{2, 4, 6\}$ by the Friedman test \cite{DerracGMH11}, respectively.
We used the \texttt{CONTROLTEST} package (\url{https://sci2s.ugr.es/sicidm}) to calculate the rankings based on the IGD$^+$-C values.
We computed the rankings at $1\,000$, $3\,000$, $5\,000$, $8\,000$, $10\,000$, $15\,000$, $\ldots$,  $50\,000$ function evaluations.
In addition to the three normalization methods (PP, BP, and BA), Figures \ref{fig:rnsga2_franking}--\ref{fig:nums_franking} show the results of the three PBEMO algorithms with \underline{no} normalization method (NO).
We investigate the effectiveness of PP, BP, and BA by comparing them with NO.
Recall that the original MOEA/D-NUMS does not use any normalization method.
Thus, the performance of MOEA/D-NUMS with NO can be viewed as that of the original MOEA/D-NUMS.
Tables \ref{tab:sup_rnsga2_type1}--\ref{tab:sup_nums_type1} show the results of the three PBEMO algorithms on each test problem at $50\,000$ function evaluations.
Below, we refer to ``a PBEMO algorithm with a normalization method'' as ``a normalization method'' for simplicity.

\subsubsection{Results of R-NSGA-II}

As shown in the results of R-NSGA-II in Figures \ref{fig:rnsga2_franking}(a)--(c) and (g)--(i), NO shows the best performance in most cases on the DTLZ1--7 and IDTLZ1--4 problems.
Since the scales of objective functions in these test problems are almost the same, R-NSGA-II does not require normalization of objectives.
As shown in Figures \ref{fig:rnsga2_franking}(a)--(c), (g), and (h), BA performs the best in some cases among the three normalization methods.
However, as shown in Figure \ref{fig:rnsga2_franking}(i), PP is competitive for BA for the IDTLZ1--4 problems with $m=6$ at a later stage of the search.

In contrast to the results on the DTLZ1--7 and IDTLZ1--4 problems, NO performs the worst on the SDTLZ1--4 problems at any number of function evaluations.
Our results show that the best normalization method depends on the number of objectives $m$ and a budget of function evaluations.
For example, as seen from the results for $m=2$ in Figure \ref{fig:rnsga2_franking}(d), BP is the best at an early stage of the search, but BA performs the best a later stage of the search.
As seen from the results for $m=6$ in Figure \ref{fig:rnsga2_franking}(f), the rankings of PP, BP, and BA are almost the same.
\pref{sec:est_pbemo} shows that BA performs poorly in terms of ORE.
However, the results in this section show that the performance of R-NSGA-II with BA is competitive with that with PP and BP.


\subsubsection{Results of r-NSGA-II}

As shown in Figures \ref{fig:r2nsga2_franking}(b) and (c), in contrast to the results of R-NSGA-II, NO performs the worst on the DTLZ1--7 problems with $m \in \{4, 6\}$.
However, as shown in Figures \ref{fig:r2nsga2_franking}(a) and (g)--(i), NO is competitive with the three normalization methods on other non-scaled DTLZ problems.

As seen from \pref{fig:r2nsga2_franking}(d), PP is the best performer on the SDTLZ1--4 problems with $m=2$.
In contrast, BA performs significantly worse than even NO.
However, as shown in Figures \ref{fig:r2nsga2_franking}(e) and (f), BA shows the best performance in most cases for $m \geq 4$.

\subsubsection{Results of MOEA/D-NUMS}

As shown in Figures \ref{fig:nums_franking}(a)--(c), NO performs the best on the DTLZ1--7 problems at an early stage of the search. 
However, BP and BA perform the best at a later stage of the search.
As seen from Figures \ref{fig:nums_franking}(g)--(i), the performance of NO is the best on the IDTLZ1--4 problems in most cases, followed by BA. 

As shown in Figures \ref{fig:nums_franking}(d)--(f), BA performs the best on the SDTLZ1--4 problems for $m \in \{2, 4, 6\}$.
In the previous study \cite{LiCMY18}, MOEA/D-NUMS does not use any normalization method.
However, our results show that the performance of MOEA/D-NUMS on the test problems with differently scaled objective functions can be significantly improved by using BA.
As seen from Figures \ref{fig:nums_franking}(d)--(f), NO outperforms PP and BP even on the test problems with differently scaled objective functions.
Although this kind of negative effect of normalization methods in MOEA/D has been reported in \cite{IshibuchiDN17}, our results suggest that it can be applied to a preference-based version of MOEA/D.

\begin{tcolorbox}[title=Answers to RQ2, sharpish corners, top=2pt, bottom=2pt, left=4pt, right=4pt, boxrule=0.5pt]
With a few exceptions, our results show the importance of normalization methods for PBEMO algorithms on test problems with differently scaled objectives.
We observed that the best normalization method depends on various factors, including the number of objectives $m$, the type of problems, a budget of function evaluations, and the type of PBEMO algorithm.
Thus, there is no clear winner among the three normalization methods (PP, BP, and BA). 
However, roughly speaking, our results indicate that BA is generally suitable for PBEMO algorithms, especially for MOEA/D-NUMS.
We found that the performance of normalization methods in terms of ORE and IGD$^+$-C is not strongly related to each other.
This observation suggests that the positions of the approximated ideal and nadir points may play an important role in PBEMO.

\end{tcolorbox}

\section{Conclusion}
\label{sec:conclusion}

In this paper, we first described the three normalization methods (PP, BP, and BA) for PBEMO in \pref{sec:review}.
We presented the computationally cheap archiving method for approximating the nadir point in \pref{sec:nm_apd}.
Then, we investigated the effects of normalization methods in PBEMO on the DTLZ, SDTLZ, and IDTLZ problems.
In \pref{sec:est_pbemo}, the results showed that the normalization methods in PBEMO perform significantly worse than that in conventional EMO (NSGA-II) in terms of approximating the ideal point, nadir point, and range of the PF.
In \pref{sec:comparison_nm}, the results showed the importance of normalization methods in PBEMO to handle objective functions with different scales.
The results also showed that the best normalization method depends on various factors, but BA is generally effective for PBEMO.

We believe that our findings are helpful in selecting the normalization method for a PBEMO algorithm.
Our findings are also helpful in designing a new PBEMO algorithm.
In this paper, we focused on PBEMO using the reference point.
An investigation of the effects of normalization methods in PBEMO using other preference elicitation methods is an avenue for future work.
It may also be interesting to analyze the influence of normalization methods on the performance of interactive PBEMO.

 \section*{Acknowledgement}

This work was supported by JSPS KAKENHI Grant Number 21K17824 and LEADER, MEXT, Japan.

\bibliographystyle{plain}
\bibliography{reference} 
\clearpage

\begin{figure*}
\centering
\fontsize{30pt}{100pt}\selectfont{\textbf{Supplement}}
\end{figure*}

\setcounter{figure}{0}
\setcounter{table}{0}

\renewcommand{\thesection}{S.\arabic{section}}
\renewcommand{\thetable}{S.\arabic{table}}
\renewcommand{\thefigure}{S.\arabic{figure}}
\renewcommand\thealgocf{S.\arabic{algocf}} 
\renewcommand{\theequation}{S.\arabic{equation}}


\definecolor{c1}{RGB}{150,150,150}
\definecolor{c2}{RGB}{220,220,220}

\section{On the performance of the normalization method in MOEA/D-NUMS in terms of $e^{\mathrm{ideal}}$}
\label{supsec:nums_dtlz}

Here, we discuss why the normalization methods in MOEA/D-NUMS can quickly find the ideal point in Section \ref{sec:est_pbemo}.
As seen from Figure \ref{fig:3error_SDTLZ1_m4}(c), PP, BP, and BA achieve $e^{\mathrm{ideal}} = 0$ at 200 function evaluations.
Similar results can be found on other test problems, e.g., DTLZ2--4.
However, this amazing performance is simply due to the synergy of the properties of MOEA/D-NUMS and some DTLZ problems.
As described in \pref{sec:setting}, MOEA/D-NUMS used the DE operator.
Compared to the GA operator, the DE operator is likely to generate a solution outside the bound constraints due to differential mutation.
Such an infeasible solution is repaired by polynomial mutation using the replacement operator.
Let $x_i$ be the $i$-th element in the infeasible solution $\vec{x}$, where $i \in \{1, \ldots, n\}$, and $n$ is the number of variables.
In the replacement operator, $x_i$ is replaced with the corresponding minimum or maximum value.
In addition, an objective function in some DTLZ problems (including SDTLZ1) can be minimized by solutions on bounds.
For example, in the (S)DTLZ1 problem with the bounds $[0, 1]^n$, $f_1(\vec{x}) = 0$ if $\vec{x} = (0, *, \ldots, *)^{\top}$.
Similarly, $f_2(\vec{x}) = 0$ if $\vec{x} = (*, 1, \ldots, *)^{\top}$.

Similar to other PBEMO algorithms, normalization methods in MOEA/D-NUMS perform poorly on test problems without this kind of property.
For example, an objective function in IDTLZ1 cannot be minimized by solutions on bounds.
For this reason, as shown in \pref{fig:3error_IDTLZ1_m4}(c), the three normalization methods in MOEA/D-NUMS perform significantly worse than that in NSGA-II in terms of $e^{\mathrm{ideal}}$.

\begin{table}[t]  
\setlength{\tabcolsep}{4.5pt} 
\renewcommand{\arraystretch}{0.75}
\centering
\caption{\small Balanced and extreme settings of the reference point $\vec{z}$ for each DTLZ problem. For SDTLZ*, for $i \in \{1, \ldots, m\}$, $z_i$ in SDTLZ$*$ was $z_i$ in DTLZ$*$ multiplied by $10^{i-1}$. For IDTLZ*, the same $\vec{z}$ in DTLZ$*$ was used.
}
{\small
  \label{tab:ref_points_dtlz}
\subfloat[Balanced reference point]{    
\begin{tabular}{lllllcccc}
  \toprule
$m$ & Problem & the reference point $\vec{z}$\\
\midrule  
& DTLZ1 & $\vec{z} = (0.24, 0.18, 0.18)^{\top}$\\
$3$ & DTLZ2, 3, 4 & $\vec{z} = (0.8, 0.6, 0.6)^{\top}$\\
& DTLZ5, 6 & $\vec{z} = (0.65, 0.65, 0.74)^{\top}$\\
& DTLZ7 & $\vec{z} = (0.75, 0.15, 6.0)^{\top}$\\
\midrule
& DTLZ1 & $\vec{z} = (0.134, 0.12, 0.16, 0.12, 0.134)^{\top}$\\
$5$ & DTLZ2, 3, 4 & $\vec{z} = (0.556, 0.5, 0.666, 0.5, 0.556)^{\top}$\\
& DTLZ5, 6 & $\vec{z} = (0.4, 0.4, 0.56, 0.8, 0.7)^{\top}$\\
\midrule
& DTLZ1 & $\vec{z} = (0.08, 0.08, 0.074, 0.08, 0.086, 0.074, 0.068, 0.068)^{\top}$\\
$8$ & DTLZ2, 3, 4 & $\vec{z} = (0.45, 0.45, 0.415, 0.45, 0.486, 0.415, 0.381, 0.381)^{\top}$\\
& DTLZ5, 6 & $\vec{z} = (0.12, 0.12, 0.17, 0.24, 0.34, 0.48, 0.68, 0.42)^{\top}$\\
\midrule
& DTLZ1 & $\vec{z} = (0.06, 0.065, 0.06, 0.0436, 0.0545, 0.049, 0.0545, 0.049, 0.06, 0.049)^{\top}$\\
$10$ & DTLZ2, 3, 4 & $\vec{z} = (0.4, 0.437, 0.4, 0.29, 0.364, 0.328, 0.364, 0.328, 0.4, 0.328)^{\top}$\\
& DTLZ5, 6 & $\vec{z} = (0, 0, 0, 0.0035, 0.01, 0.031, 0.0963, 0.29, 0.88, 0.7)^{\top}$\\
\bottomrule 
\end{tabular}
}
\\
\subfloat[Extreme reference point]{    
\begin{tabular}{lllllcccc}
  \toprule
$m$ & Problem & the reference point $\vec{z}$\\
\midrule  
& DTLZ1 & $\vec{z} = (0.15, 0.15, 0.45)^{\top}$\\
$3$ & DTLZ2, 3, 4 & $\vec{z} = (0.4, 1.2, 0.4)^{\top}$\\
& DTLZ5, 6 & $\vec{z} = (0.4, 0.4, 1.2)^{\top}$\\
\midrule
& DTLZ1 & $\vec{z} = (0.03, 0.18, 0.33, 0.03, 0.03)^{\top}$\\
$5$ & DTLZ2, 3, 4 & $\vec{z} = (0.15, 1.2, 0.187, 0.168, 0.15)^{\top}$\\
& DTLZ5, 6 & $\vec{z} = (0.18, 0.18, 0.255, 0.36, 1.05)^{\top}$\\
\midrule
& DTLZ1 & $\vec{z} = (0.3, 0.042, 0.048, 0.042, 0.042, 0.036, 0.048, 0.042)^{\top}$\\
$8$ & DTLZ2, 3, 4 & $\vec{z} = (0.15, 0.128, 0.173, 0.15, 1.071, 0.15, 0.173, 0.15)^{\top}$\\
& DTLZ5, 6 & $\vec{z} = (0.07, 0.07, 0.1, 0.1415, 0.2, 0.283, 0.4, 1.2)^{\top}$\\
\midrule
& DTLZ1 & $\vec{z} = (0.03, 0.036, 0.03, 0.036, 0.036, 0.3, 0.03, 0.036, 0.03, 0.036)^{\top}$\\
$10$ & DTLZ2, 3, 4 & $\vec{z} = (0.14, 0.14, 1.164, 0.117, 0.14, 0.117, 0.14, 0.117, 0.14, 0.117)^{\top}$\\
& DTLZ5, 6 & $\vec{z} = (0, 0, 0, 0, 0.0144, 0.04, 0.12, 0.37, 1.13, 0.12)^{\top}$\\
\bottomrule 
\end{tabular}
}
}
\end{table}

  


\begin{figure*}[t]
\centering
  \subfloat{\includegraphics[width=0.7\textwidth]{./figs/legend/legend_3.pdf}}
\vspace{-3.9mm}
   \\
   \subfloat[$e^{\mathrm{ideal}}$ ($m=2$)]{\includegraphics[width=0.32\textwidth]{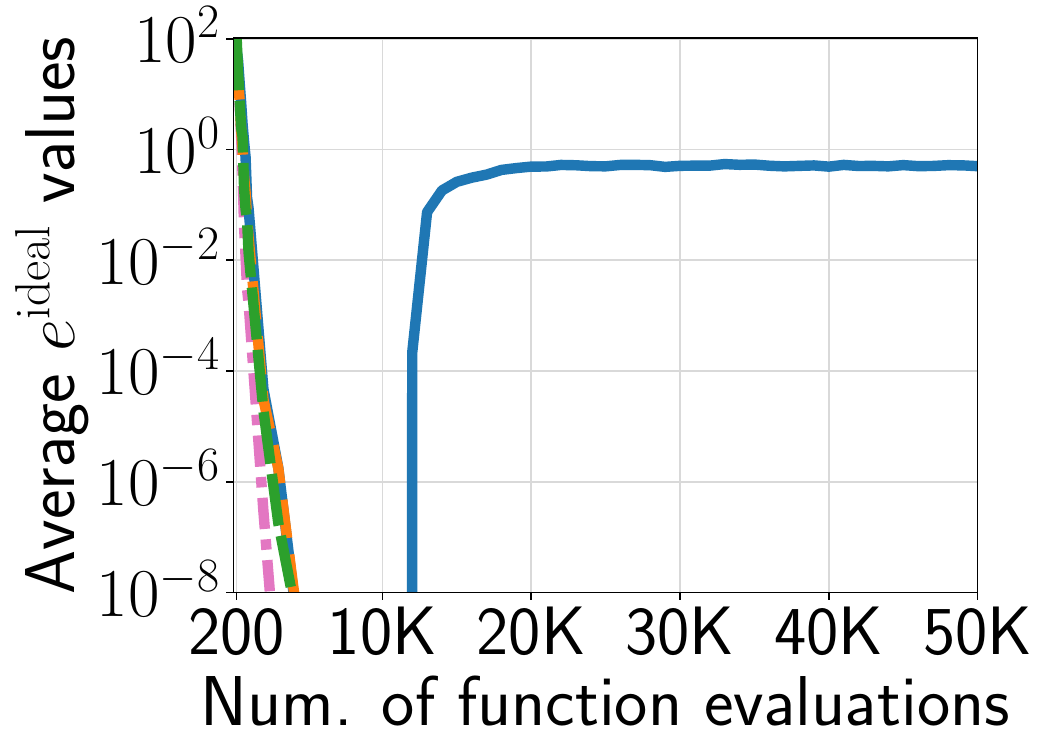}}
   \subfloat[$e^{\mathrm{ideal}}$ ($m=4$)]{\includegraphics[width=0.32\textwidth]{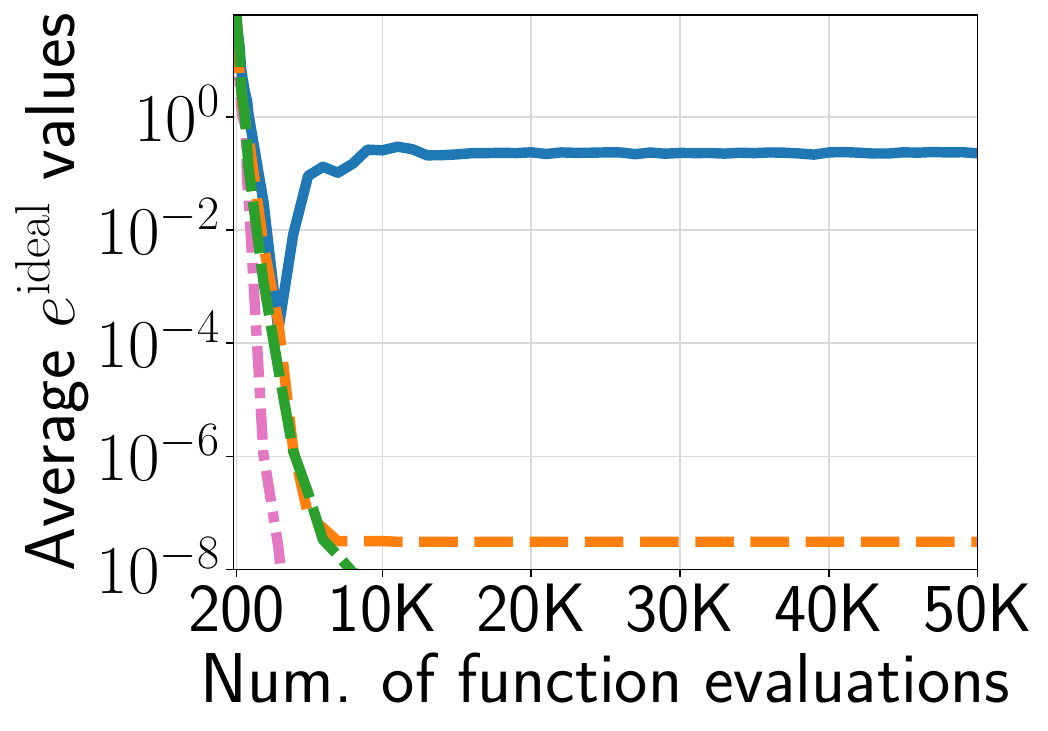}}
   \subfloat[$e^{\mathrm{ideal}}$ ($m=6$)]{\includegraphics[width=0.32\textwidth]{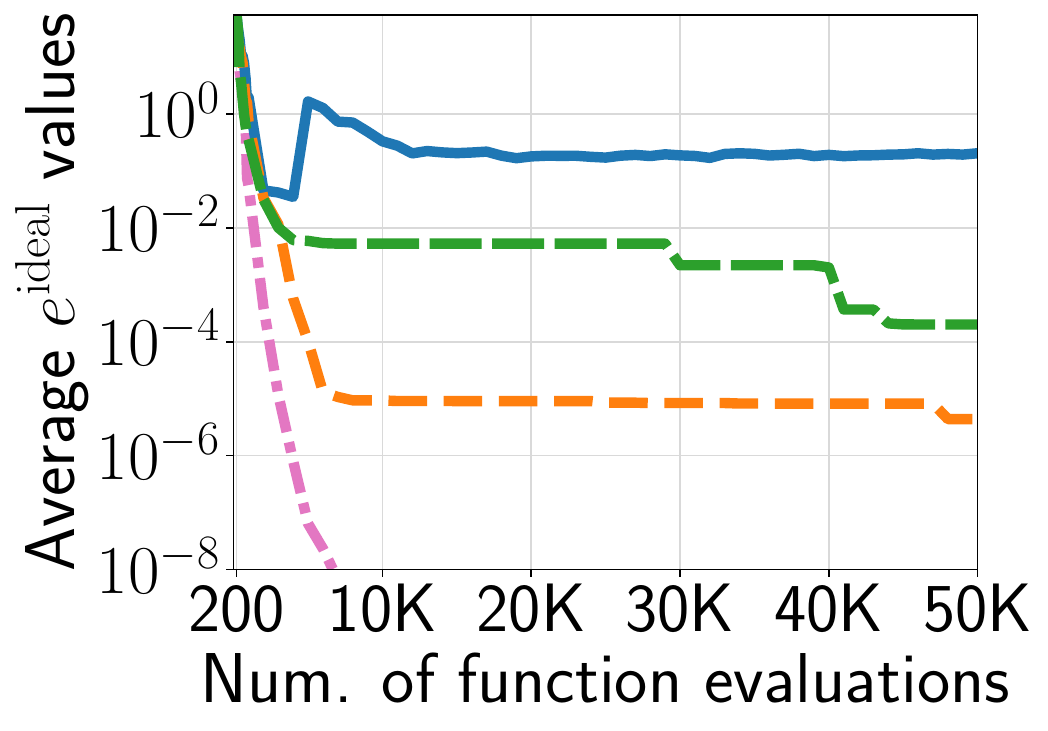}}
\\
   \subfloat[$e^{\mathrm{nadir}}$ ($m=2$)]{\includegraphics[width=0.32\textwidth]{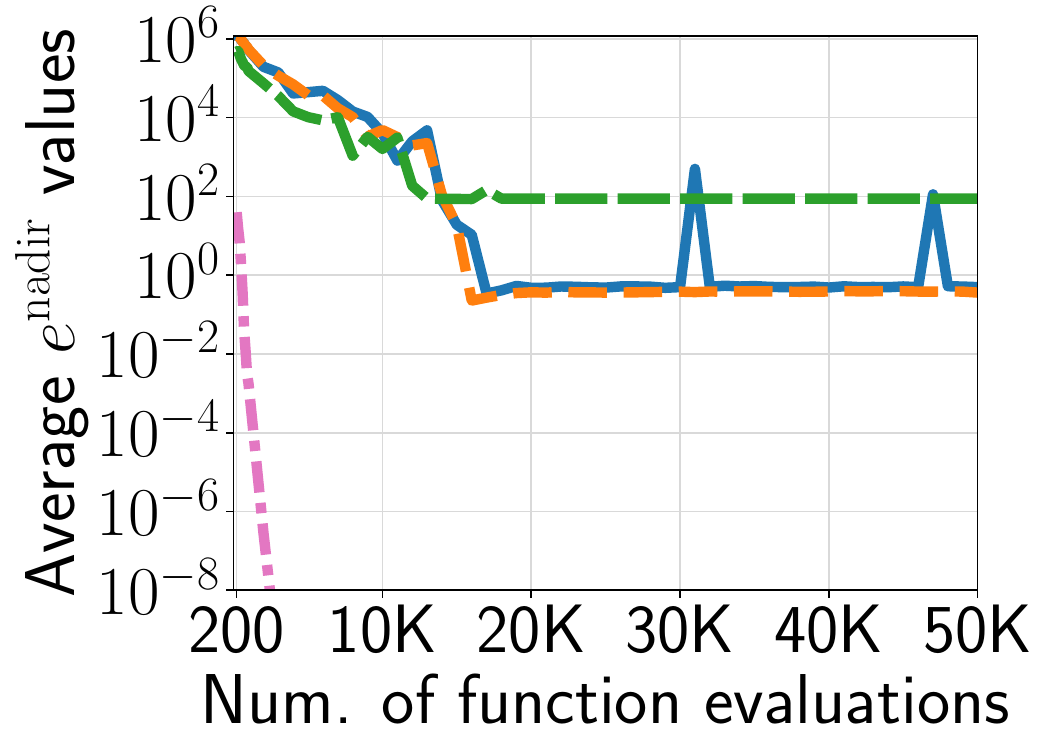}}
   \subfloat[$e^{\mathrm{nadir}}$ ($m=4$)]{\includegraphics[width=0.32\textwidth]{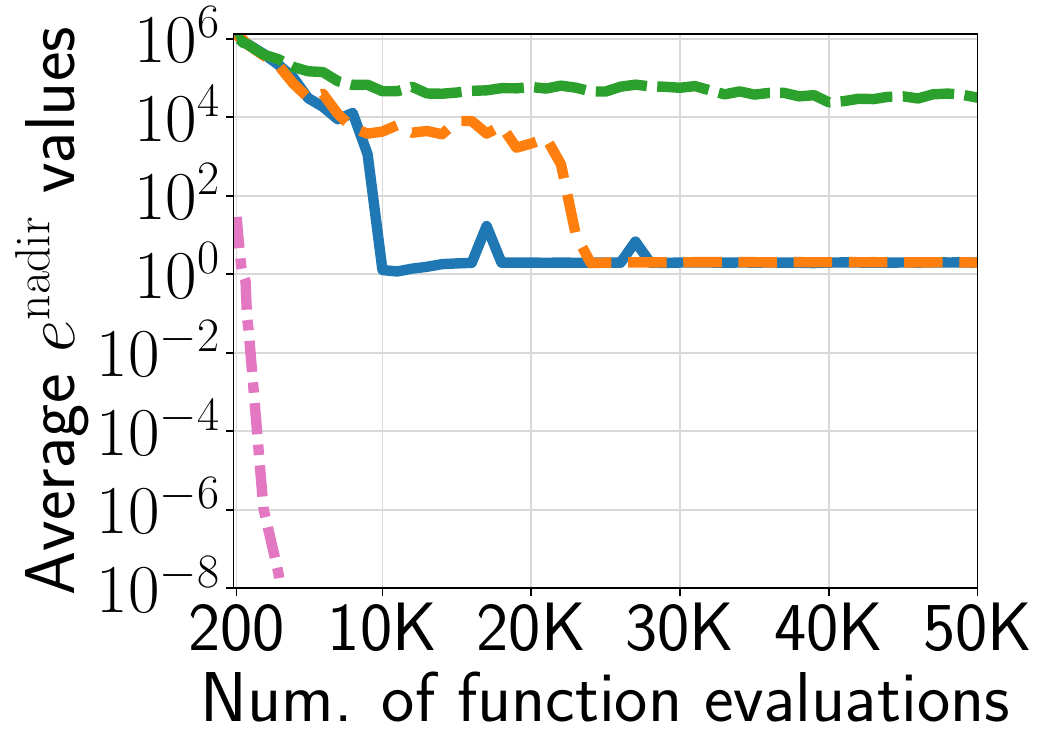}}
   \subfloat[$e^{\mathrm{nadir}}$ ($m=6$)]{\includegraphics[width=0.32\textwidth]{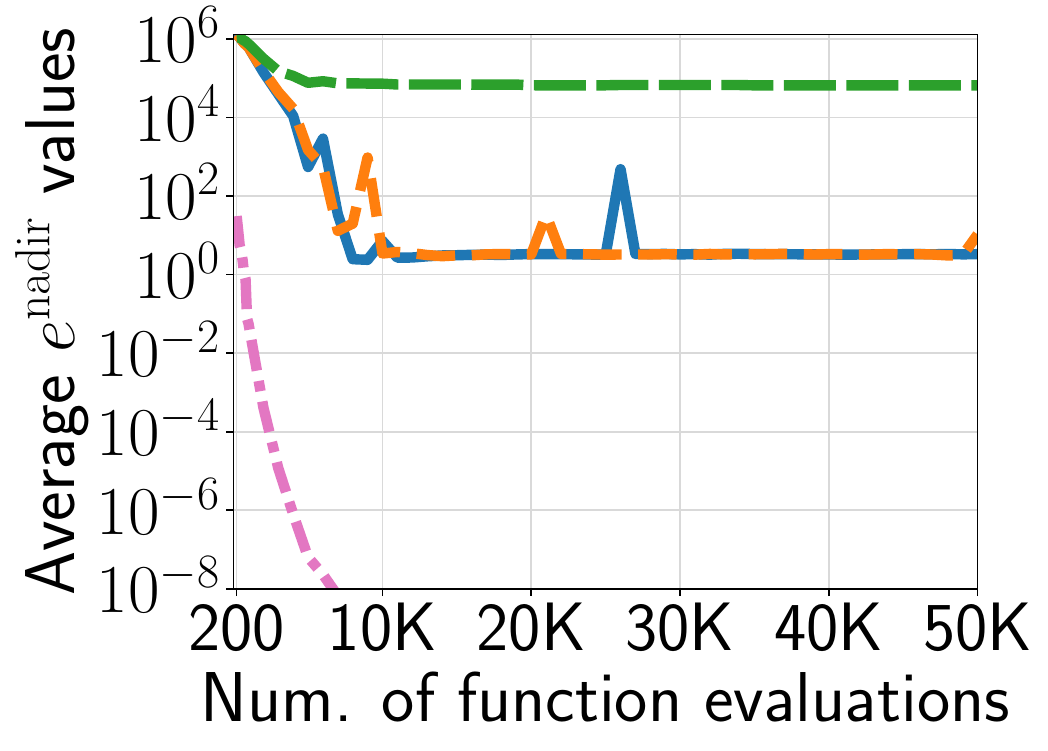}}
\\
   \subfloat[ORE ($m=2$)]{\includegraphics[width=0.32\textwidth]{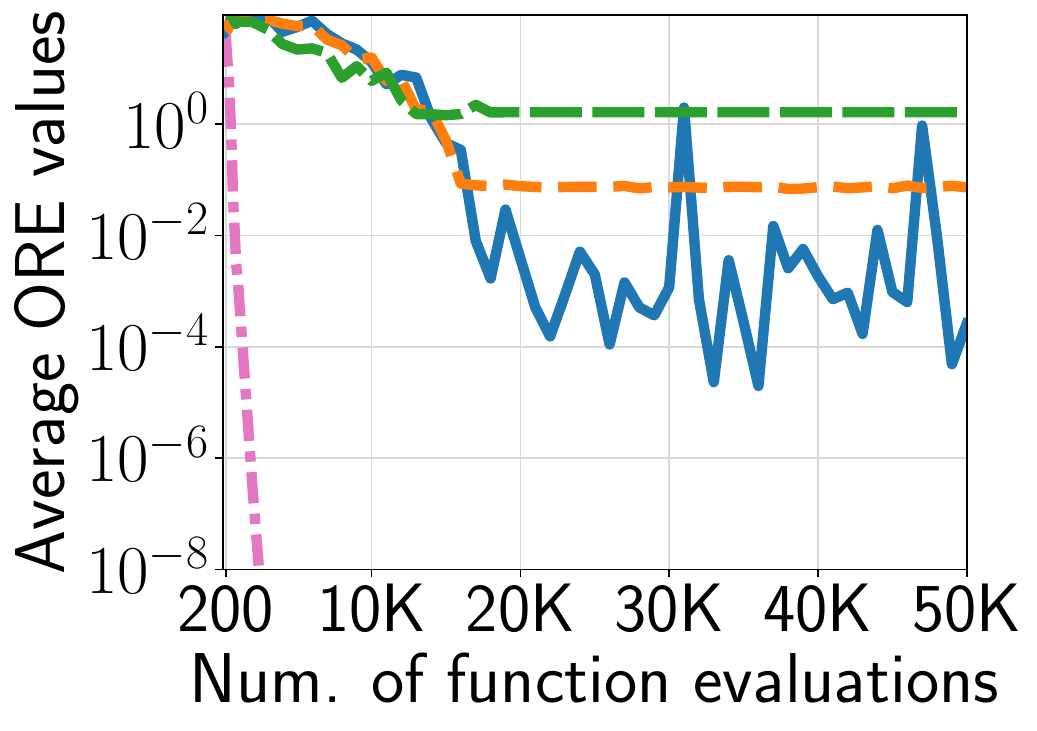}}
   \subfloat[ORE ($m=4$)]{\includegraphics[width=0.32\textwidth]{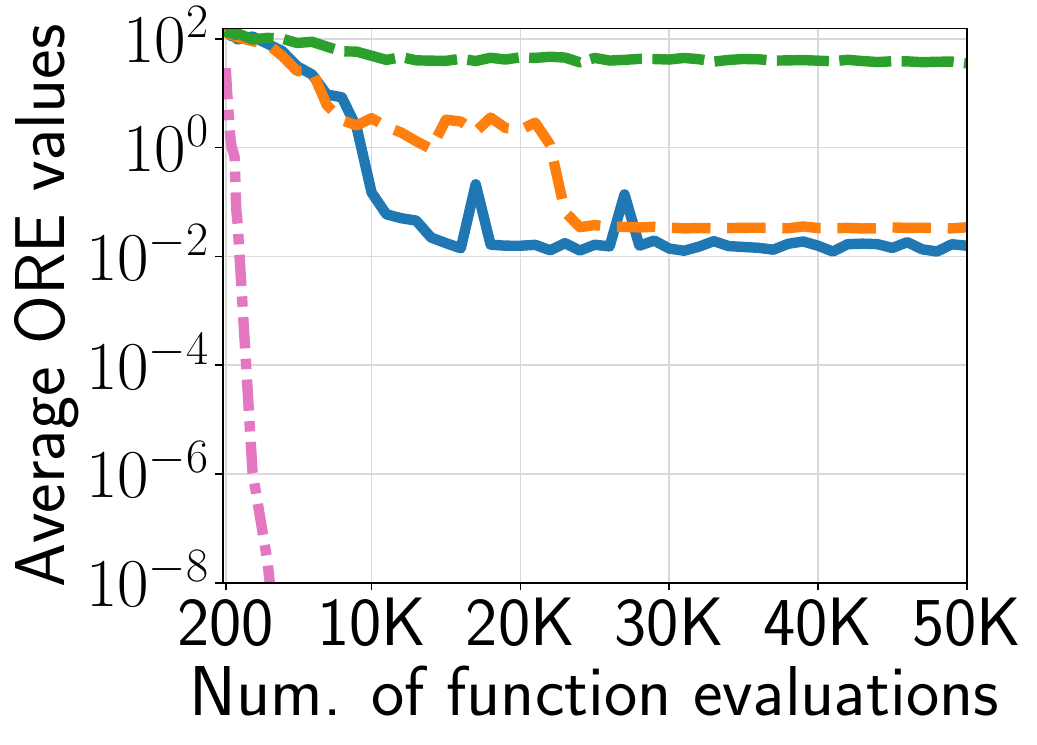}}
   \subfloat[ORE ($m=6$)]{\includegraphics[width=0.32\textwidth]{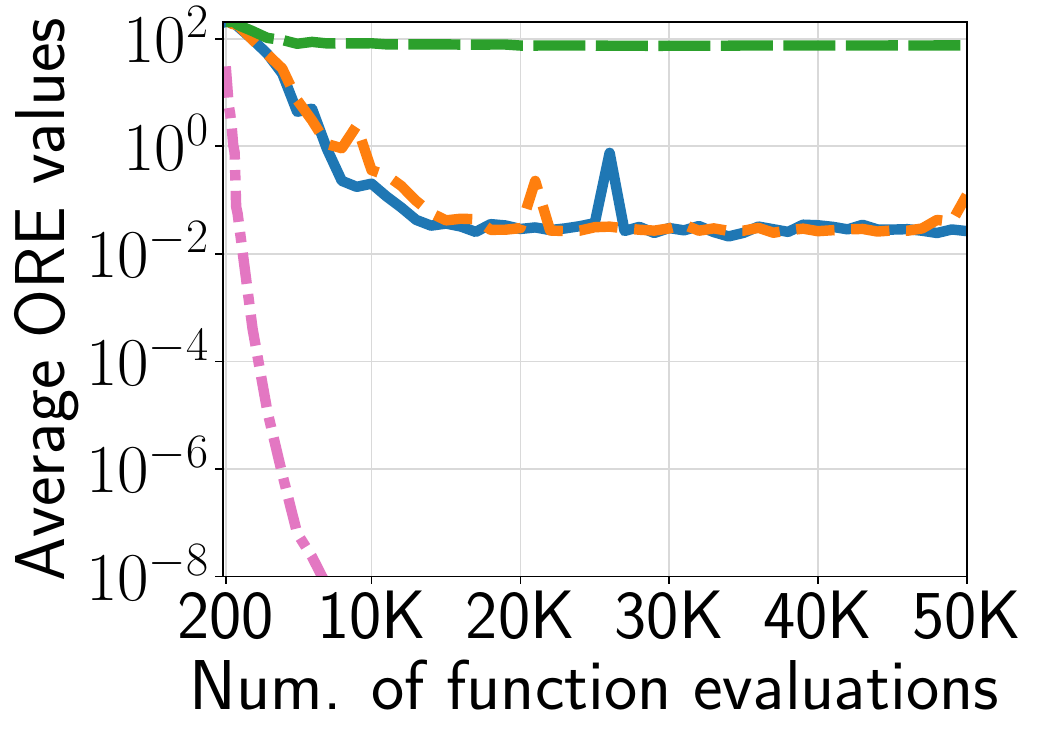}}
\\
\caption{Average $e^{\mathrm{ideal}}$, $e^{\mathrm{nadir}}$, and ORE values of the three normalization methods in R-NSGA-II on DTLZ1.}
\label{supfig:3error_RNSGA2_DTLZ1}
\end{figure*}

\begin{figure*}[t]
\centering
  \subfloat{\includegraphics[width=0.7\textwidth]{./figs/legend/legend_3.pdf}}
\vspace{-3.9mm}
   \\
   \subfloat[$e^{\mathrm{ideal}}$ ($m=2$)]{\includegraphics[width=0.32\textwidth]{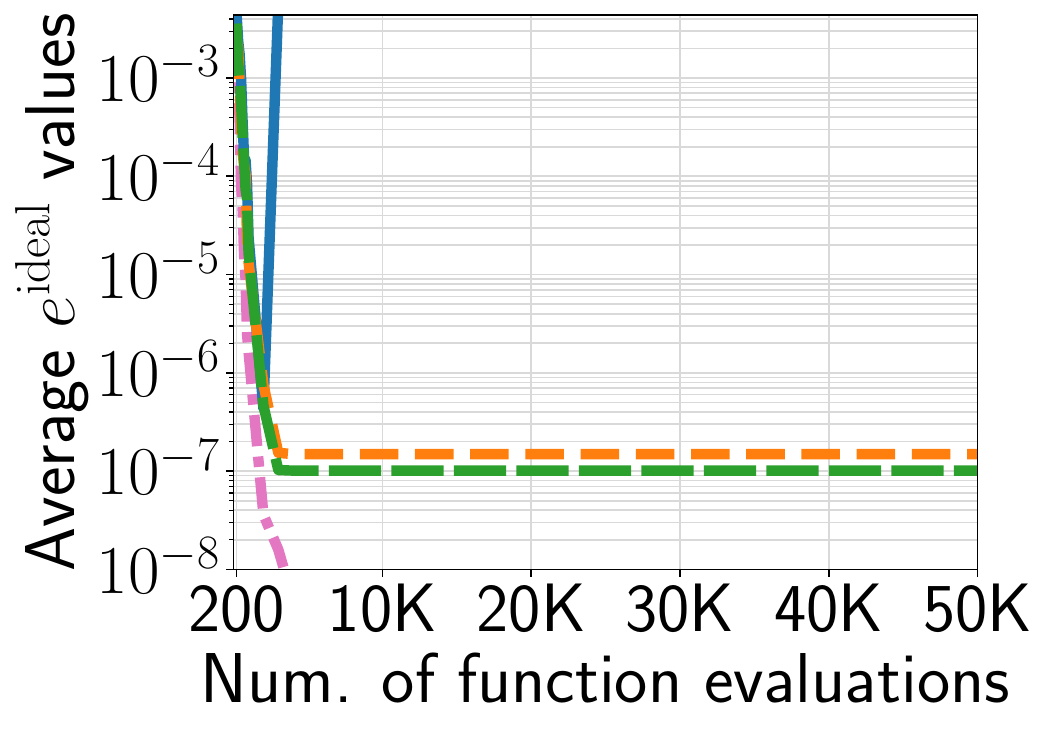}}
   \subfloat[$e^{\mathrm{ideal}}$ ($m=4$)]{\includegraphics[width=0.32\textwidth]{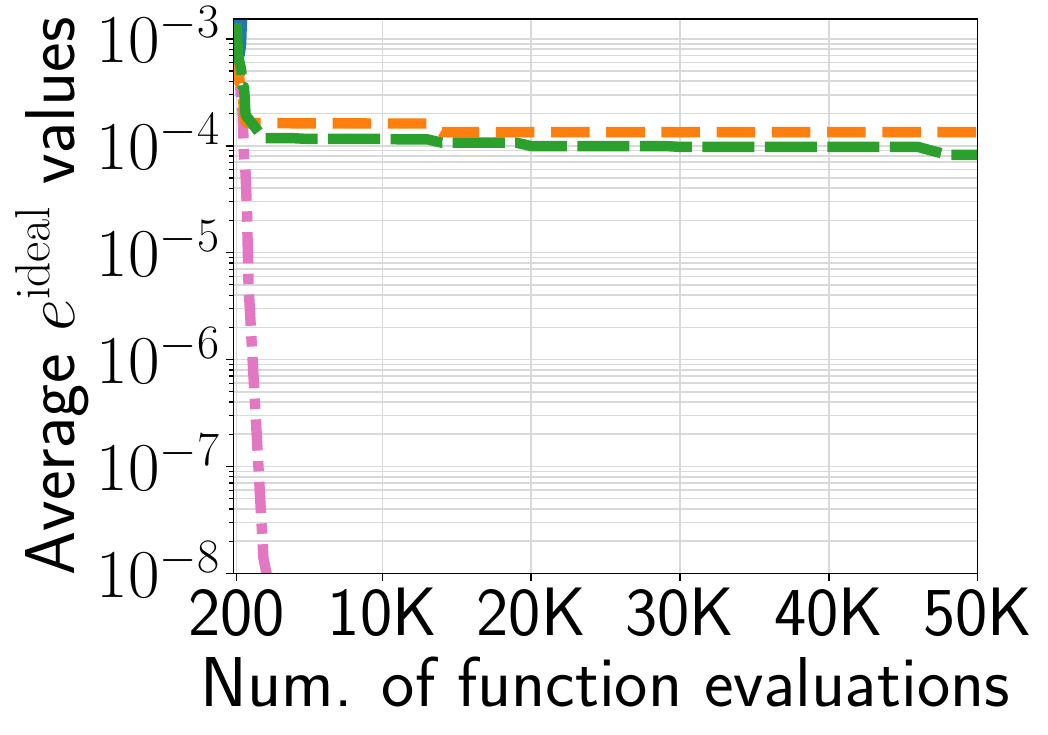}}
   \subfloat[$e^{\mathrm{ideal}}$ ($m=6$)]{\includegraphics[width=0.32\textwidth]{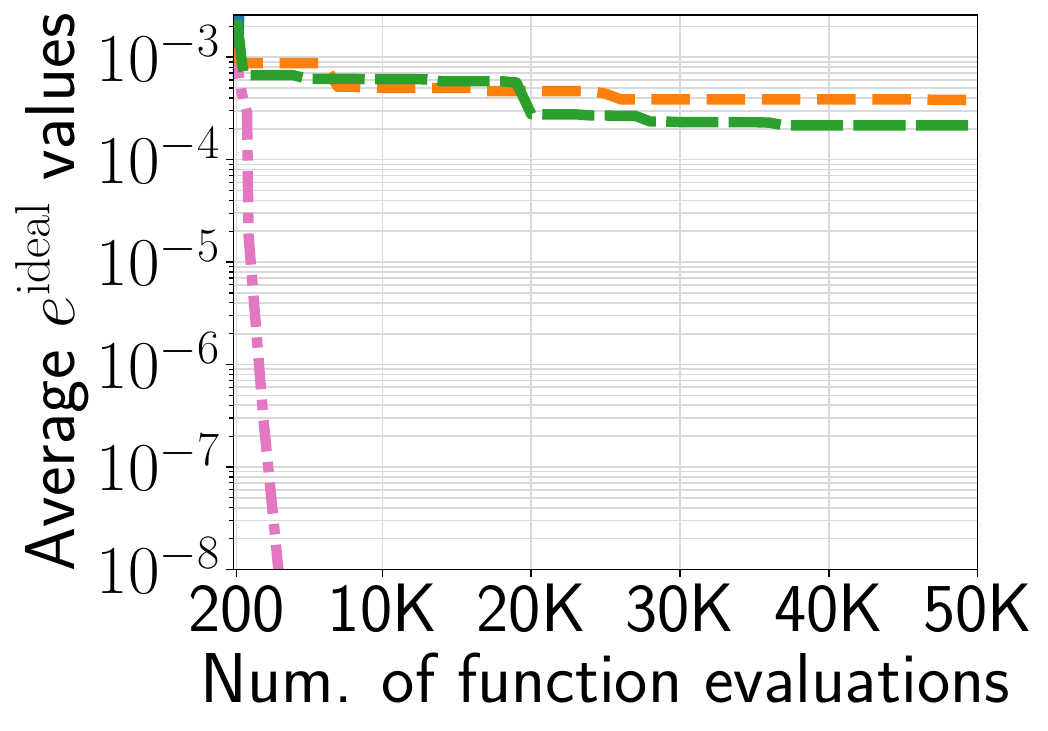}}
\\
   \subfloat[$e^{\mathrm{nadir}}$ ($m=2$)]{\includegraphics[width=0.32\textwidth]{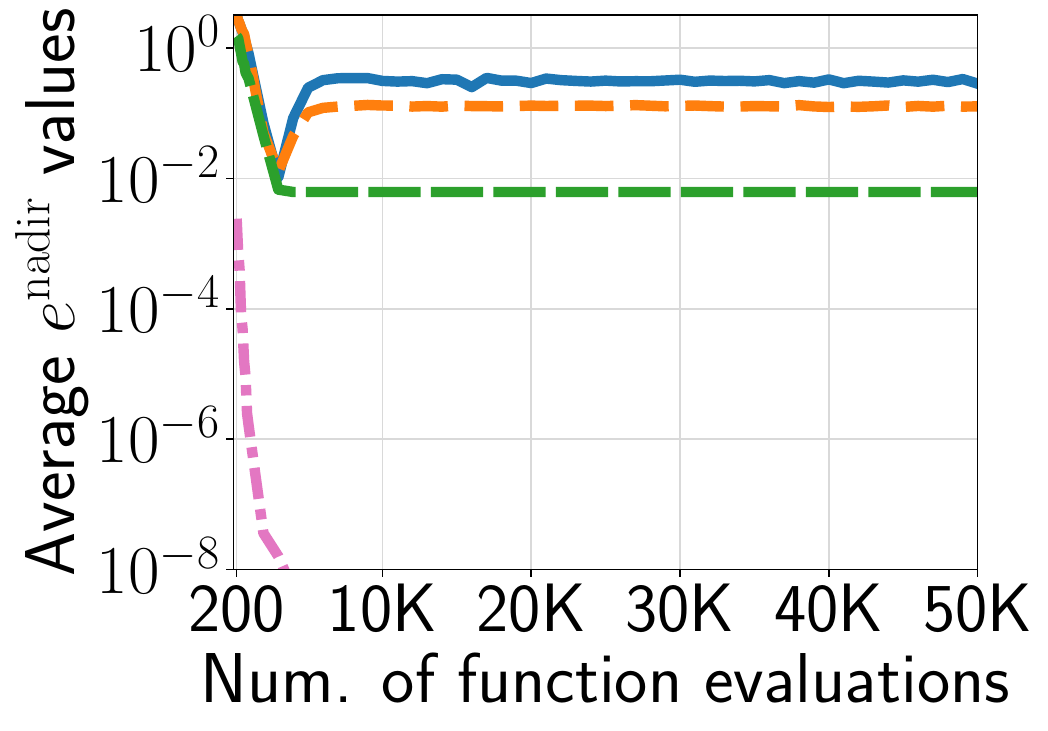}}
   \subfloat[$e^{\mathrm{nadir}}$ ($m=4$)]{\includegraphics[width=0.32\textwidth]{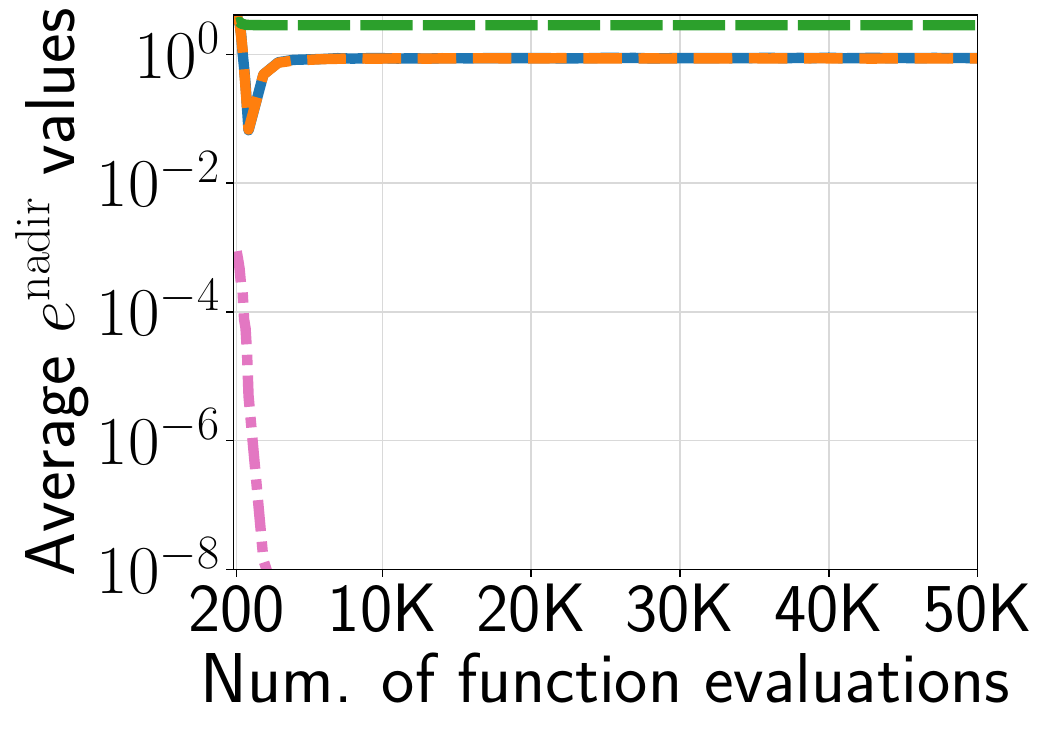}}
   \subfloat[$e^{\mathrm{nadir}}$ ($m=6$)]{\includegraphics[width=0.32\textwidth]{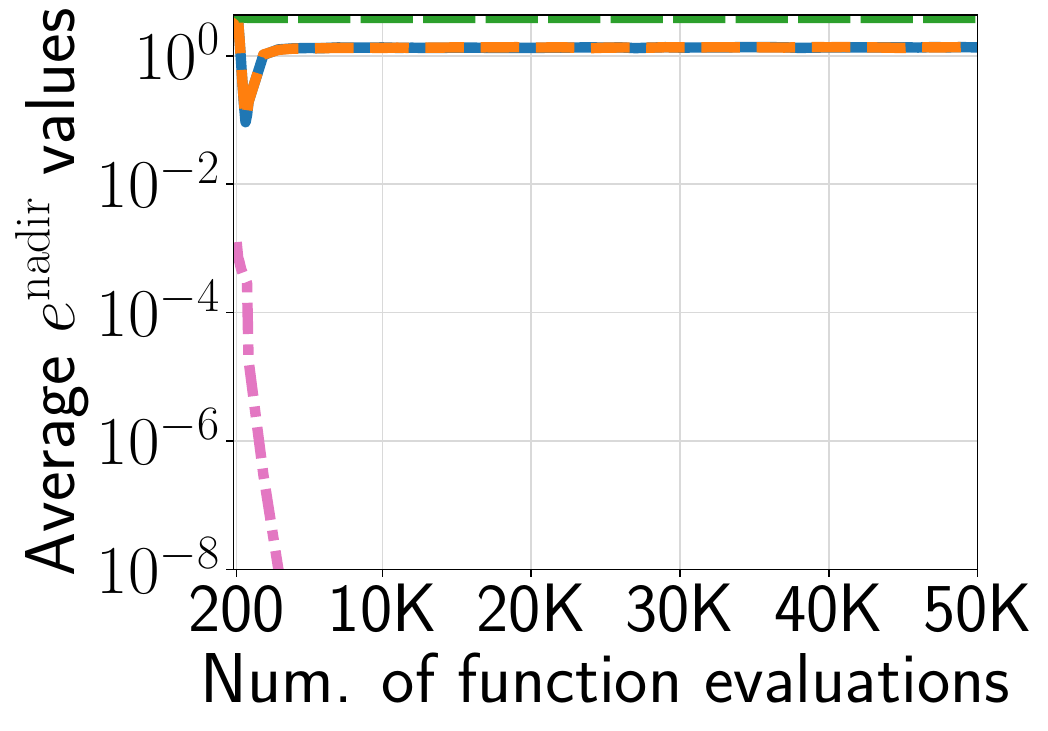}}
\\
   \subfloat[ORE ($m=2$)]{\includegraphics[width=0.32\textwidth]{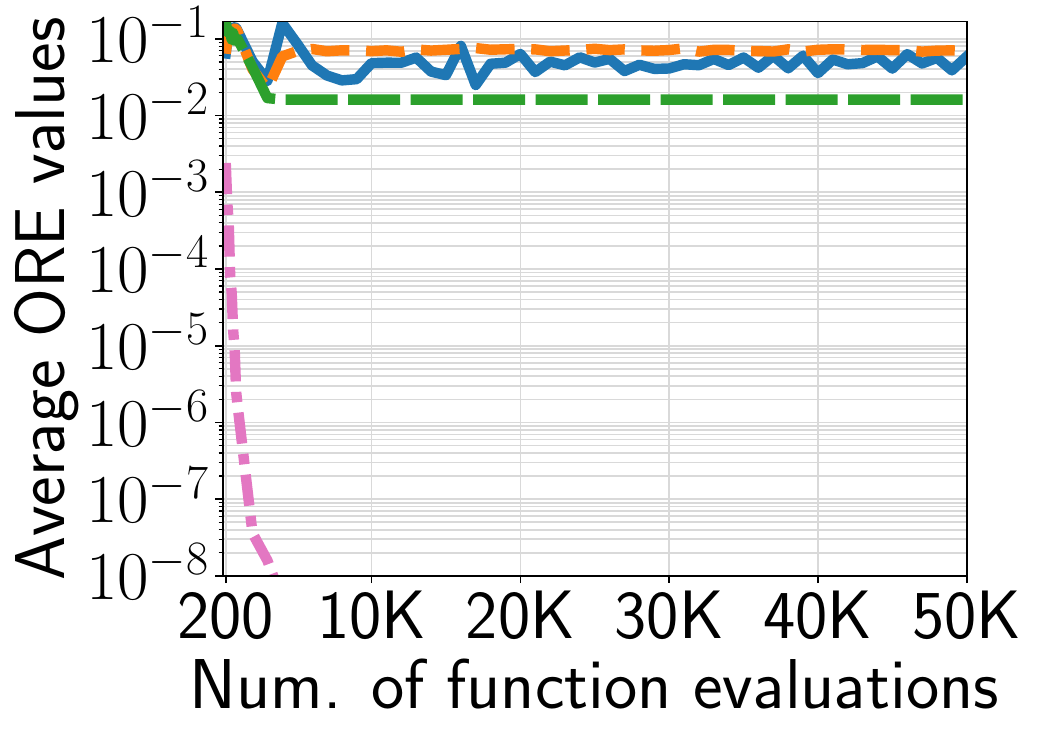}}
   \subfloat[ORE ($m=4$)]{\includegraphics[width=0.32\textwidth]{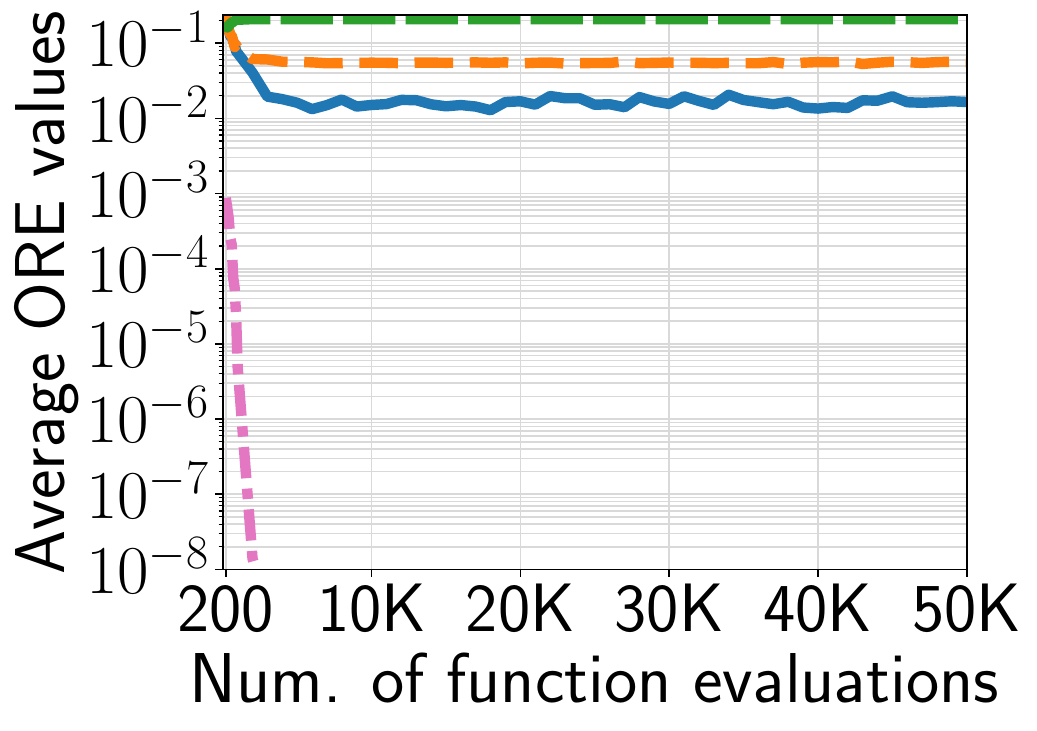}}
   \subfloat[ORE ($m=6$)]{\includegraphics[width=0.32\textwidth]{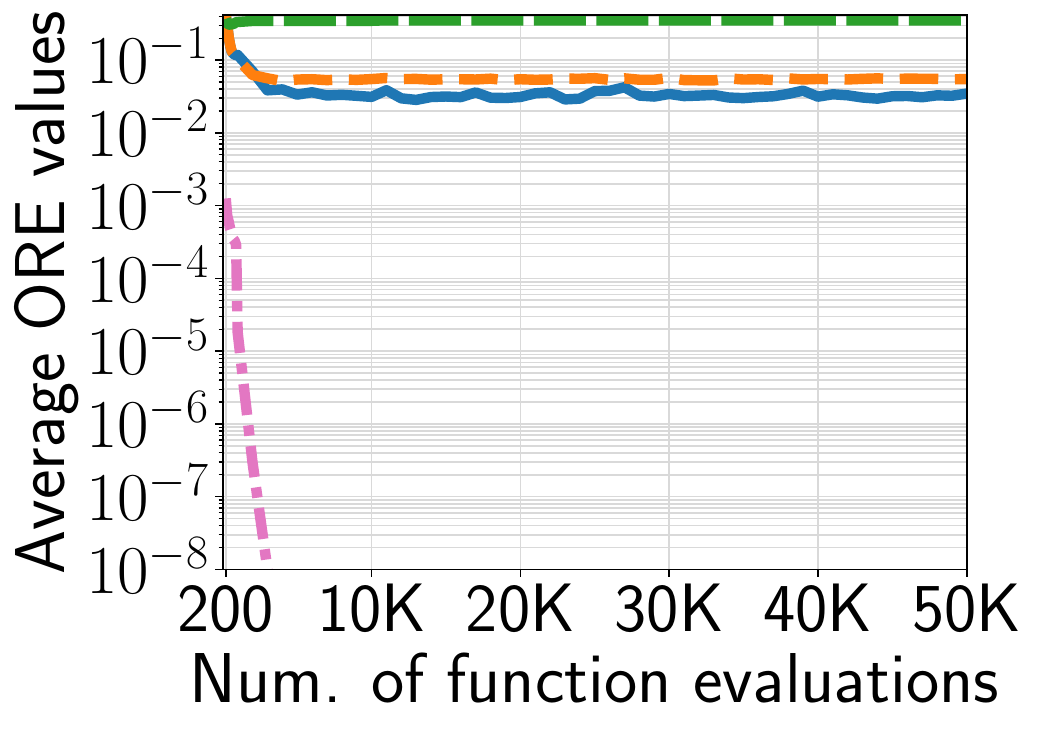}}
\\
\caption{Average $e^{\mathrm{ideal}}$, $e^{\mathrm{nadir}}$, and ORE values of the three normalization methods in R-NSGA-II on DTLZ2.}
\label{supfig:3error_RNSGA2_DTLZ2}
\end{figure*}

\begin{figure*}[t]
\centering
  \subfloat{\includegraphics[width=0.7\textwidth]{./figs/legend/legend_3.pdf}}
\vspace{-3.9mm}
   \\
   \subfloat[$e^{\mathrm{ideal}}$ ($m=2$)]{\includegraphics[width=0.32\textwidth]{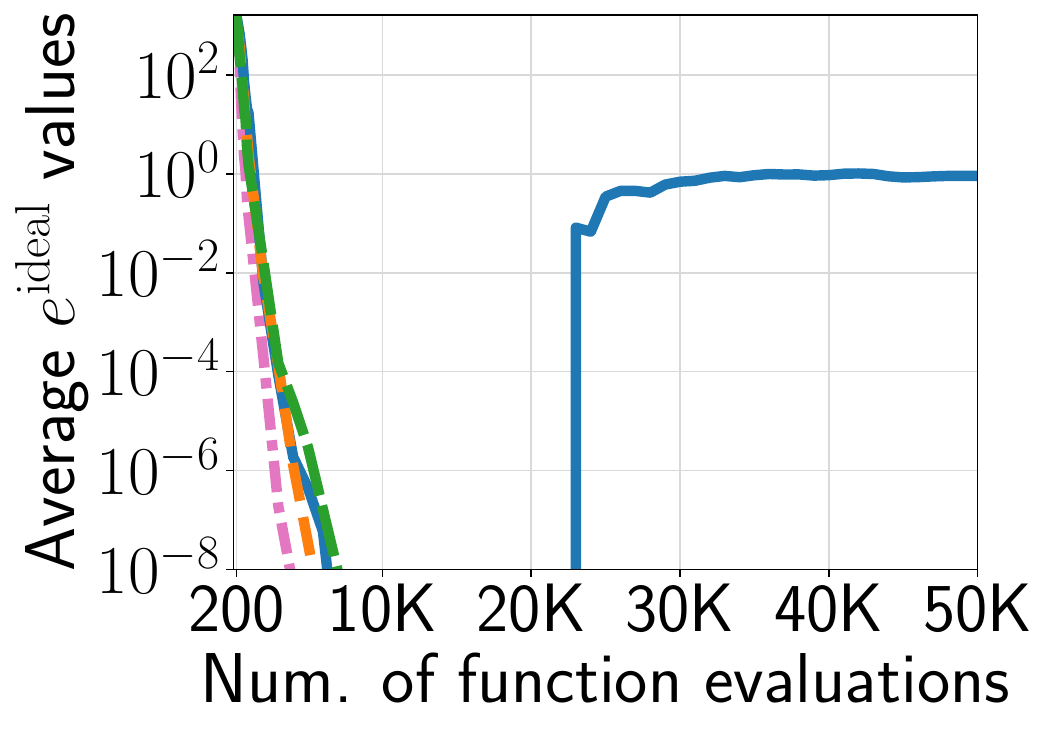}}
   \subfloat[$e^{\mathrm{ideal}}$ ($m=4$)]{\includegraphics[width=0.32\textwidth]{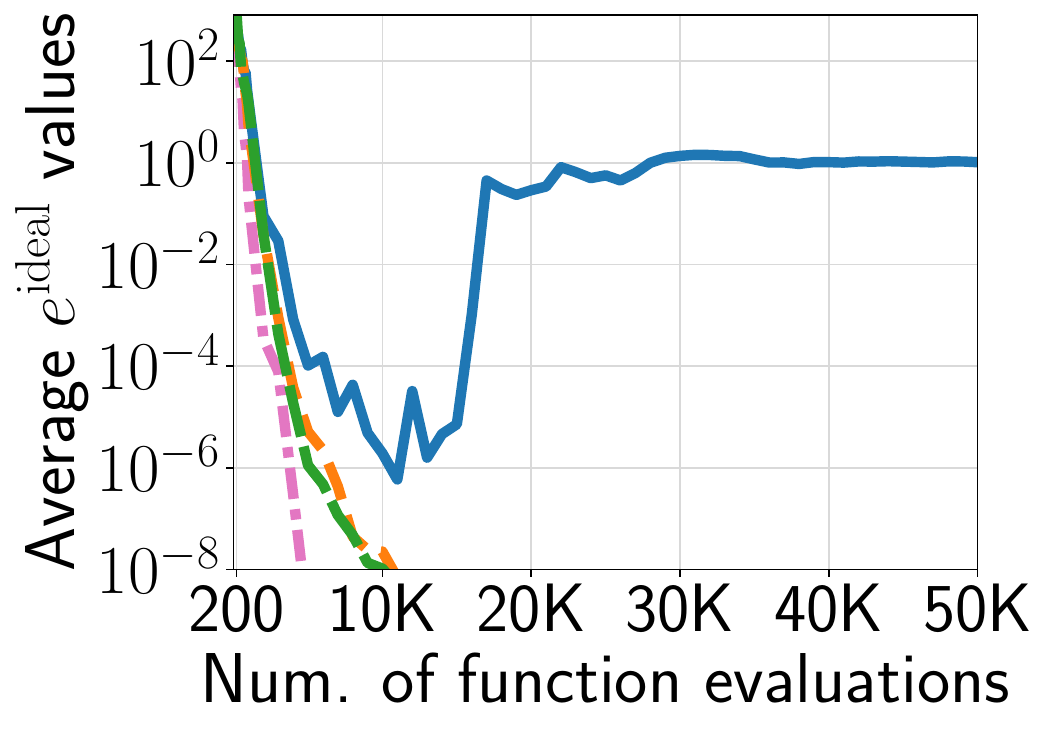}}
   \subfloat[$e^{\mathrm{ideal}}$ ($m=6$)]{\includegraphics[width=0.32\textwidth]{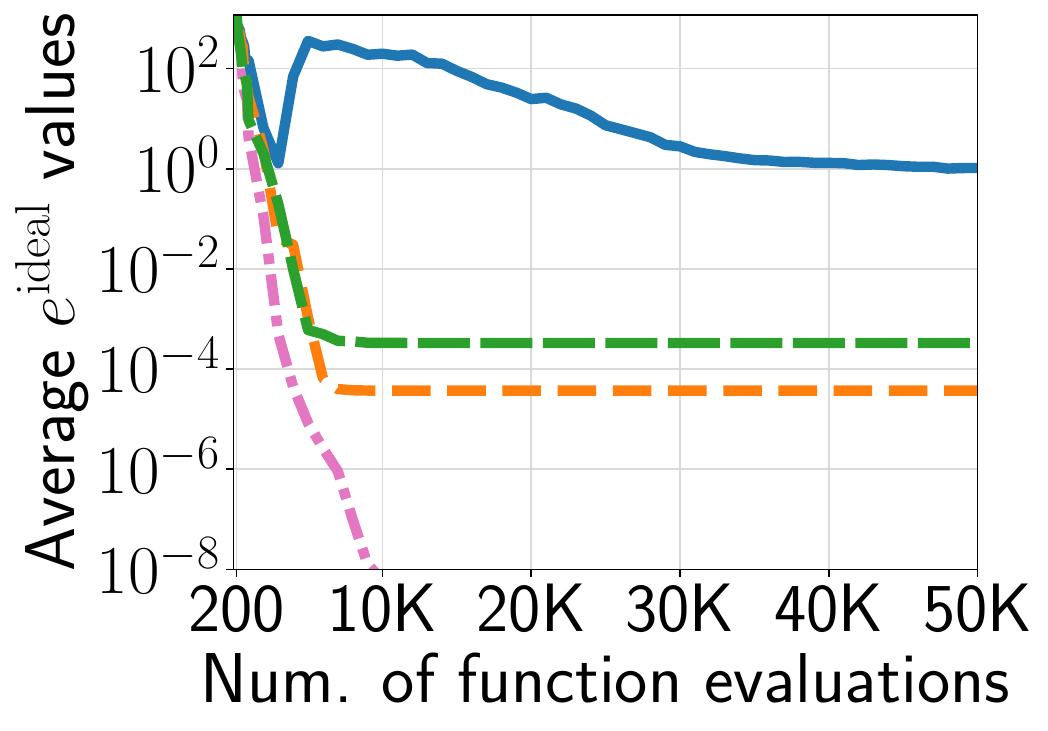}}
\\
   \subfloat[$e^{\mathrm{nadir}}$ ($m=2$)]{\includegraphics[width=0.32\textwidth]{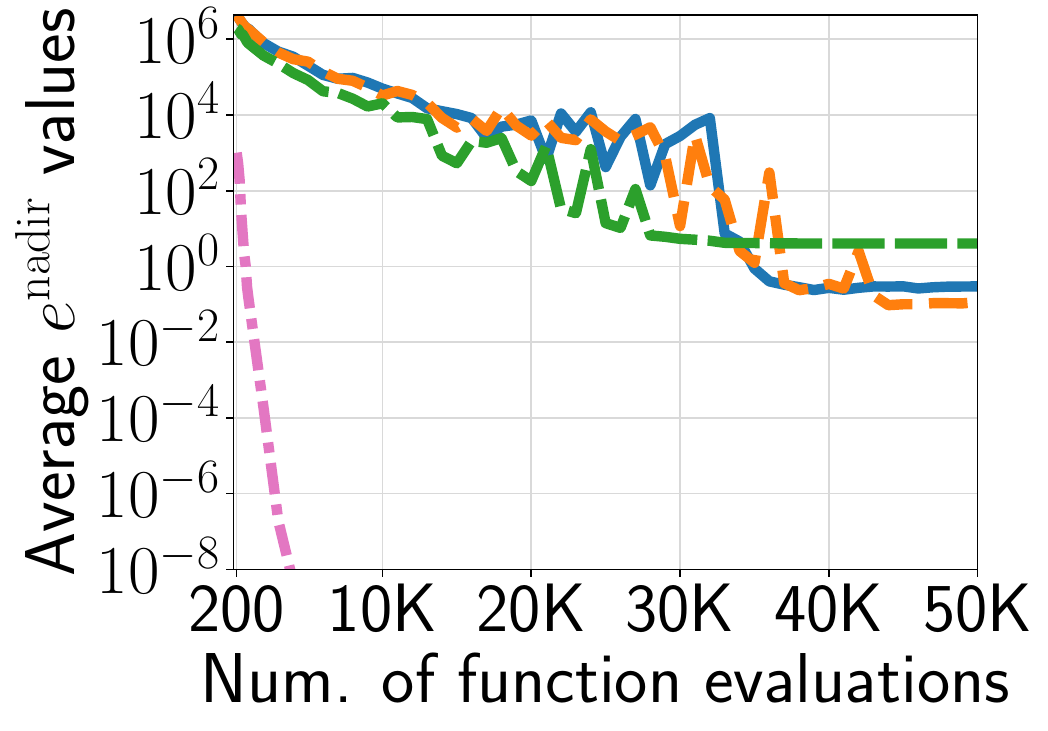}}
   \subfloat[$e^{\mathrm{nadir}}$ ($m=4$)]{\includegraphics[width=0.32\textwidth]{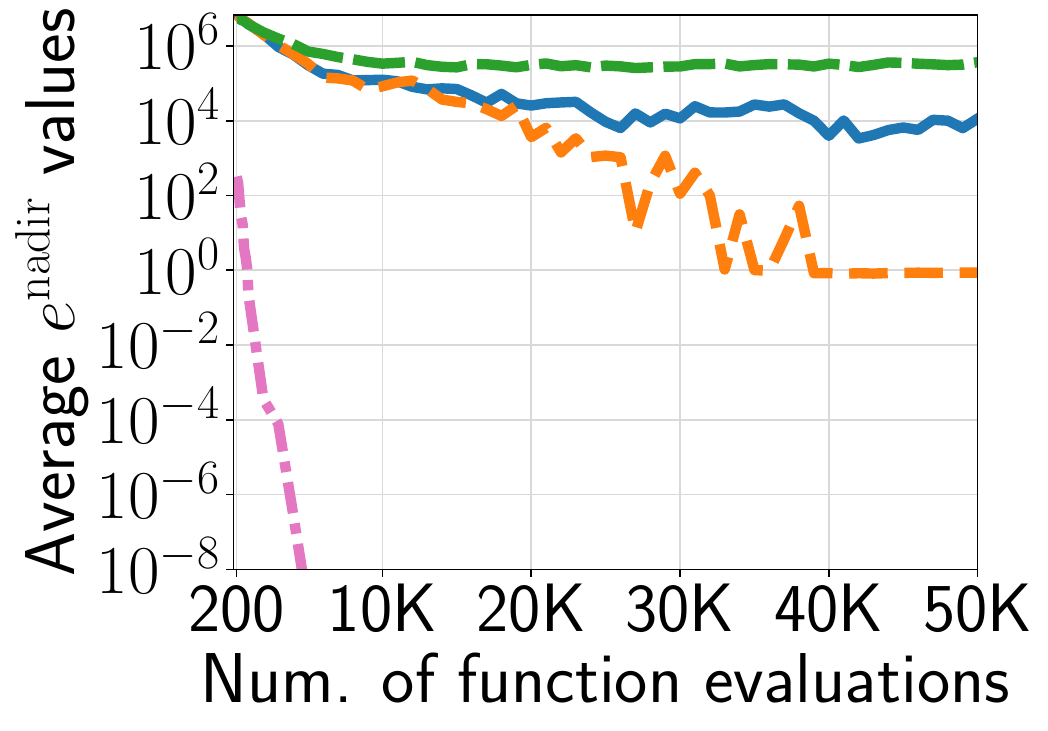}}
   \subfloat[$e^{\mathrm{nadir}}$ ($m=6$)]{\includegraphics[width=0.32\textwidth]{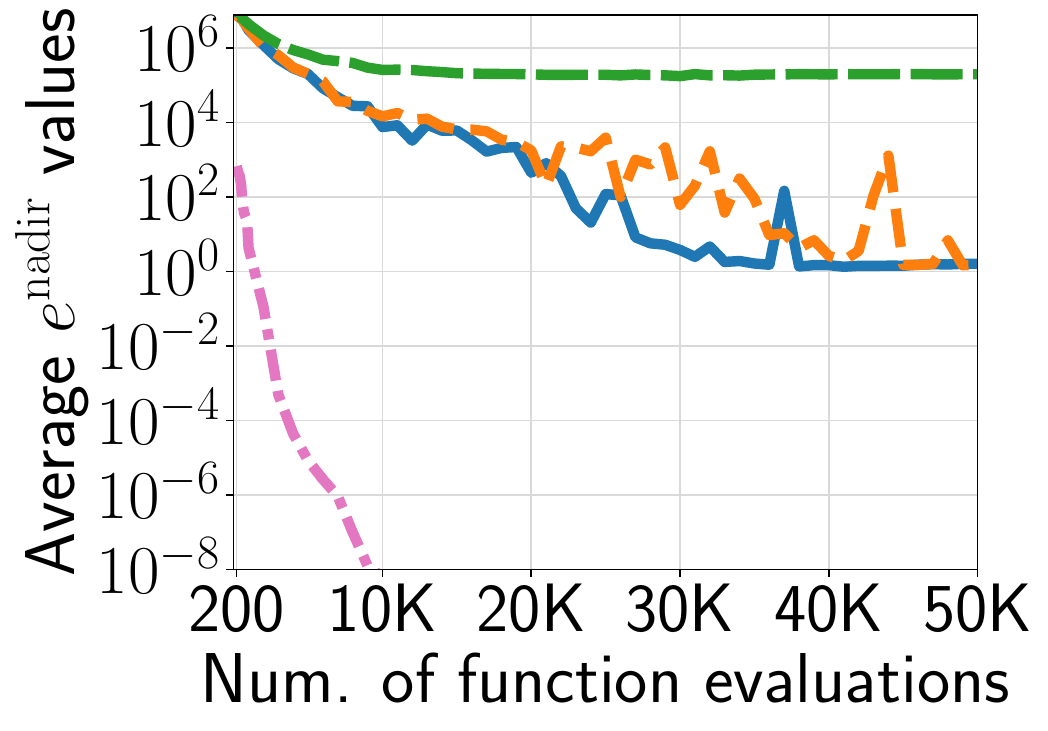}}
\\
   \subfloat[ORE ($m=2$)]{\includegraphics[width=0.32\textwidth]{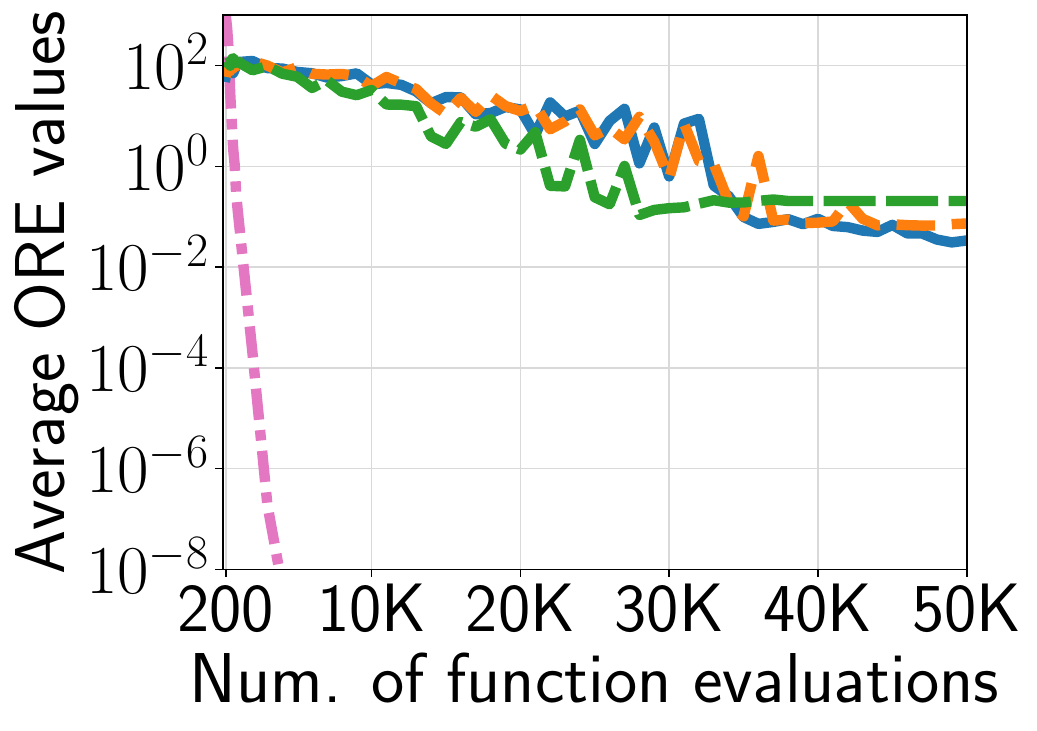}}
   \subfloat[ORE ($m=4$)]{\includegraphics[width=0.32\textwidth]{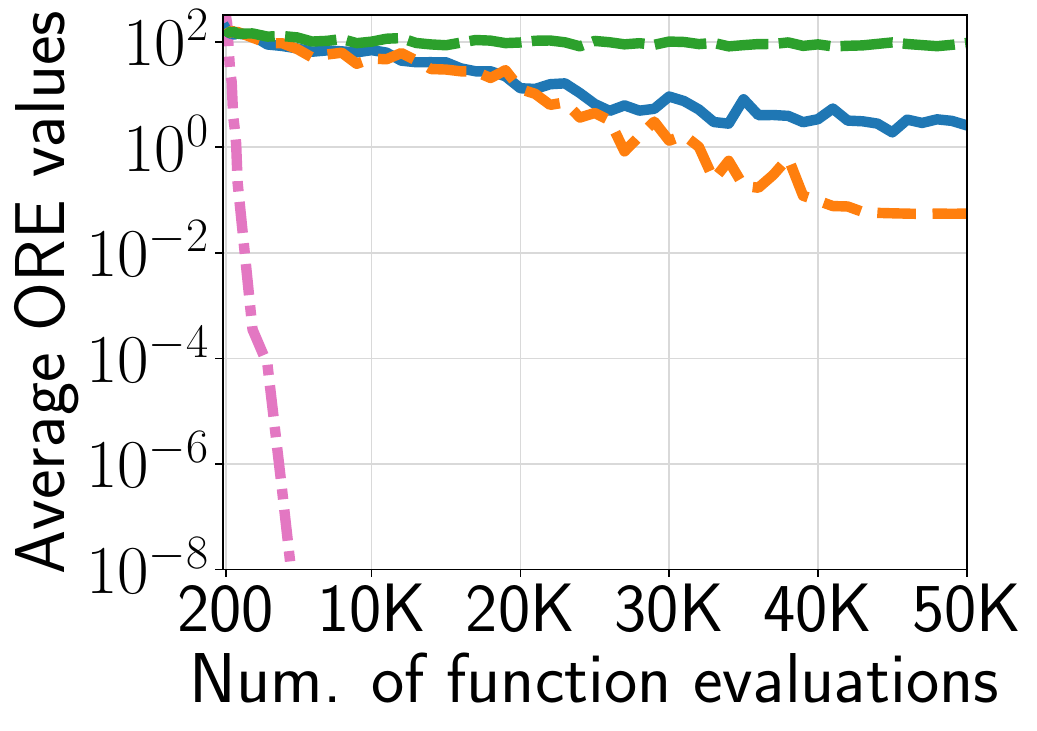}}
   \subfloat[ORE ($m=6$)]{\includegraphics[width=0.32\textwidth]{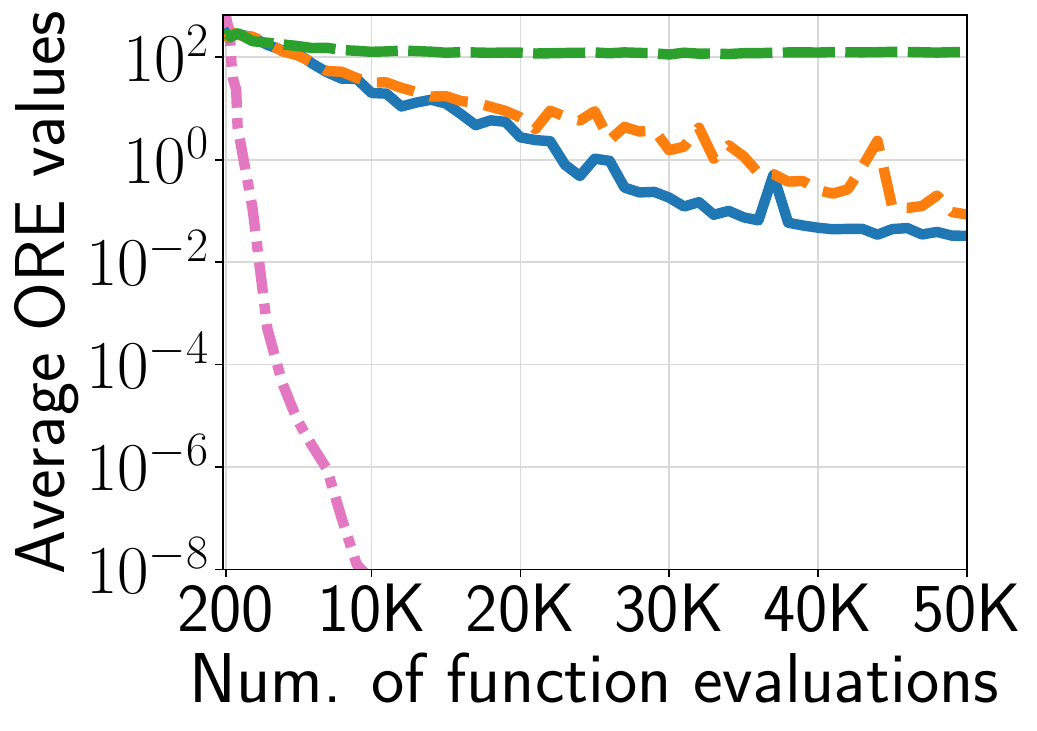}}
\\
\caption{Average $e^{\mathrm{ideal}}$, $e^{\mathrm{nadir}}$, and ORE values of the three normalization methods in R-NSGA-II on DTLZ3.}
\label{supfig:3error_RNSGA2_DTLZ3}
\end{figure*}

\begin{figure*}[t]
\centering
  \subfloat{\includegraphics[width=0.7\textwidth]{./figs/legend/legend_3.pdf}}
\vspace{-3.9mm}
   \\
   \subfloat[$e^{\mathrm{ideal}}$ ($m=2$)]{\includegraphics[width=0.32\textwidth]{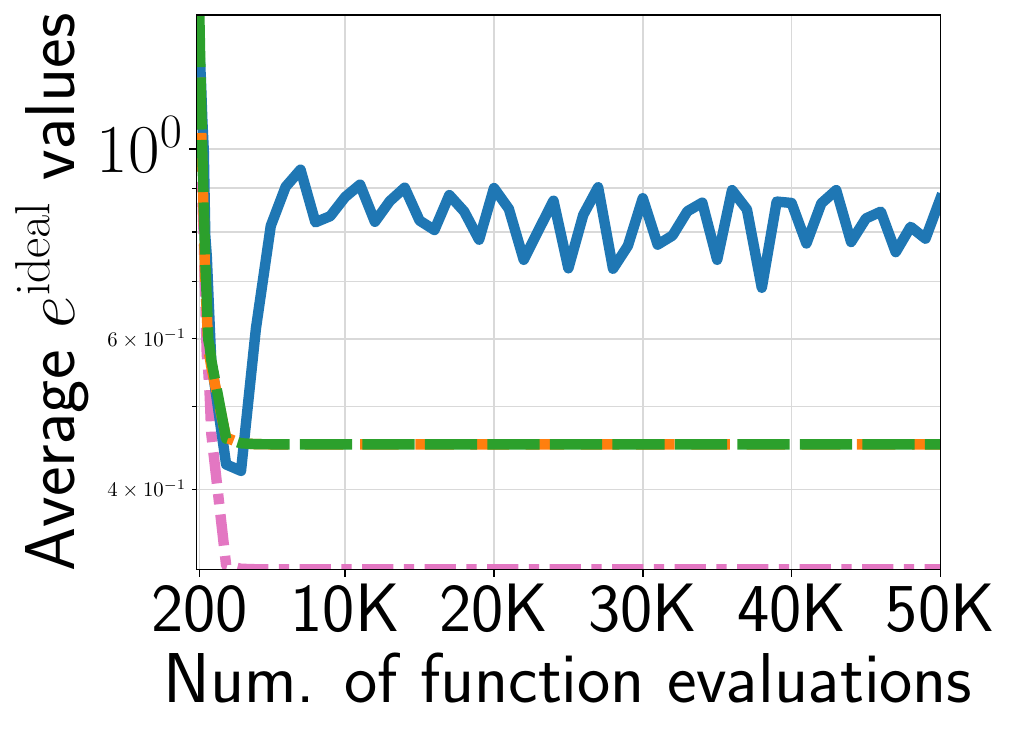}}
   \subfloat[$e^{\mathrm{ideal}}$ ($m=4$)]{\includegraphics[width=0.32\textwidth]{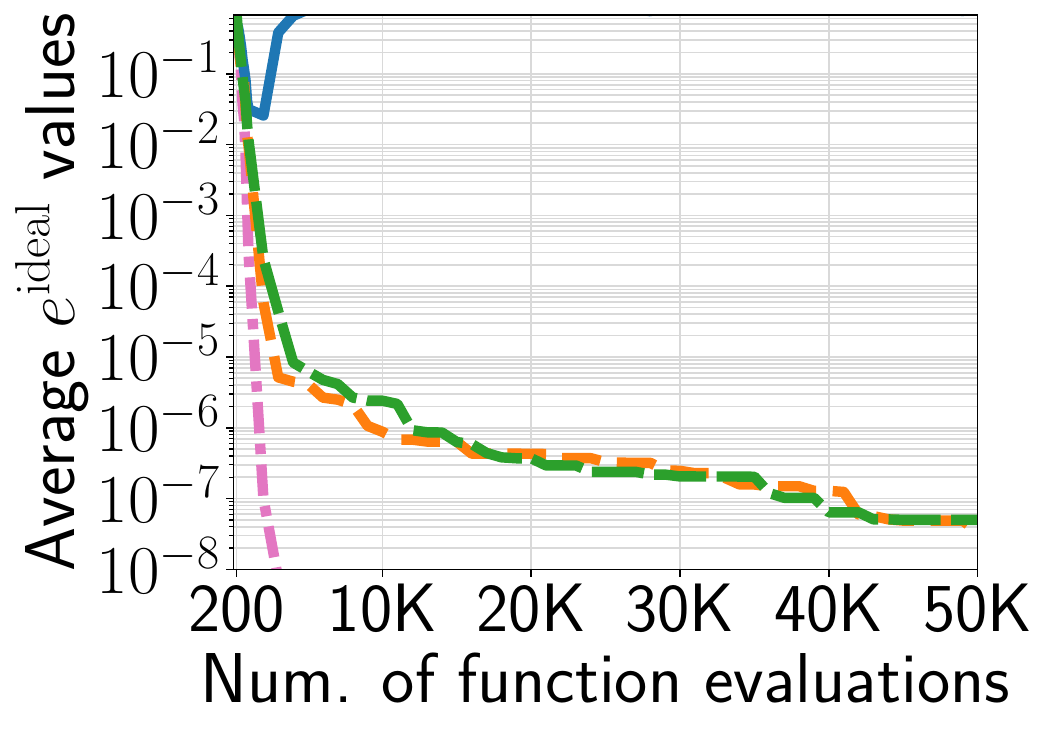}}
   \subfloat[$e^{\mathrm{ideal}}$ ($m=6$)]{\includegraphics[width=0.32\textwidth]{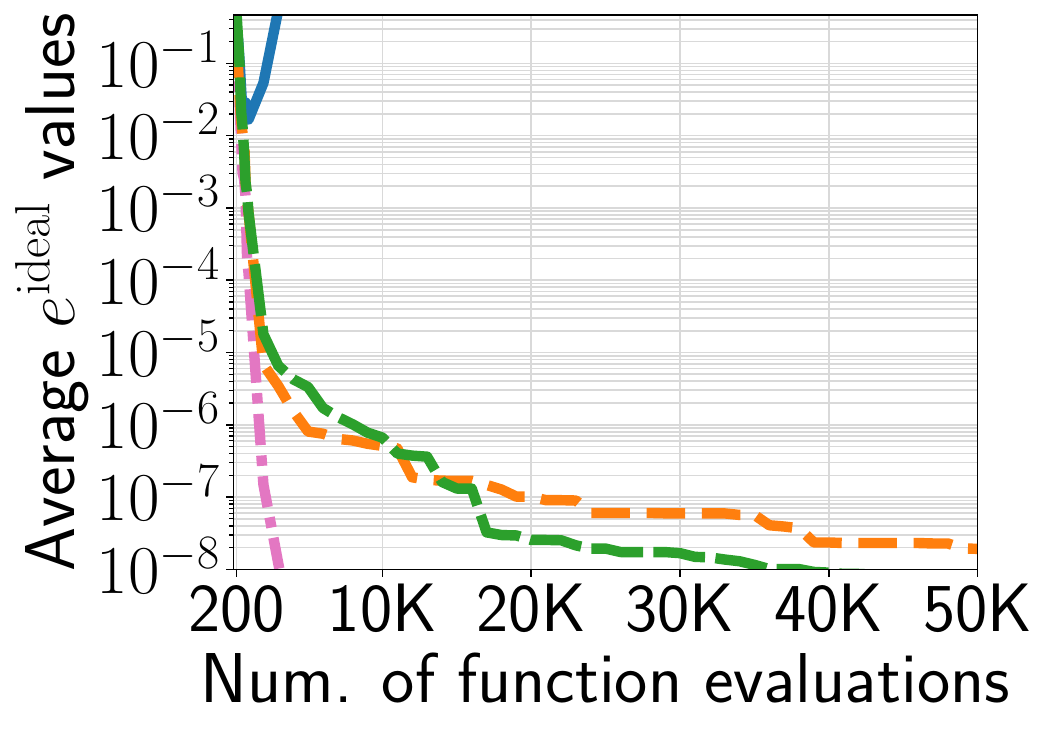}}
\\
   \subfloat[$e^{\mathrm{nadir}}$ ($m=2$)]{\includegraphics[width=0.32\textwidth]{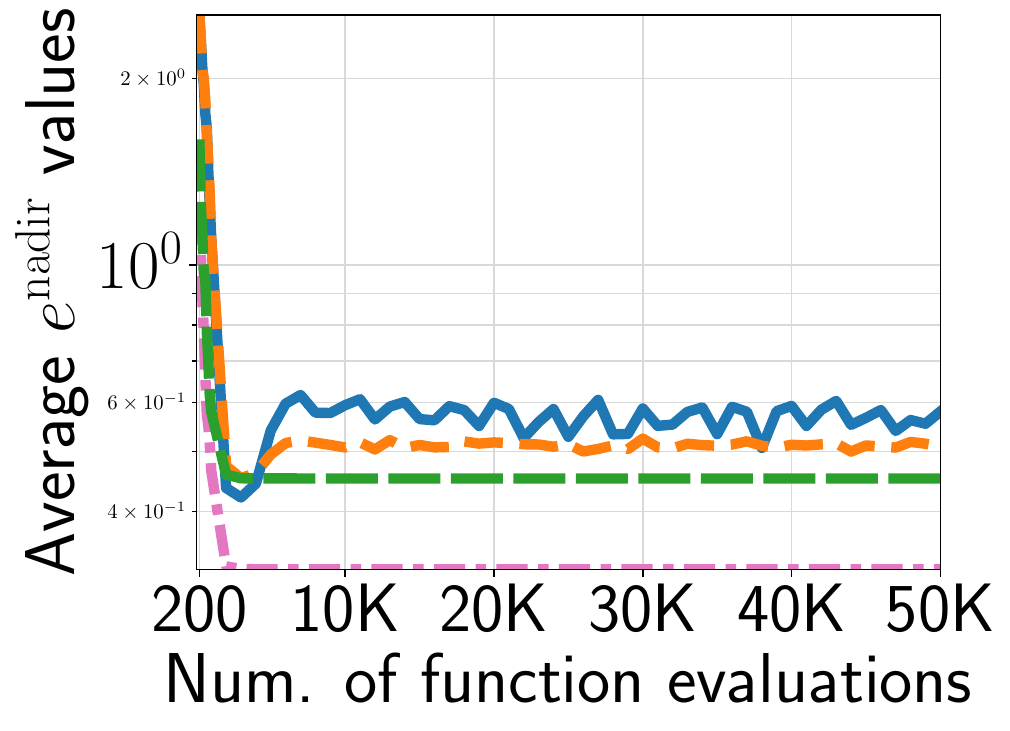}}
   \subfloat[$e^{\mathrm{nadir}}$ ($m=4$)]{\includegraphics[width=0.32\textwidth]{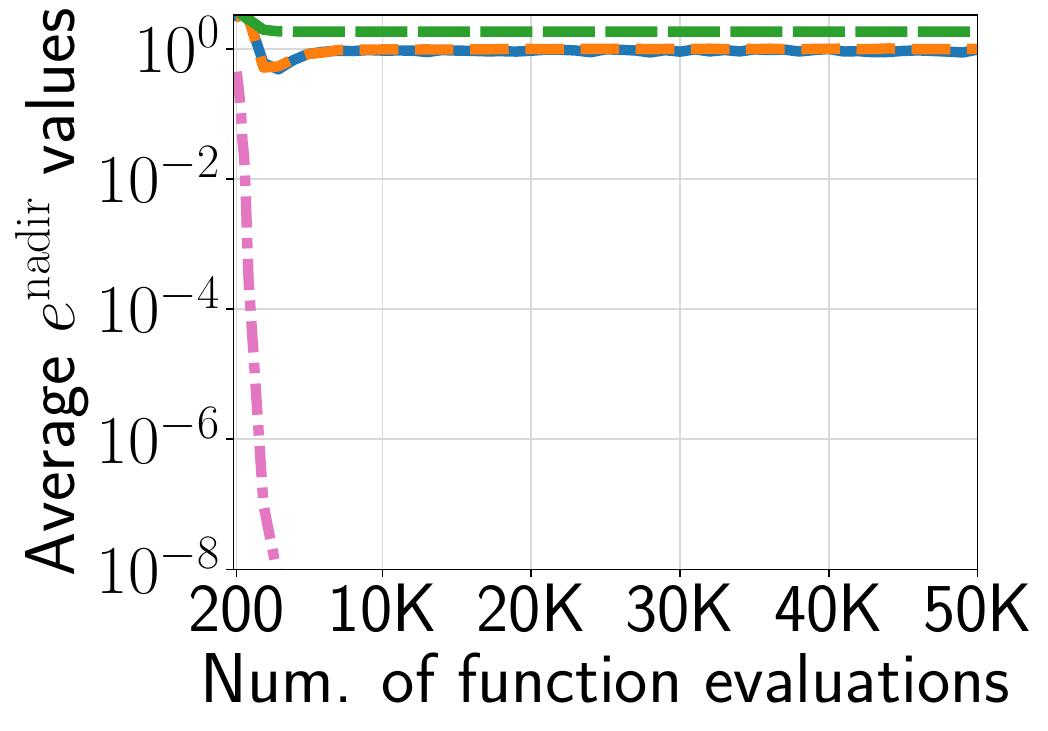}}
   \subfloat[$e^{\mathrm{nadir}}$ ($m=6$)]{\includegraphics[width=0.32\textwidth]{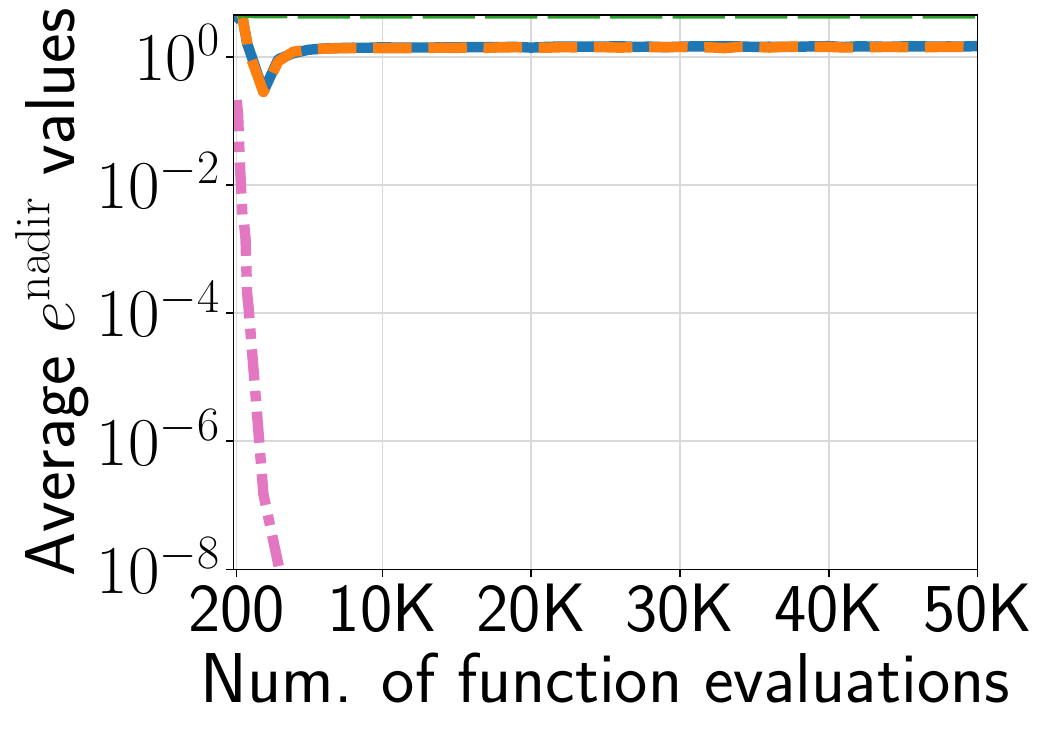}}
\\
   \subfloat[ORE ($m=2$)]{\includegraphics[width=0.32\textwidth]{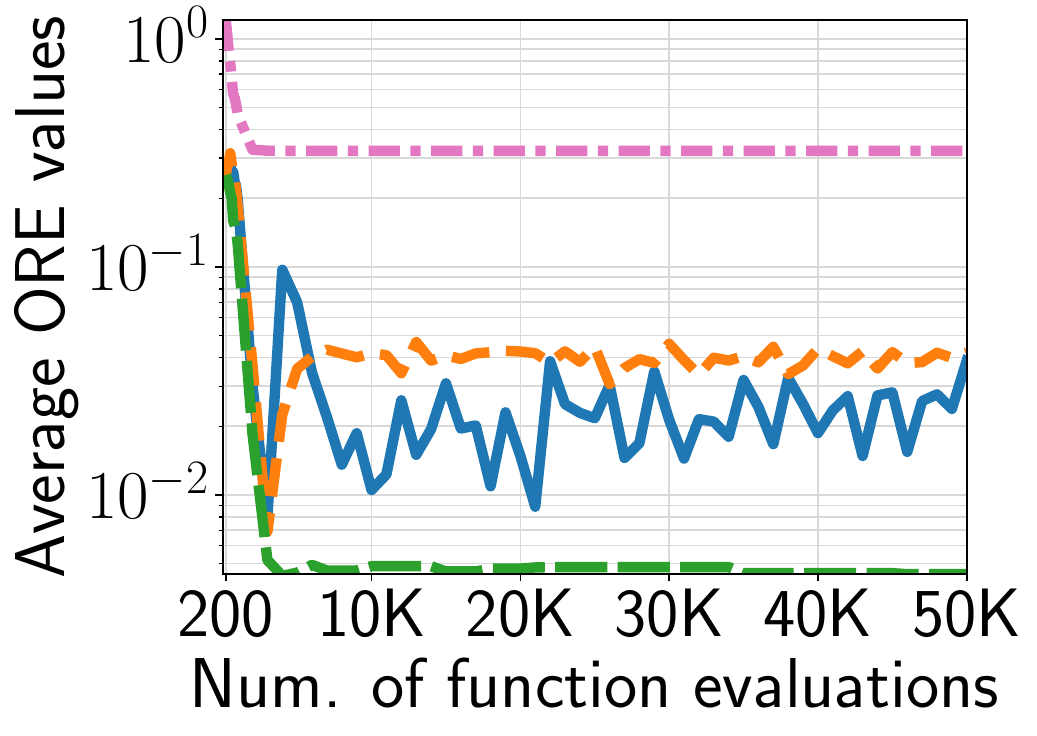}}
   \subfloat[ORE ($m=4$)]{\includegraphics[width=0.32\textwidth]{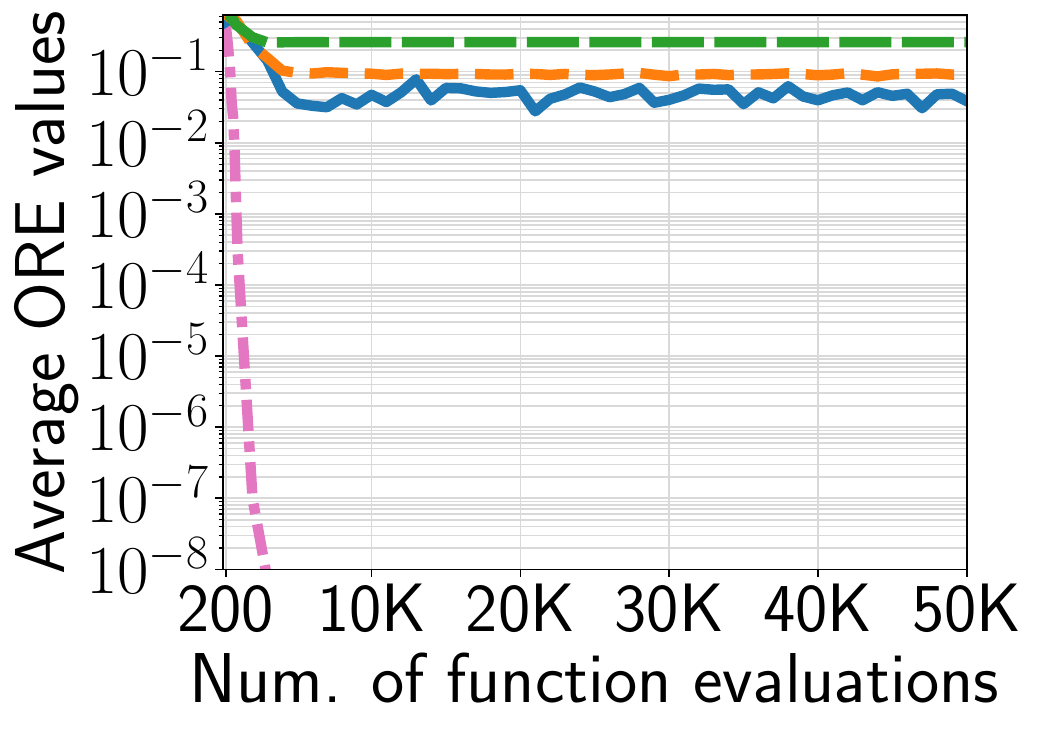}}
   \subfloat[ORE ($m=6$)]{\includegraphics[width=0.32\textwidth]{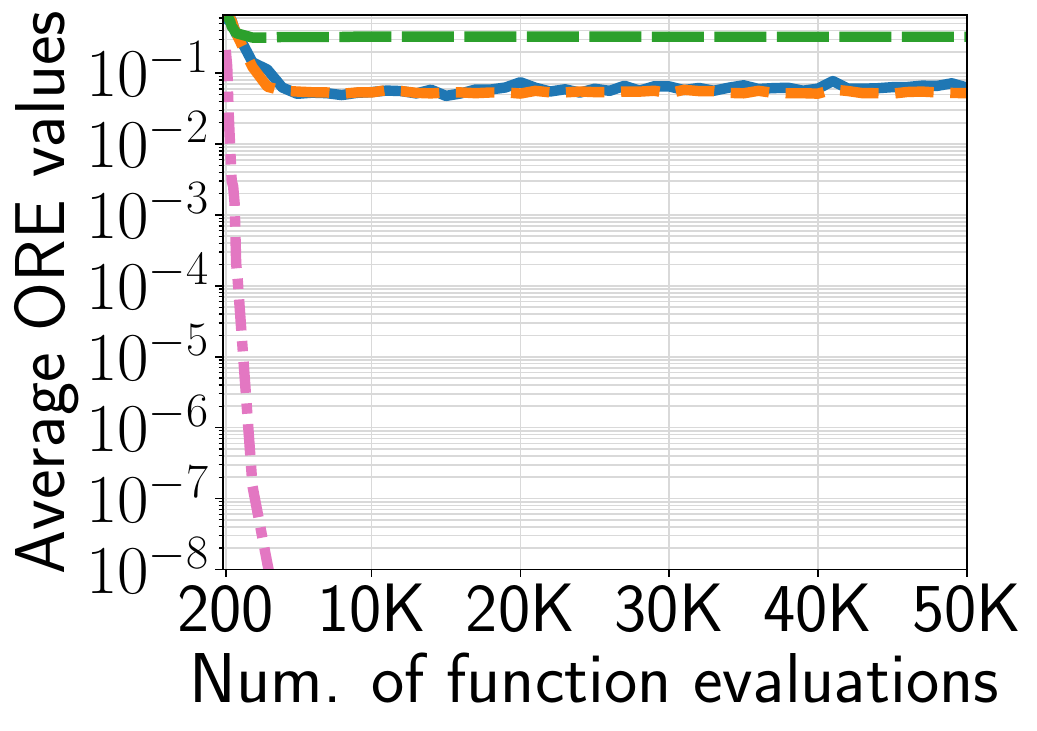}}
\\
\caption{Average $e^{\mathrm{ideal}}$, $e^{\mathrm{nadir}}$, and ORE values of the three normalization methods in R-NSGA-II on DTLZ4.}
\label{supfig:3error_RNSGA2_DTLZ4}
\end{figure*}

\begin{figure*}[t]
\centering
  \subfloat{\includegraphics[width=0.7\textwidth]{./figs/legend/legend_3.pdf}}
\vspace{-3.9mm}
   \\
   \subfloat[$e^{\mathrm{ideal}}$ ($m=2$)]{\includegraphics[width=0.32\textwidth]{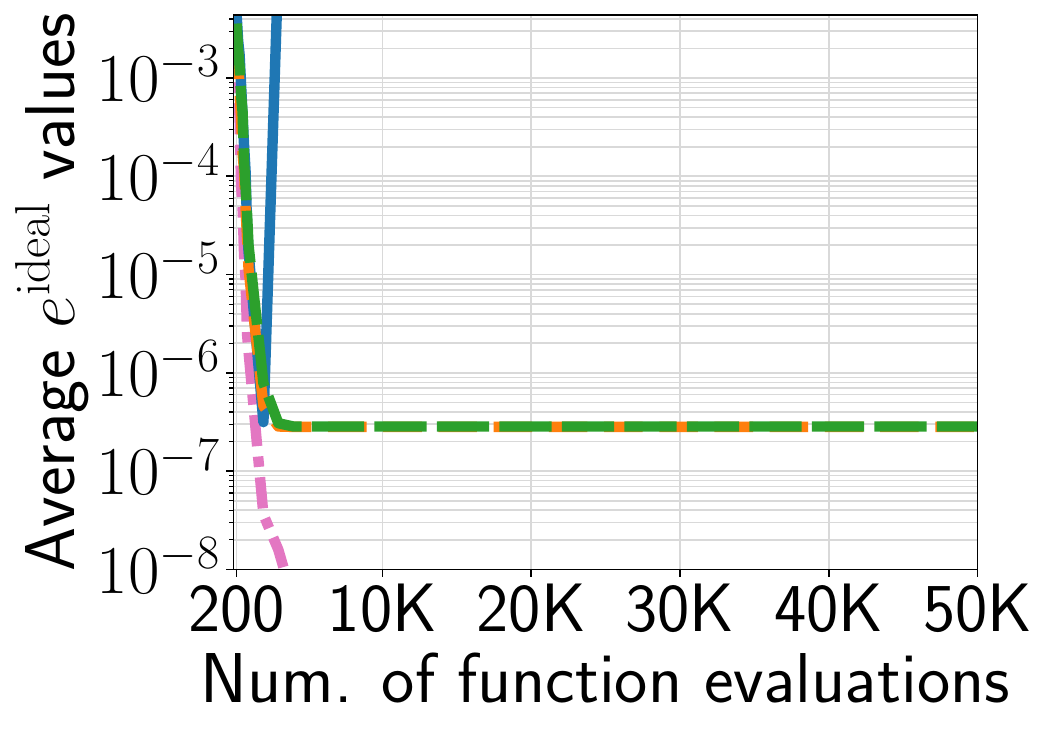}}
   \subfloat[$e^{\mathrm{ideal}}$ ($m=4$)]{\includegraphics[width=0.32\textwidth]{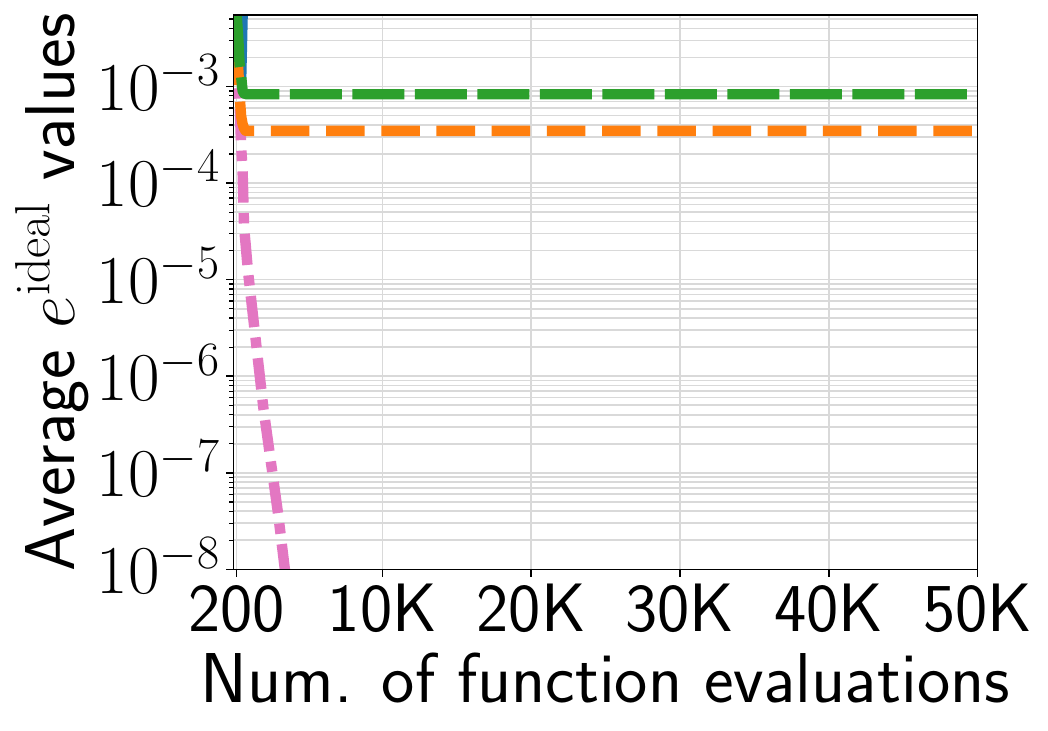}}
   \subfloat[$e^{\mathrm{ideal}}$ ($m=6$)]{\includegraphics[width=0.32\textwidth]{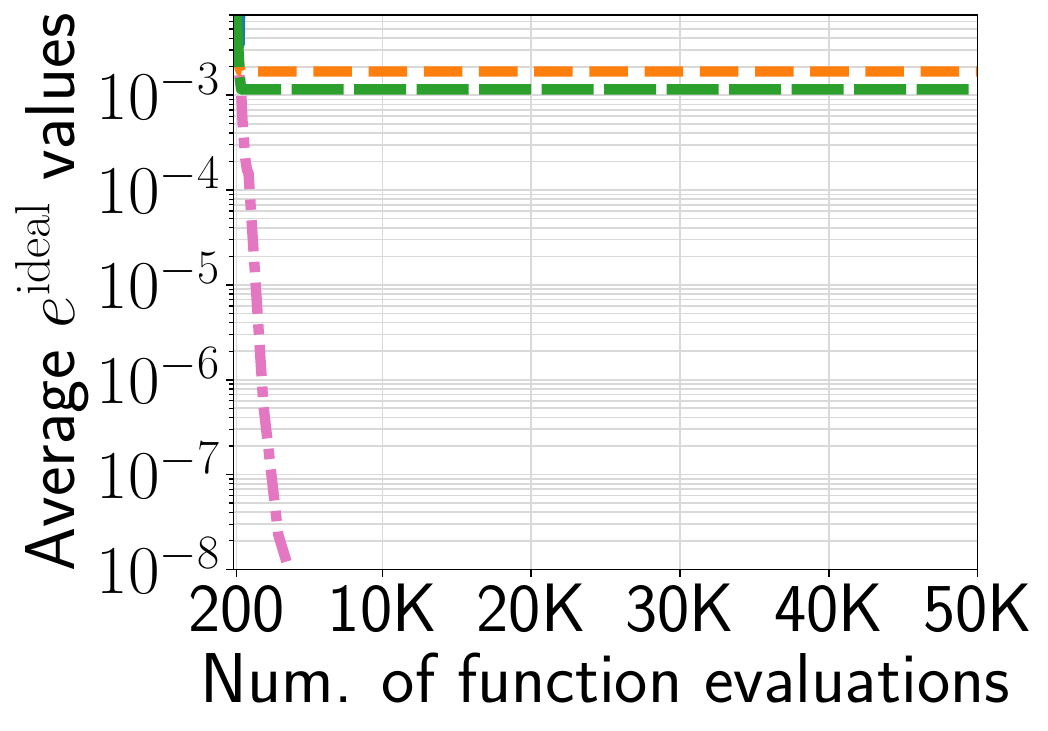}}
\\
   \subfloat[$e^{\mathrm{nadir}}$ ($m=2$)]{\includegraphics[width=0.32\textwidth]{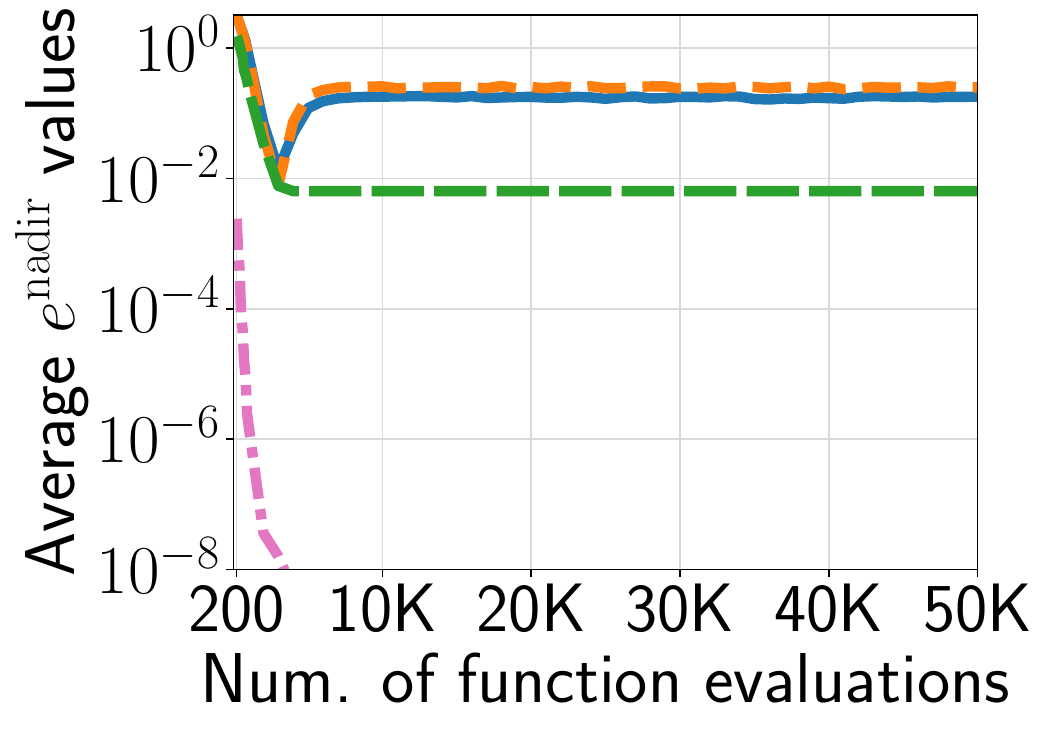}}
   \subfloat[$e^{\mathrm{nadir}}$ ($m=4$)]{\includegraphics[width=0.32\textwidth]{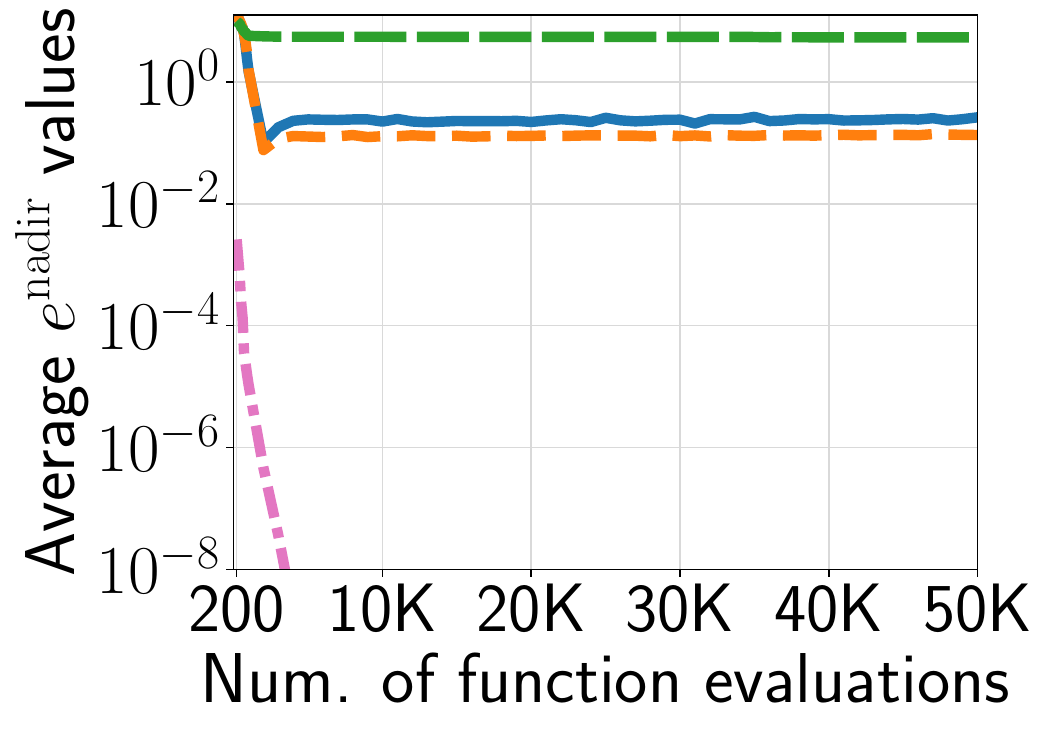}}
   \subfloat[$e^{\mathrm{nadir}}$ ($m=6$)]{\includegraphics[width=0.32\textwidth]{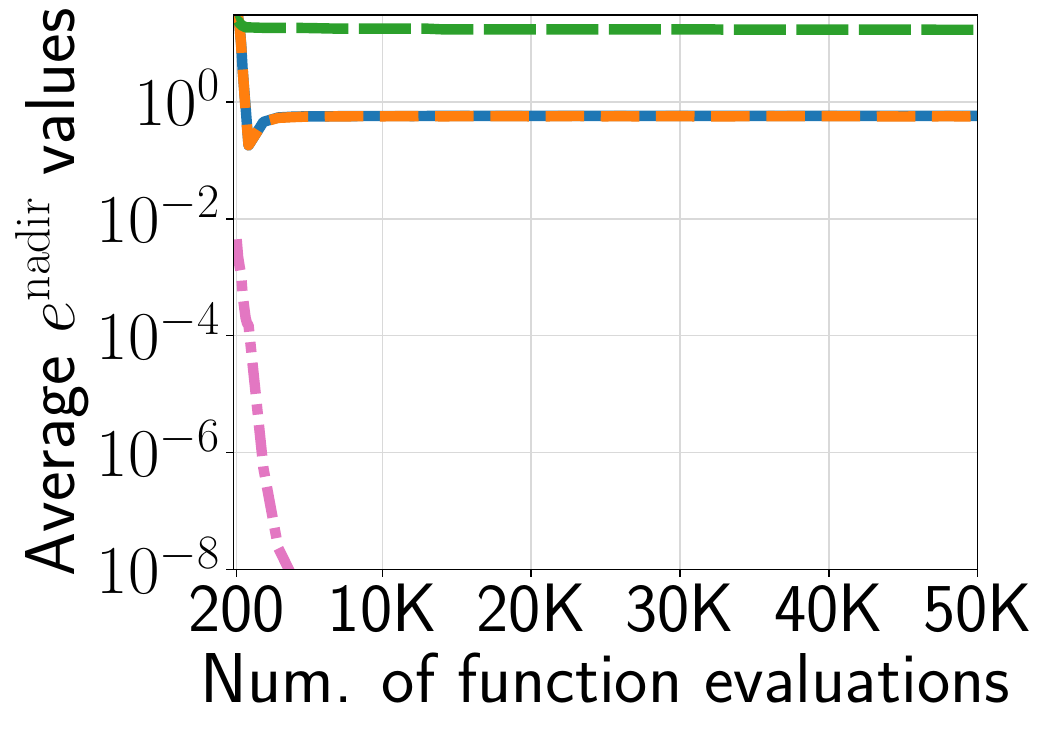}}
\\
   \subfloat[ORE ($m=2$)]{\includegraphics[width=0.32\textwidth]{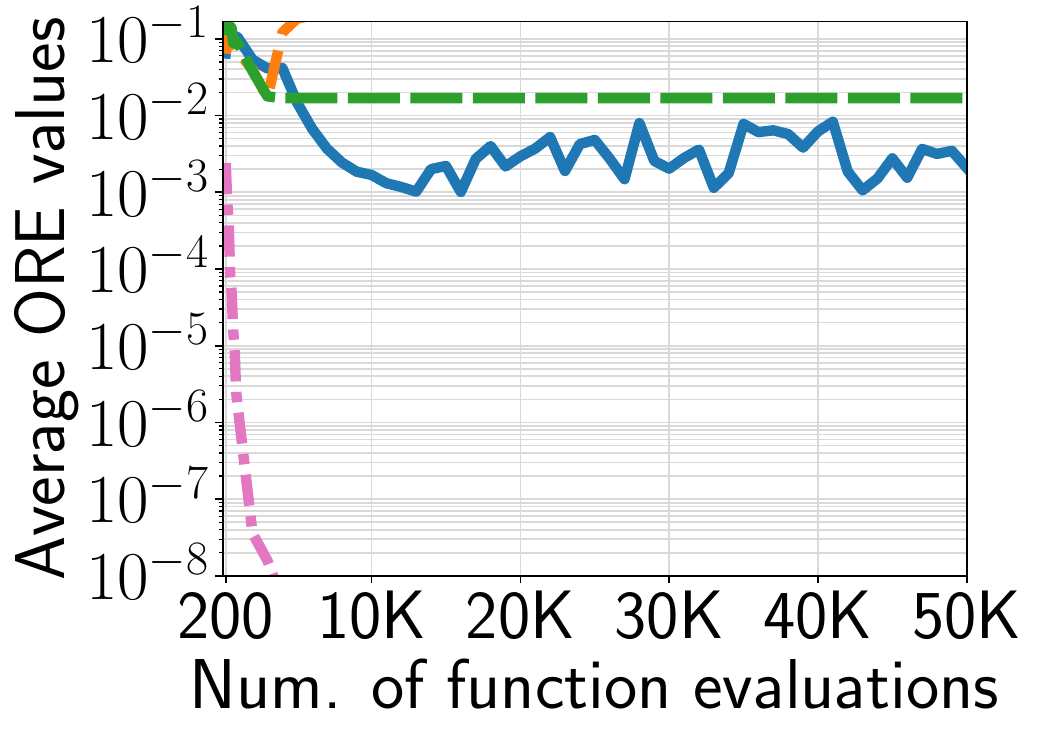}}
   \subfloat[ORE ($m=4$)]{\includegraphics[width=0.32\textwidth]{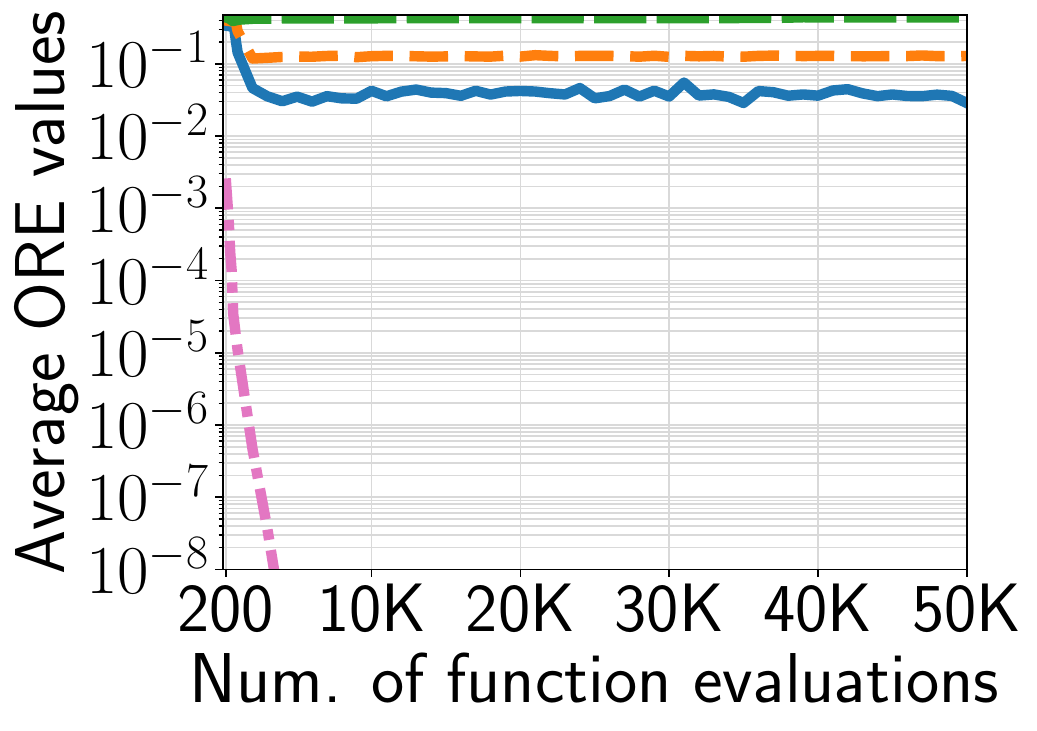}}
   \subfloat[ORE ($m=6$)]{\includegraphics[width=0.32\textwidth]{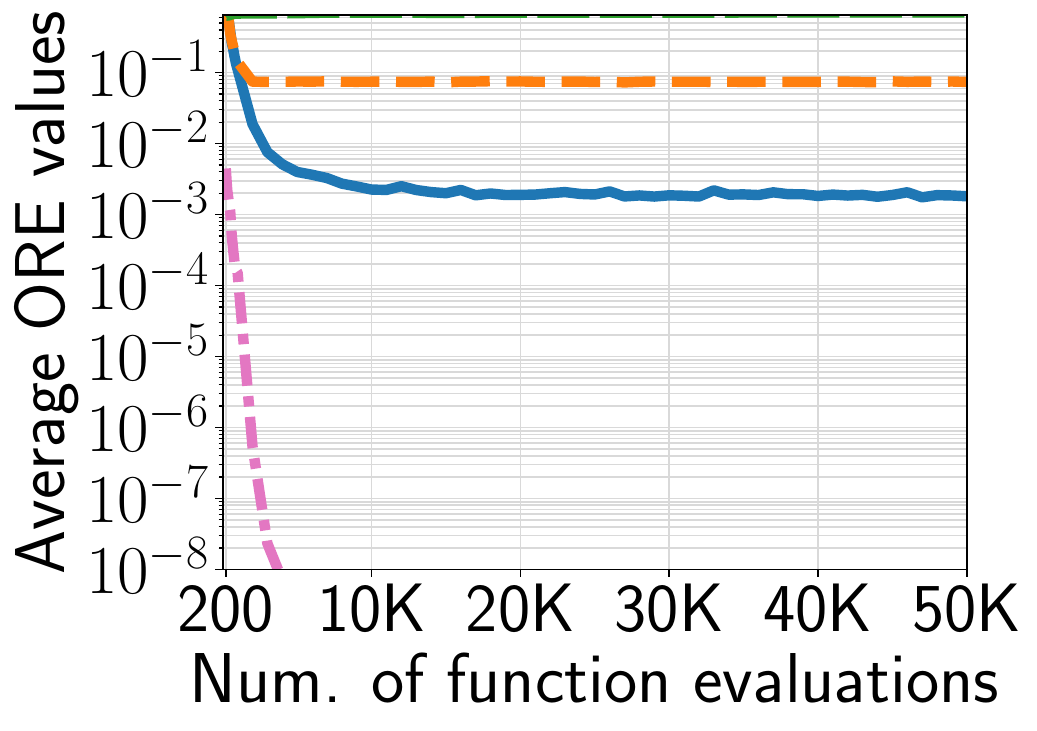}}
\\
\caption{Average $e^{\mathrm{ideal}}$, $e^{\mathrm{nadir}}$, and ORE values of the three normalization methods in R-NSGA-II on DTLZ5.}
\label{supfig:3error_RNSGA2_DTLZ5}
\end{figure*}

\begin{figure*}[t]
\centering
  \subfloat{\includegraphics[width=0.7\textwidth]{./figs/legend/legend_3.pdf}}
\vspace{-3.9mm}
   \\
   \subfloat[$e^{\mathrm{ideal}}$ ($m=2$)]{\includegraphics[width=0.32\textwidth]{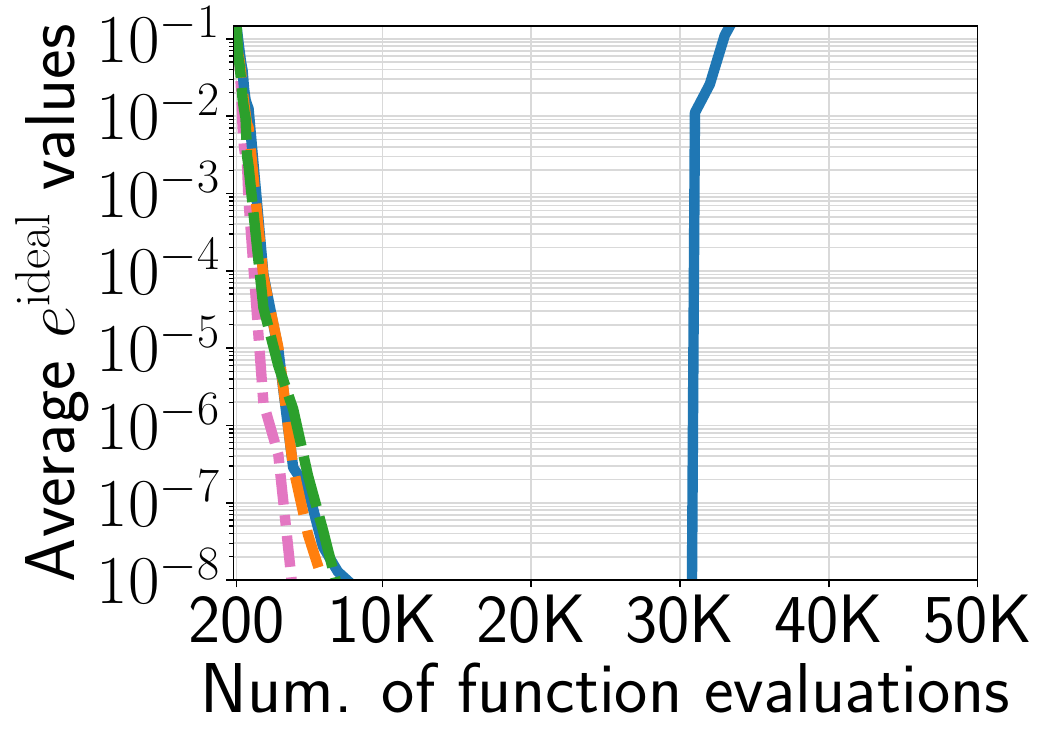}}
   \subfloat[$e^{\mathrm{ideal}}$ ($m=4$)]{\includegraphics[width=0.32\textwidth]{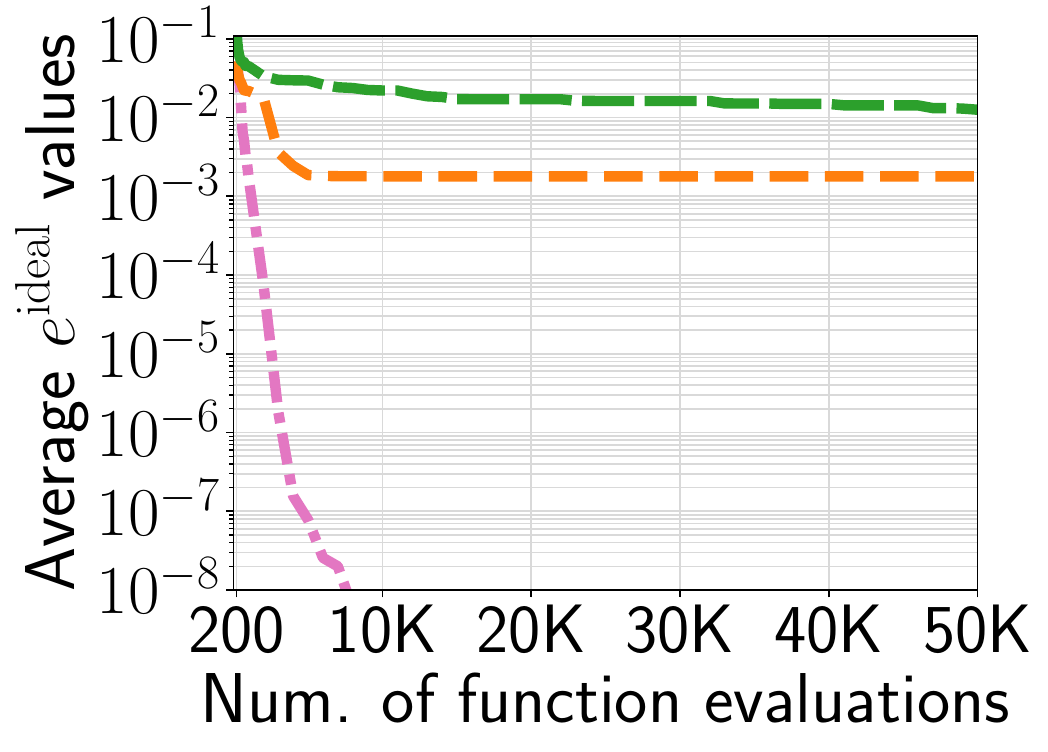}}
   \subfloat[$e^{\mathrm{ideal}}$ ($m=6$)]{\includegraphics[width=0.32\textwidth]{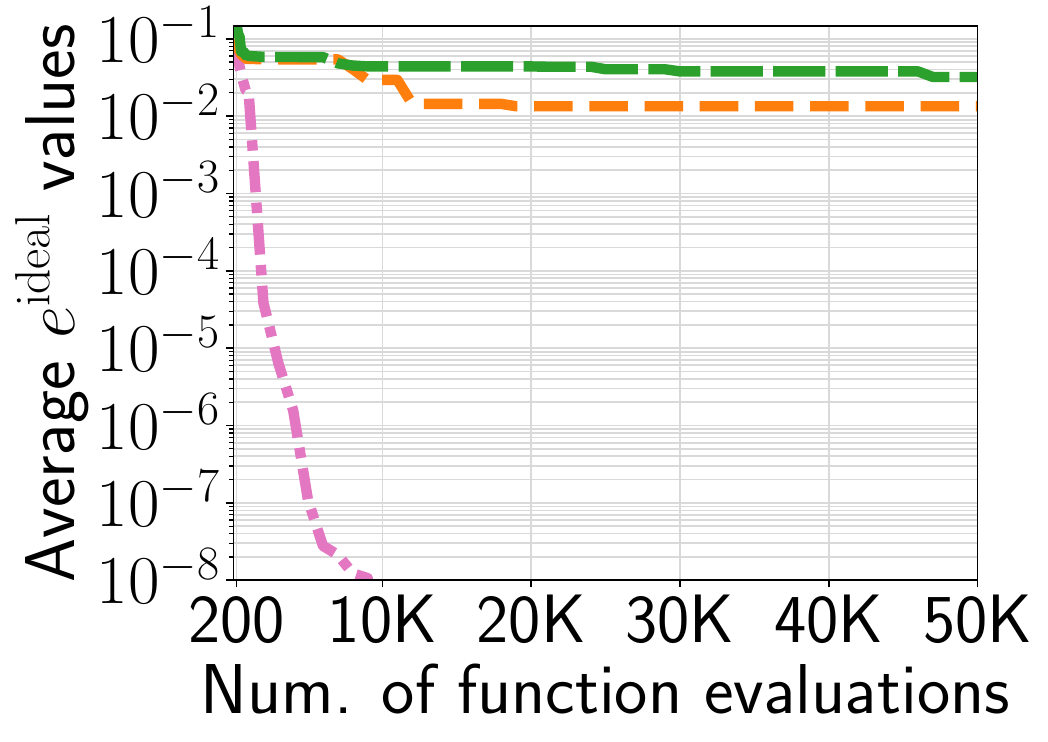}}
\\
   \subfloat[$e^{\mathrm{nadir}}$ ($m=2$)]{\includegraphics[width=0.32\textwidth]{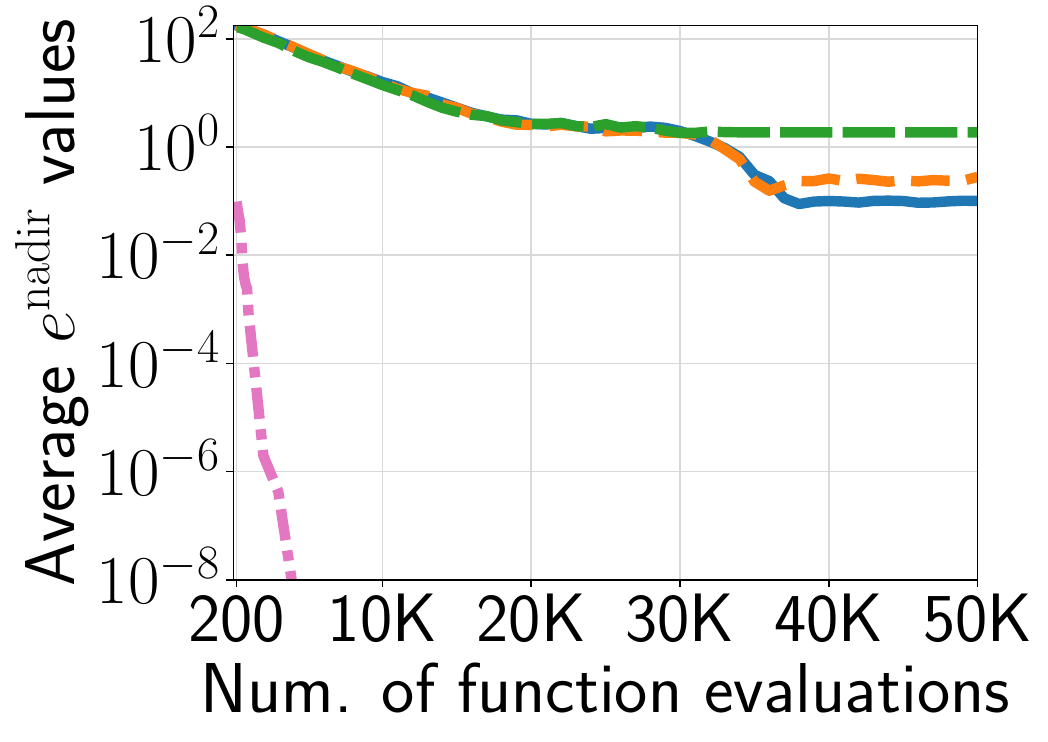}}
   \subfloat[$e^{\mathrm{nadir}}$ ($m=4$)]{\includegraphics[width=0.32\textwidth]{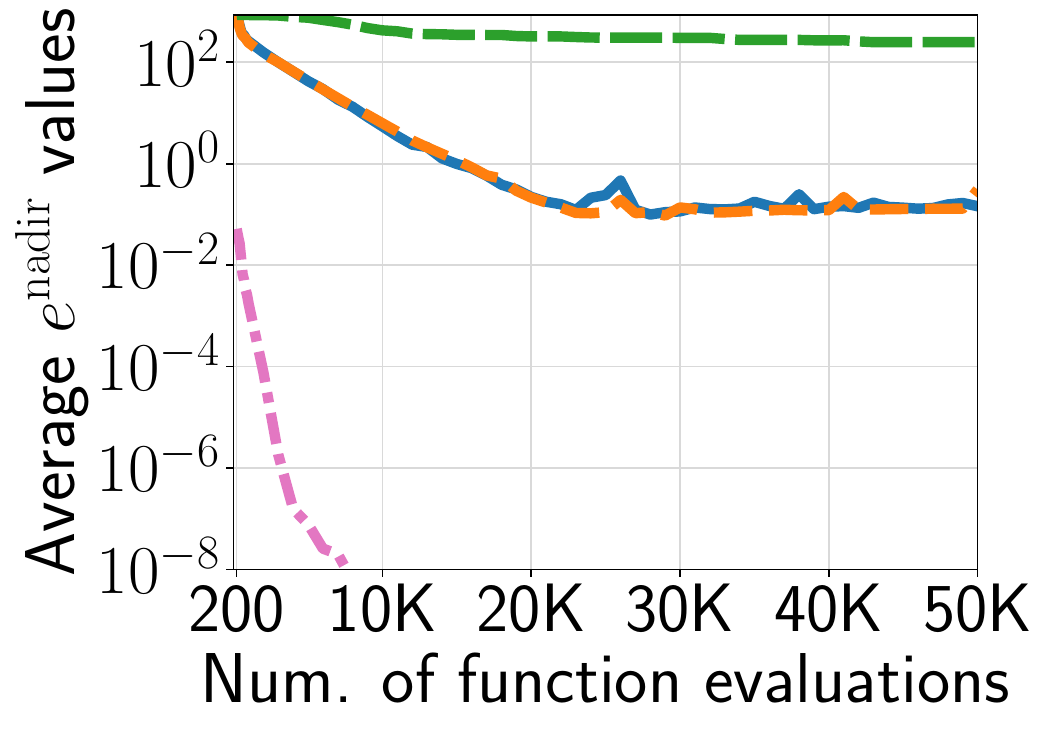}}
   \subfloat[$e^{\mathrm{nadir}}$ ($m=6$)]{\includegraphics[width=0.32\textwidth]{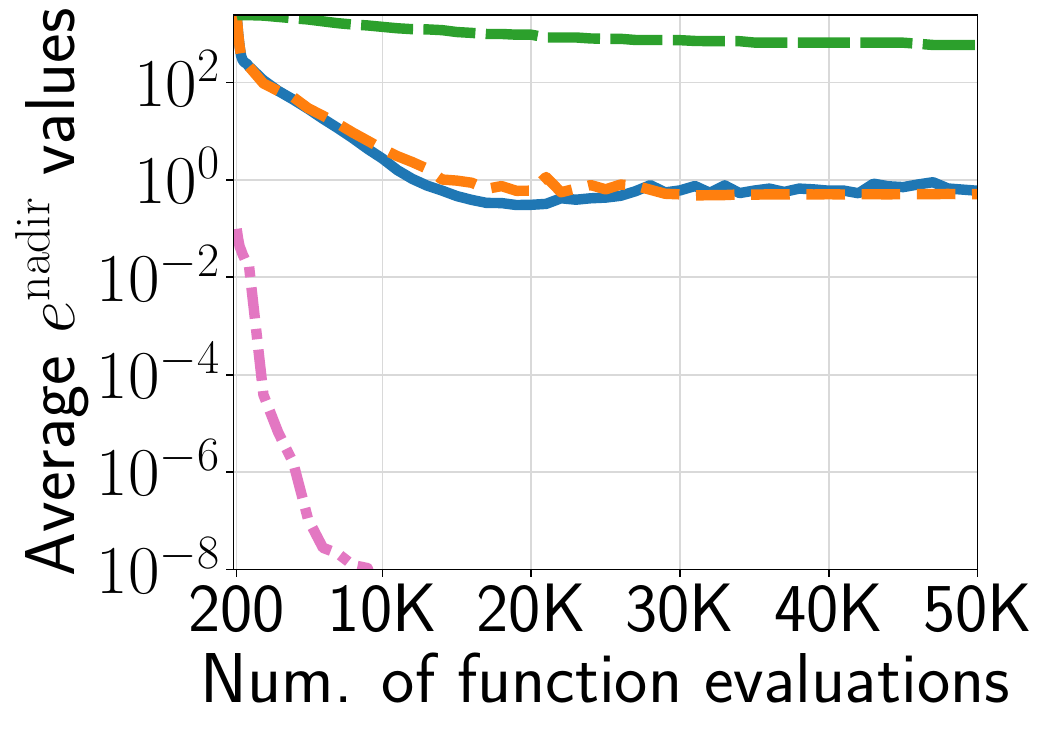}}
\\
   \subfloat[ORE ($m=2$)]{\includegraphics[width=0.32\textwidth]{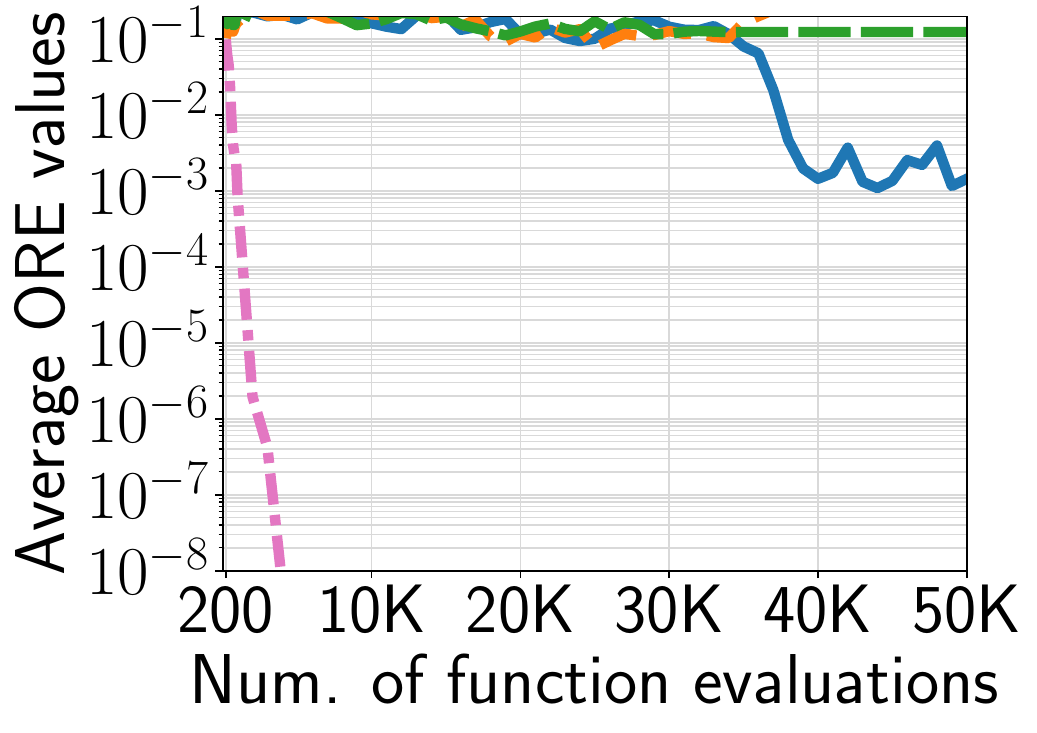}}
   \subfloat[ORE ($m=4$)]{\includegraphics[width=0.32\textwidth]{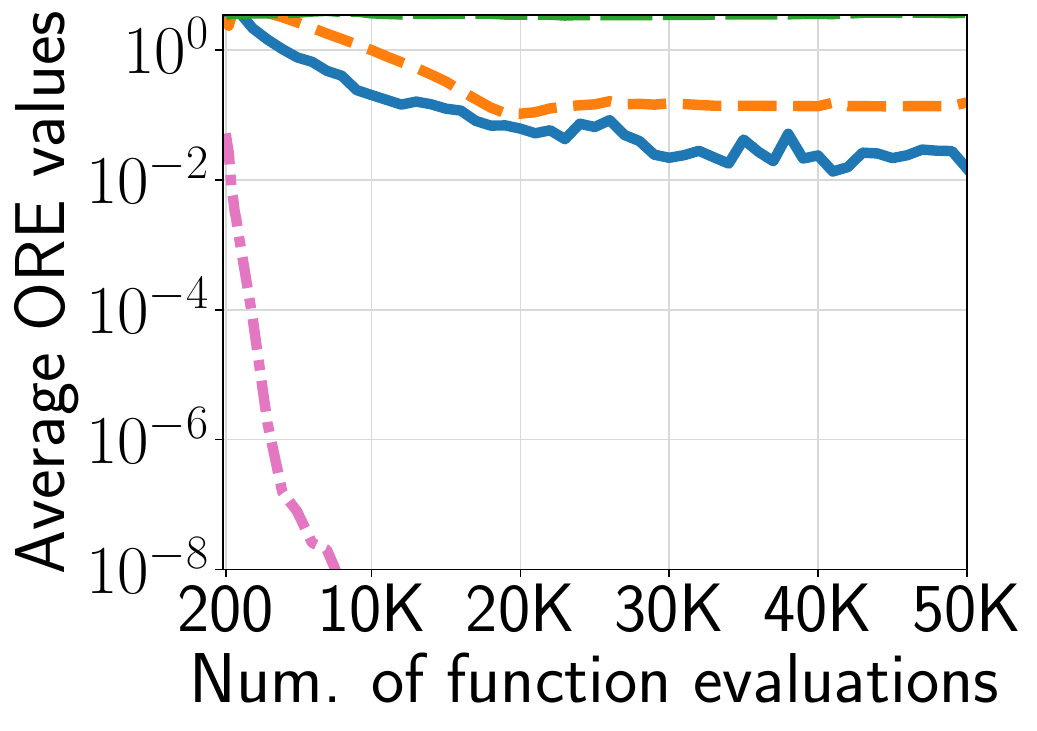}}
   \subfloat[ORE ($m=6$)]{\includegraphics[width=0.32\textwidth]{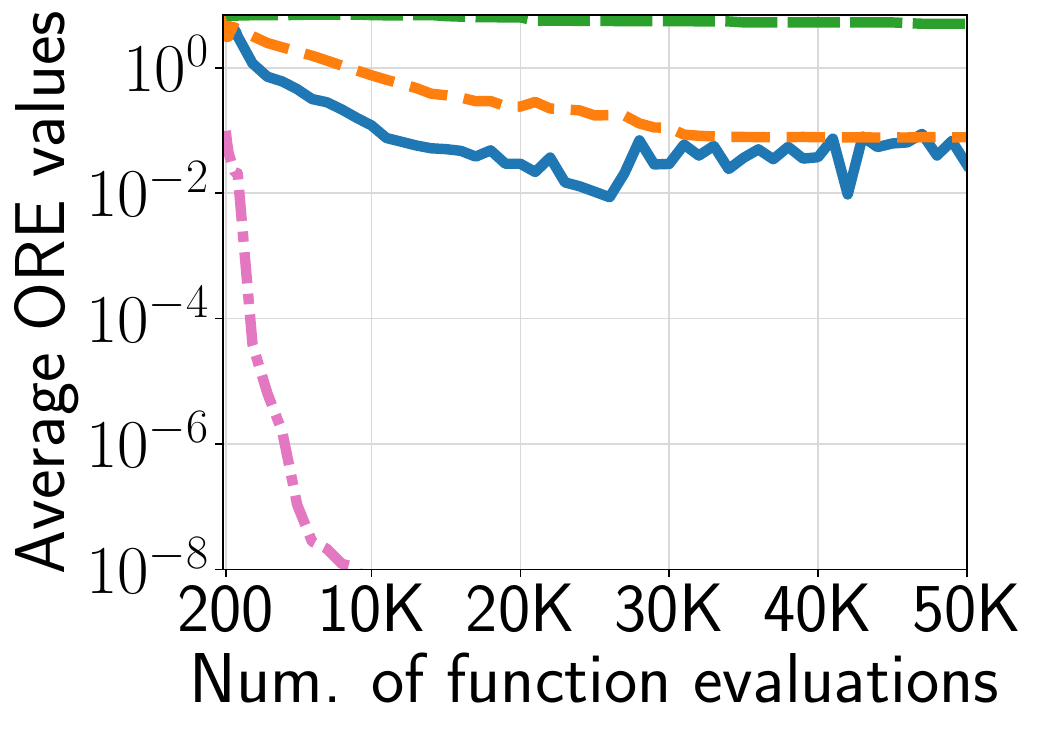}}
\\
\caption{Average $e^{\mathrm{ideal}}$, $e^{\mathrm{nadir}}$, and ORE values of the three normalization methods in R-NSGA-II on DTLZ6.}
\label{supfig:3error_RNSGA2_DTLZ6}
\end{figure*}

\begin{figure*}[t]
\centering
  \subfloat{\includegraphics[width=0.7\textwidth]{./figs/legend/legend_3.pdf}}
\vspace{-3.9mm}
   \\
   \subfloat[$e^{\mathrm{ideal}}$ ($m=2$)]{\includegraphics[width=0.32\textwidth]{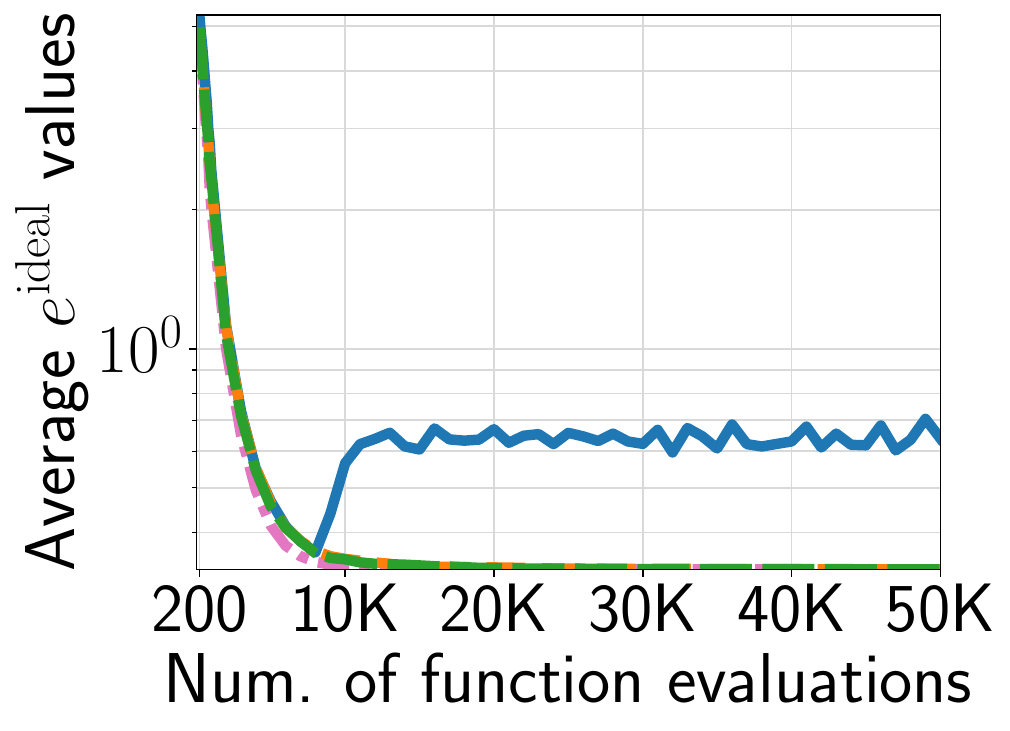}}
   \subfloat[$e^{\mathrm{ideal}}$ ($m=4$)]{\includegraphics[width=0.32\textwidth]{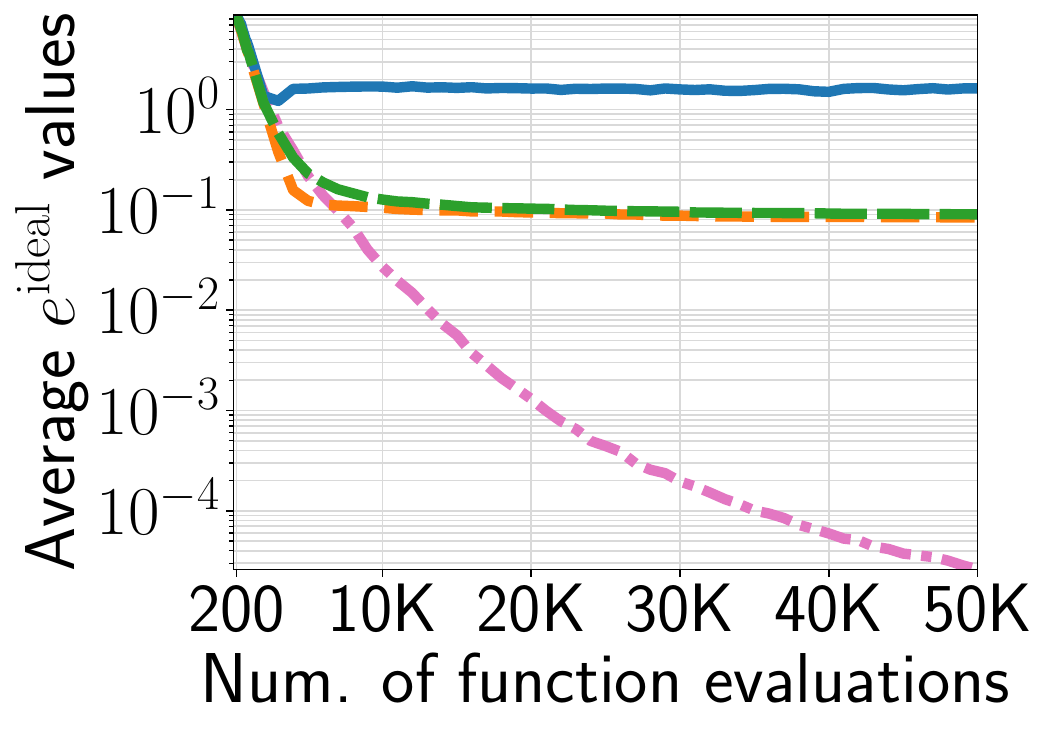}}
   \subfloat[$e^{\mathrm{ideal}}$ ($m=6$)]{\includegraphics[width=0.32\textwidth]{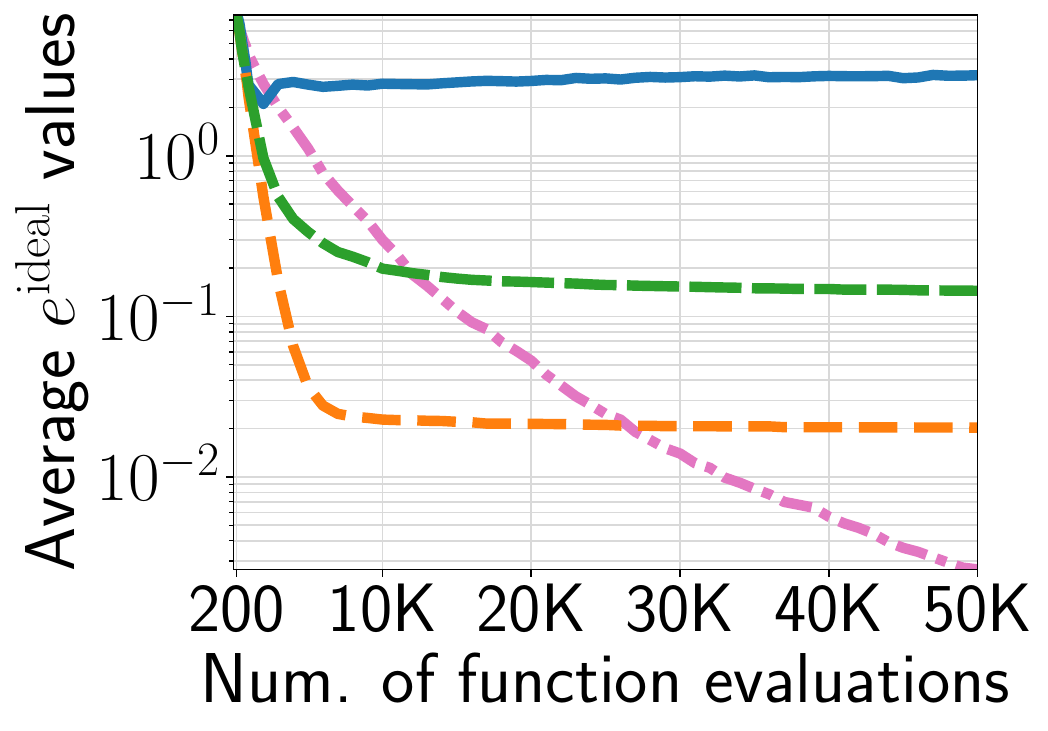}}
\\
   \subfloat[$e^{\mathrm{nadir}}$ ($m=2$)]{\includegraphics[width=0.32\textwidth]{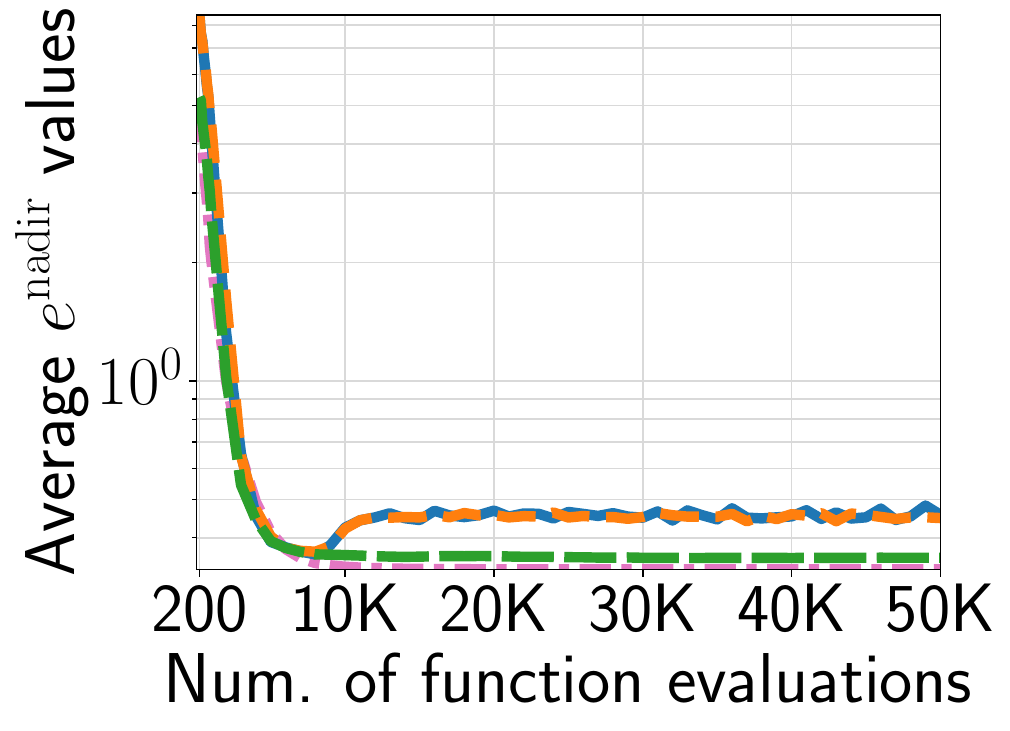}}
   \subfloat[$e^{\mathrm{nadir}}$ ($m=4$)]{\includegraphics[width=0.32\textwidth]{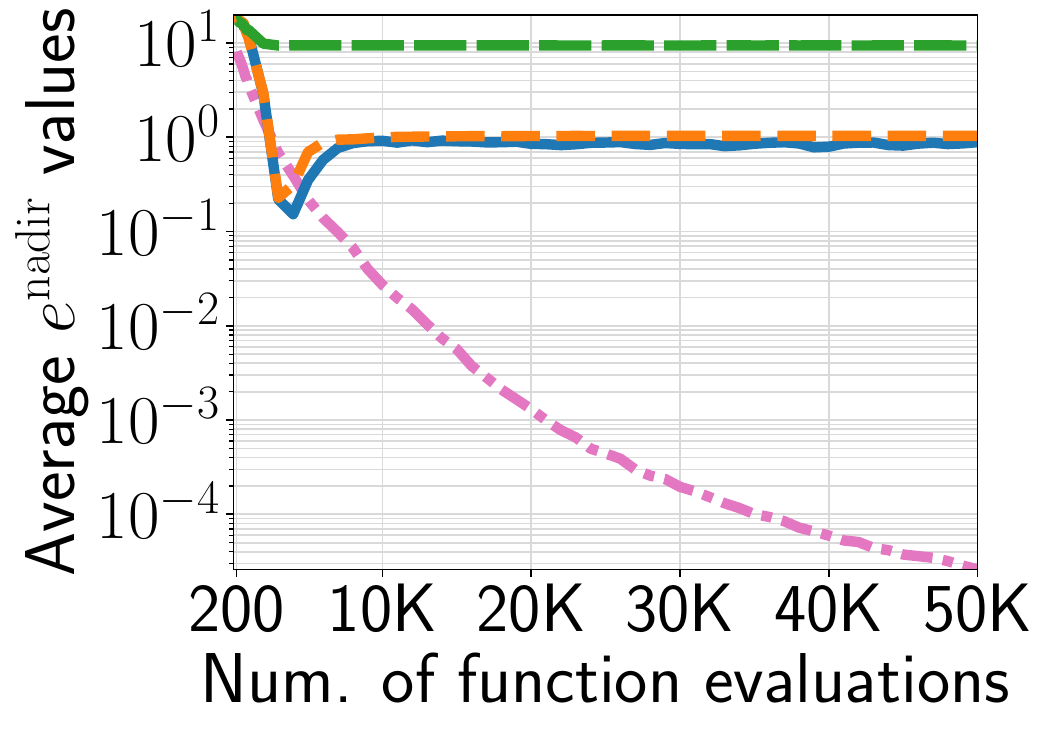}}
   \subfloat[$e^{\mathrm{nadir}}$ ($m=6$)]{\includegraphics[width=0.32\textwidth]{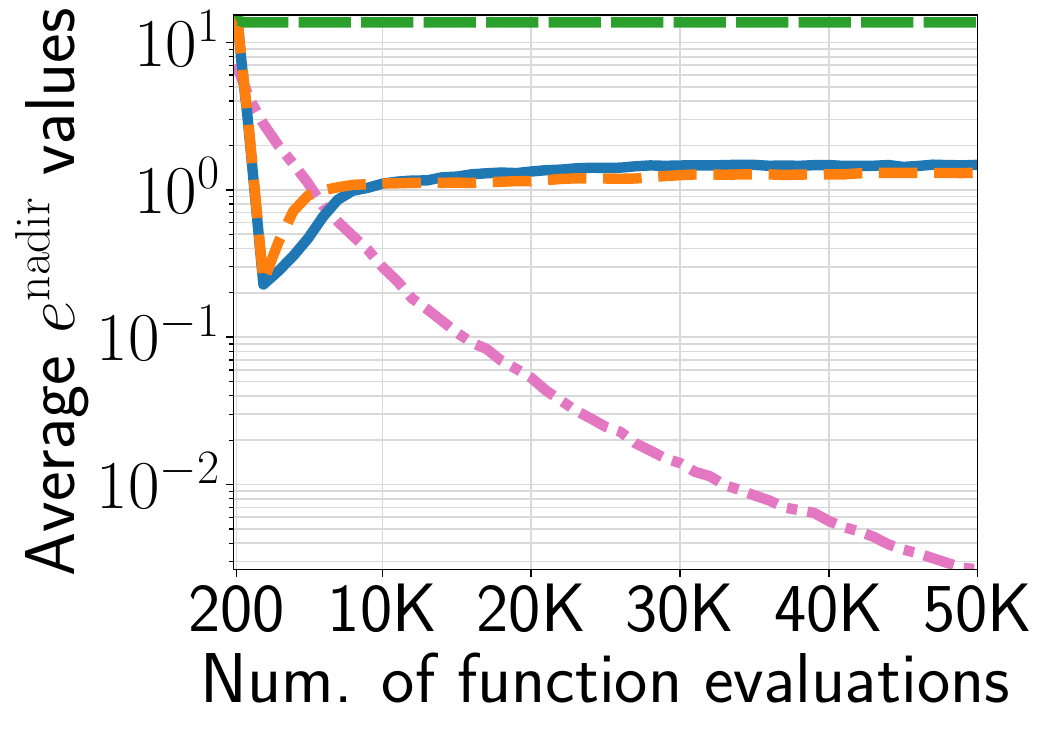}}
\\
   \subfloat[ORE ($m=2$)]{\includegraphics[width=0.32\textwidth]{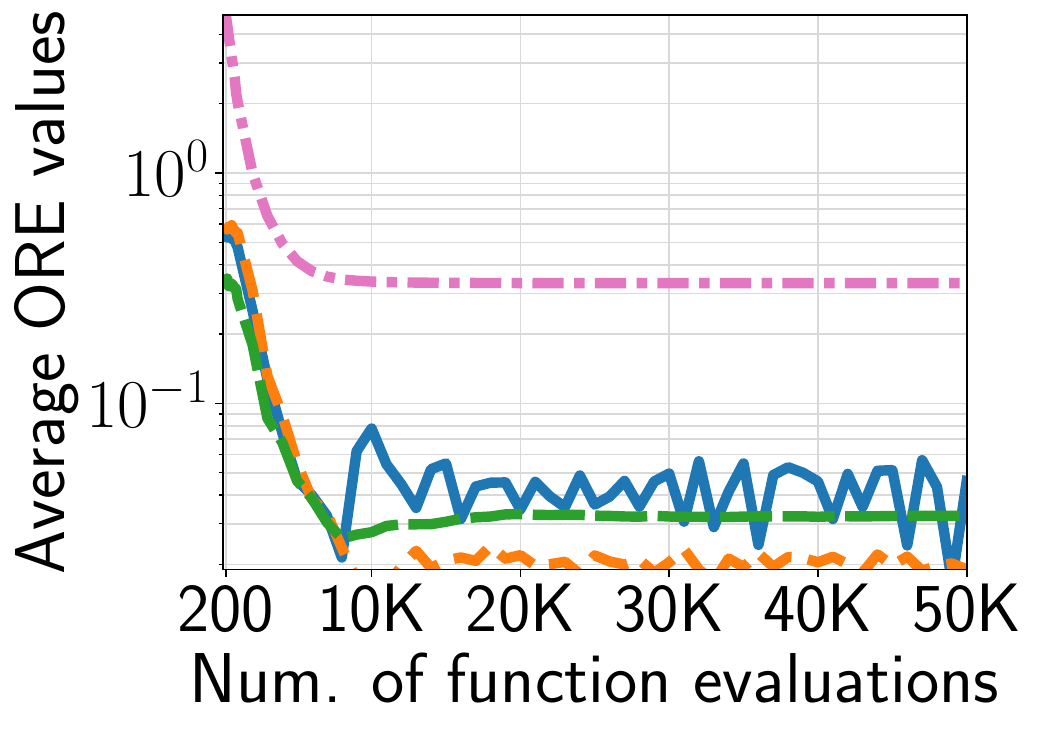}}
   \subfloat[ORE ($m=4$)]{\includegraphics[width=0.32\textwidth]{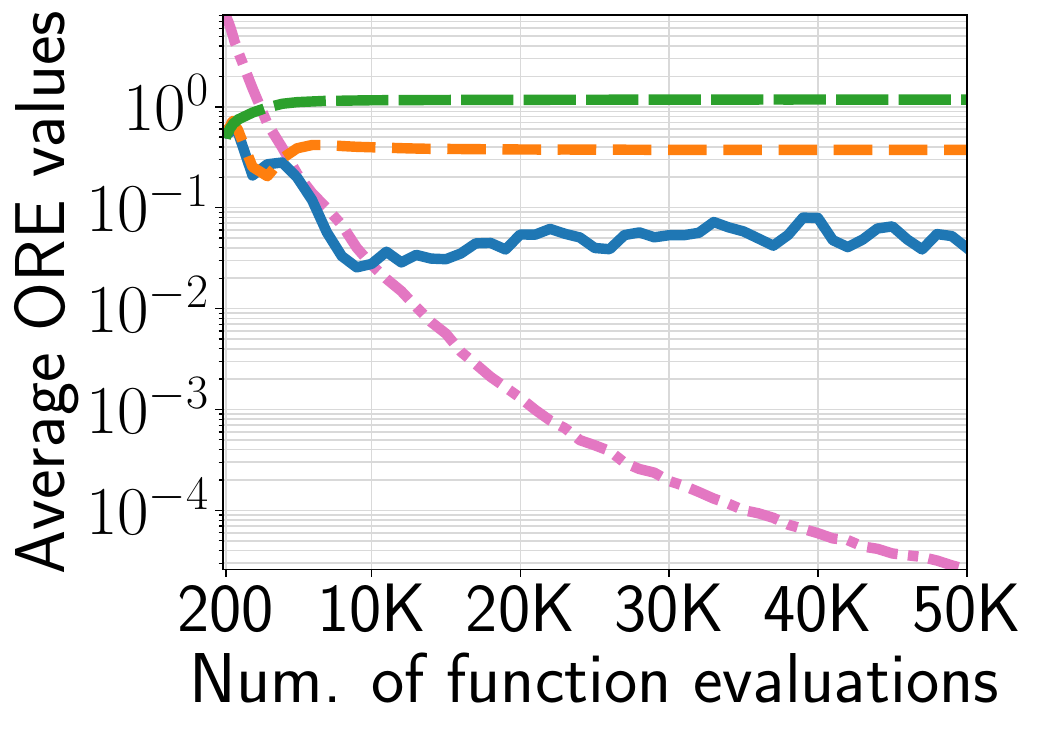}}
   \subfloat[ORE ($m=6$)]{\includegraphics[width=0.32\textwidth]{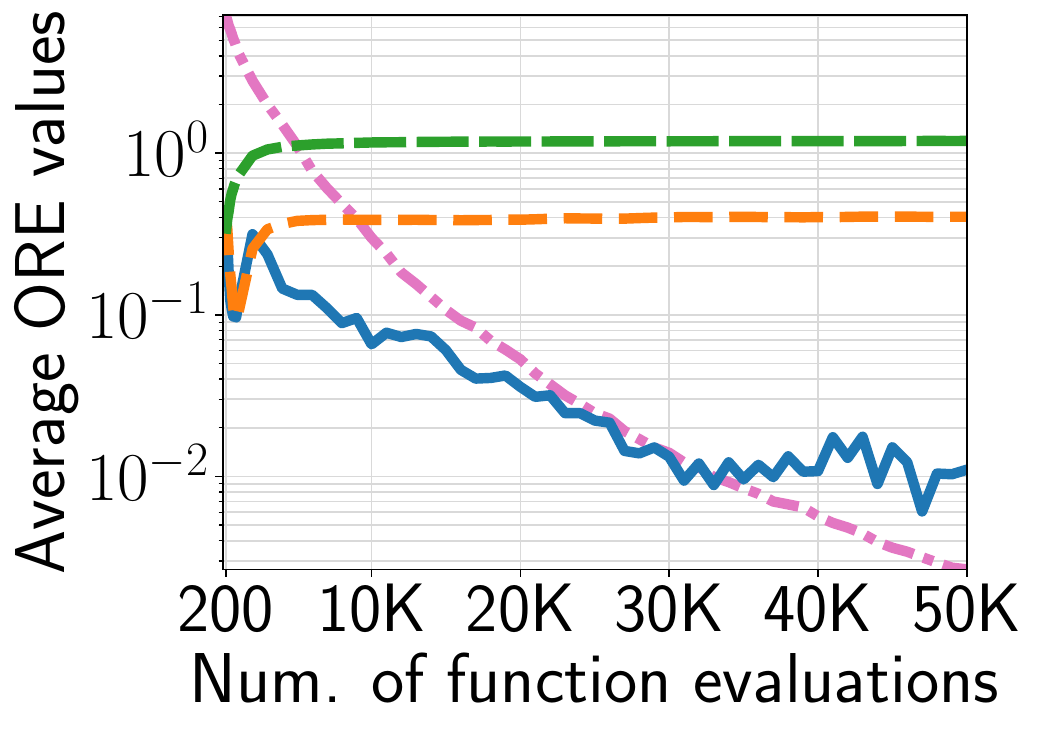}}
\\
\caption{Average $e^{\mathrm{ideal}}$, $e^{\mathrm{nadir}}$, and ORE values of the three normalization methods in R-NSGA-II on DTLZ7.}
\label{supfig:3error_RNSGA2_DTLZ7}
\end{figure*}

\begin{figure*}[t]
\centering
  \subfloat{\includegraphics[width=0.7\textwidth]{./figs/legend/legend_3.pdf}}
\vspace{-3.9mm}
   \\
   \subfloat[$e^{\mathrm{ideal}}$ ($m=2$)]{\includegraphics[width=0.32\textwidth]{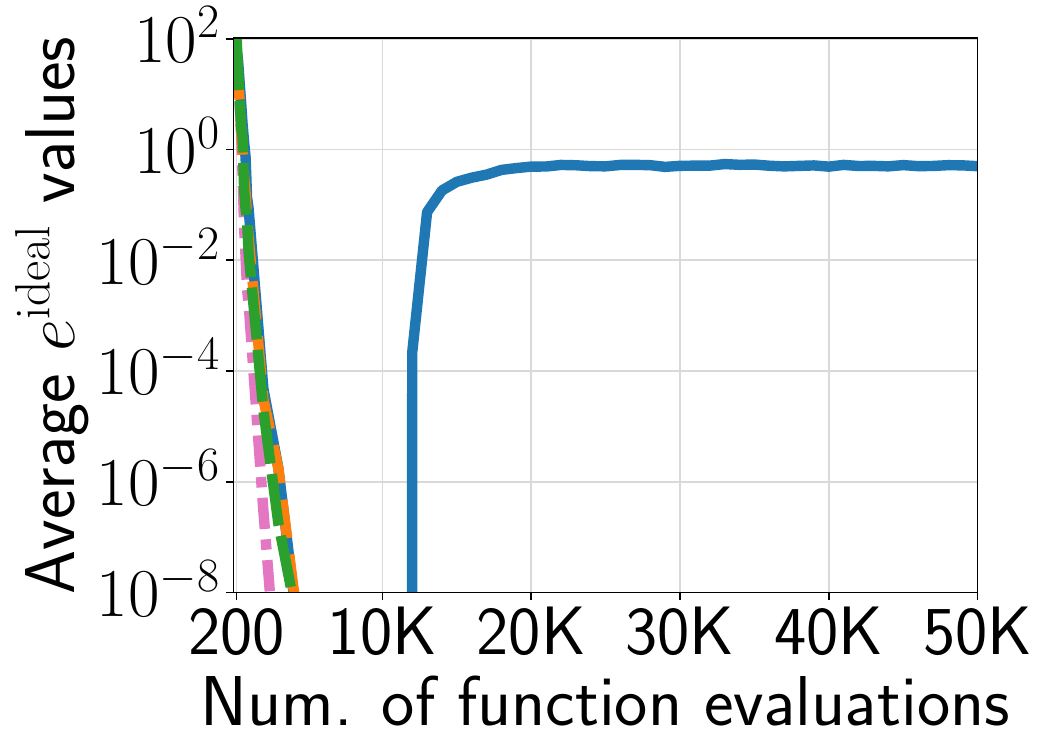}}
   \subfloat[$e^{\mathrm{ideal}}$ ($m=4$)]{\includegraphics[width=0.32\textwidth]{./figs/qi_error_ideal/RNSGA2_mu100/SDTLZ1_m4_r0.1_z-type1.pdf}}
   \subfloat[$e^{\mathrm{ideal}}$ ($m=6$)]{\includegraphics[width=0.32\textwidth]{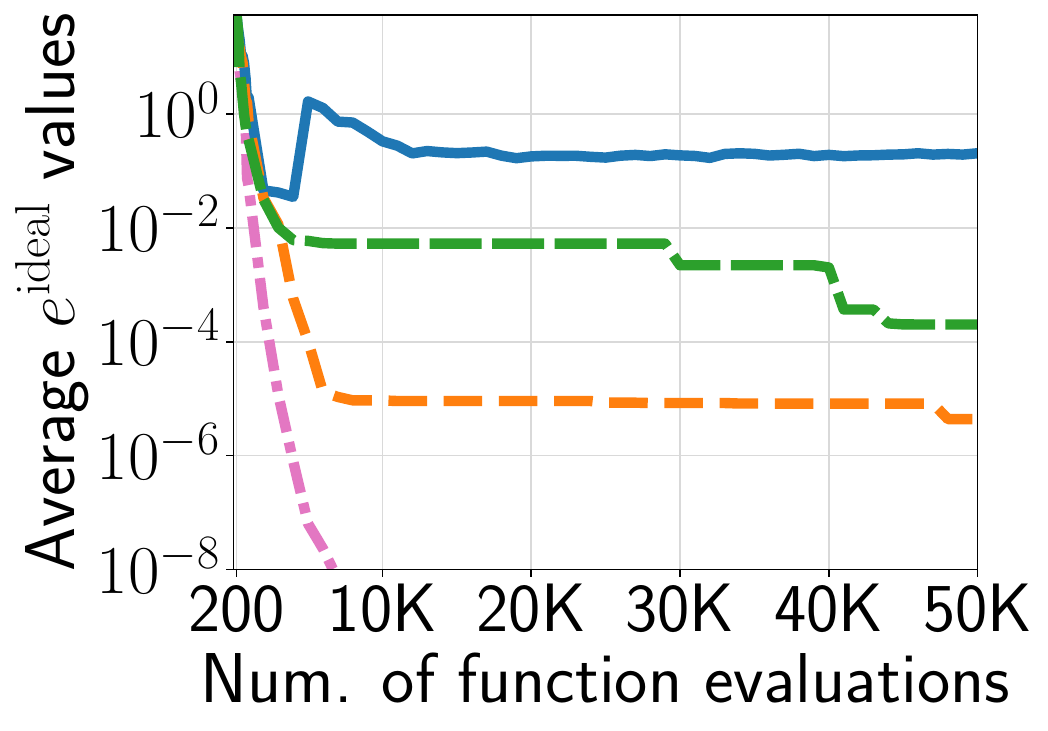}}
\\
   \subfloat[$e^{\mathrm{nadir}}$ ($m=2$)]{\includegraphics[width=0.32\textwidth]{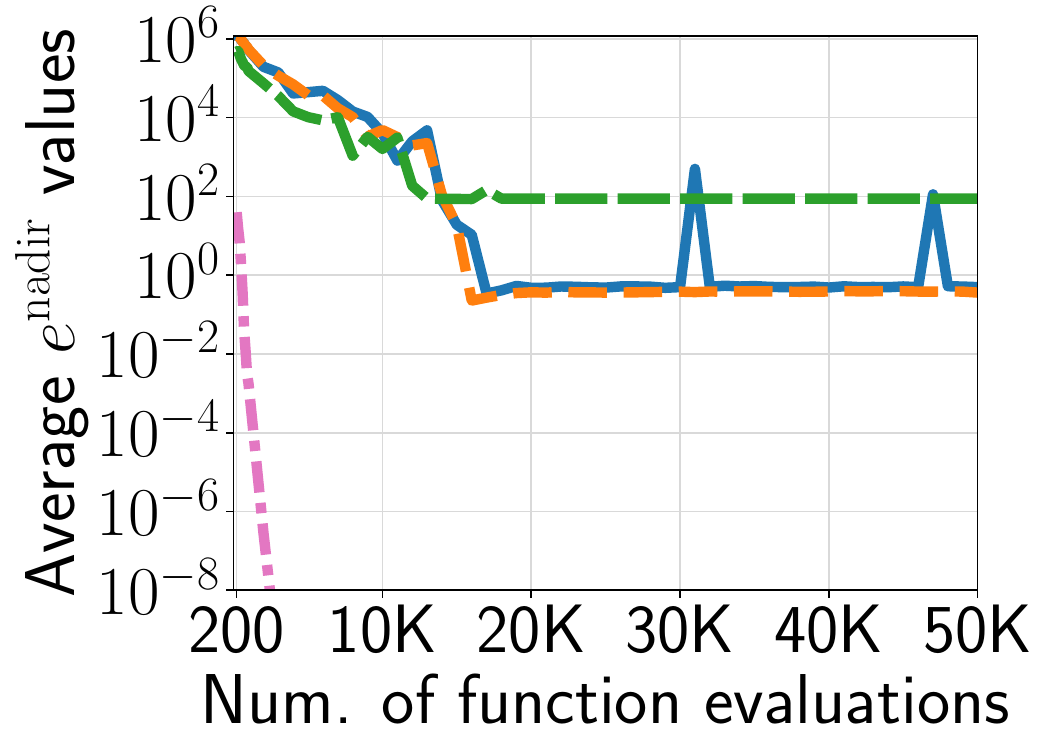}}
   \subfloat[$e^{\mathrm{nadir}}$ ($m=4$)]{\includegraphics[width=0.32\textwidth]{./figs/qi_error_nadir/RNSGA2_mu100/SDTLZ1_m4_r0.1_z-type1.pdf}}
   \subfloat[$e^{\mathrm{nadir}}$ ($m=6$)]{\includegraphics[width=0.32\textwidth]{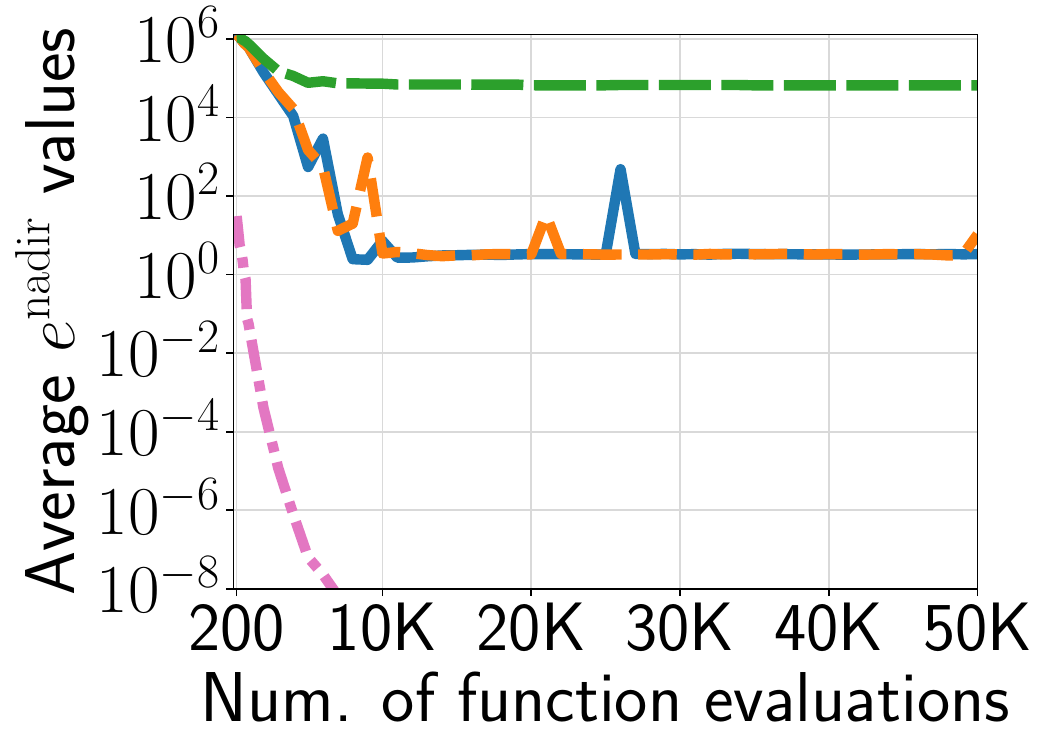}}
\\
   \subfloat[ORE ($m=2$)]{\includegraphics[width=0.32\textwidth]{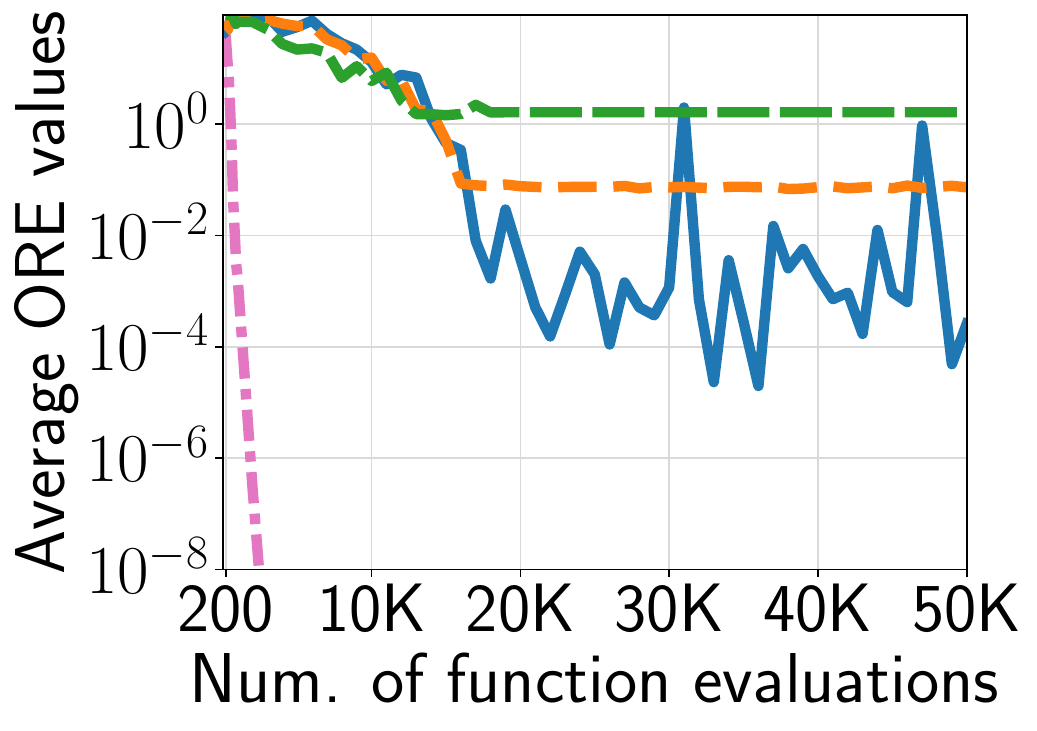}}
   \subfloat[ORE ($m=4$)]{\includegraphics[width=0.32\textwidth]{./figs/qi_ore/RNSGA2_mu100/SDTLZ1_m4_r0.1_z-type1.pdf}}
   \subfloat[ORE ($m=6$)]{\includegraphics[width=0.32\textwidth]{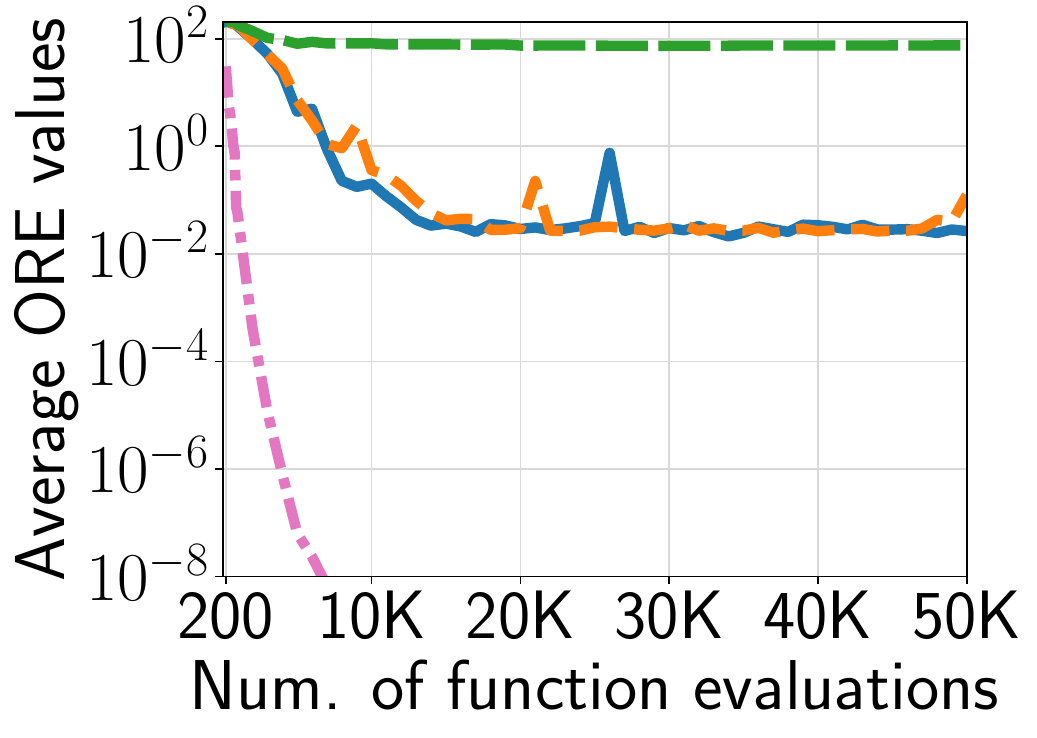}}
\\
\caption{Average $e^{\mathrm{ideal}}$, $e^{\mathrm{nadir}}$, and ORE values of the three normalization methods in R-NSGA-II on SDTLZ1.}
\label{supfig:3error_RNSGA2_SDTLZ1}
\end{figure*}

\begin{figure*}[t]
\centering
  \subfloat{\includegraphics[width=0.7\textwidth]{./figs/legend/legend_3.pdf}}
\vspace{-3.9mm}
   \\
   \subfloat[$e^{\mathrm{ideal}}$ ($m=2$)]{\includegraphics[width=0.32\textwidth]{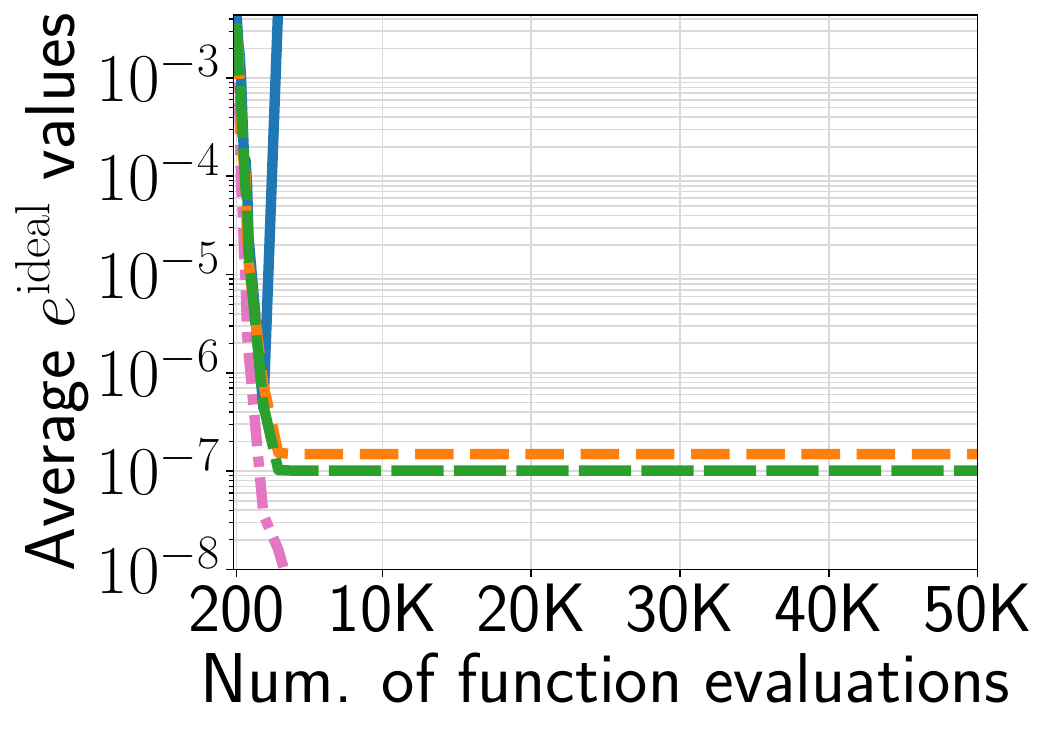}}
   \subfloat[$e^{\mathrm{ideal}}$ ($m=4$)]{\includegraphics[width=0.32\textwidth]{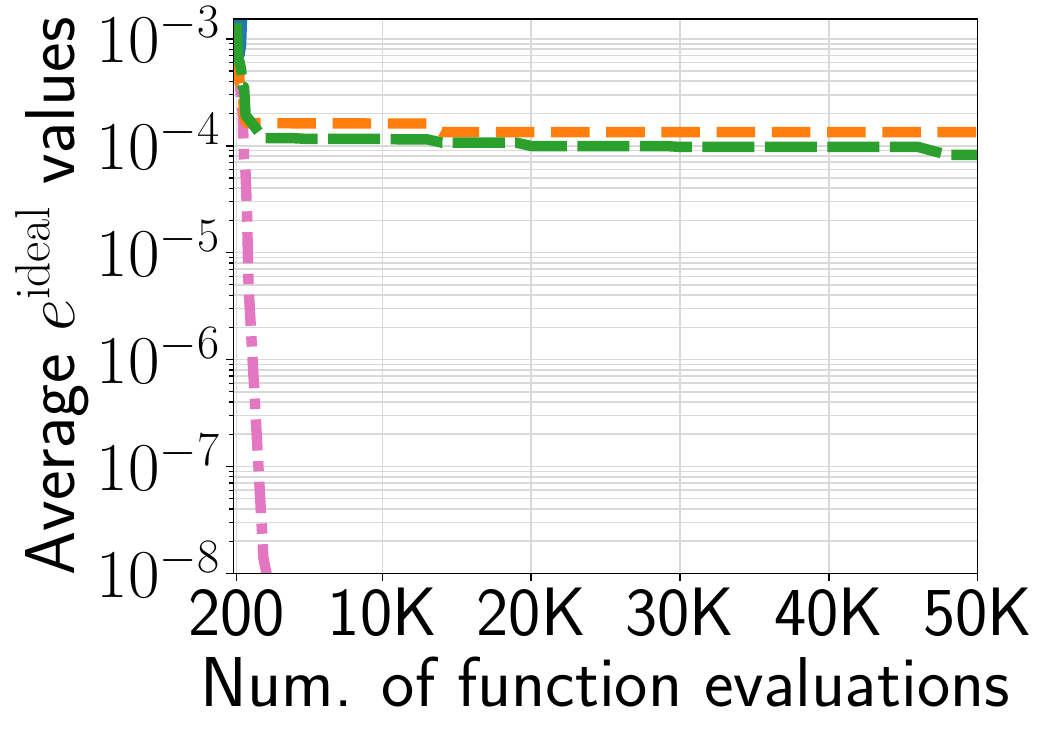}}
   \subfloat[$e^{\mathrm{ideal}}$ ($m=6$)]{\includegraphics[width=0.32\textwidth]{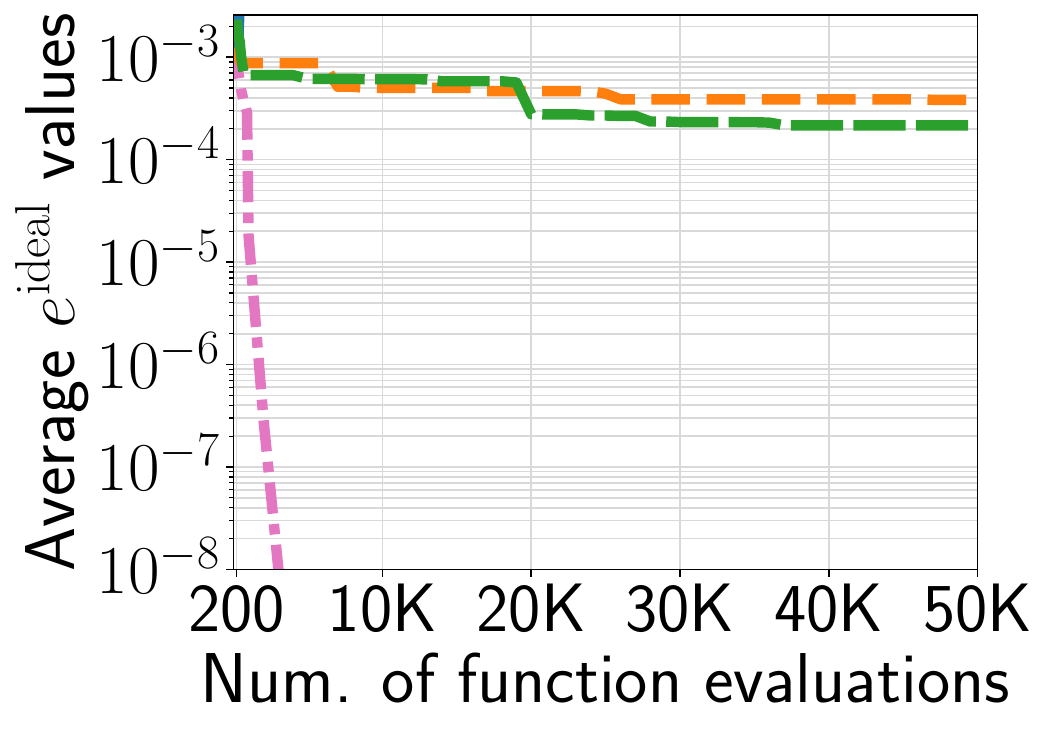}}
\\
   \subfloat[$e^{\mathrm{nadir}}$ ($m=2$)]{\includegraphics[width=0.32\textwidth]{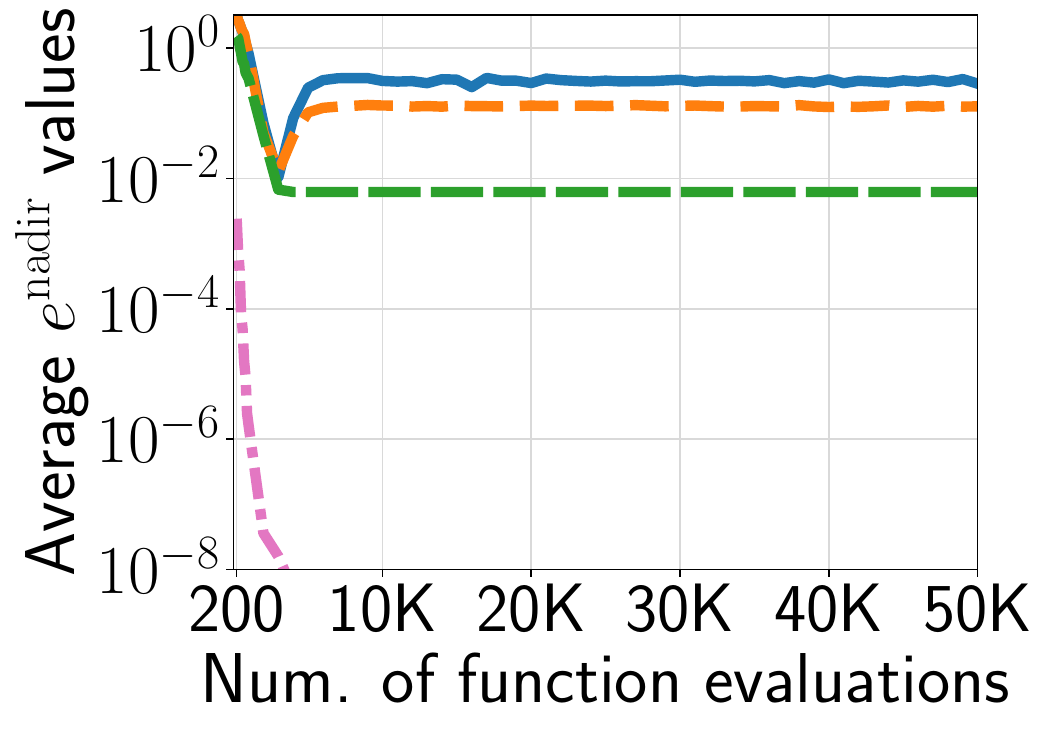}}
   \subfloat[$e^{\mathrm{nadir}}$ ($m=4$)]{\includegraphics[width=0.32\textwidth]{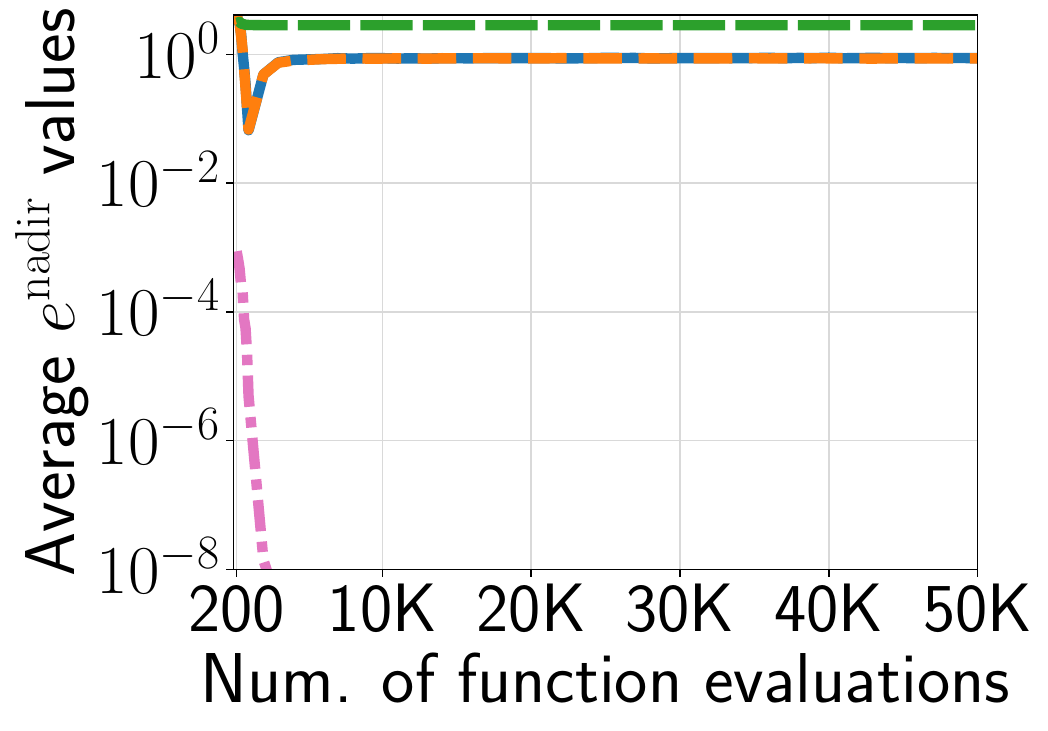}}
   \subfloat[$e^{\mathrm{nadir}}$ ($m=6$)]{\includegraphics[width=0.32\textwidth]{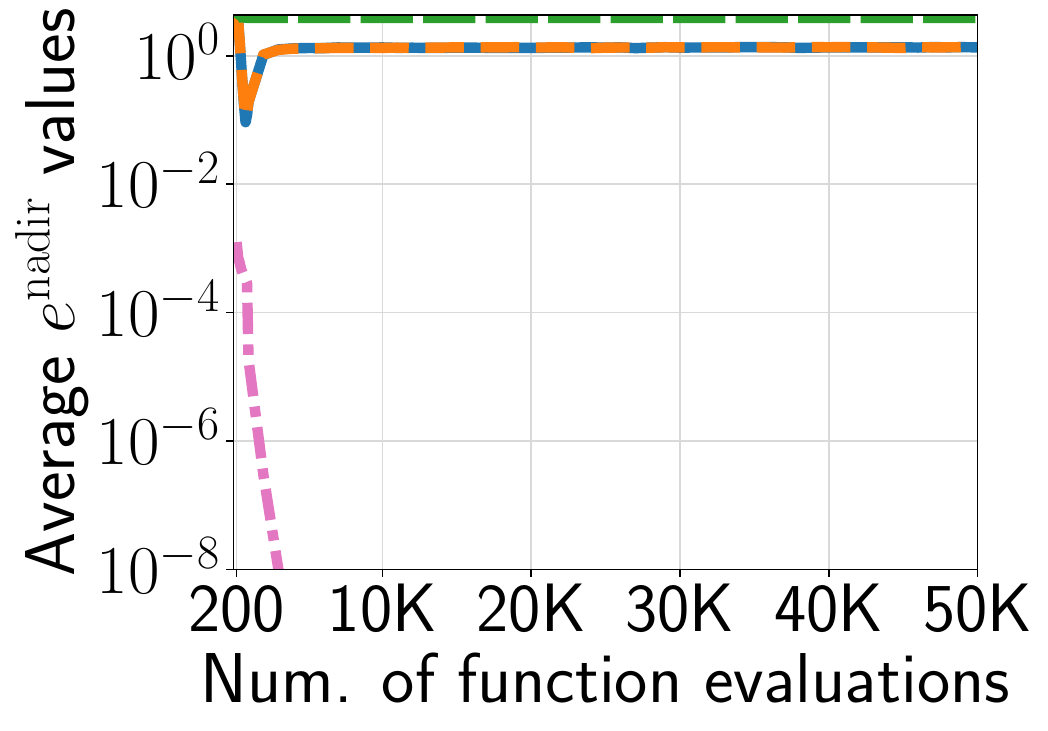}}
\\
   \subfloat[ORE ($m=2$)]{\includegraphics[width=0.32\textwidth]{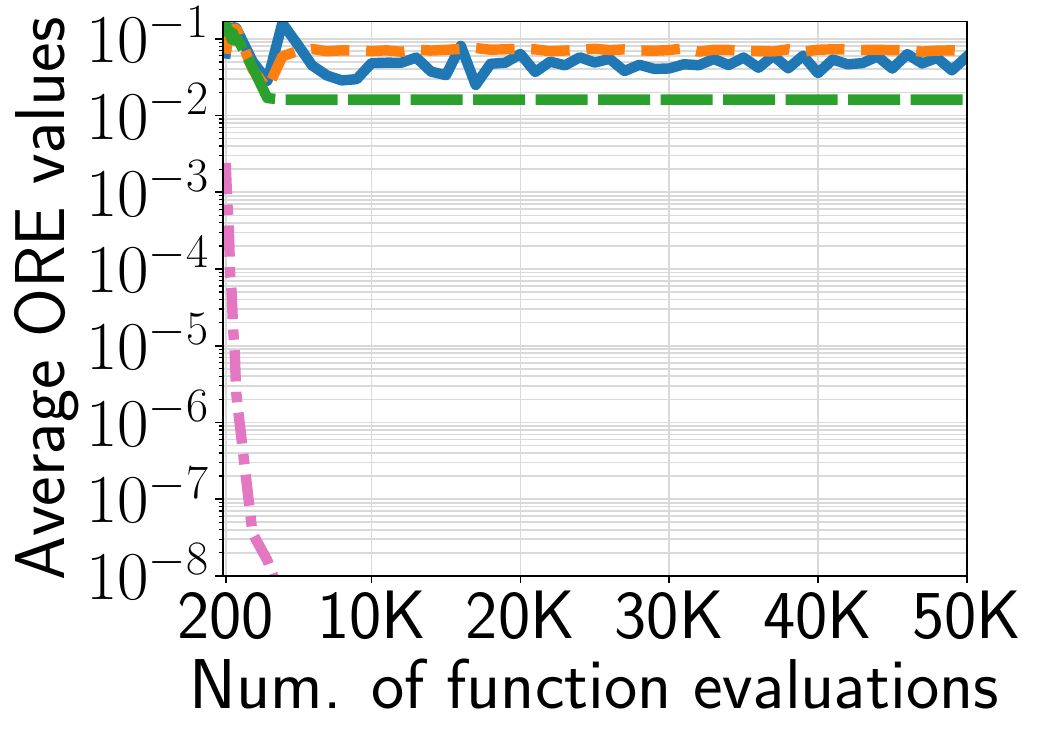}}
   \subfloat[ORE ($m=4$)]{\includegraphics[width=0.32\textwidth]{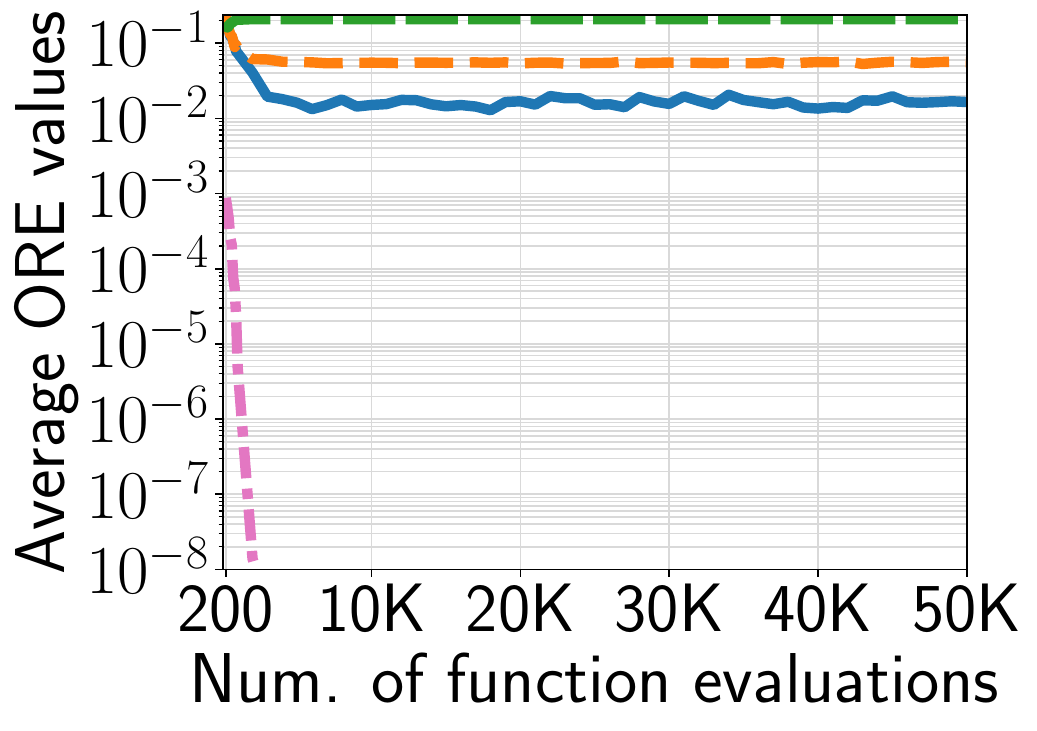}}
   \subfloat[ORE ($m=6$)]{\includegraphics[width=0.32\textwidth]{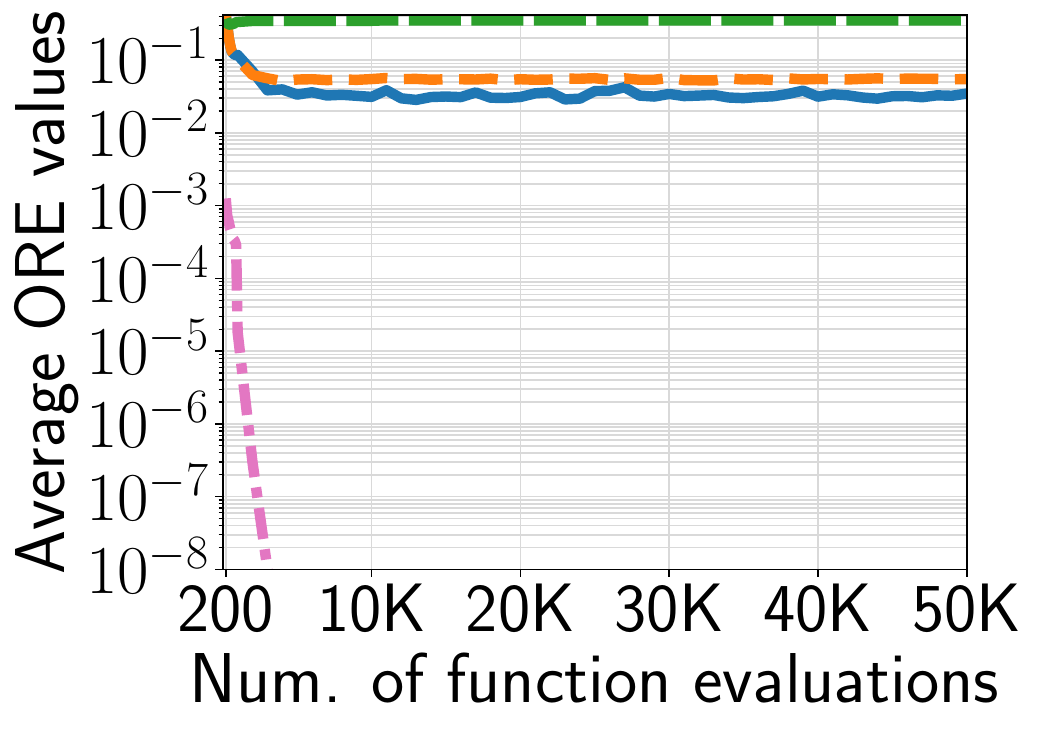}}
\\
\caption{Average $e^{\mathrm{ideal}}$, $e^{\mathrm{nadir}}$, and ORE values of the three normalization methods in R-NSGA-II on SDTLZ2.}
\label{supfig:3error_RNSGA2_SDTLZ2}
\end{figure*}

\begin{figure*}[t]
\centering
  \subfloat{\includegraphics[width=0.7\textwidth]{./figs/legend/legend_3.pdf}}
\vspace{-3.9mm}
   \\
   \subfloat[$e^{\mathrm{ideal}}$ ($m=2$)]{\includegraphics[width=0.32\textwidth]{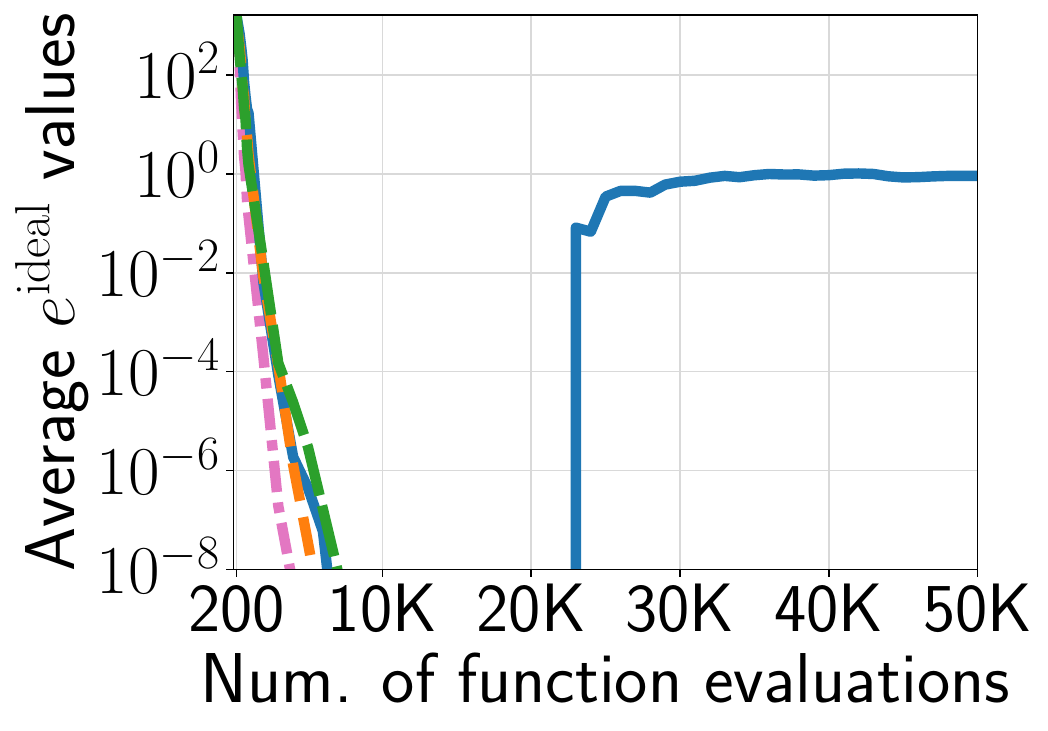}}
   \subfloat[$e^{\mathrm{ideal}}$ ($m=4$)]{\includegraphics[width=0.32\textwidth]{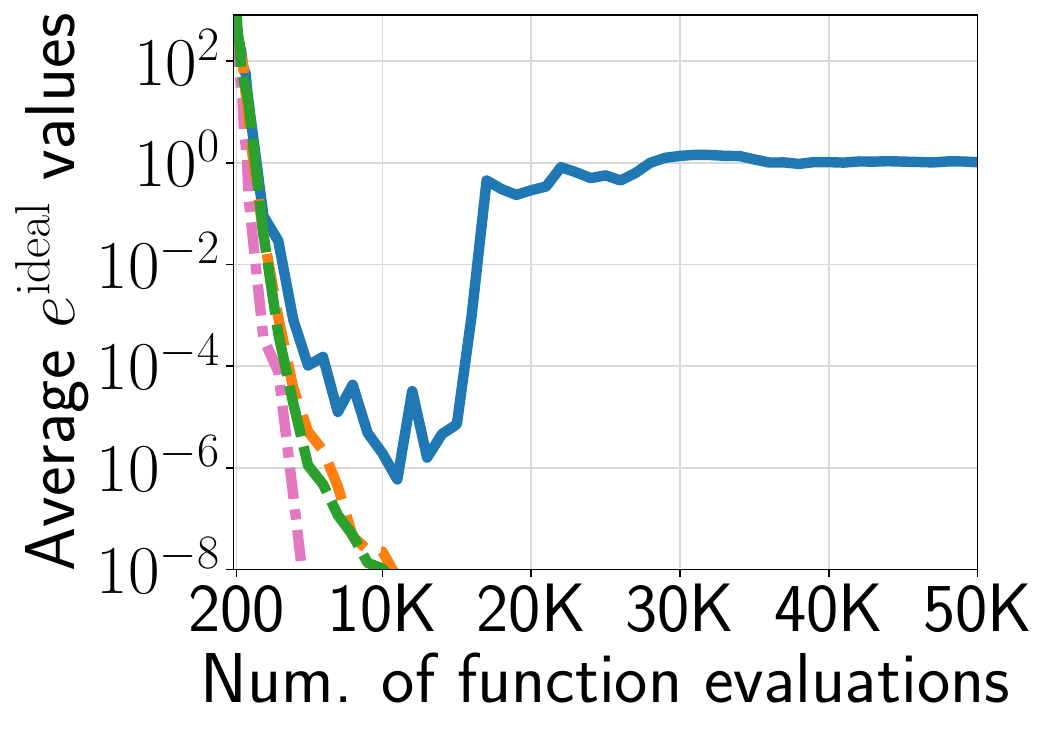}}
   \subfloat[$e^{\mathrm{ideal}}$ ($m=6$)]{\includegraphics[width=0.32\textwidth]{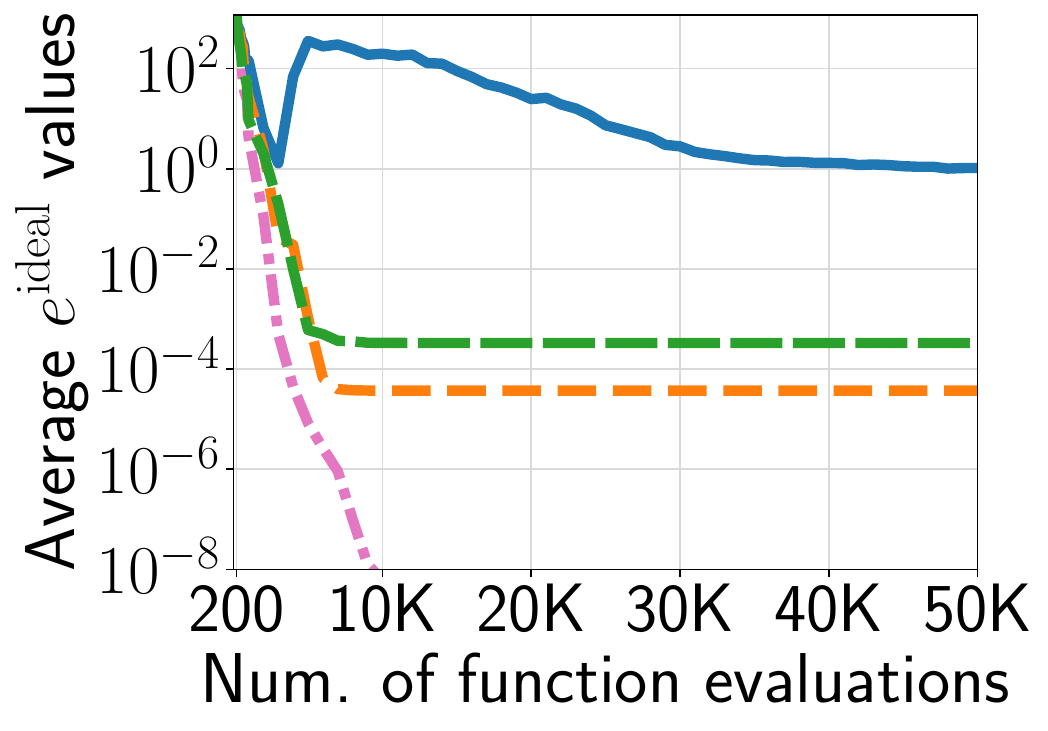}}
\\
   \subfloat[$e^{\mathrm{nadir}}$ ($m=2$)]{\includegraphics[width=0.32\textwidth]{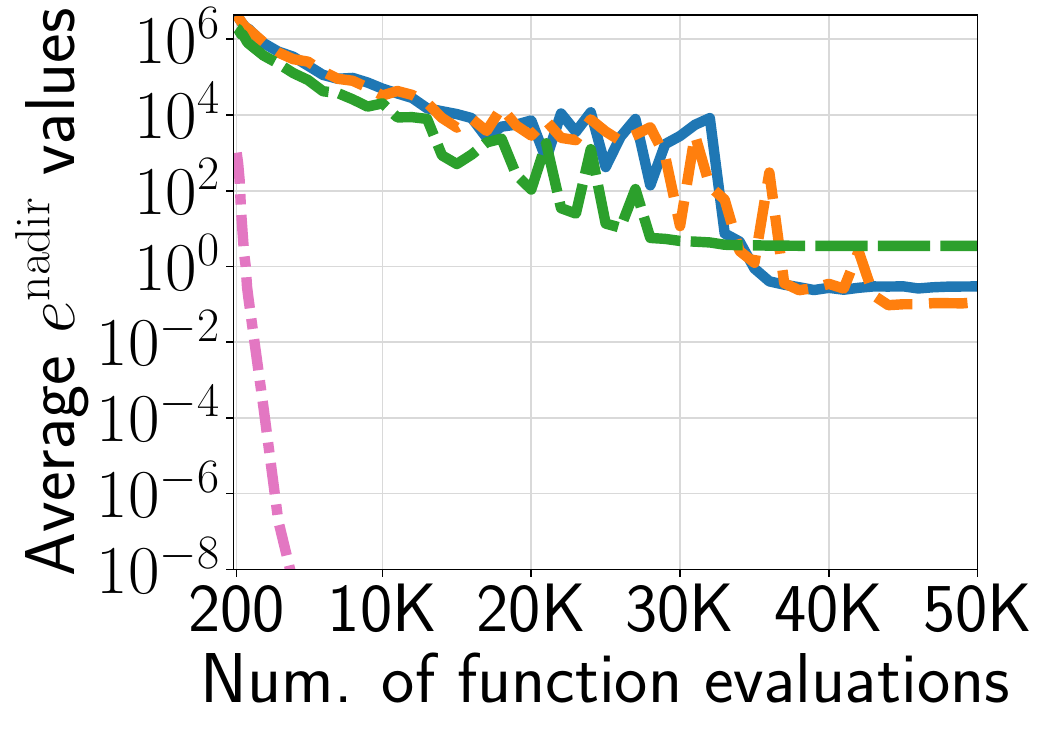}}
   \subfloat[$e^{\mathrm{nadir}}$ ($m=4$)]{\includegraphics[width=0.32\textwidth]{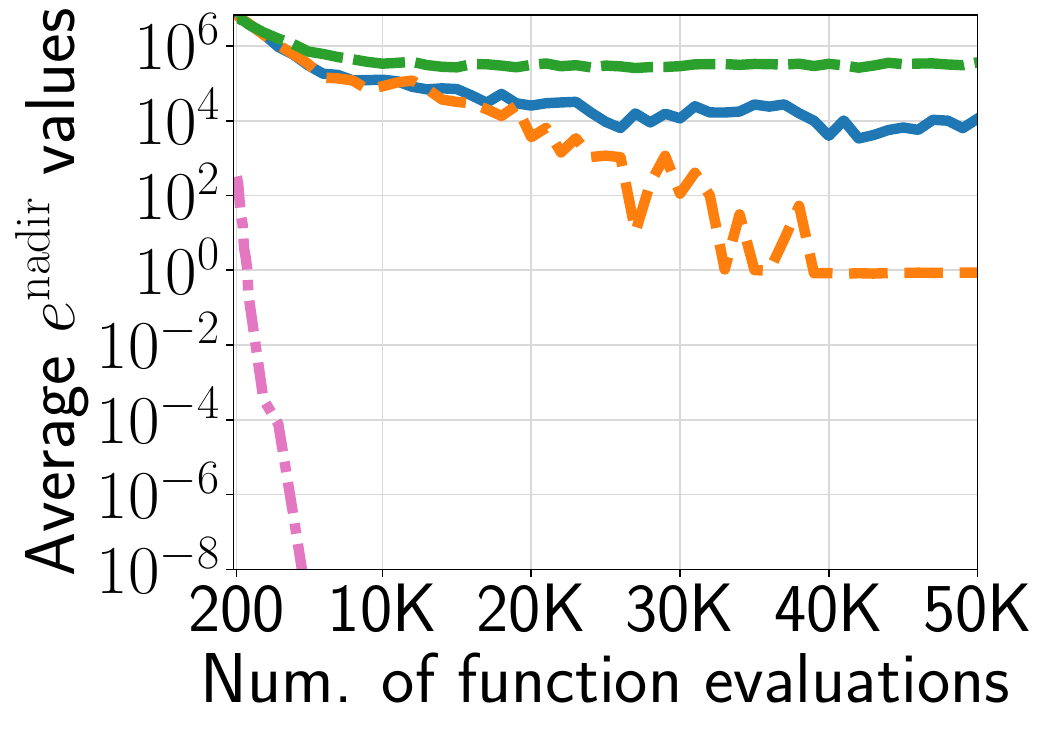}}
   \subfloat[$e^{\mathrm{nadir}}$ ($m=6$)]{\includegraphics[width=0.32\textwidth]{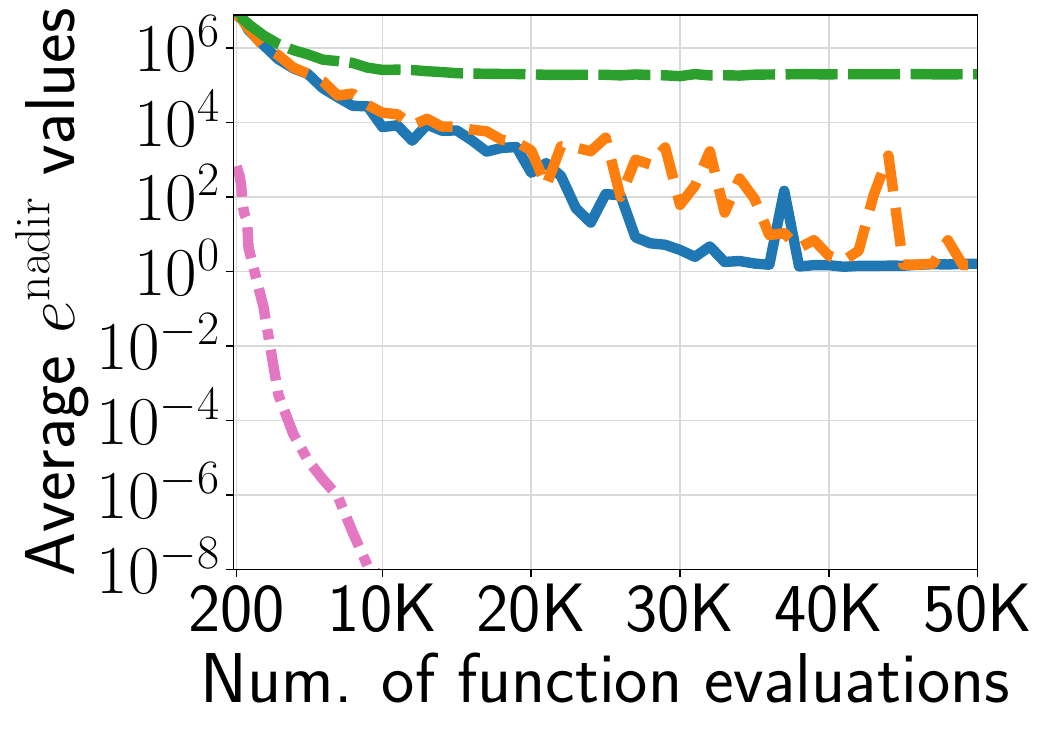}}
\\
   \subfloat[ORE ($m=2$)]{\includegraphics[width=0.32\textwidth]{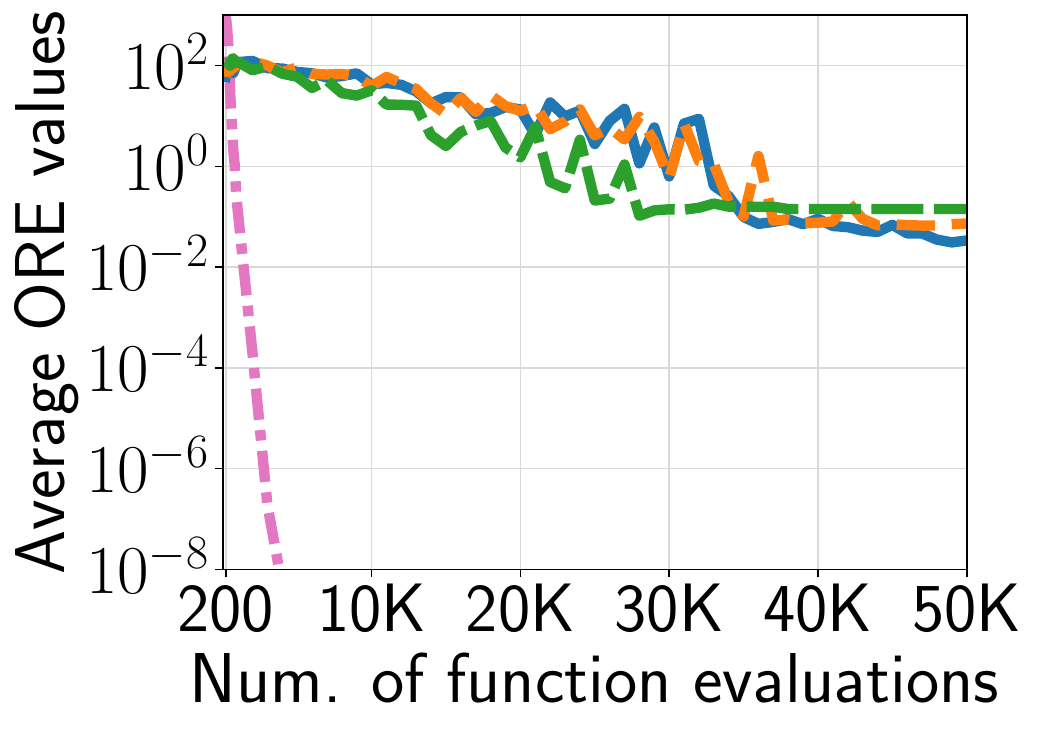}}
   \subfloat[ORE ($m=4$)]{\includegraphics[width=0.32\textwidth]{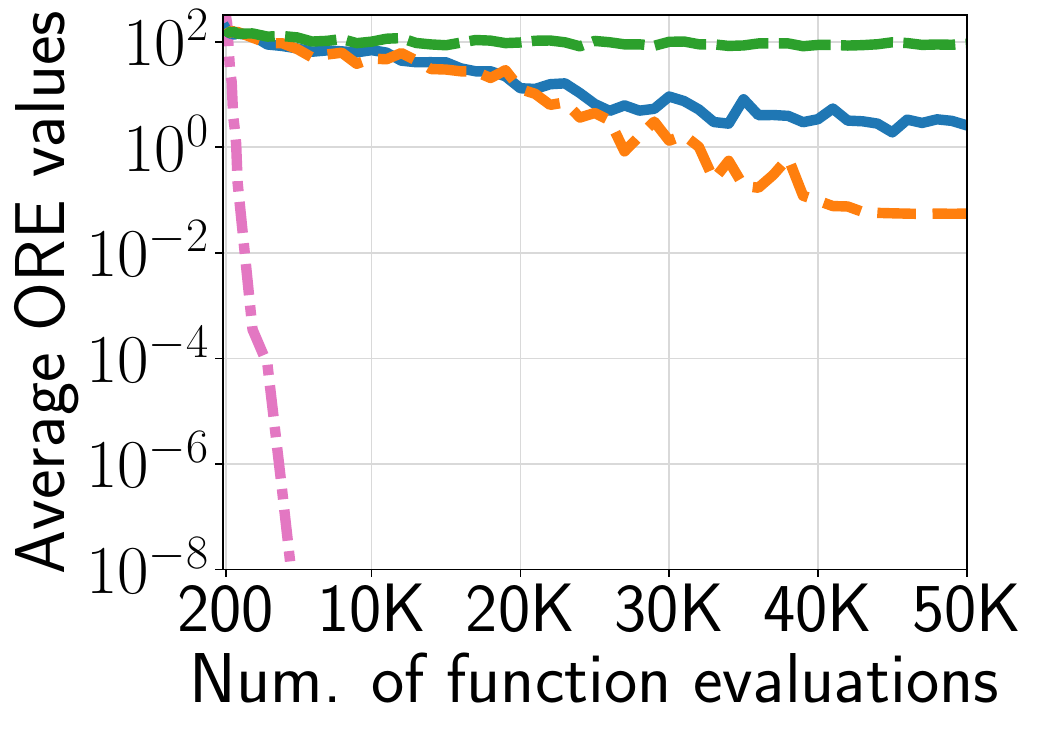}}
   \subfloat[ORE ($m=6$)]{\includegraphics[width=0.32\textwidth]{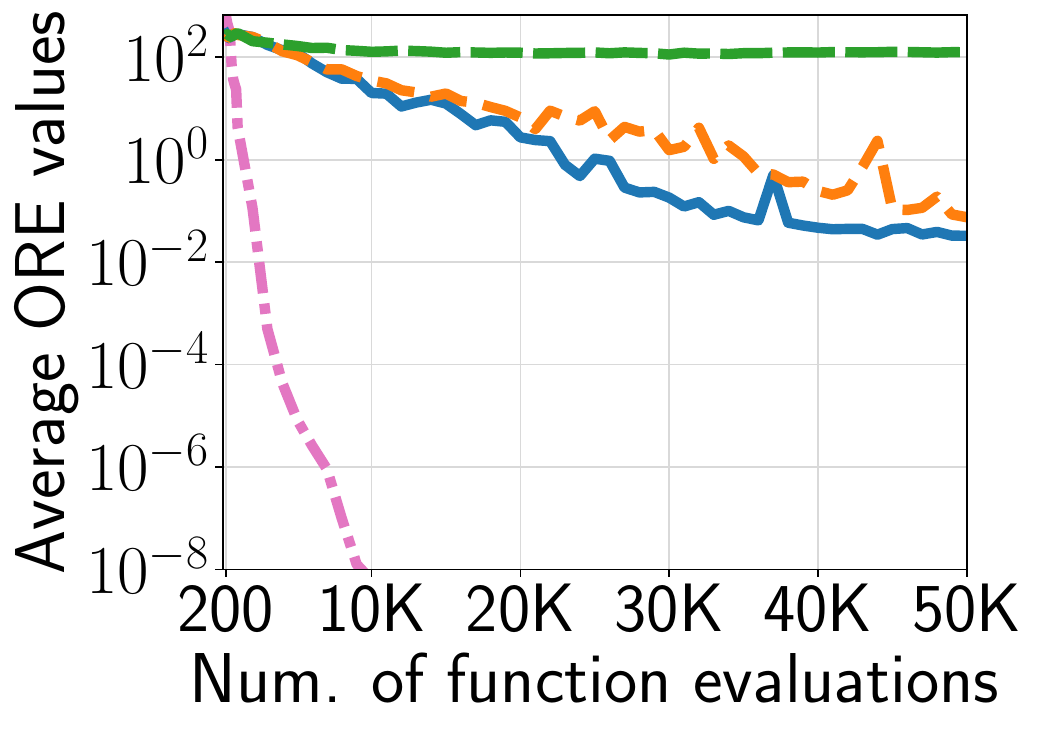}}
\\
\caption{Average $e^{\mathrm{ideal}}$, $e^{\mathrm{nadir}}$, and ORE values of the three normalization methods in R-NSGA-II on SDTLZ3.}
\label{supfig:3error_RNSGA2_SDTLZ3}
\end{figure*}

\begin{figure*}[t]
\centering
  \subfloat{\includegraphics[width=0.7\textwidth]{./figs/legend/legend_3.pdf}}
\vspace{-3.9mm}
   \\
   \subfloat[$e^{\mathrm{ideal}}$ ($m=2$)]{\includegraphics[width=0.32\textwidth]{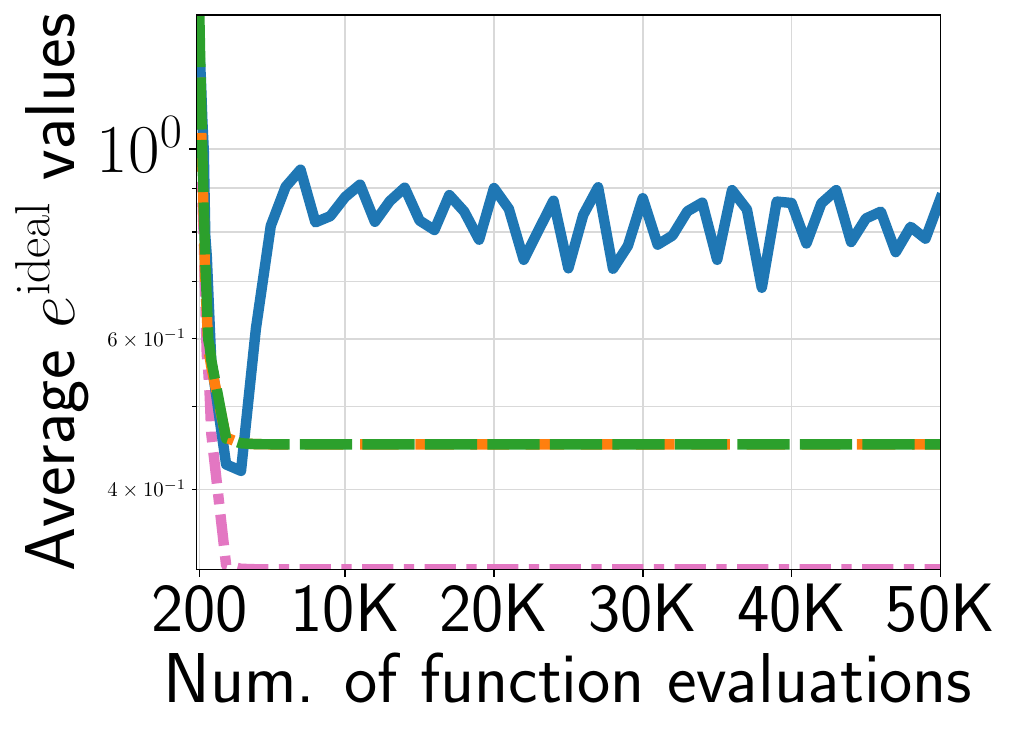}}
   \subfloat[$e^{\mathrm{ideal}}$ ($m=4$)]{\includegraphics[width=0.32\textwidth]{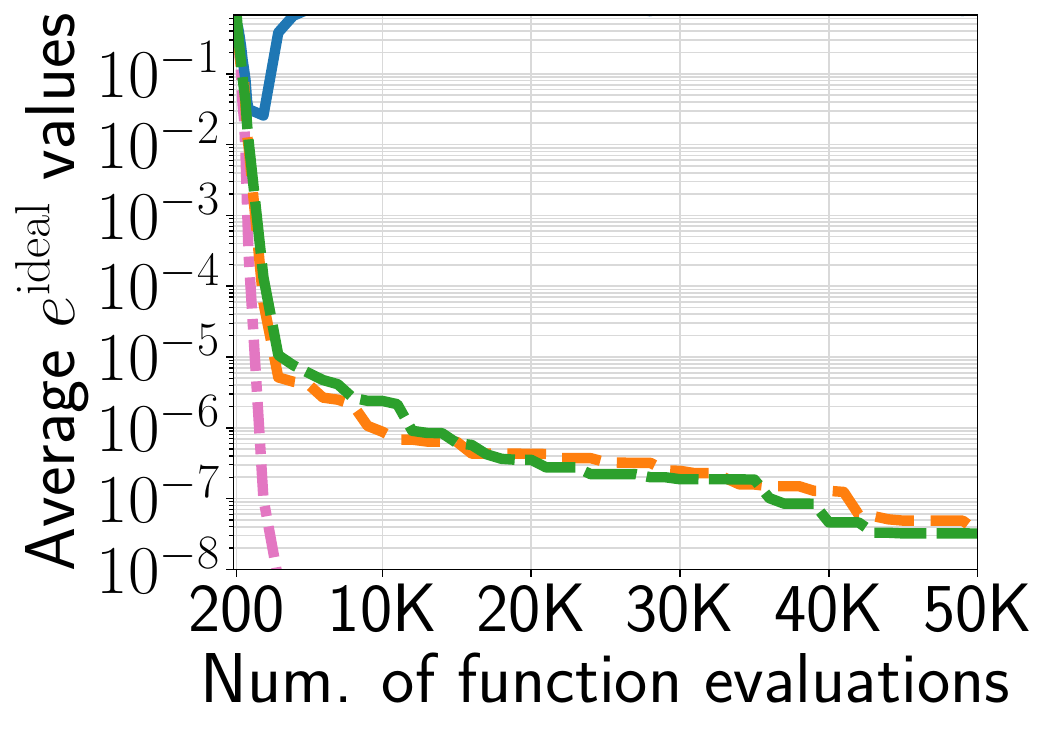}}
   \subfloat[$e^{\mathrm{ideal}}$ ($m=6$)]{\includegraphics[width=0.32\textwidth]{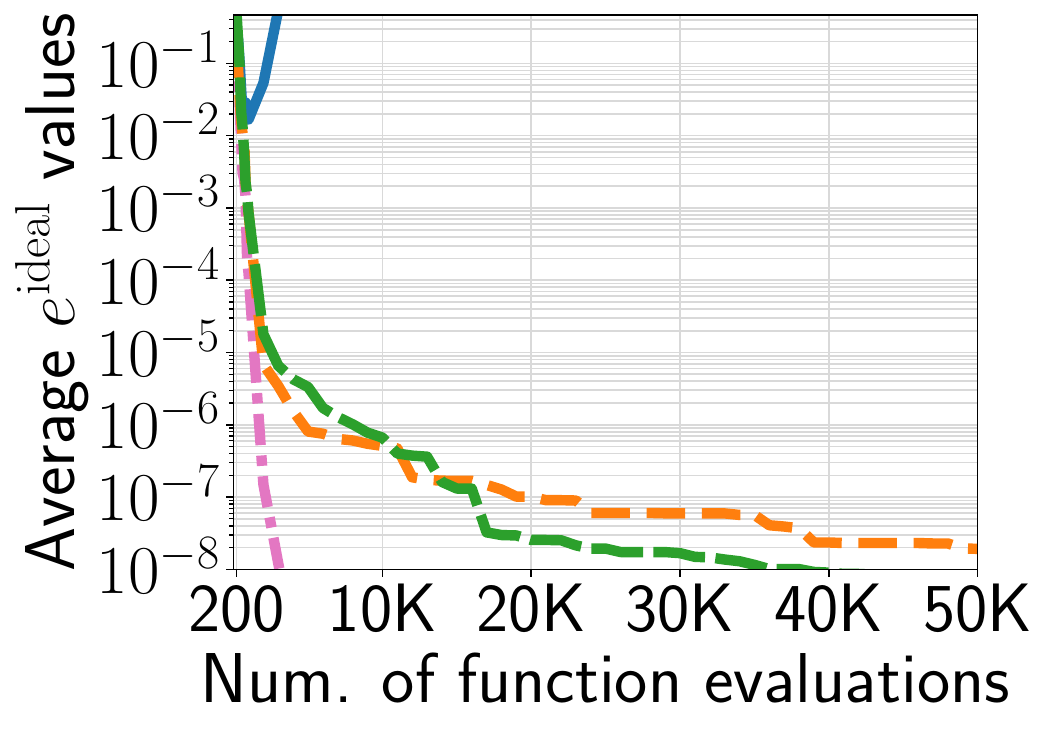}}
\\
   \subfloat[$e^{\mathrm{nadir}}$ ($m=2$)]{\includegraphics[width=0.32\textwidth]{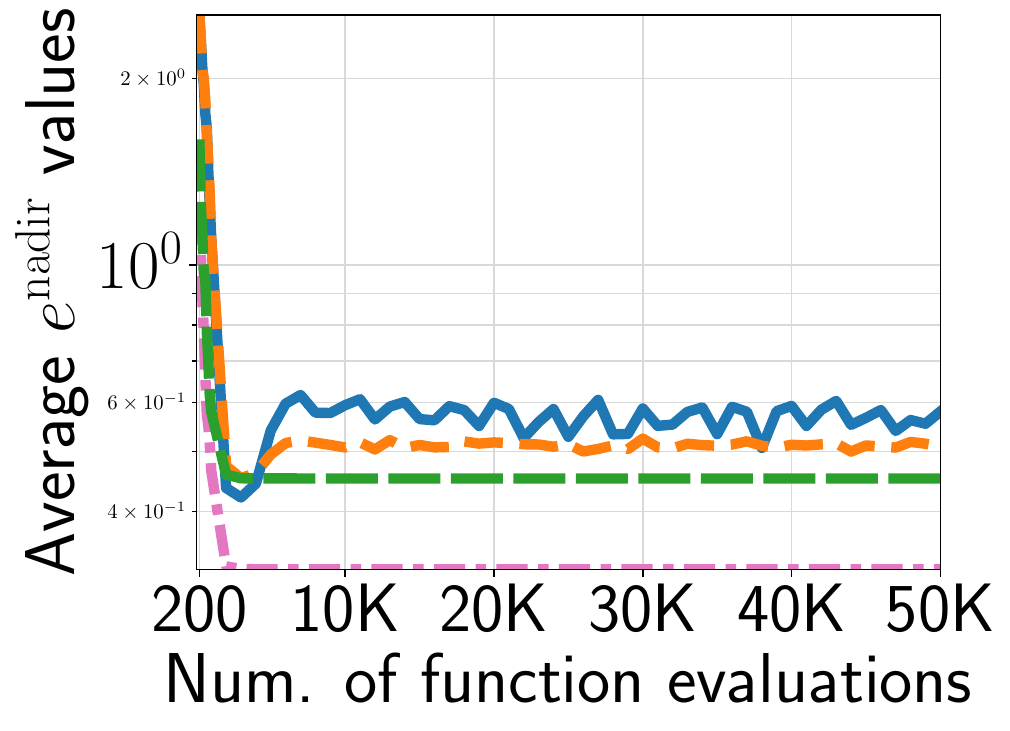}}
   \subfloat[$e^{\mathrm{nadir}}$ ($m=4$)]{\includegraphics[width=0.32\textwidth]{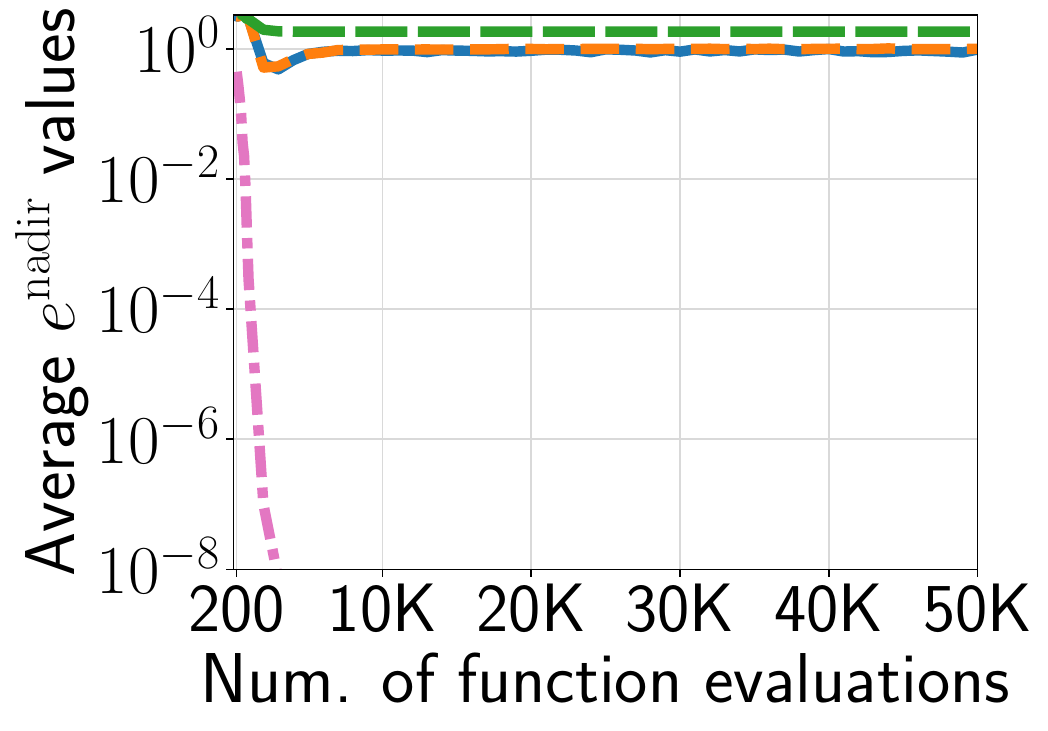}}
   \subfloat[$e^{\mathrm{nadir}}$ ($m=6$)]{\includegraphics[width=0.32\textwidth]{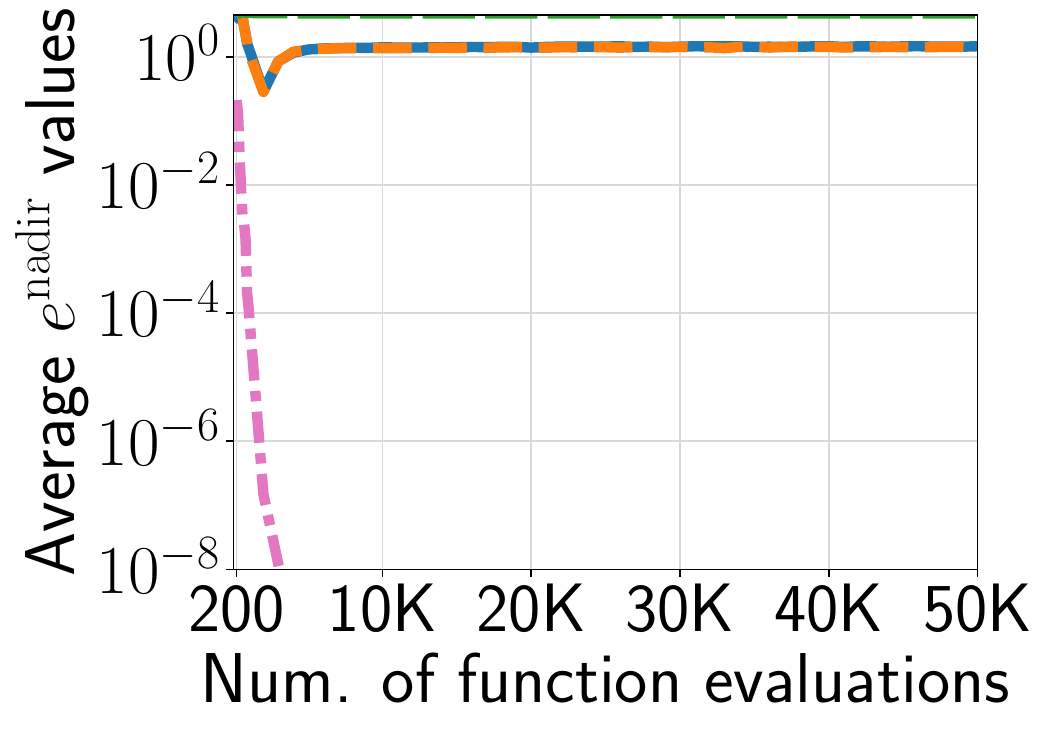}}
\\
   \subfloat[ORE ($m=2$)]{\includegraphics[width=0.32\textwidth]{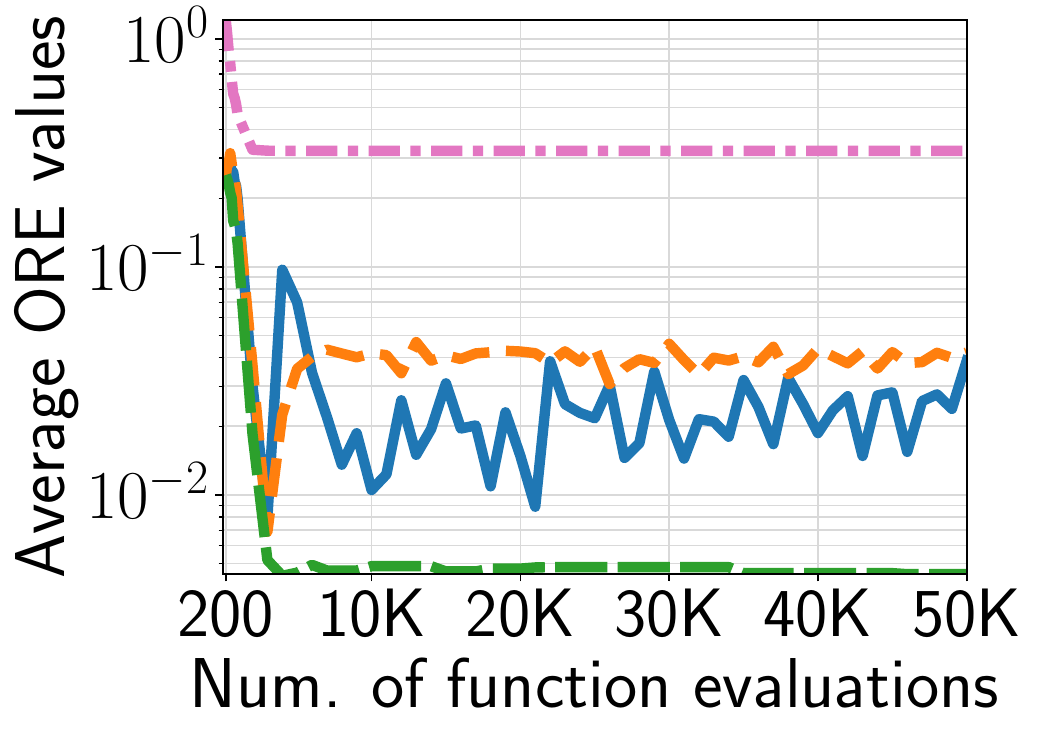}}
   \subfloat[ORE ($m=4$)]{\includegraphics[width=0.32\textwidth]{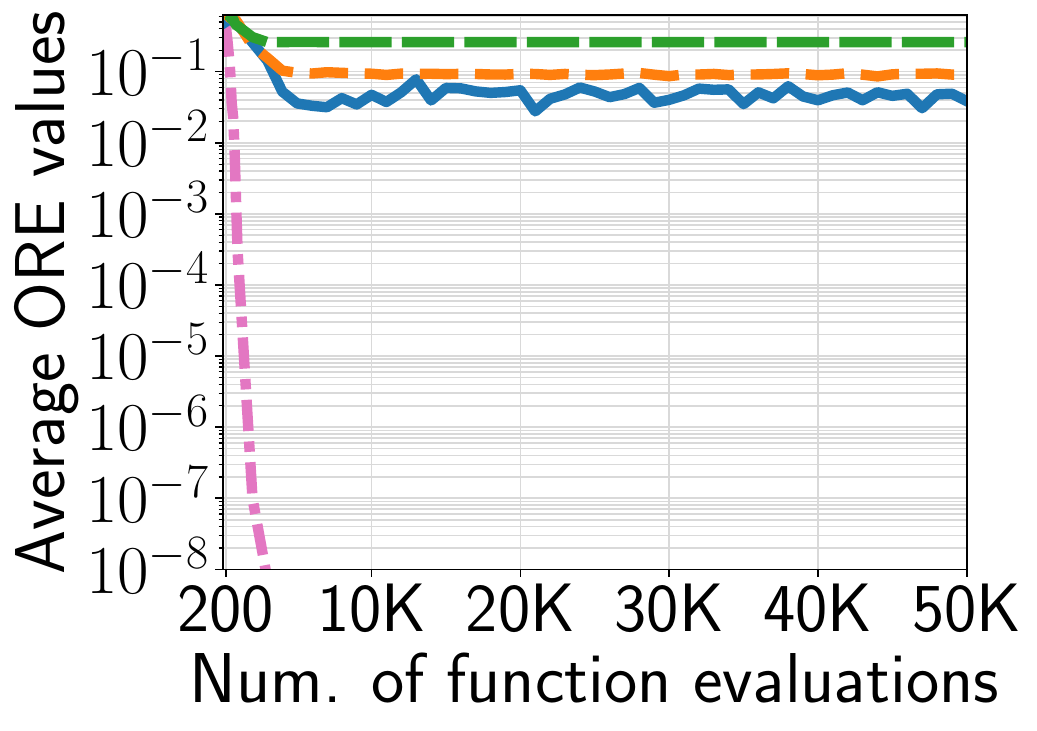}}
   \subfloat[ORE ($m=6$)]{\includegraphics[width=0.32\textwidth]{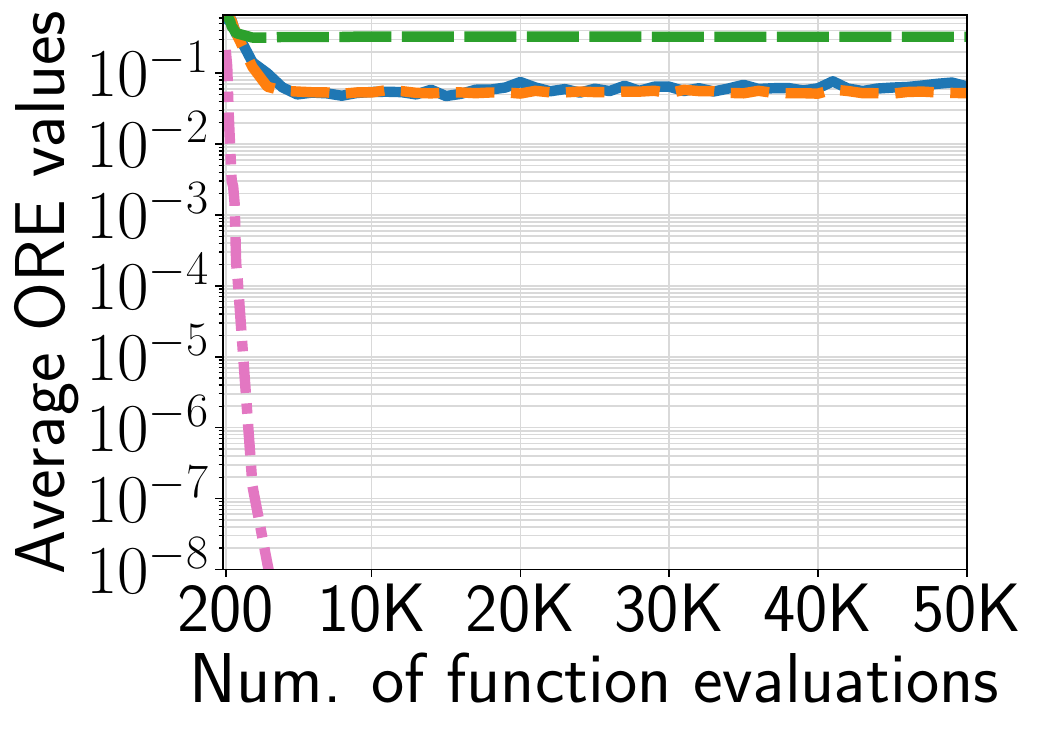}}
\\
\caption{Average $e^{\mathrm{ideal}}$, $e^{\mathrm{nadir}}$, and ORE values of the three normalization methods in R-NSGA-II on SDTLZ4.}
\label{supfig:3error_RNSGA2_SDTLZ4}
\end{figure*}

\begin{figure*}[t]
\centering
  \subfloat{\includegraphics[width=0.7\textwidth]{./figs/legend/legend_3.pdf}}
\vspace{-3.9mm}
   \\
   \subfloat[$e^{\mathrm{ideal}}$ ($m=2$)]{\includegraphics[width=0.32\textwidth]{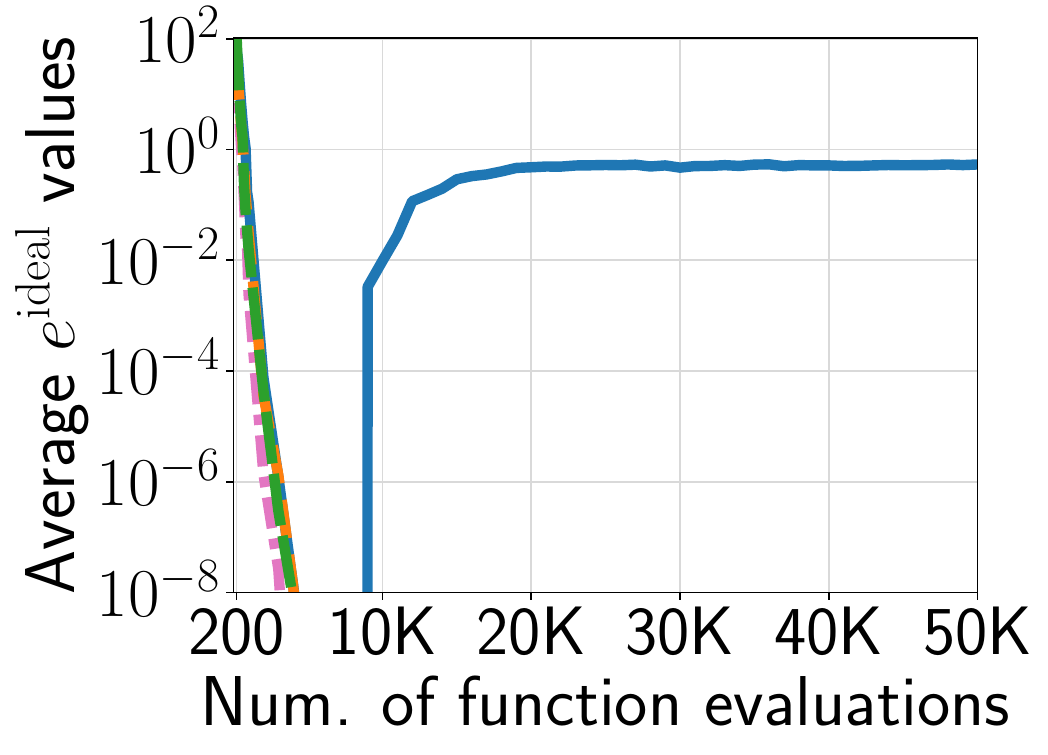}}
   \subfloat[$e^{\mathrm{ideal}}$ ($m=4$)]{\includegraphics[width=0.32\textwidth]{./figs/qi_error_ideal/RNSGA2_mu100/IDTLZ1_m4_r0.1_z-type1.pdf}}
   \subfloat[$e^{\mathrm{ideal}}$ ($m=6$)]{\includegraphics[width=0.32\textwidth]{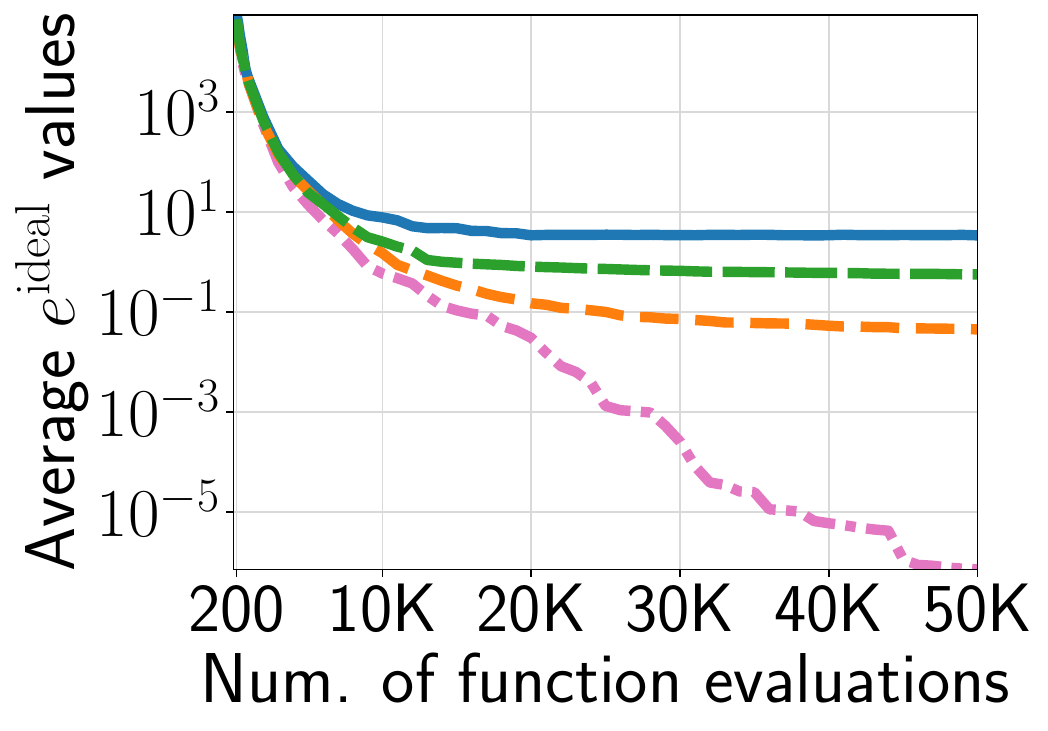}}
\\
   \subfloat[$e^{\mathrm{nadir}}$ ($m=2$)]{\includegraphics[width=0.32\textwidth]{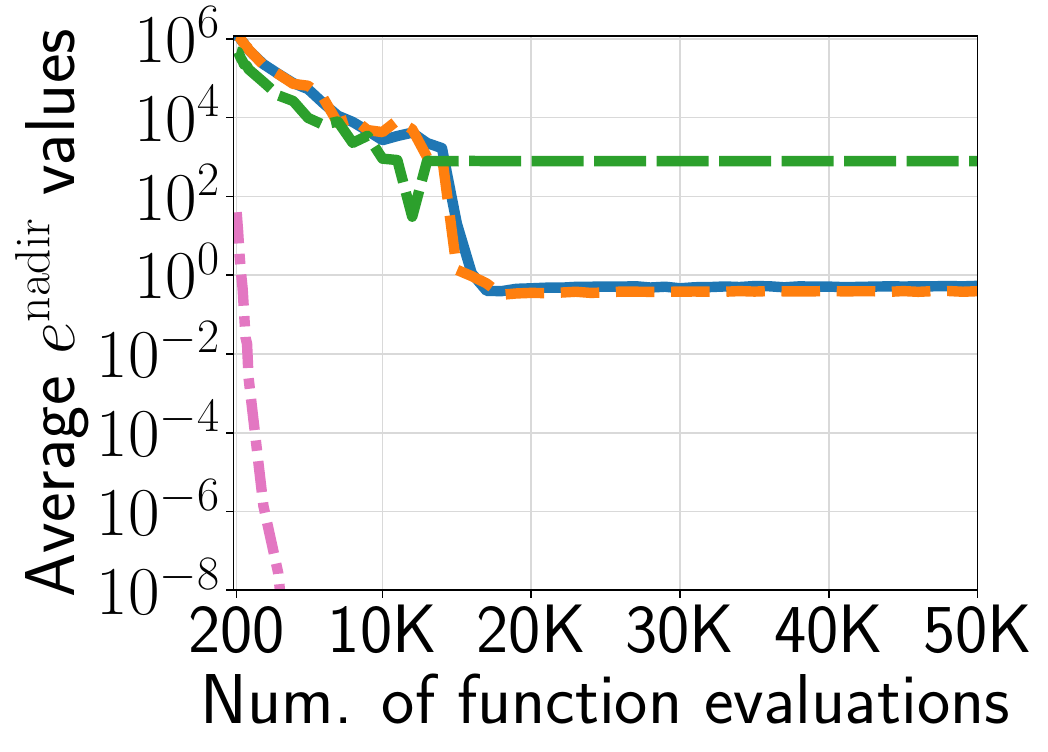}}
   \subfloat[$e^{\mathrm{nadir}}$ ($m=4$)]{\includegraphics[width=0.32\textwidth]{./figs/qi_error_nadir/RNSGA2_mu100/IDTLZ1_m4_r0.1_z-type1.pdf}}
   \subfloat[$e^{\mathrm{nadir}}$ ($m=6$)]{\includegraphics[width=0.32\textwidth]{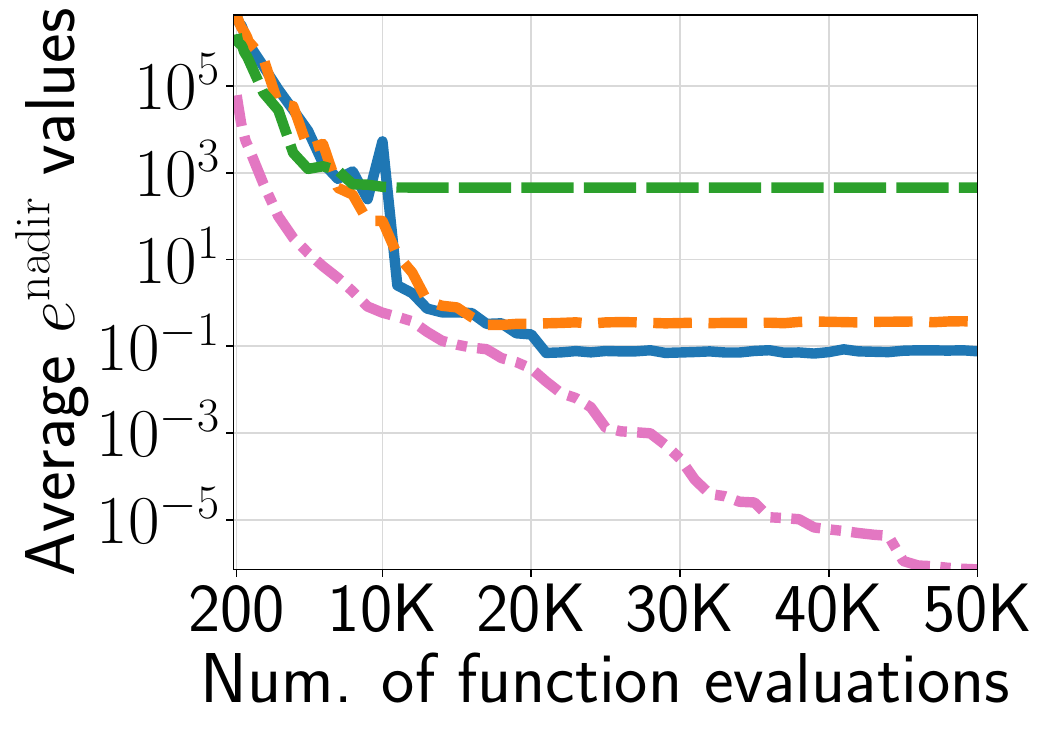}}
\\
   \subfloat[ORE ($m=2$)]{\includegraphics[width=0.32\textwidth]{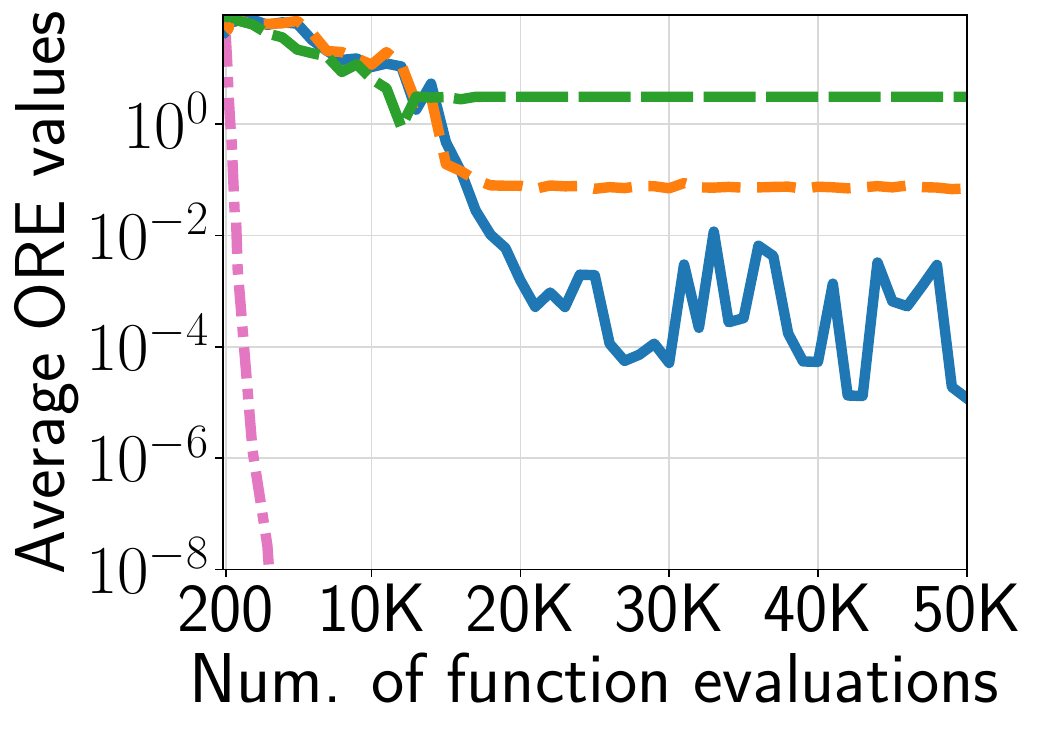}}
   \subfloat[ORE ($m=4$)]{\includegraphics[width=0.32\textwidth]{./figs/qi_ore/RNSGA2_mu100/IDTLZ1_m4_r0.1_z-type1.pdf}}
   \subfloat[ORE ($m=6$)]{\includegraphics[width=0.32\textwidth]{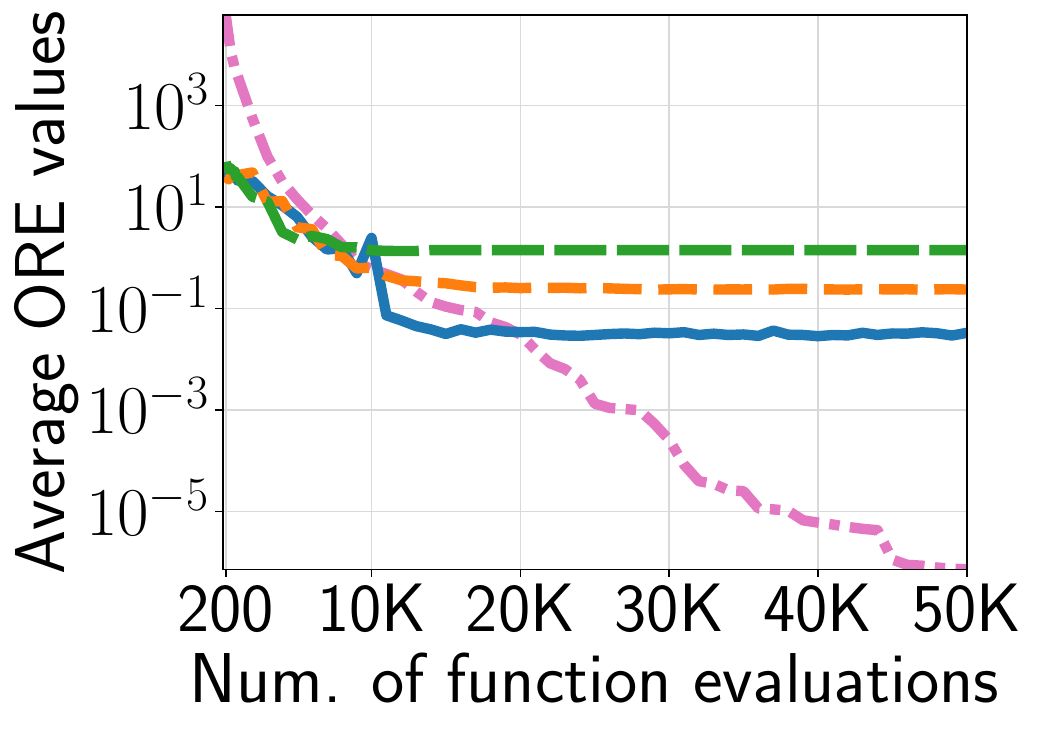}}
\\
\caption{Average $e^{\mathrm{ideal}}$, $e^{\mathrm{nadir}}$, and ORE values of the three normalization methods in R-NSGA-II on IDTLZ1.}
\label{supfig:3error_RNSGA2_IDTLZ1}
\end{figure*}

\begin{figure*}[t]
\centering
  \subfloat{\includegraphics[width=0.7\textwidth]{./figs/legend/legend_3.pdf}}
\vspace{-3.9mm}
   \\
   \subfloat[$e^{\mathrm{ideal}}$ ($m=2$)]{\includegraphics[width=0.32\textwidth]{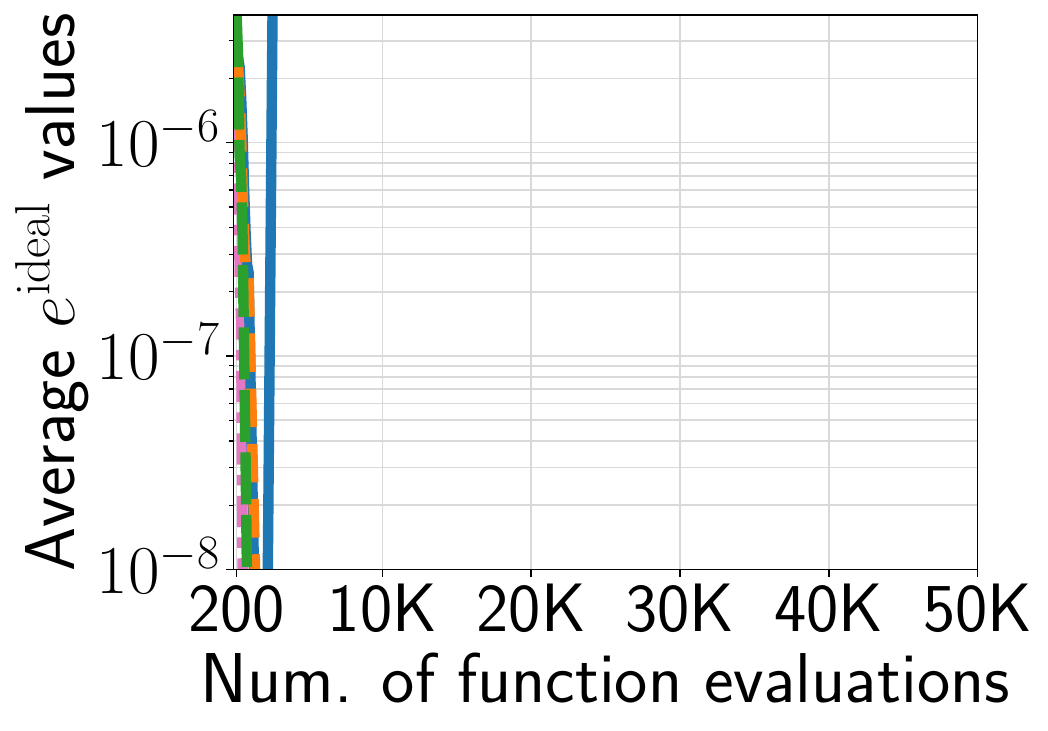}}
   \subfloat[$e^{\mathrm{ideal}}$ ($m=4$)]{\includegraphics[width=0.32\textwidth]{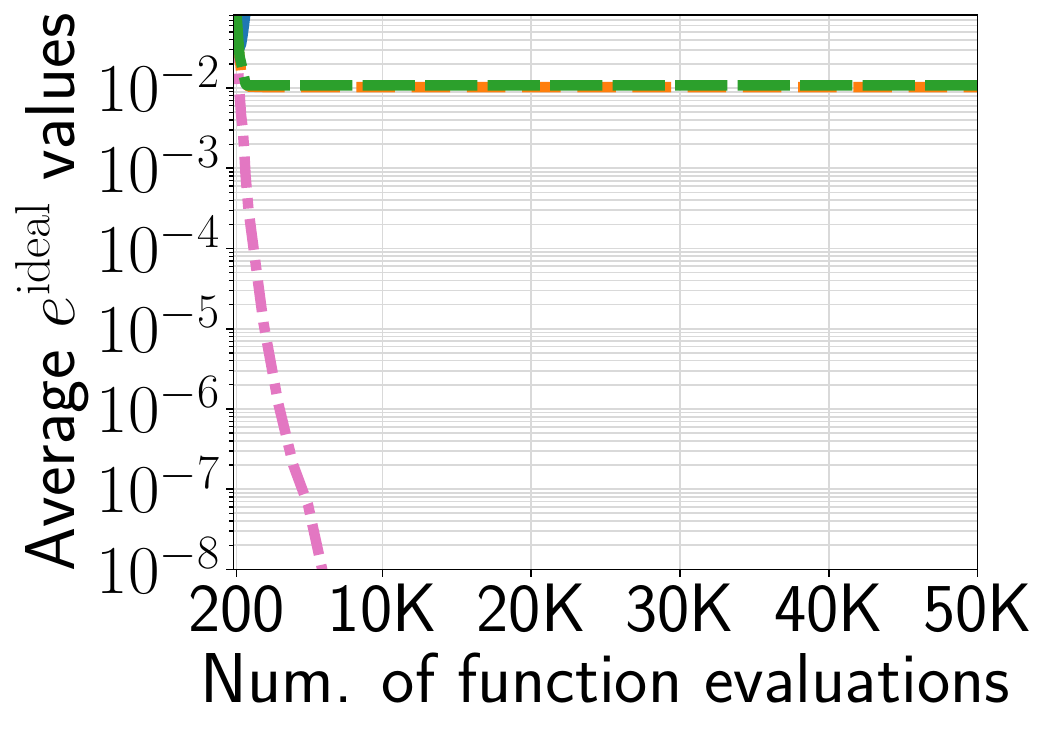}}
   \subfloat[$e^{\mathrm{ideal}}$ ($m=6$)]{\includegraphics[width=0.32\textwidth]{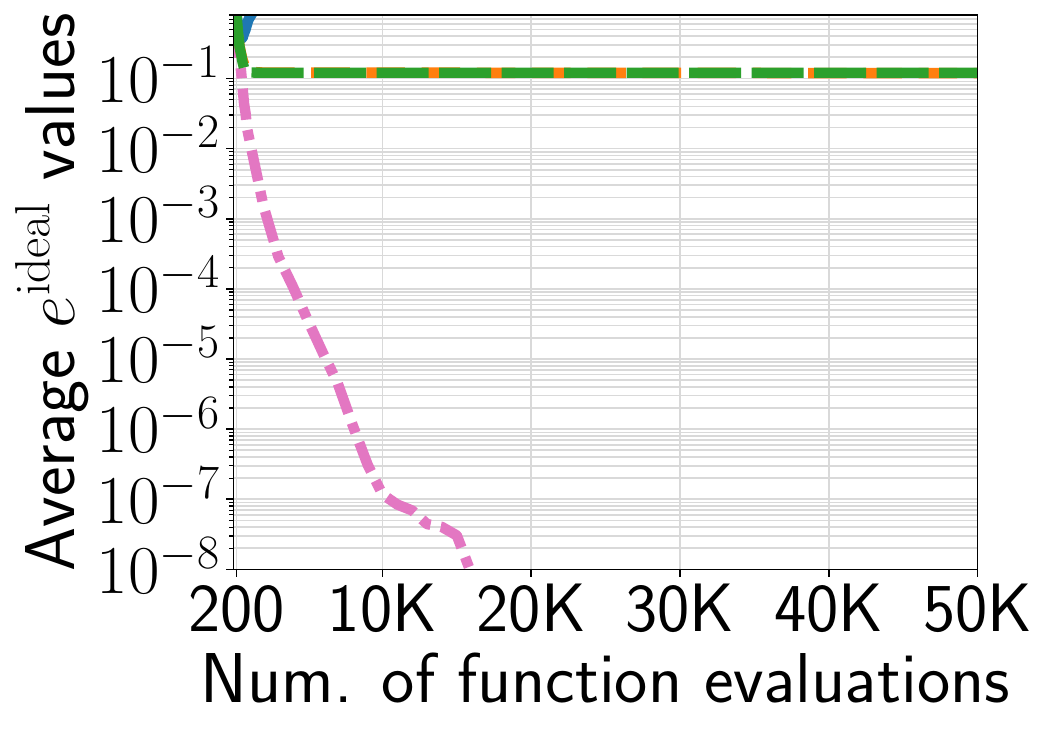}}
\\
   \subfloat[$e^{\mathrm{nadir}}$ ($m=2$)]{\includegraphics[width=0.32\textwidth]{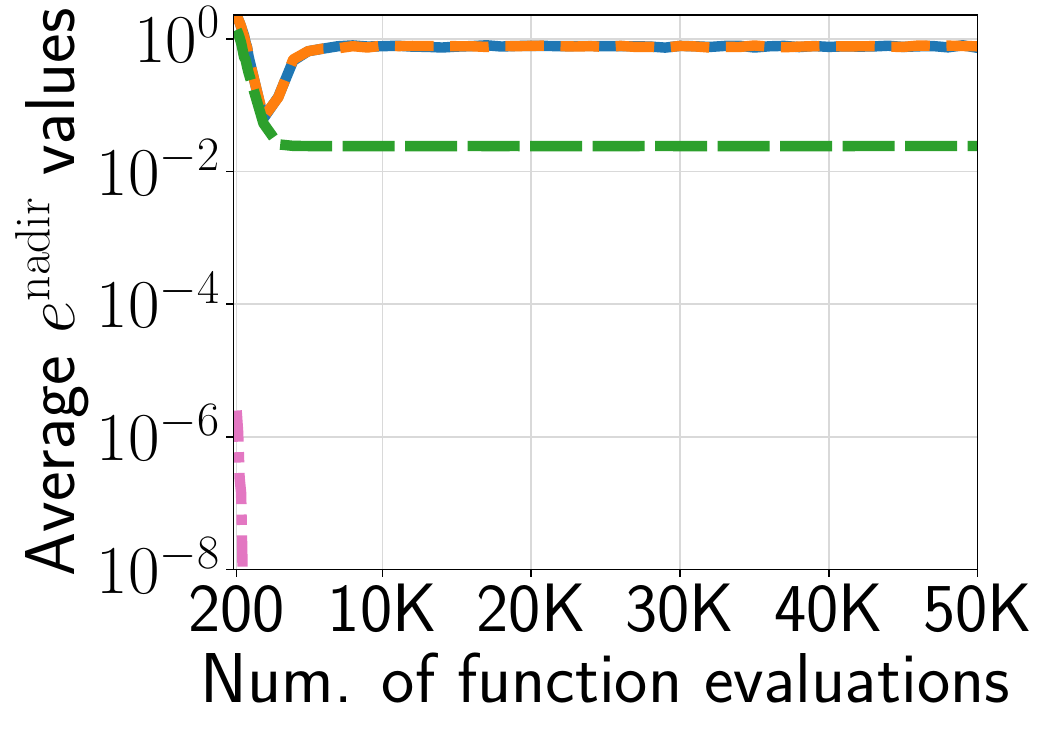}}
   \subfloat[$e^{\mathrm{nadir}}$ ($m=4$)]{\includegraphics[width=0.32\textwidth]{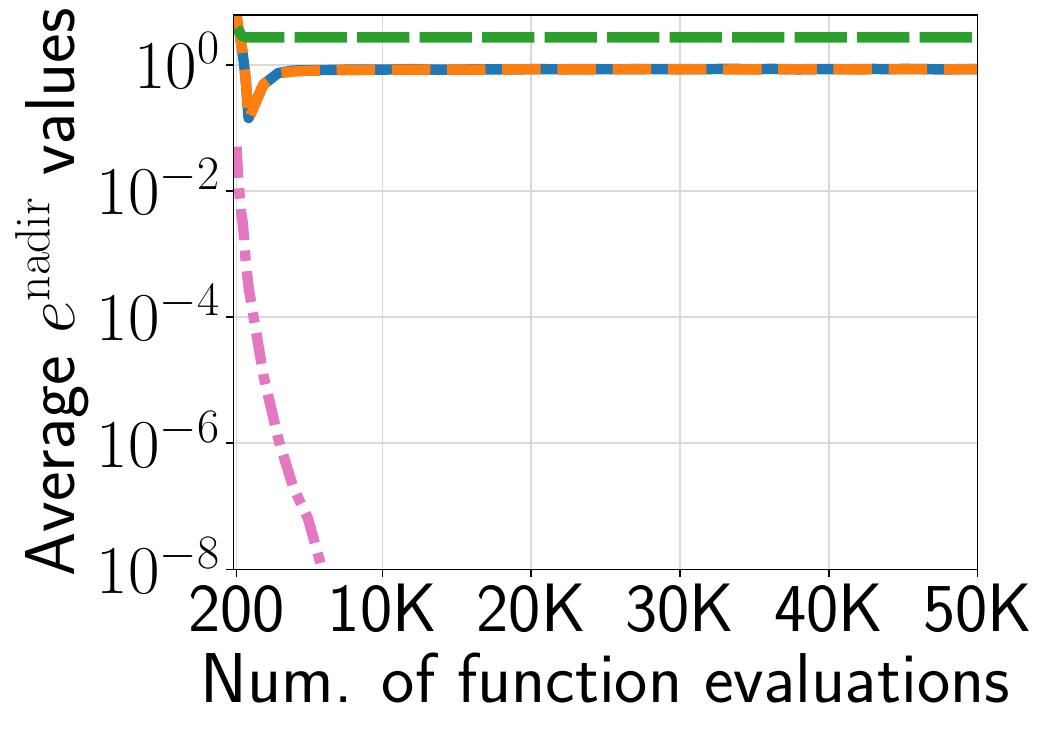}}
   \subfloat[$e^{\mathrm{nadir}}$ ($m=6$)]{\includegraphics[width=0.32\textwidth]{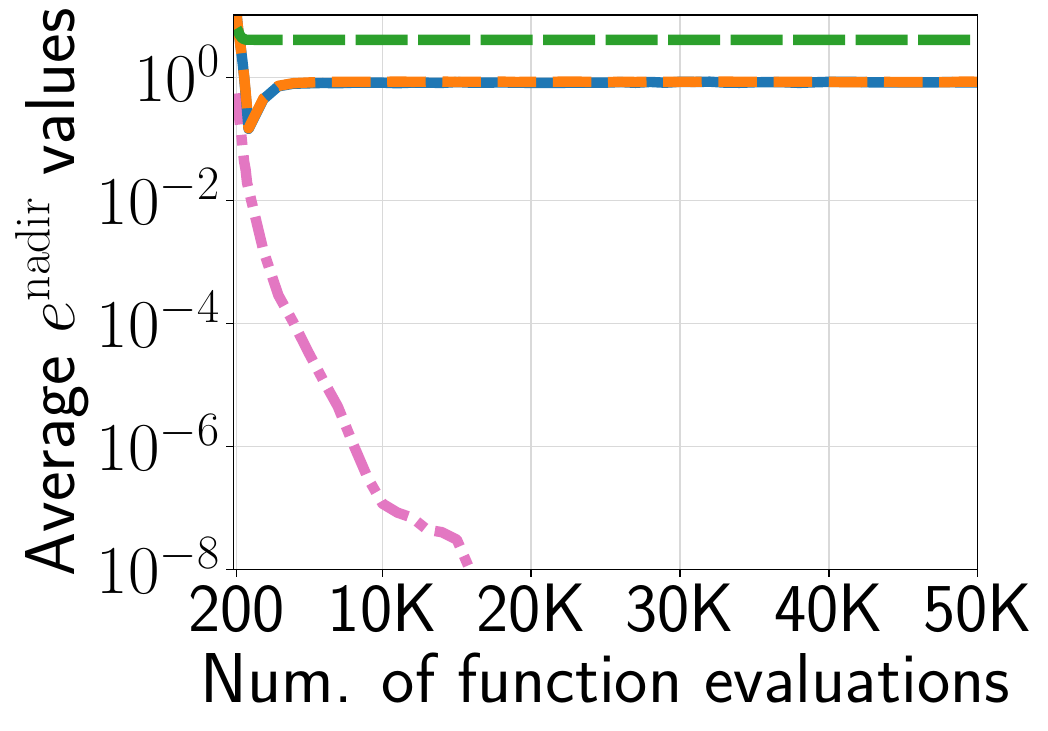}}
\\
   \subfloat[ORE ($m=2$)]{\includegraphics[width=0.32\textwidth]{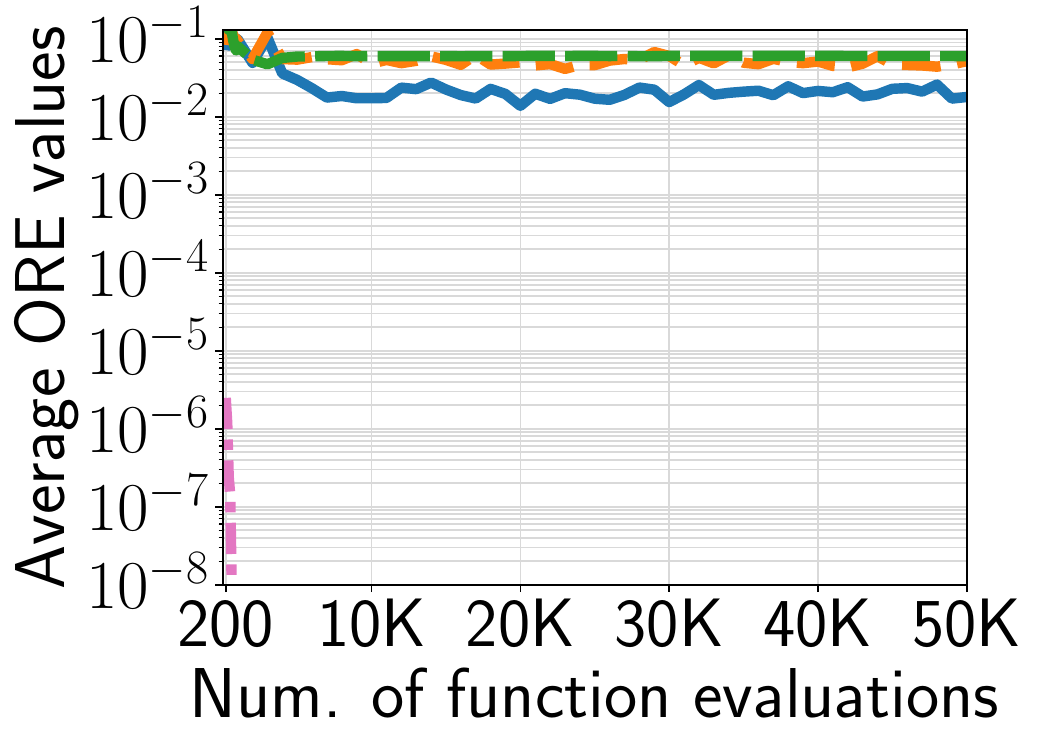}}
   \subfloat[ORE ($m=4$)]{\includegraphics[width=0.32\textwidth]{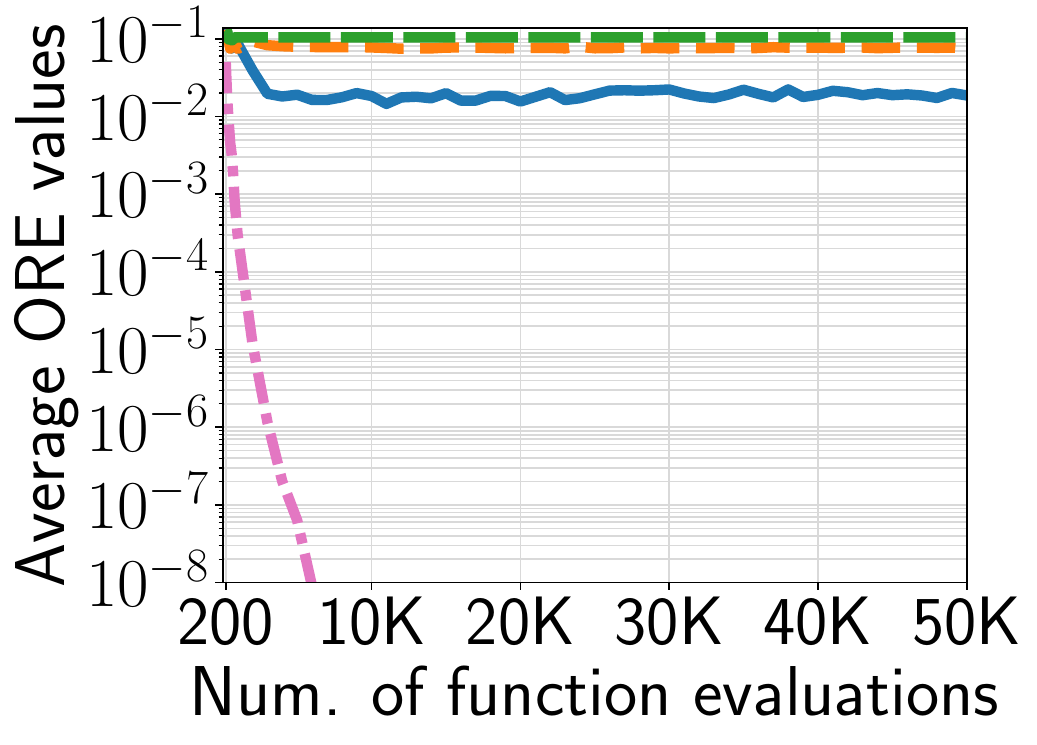}}
   \subfloat[ORE ($m=6$)]{\includegraphics[width=0.32\textwidth]{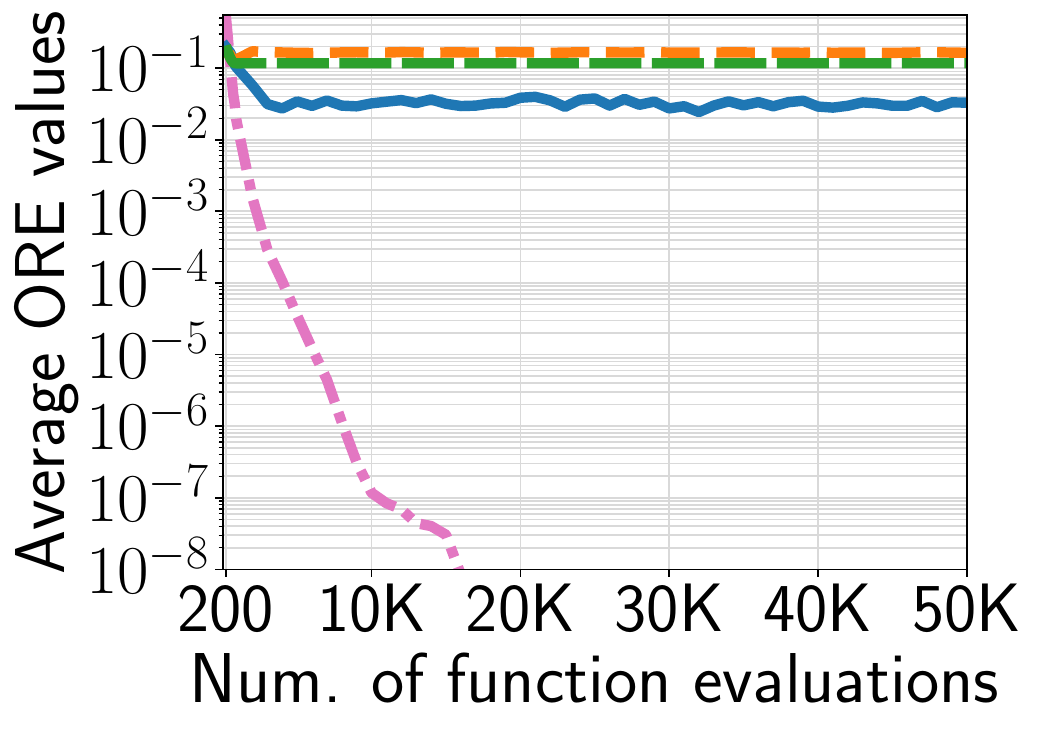}}
\\
\caption{Average $e^{\mathrm{ideal}}$, $e^{\mathrm{nadir}}$, and ORE values of the three normalization methods in R-NSGA-II on IDTLZ2.}
\label{supfig:3error_RNSGA2_IDTLZ2}
\end{figure*}

\begin{figure*}[t]
\centering
  \subfloat{\includegraphics[width=0.7\textwidth]{./figs/legend/legend_3.pdf}}
\vspace{-3.9mm}
   \\
   \subfloat[$e^{\mathrm{ideal}}$ ($m=2$)]{\includegraphics[width=0.32\textwidth]{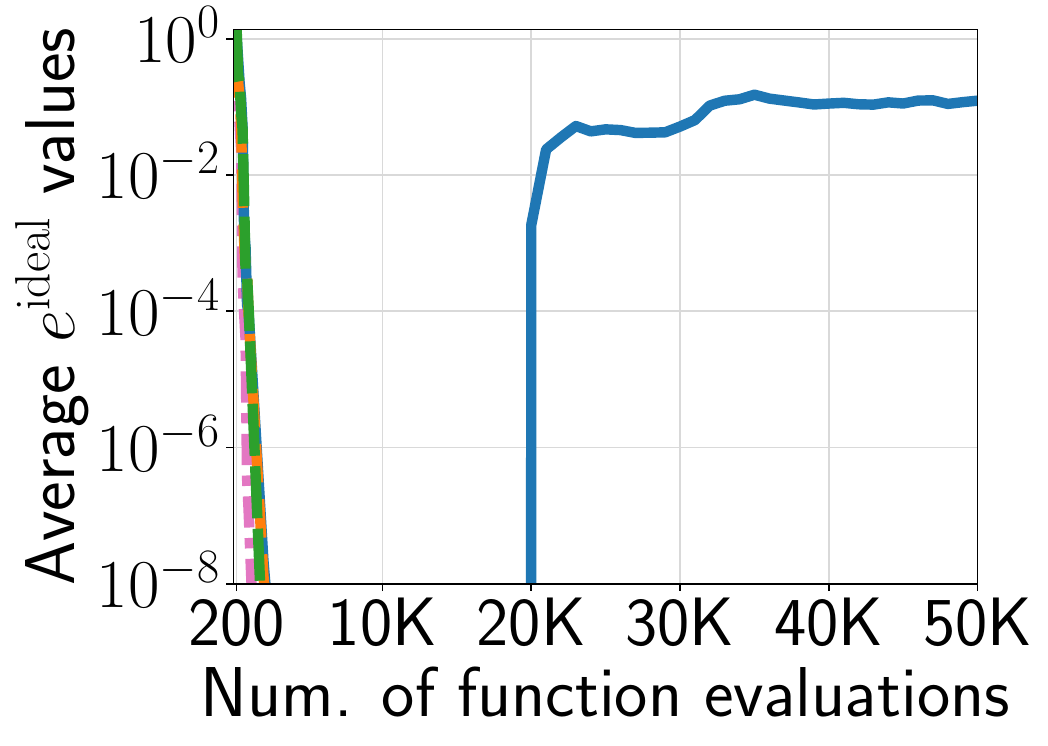}}
   \subfloat[$e^{\mathrm{ideal}}$ ($m=4$)]{\includegraphics[width=0.32\textwidth]{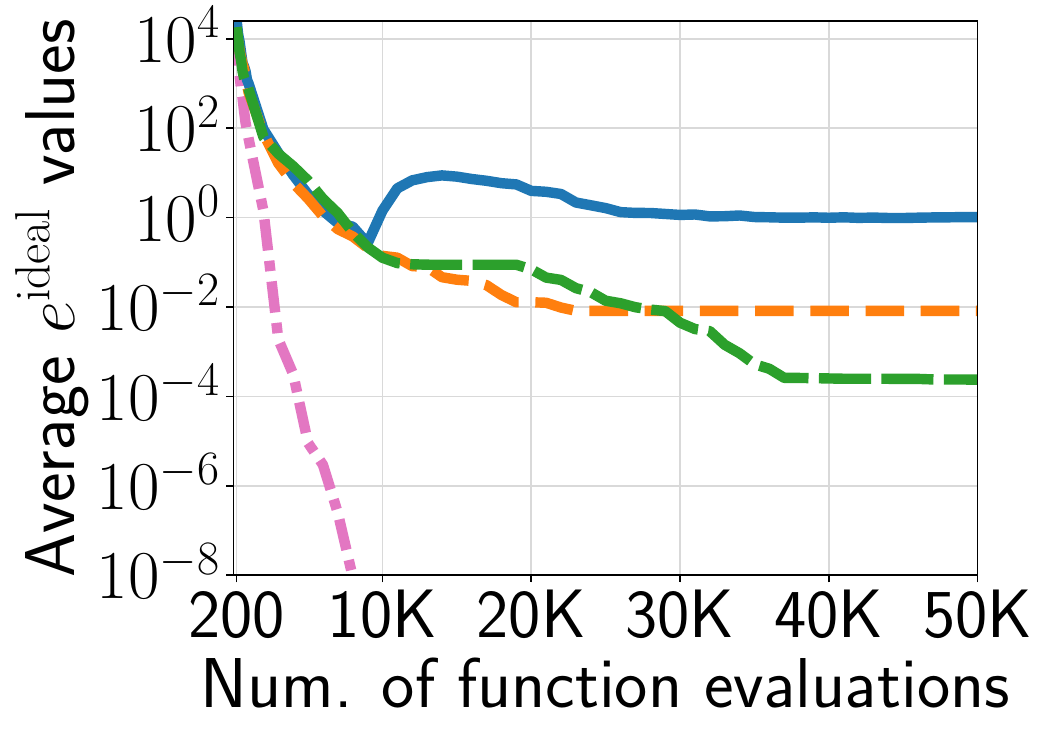}}
   \subfloat[$e^{\mathrm{ideal}}$ ($m=6$)]{\includegraphics[width=0.32\textwidth]{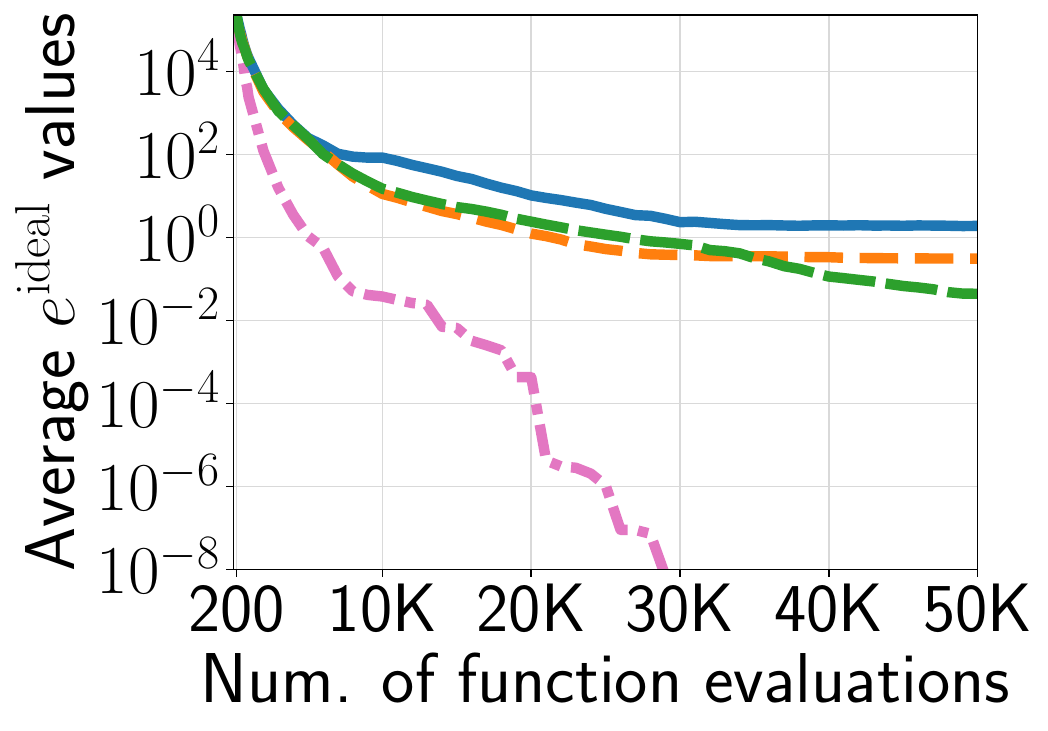}}
\\
   \subfloat[$e^{\mathrm{nadir}}$ ($m=2$)]{\includegraphics[width=0.32\textwidth]{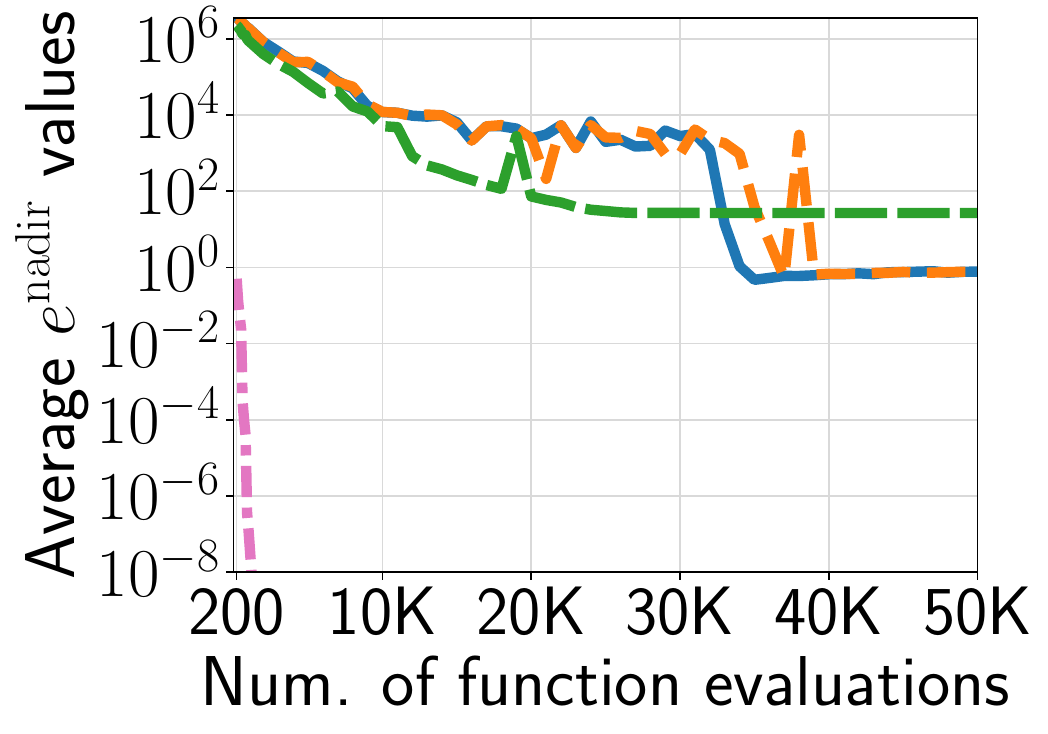}}
   \subfloat[$e^{\mathrm{nadir}}$ ($m=4$)]{\includegraphics[width=0.32\textwidth]{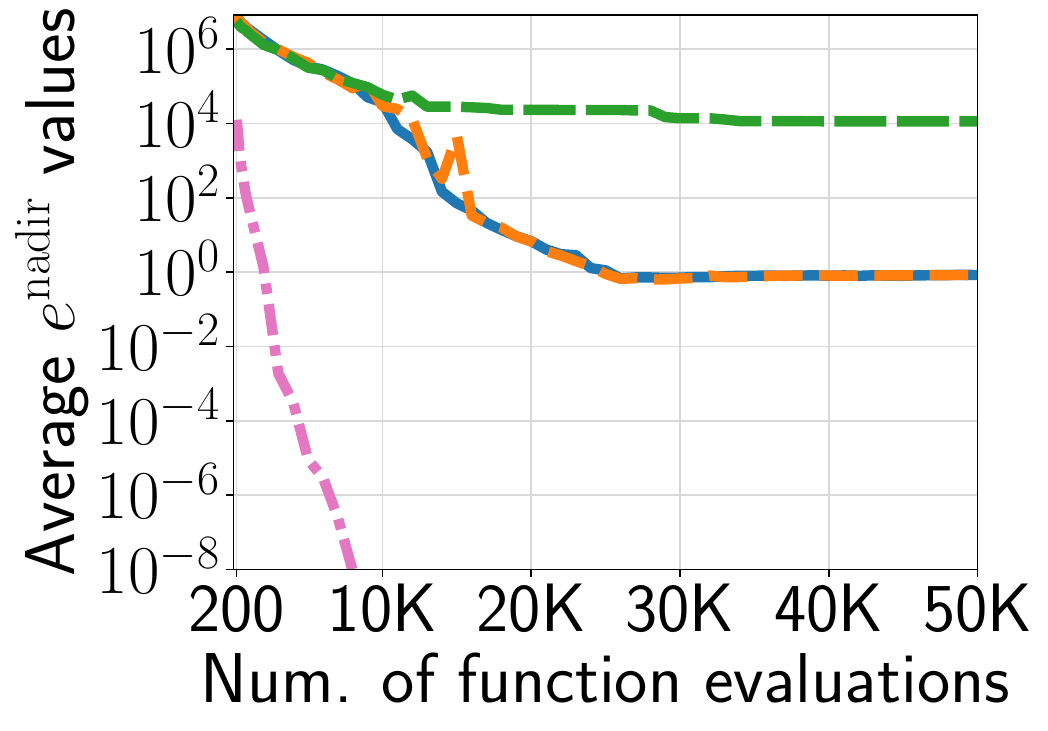}}
   \subfloat[$e^{\mathrm{nadir}}$ ($m=6$)]{\includegraphics[width=0.32\textwidth]{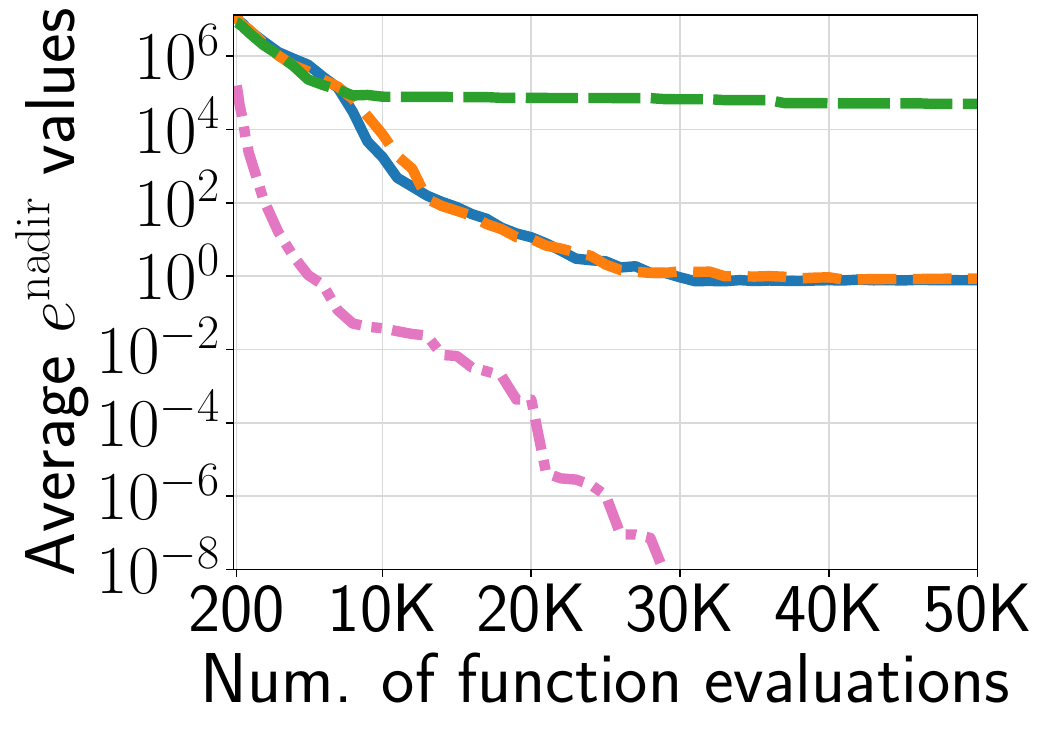}}
\\
   \subfloat[ORE ($m=2$)]{\includegraphics[width=0.32\textwidth]{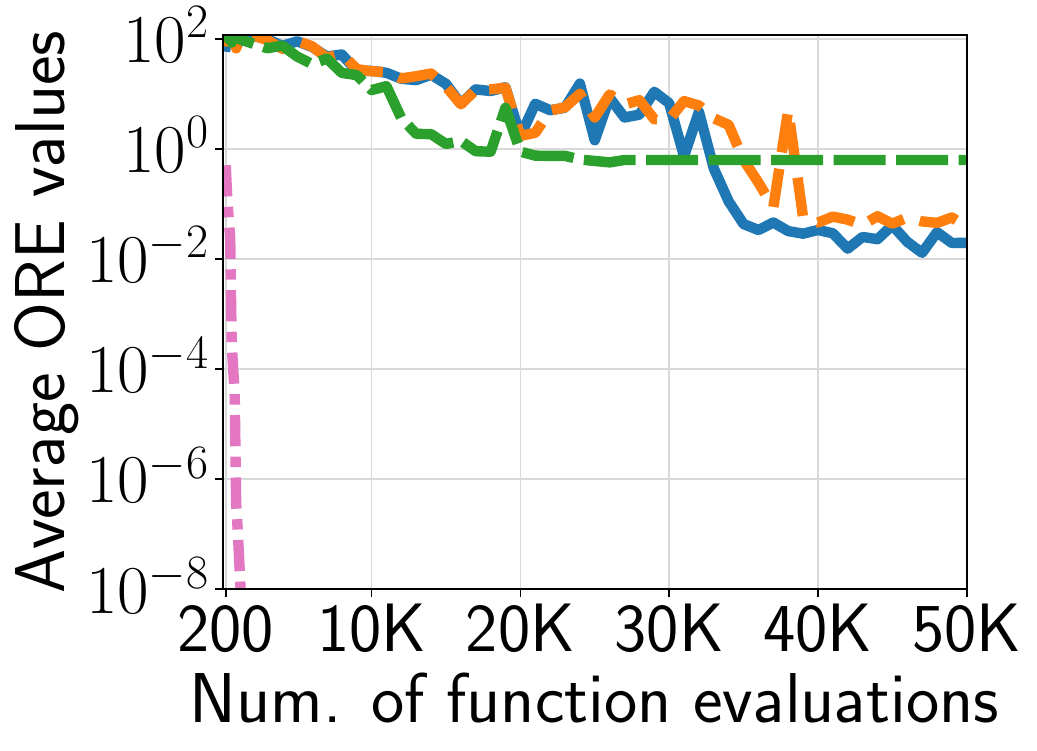}}
   \subfloat[ORE ($m=4$)]{\includegraphics[width=0.32\textwidth]{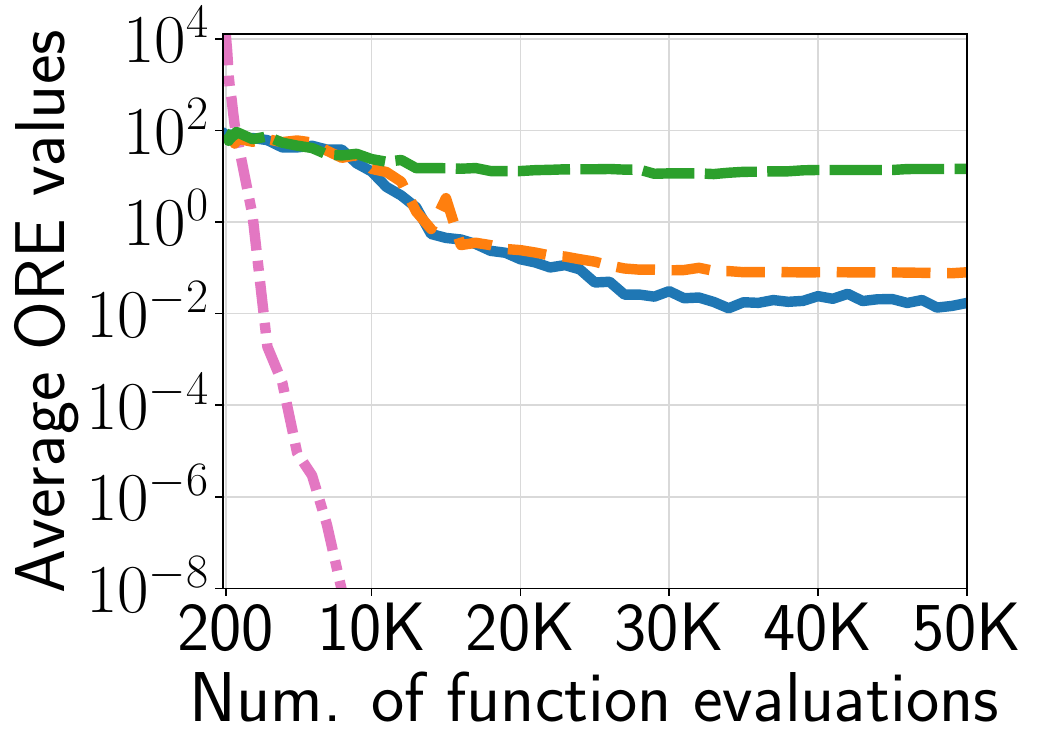}}
   \subfloat[ORE ($m=6$)]{\includegraphics[width=0.32\textwidth]{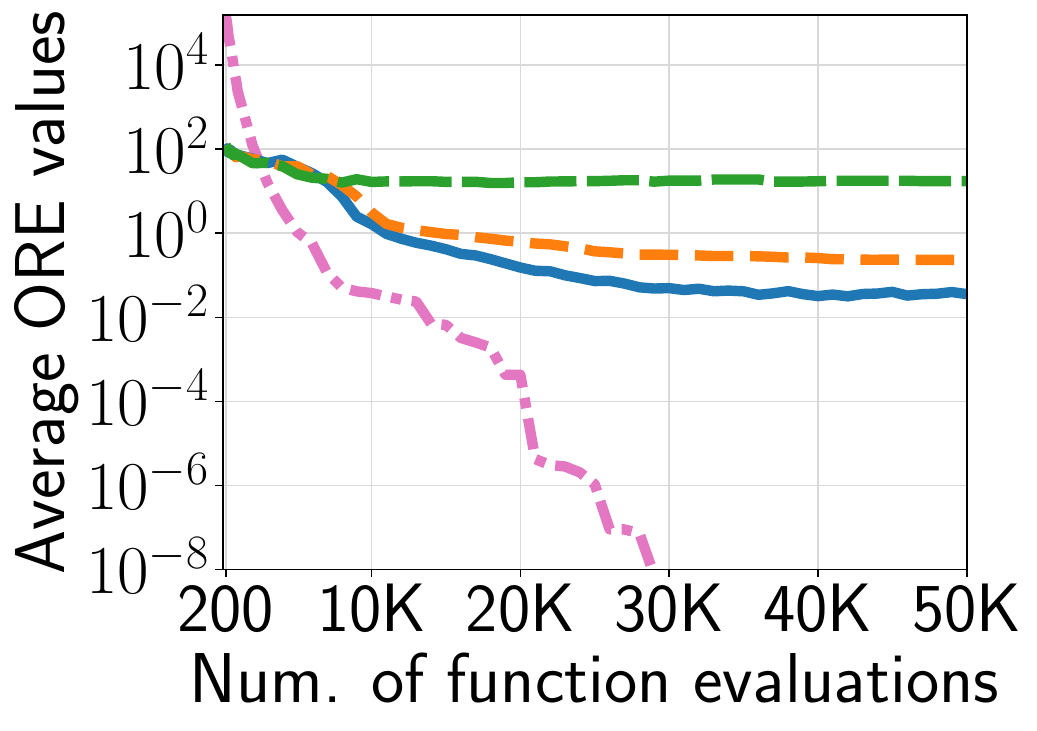}}
\\
\caption{Average $e^{\mathrm{ideal}}$, $e^{\mathrm{nadir}}$, and ORE values of the three normalization methods in R-NSGA-II on IDTLZ3.}
\label{supfig:3error_RNSGA2_IDTLZ3}
\end{figure*}

\begin{figure*}[t]
\centering
  \subfloat{\includegraphics[width=0.7\textwidth]{./figs/legend/legend_3.pdf}}
\vspace{-3.9mm}
   \\
   \subfloat[$e^{\mathrm{ideal}}$ ($m=2$)]{\includegraphics[width=0.32\textwidth]{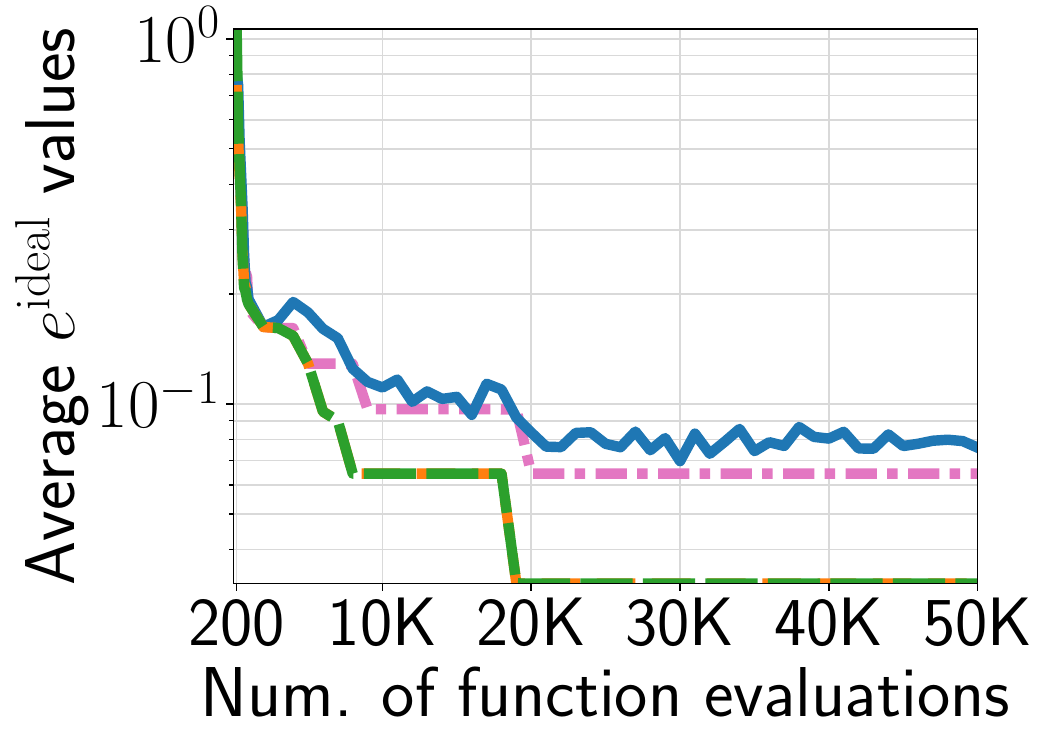}}
   \subfloat[$e^{\mathrm{ideal}}$ ($m=4$)]{\includegraphics[width=0.32\textwidth]{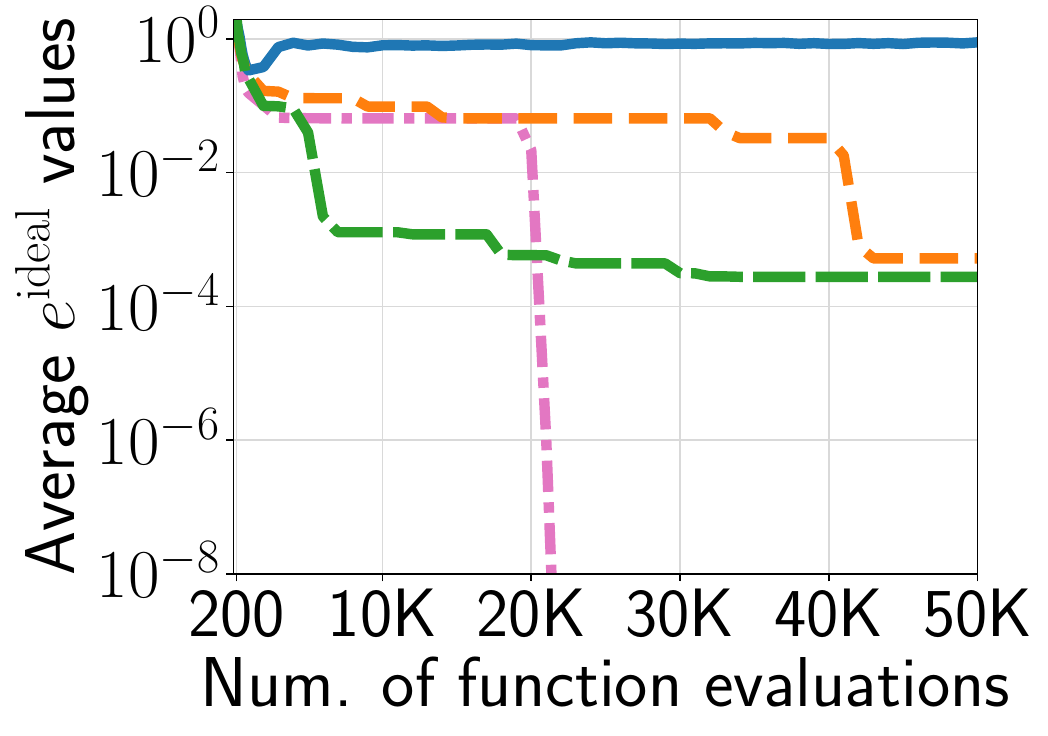}}
   \subfloat[$e^{\mathrm{ideal}}$ ($m=6$)]{\includegraphics[width=0.32\textwidth]{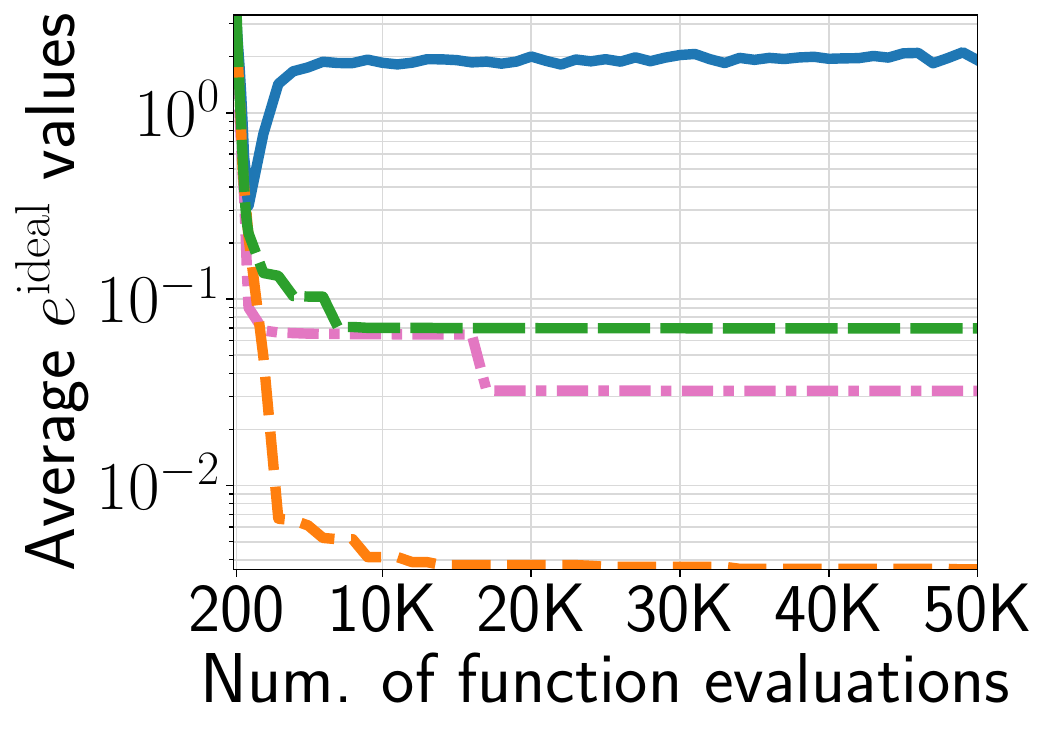}}
\\
   \subfloat[$e^{\mathrm{nadir}}$ ($m=2$)]{\includegraphics[width=0.32\textwidth]{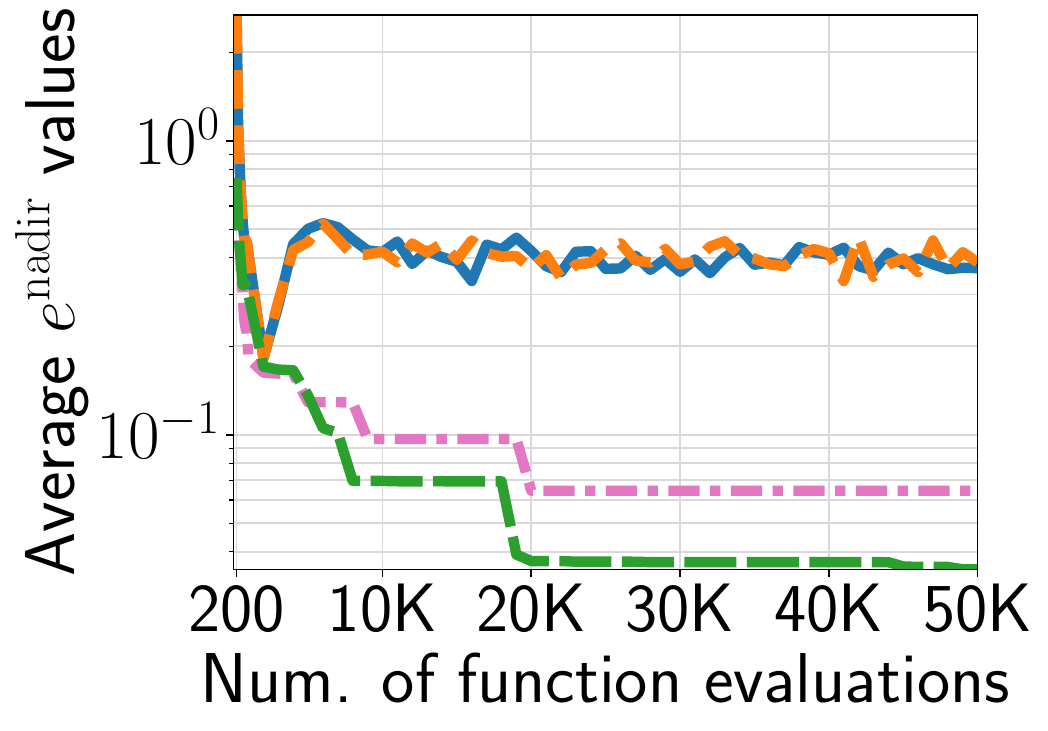}}
   \subfloat[$e^{\mathrm{nadir}}$ ($m=4$)]{\includegraphics[width=0.32\textwidth]{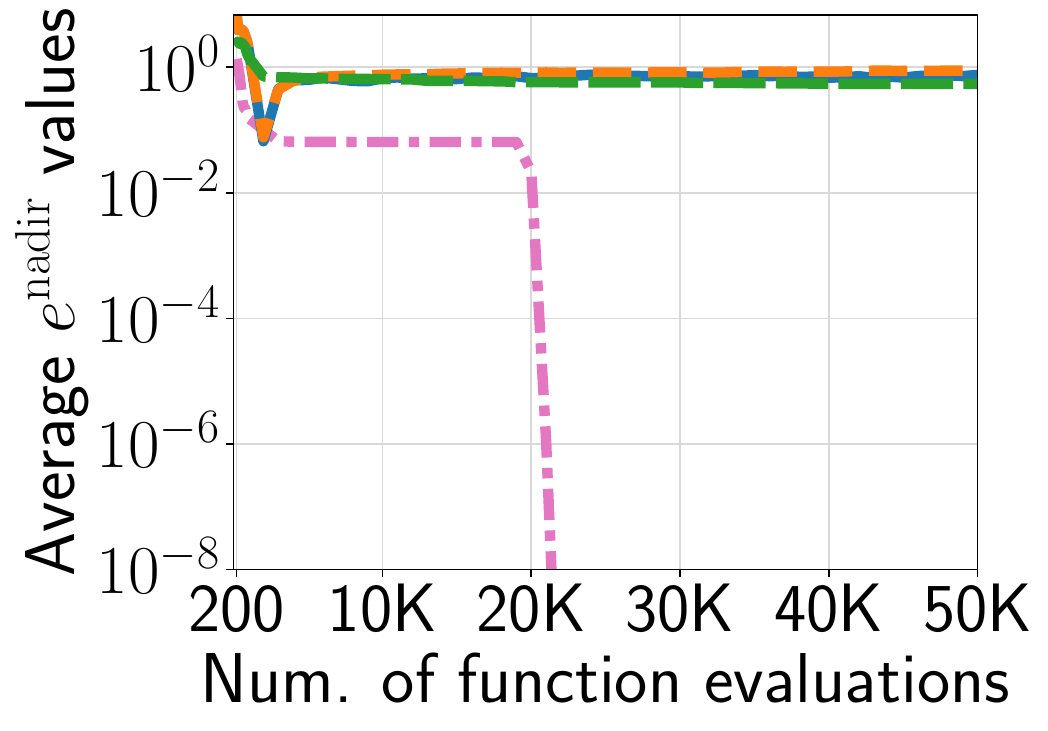}}
   \subfloat[$e^{\mathrm{nadir}}$ ($m=6$)]{\includegraphics[width=0.32\textwidth]{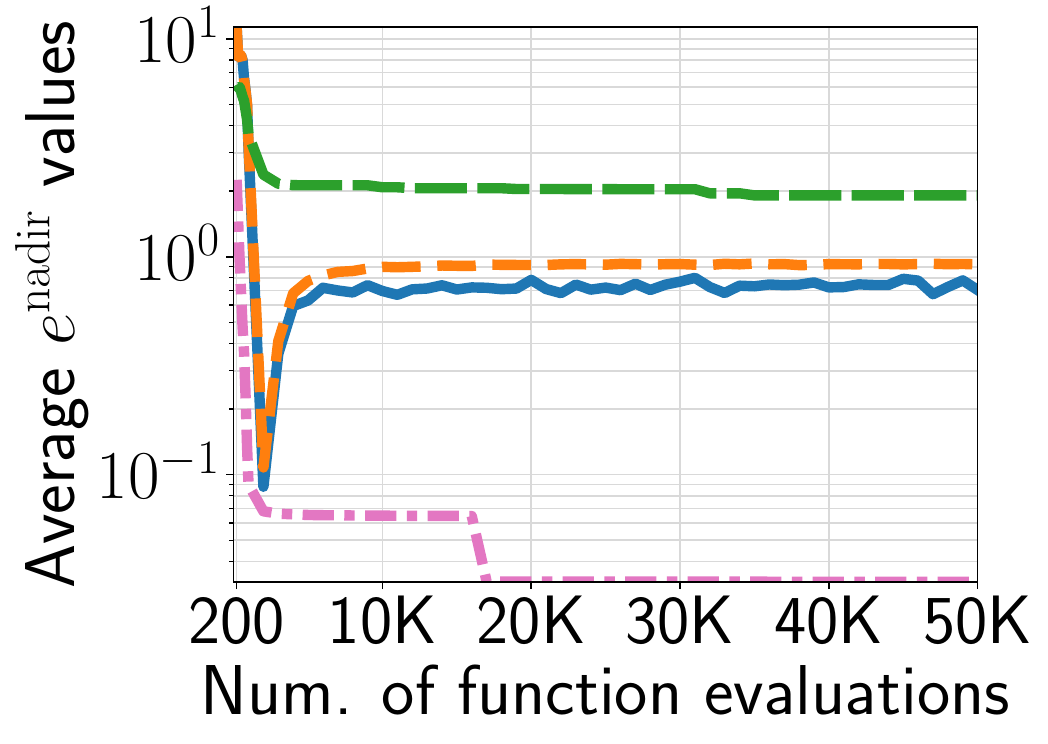}}
\\
   \subfloat[ORE ($m=2$)]{\includegraphics[width=0.32\textwidth]{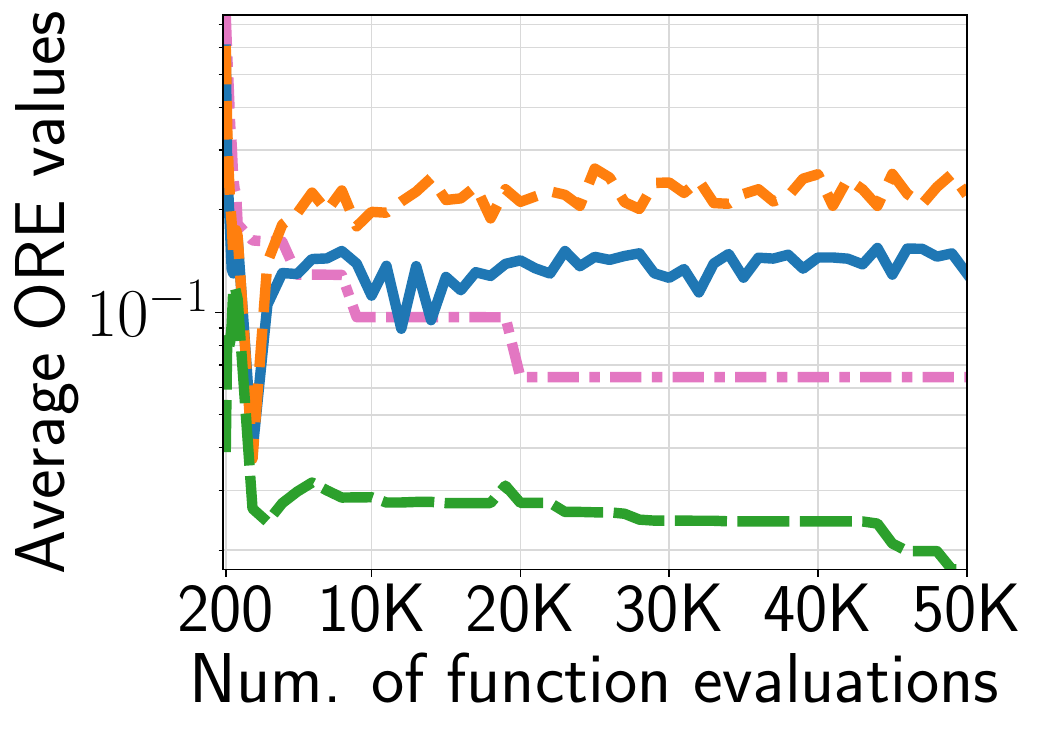}}
   \subfloat[ORE ($m=4$)]{\includegraphics[width=0.32\textwidth]{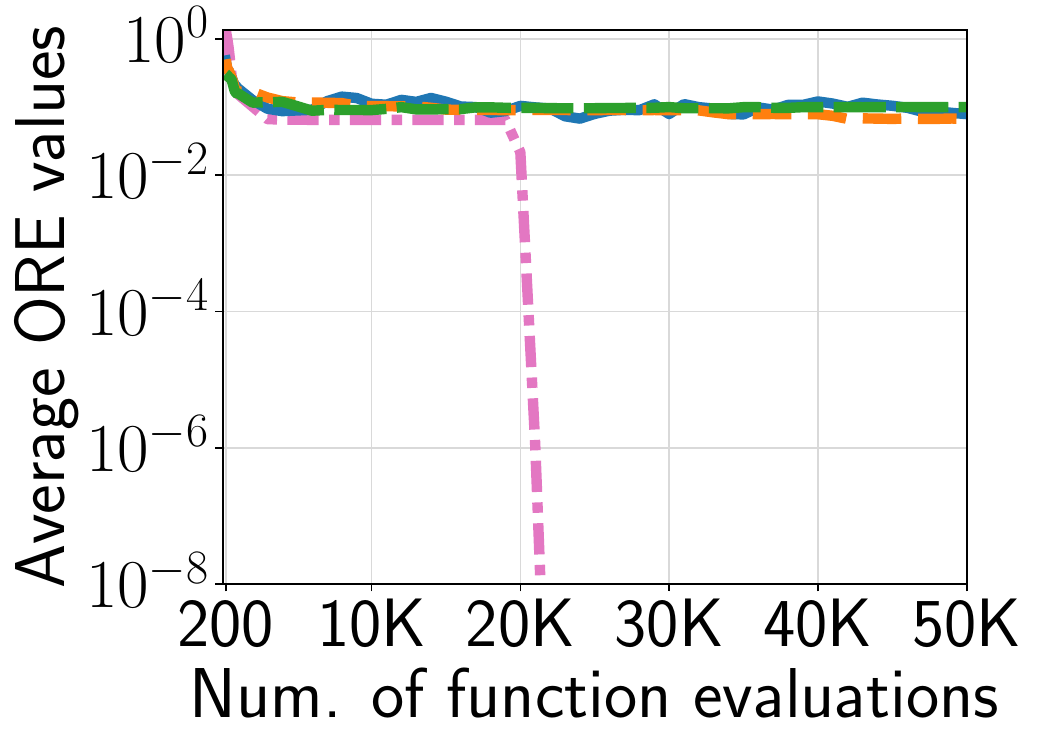}}
   \subfloat[ORE ($m=6$)]{\includegraphics[width=0.32\textwidth]{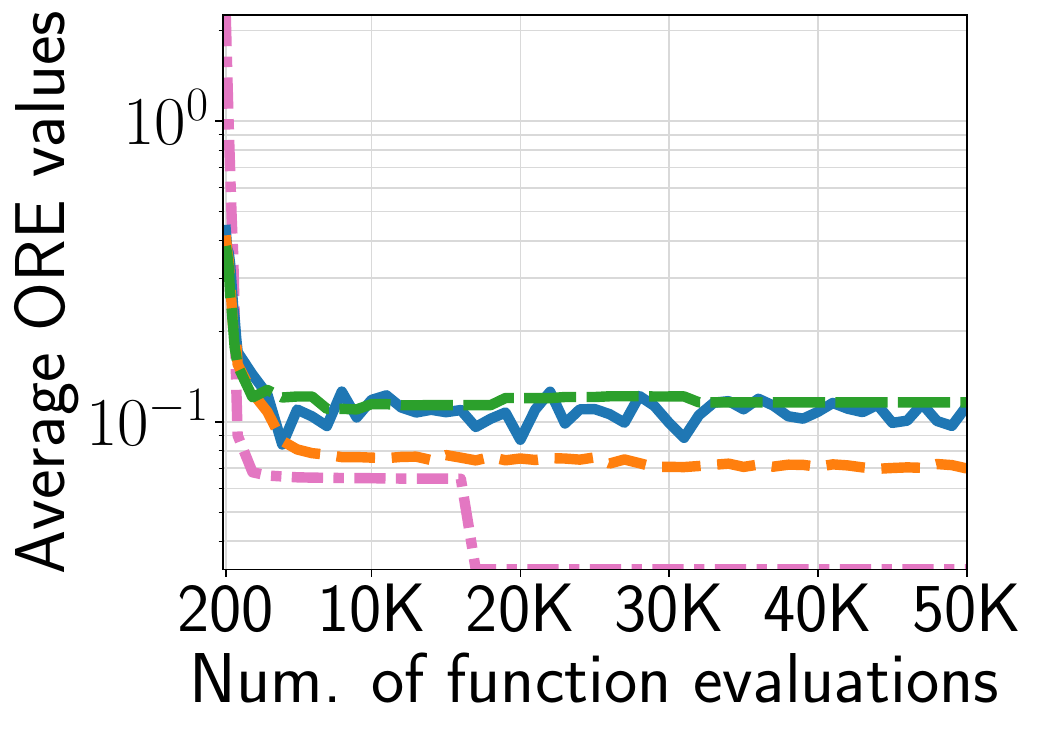}}
\\
\caption{Average $e^{\mathrm{ideal}}$, $e^{\mathrm{nadir}}$, and ORE values of the three normalization methods in R-NSGA-II on IDTLZ4.}
\label{supfig:3error_RNSGA2_IDTLZ4}
\end{figure*}

\clearpage

\begin{figure*}[t]
\centering
  \subfloat{\includegraphics[width=0.7\textwidth]{./figs/legend/legend_3.pdf}}
\vspace{-3.9mm}
   \\
   \subfloat[$e^{\mathrm{ideal}}$ ($m=2$)]{\includegraphics[width=0.32\textwidth]{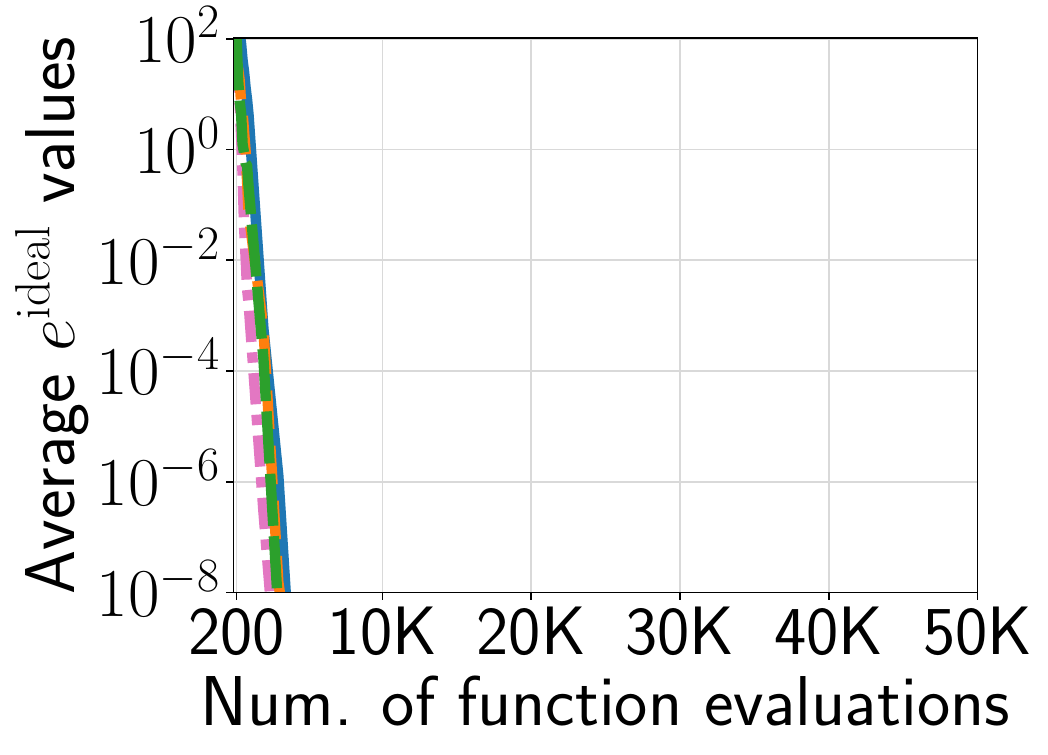}}
   \subfloat[$e^{\mathrm{ideal}}$ ($m=4$)]{\includegraphics[width=0.32\textwidth]{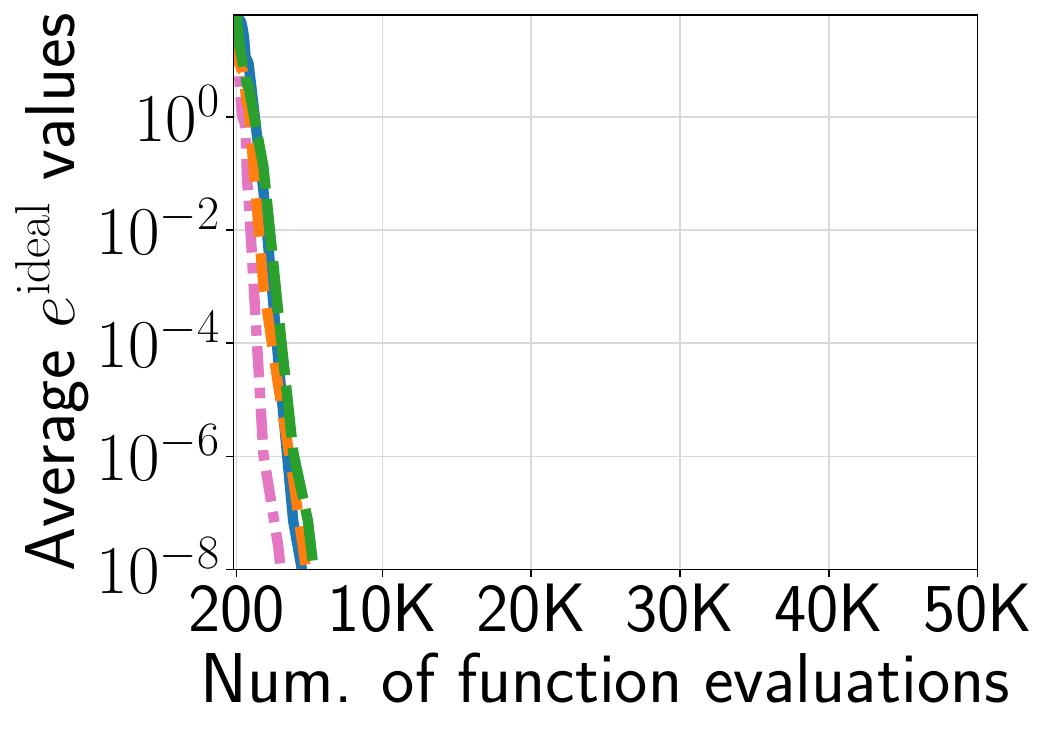}}
   \subfloat[$e^{\mathrm{ideal}}$ ($m=6$)]{\includegraphics[width=0.32\textwidth]{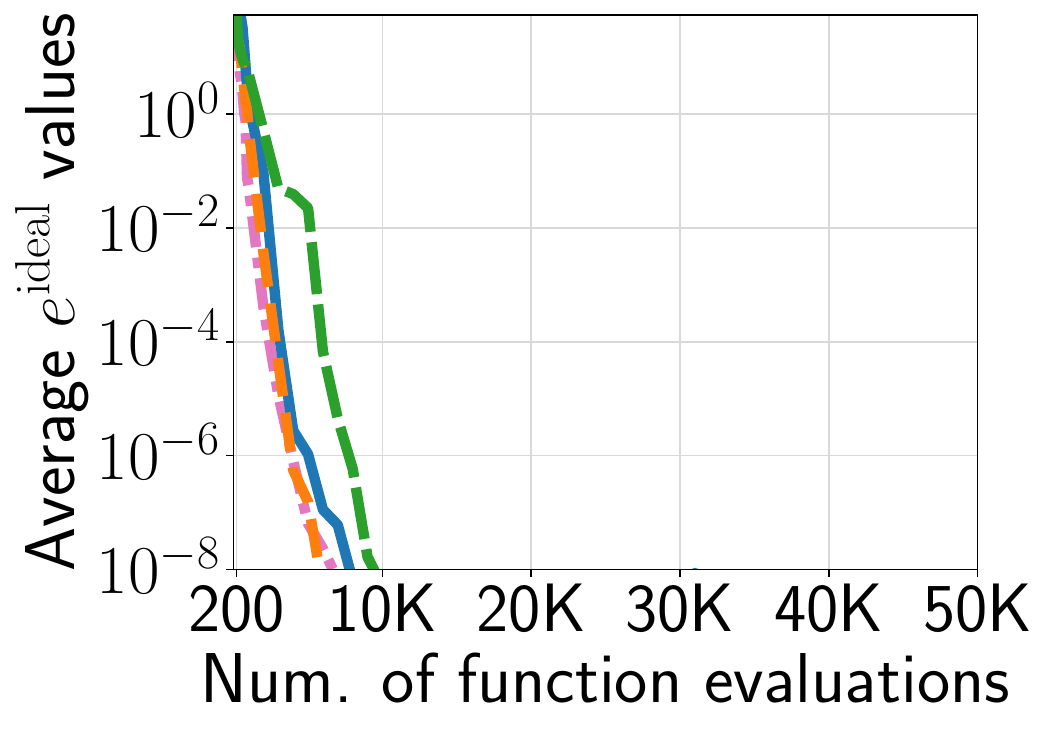}}
\\
   \subfloat[$e^{\mathrm{nadir}}$ ($m=2$)]{\includegraphics[width=0.32\textwidth]{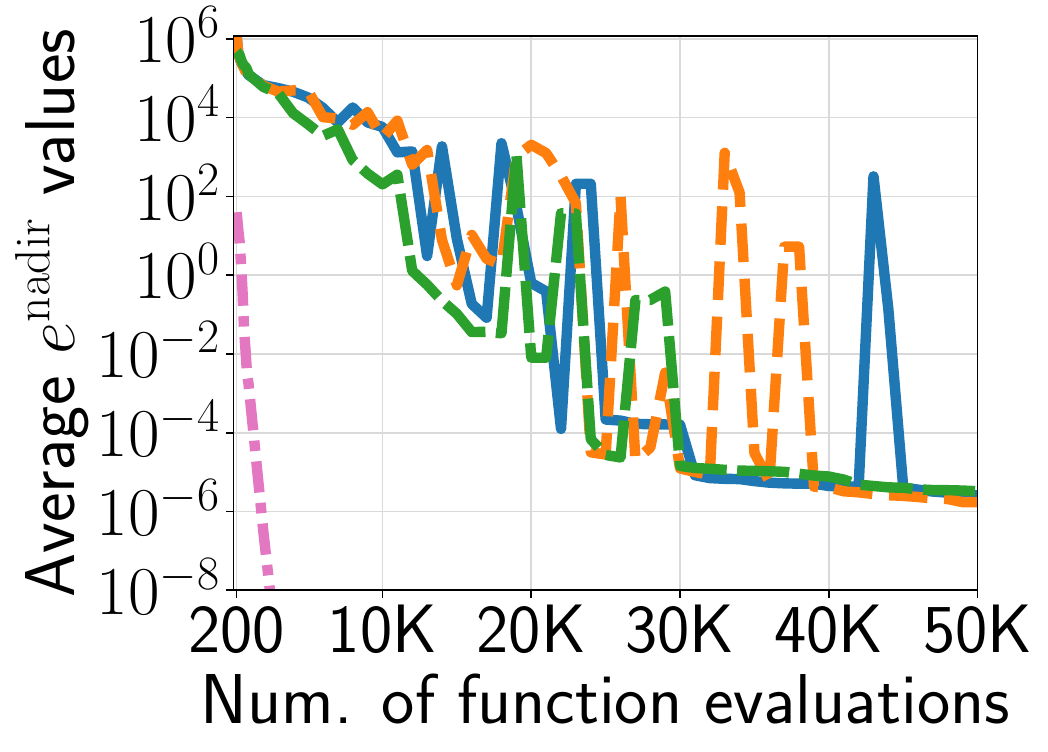}}
   \subfloat[$e^{\mathrm{nadir}}$ ($m=4$)]{\includegraphics[width=0.32\textwidth]{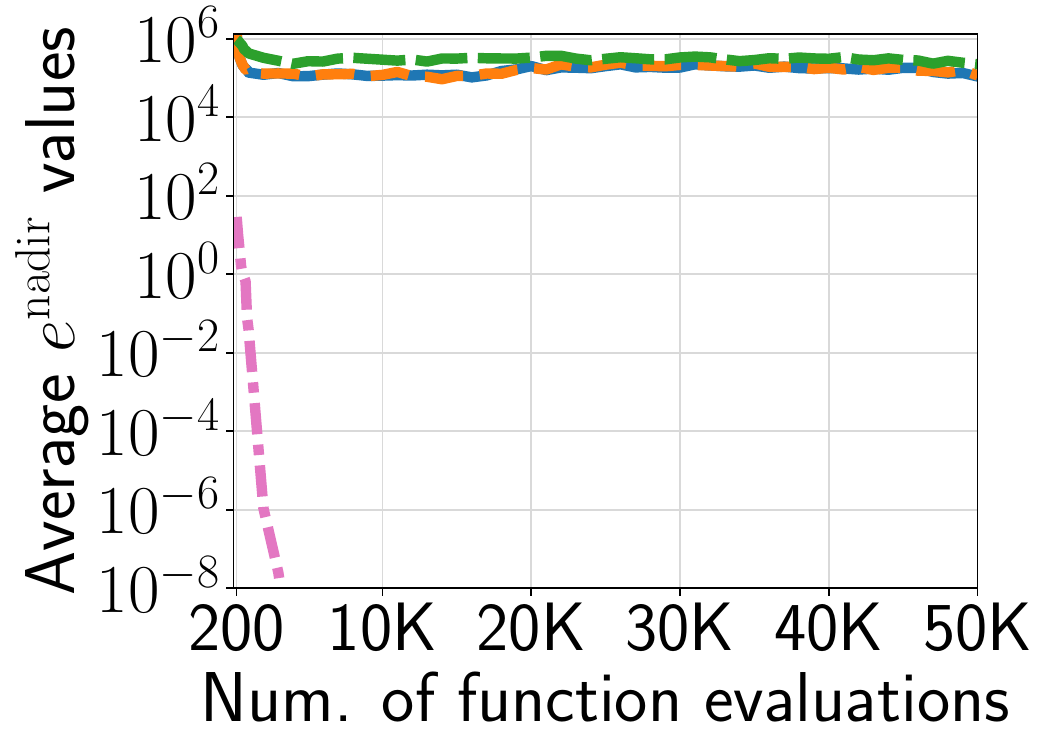}}
   \subfloat[$e^{\mathrm{nadir}}$ ($m=6$)]{\includegraphics[width=0.32\textwidth]{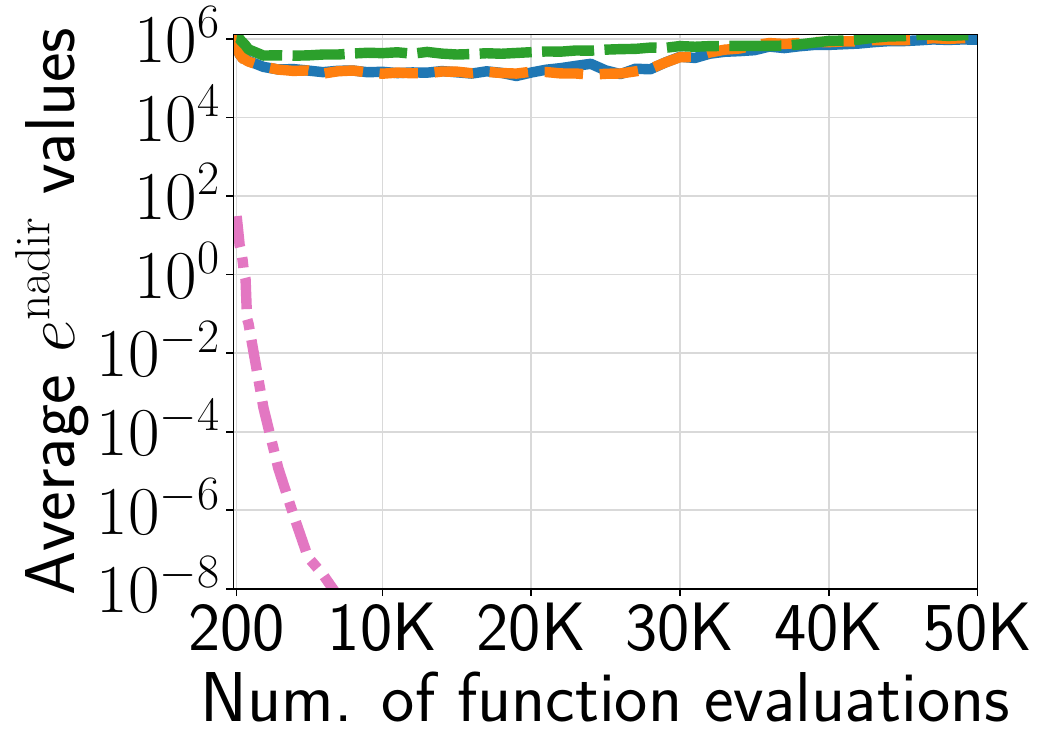}}
\\
   \subfloat[ORE ($m=2$)]{\includegraphics[width=0.32\textwidth]{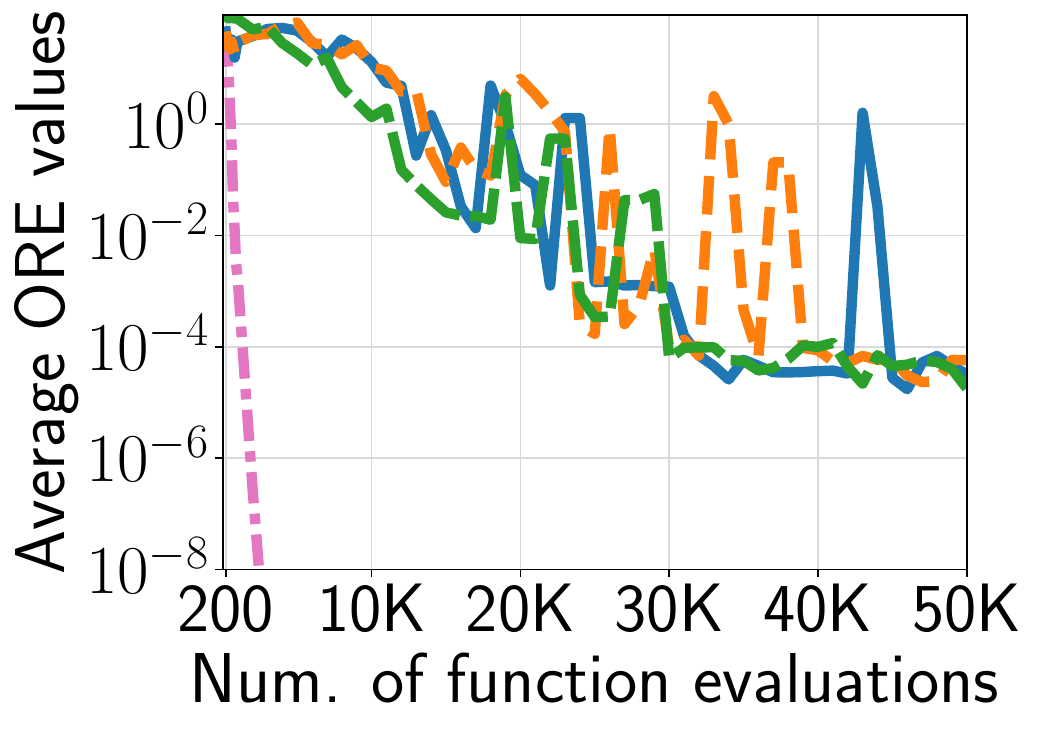}}
   \subfloat[ORE ($m=4$)]{\includegraphics[width=0.32\textwidth]{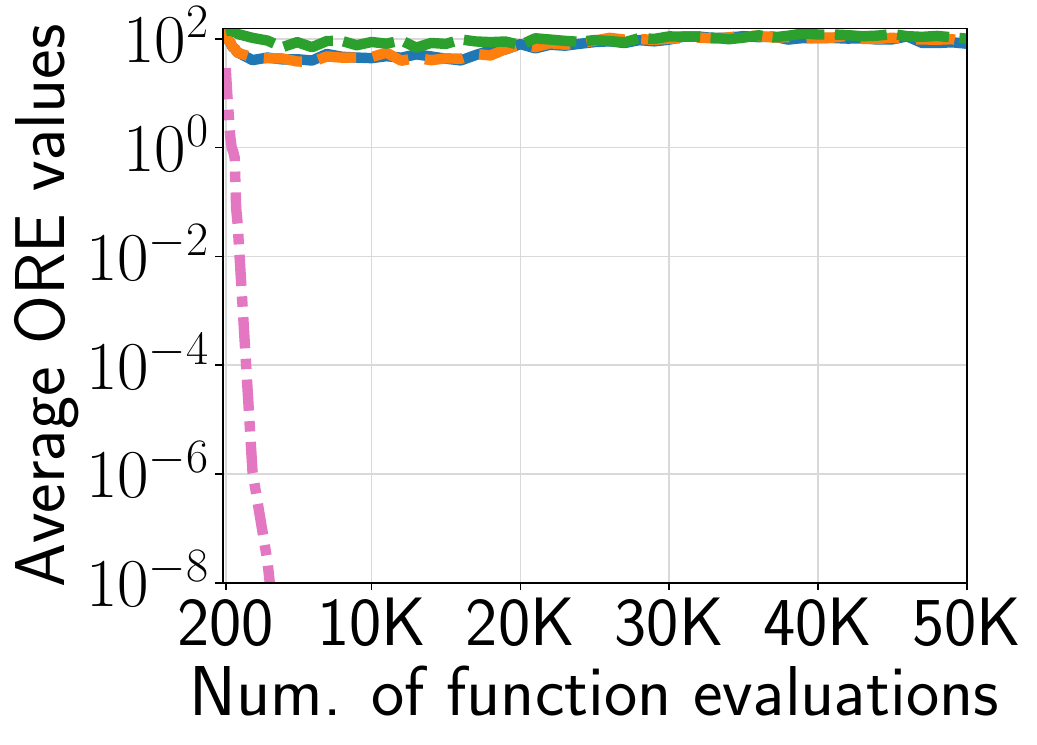}}
   \subfloat[ORE ($m=6$)]{\includegraphics[width=0.32\textwidth]{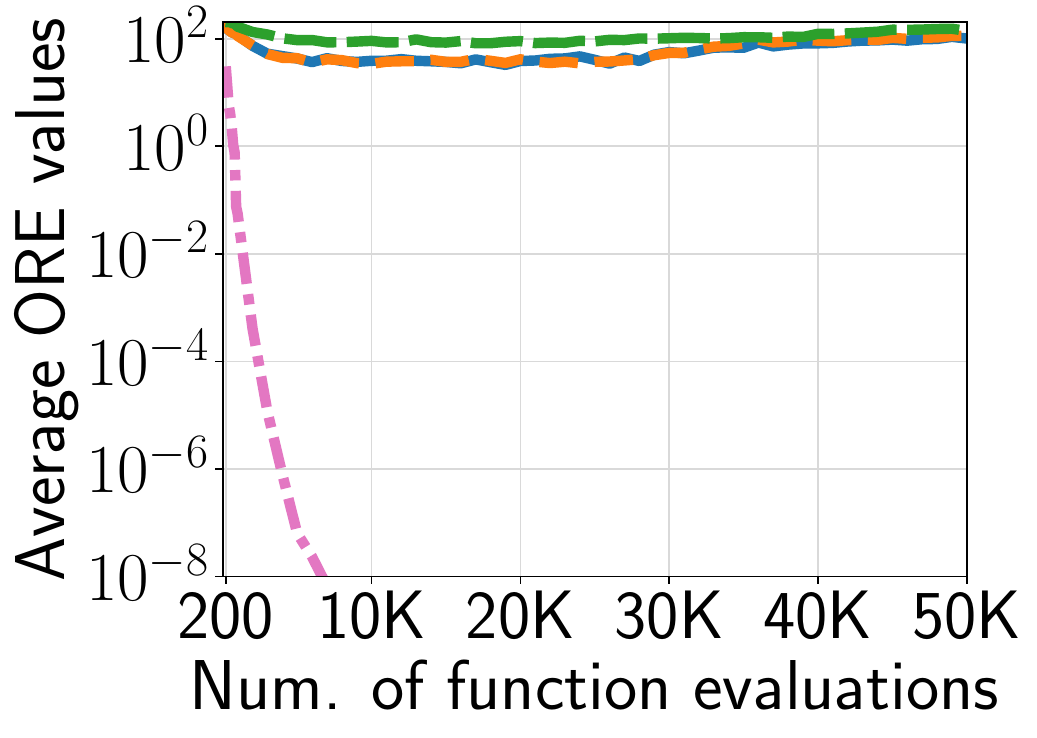}}
\\
\caption{Average $e^{\mathrm{ideal}}$, $e^{\mathrm{nadir}}$, and ORE values of the three normalization methods in r-NSGA-II on DTLZ1.}
\label{supfig:3error_rNSGA2_DTLZ1}
\end{figure*}

\begin{figure*}[t]
\centering
  \subfloat{\includegraphics[width=0.7\textwidth]{./figs/legend/legend_3.pdf}}
\vspace{-3.9mm}
   \\
   \subfloat[$e^{\mathrm{ideal}}$ ($m=2$)]{\includegraphics[width=0.32\textwidth]{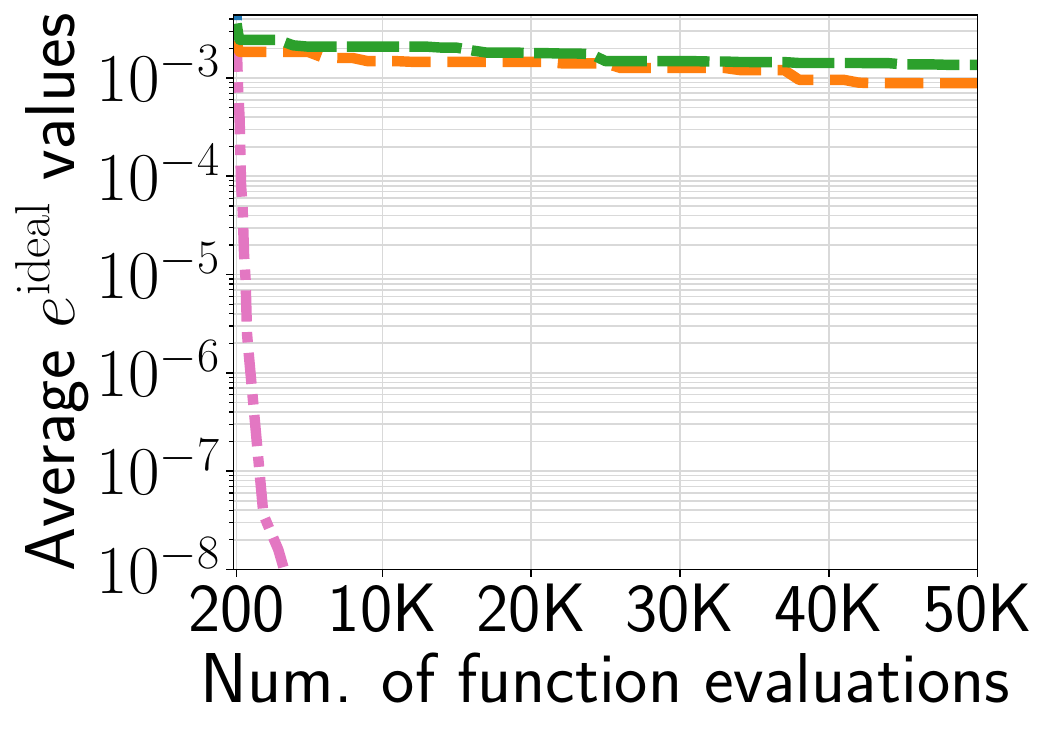}}
   \subfloat[$e^{\mathrm{ideal}}$ ($m=4$)]{\includegraphics[width=0.32\textwidth]{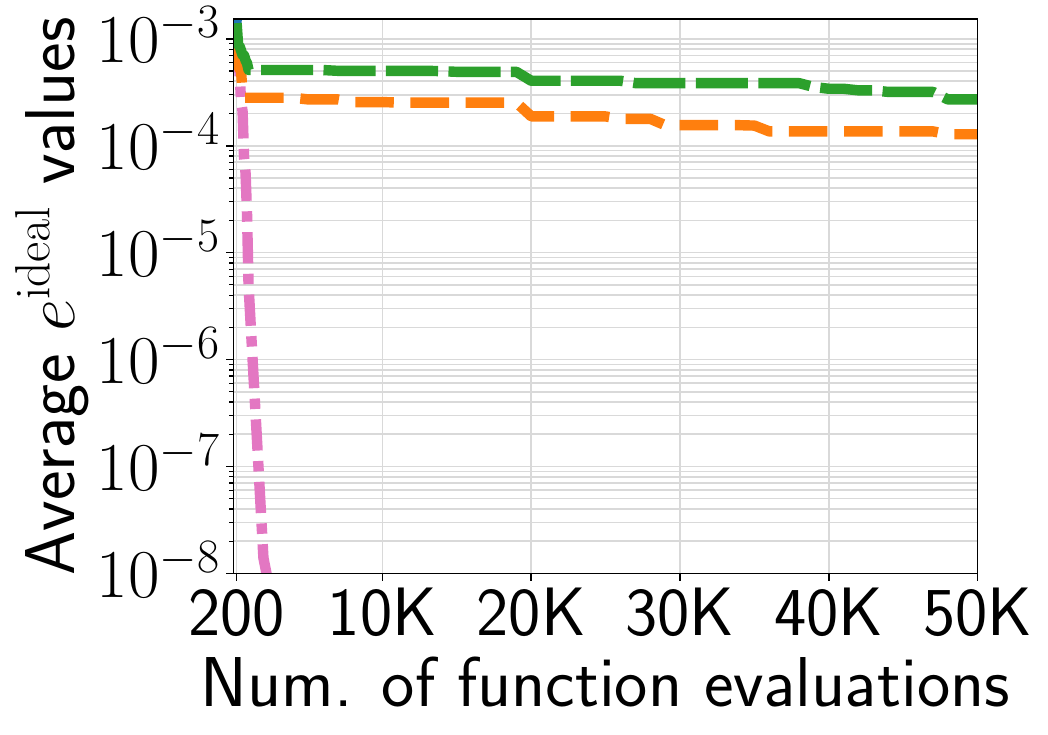}}
   \subfloat[$e^{\mathrm{ideal}}$ ($m=6$)]{\includegraphics[width=0.32\textwidth]{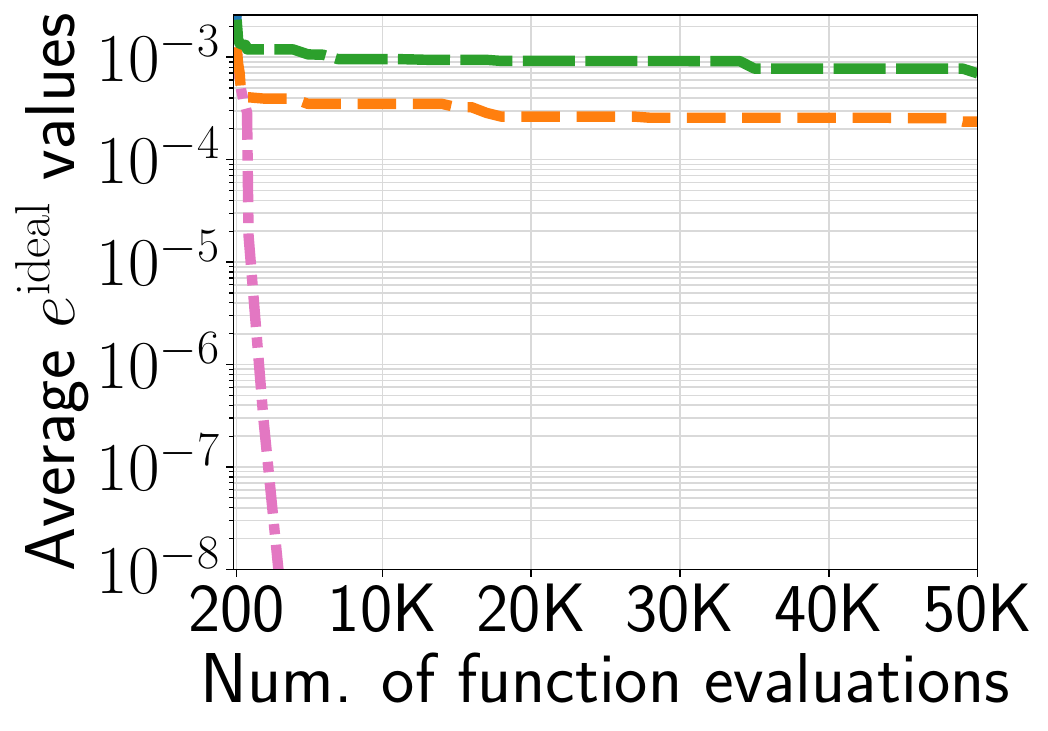}}
\\
   \subfloat[$e^{\mathrm{nadir}}$ ($m=2$)]{\includegraphics[width=0.32\textwidth]{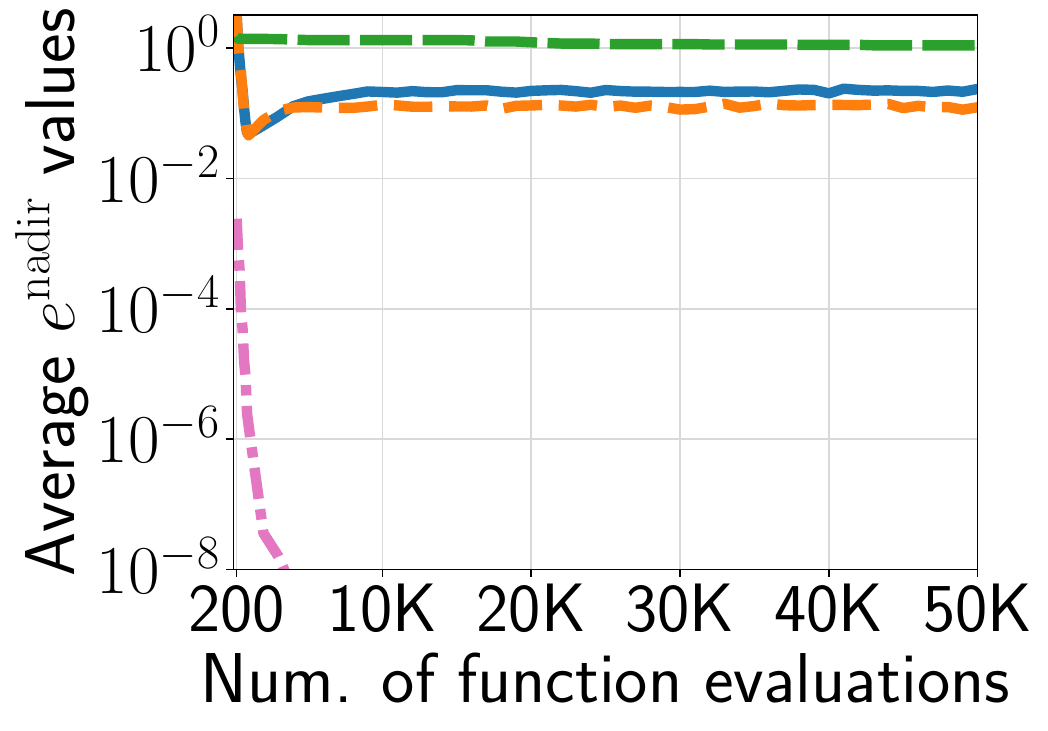}}
   \subfloat[$e^{\mathrm{nadir}}$ ($m=4$)]{\includegraphics[width=0.32\textwidth]{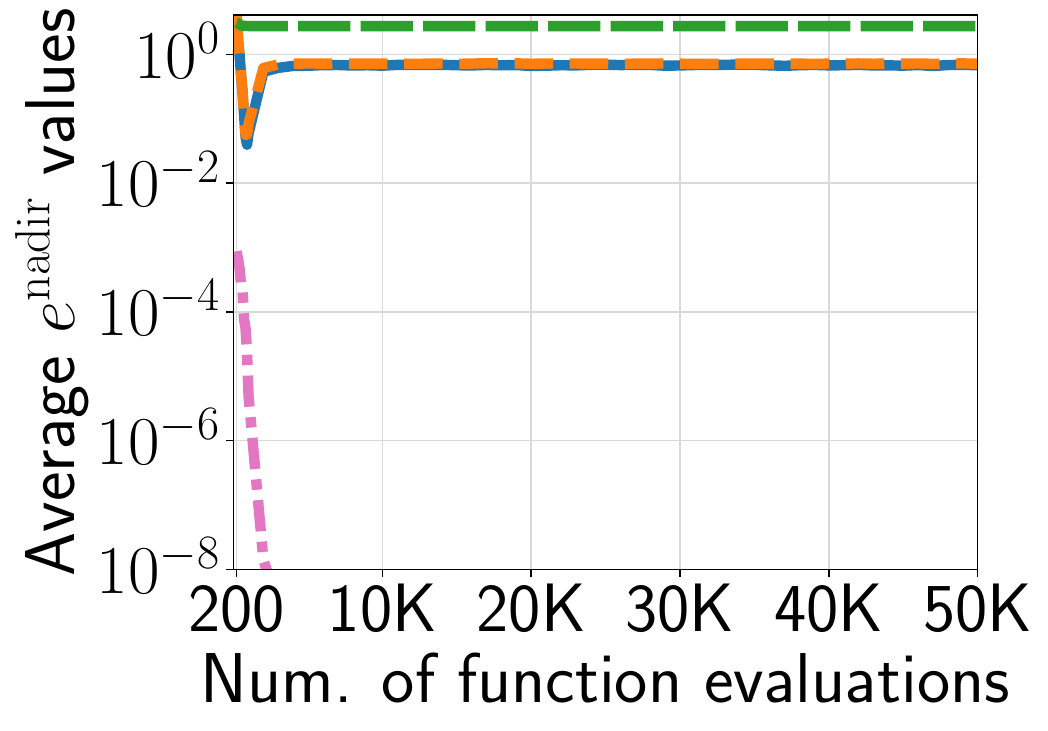}}
   \subfloat[$e^{\mathrm{nadir}}$ ($m=6$)]{\includegraphics[width=0.32\textwidth]{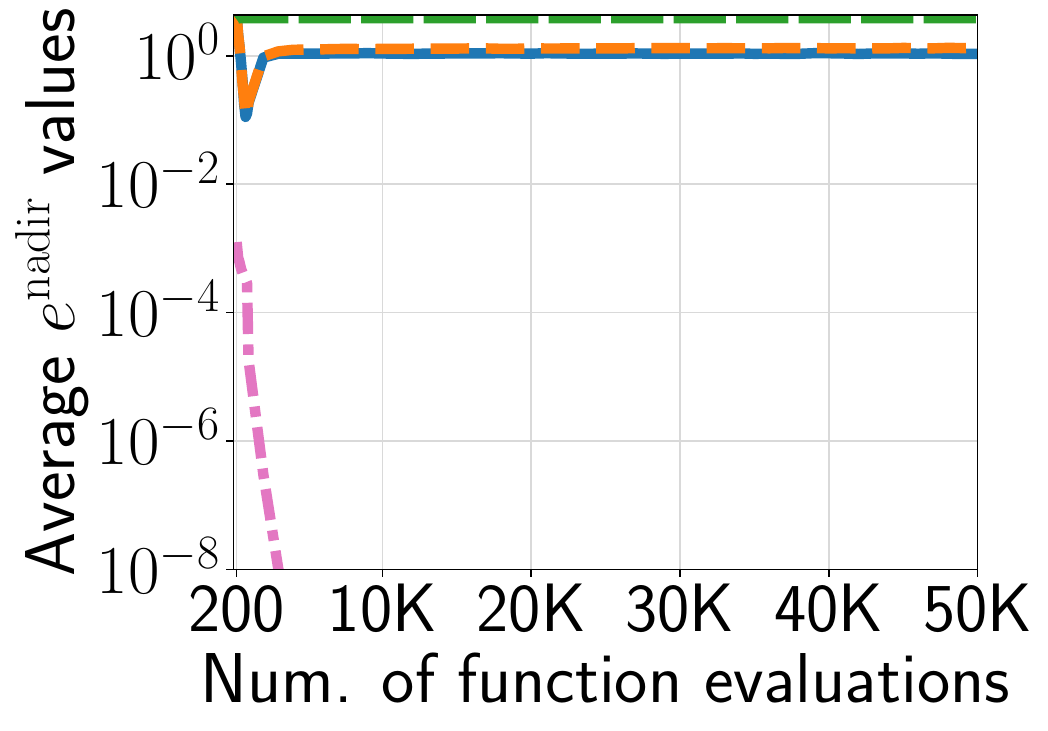}}
\\
   \subfloat[ORE ($m=2$)]{\includegraphics[width=0.32\textwidth]{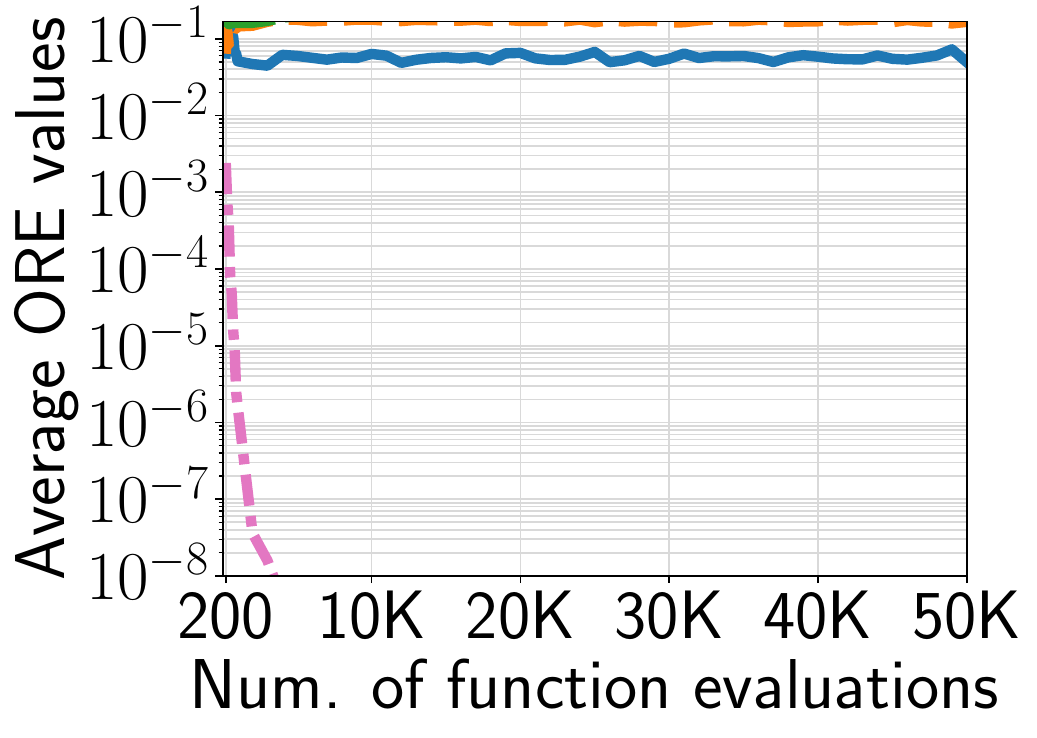}}
   \subfloat[ORE ($m=4$)]{\includegraphics[width=0.32\textwidth]{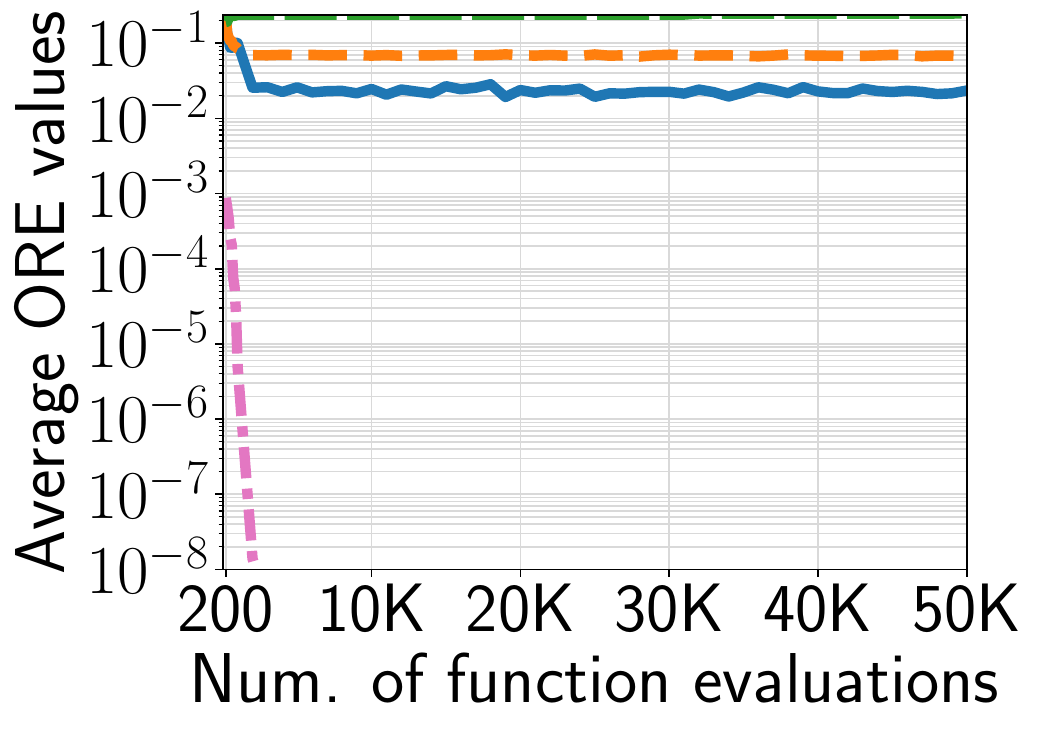}}
   \subfloat[ORE ($m=6$)]{\includegraphics[width=0.32\textwidth]{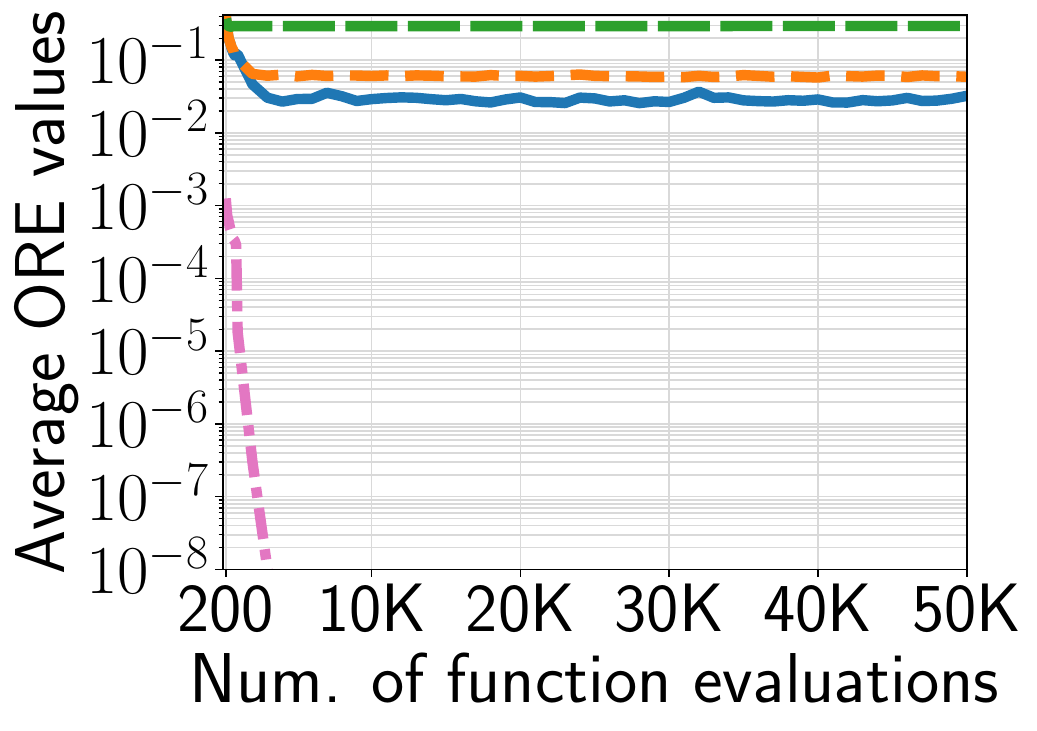}}
\\
\caption{Average $e^{\mathrm{ideal}}$, $e^{\mathrm{nadir}}$, and ORE values of the three normalization methods in r-NSGA-II on DTLZ2.}
\label{supfig:3error_rNSGA2_DTLZ2}
\end{figure*}

\begin{figure*}[t]
\centering
  \subfloat{\includegraphics[width=0.7\textwidth]{./figs/legend/legend_3.pdf}}
\vspace{-3.9mm}
   \\
   \subfloat[$e^{\mathrm{ideal}}$ ($m=2$)]{\includegraphics[width=0.32\textwidth]{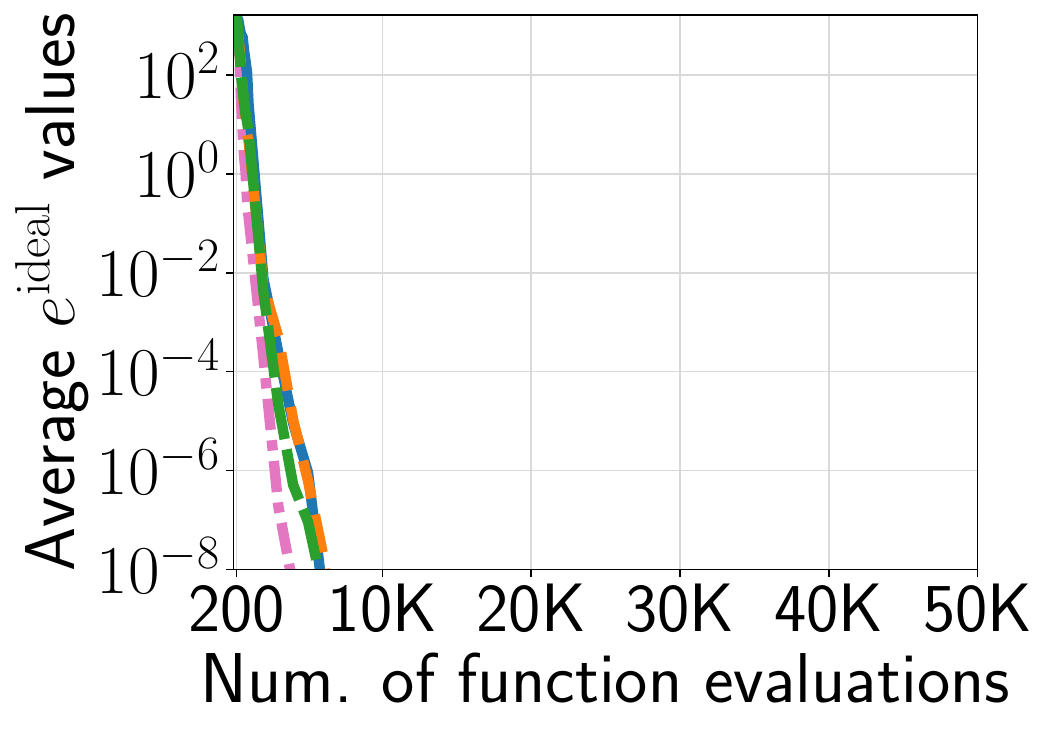}}
   \subfloat[$e^{\mathrm{ideal}}$ ($m=4$)]{\includegraphics[width=0.32\textwidth]{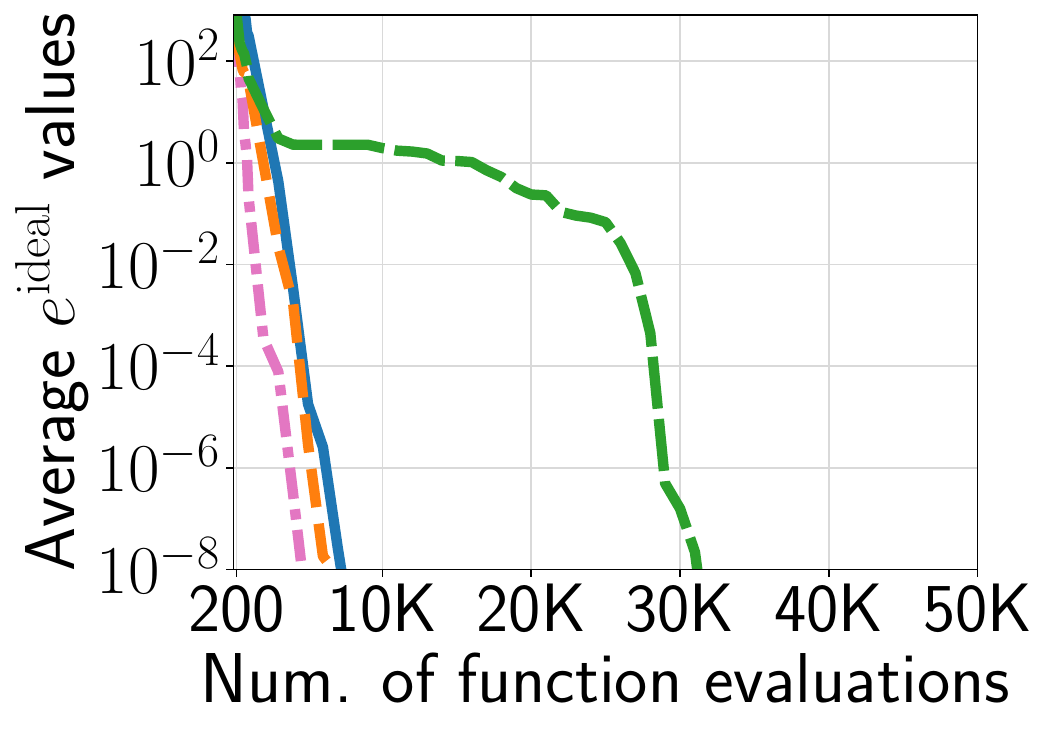}}
   \subfloat[$e^{\mathrm{ideal}}$ ($m=6$)]{\includegraphics[width=0.32\textwidth]{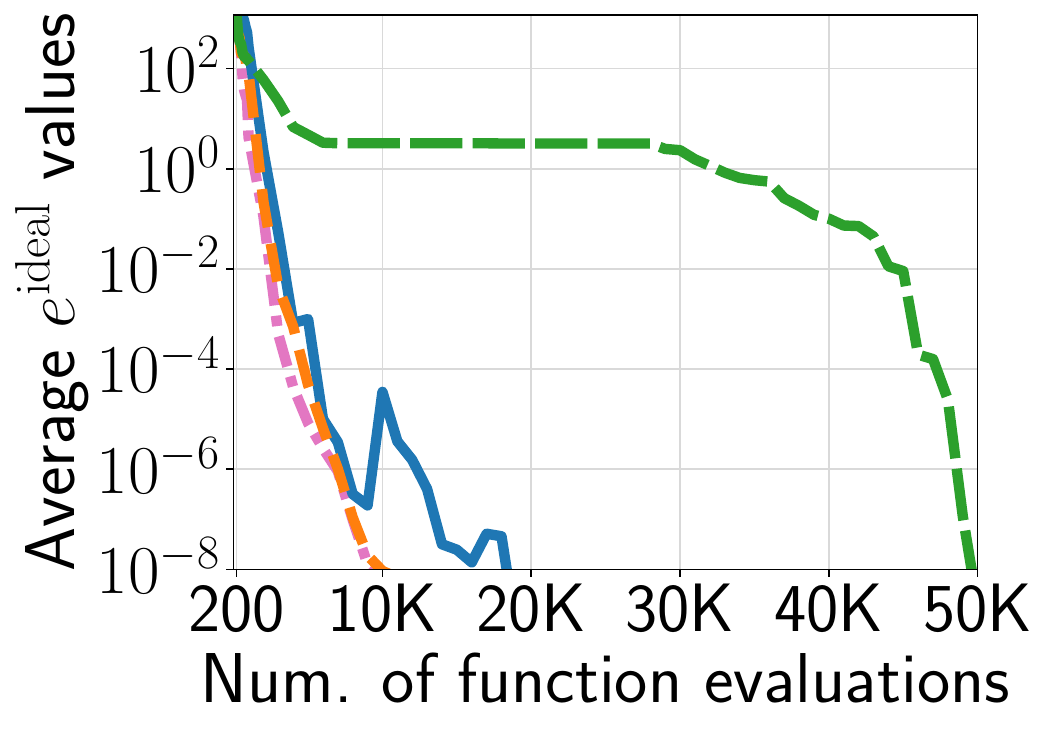}}
\\
   \subfloat[$e^{\mathrm{nadir}}$ ($m=2$)]{\includegraphics[width=0.32\textwidth]{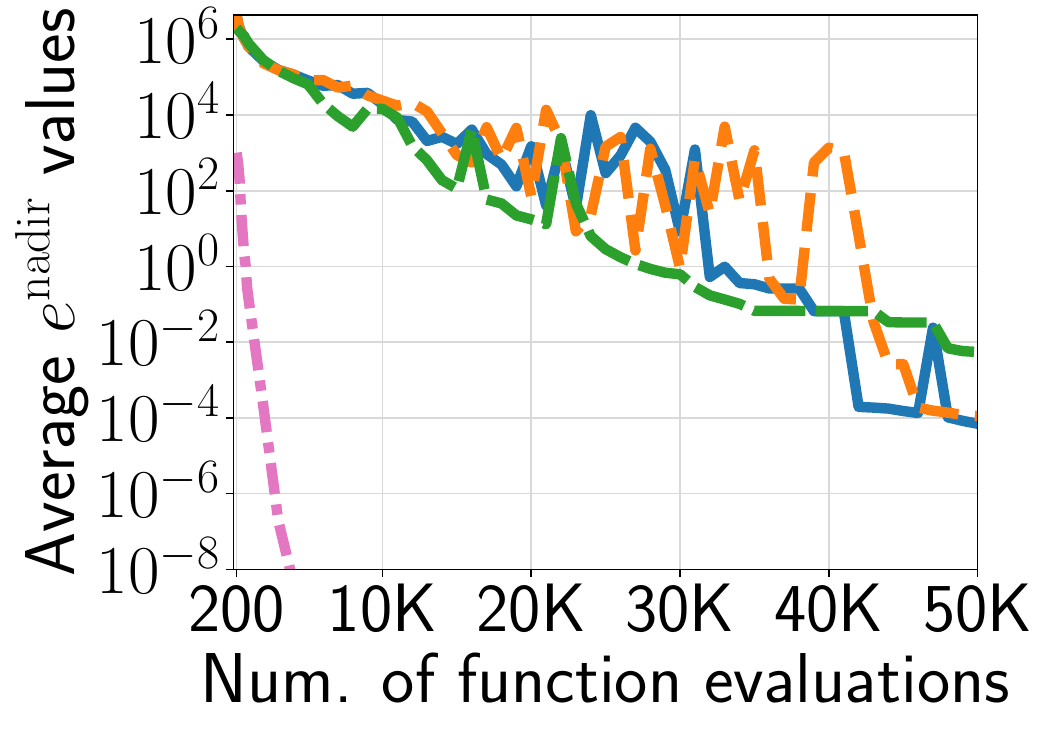}}
   \subfloat[$e^{\mathrm{nadir}}$ ($m=4$)]{\includegraphics[width=0.32\textwidth]{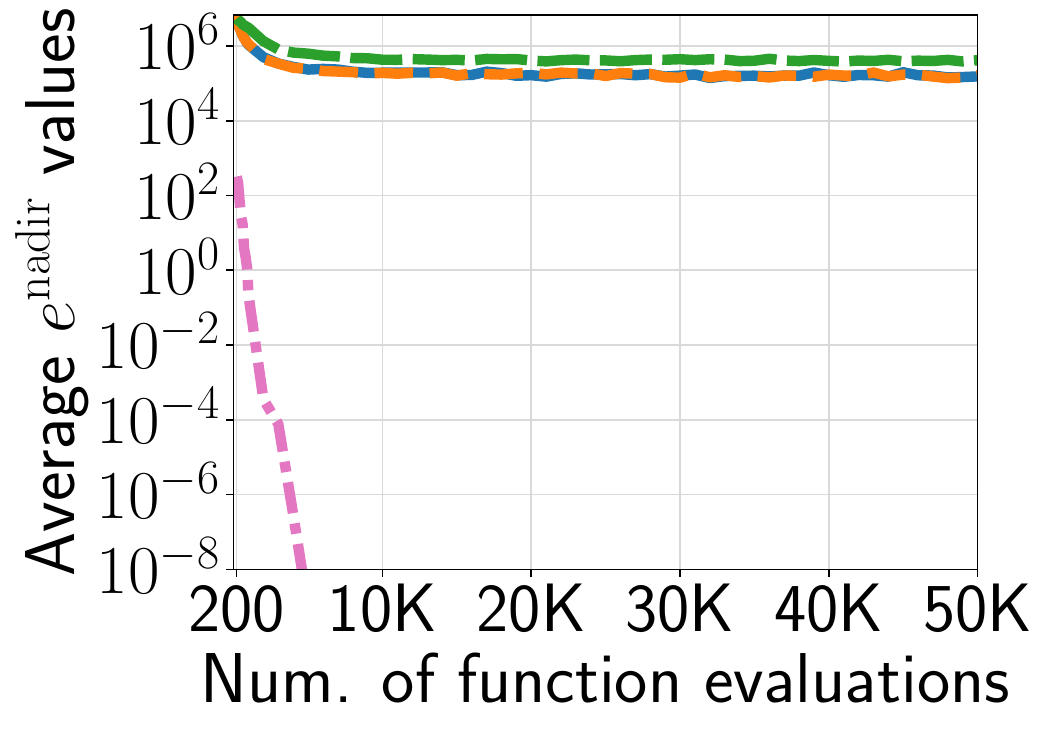}}
   \subfloat[$e^{\mathrm{nadir}}$ ($m=6$)]{\includegraphics[width=0.32\textwidth]{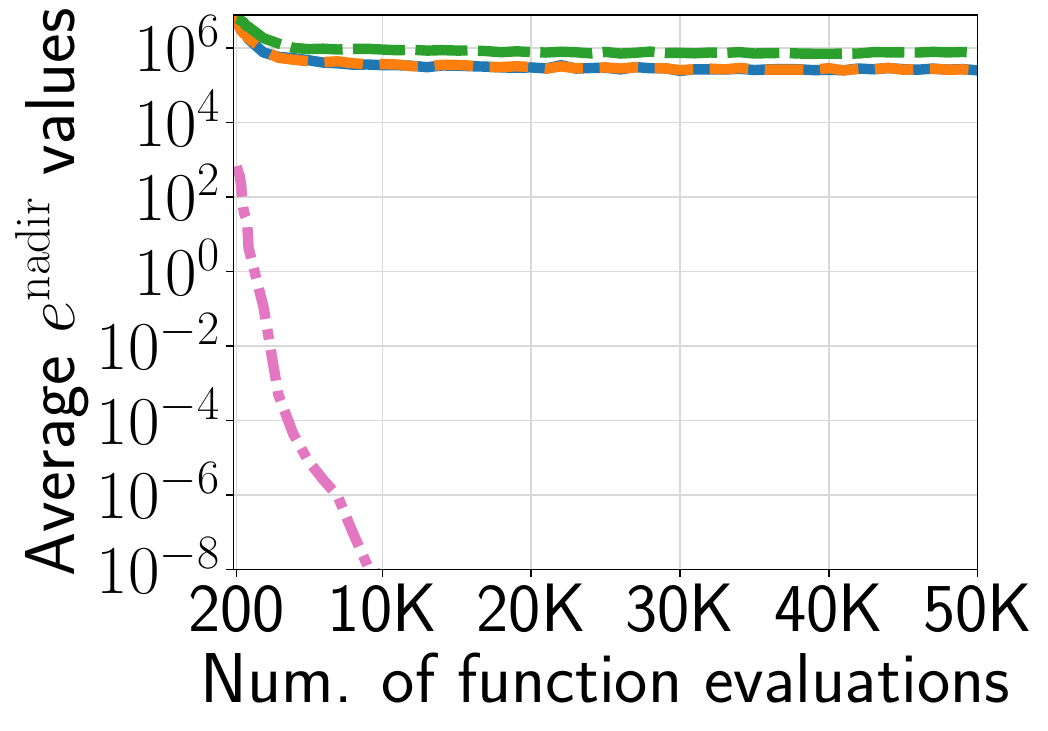}}
\\
   \subfloat[ORE ($m=2$)]{\includegraphics[width=0.32\textwidth]{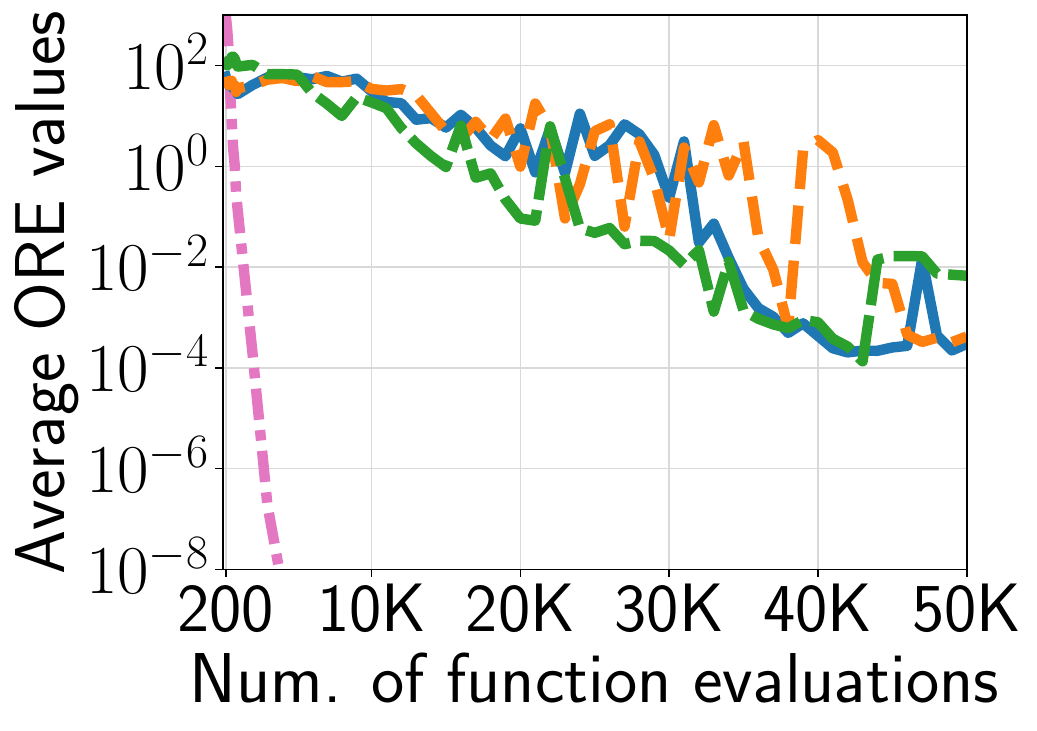}}
   \subfloat[ORE ($m=4$)]{\includegraphics[width=0.32\textwidth]{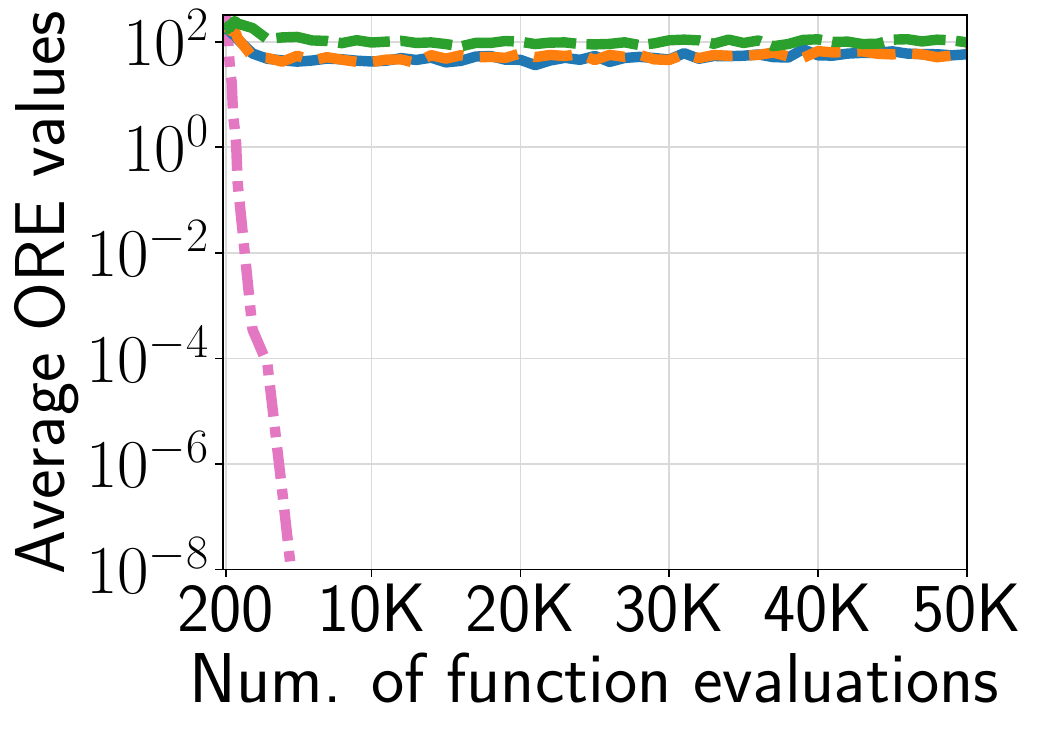}}
   \subfloat[ORE ($m=6$)]{\includegraphics[width=0.32\textwidth]{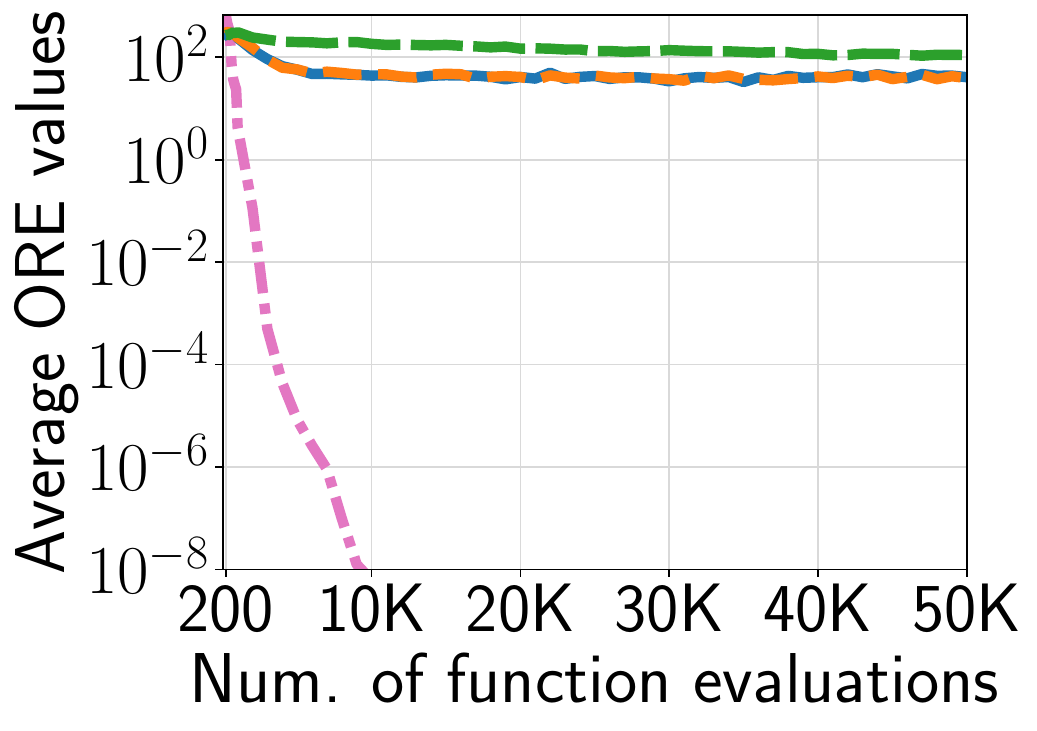}}
\\
\caption{Average $e^{\mathrm{ideal}}$, $e^{\mathrm{nadir}}$, and ORE values of the three normalization methods in r-NSGA-II on DTLZ3.}
\label{supfig:3error_rNSGA2_DTLZ3}
\end{figure*}

\begin{figure*}[t]
\centering
  \subfloat{\includegraphics[width=0.7\textwidth]{./figs/legend/legend_3.pdf}}
\vspace{-3.9mm}
   \\
   \subfloat[$e^{\mathrm{ideal}}$ ($m=2$)]{\includegraphics[width=0.32\textwidth]{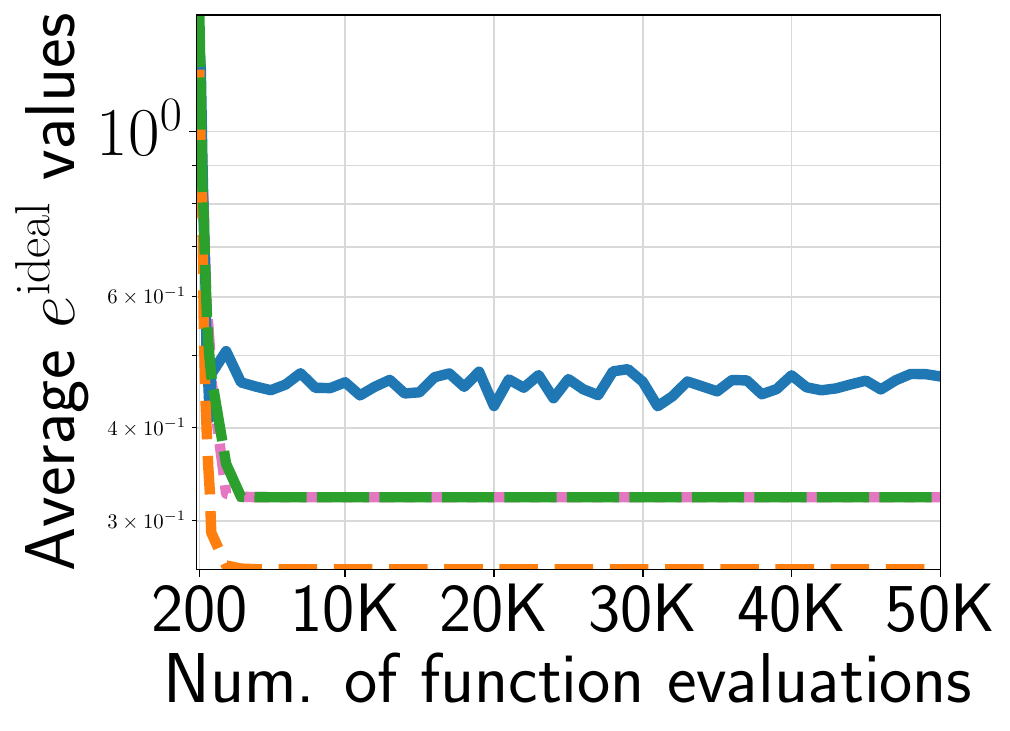}}
   \subfloat[$e^{\mathrm{ideal}}$ ($m=4$)]{\includegraphics[width=0.32\textwidth]{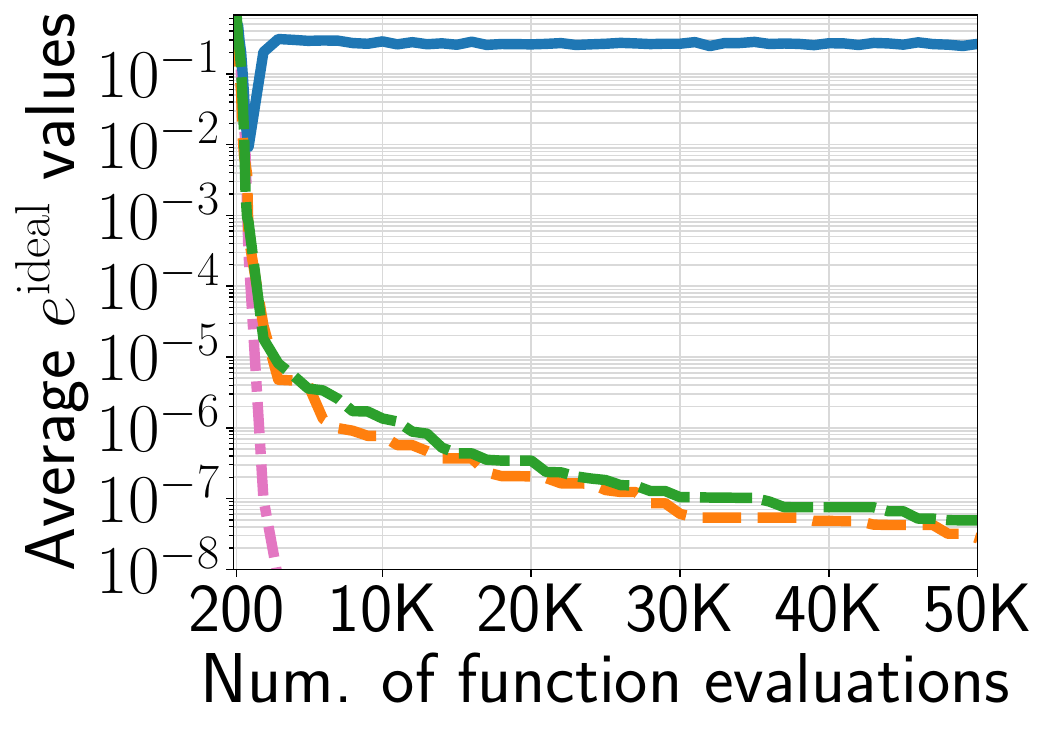}}
   \subfloat[$e^{\mathrm{ideal}}$ ($m=6$)]{\includegraphics[width=0.32\textwidth]{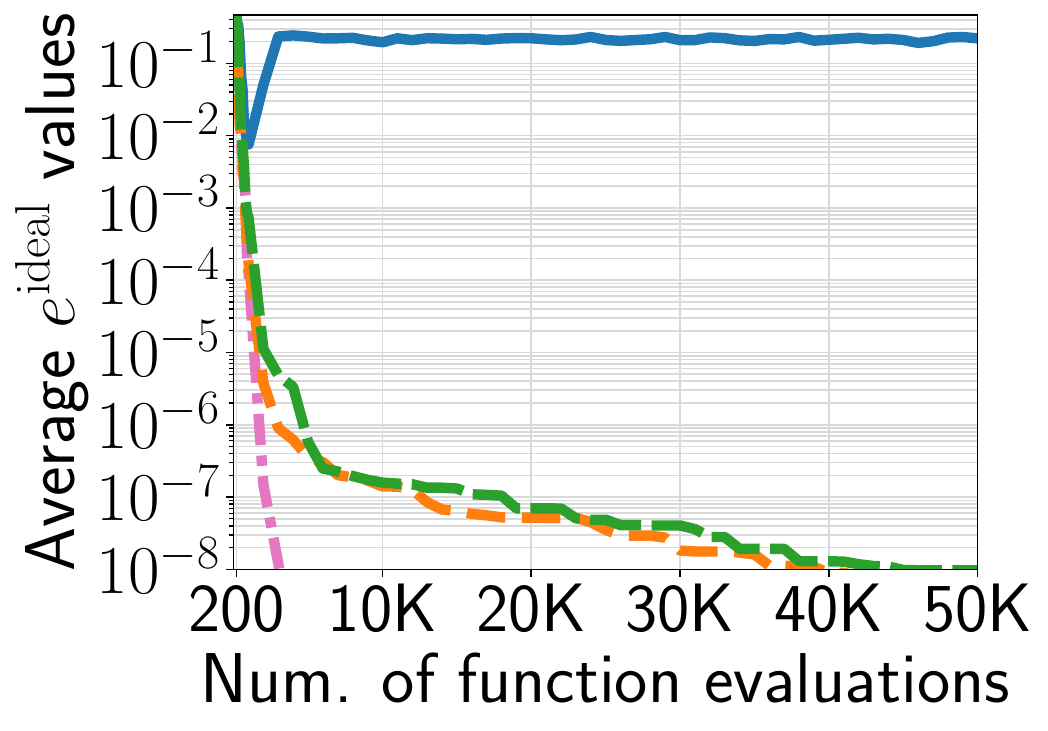}}
\\
   \subfloat[$e^{\mathrm{nadir}}$ ($m=2$)]{\includegraphics[width=0.32\textwidth]{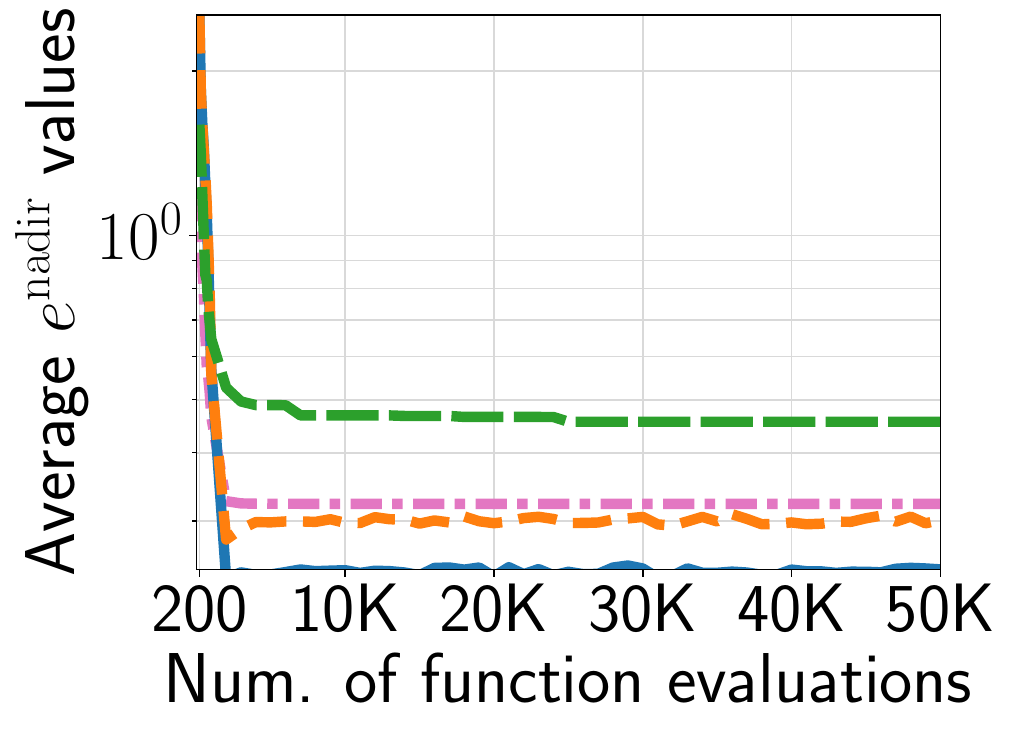}}
   \subfloat[$e^{\mathrm{nadir}}$ ($m=4$)]{\includegraphics[width=0.32\textwidth]{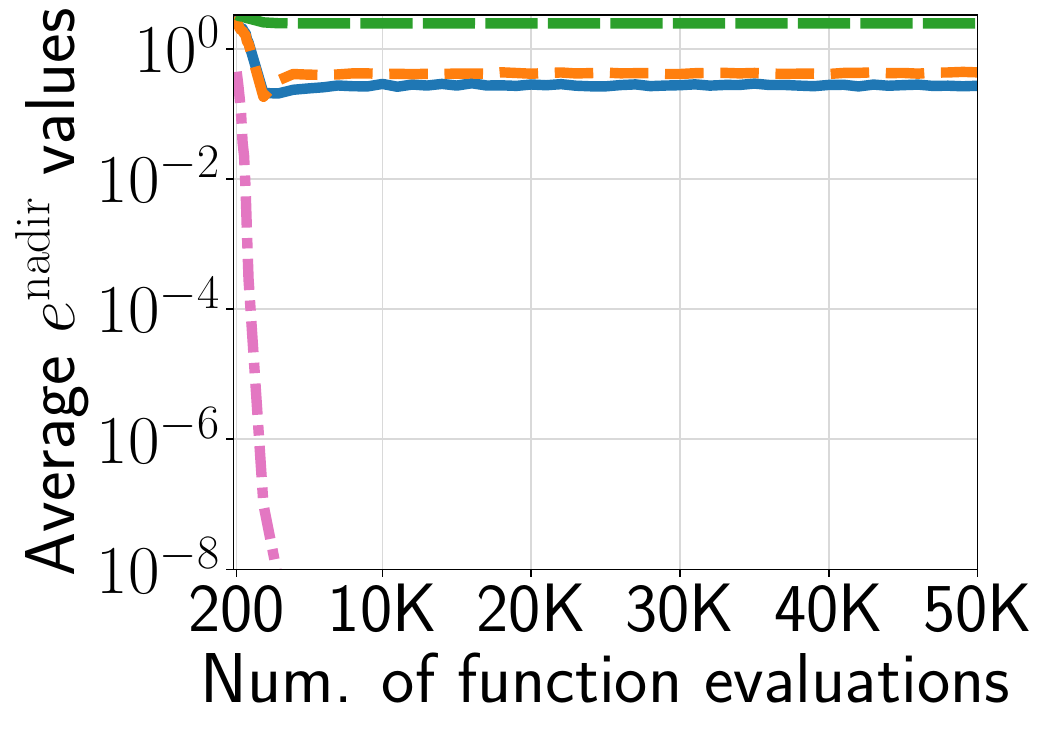}}
   \subfloat[$e^{\mathrm{nadir}}$ ($m=6$)]{\includegraphics[width=0.32\textwidth]{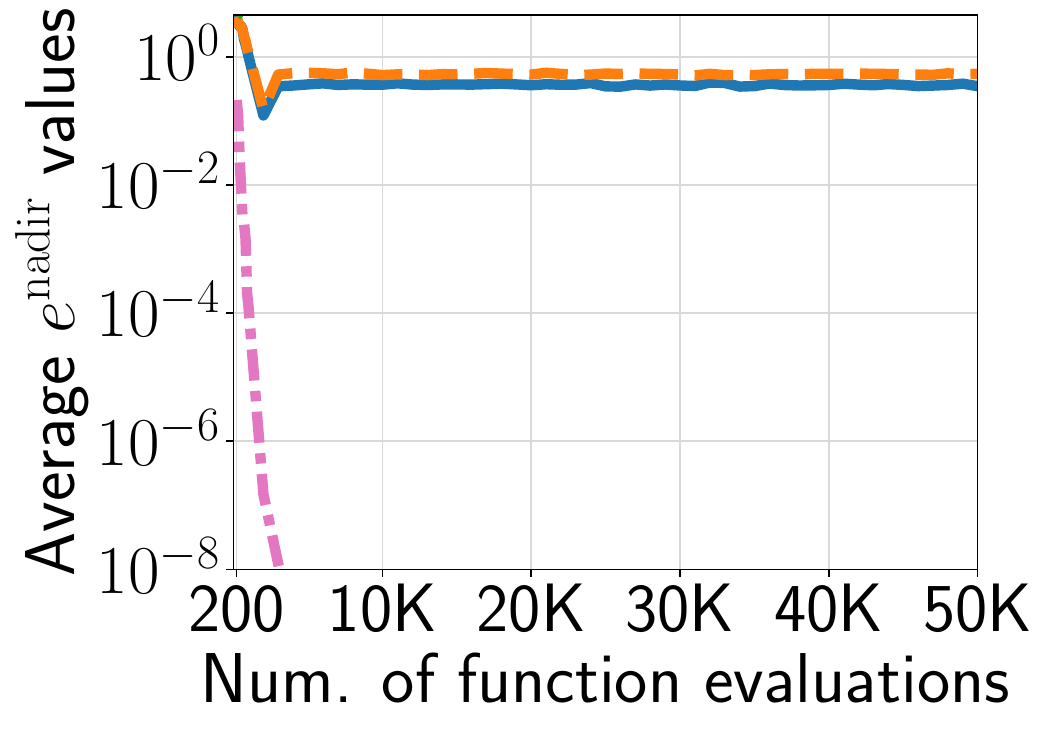}}
\\
   \subfloat[ORE ($m=2$)]{\includegraphics[width=0.32\textwidth]{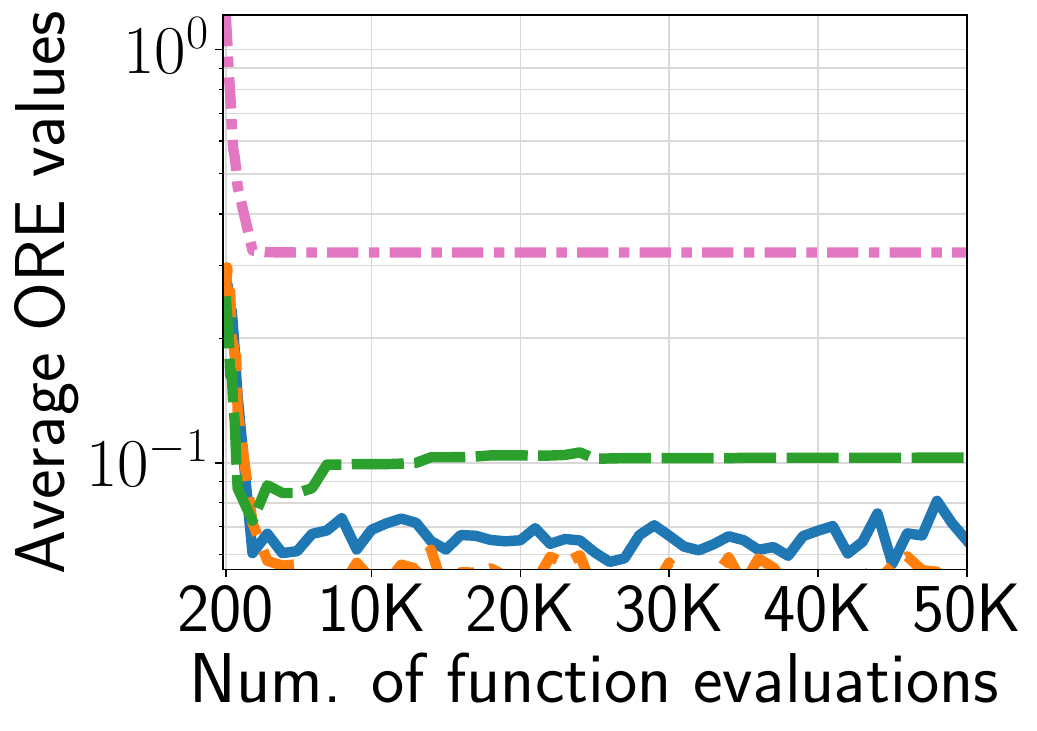}}
   \subfloat[ORE ($m=4$)]{\includegraphics[width=0.32\textwidth]{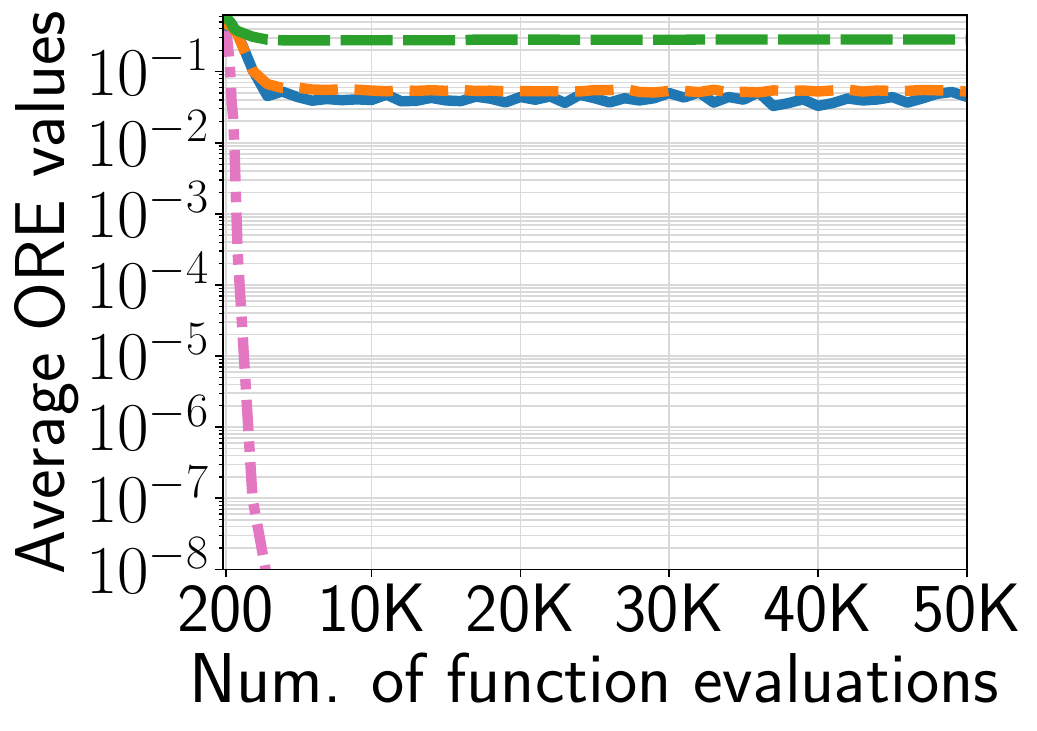}}
   \subfloat[ORE ($m=6$)]{\includegraphics[width=0.32\textwidth]{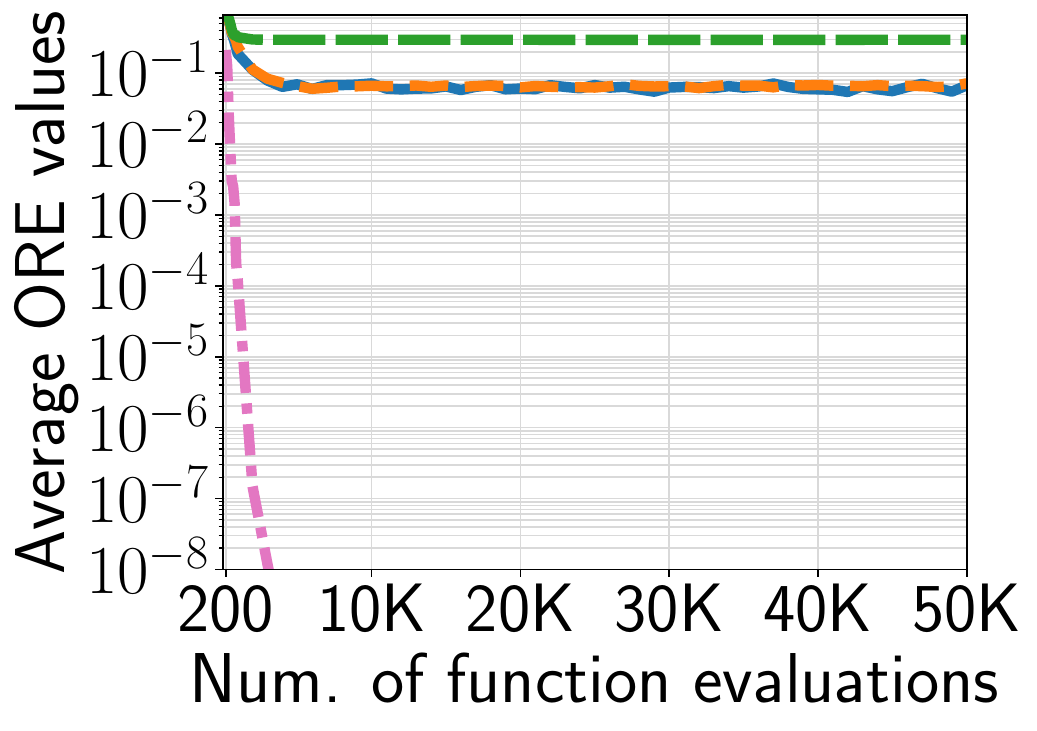}}
\\
\caption{Average $e^{\mathrm{ideal}}$, $e^{\mathrm{nadir}}$, and ORE values of the three normalization methods in r-NSGA-II on DTLZ4.}
\label{supfig:3error_rNSGA2_DTLZ4}
\end{figure*}

\begin{figure*}[t]
\centering
  \subfloat{\includegraphics[width=0.7\textwidth]{./figs/legend/legend_3.pdf}}
\vspace{-3.9mm}
   \\
   \subfloat[$e^{\mathrm{ideal}}$ ($m=2$)]{\includegraphics[width=0.32\textwidth]{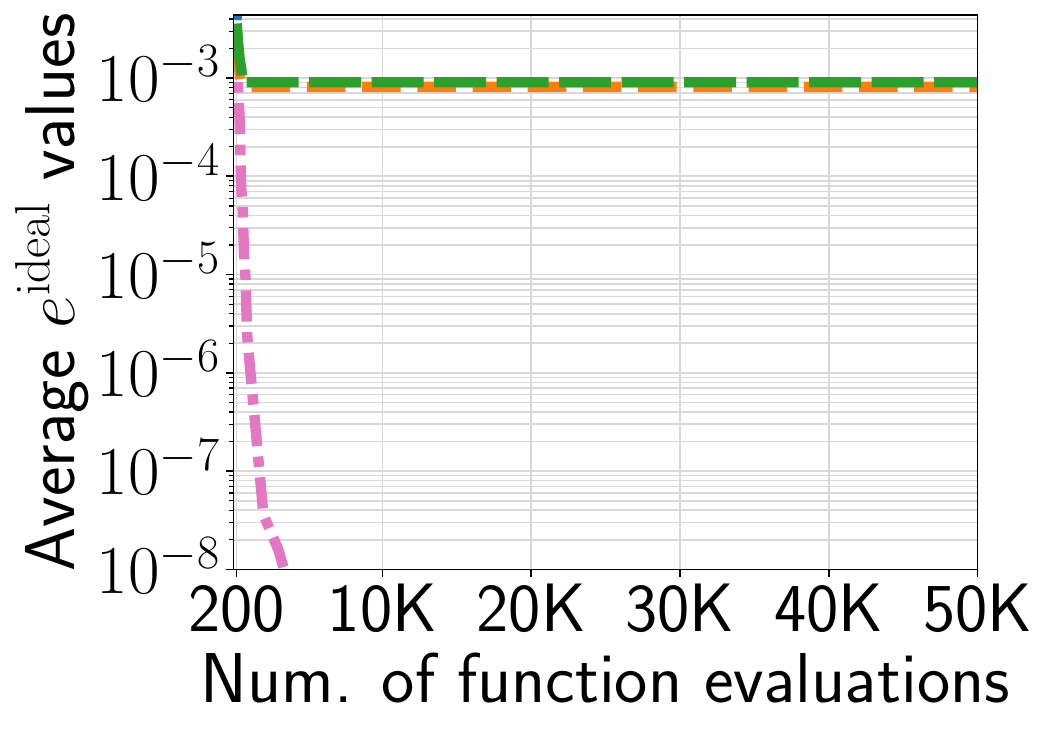}}
   \subfloat[$e^{\mathrm{ideal}}$ ($m=4$)]{\includegraphics[width=0.32\textwidth]{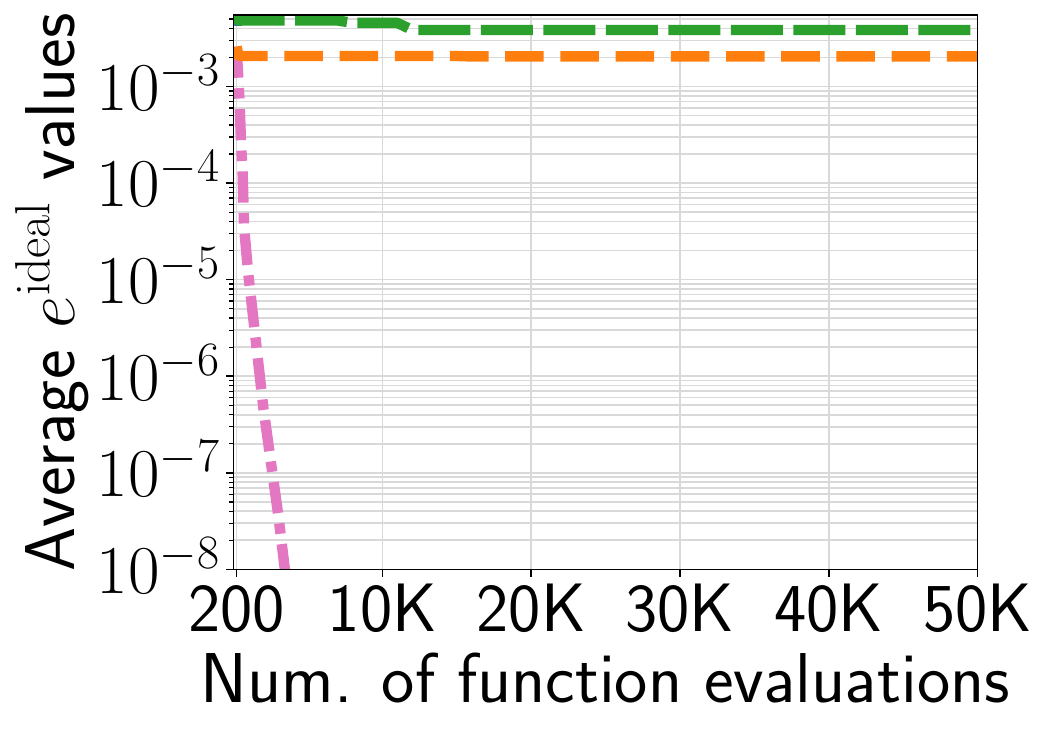}}
   \subfloat[$e^{\mathrm{ideal}}$ ($m=6$)]{\includegraphics[width=0.32\textwidth]{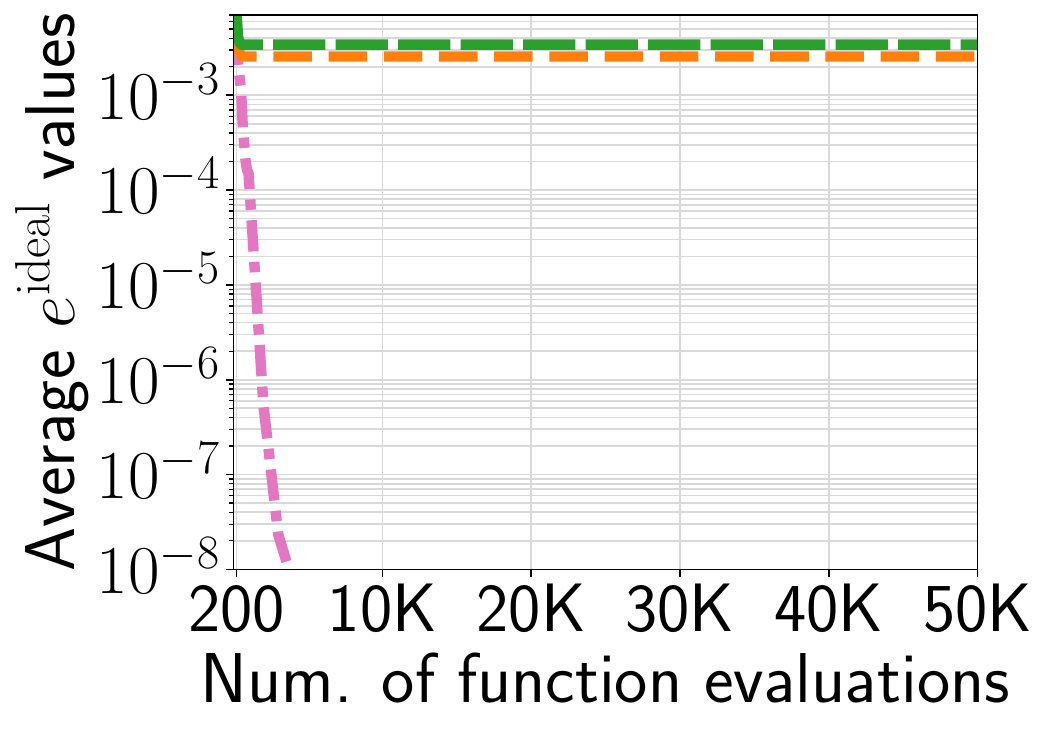}}
\\
   \subfloat[$e^{\mathrm{nadir}}$ ($m=2$)]{\includegraphics[width=0.32\textwidth]{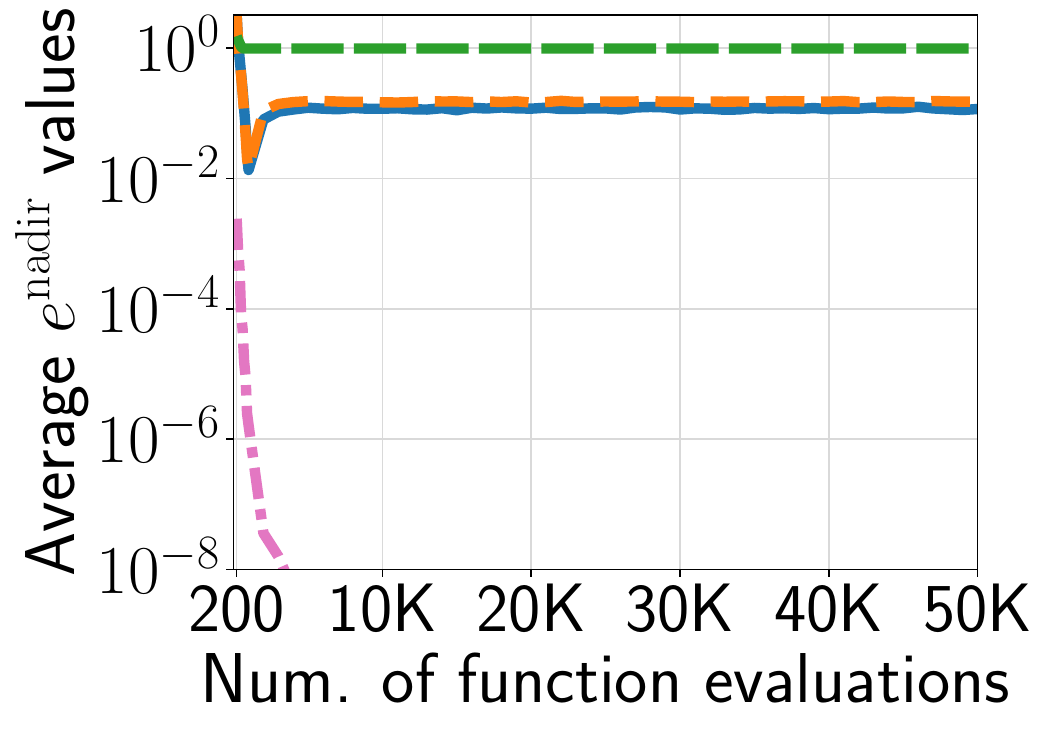}}
   \subfloat[$e^{\mathrm{nadir}}$ ($m=4$)]{\includegraphics[width=0.32\textwidth]{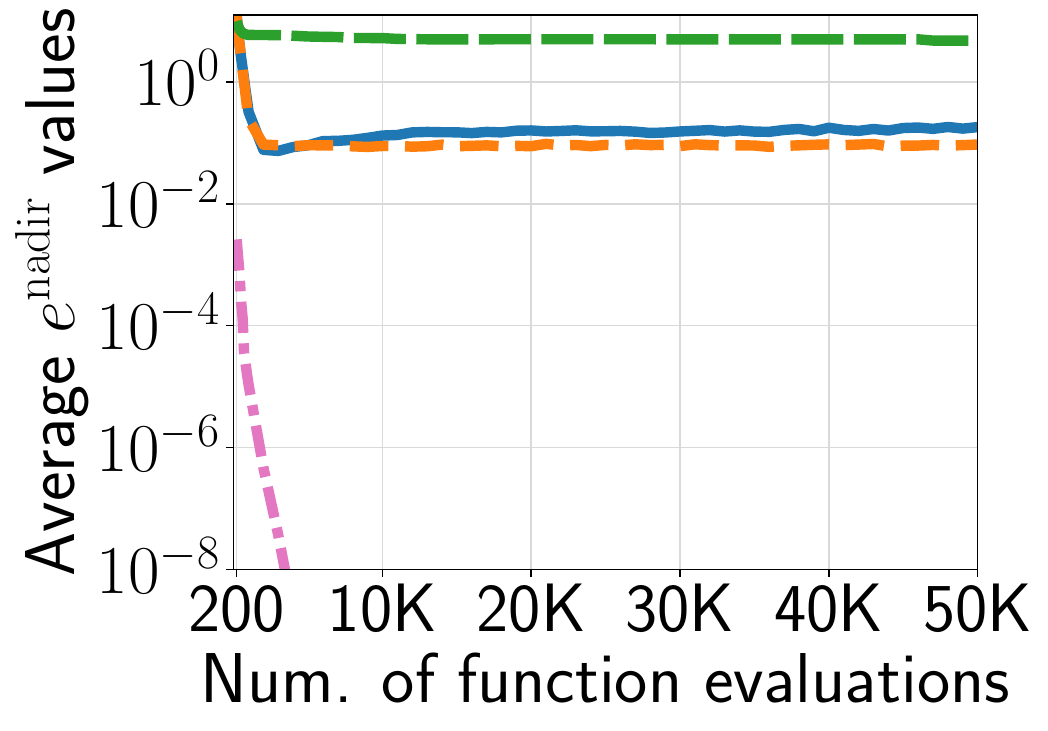}}
   \subfloat[$e^{\mathrm{nadir}}$ ($m=6$)]{\includegraphics[width=0.32\textwidth]{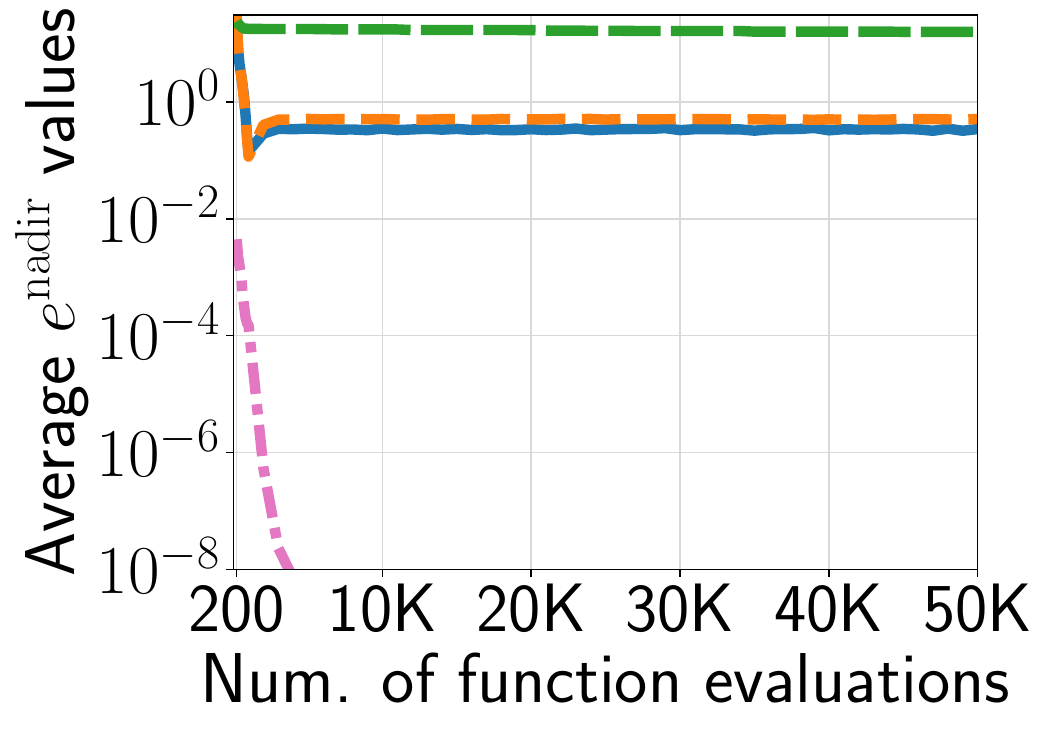}}
\\
   \subfloat[ORE ($m=2$)]{\includegraphics[width=0.32\textwidth]{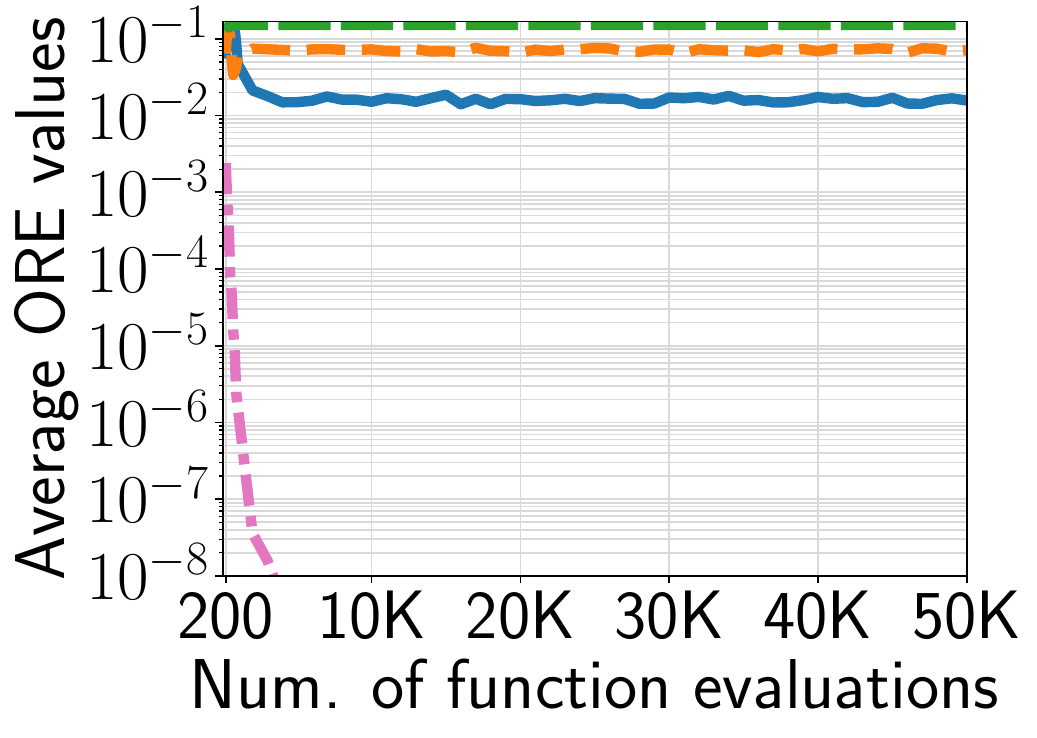}}
   \subfloat[ORE ($m=4$)]{\includegraphics[width=0.32\textwidth]{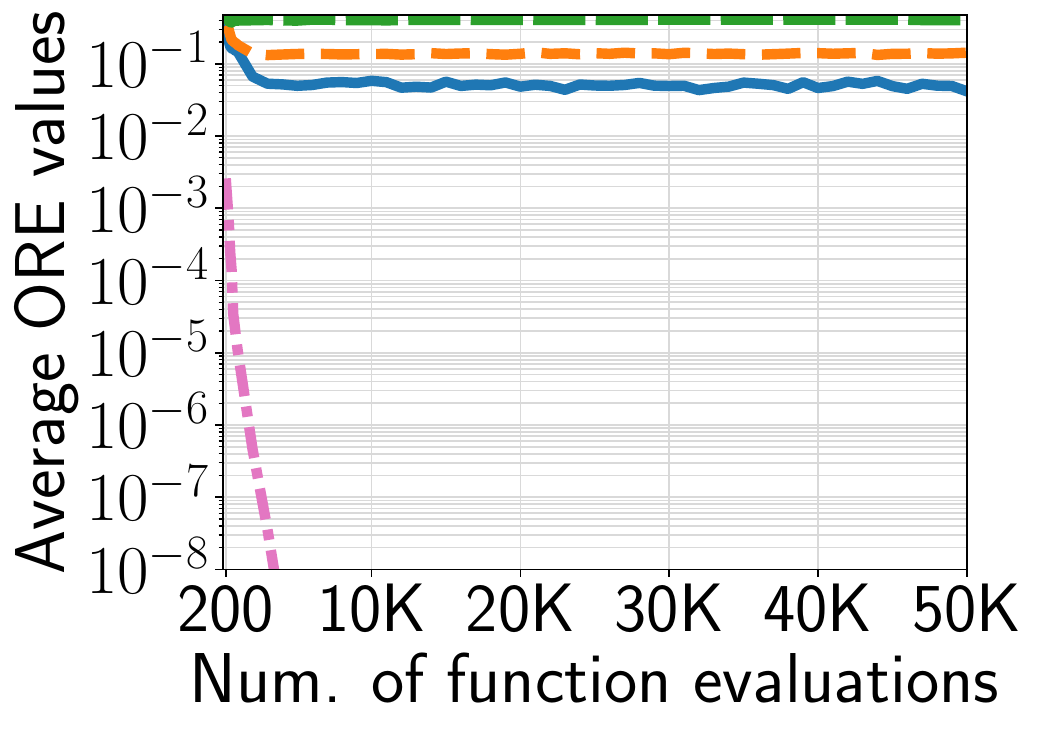}}
   \subfloat[ORE ($m=6$)]{\includegraphics[width=0.32\textwidth]{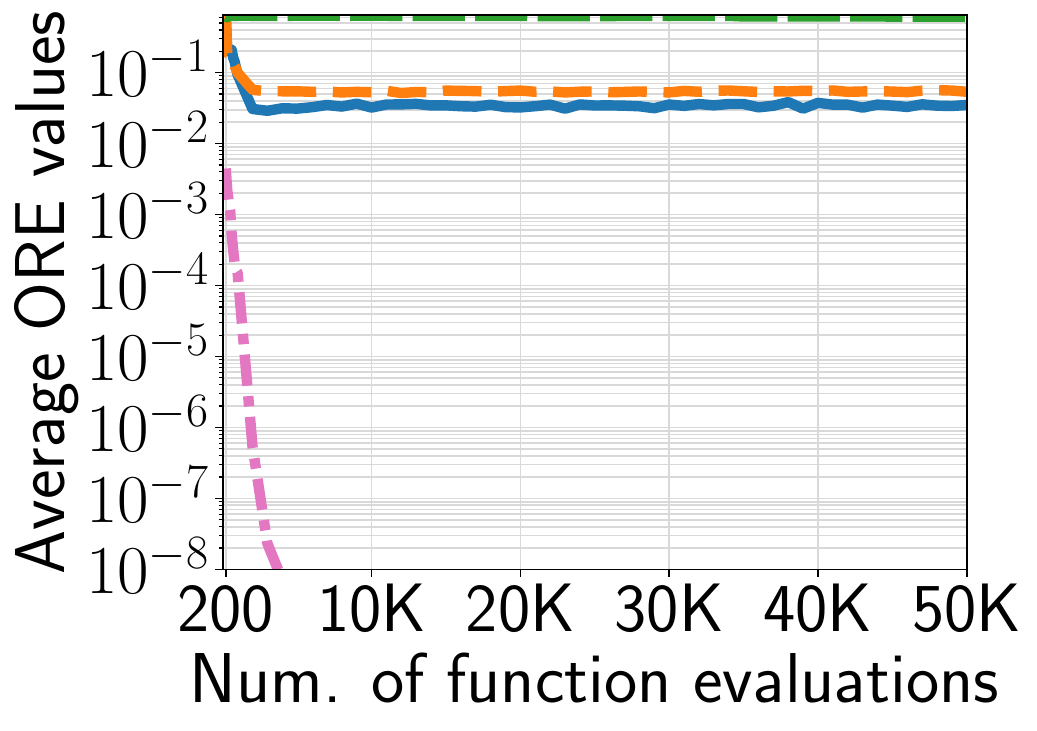}}
\\
\caption{Average $e^{\mathrm{ideal}}$, $e^{\mathrm{nadir}}$, and ORE values of the three normalization methods in r-NSGA-II on DTLZ5.}
\label{supfig:3error_rNSGA2_DTLZ5}
\end{figure*}

\begin{figure*}[t]
\centering
  \subfloat{\includegraphics[width=0.7\textwidth]{./figs/legend/legend_3.pdf}}
\vspace{-3.9mm}
   \\
   \subfloat[$e^{\mathrm{ideal}}$ ($m=2$)]{\includegraphics[width=0.32\textwidth]{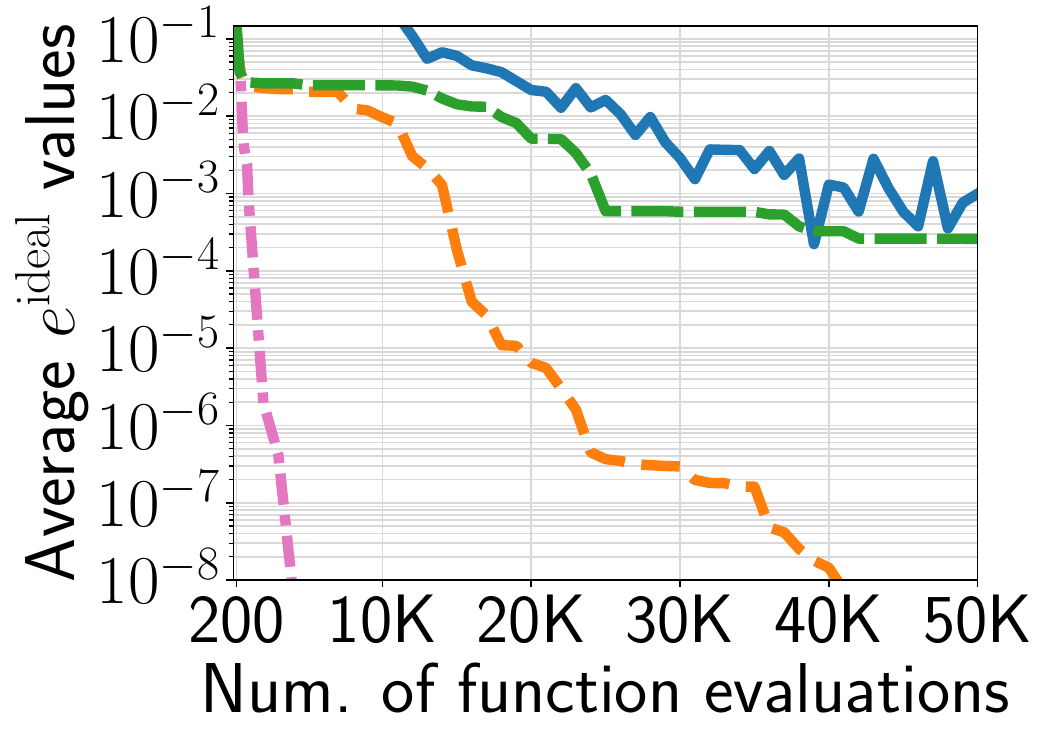}}
   \subfloat[$e^{\mathrm{ideal}}$ ($m=4$)]{\includegraphics[width=0.32\textwidth]{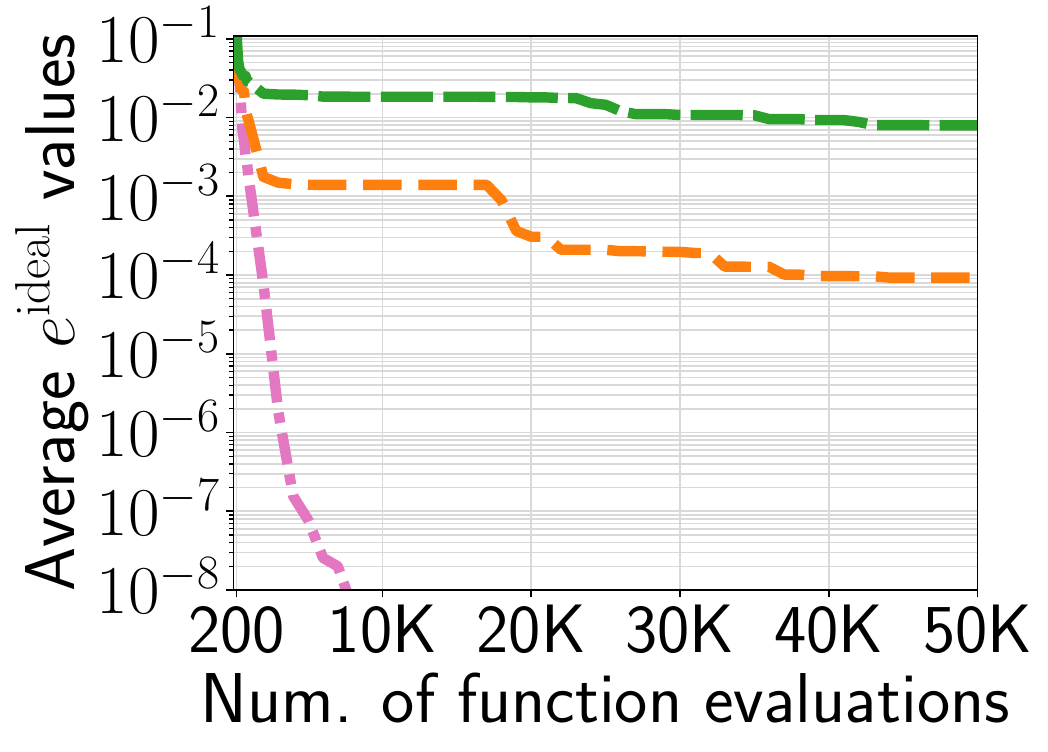}}
   \subfloat[$e^{\mathrm{ideal}}$ ($m=6$)]{\includegraphics[width=0.32\textwidth]{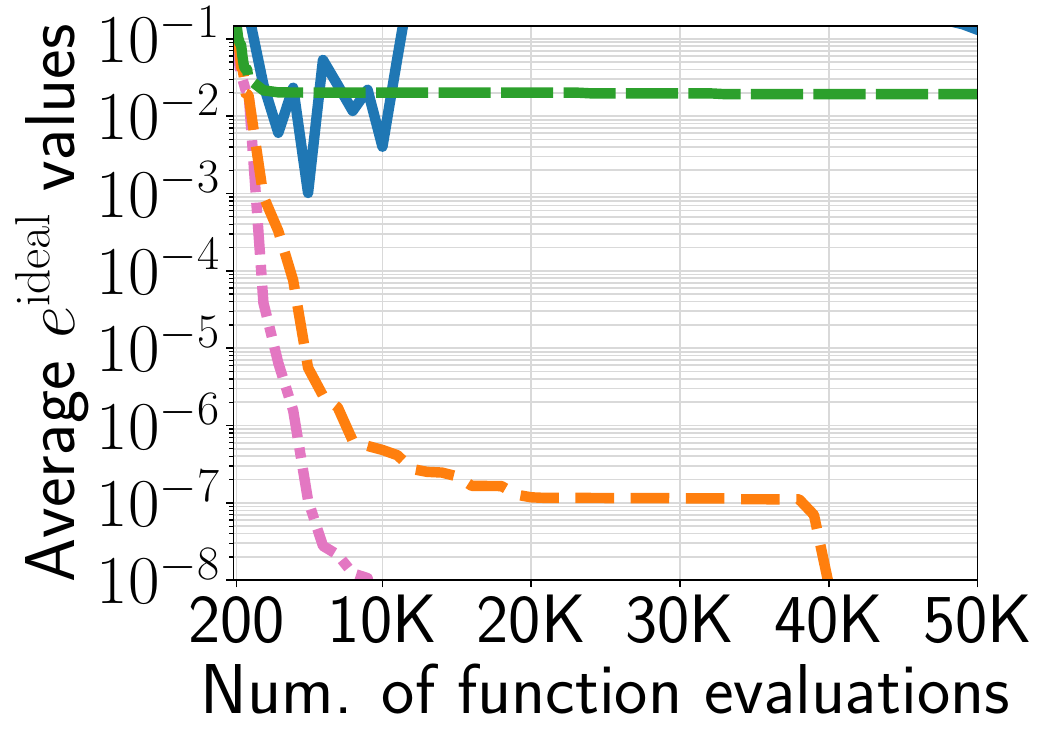}}
\\
   \subfloat[$e^{\mathrm{nadir}}$ ($m=2$)]{\includegraphics[width=0.32\textwidth]{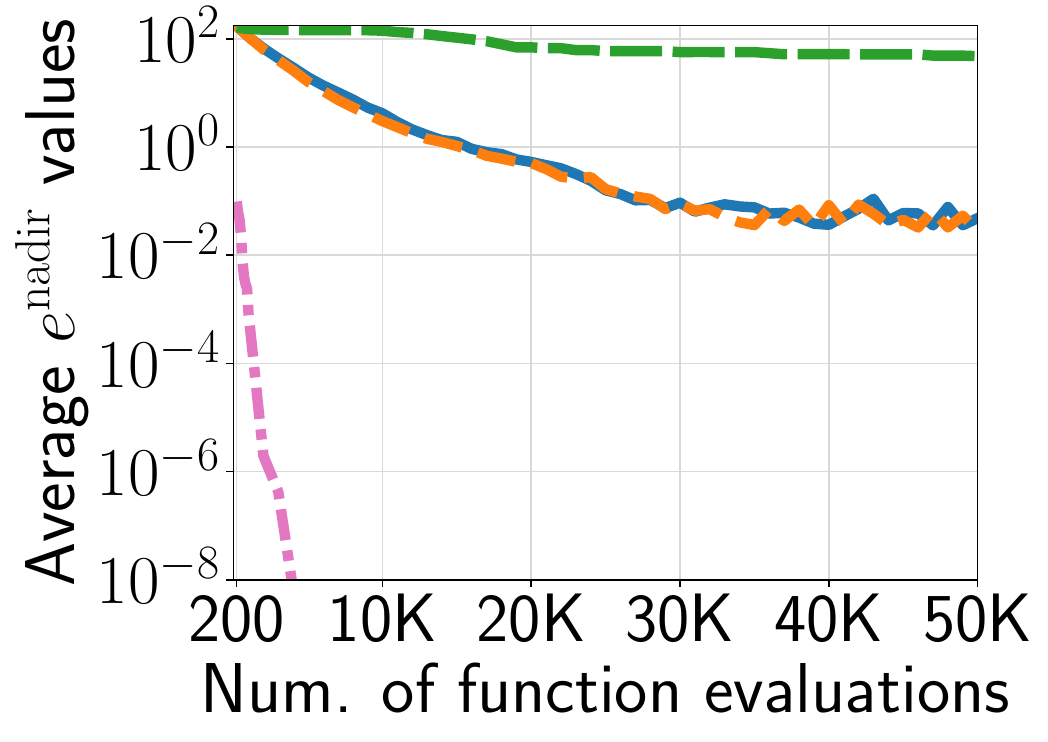}}
   \subfloat[$e^{\mathrm{nadir}}$ ($m=4$)]{\includegraphics[width=0.32\textwidth]{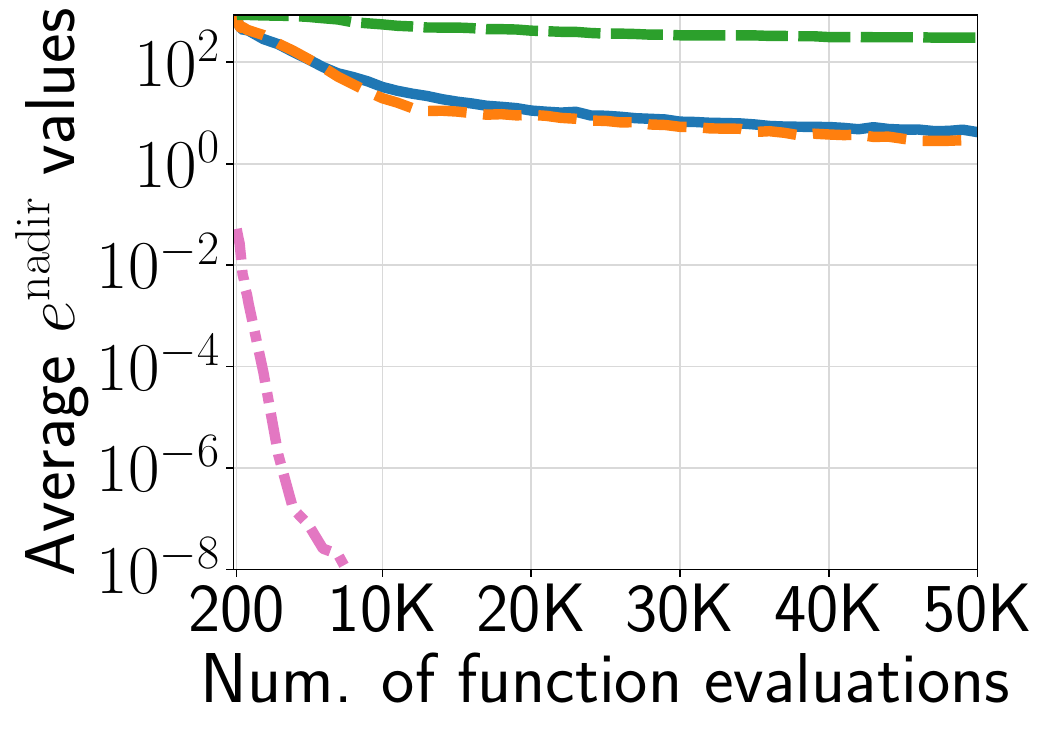}}
   \subfloat[$e^{\mathrm{nadir}}$ ($m=6$)]{\includegraphics[width=0.32\textwidth]{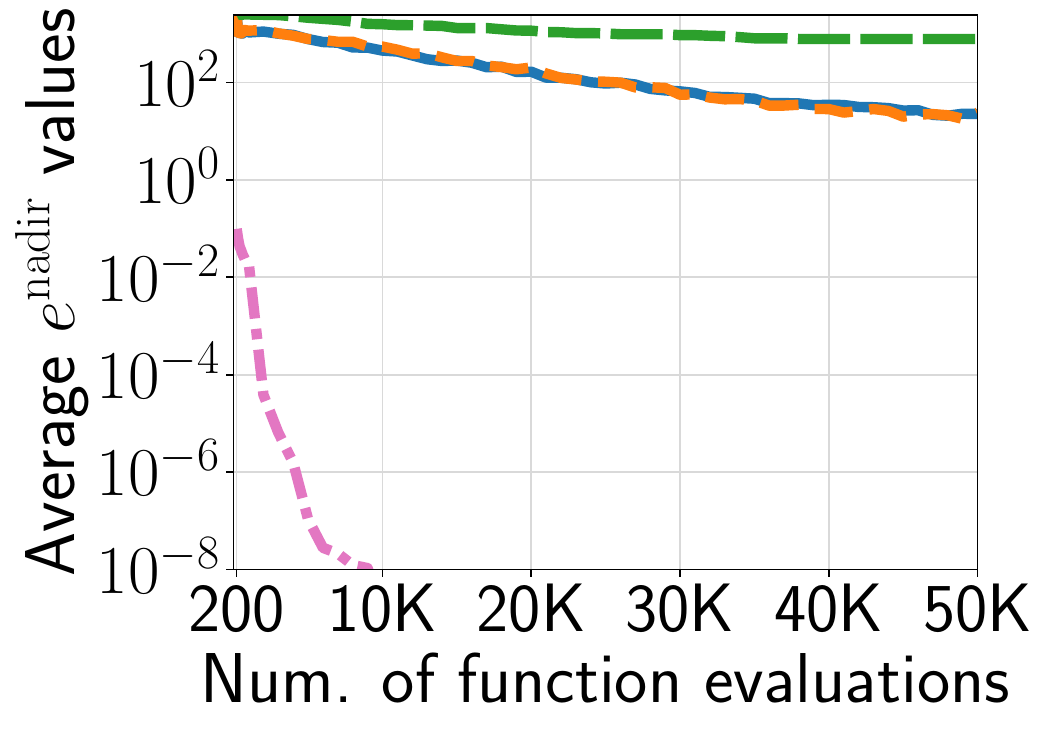}}
\\
   \subfloat[ORE ($m=2$)]{\includegraphics[width=0.32\textwidth]{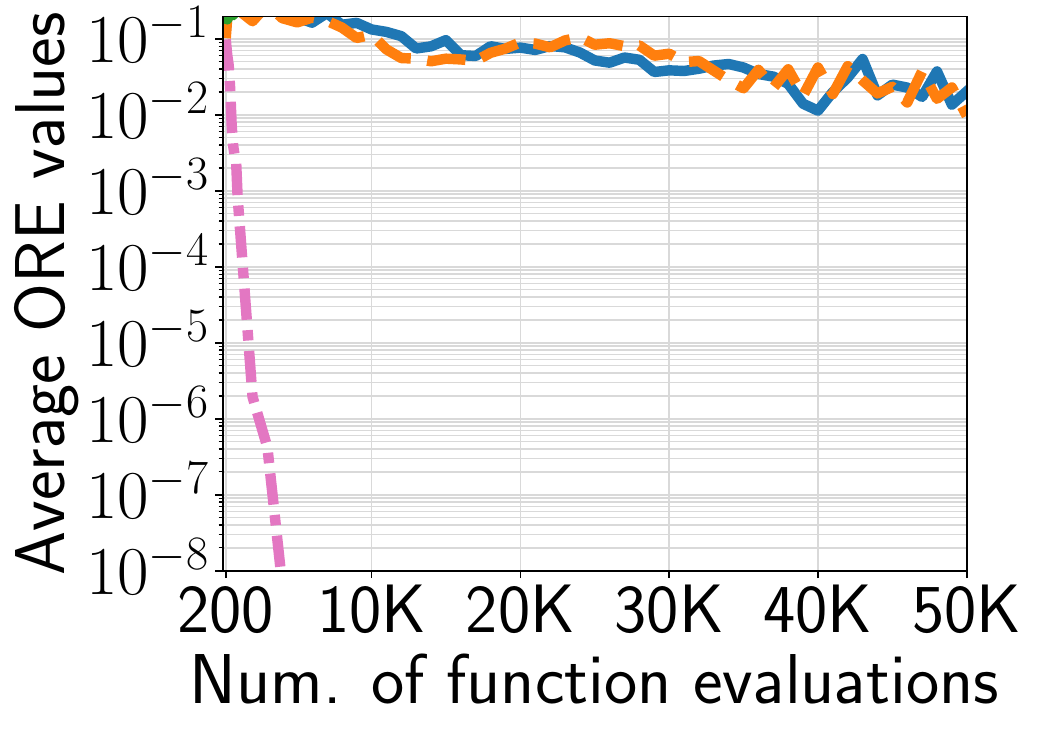}}
   \subfloat[ORE ($m=4$)]{\includegraphics[width=0.32\textwidth]{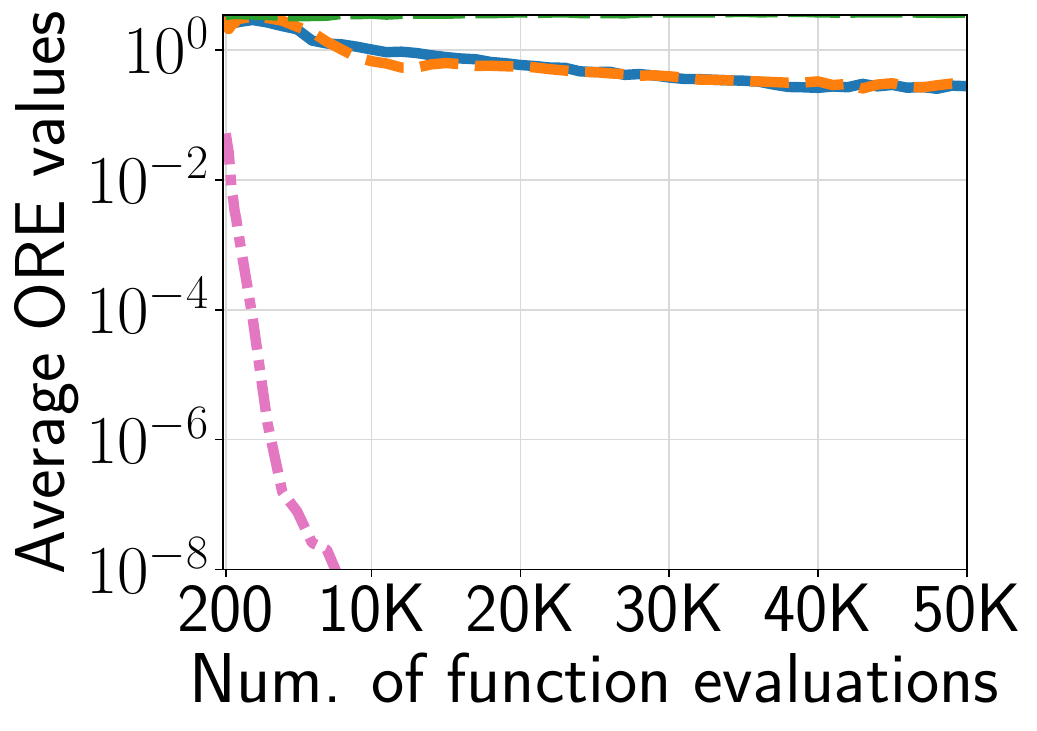}}
   \subfloat[ORE ($m=6$)]{\includegraphics[width=0.32\textwidth]{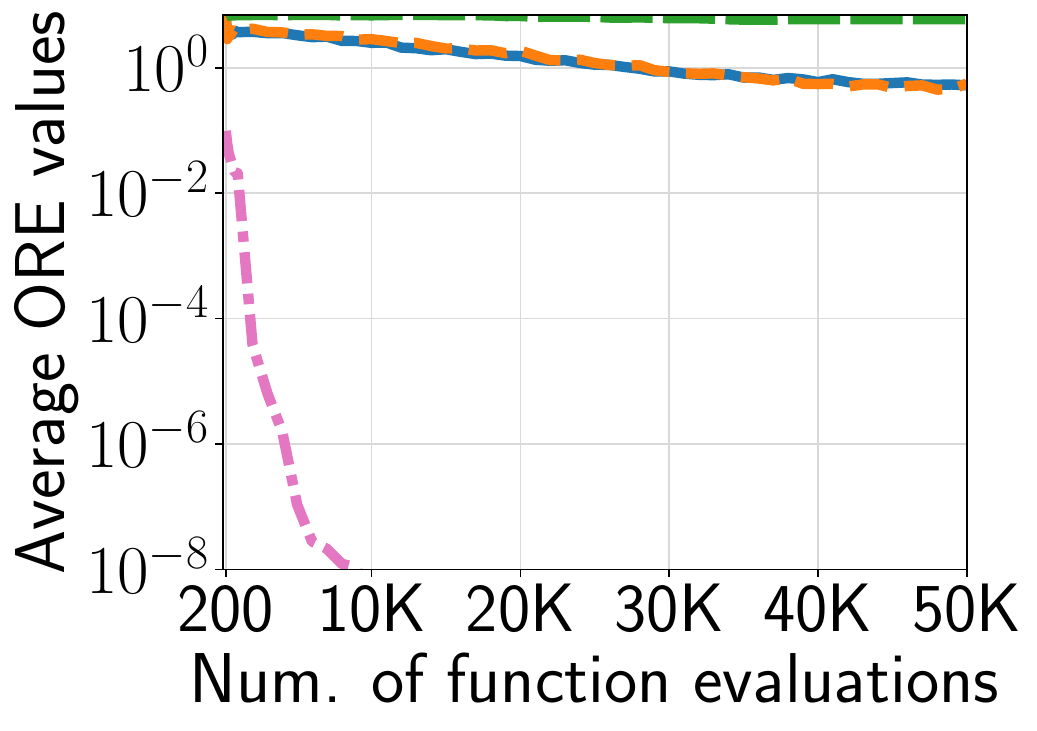}}
\\
\caption{Average $e^{\mathrm{ideal}}$, $e^{\mathrm{nadir}}$, and ORE values of the three normalization methods in r-NSGA-II on DTLZ6.}
\label{supfig:3error_rNSGA2_DTLZ6}
\end{figure*}

\begin{figure*}[t]
\centering
  \subfloat{\includegraphics[width=0.7\textwidth]{./figs/legend/legend_3.pdf}}
\vspace{-3.9mm}
   \\
   \subfloat[$e^{\mathrm{ideal}}$ ($m=2$)]{\includegraphics[width=0.32\textwidth]{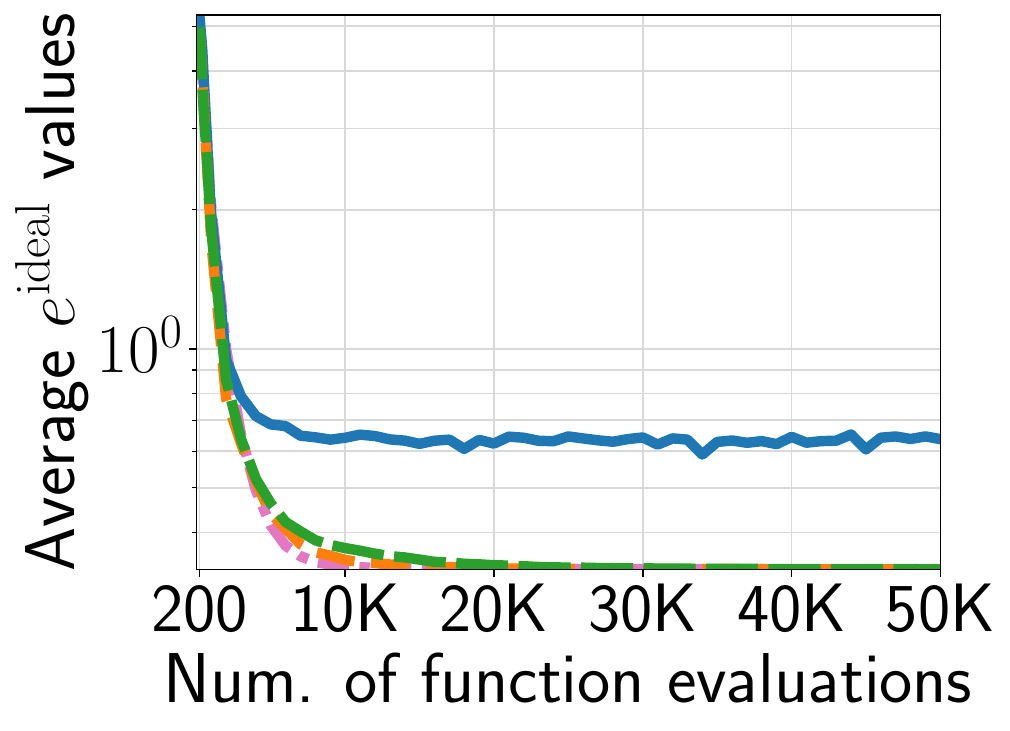}}
   \subfloat[$e^{\mathrm{ideal}}$ ($m=4$)]{\includegraphics[width=0.32\textwidth]{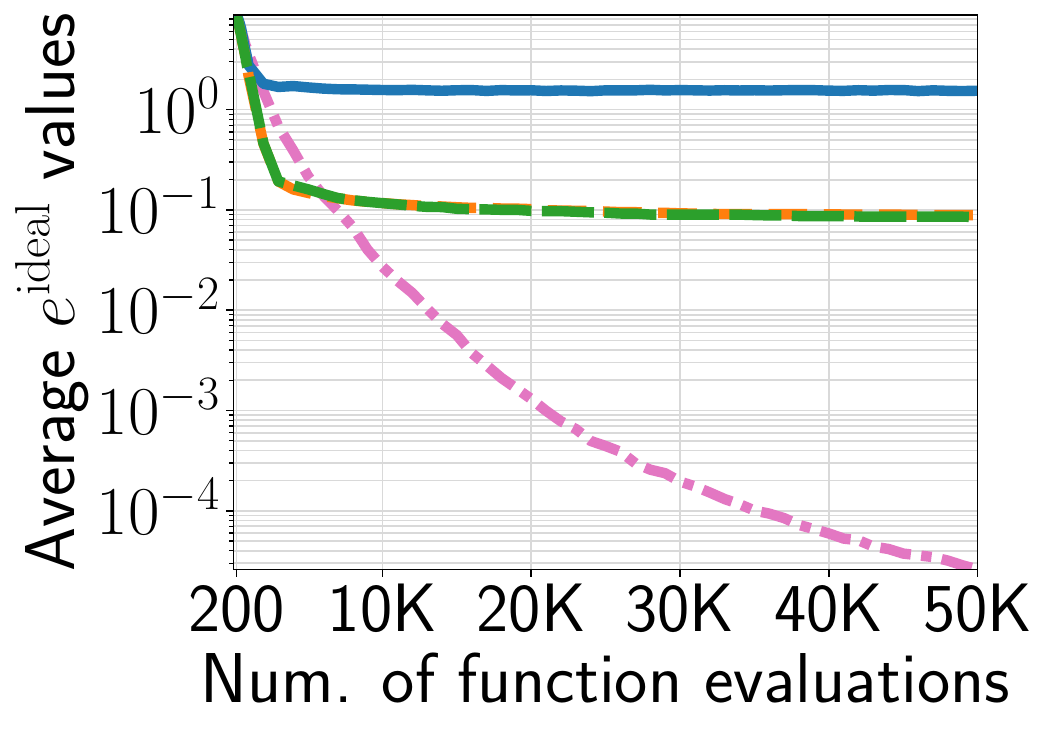}}
   \subfloat[$e^{\mathrm{ideal}}$ ($m=6$)]{\includegraphics[width=0.32\textwidth]{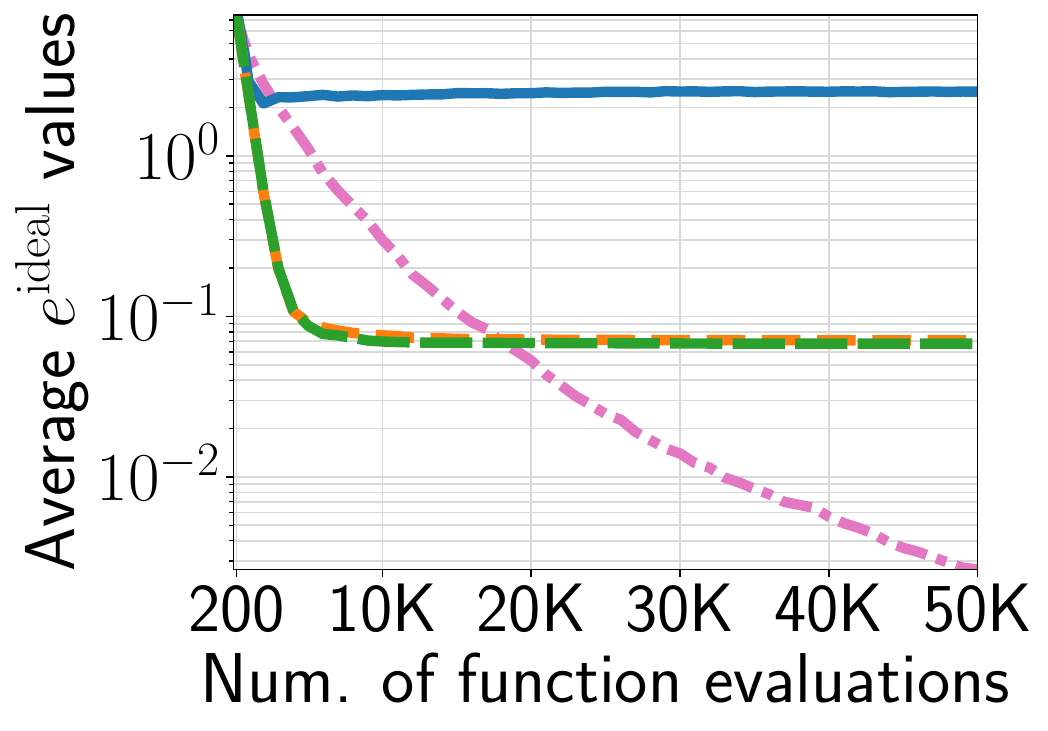}}
\\
   \subfloat[$e^{\mathrm{nadir}}$ ($m=2$)]{\includegraphics[width=0.32\textwidth]{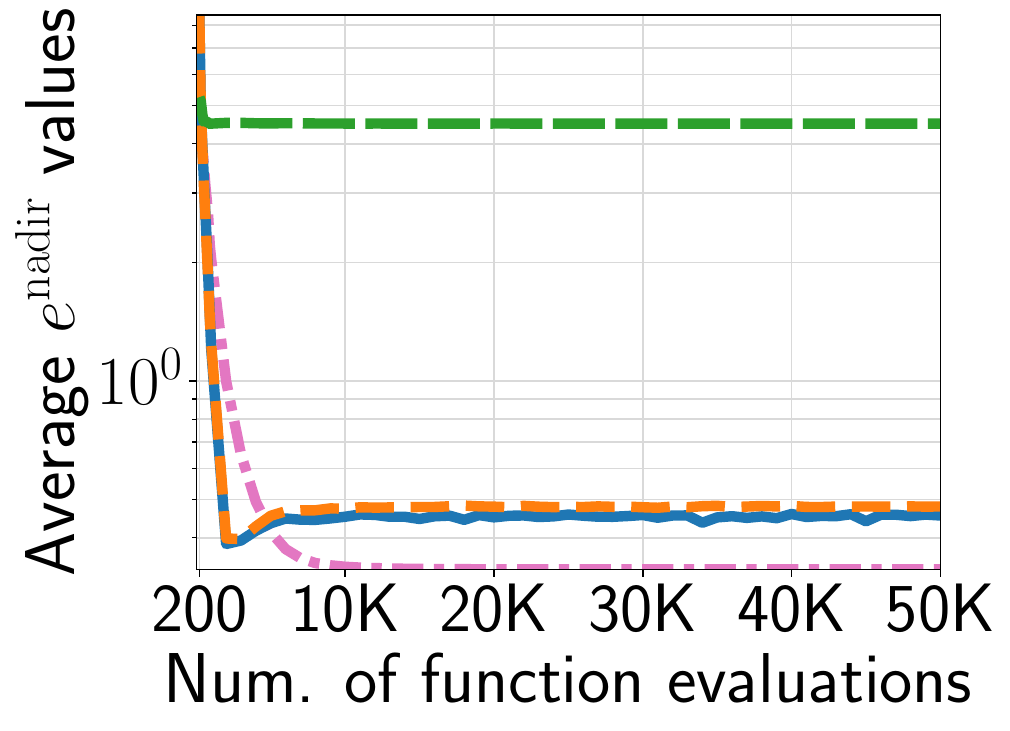}}
   \subfloat[$e^{\mathrm{nadir}}$ ($m=4$)]{\includegraphics[width=0.32\textwidth]{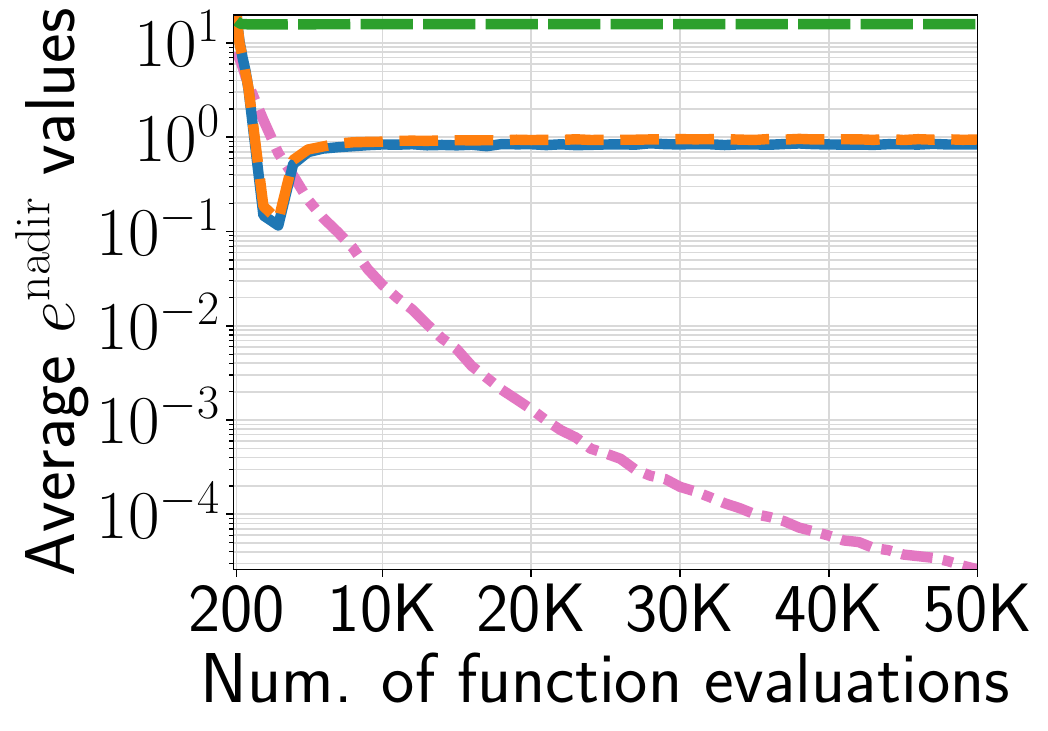}}
   \subfloat[$e^{\mathrm{nadir}}$ ($m=6$)]{\includegraphics[width=0.32\textwidth]{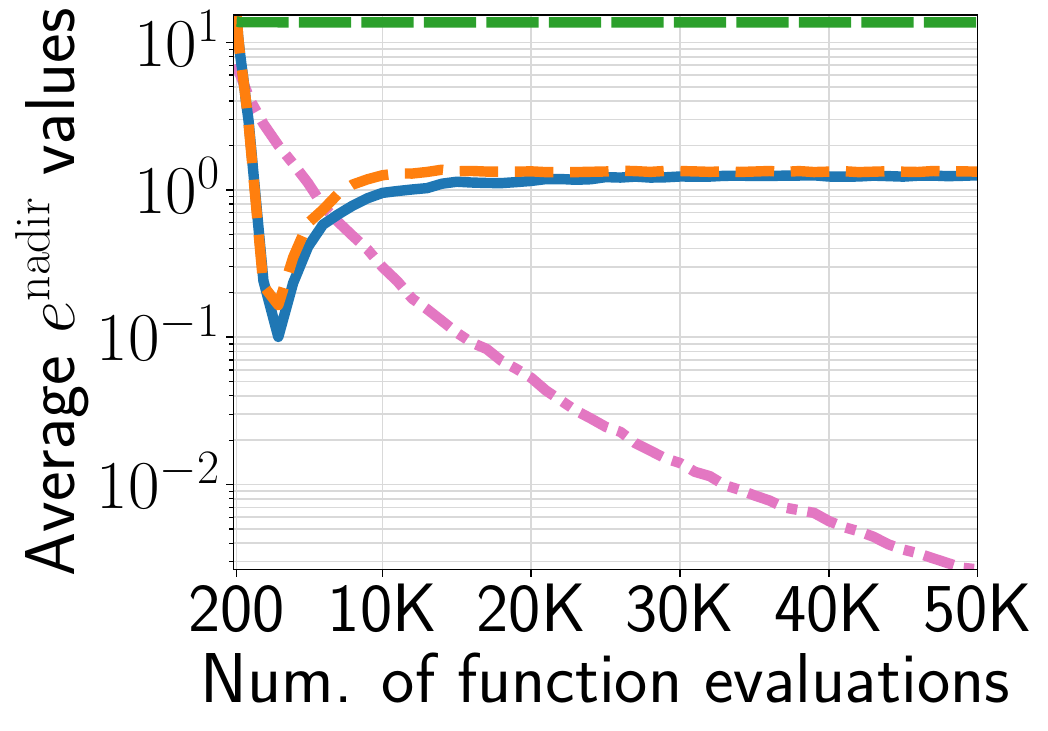}}
\\
   \subfloat[ORE ($m=2$)]{\includegraphics[width=0.32\textwidth]{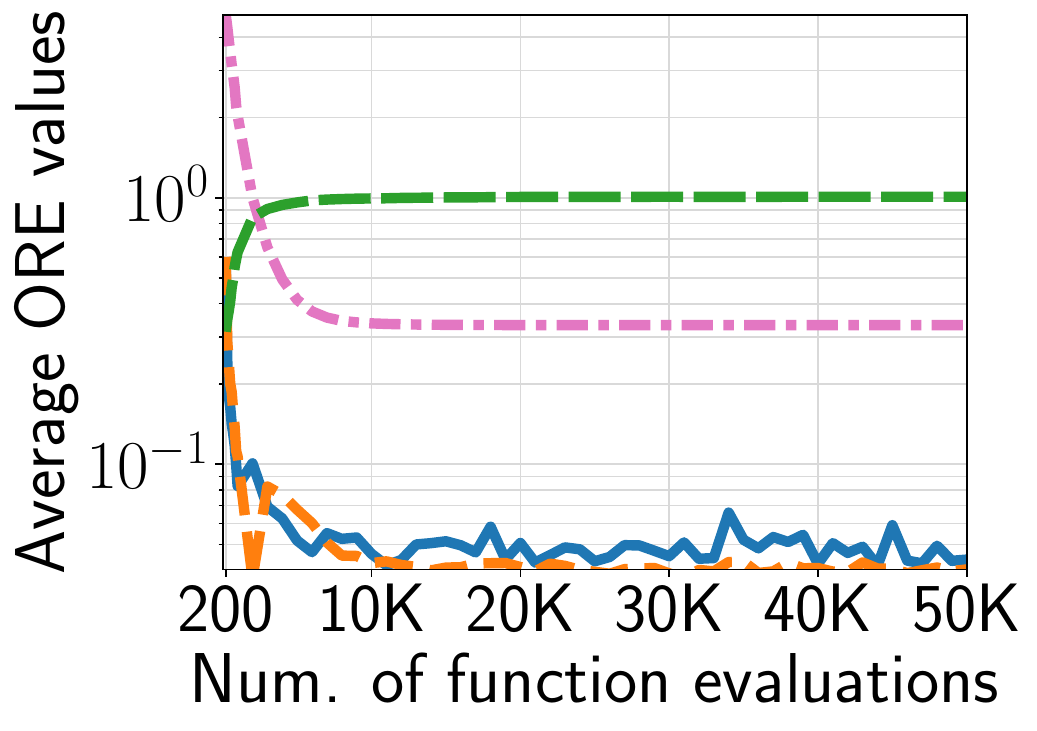}}
   \subfloat[ORE ($m=4$)]{\includegraphics[width=0.32\textwidth]{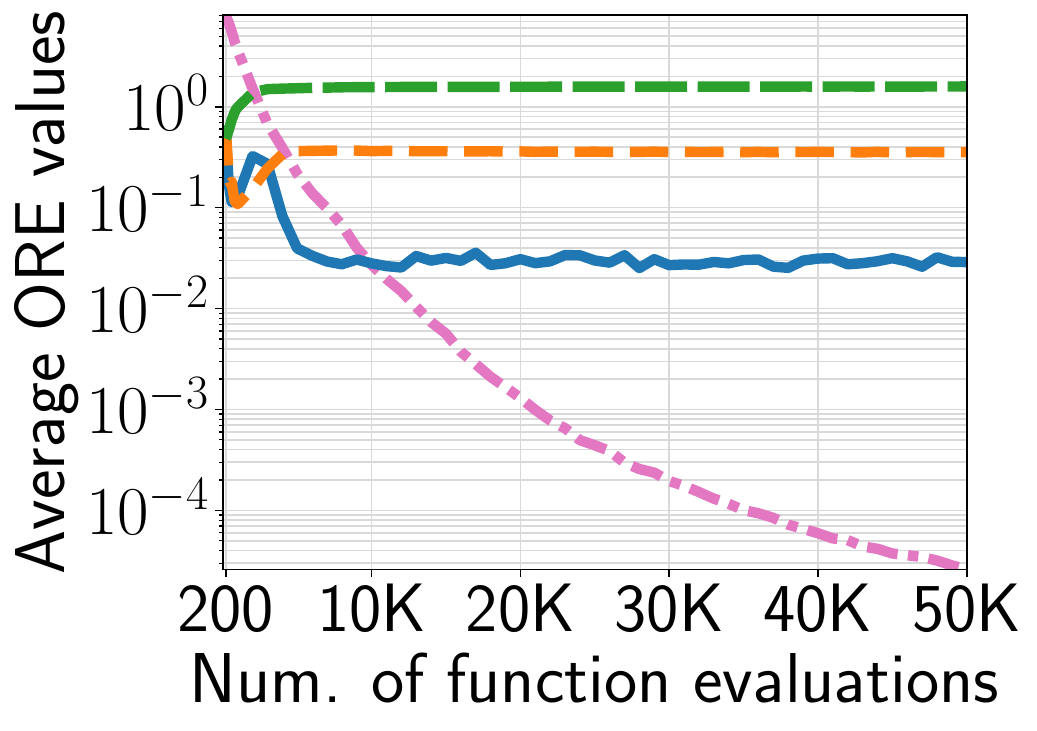}}
   \subfloat[ORE ($m=6$)]{\includegraphics[width=0.32\textwidth]{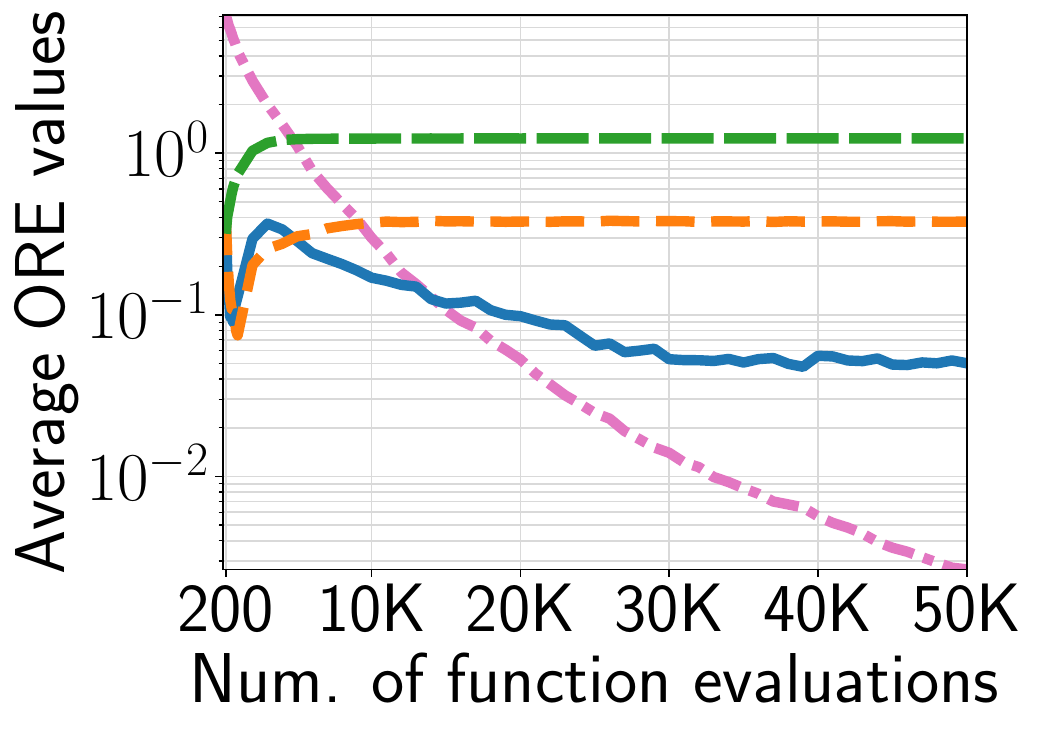}}
\\
\caption{Average $e^{\mathrm{ideal}}$, $e^{\mathrm{nadir}}$, and ORE values of the three normalization methods in r-NSGA-II on DTLZ7.}
\label{supfig:3error_rNSGA2_DTLZ7}
\end{figure*}

\begin{figure*}[t]
\centering
  \subfloat{\includegraphics[width=0.7\textwidth]{./figs/legend/legend_3.pdf}}
\vspace{-3.9mm}
   \\
   \subfloat[$e^{\mathrm{ideal}}$ ($m=2$)]{\includegraphics[width=0.32\textwidth]{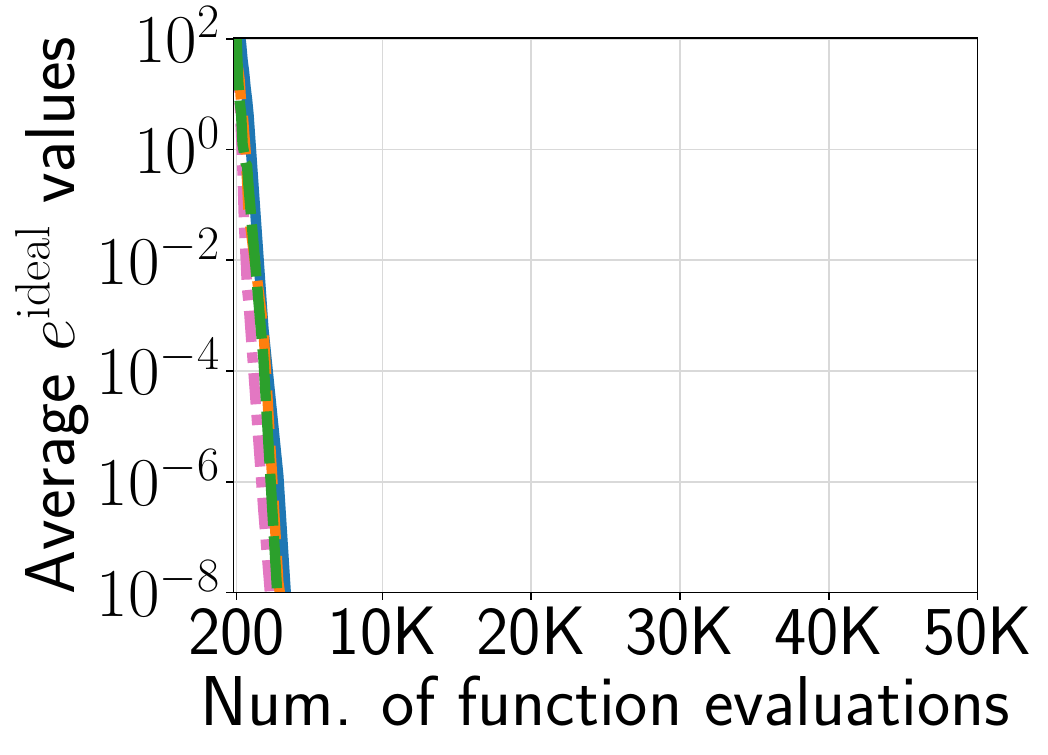}}
   \subfloat[$e^{\mathrm{ideal}}$ ($m=4$)]{\includegraphics[width=0.32\textwidth]{./figs/qi_error_ideal/rNSGA2_mu100/SDTLZ1_m4_r0.1_z-type1.pdf}}
   \subfloat[$e^{\mathrm{ideal}}$ ($m=6$)]{\includegraphics[width=0.32\textwidth]{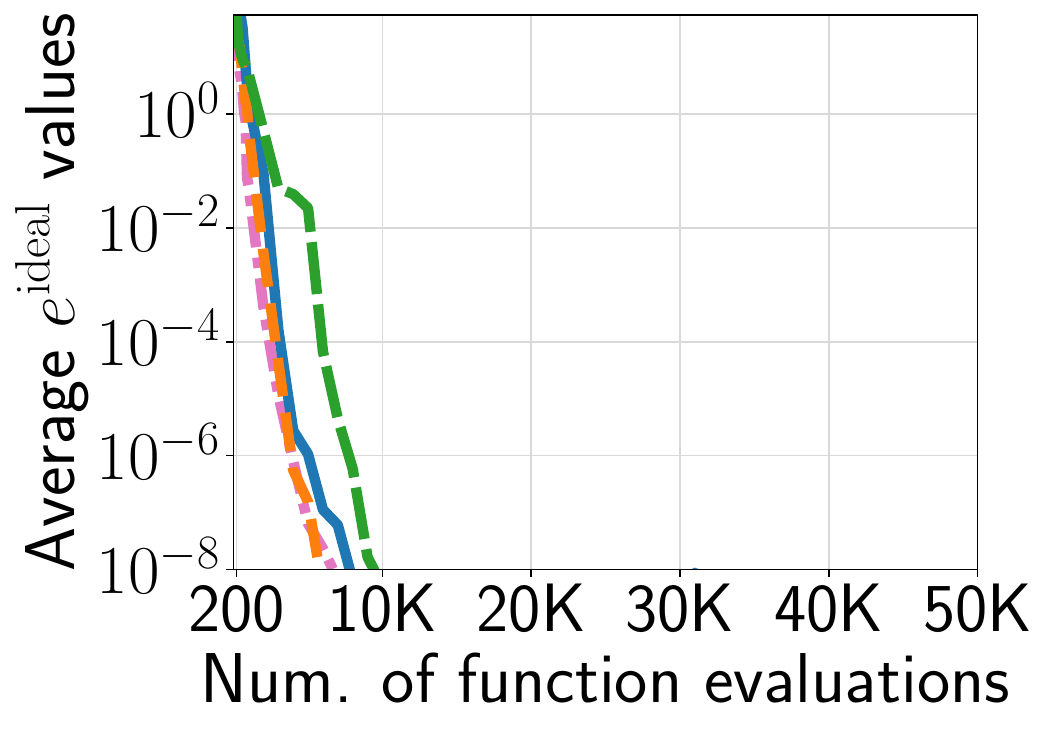}}
\\
   \subfloat[$e^{\mathrm{nadir}}$ ($m=2$)]{\includegraphics[width=0.32\textwidth]{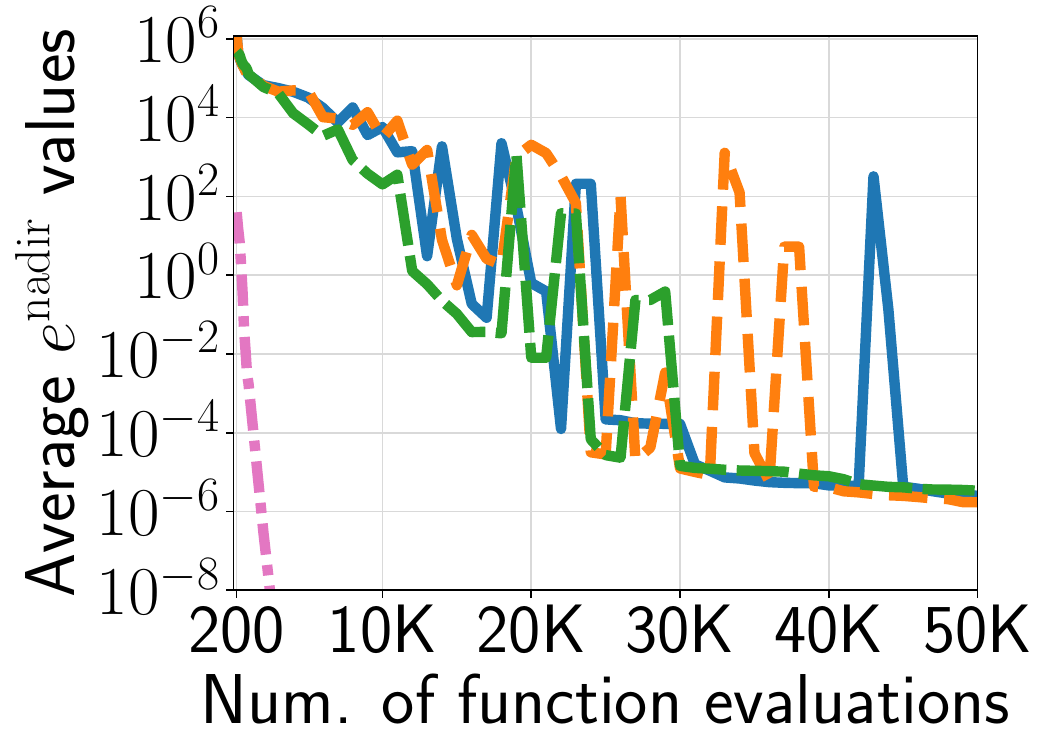}}
   \subfloat[$e^{\mathrm{nadir}}$ ($m=4$)]{\includegraphics[width=0.32\textwidth]{./figs/qi_error_nadir/rNSGA2_mu100/SDTLZ1_m4_r0.1_z-type1.pdf}}
   \subfloat[$e^{\mathrm{nadir}}$ ($m=6$)]{\includegraphics[width=0.32\textwidth]{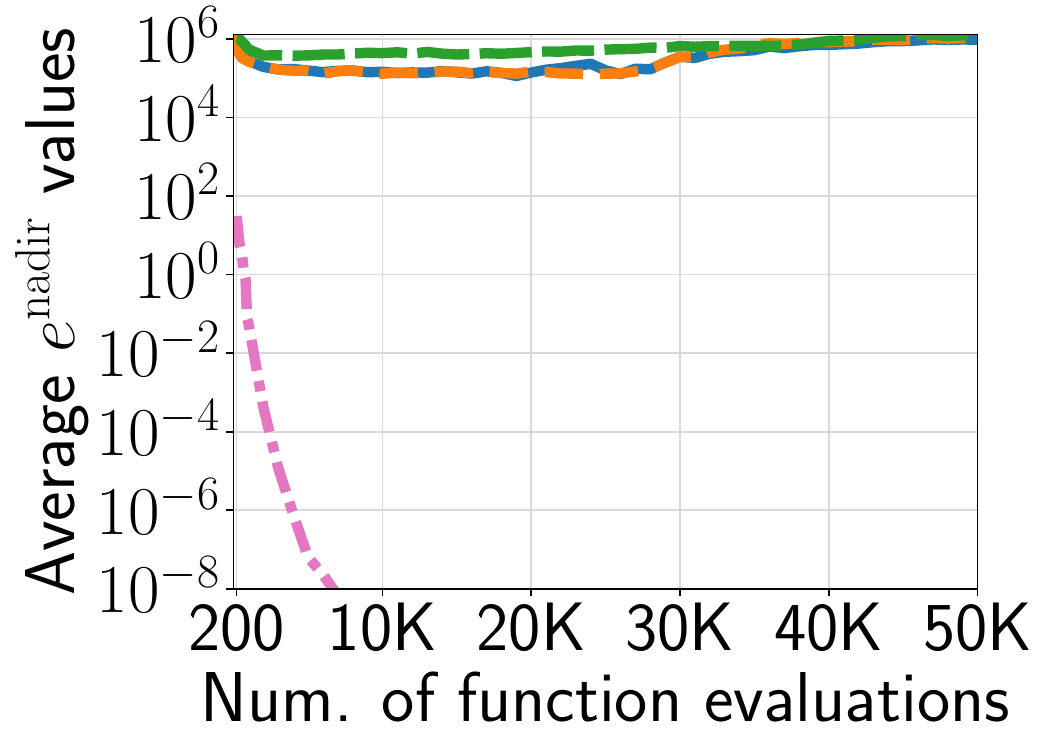}}
\\
   \subfloat[ORE ($m=2$)]{\includegraphics[width=0.32\textwidth]{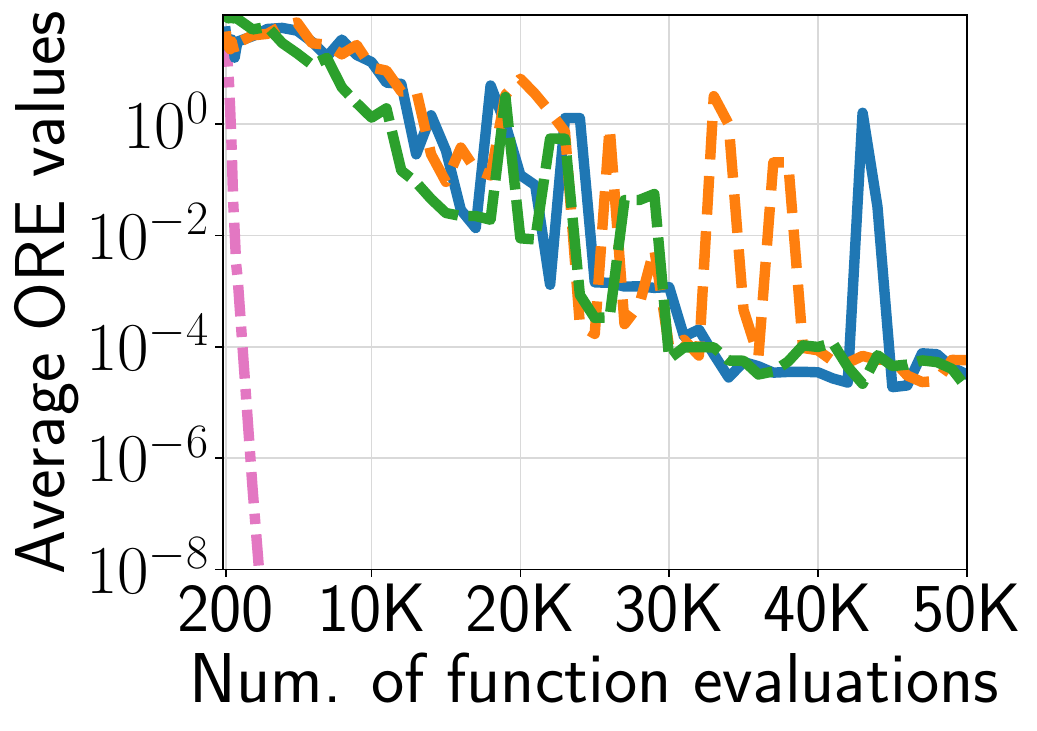}}
   \subfloat[ORE ($m=4$)]{\includegraphics[width=0.32\textwidth]{./figs/qi_ore/rNSGA2_mu100/SDTLZ1_m4_r0.1_z-type1.pdf}}
   \subfloat[ORE ($m=6$)]{\includegraphics[width=0.32\textwidth]{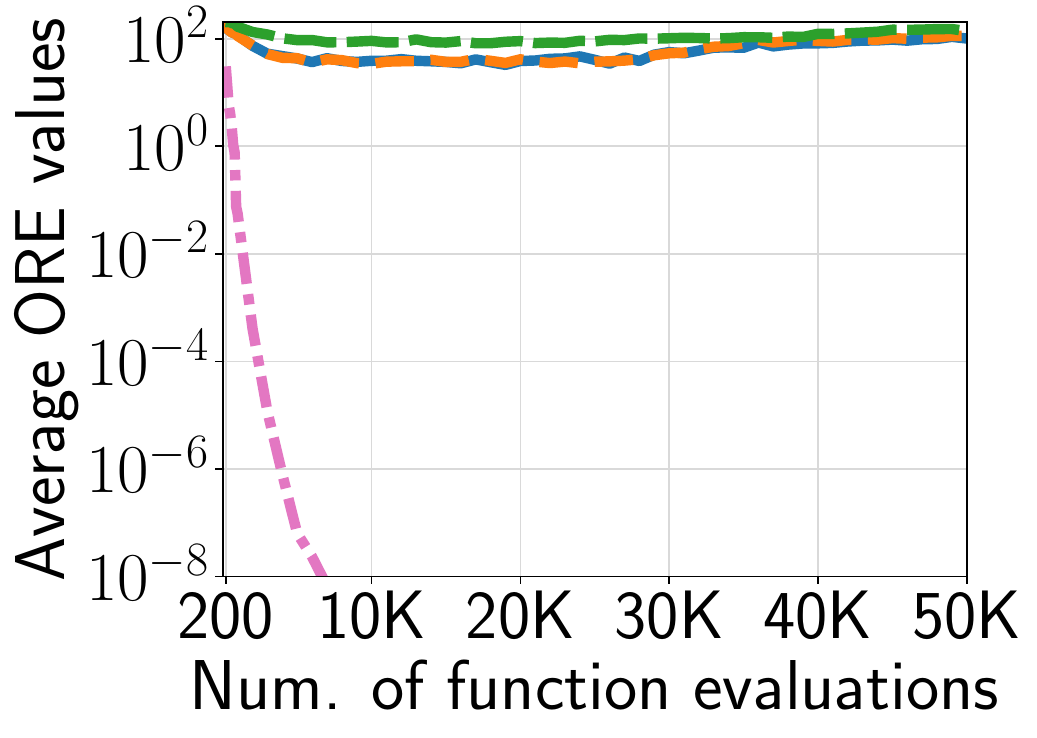}}
\\
\caption{Average $e^{\mathrm{ideal}}$, $e^{\mathrm{nadir}}$, and ORE values of the three normalization methods in r-NSGA-II on SDTLZ1.}
\label{supfig:3error_rNSGA2_SDTLZ1}
\end{figure*}

\begin{figure*}[t]
\centering
  \subfloat{\includegraphics[width=0.7\textwidth]{./figs/legend/legend_3.pdf}}
\vspace{-3.9mm}
   \\
   \subfloat[$e^{\mathrm{ideal}}$ ($m=2$)]{\includegraphics[width=0.32\textwidth]{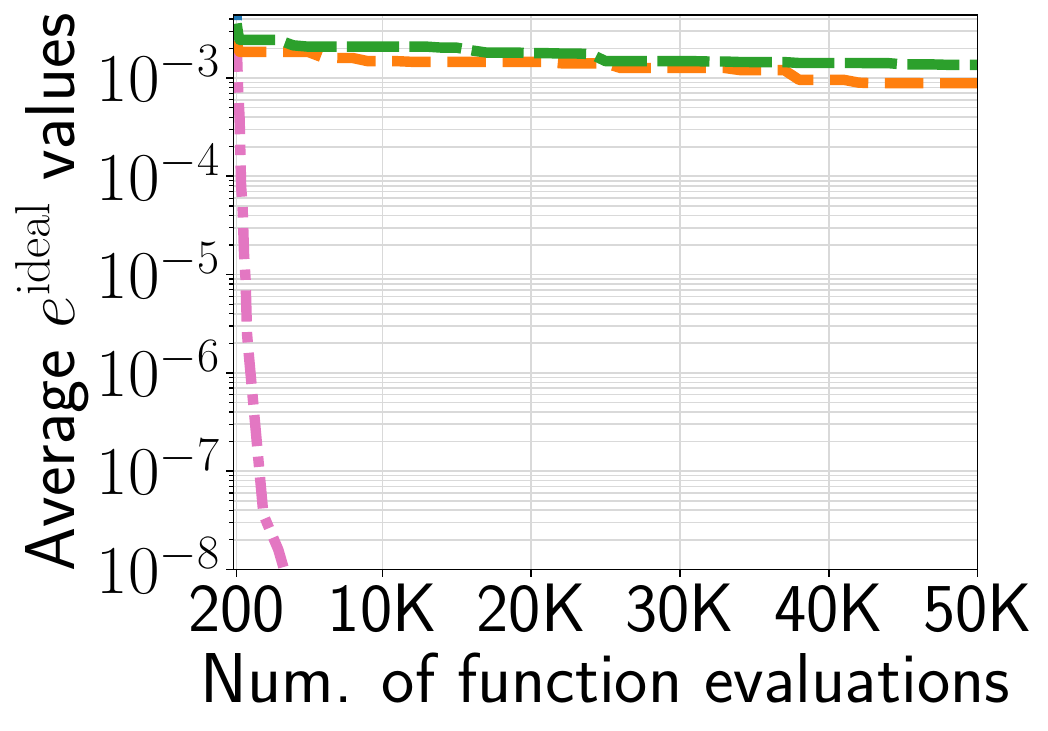}}
   \subfloat[$e^{\mathrm{ideal}}$ ($m=4$)]{\includegraphics[width=0.32\textwidth]{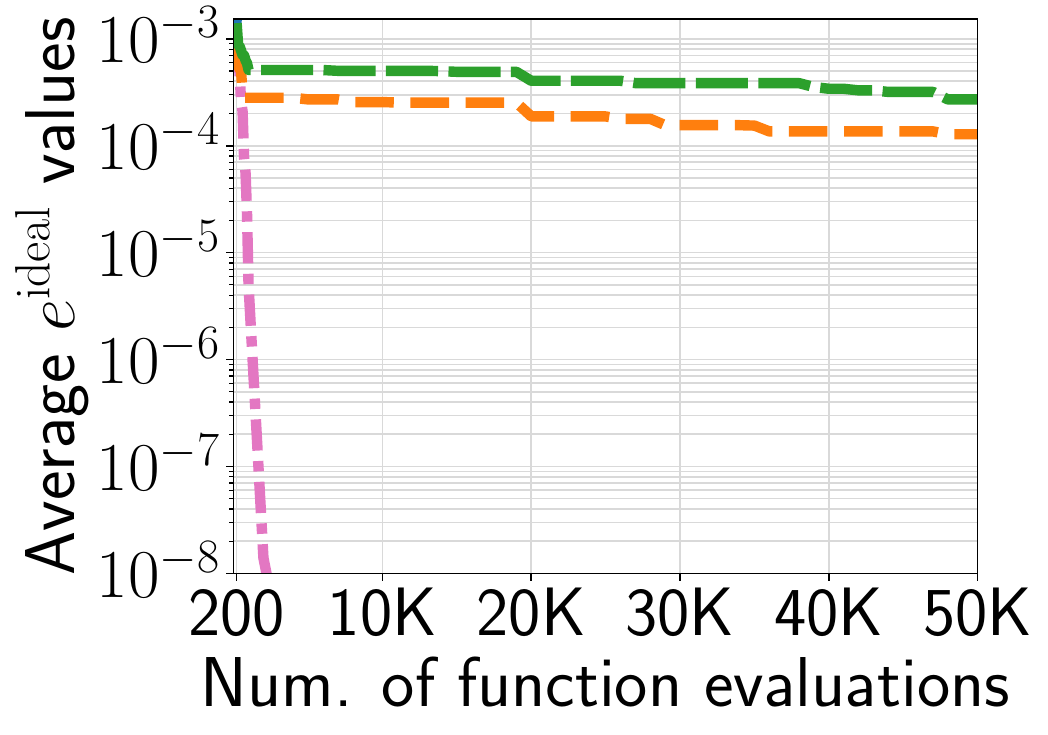}}
   \subfloat[$e^{\mathrm{ideal}}$ ($m=6$)]{\includegraphics[width=0.32\textwidth]{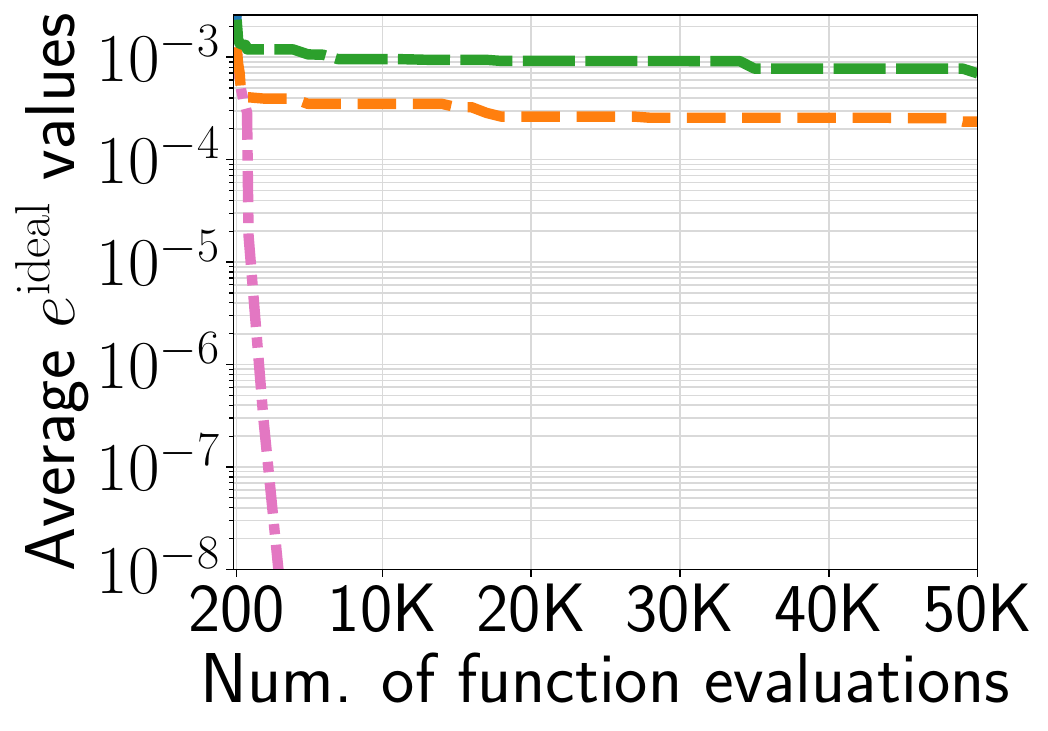}}
\\
   \subfloat[$e^{\mathrm{nadir}}$ ($m=2$)]{\includegraphics[width=0.32\textwidth]{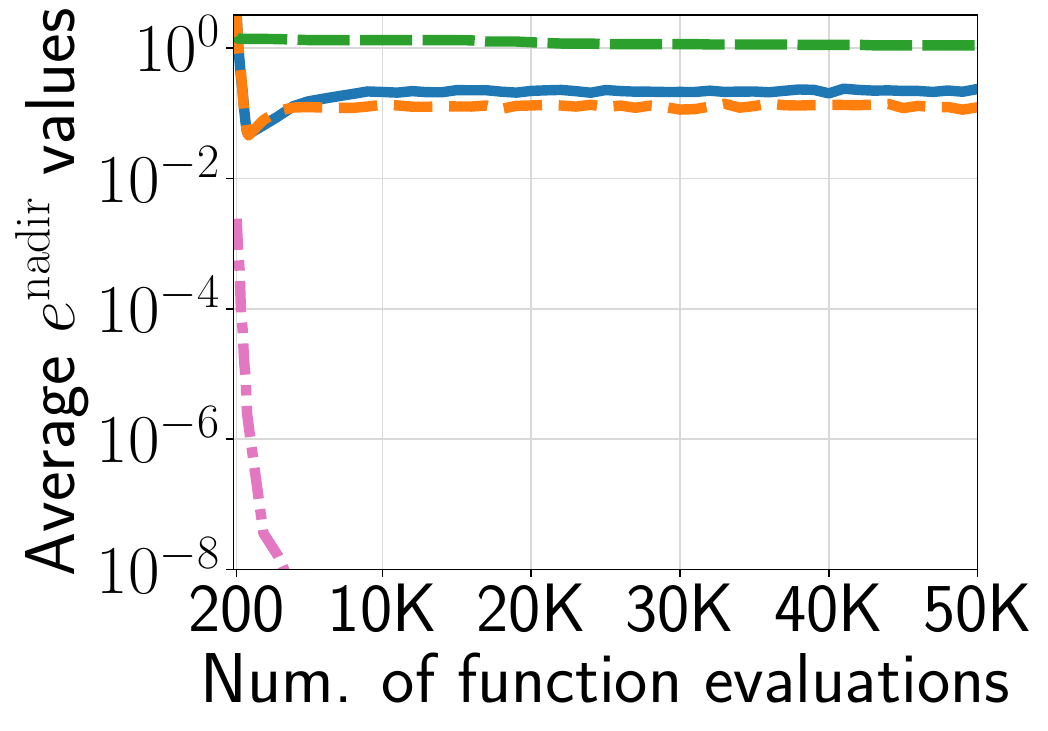}}
   \subfloat[$e^{\mathrm{nadir}}$ ($m=4$)]{\includegraphics[width=0.32\textwidth]{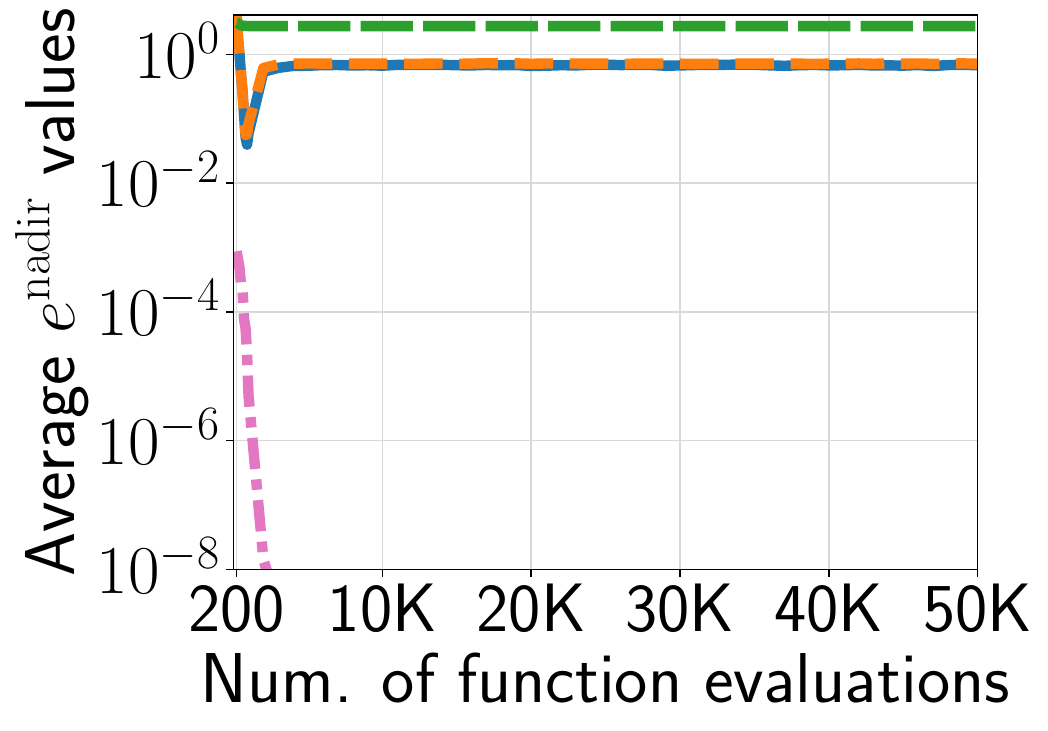}}
   \subfloat[$e^{\mathrm{nadir}}$ ($m=6$)]{\includegraphics[width=0.32\textwidth]{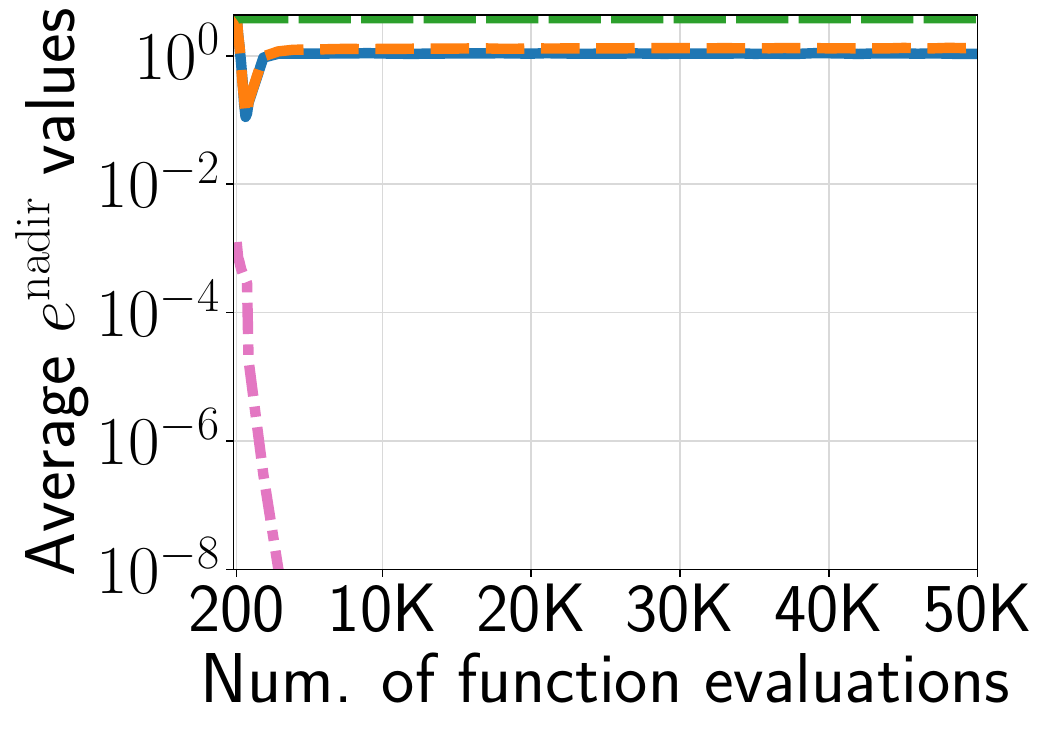}}
\\
   \subfloat[ORE ($m=2$)]{\includegraphics[width=0.32\textwidth]{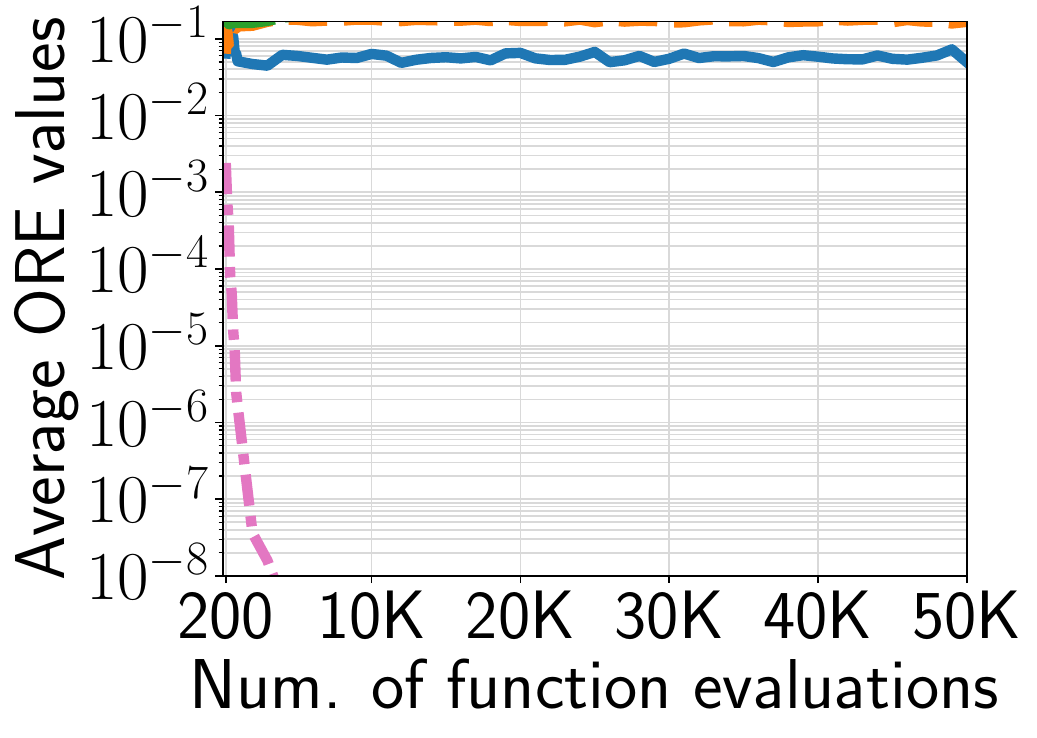}}
   \subfloat[ORE ($m=4$)]{\includegraphics[width=0.32\textwidth]{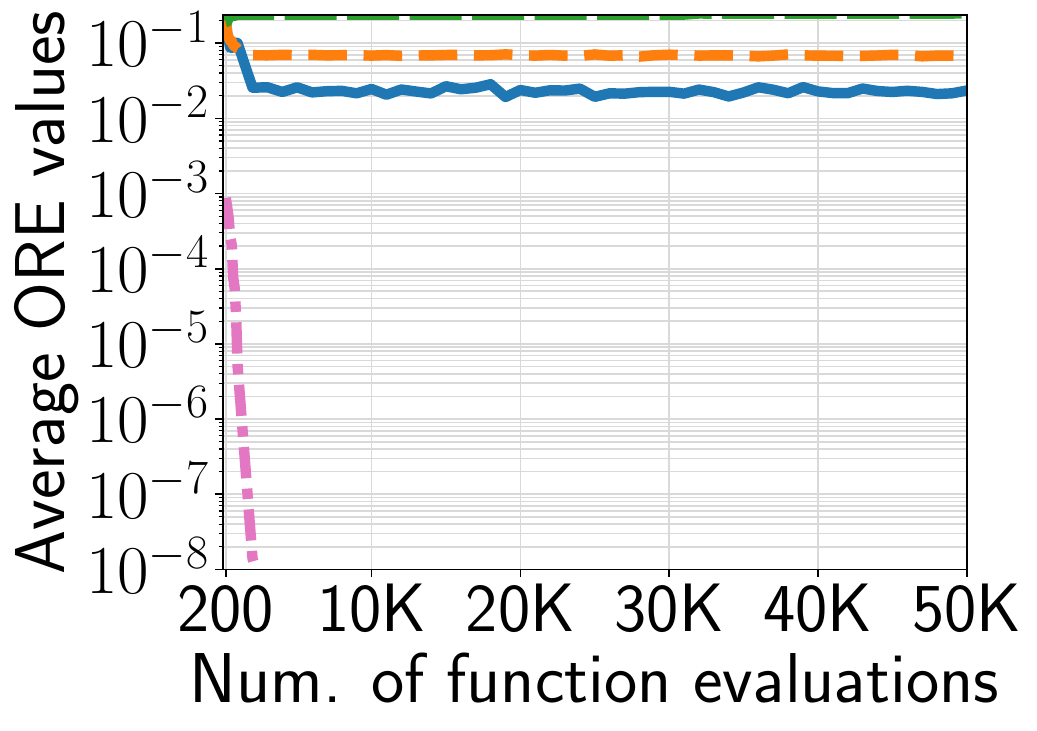}}
   \subfloat[ORE ($m=6$)]{\includegraphics[width=0.32\textwidth]{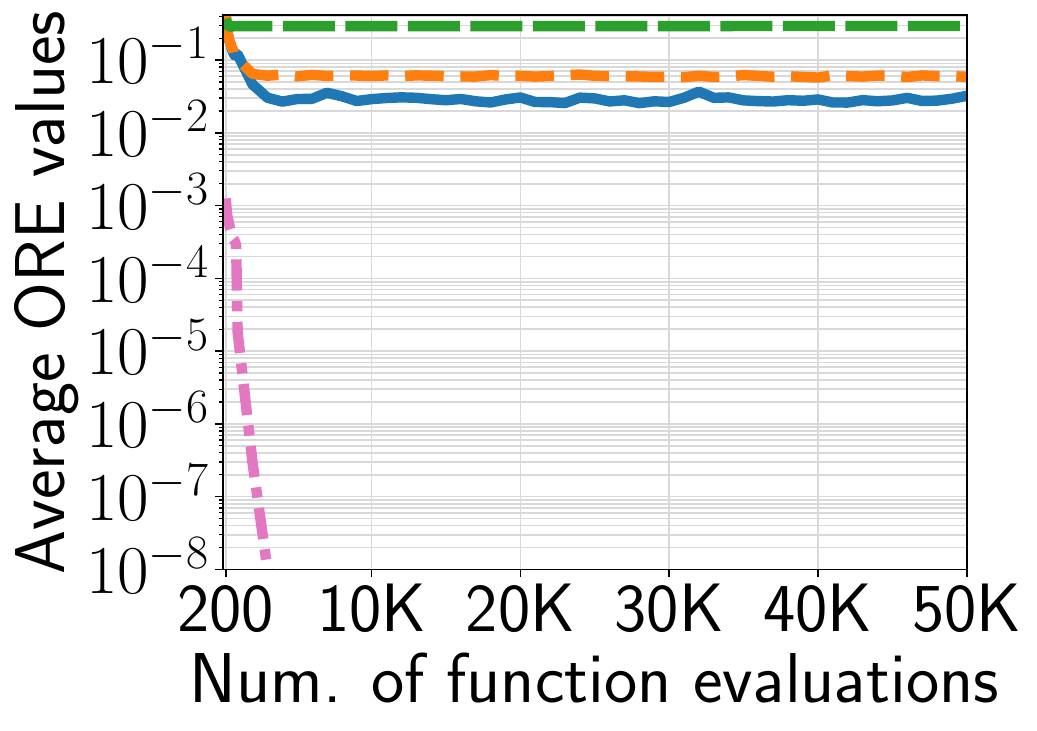}}
\\
\caption{Average $e^{\mathrm{ideal}}$, $e^{\mathrm{nadir}}$, and ORE values of the three normalization methods in r-NSGA-II on SDTLZ2.}
\label{supfig:3error_rNSGA2_SDTLZ2}
\end{figure*}

\begin{figure*}[t]
\centering
  \subfloat{\includegraphics[width=0.7\textwidth]{./figs/legend/legend_3.pdf}}
\vspace{-3.9mm}
   \\
   \subfloat[$e^{\mathrm{ideal}}$ ($m=2$)]{\includegraphics[width=0.32\textwidth]{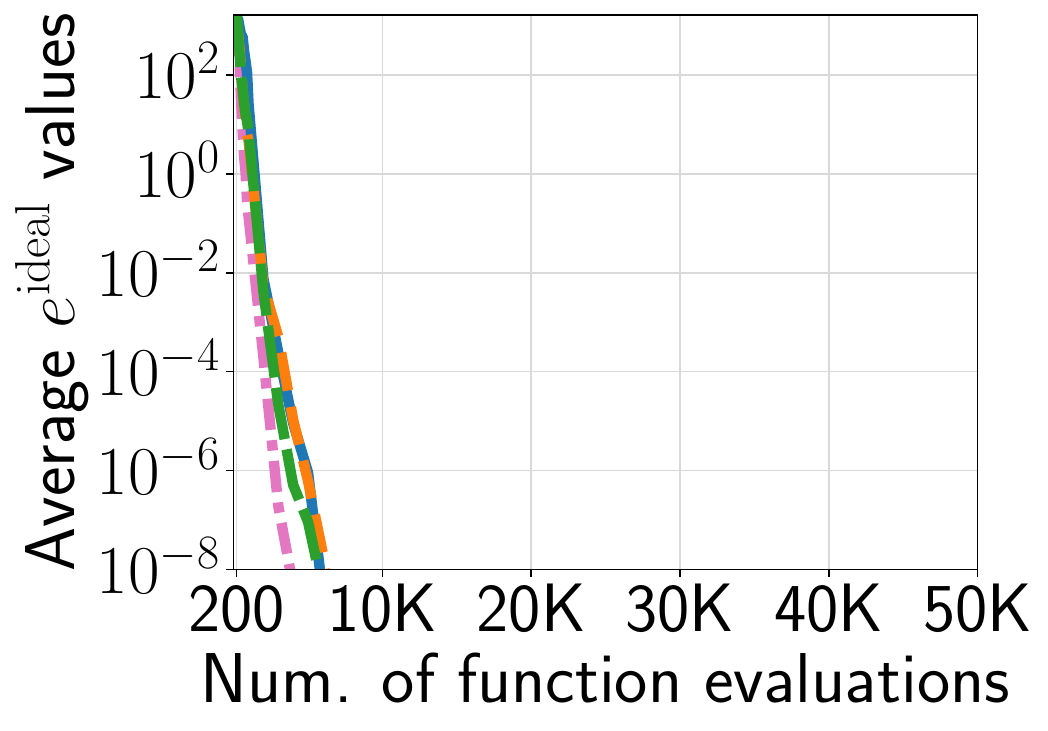}}
   \subfloat[$e^{\mathrm{ideal}}$ ($m=4$)]{\includegraphics[width=0.32\textwidth]{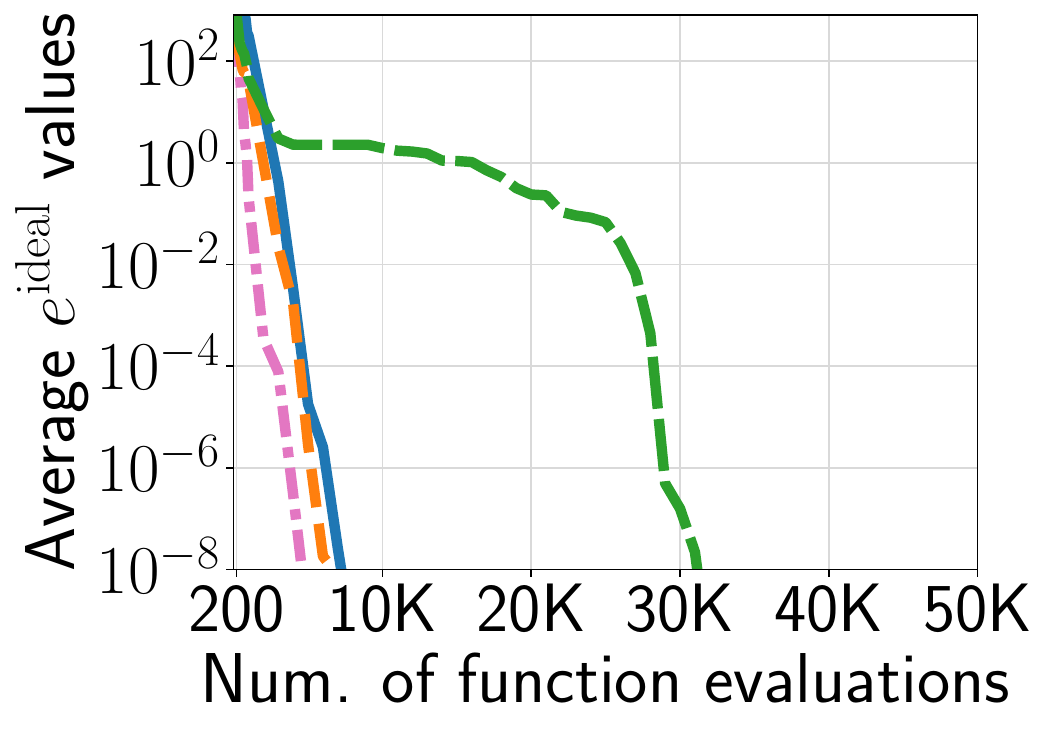}}
   \subfloat[$e^{\mathrm{ideal}}$ ($m=6$)]{\includegraphics[width=0.32\textwidth]{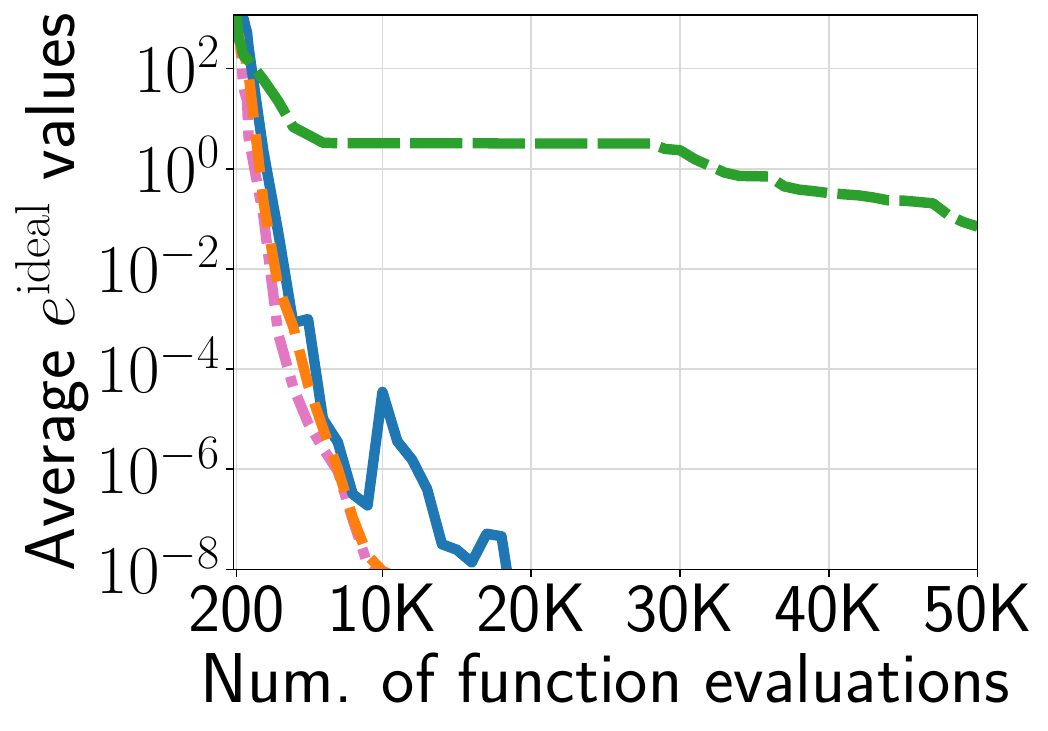}}
\\
   \subfloat[$e^{\mathrm{nadir}}$ ($m=2$)]{\includegraphics[width=0.32\textwidth]{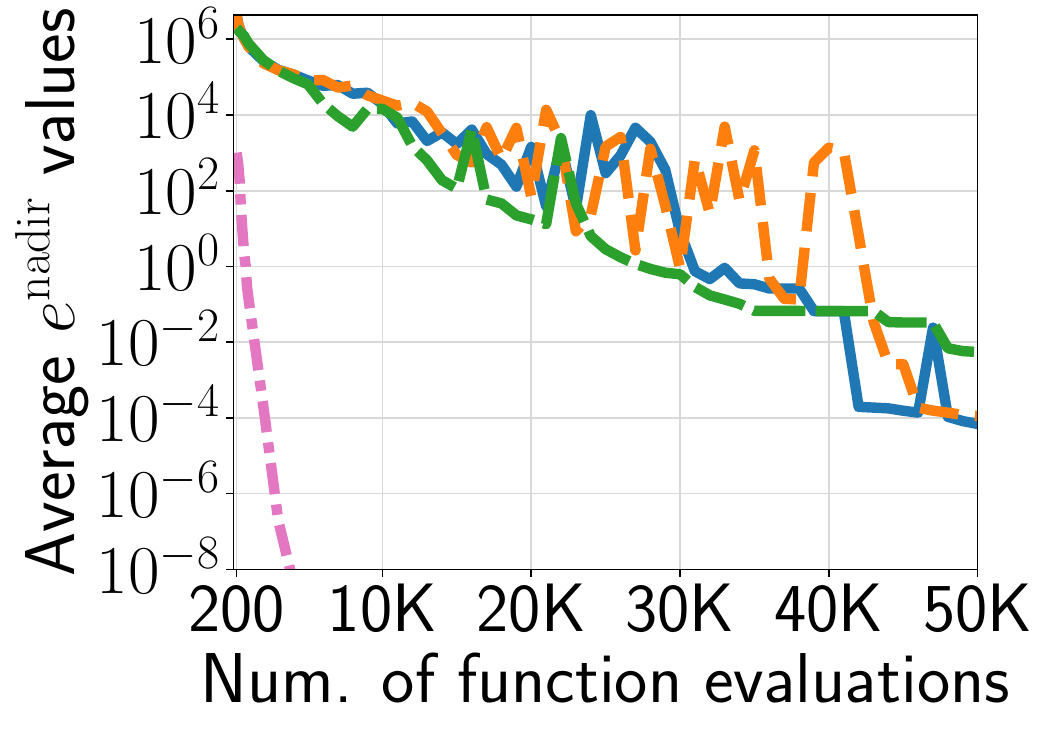}}
   \subfloat[$e^{\mathrm{nadir}}$ ($m=4$)]{\includegraphics[width=0.32\textwidth]{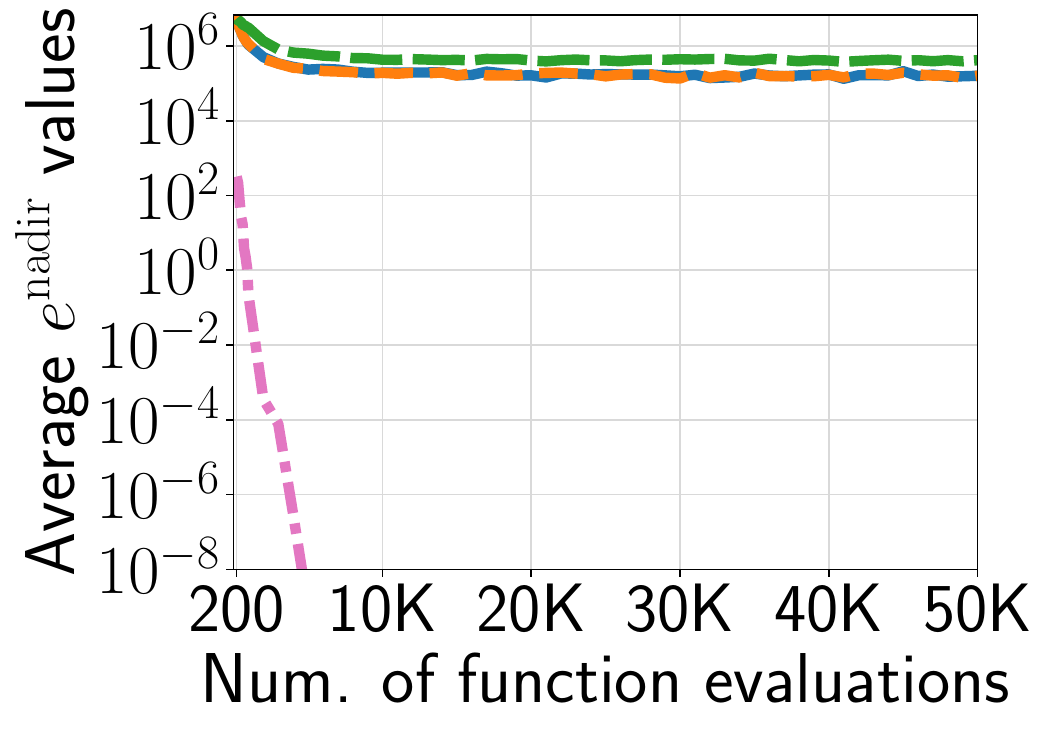}}
   \subfloat[$e^{\mathrm{nadir}}$ ($m=6$)]{\includegraphics[width=0.32\textwidth]{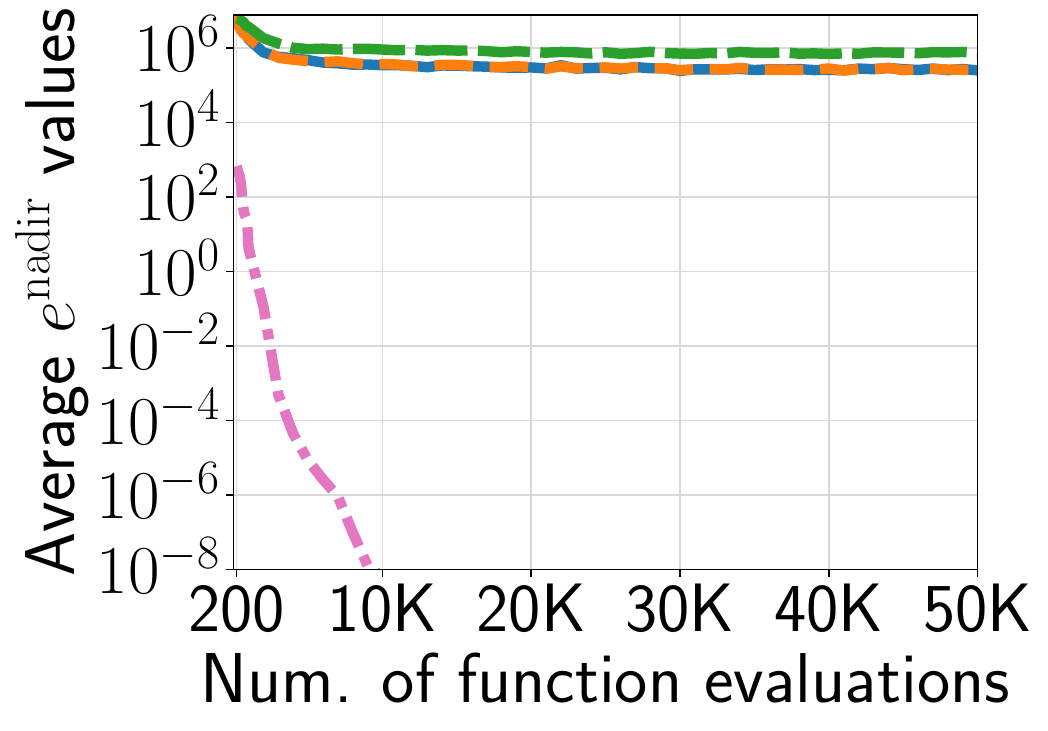}}
\\
   \subfloat[ORE ($m=2$)]{\includegraphics[width=0.32\textwidth]{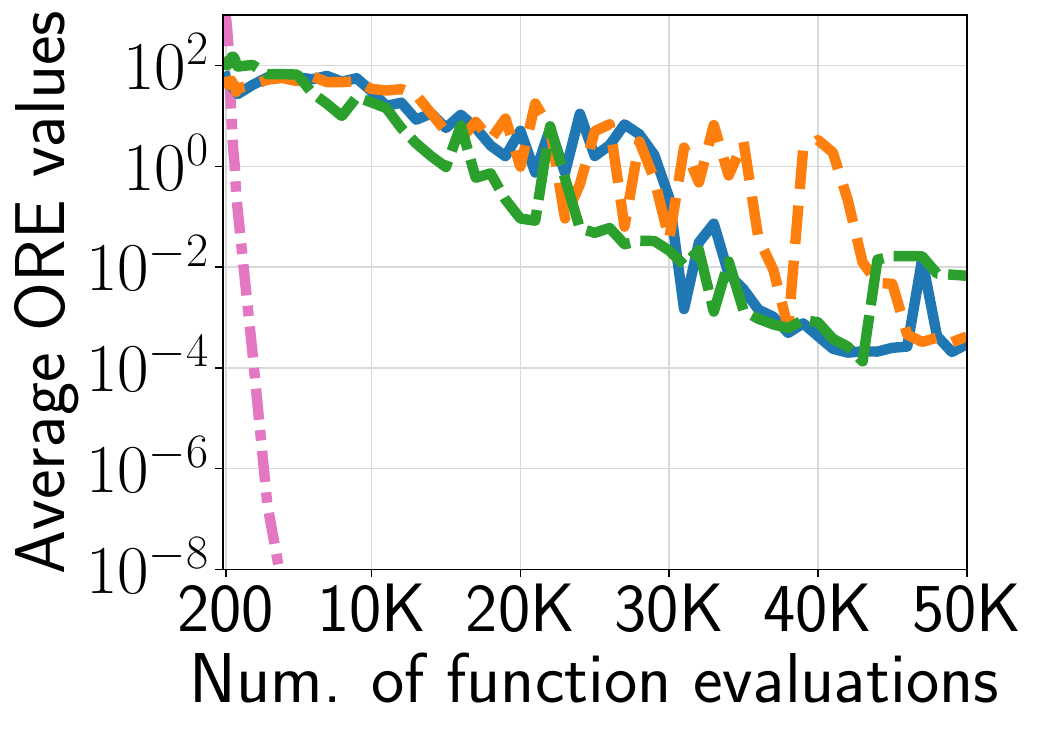}}
   \subfloat[ORE ($m=4$)]{\includegraphics[width=0.32\textwidth]{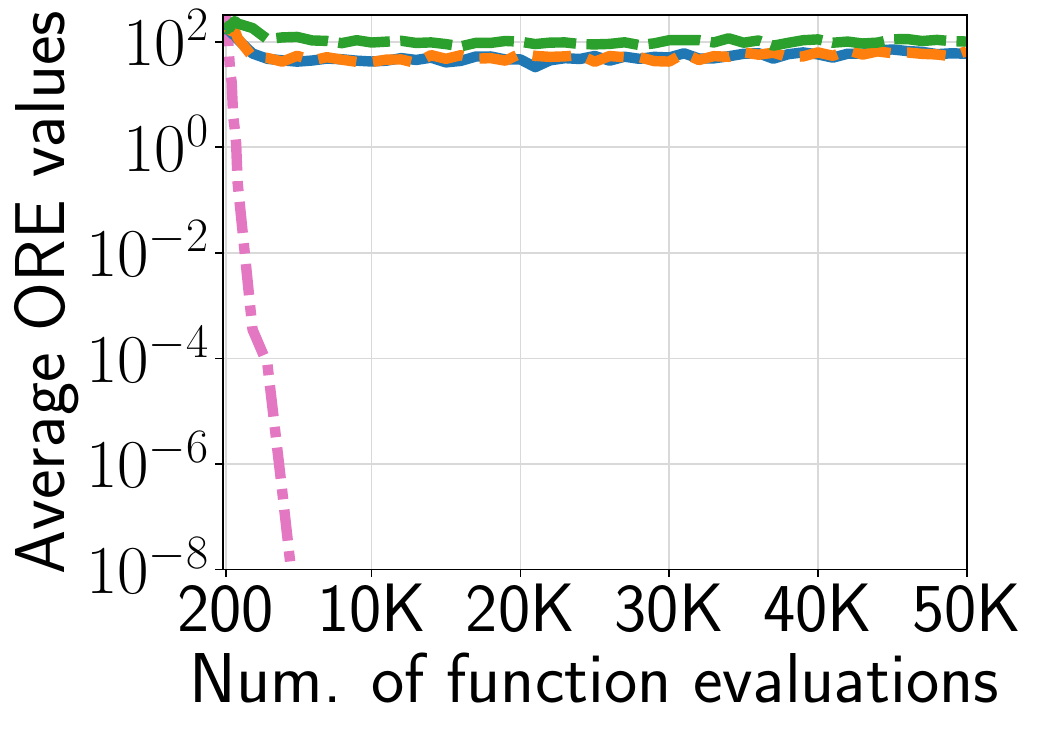}}
   \subfloat[ORE ($m=6$)]{\includegraphics[width=0.32\textwidth]{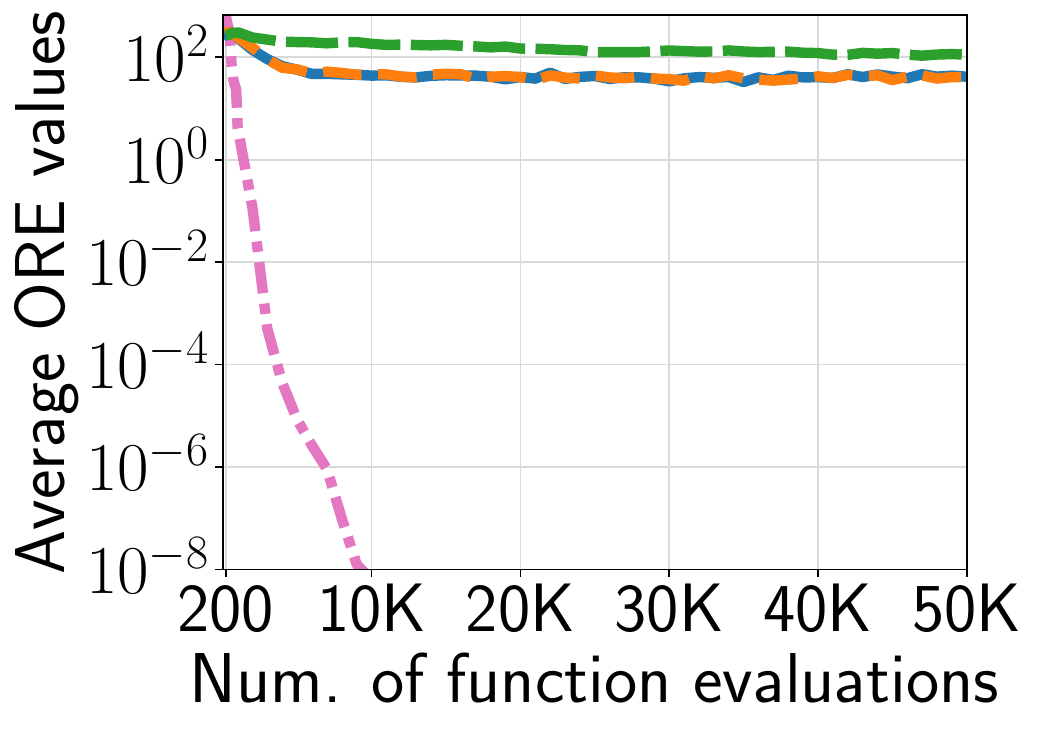}}
\\
\caption{Average $e^{\mathrm{ideal}}$, $e^{\mathrm{nadir}}$, and ORE values of the three normalization methods in r-NSGA-II on SDTLZ3.}
\label{supfig:3error_rNSGA2_SDTLZ3}
\end{figure*}

\begin{figure*}[t]
\centering
  \subfloat{\includegraphics[width=0.7\textwidth]{./figs/legend/legend_3.pdf}}
\vspace{-3.9mm}
   \\
   \subfloat[$e^{\mathrm{ideal}}$ ($m=2$)]{\includegraphics[width=0.32\textwidth]{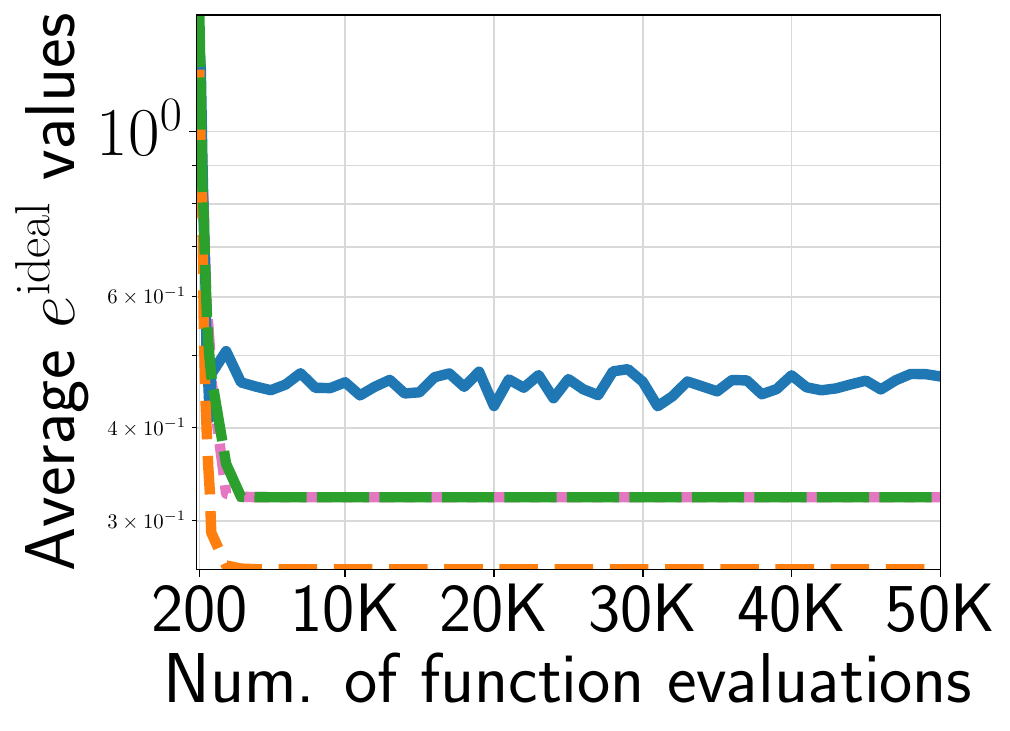}}
   \subfloat[$e^{\mathrm{ideal}}$ ($m=4$)]{\includegraphics[width=0.32\textwidth]{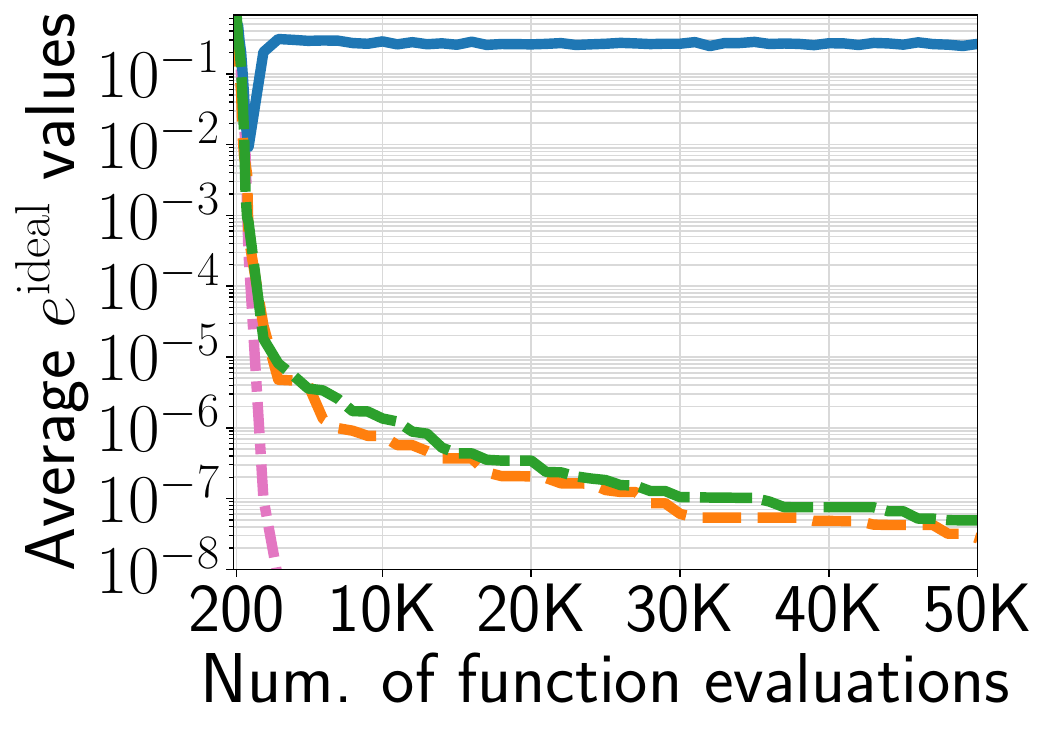}}
   \subfloat[$e^{\mathrm{ideal}}$ ($m=6$)]{\includegraphics[width=0.32\textwidth]{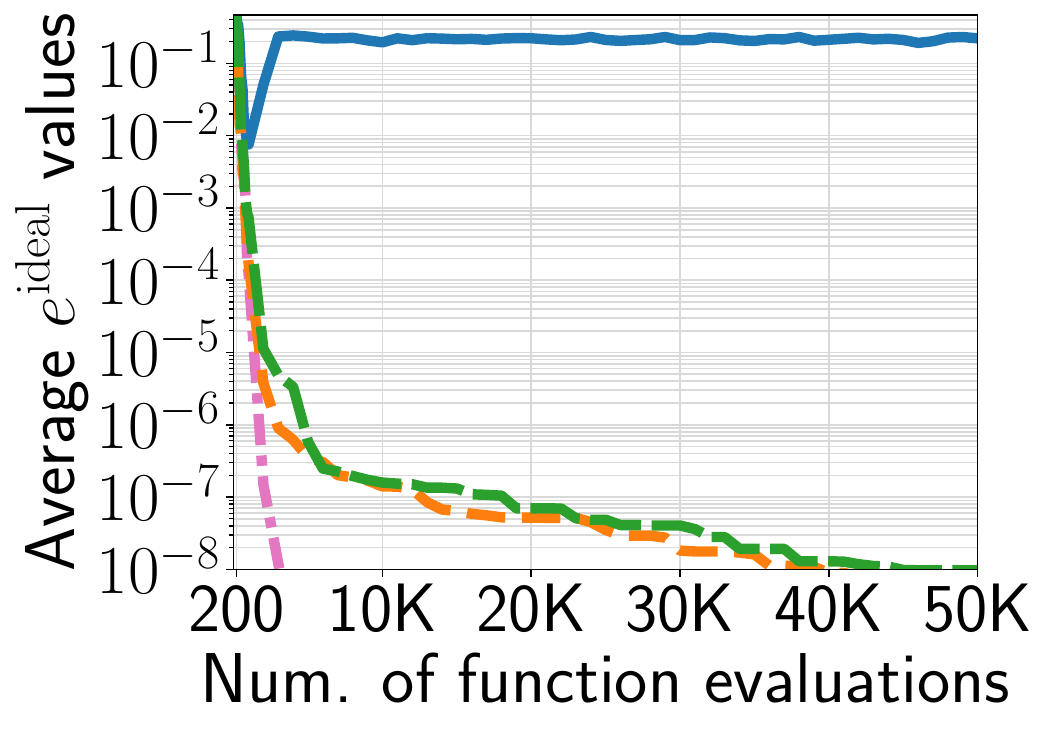}}
\\
   \subfloat[$e^{\mathrm{nadir}}$ ($m=2$)]{\includegraphics[width=0.32\textwidth]{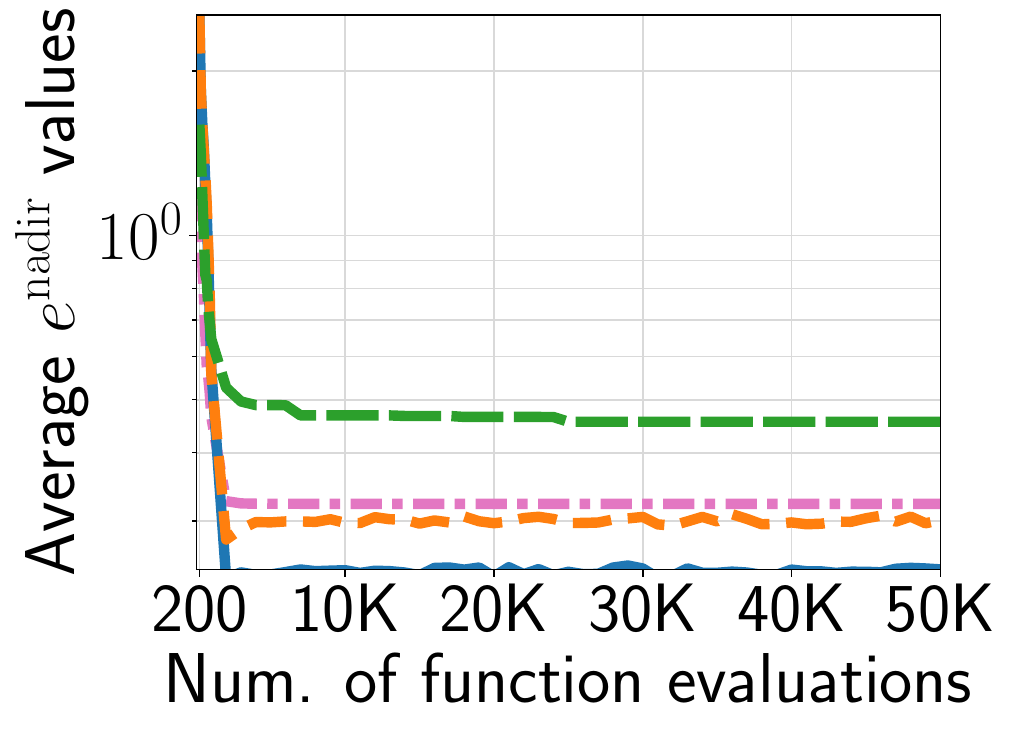}}
   \subfloat[$e^{\mathrm{nadir}}$ ($m=4$)]{\includegraphics[width=0.32\textwidth]{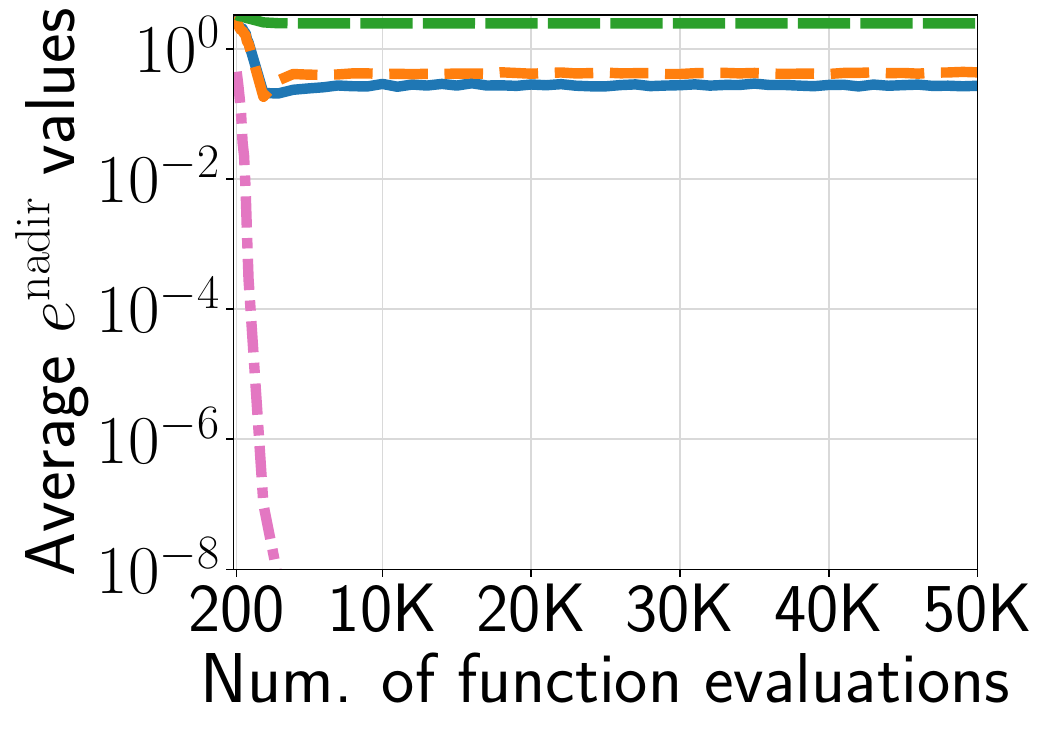}}
   \subfloat[$e^{\mathrm{nadir}}$ ($m=6$)]{\includegraphics[width=0.32\textwidth]{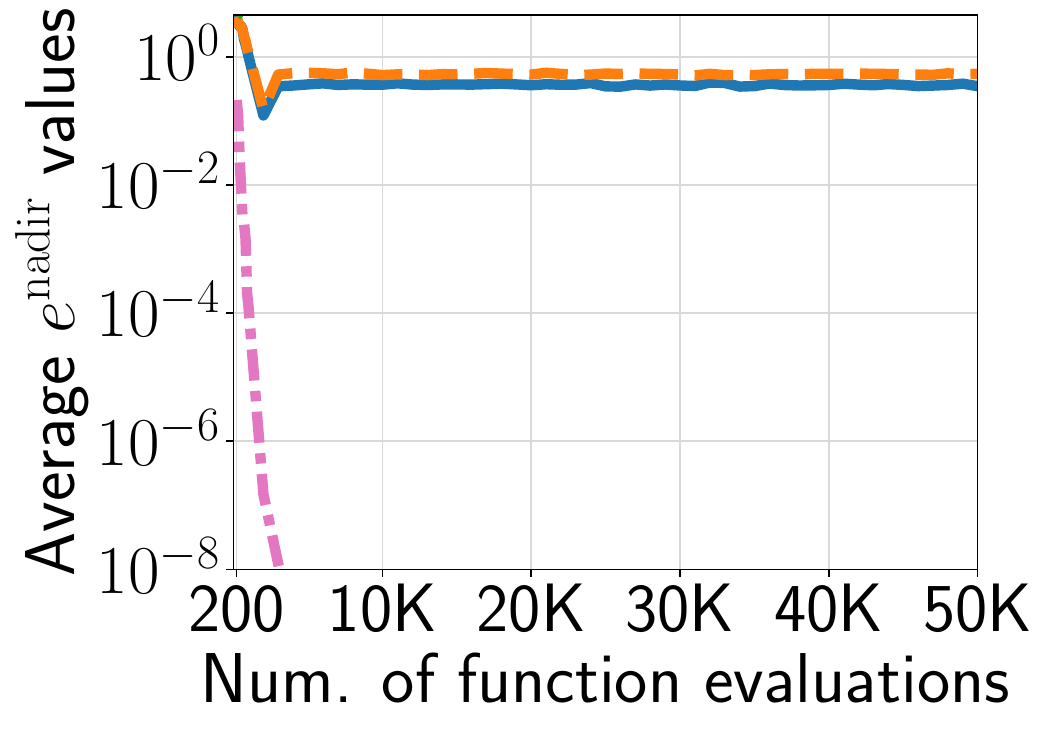}}
\\
   \subfloat[ORE ($m=2$)]{\includegraphics[width=0.32\textwidth]{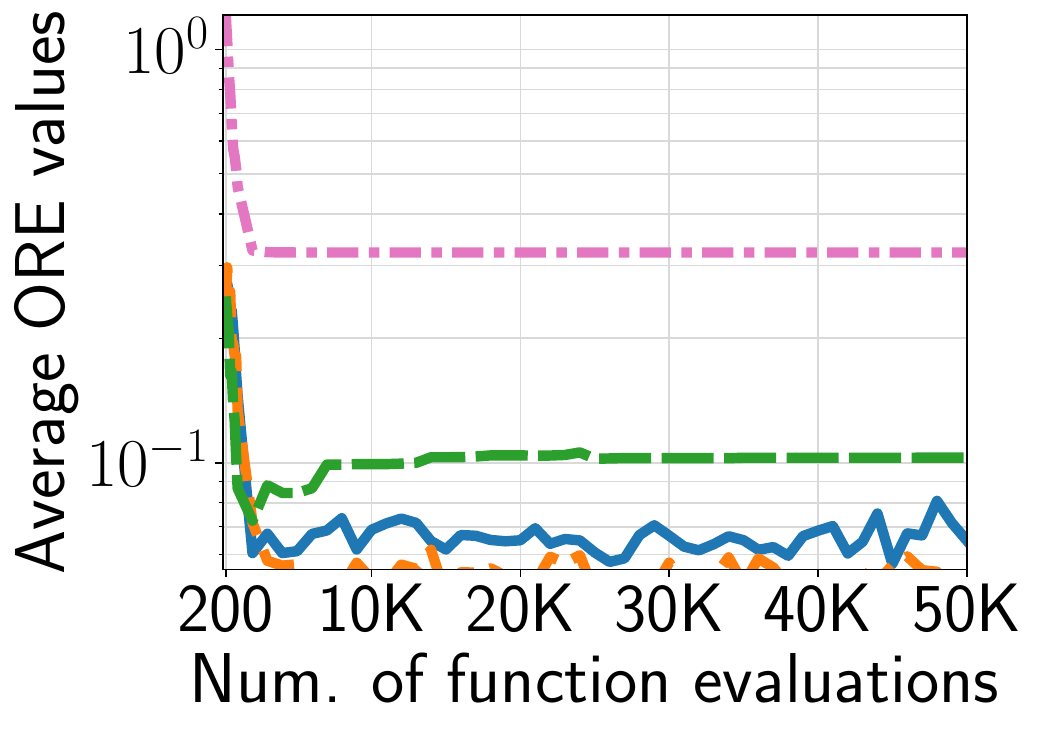}}
   \subfloat[ORE ($m=4$)]{\includegraphics[width=0.32\textwidth]{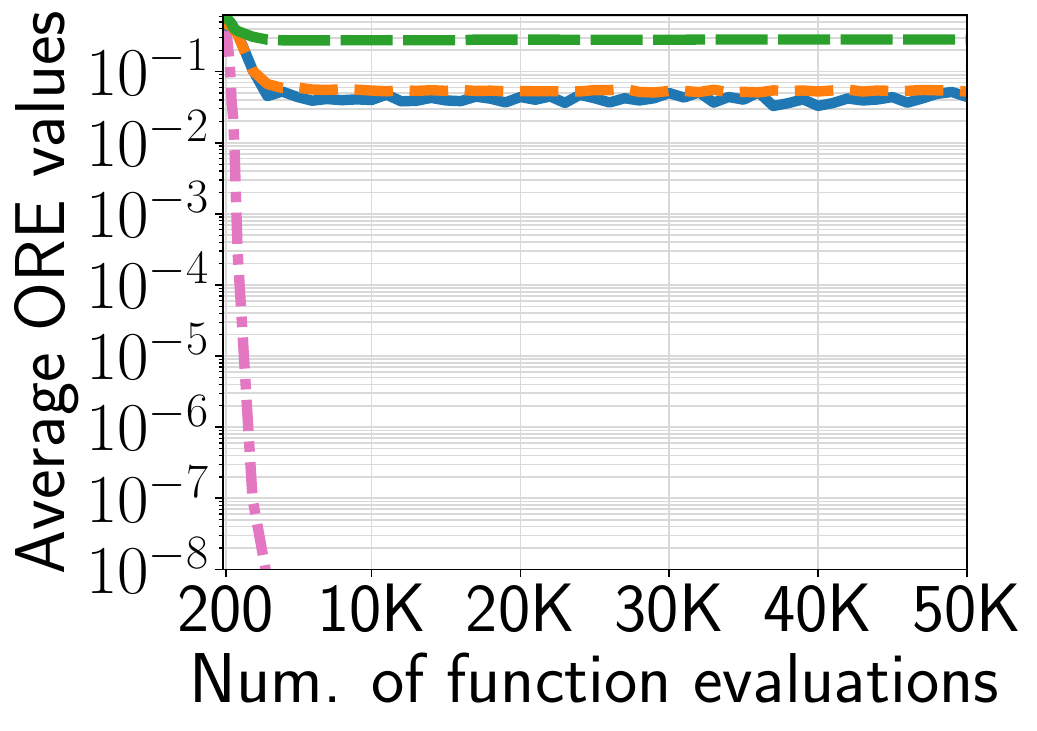}}
   \subfloat[ORE ($m=6$)]{\includegraphics[width=0.32\textwidth]{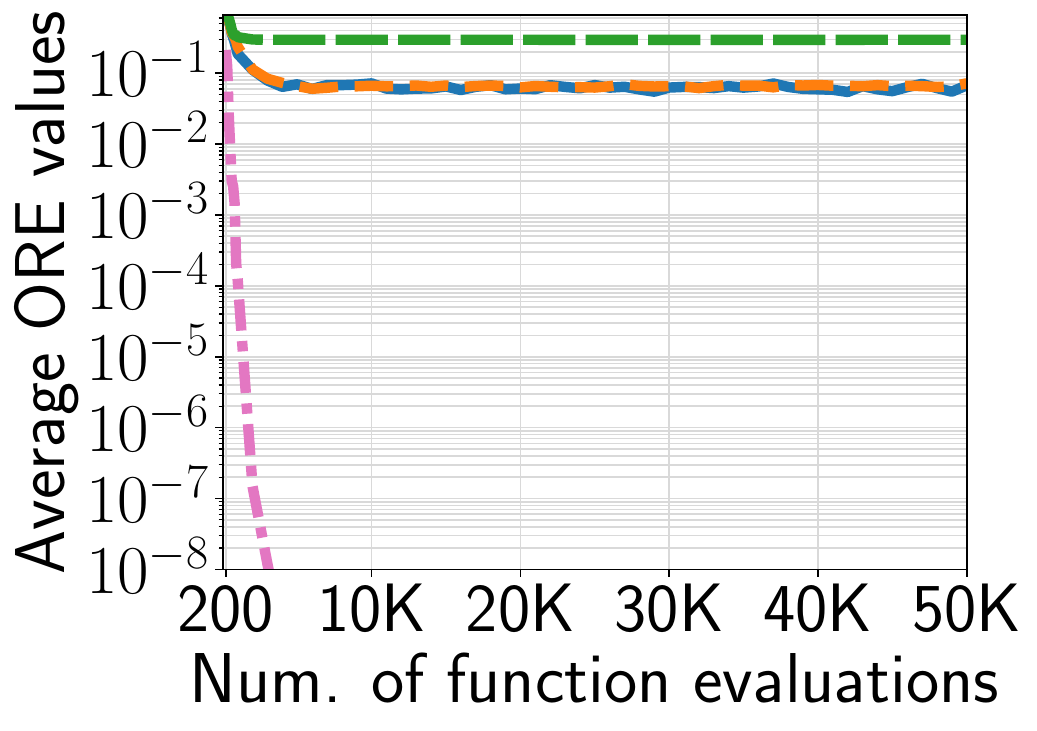}}
\\
\caption{Average $e^{\mathrm{ideal}}$, $e^{\mathrm{nadir}}$, and ORE values of the three normalization methods in r-NSGA-II on SDTLZ4.}
\label{supfig:3error_rNSGA2_SDTLZ4}
\end{figure*}

\begin{figure*}[t]
\centering
  \subfloat{\includegraphics[width=0.7\textwidth]{./figs/legend/legend_3.pdf}}
\vspace{-3.9mm}
   \\
   \subfloat[$e^{\mathrm{ideal}}$ ($m=2$)]{\includegraphics[width=0.32\textwidth]{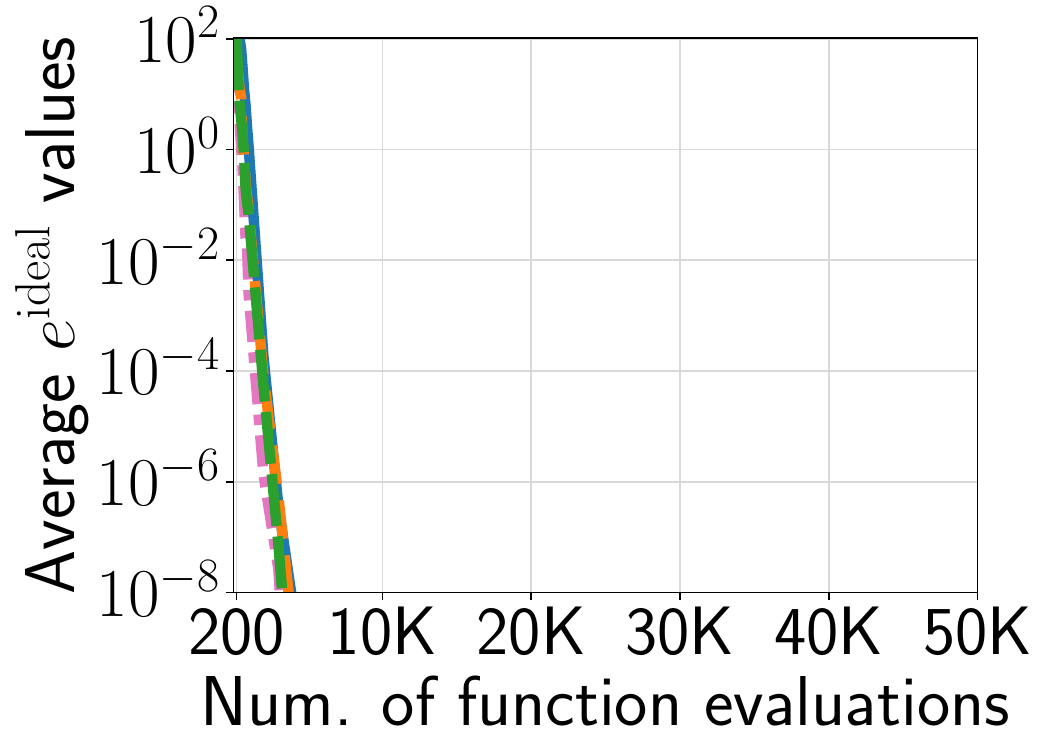}}
   \subfloat[$e^{\mathrm{ideal}}$ ($m=4$)]{\includegraphics[width=0.32\textwidth]{./figs/qi_error_ideal/rNSGA2_mu100/IDTLZ1_m4_r0.1_z-type1.pdf}}
   \subfloat[$e^{\mathrm{ideal}}$ ($m=6$)]{\includegraphics[width=0.32\textwidth]{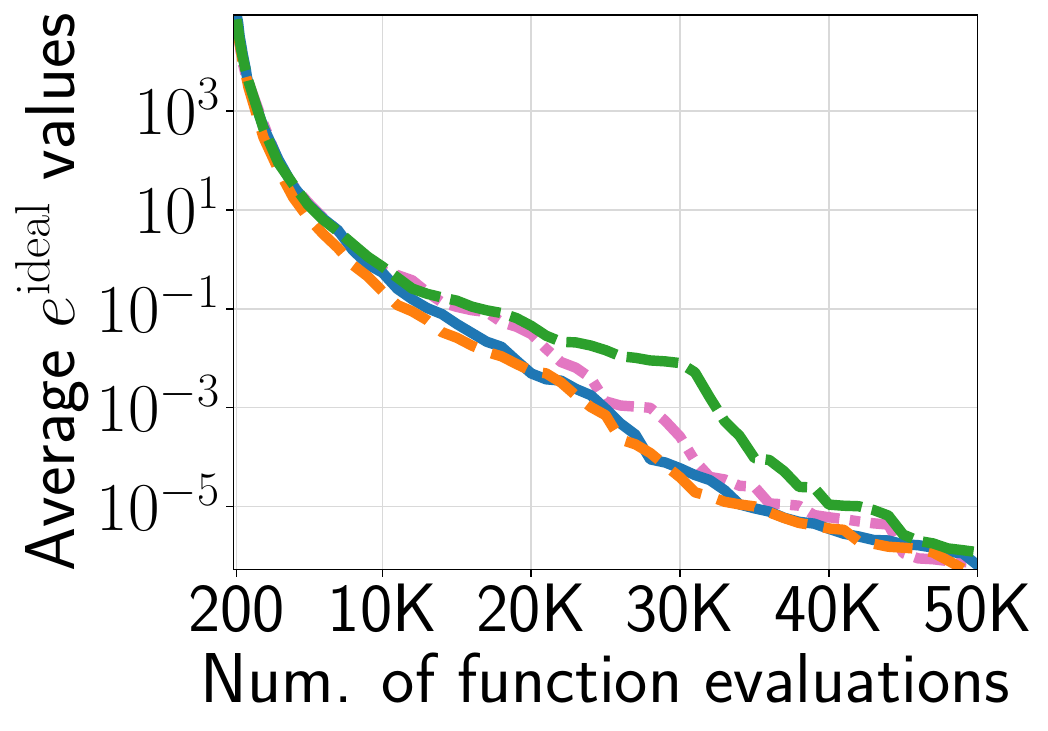}}
\\
   \subfloat[$e^{\mathrm{nadir}}$ ($m=2$)]{\includegraphics[width=0.32\textwidth]{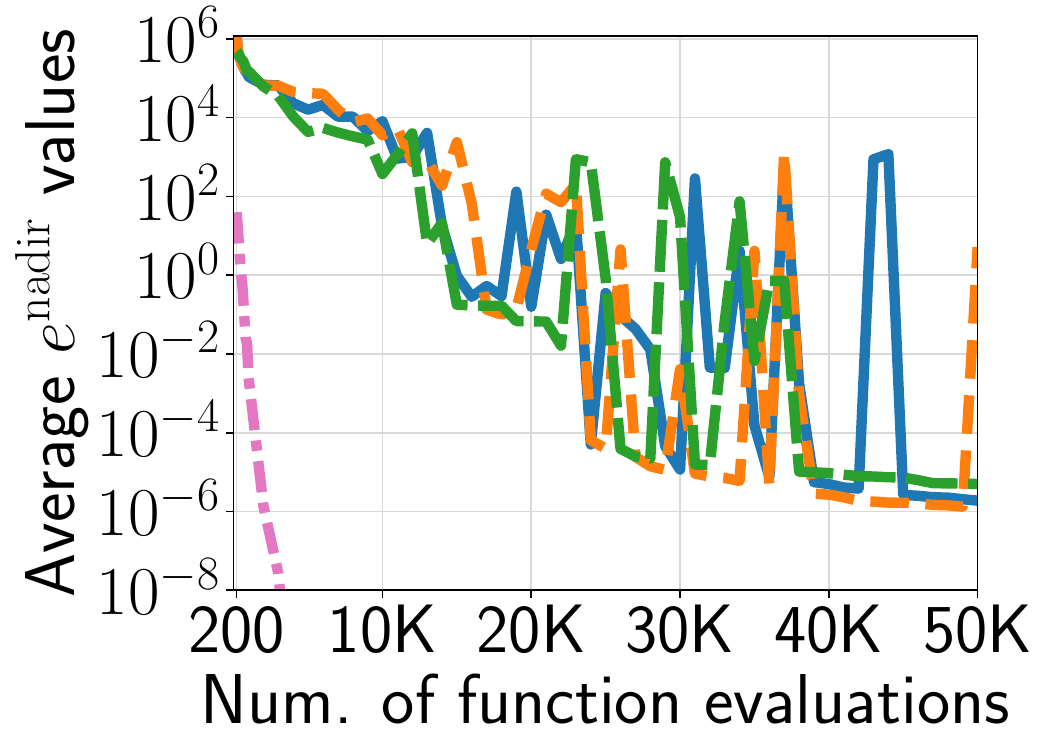}}
   \subfloat[$e^{\mathrm{nadir}}$ ($m=4$)]{\includegraphics[width=0.32\textwidth]{./figs/qi_error_nadir/rNSGA2_mu100/IDTLZ1_m4_r0.1_z-type1.pdf}}
   \subfloat[$e^{\mathrm{nadir}}$ ($m=6$)]{\includegraphics[width=0.32\textwidth]{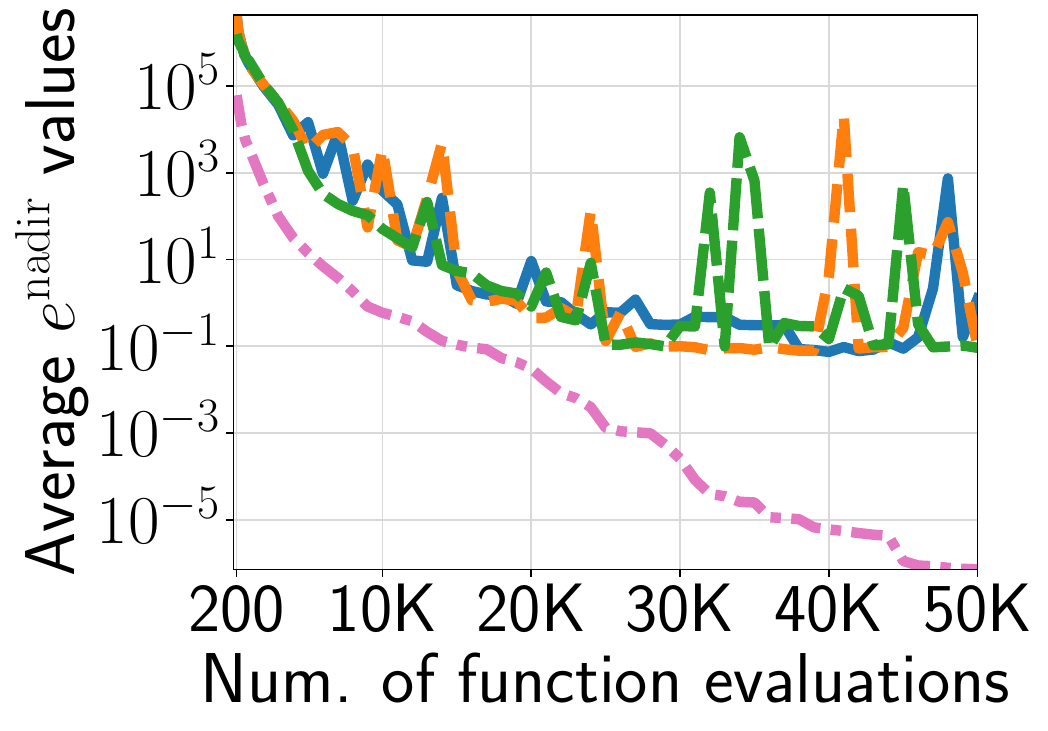}}
\\
   \subfloat[ORE ($m=2$)]{\includegraphics[width=0.32\textwidth]{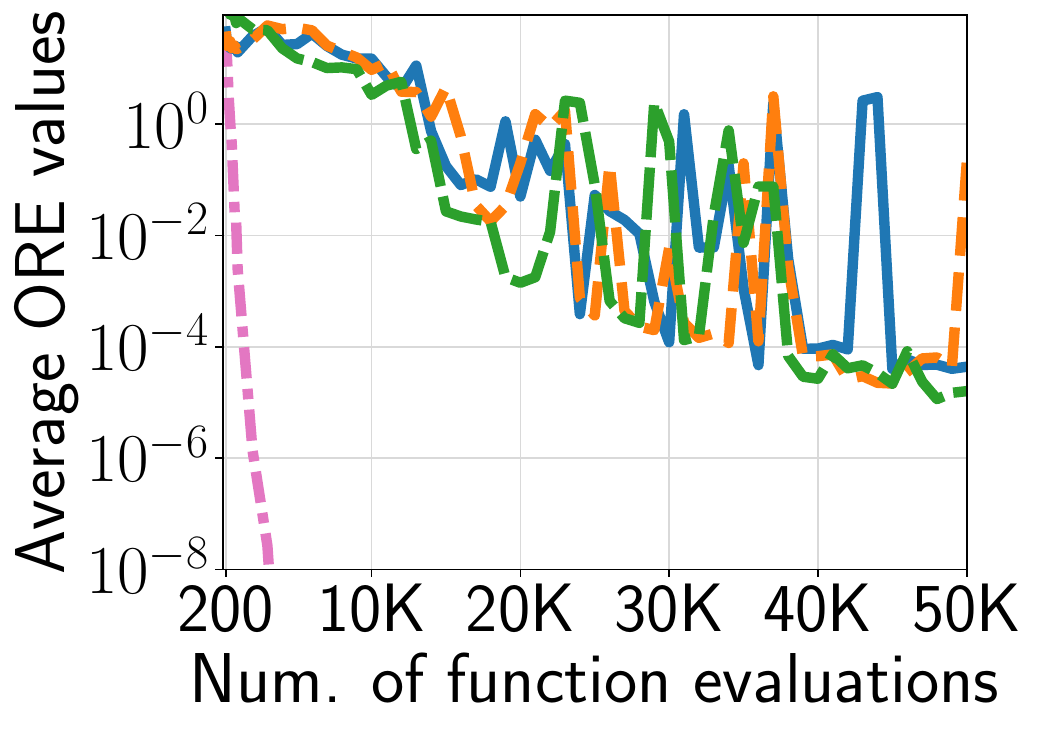}}
   \subfloat[ORE ($m=4$)]{\includegraphics[width=0.32\textwidth]{./figs/qi_ore/rNSGA2_mu100/IDTLZ1_m4_r0.1_z-type1.pdf}}
   \subfloat[ORE ($m=6$)]{\includegraphics[width=0.32\textwidth]{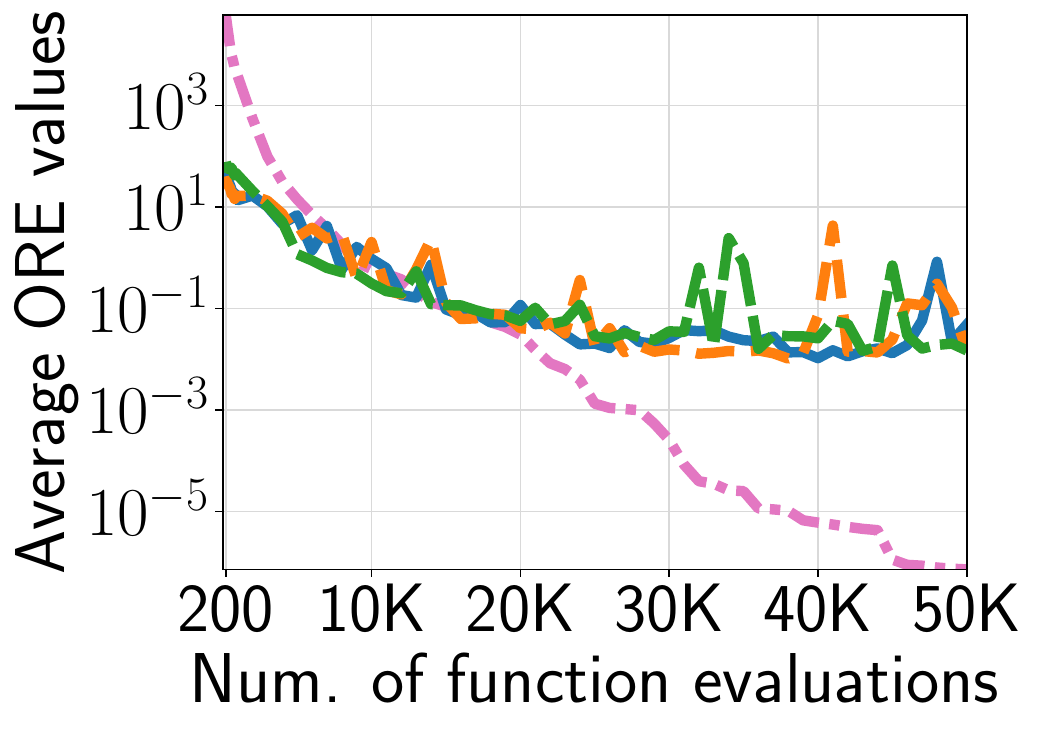}}
\\
\caption{Average $e^{\mathrm{ideal}}$, $e^{\mathrm{nadir}}$, and ORE values of the three normalization methods in r-NSGA-II on IDTLZ1.}
\label{supfig:3error_rNSGA2_IDTLZ1}
\end{figure*}

\begin{figure*}[t]
\centering
  \subfloat{\includegraphics[width=0.7\textwidth]{./figs/legend/legend_3.pdf}}
\vspace{-3.9mm}
   \\
   \subfloat[$e^{\mathrm{ideal}}$ ($m=2$)]{\includegraphics[width=0.32\textwidth]{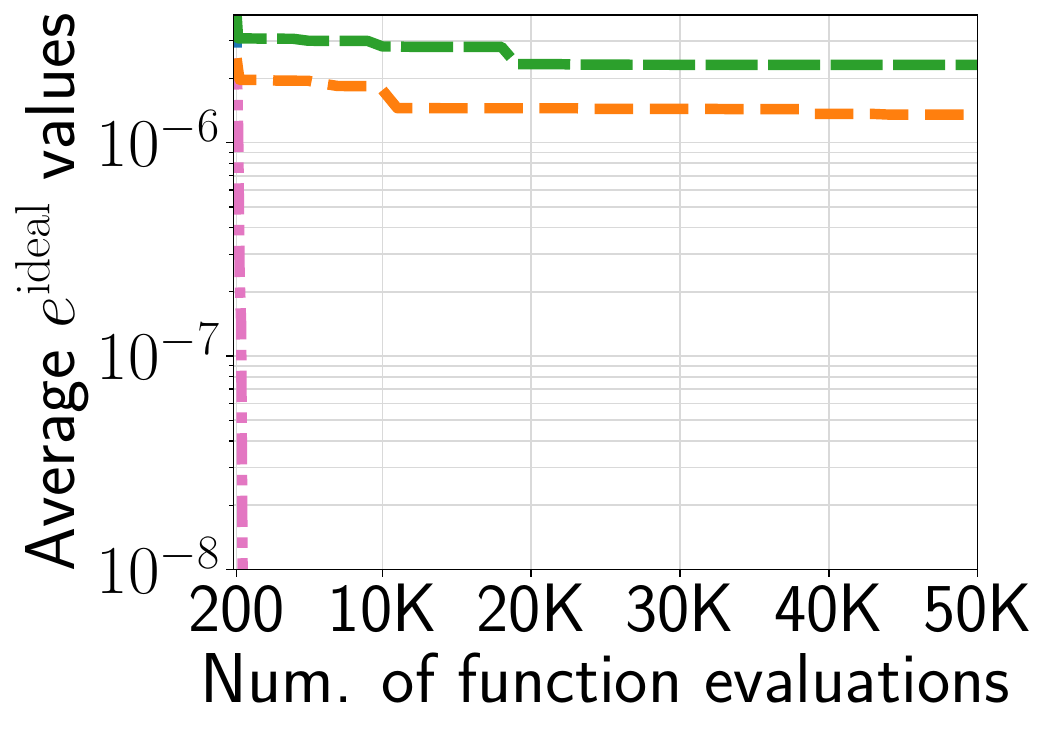}}
   \subfloat[$e^{\mathrm{ideal}}$ ($m=4$)]{\includegraphics[width=0.32\textwidth]{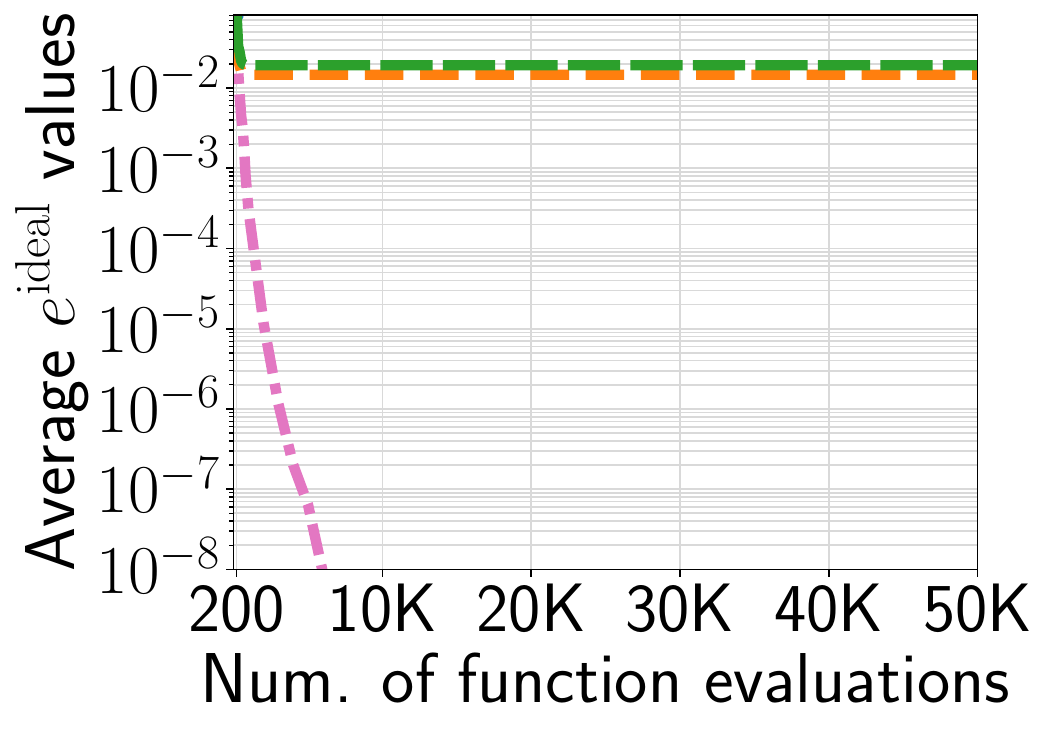}}
   \subfloat[$e^{\mathrm{ideal}}$ ($m=6$)]{\includegraphics[width=0.32\textwidth]{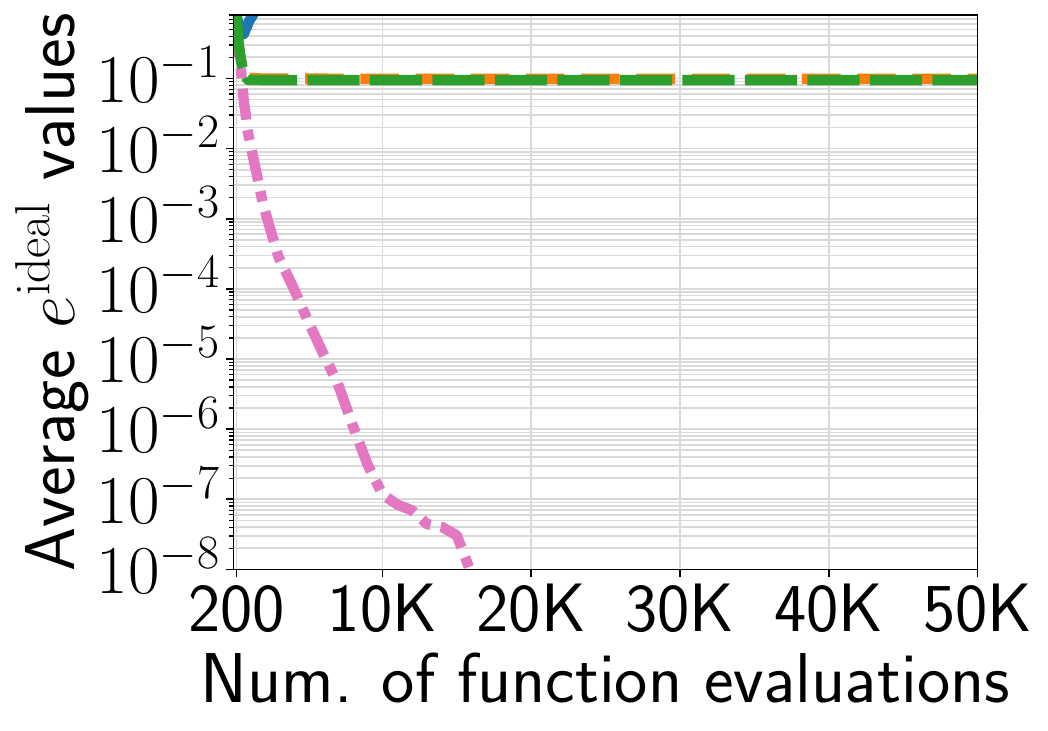}}
\\
   \subfloat[$e^{\mathrm{nadir}}$ ($m=2$)]{\includegraphics[width=0.32\textwidth]{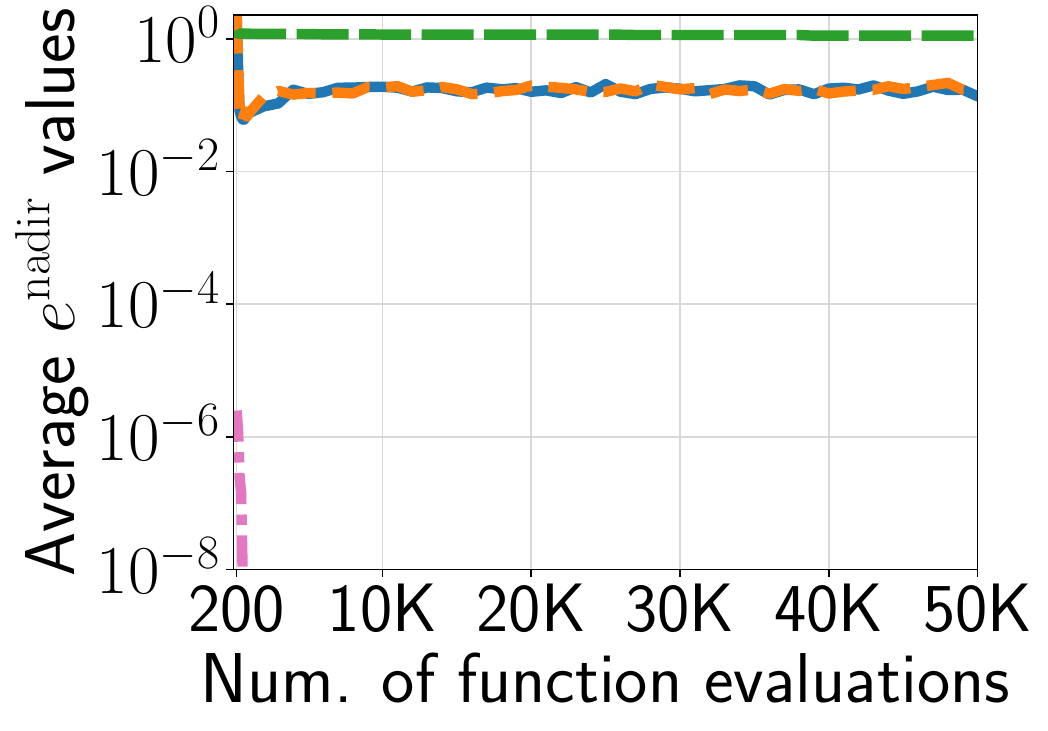}}
   \subfloat[$e^{\mathrm{nadir}}$ ($m=4$)]{\includegraphics[width=0.32\textwidth]{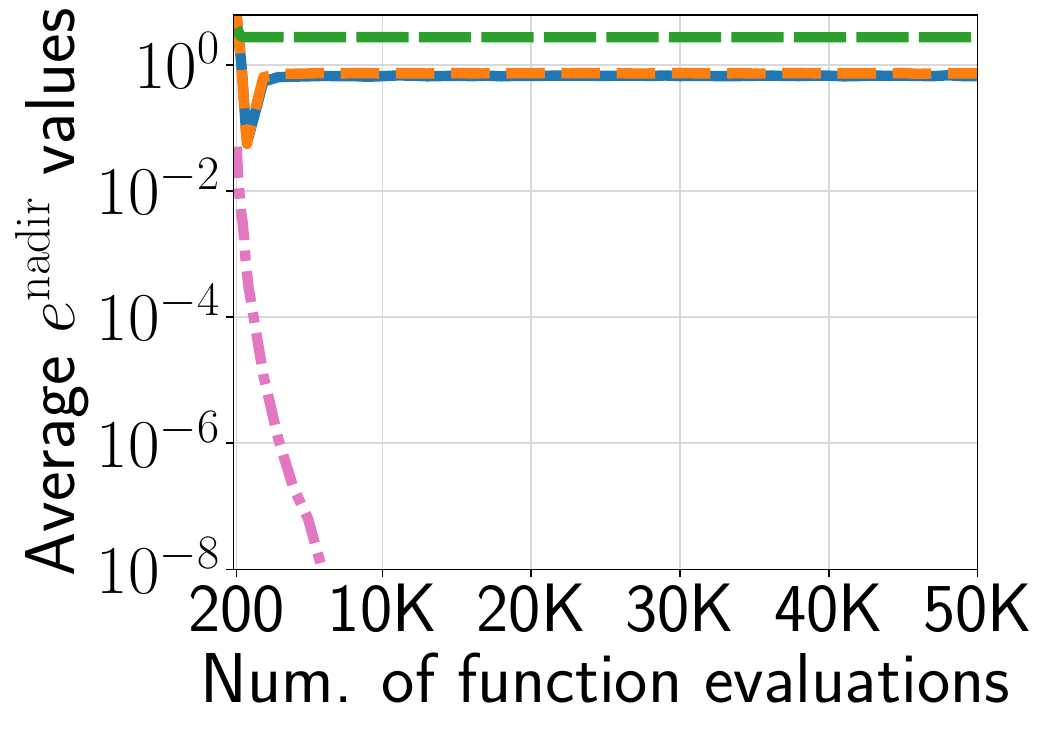}}
   \subfloat[$e^{\mathrm{nadir}}$ ($m=6$)]{\includegraphics[width=0.32\textwidth]{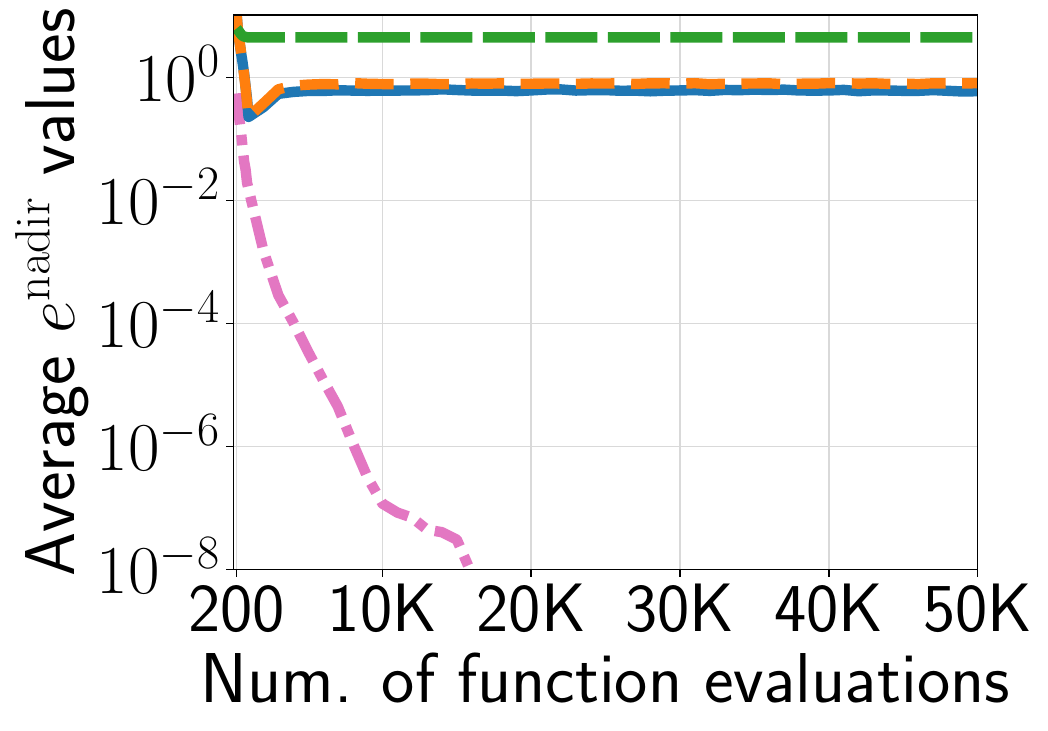}}
\\
   \subfloat[ORE ($m=2$)]{\includegraphics[width=0.32\textwidth]{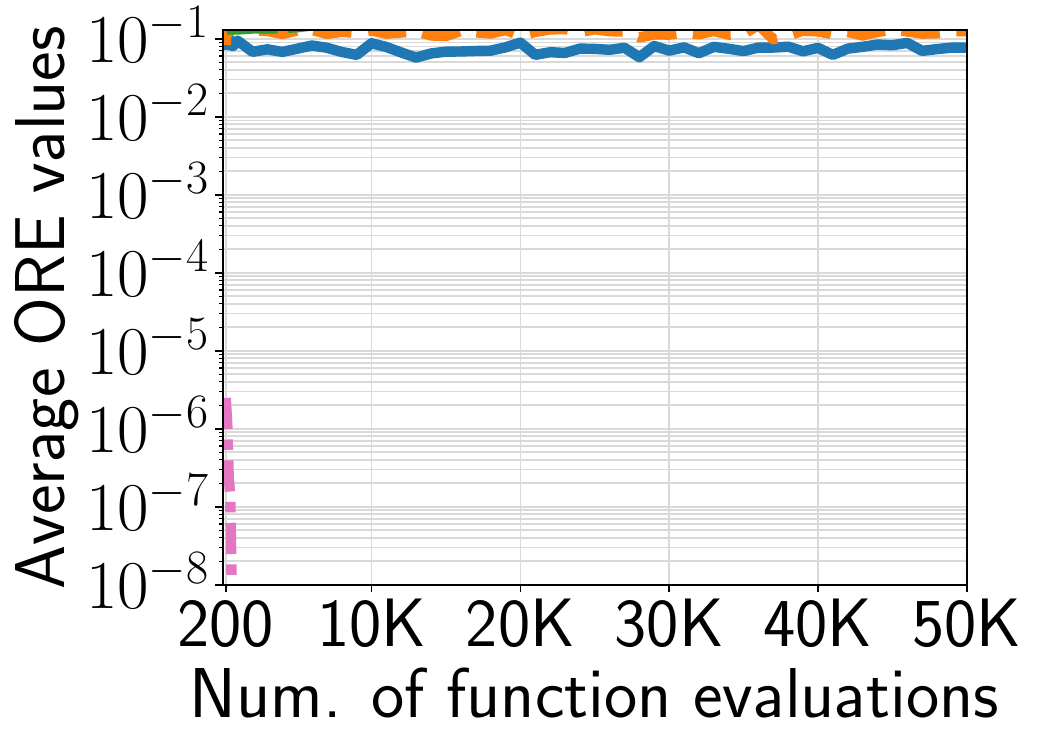}}
   \subfloat[ORE ($m=4$)]{\includegraphics[width=0.32\textwidth]{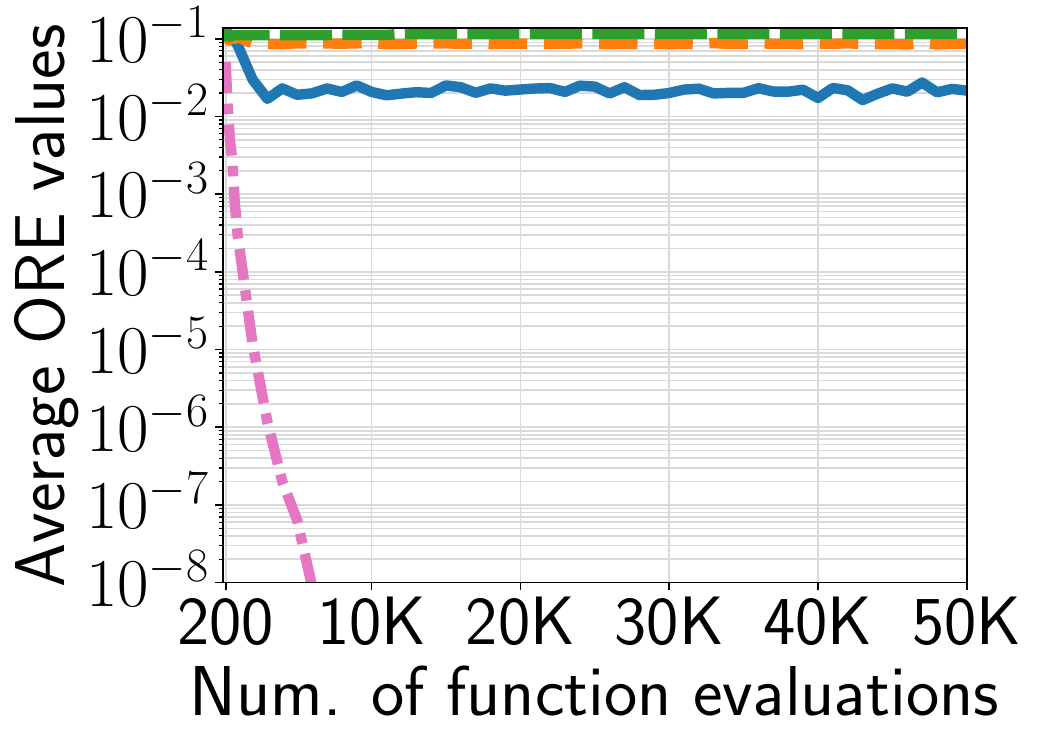}}
   \subfloat[ORE ($m=6$)]{\includegraphics[width=0.32\textwidth]{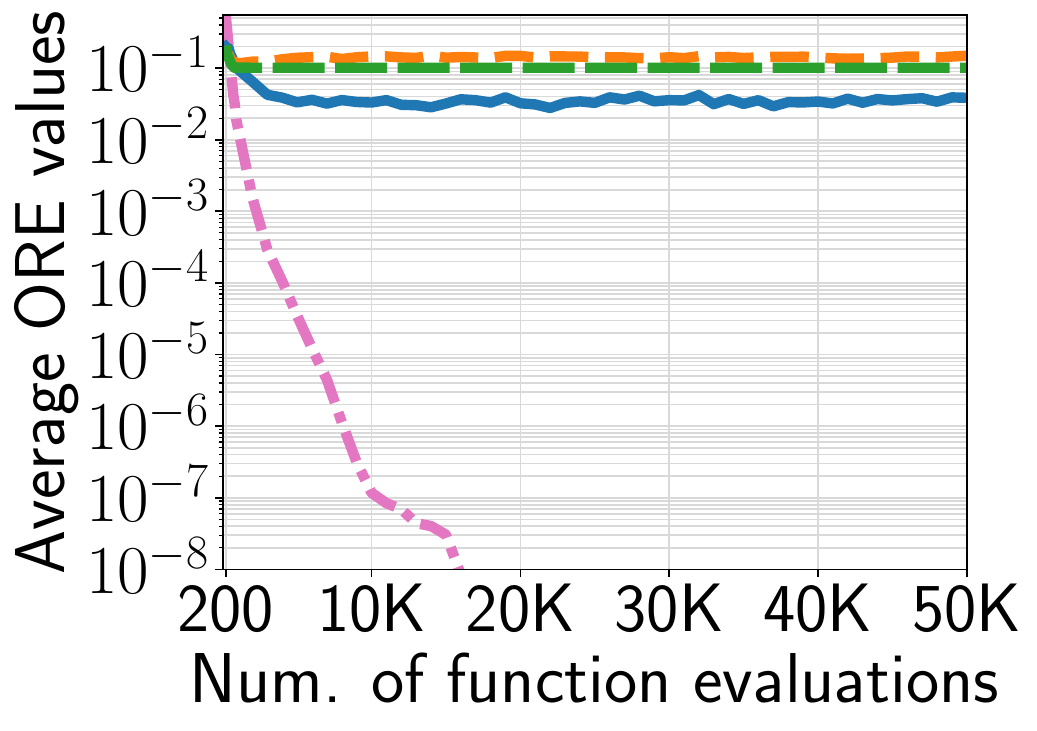}}
\\
\caption{Average $e^{\mathrm{ideal}}$, $e^{\mathrm{nadir}}$, and ORE values of the three normalization methods in r-NSGA-II on IDTLZ2.}
\label{supfig:3error_rNSGA2_IDTLZ2}
\end{figure*}

\begin{figure*}[t]
\centering
  \subfloat{\includegraphics[width=0.7\textwidth]{./figs/legend/legend_3.pdf}}
\vspace{-3.9mm}
   \\
   \subfloat[$e^{\mathrm{ideal}}$ ($m=2$)]{\includegraphics[width=0.32\textwidth]{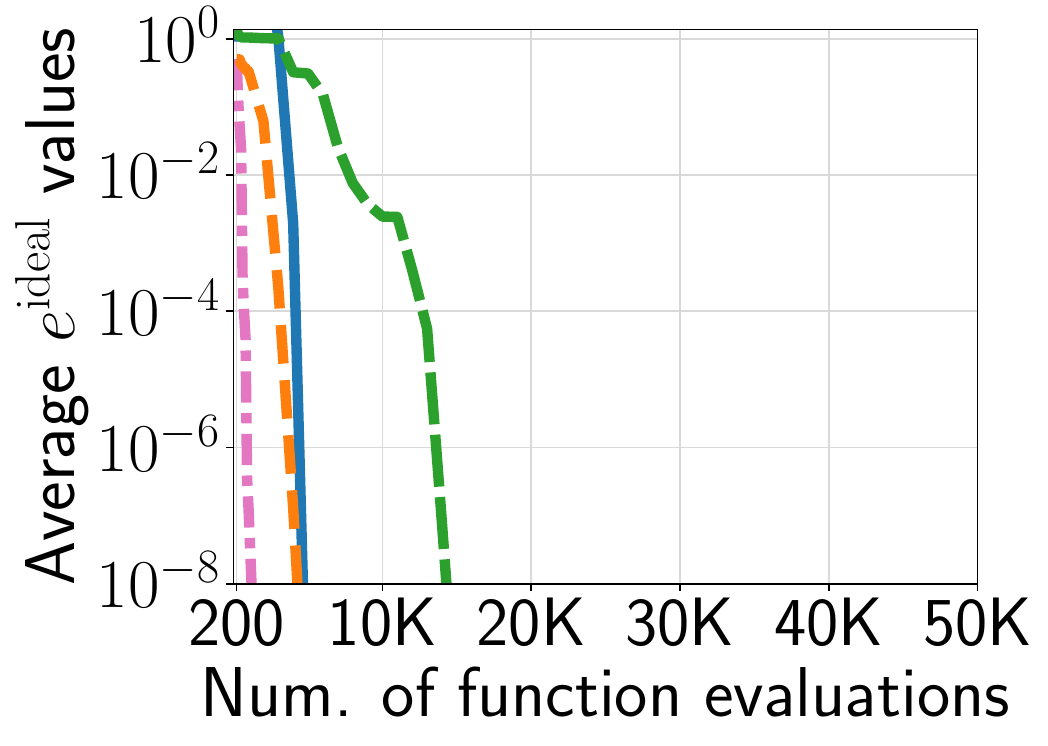}}
   \subfloat[$e^{\mathrm{ideal}}$ ($m=4$)]{\includegraphics[width=0.32\textwidth]{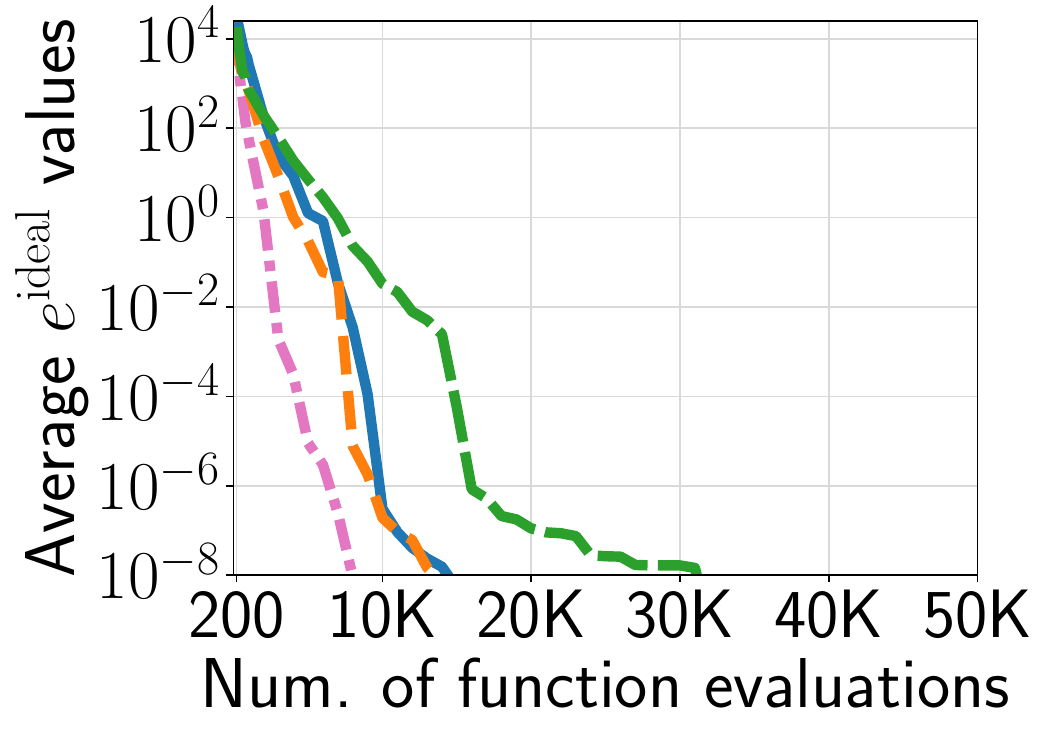}}
   \subfloat[$e^{\mathrm{ideal}}$ ($m=6$)]{\includegraphics[width=0.32\textwidth]{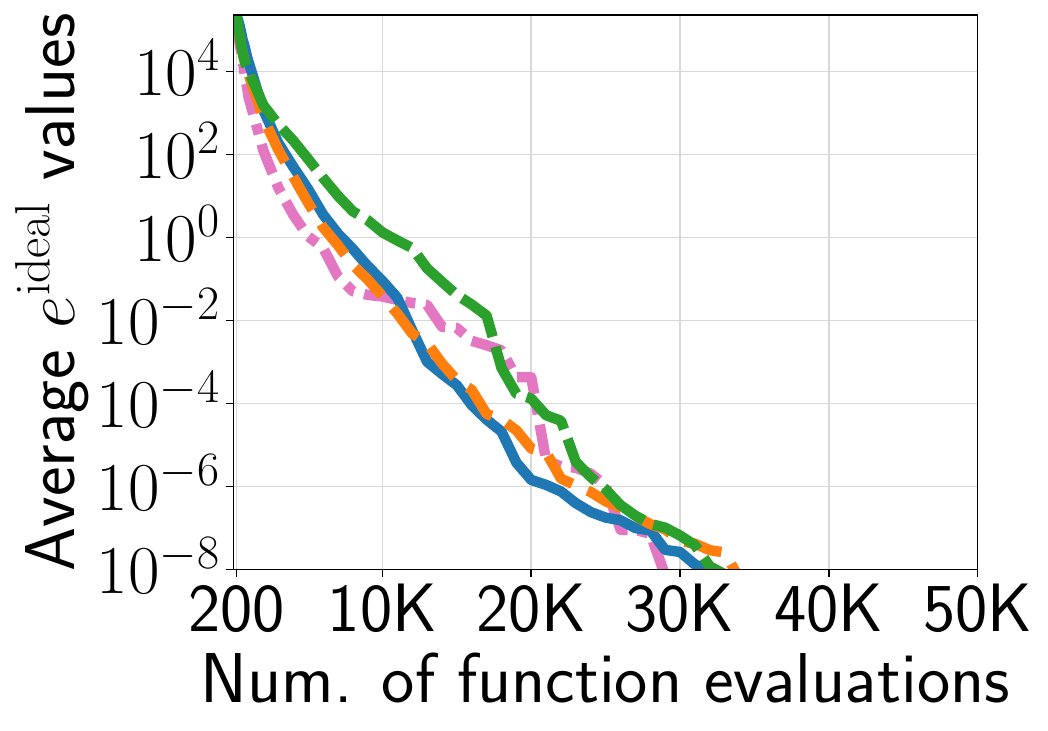}}
\\
   \subfloat[$e^{\mathrm{nadir}}$ ($m=2$)]{\includegraphics[width=0.32\textwidth]{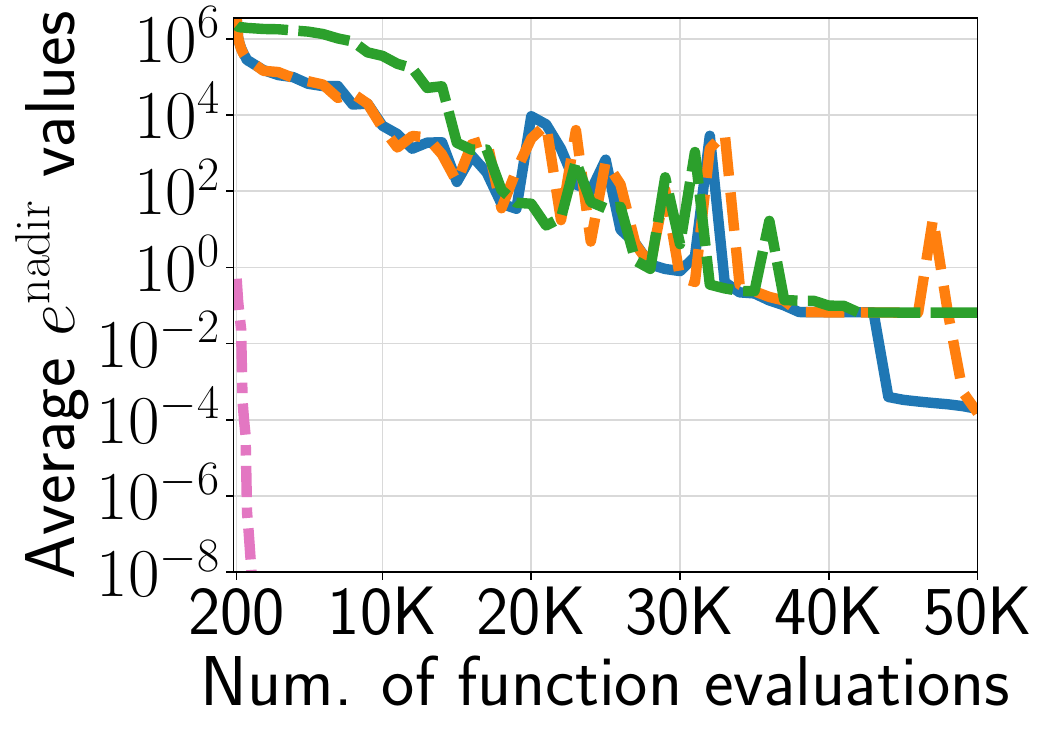}}
   \subfloat[$e^{\mathrm{nadir}}$ ($m=4$)]{\includegraphics[width=0.32\textwidth]{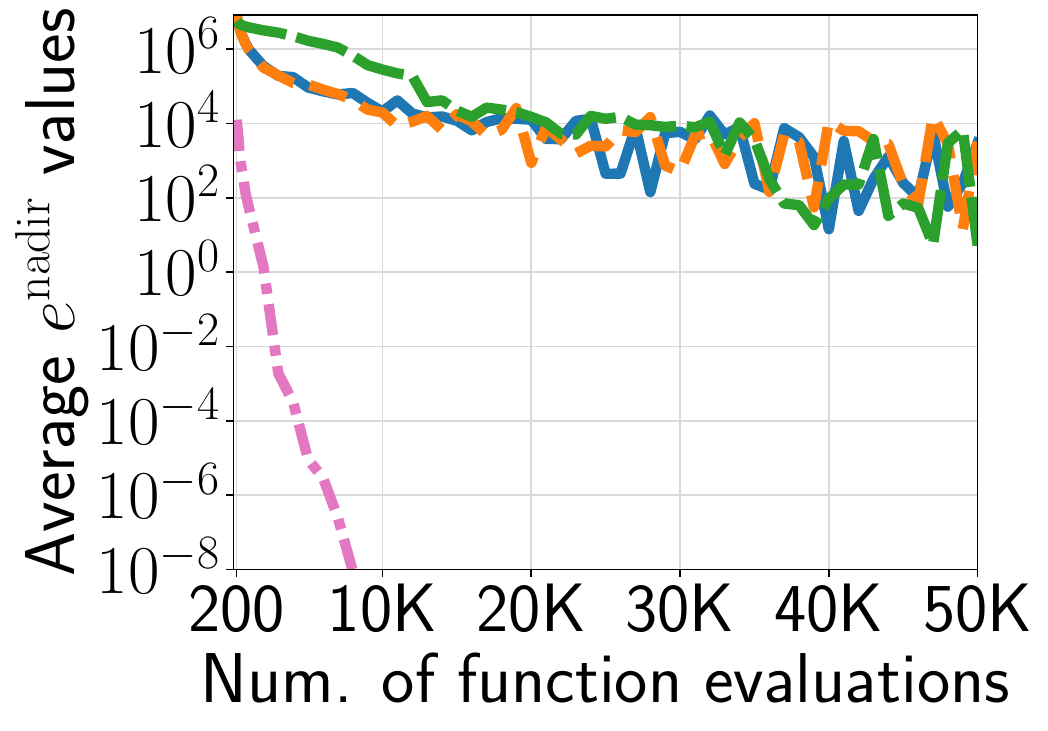}}
   \subfloat[$e^{\mathrm{nadir}}$ ($m=6$)]{\includegraphics[width=0.32\textwidth]{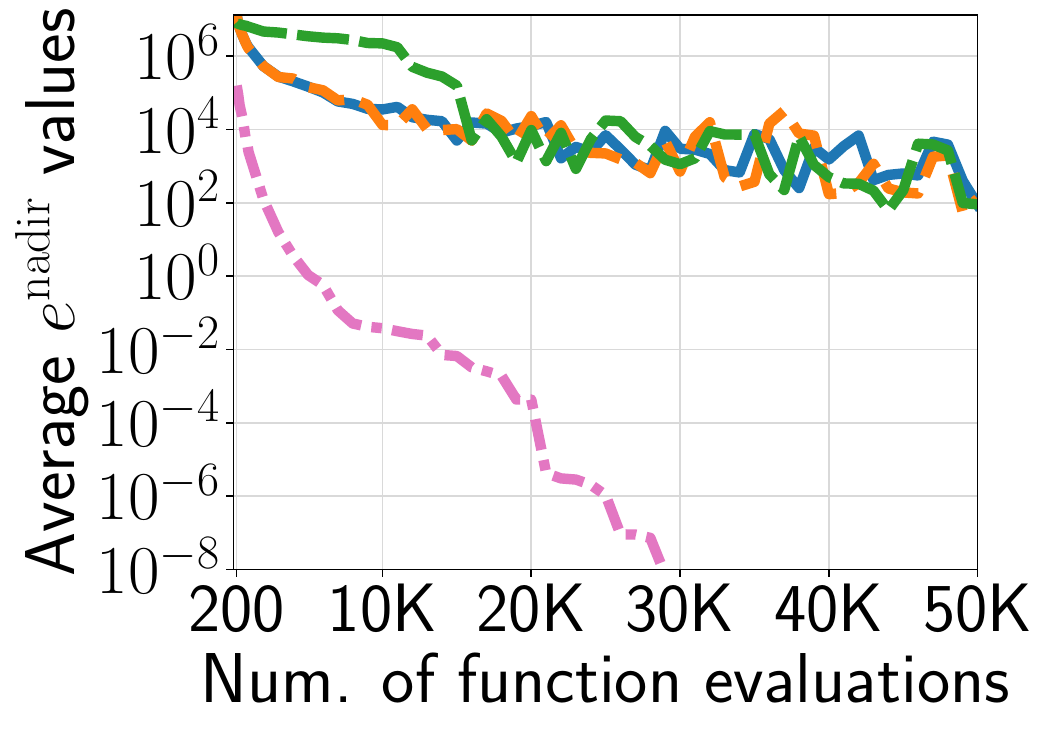}}
\\
   \subfloat[ORE ($m=2$)]{\includegraphics[width=0.32\textwidth]{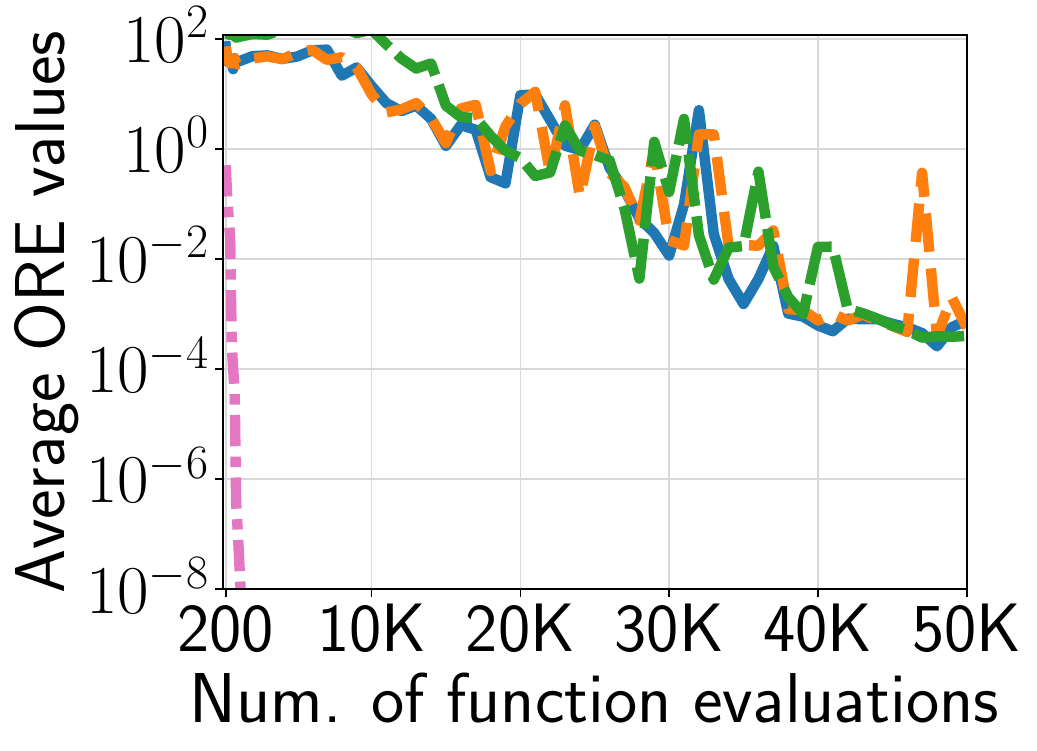}}
   \subfloat[ORE ($m=4$)]{\includegraphics[width=0.32\textwidth]{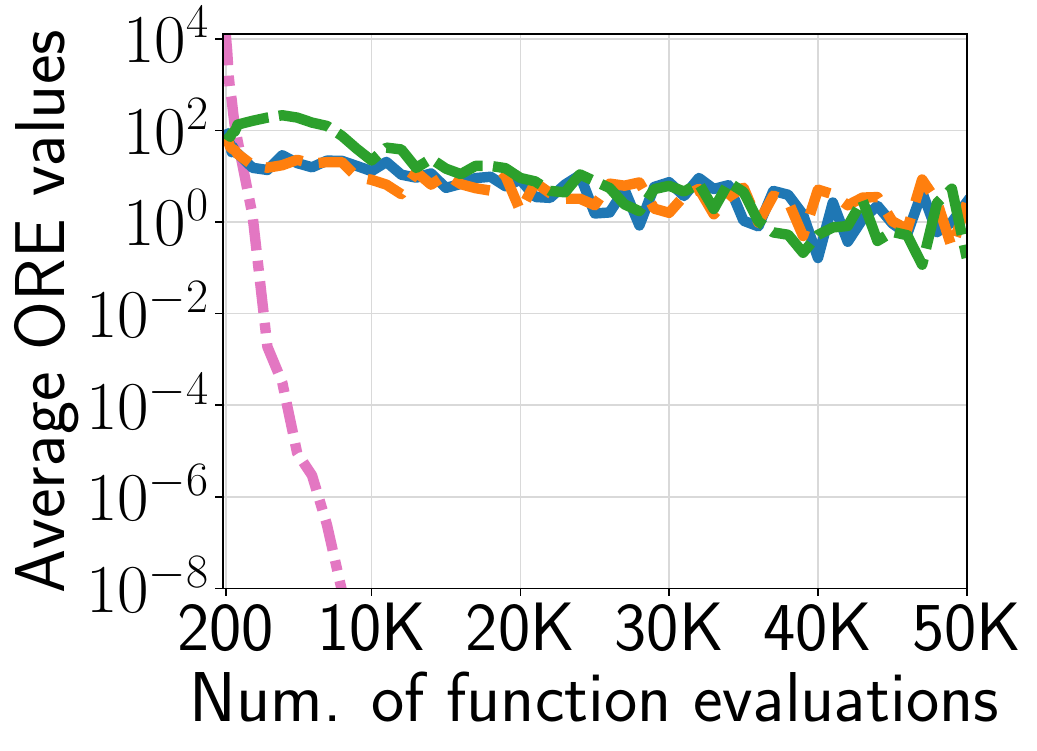}}
   \subfloat[ORE ($m=6$)]{\includegraphics[width=0.32\textwidth]{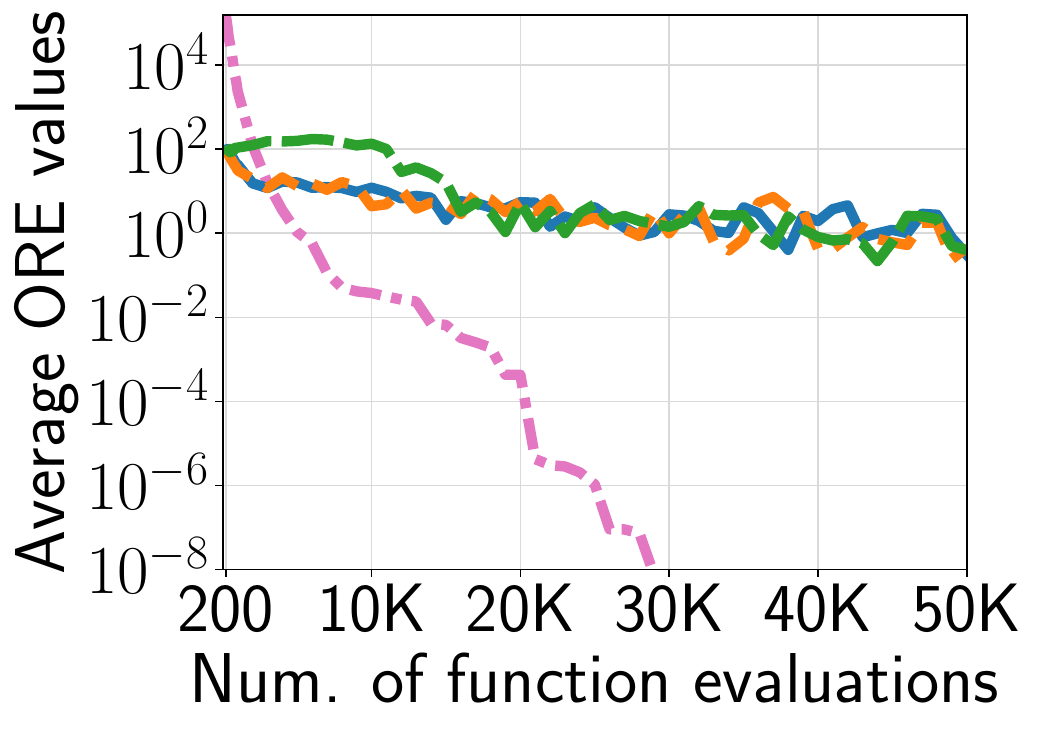}}
\\
\caption{Average $e^{\mathrm{ideal}}$, $e^{\mathrm{nadir}}$, and ORE values of the three normalization methods in r-NSGA-II on IDTLZ3.}
\label{supfig:3error_rNSGA2_IDTLZ3}
\end{figure*}

\begin{figure*}[t]
\centering
  \subfloat{\includegraphics[width=0.7\textwidth]{./figs/legend/legend_3.pdf}}
\vspace{-3.9mm}
   \\
   \subfloat[$e^{\mathrm{ideal}}$ ($m=2$)]{\includegraphics[width=0.32\textwidth]{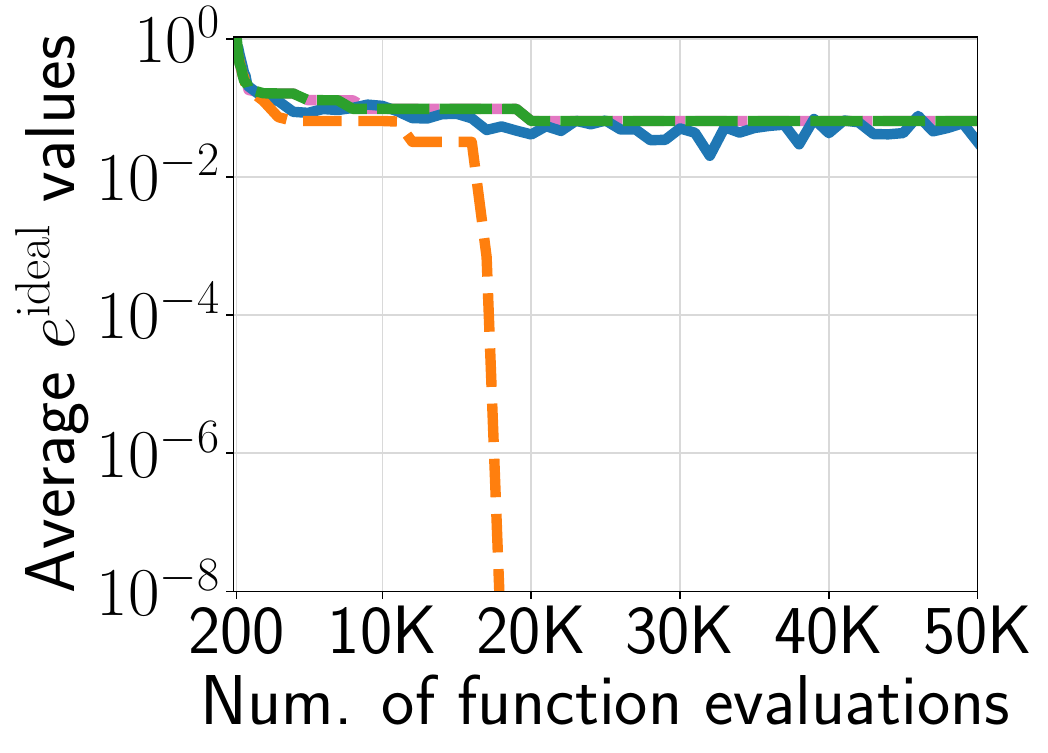}}
   \subfloat[$e^{\mathrm{ideal}}$ ($m=4$)]{\includegraphics[width=0.32\textwidth]{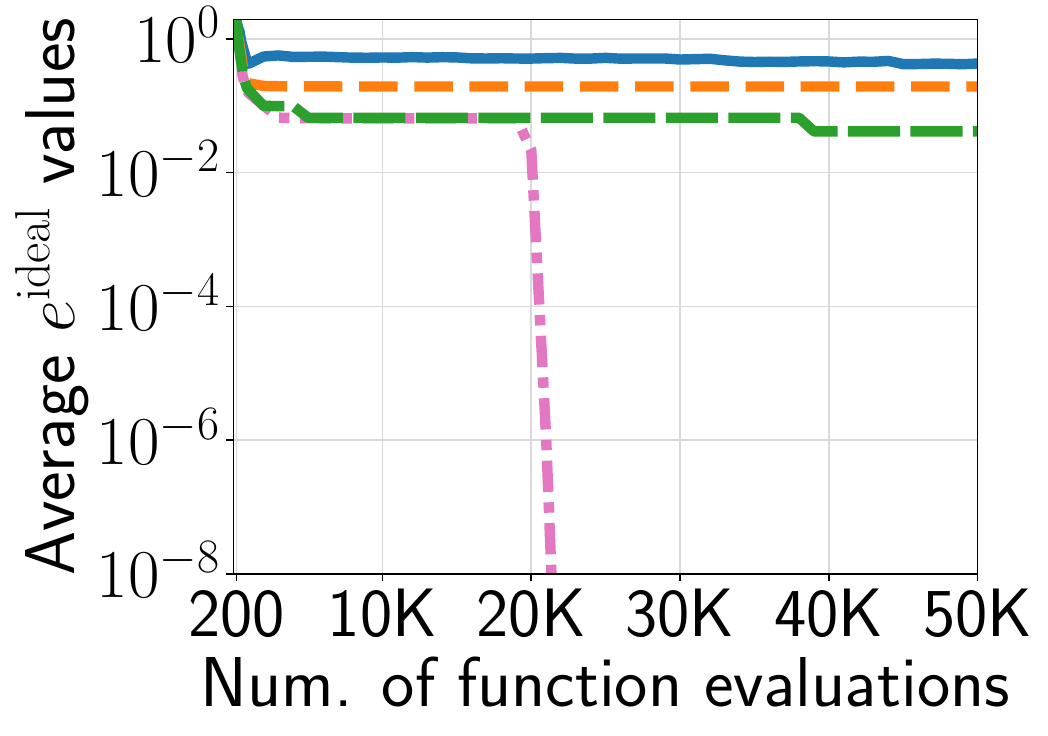}}
   \subfloat[$e^{\mathrm{ideal}}$ ($m=6$)]{\includegraphics[width=0.32\textwidth]{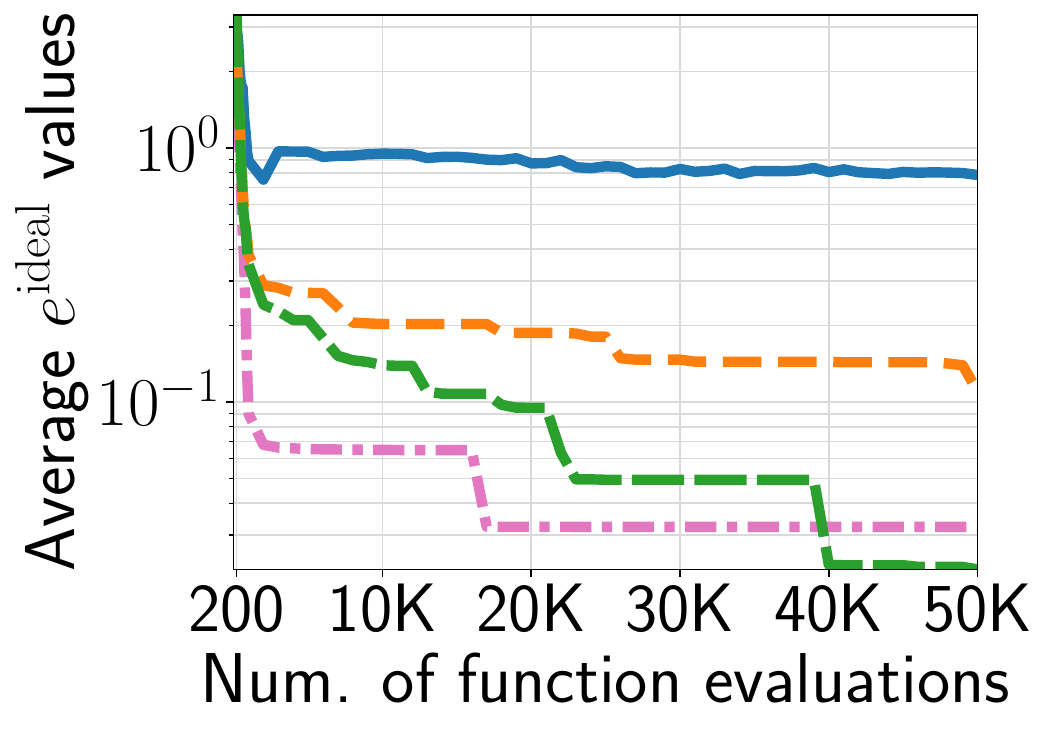}}
\\
   \subfloat[$e^{\mathrm{nadir}}$ ($m=2$)]{\includegraphics[width=0.32\textwidth]{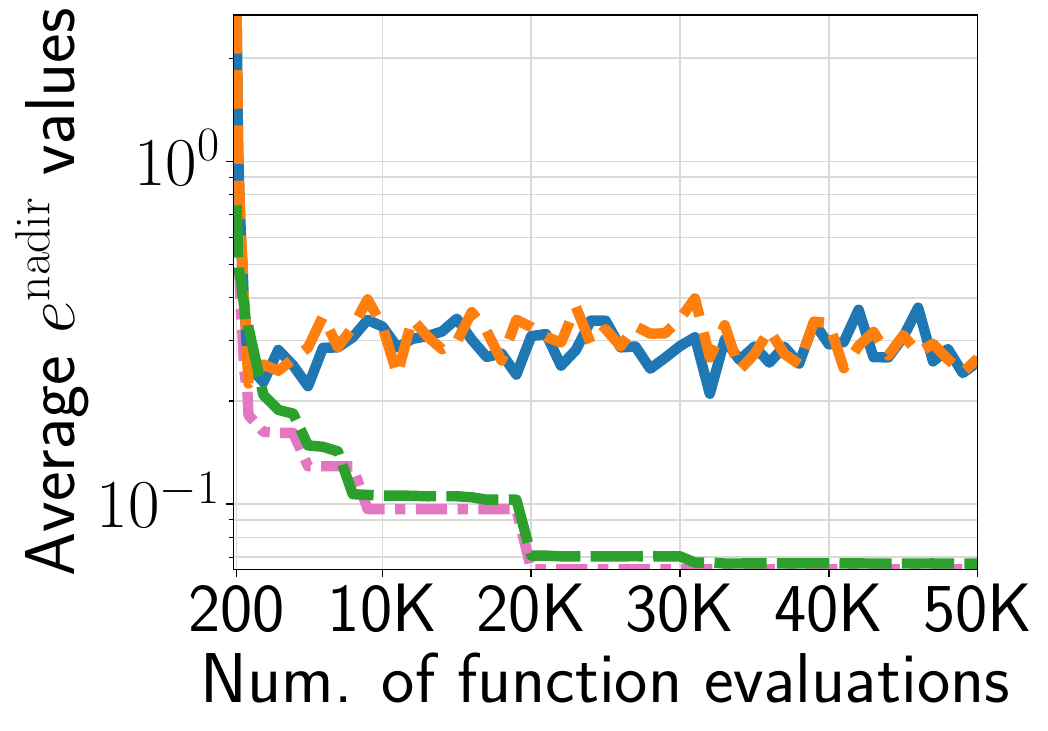}}
   \subfloat[$e^{\mathrm{nadir}}$ ($m=4$)]{\includegraphics[width=0.32\textwidth]{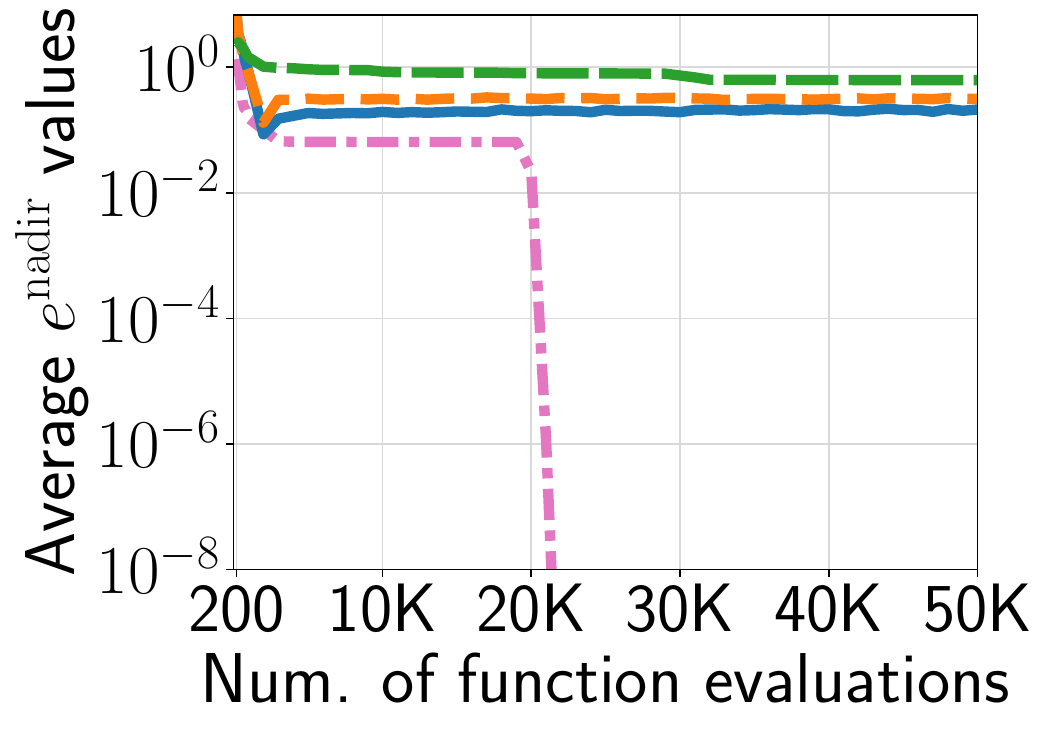}}
   \subfloat[$e^{\mathrm{nadir}}$ ($m=6$)]{\includegraphics[width=0.32\textwidth]{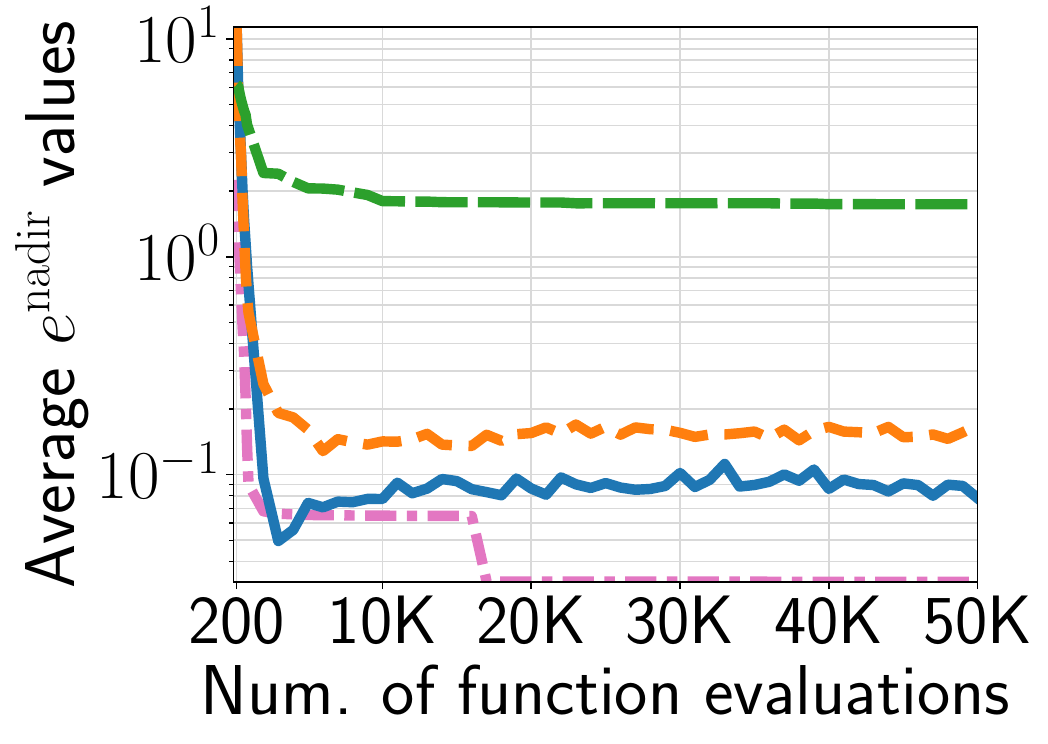}}
\\
   \subfloat[ORE ($m=2$)]{\includegraphics[width=0.32\textwidth]{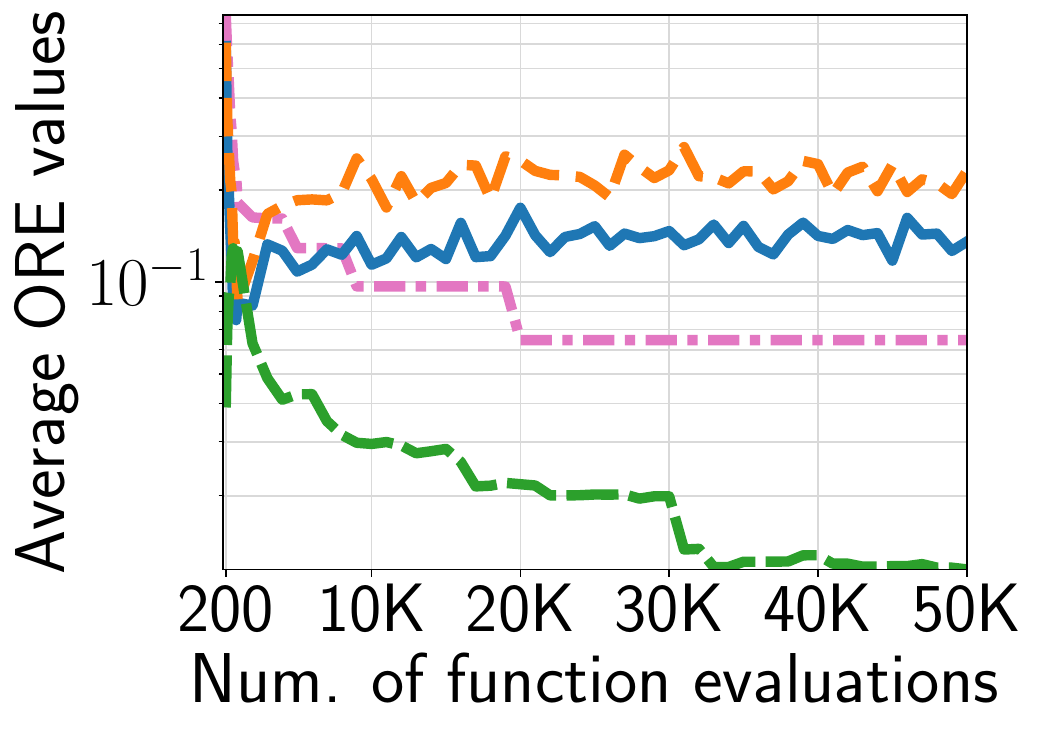}}
   \subfloat[ORE ($m=4$)]{\includegraphics[width=0.32\textwidth]{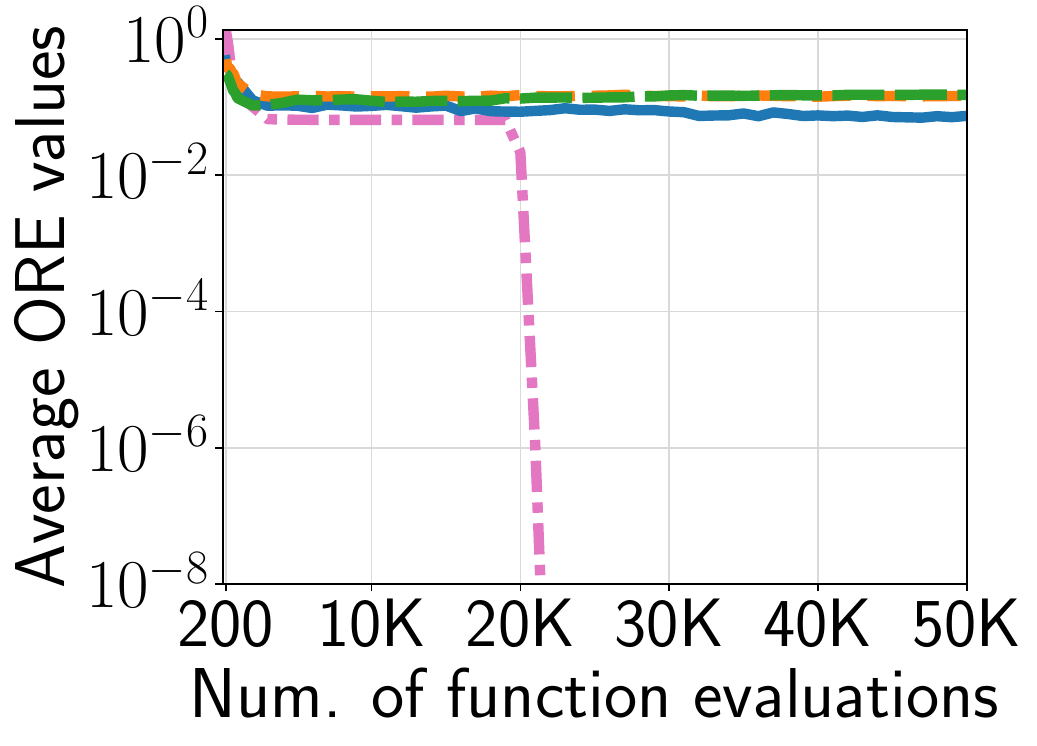}}
   \subfloat[ORE ($m=6$)]{\includegraphics[width=0.32\textwidth]{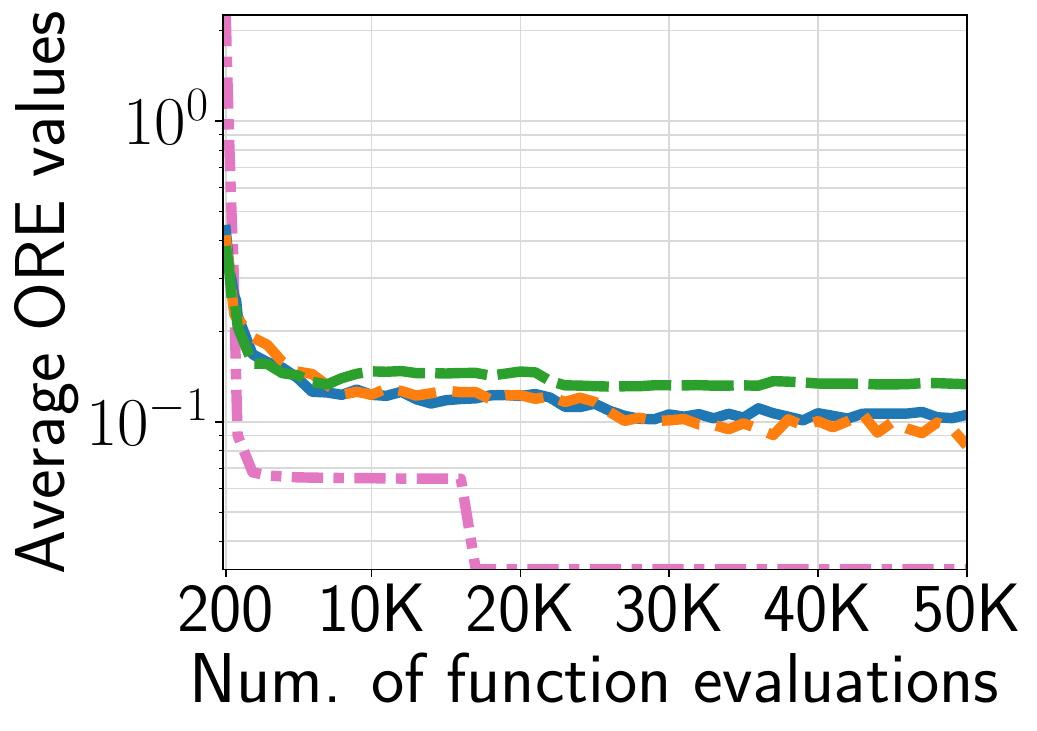}}
\caption{Average $e^{\mathrm{ideal}}$, $e^{\mathrm{nadir}}$, and ORE values of the three normalization methods in r-NSGA-II on IDTLZ4.}
\label{supfig:3error_rNSGA2_IDTLZ4}
\end{figure*}

\clearpage

\begin{figure*}[t]
\centering
  \subfloat{\includegraphics[width=0.7\textwidth]{./figs/legend/legend_3.pdf}}
\vspace{-3.9mm}
   \\
   \subfloat[$e^{\mathrm{ideal}}$ ($m=2$)]{\includegraphics[width=0.32\textwidth]{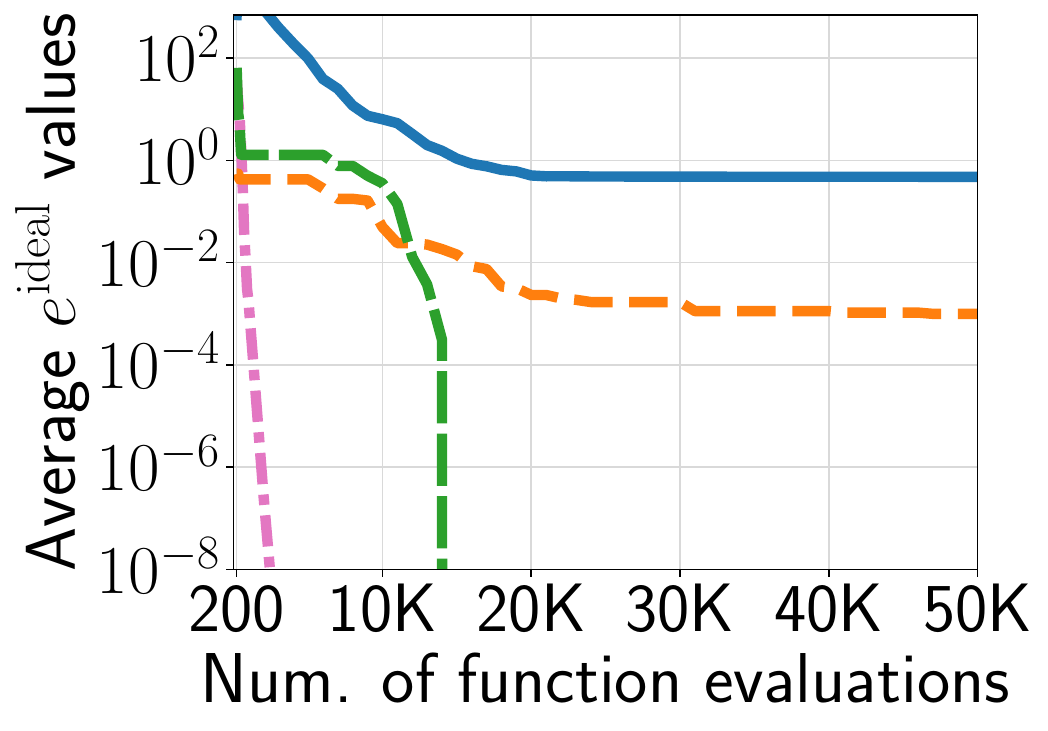}}
   \subfloat[$e^{\mathrm{ideal}}$ ($m=4$)]{\includegraphics[width=0.32\textwidth]{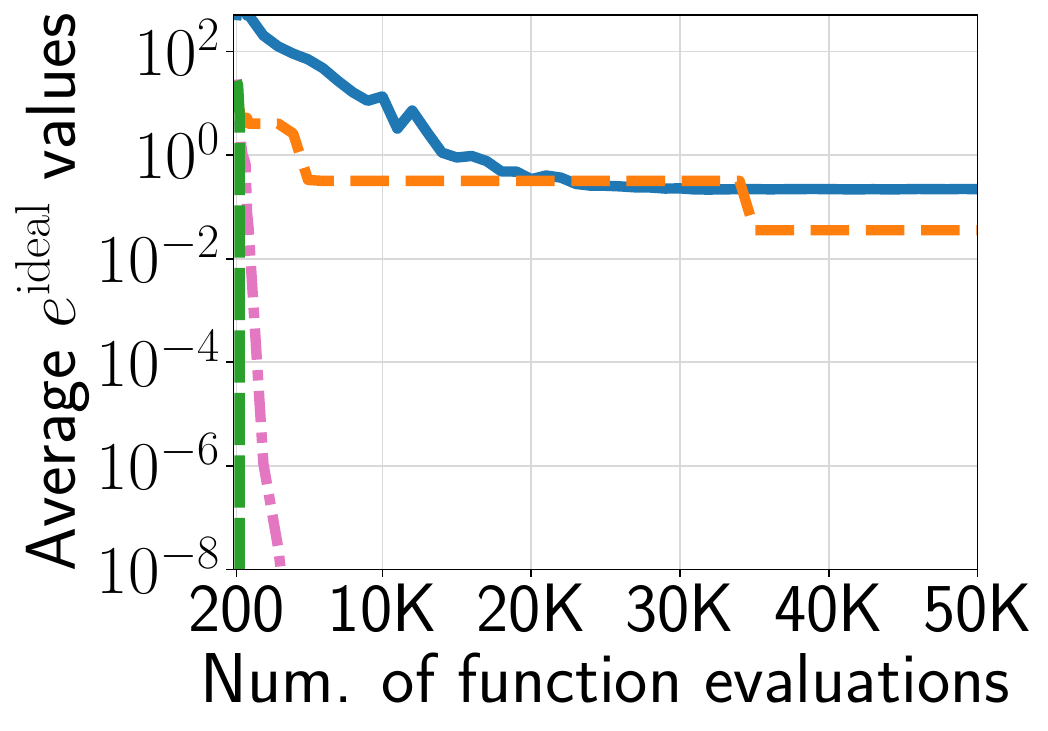}}
   \subfloat[$e^{\mathrm{ideal}}$ ($m=6$)]{\includegraphics[width=0.32\textwidth]{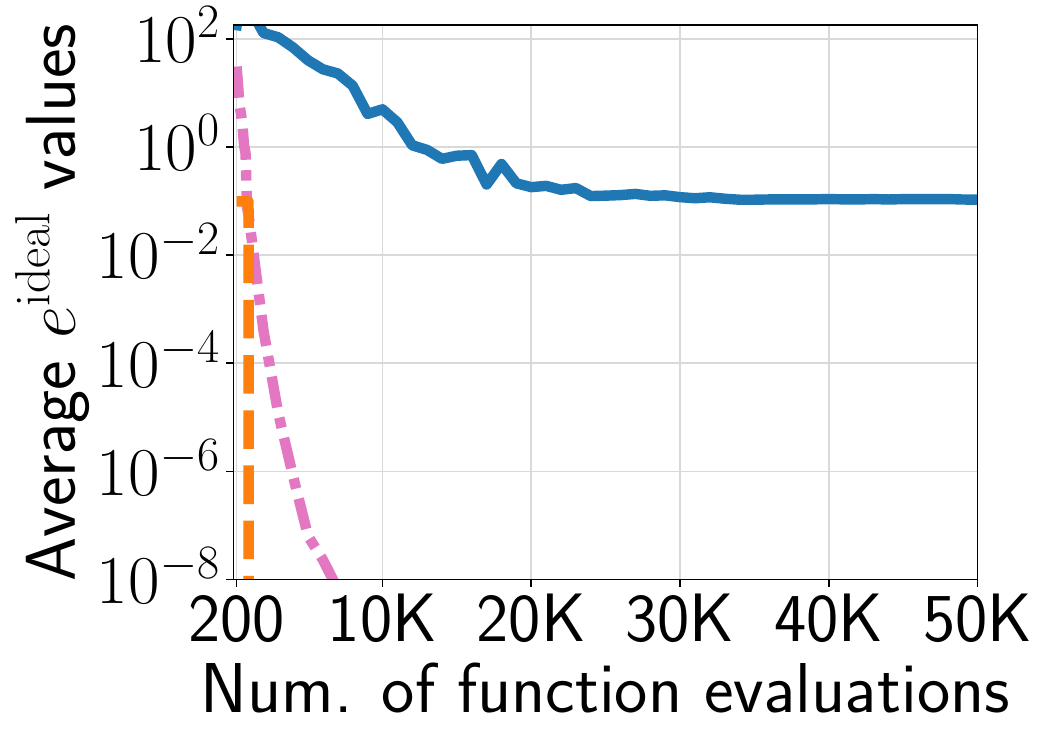}}
\\
   \subfloat[$e^{\mathrm{nadir}}$ ($m=2$)]{\includegraphics[width=0.32\textwidth]{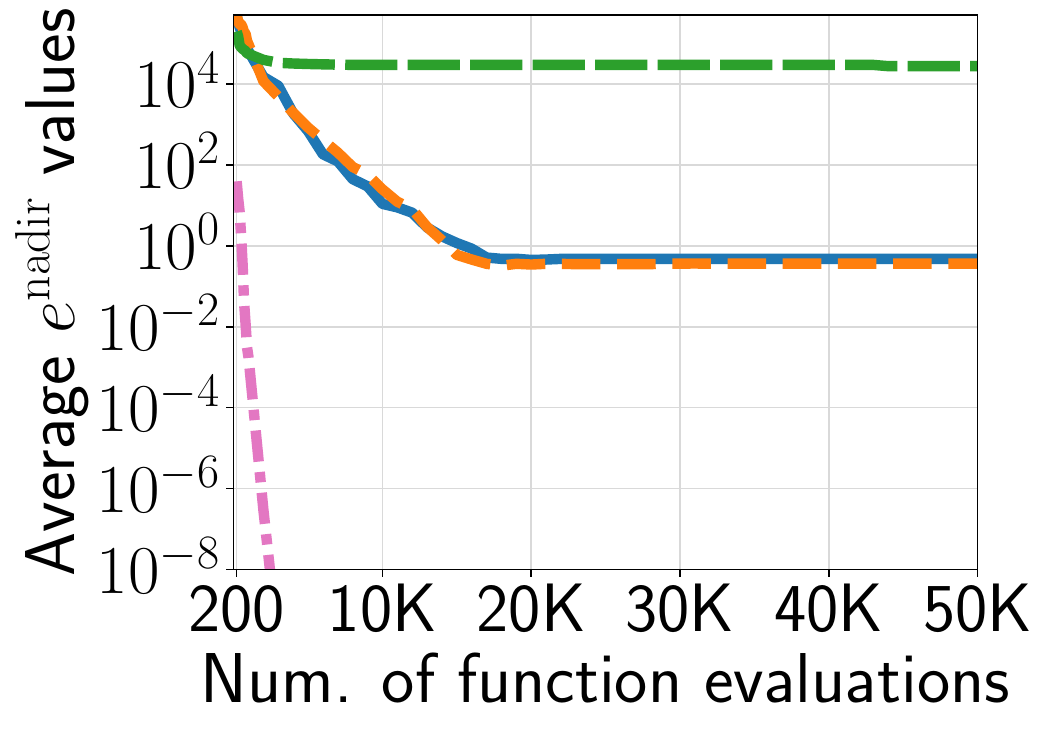}}
   \subfloat[$e^{\mathrm{nadir}}$ ($m=4$)]{\includegraphics[width=0.32\textwidth]{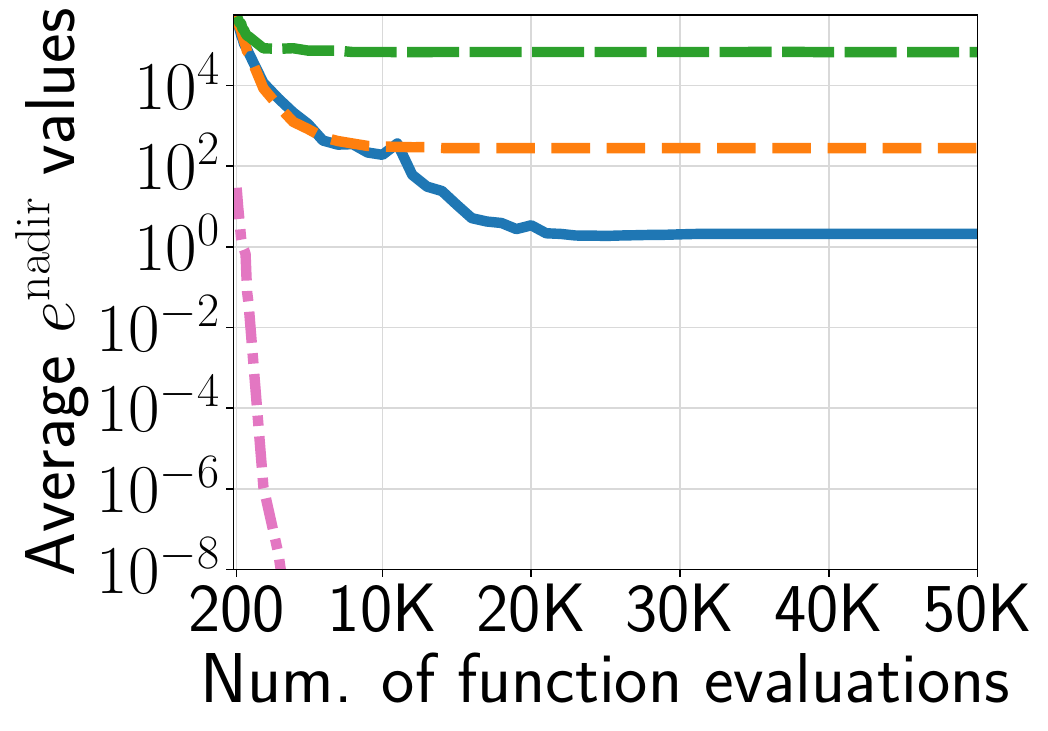}}
   \subfloat[$e^{\mathrm{nadir}}$ ($m=6$)]{\includegraphics[width=0.32\textwidth]{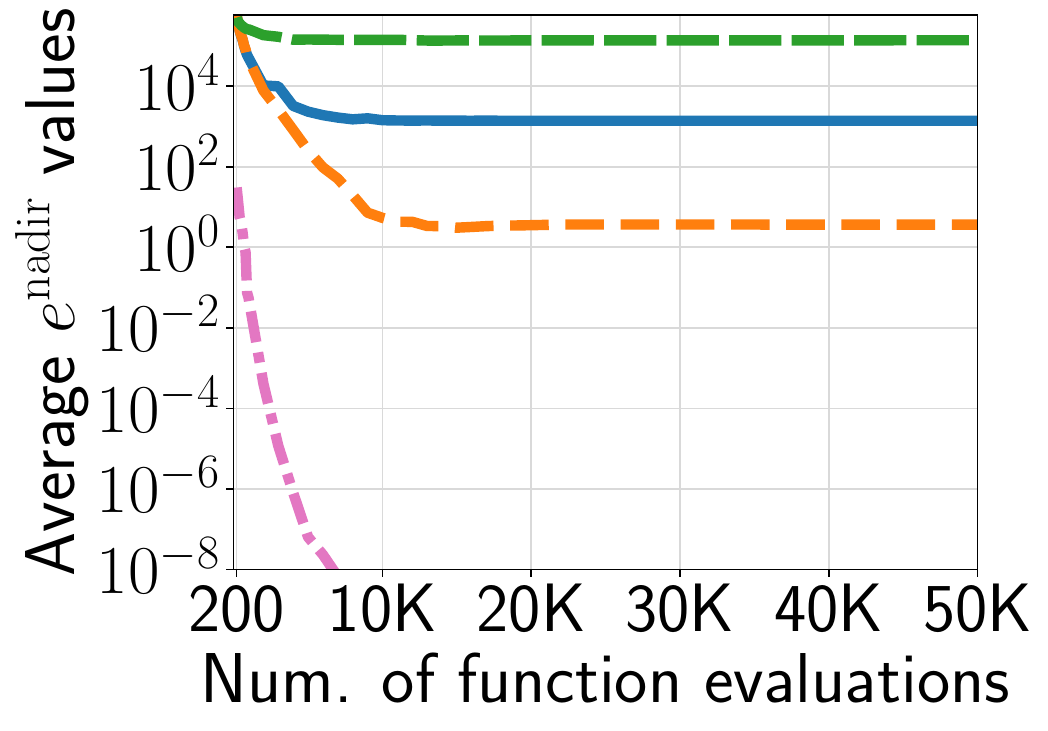}}
\\
   \subfloat[ORE ($m=2$)]{\includegraphics[width=0.32\textwidth]{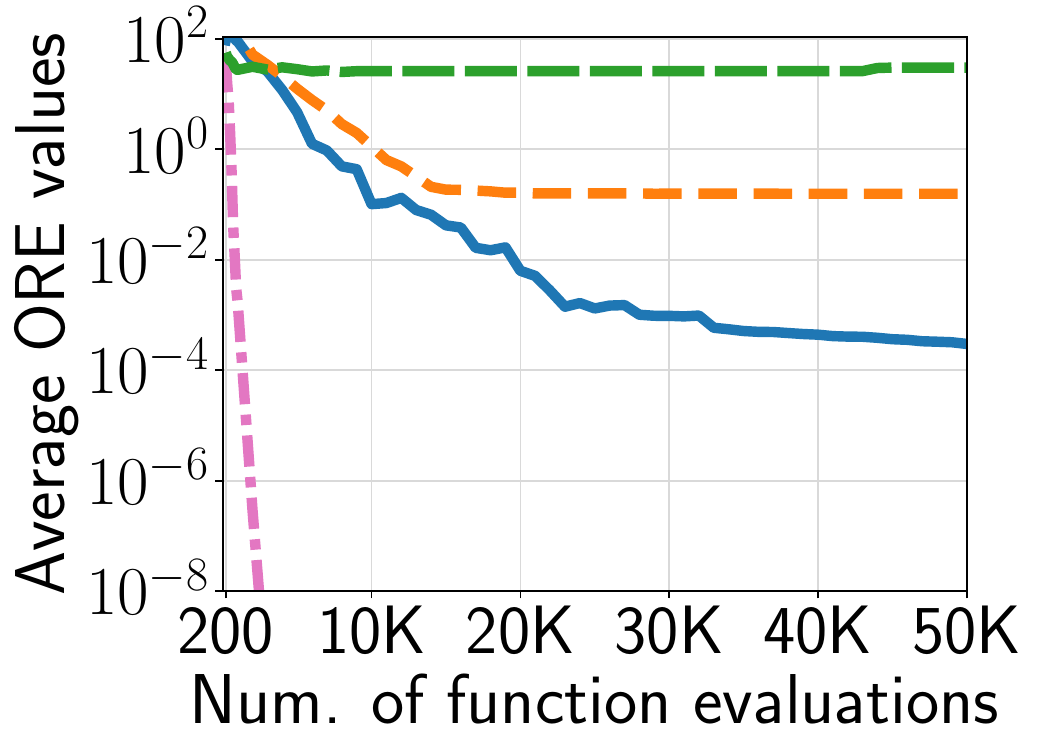}}
   \subfloat[ORE ($m=4$)]{\includegraphics[width=0.32\textwidth]{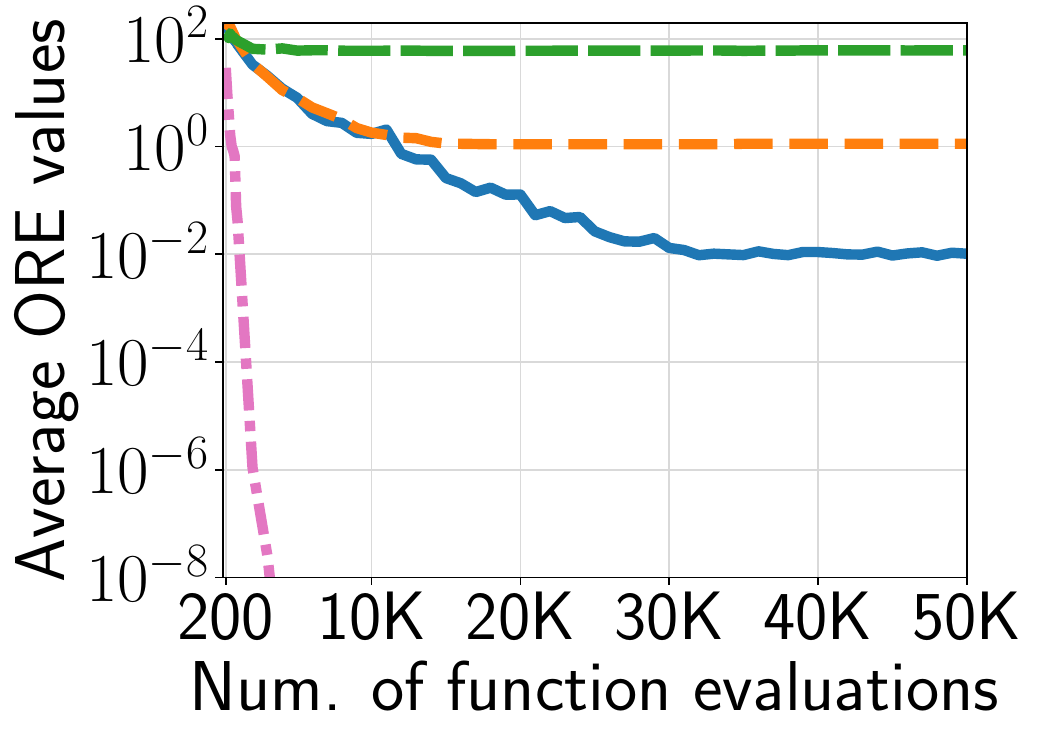}}
   \subfloat[ORE ($m=6$)]{\includegraphics[width=0.32\textwidth]{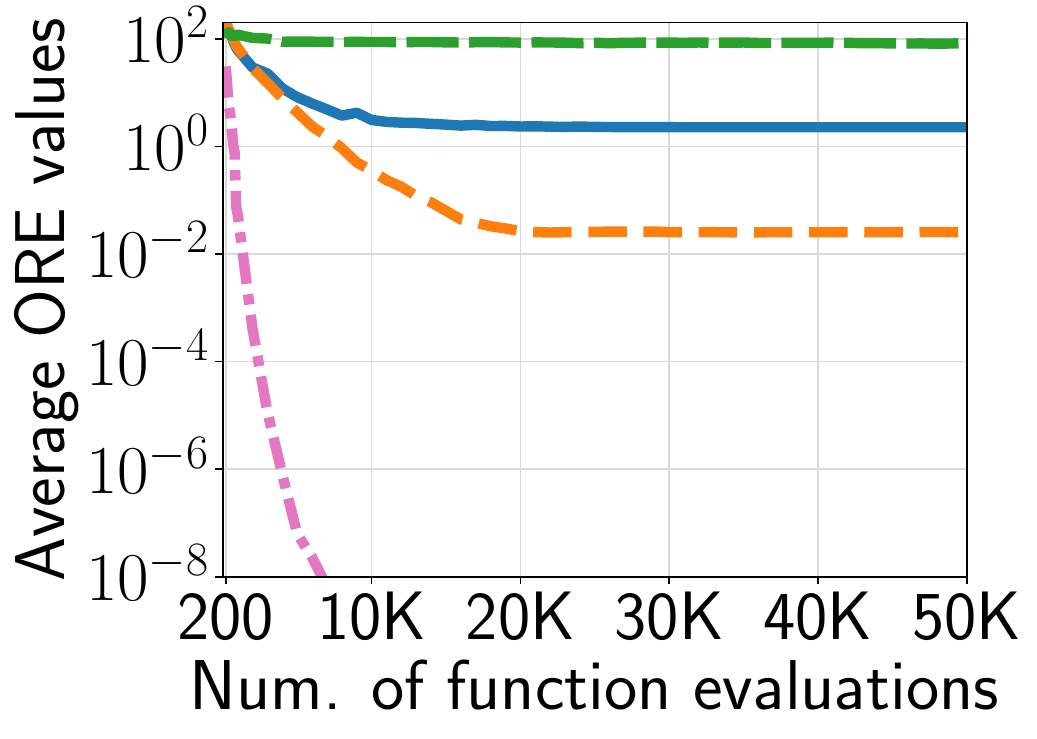}}
\\
\caption{Average $e^{\mathrm{ideal}}$, $e^{\mathrm{nadir}}$, and ORE values of the three normalization methods in MOEA/D-NUMS on DTLZ1.}
\label{supfig:3error_MOEADNUMS_DTLZ1}
\end{figure*}

\begin{figure*}[t]
\centering
  \subfloat{\includegraphics[width=0.7\textwidth]{./figs/legend/legend_3.pdf}}
\vspace{-3.9mm}
   \\
   \subfloat[$e^{\mathrm{ideal}}$ ($m=2$)]{\includegraphics[width=0.32\textwidth]{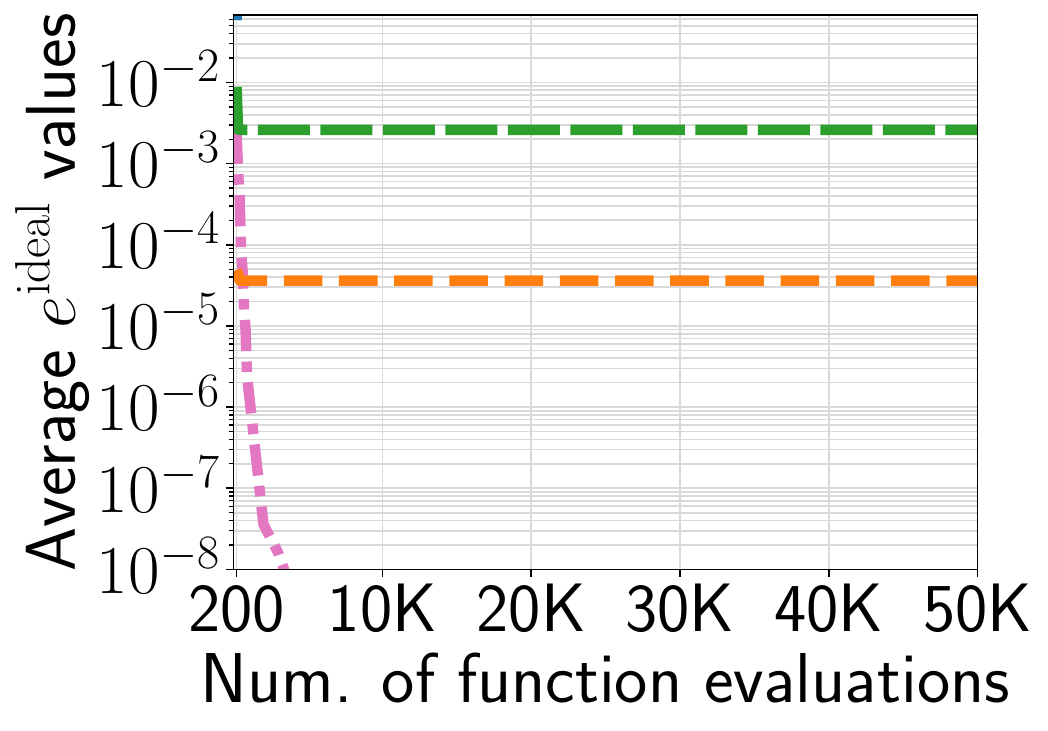}}
   \subfloat[$e^{\mathrm{ideal}}$ ($m=4$)]{\includegraphics[width=0.32\textwidth]{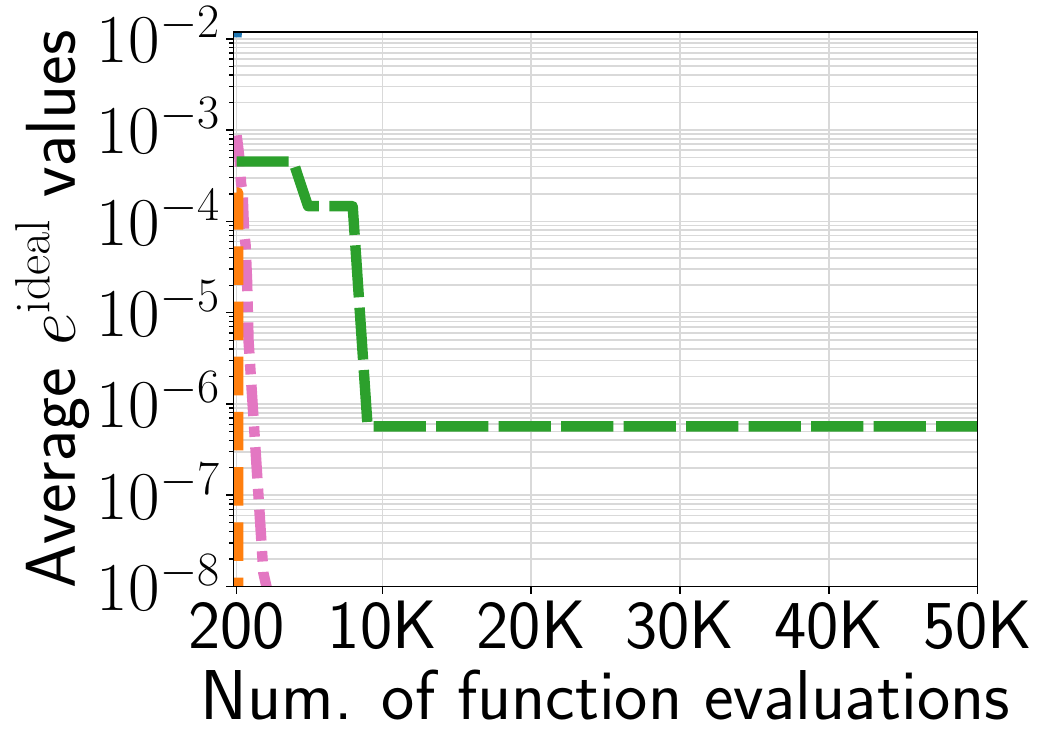}}
   \subfloat[$e^{\mathrm{ideal}}$ ($m=6$)]{\includegraphics[width=0.32\textwidth]{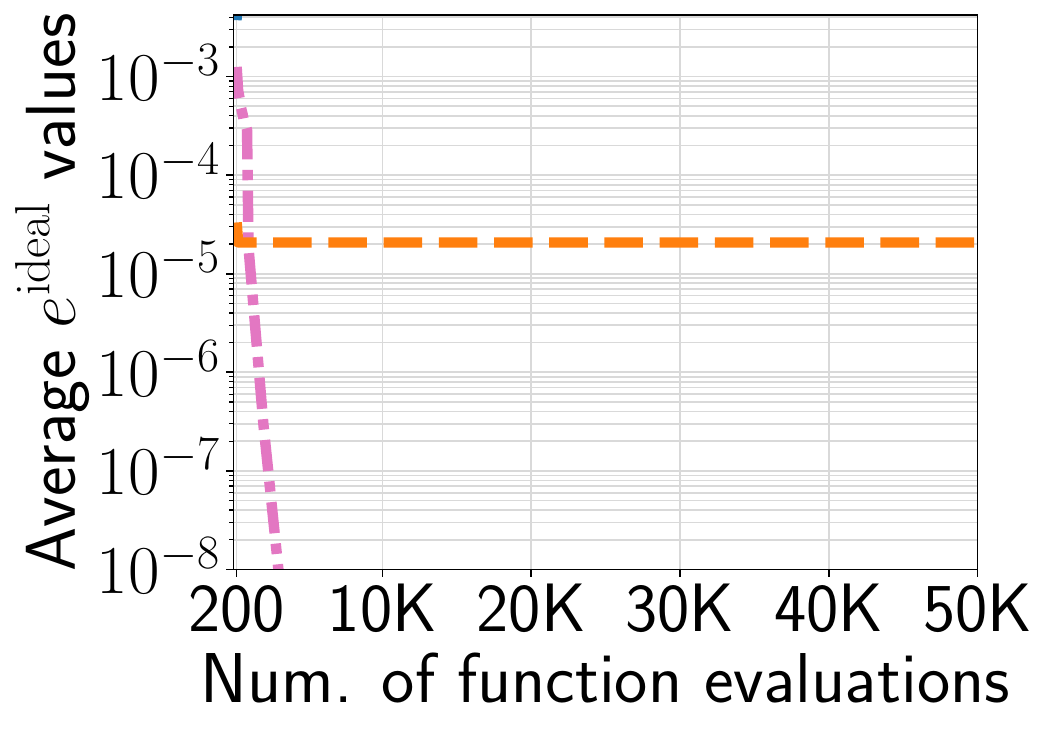}}
\\
   \subfloat[$e^{\mathrm{nadir}}$ ($m=2$)]{\includegraphics[width=0.32\textwidth]{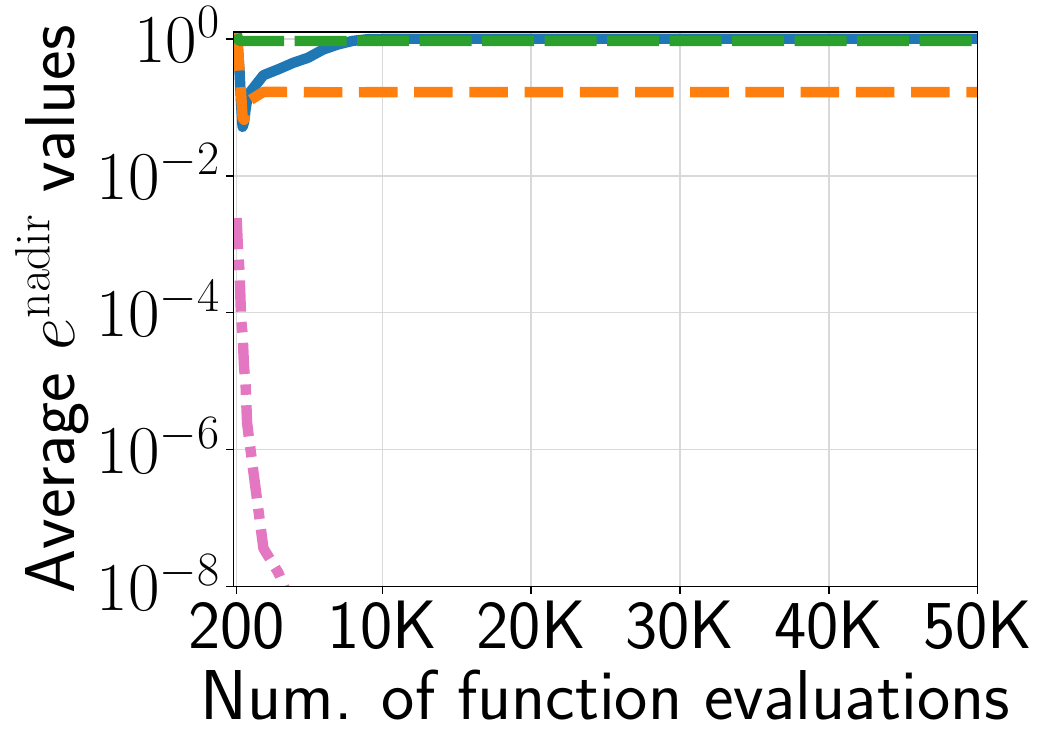}}
   \subfloat[$e^{\mathrm{nadir}}$ ($m=4$)]{\includegraphics[width=0.32\textwidth]{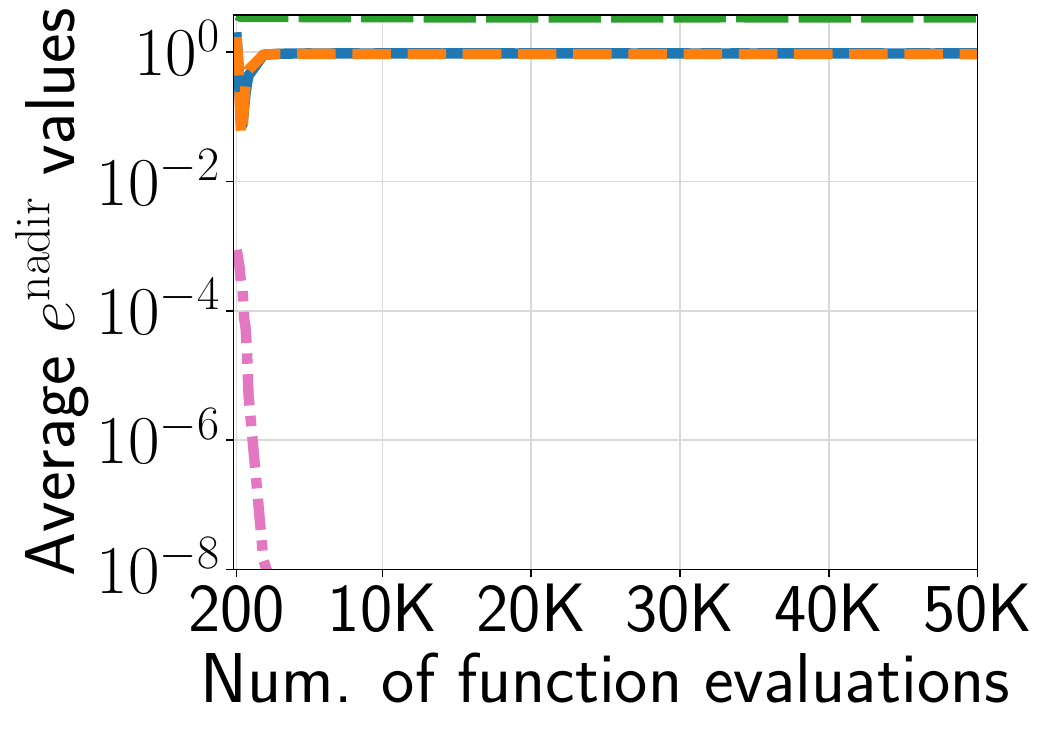}}
   \subfloat[$e^{\mathrm{nadir}}$ ($m=6$)]{\includegraphics[width=0.32\textwidth]{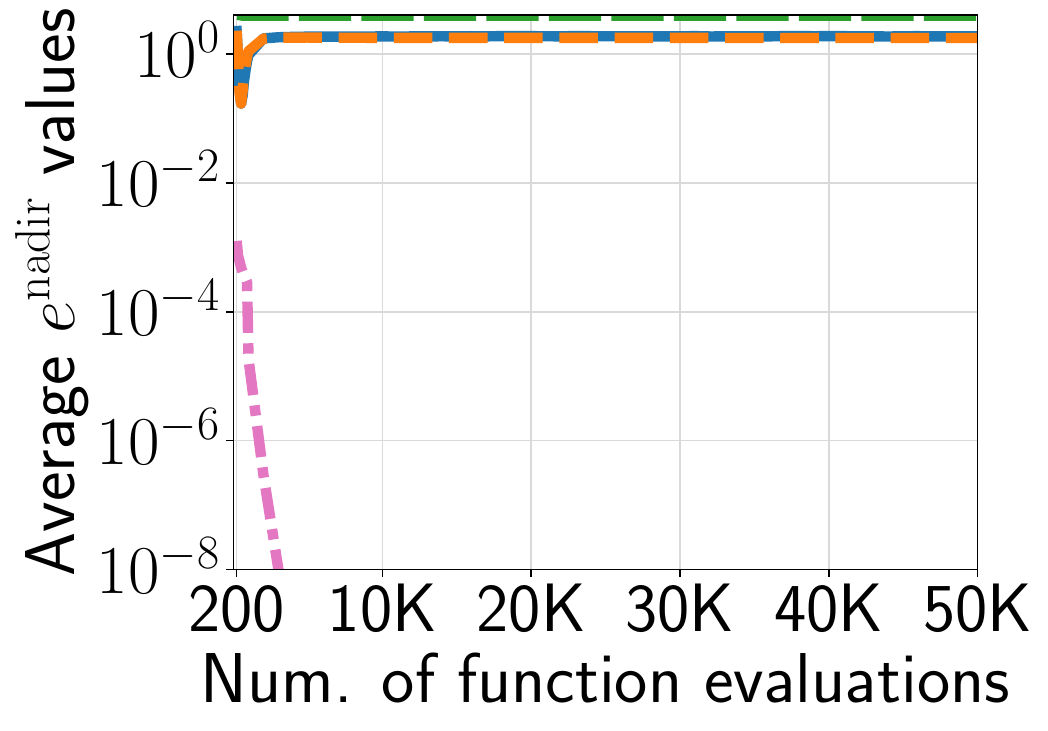}}
\\
   \subfloat[ORE ($m=2$)]{\includegraphics[width=0.32\textwidth]{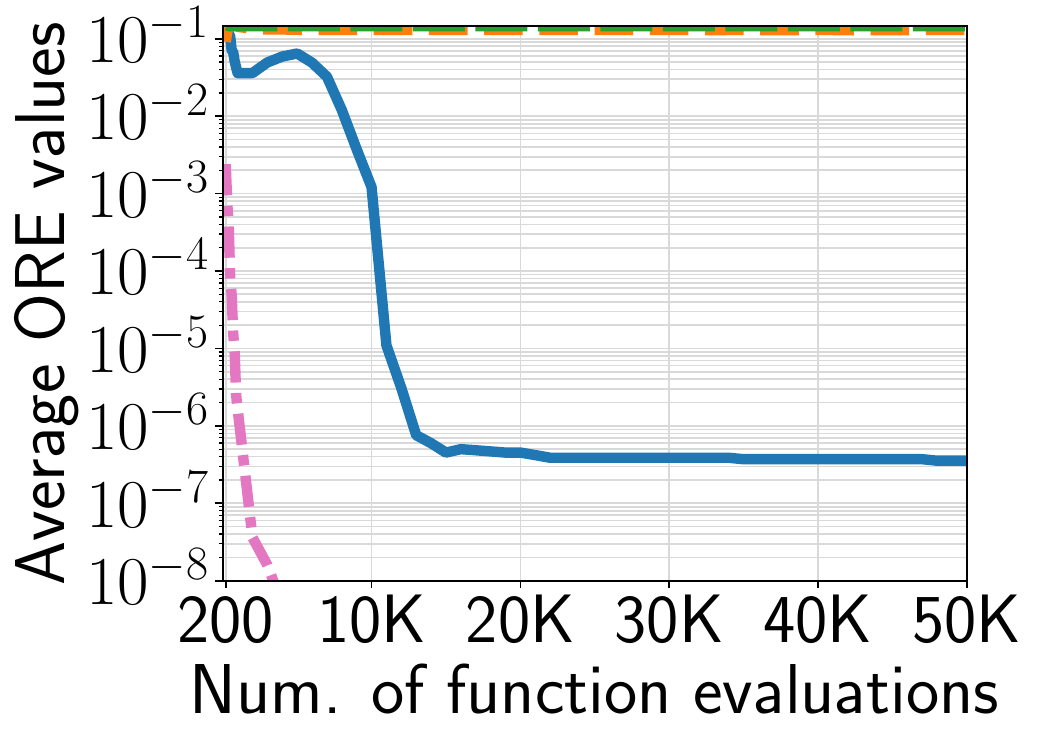}}
   \subfloat[ORE ($m=4$)]{\includegraphics[width=0.32\textwidth]{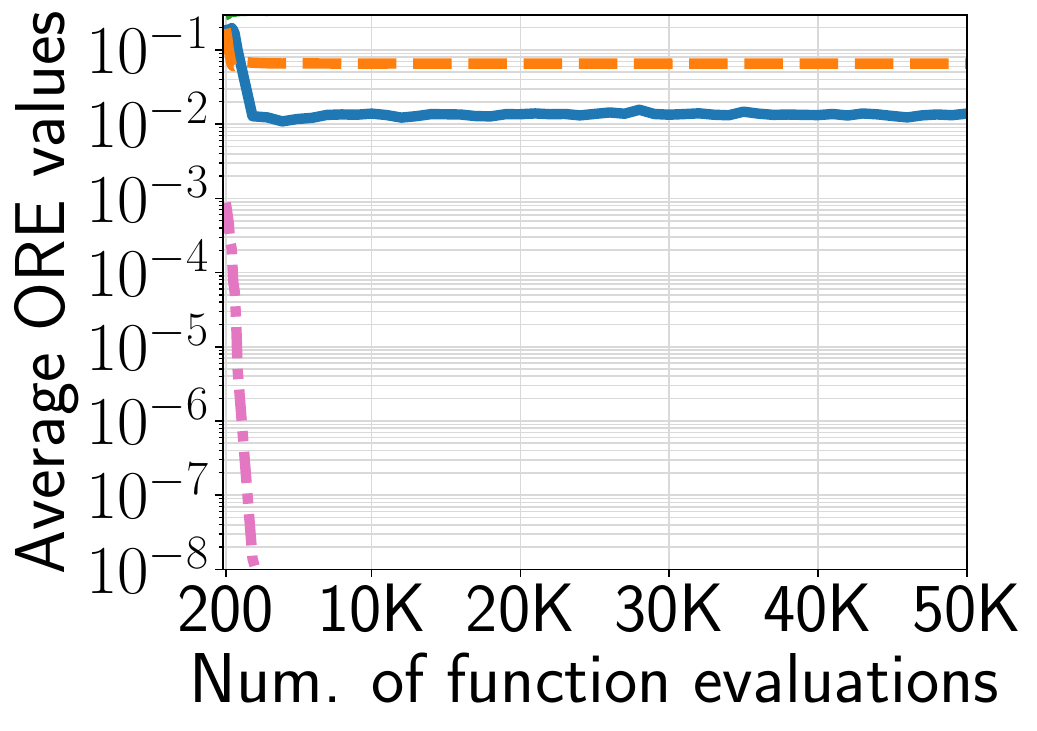}}
   \subfloat[ORE ($m=6$)]{\includegraphics[width=0.32\textwidth]{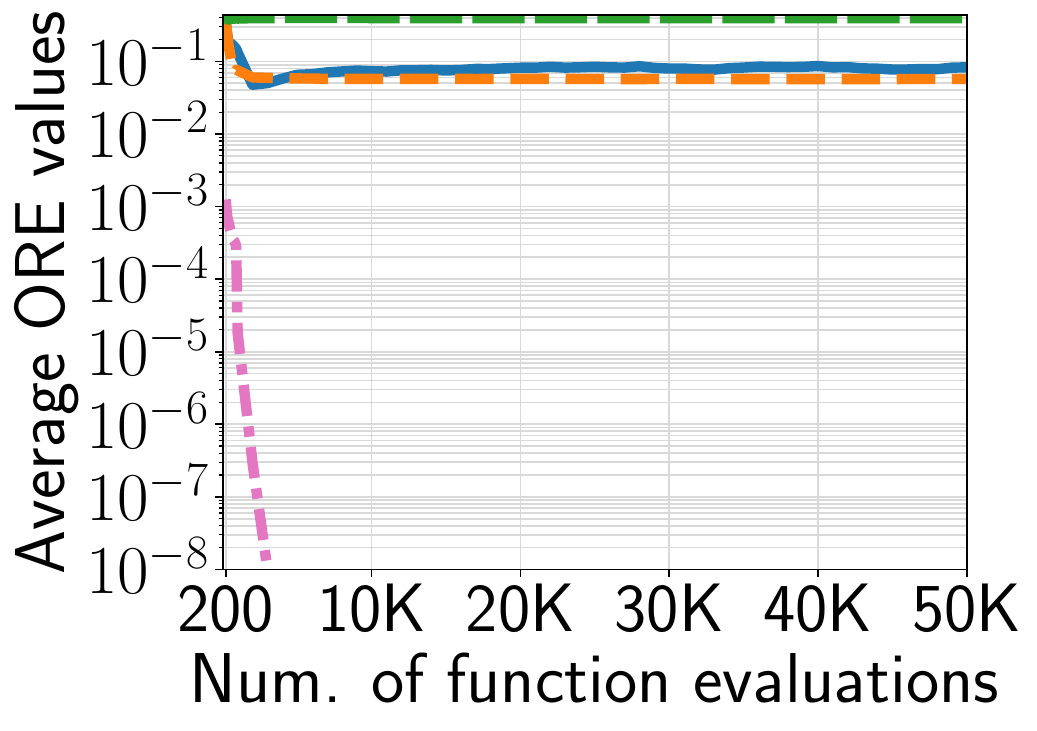}}
\\
\caption{Average $e^{\mathrm{ideal}}$, $e^{\mathrm{nadir}}$, and ORE values of the three normalization methods in MOEA/D-NUMS on DTLZ2.}
\label{supfig:3error_MOEADNUMS_DTLZ2}
\end{figure*}

\begin{figure*}[t]
\centering
  \subfloat{\includegraphics[width=0.7\textwidth]{./figs/legend/legend_3.pdf}}
\vspace{-3.9mm}
   \\
   \subfloat[$e^{\mathrm{ideal}}$ ($m=2$)]{\includegraphics[width=0.32\textwidth]{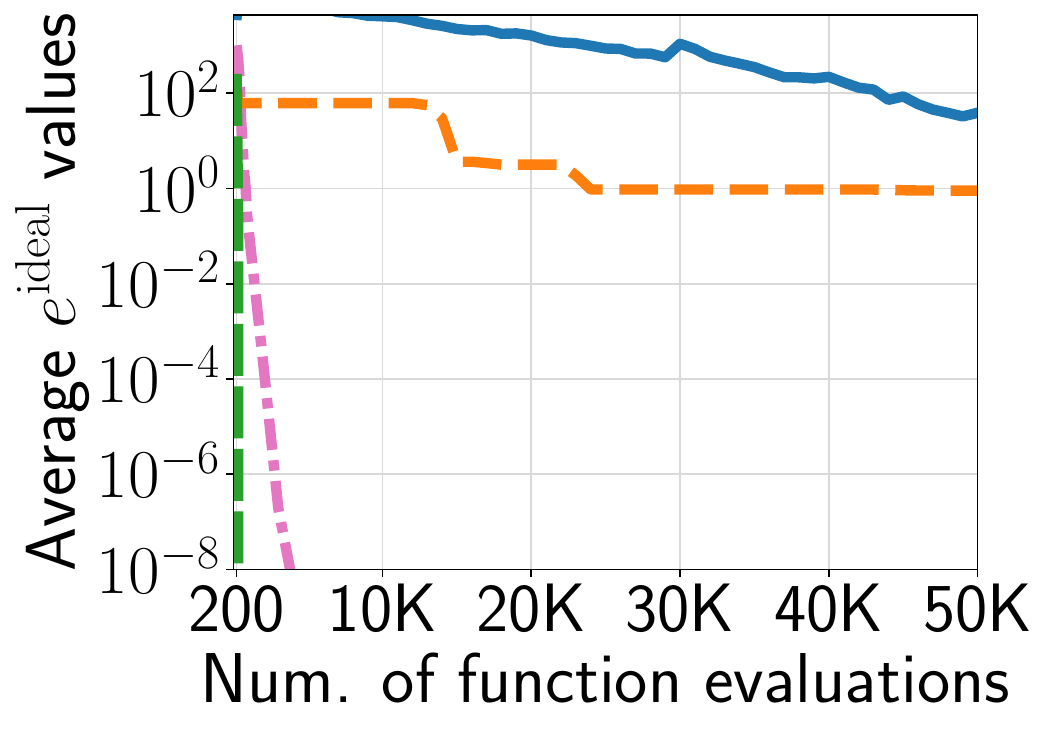}}
   \subfloat[$e^{\mathrm{ideal}}$ ($m=4$)]{\includegraphics[width=0.32\textwidth]{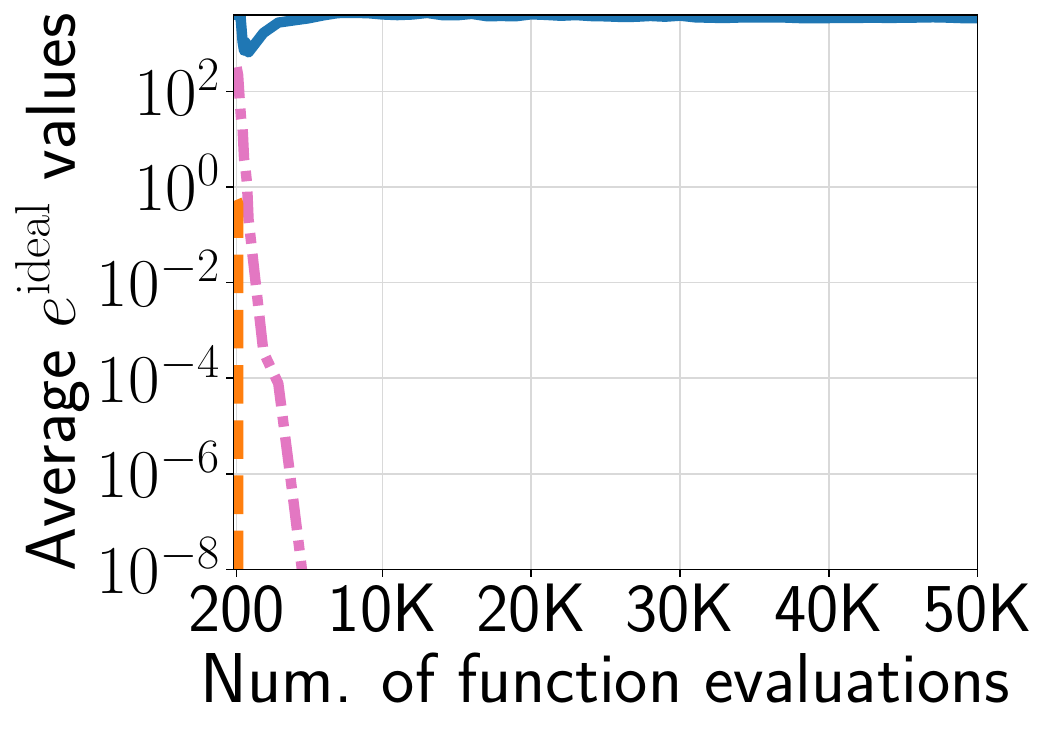}}
   \subfloat[$e^{\mathrm{ideal}}$ ($m=6$)]{\includegraphics[width=0.32\textwidth]{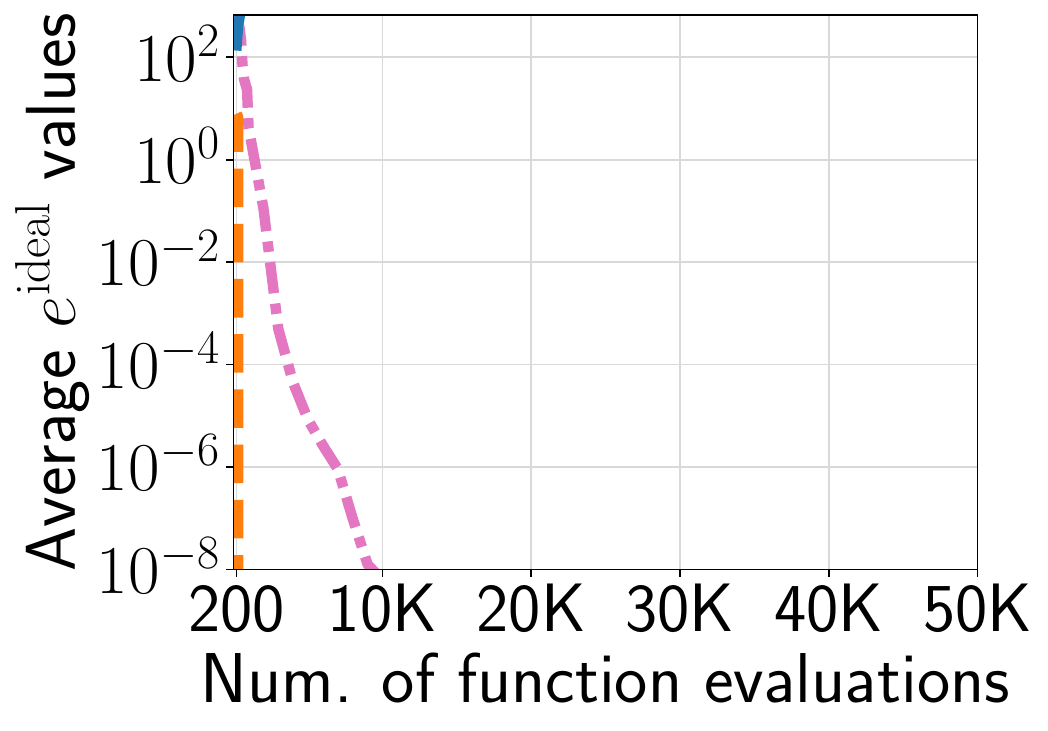}}
\\
   \subfloat[$e^{\mathrm{nadir}}$ ($m=2$)]{\includegraphics[width=0.32\textwidth]{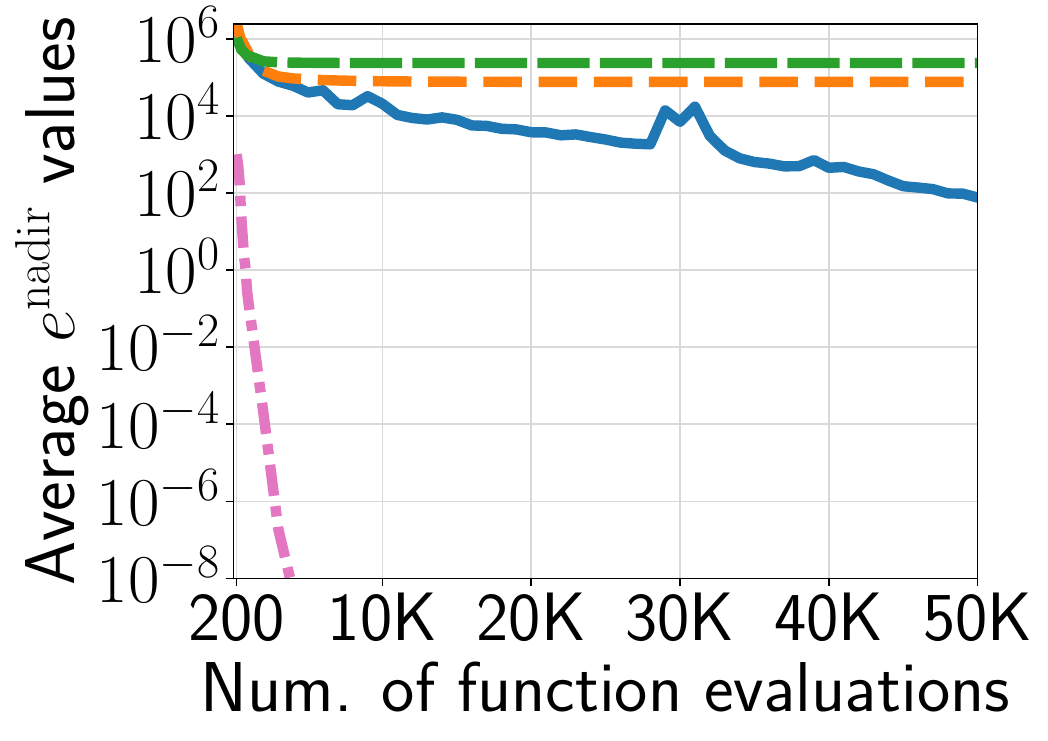}}
   \subfloat[$e^{\mathrm{nadir}}$ ($m=4$)]{\includegraphics[width=0.32\textwidth]{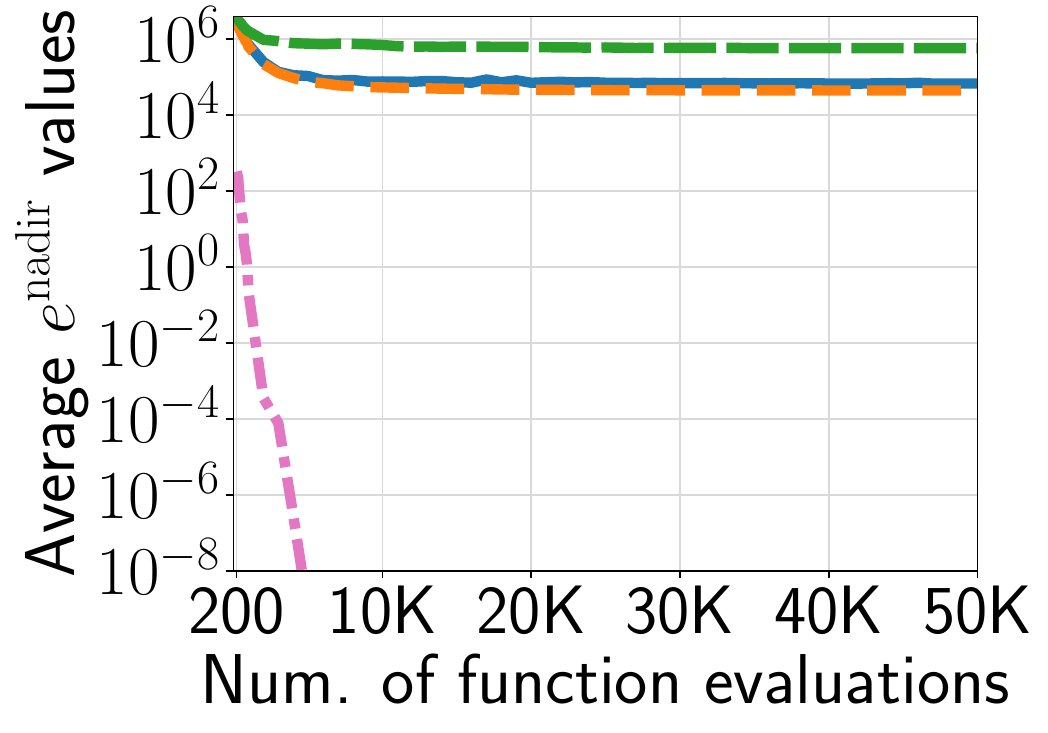}}
   \subfloat[$e^{\mathrm{nadir}}$ ($m=6$)]{\includegraphics[width=0.32\textwidth]{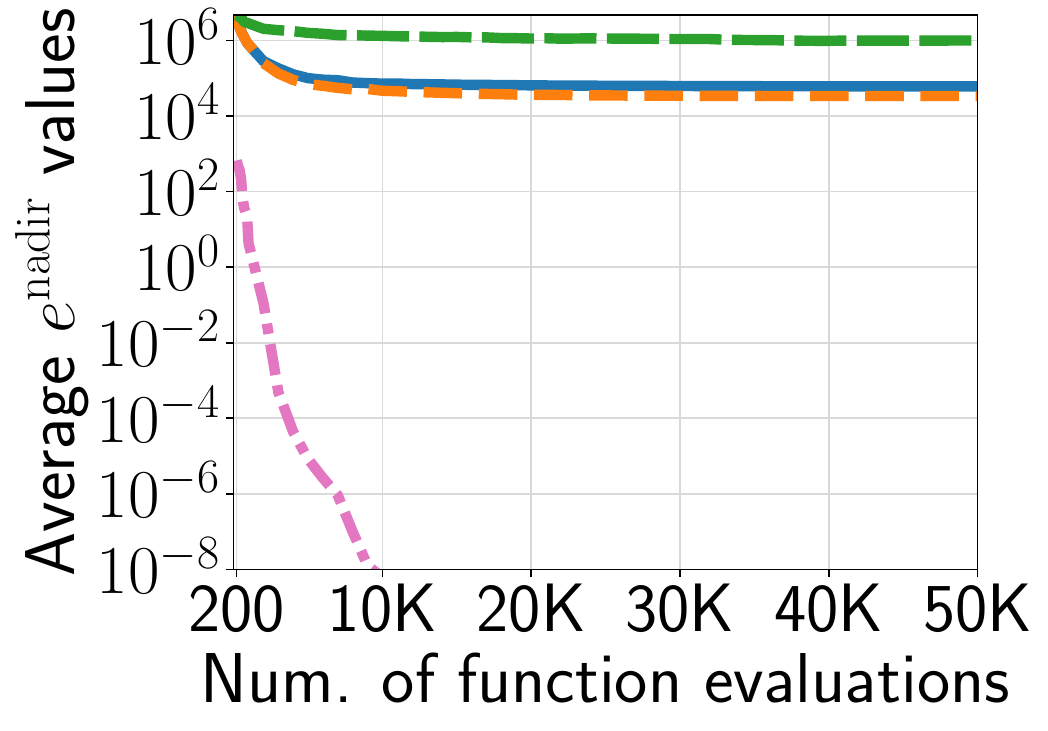}}
\\
   \subfloat[ORE ($m=2$)]{\includegraphics[width=0.32\textwidth]{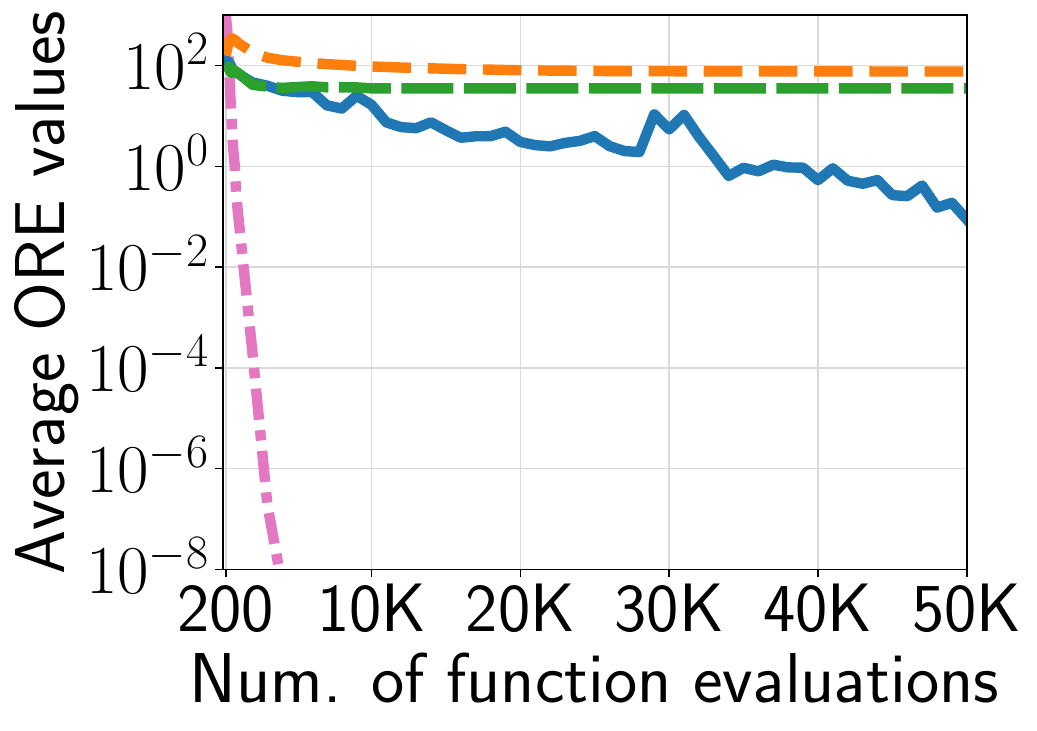}}
   \subfloat[ORE ($m=4$)]{\includegraphics[width=0.32\textwidth]{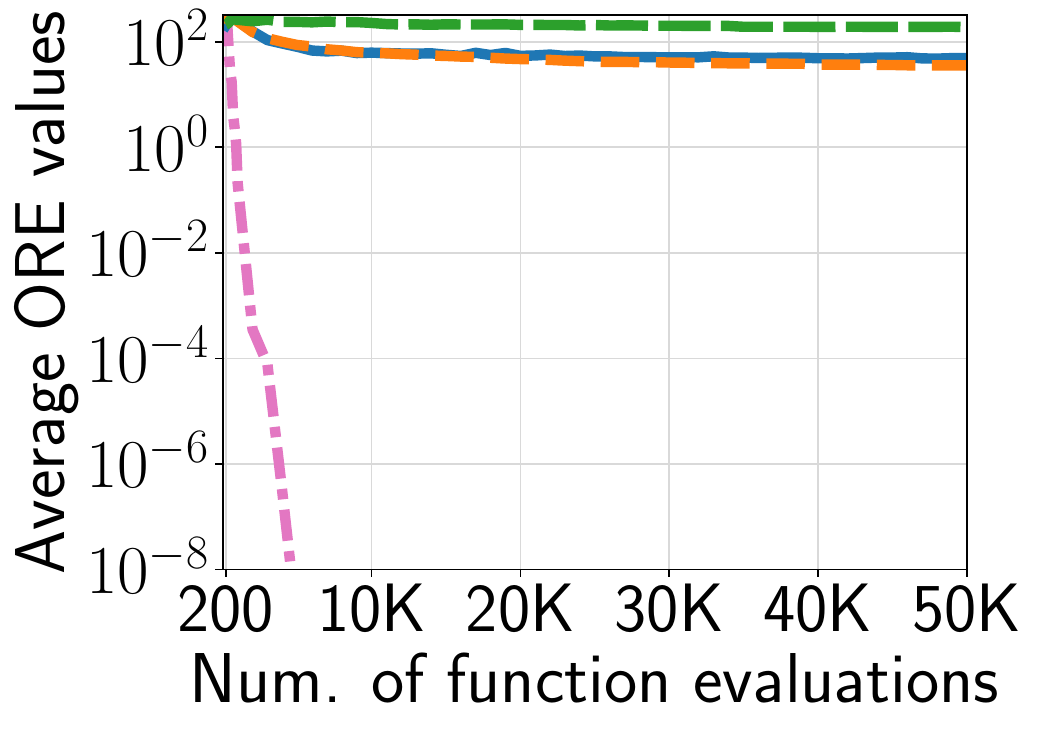}}
   \subfloat[ORE ($m=6$)]{\includegraphics[width=0.32\textwidth]{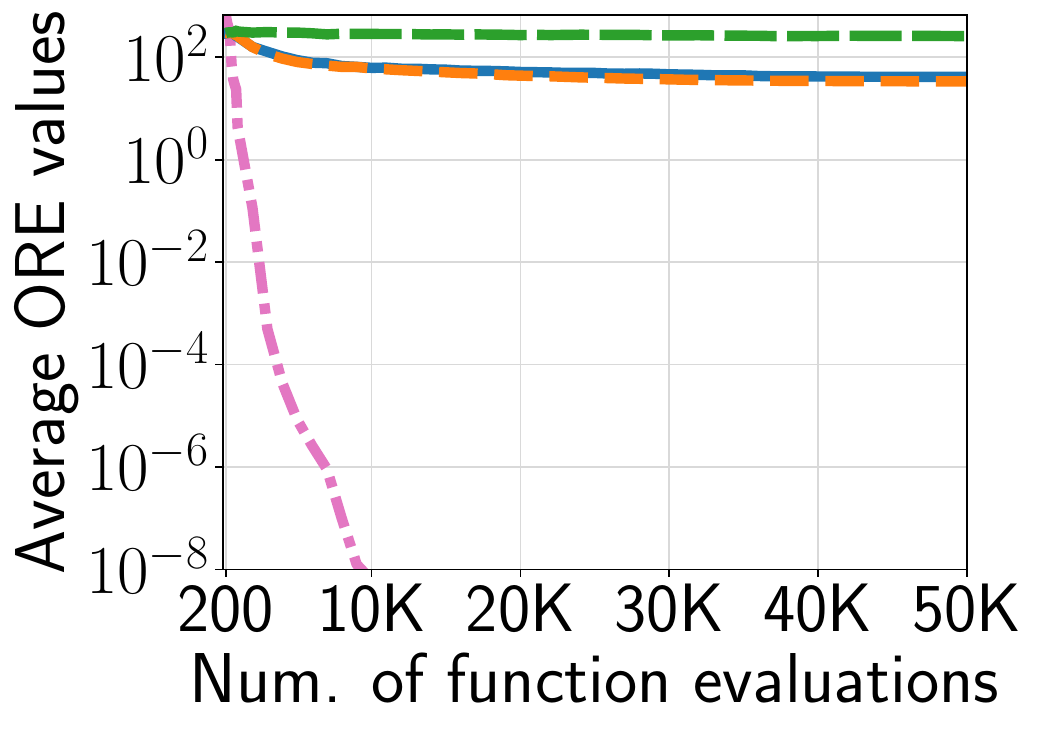}}
\\
\caption{Average $e^{\mathrm{ideal}}$, $e^{\mathrm{nadir}}$, and ORE values of the three normalization methods in MOEA/D-NUMS on DTLZ3.}
\label{supfig:3error_MOEADNUMS_DTLZ3}
\end{figure*}

\begin{figure*}[t]
\centering
  \subfloat{\includegraphics[width=0.7\textwidth]{./figs/legend/legend_3.pdf}}
\vspace{-3.9mm}
   \\
   \subfloat[$e^{\mathrm{ideal}}$ ($m=2$)]{\includegraphics[width=0.32\textwidth]{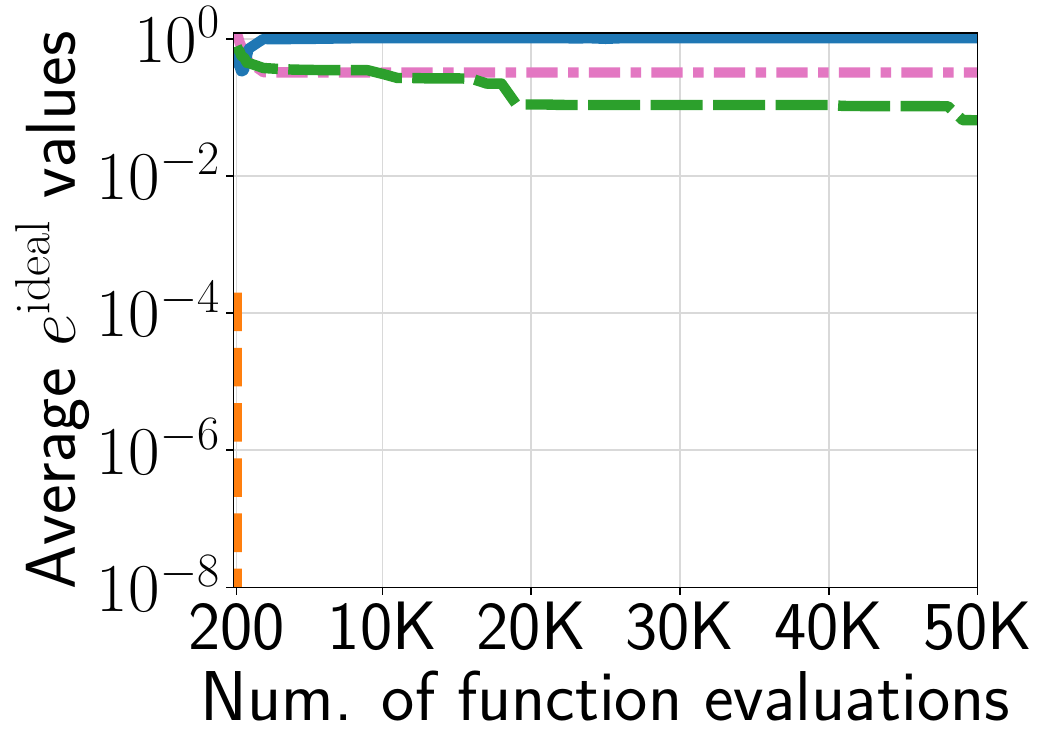}}
   \subfloat[$e^{\mathrm{ideal}}$ ($m=4$)]{\includegraphics[width=0.32\textwidth]{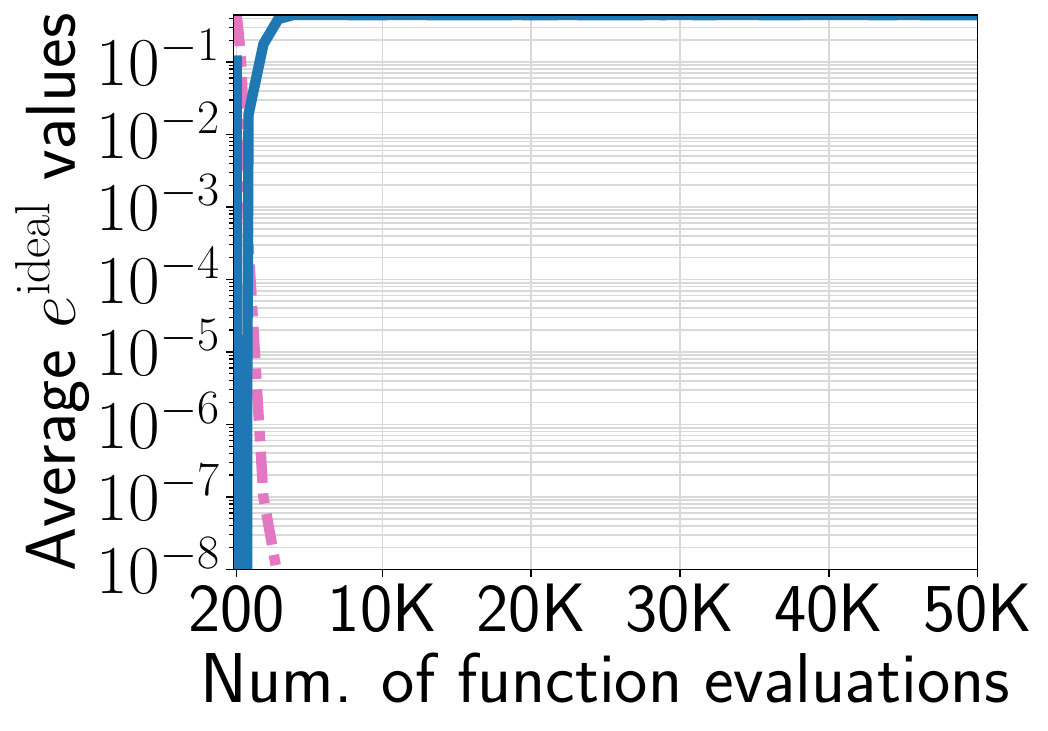}}
   \subfloat[$e^{\mathrm{ideal}}$ ($m=6$)]{\includegraphics[width=0.32\textwidth]{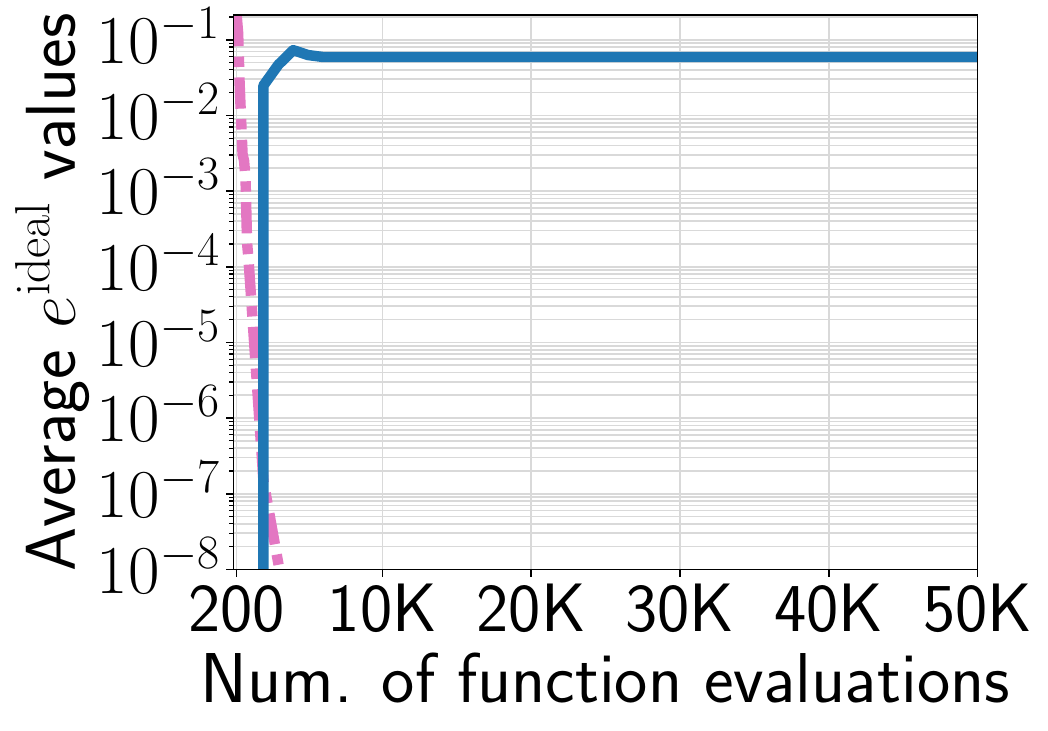}}
\\
   \subfloat[$e^{\mathrm{nadir}}$ ($m=2$)]{\includegraphics[width=0.32\textwidth]{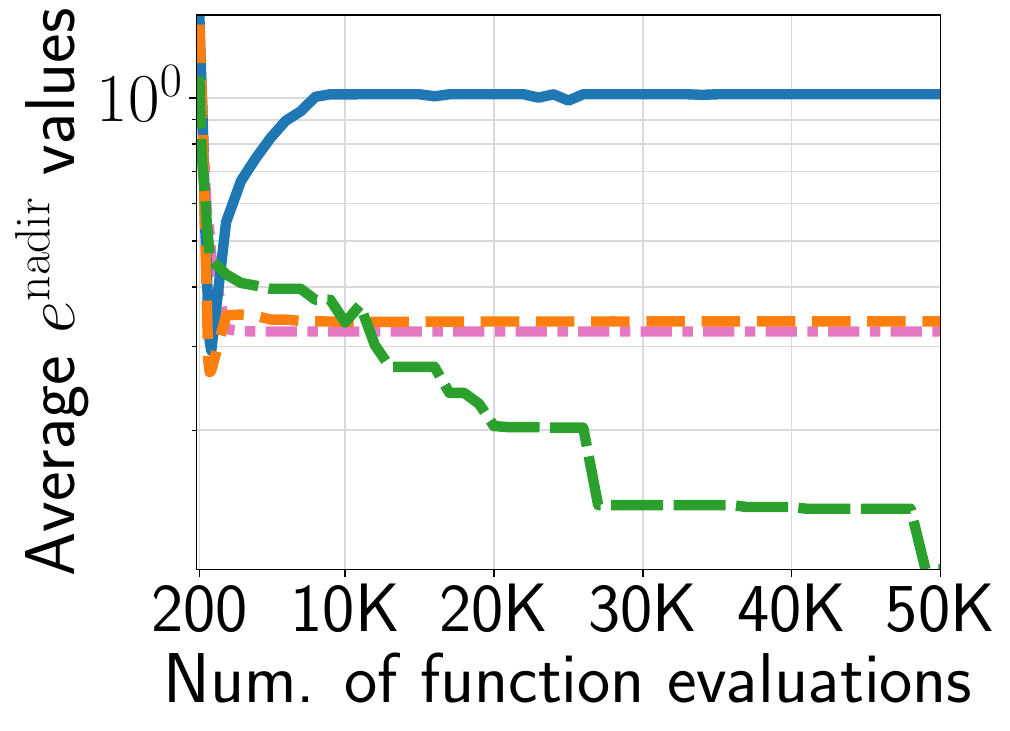}}
   \subfloat[$e^{\mathrm{nadir}}$ ($m=4$)]{\includegraphics[width=0.32\textwidth]{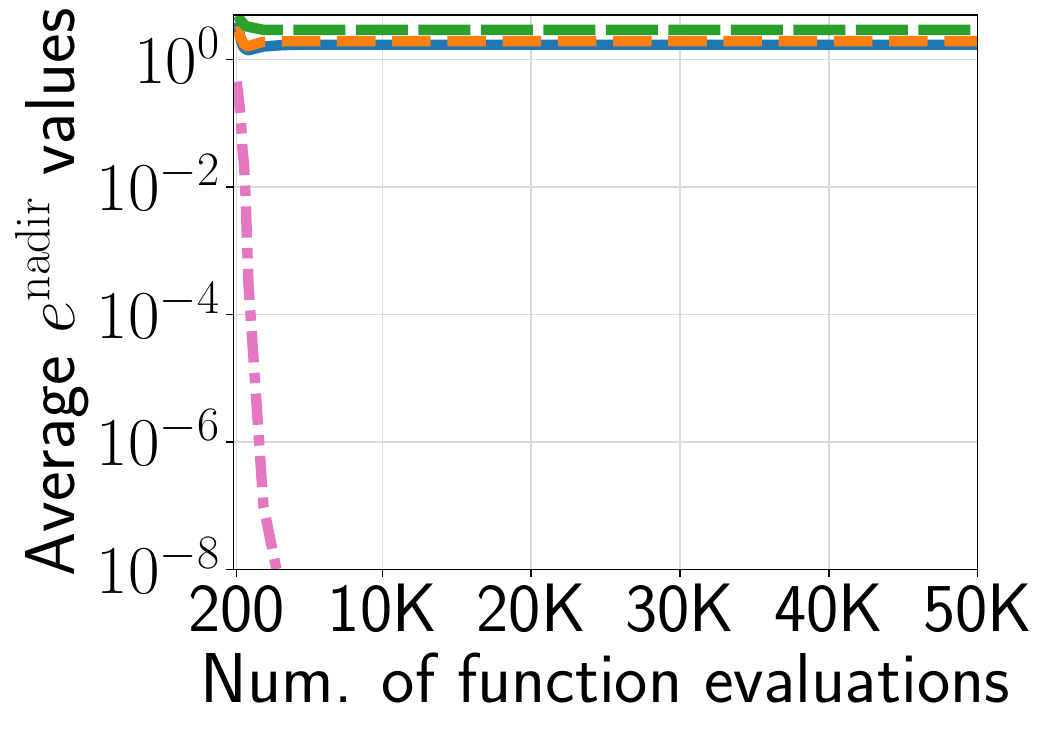}}
   \subfloat[$e^{\mathrm{nadir}}$ ($m=6$)]{\includegraphics[width=0.32\textwidth]{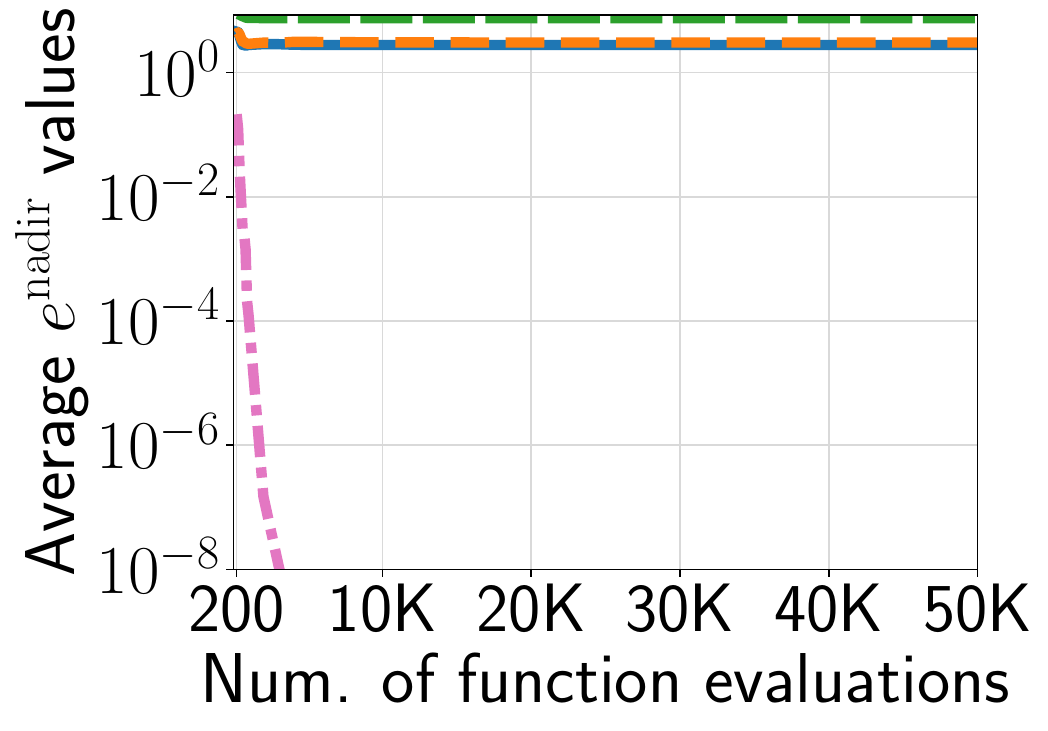}}
\\
   \subfloat[ORE ($m=2$)]{\includegraphics[width=0.32\textwidth]{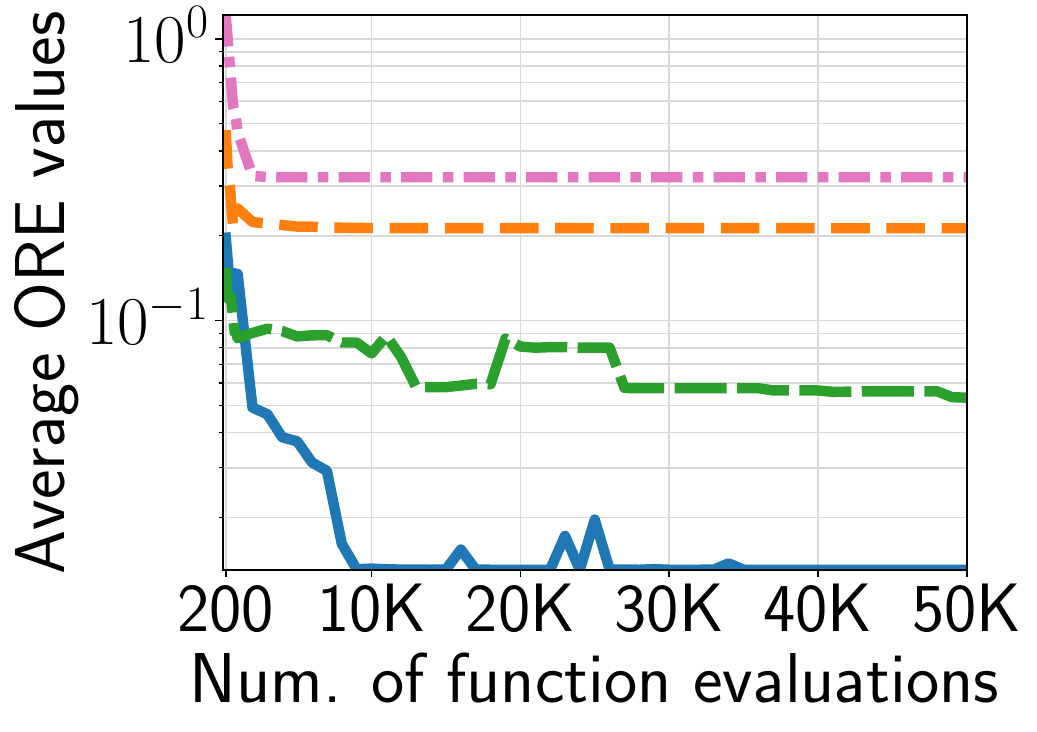}}
   \subfloat[ORE ($m=4$)]{\includegraphics[width=0.32\textwidth]{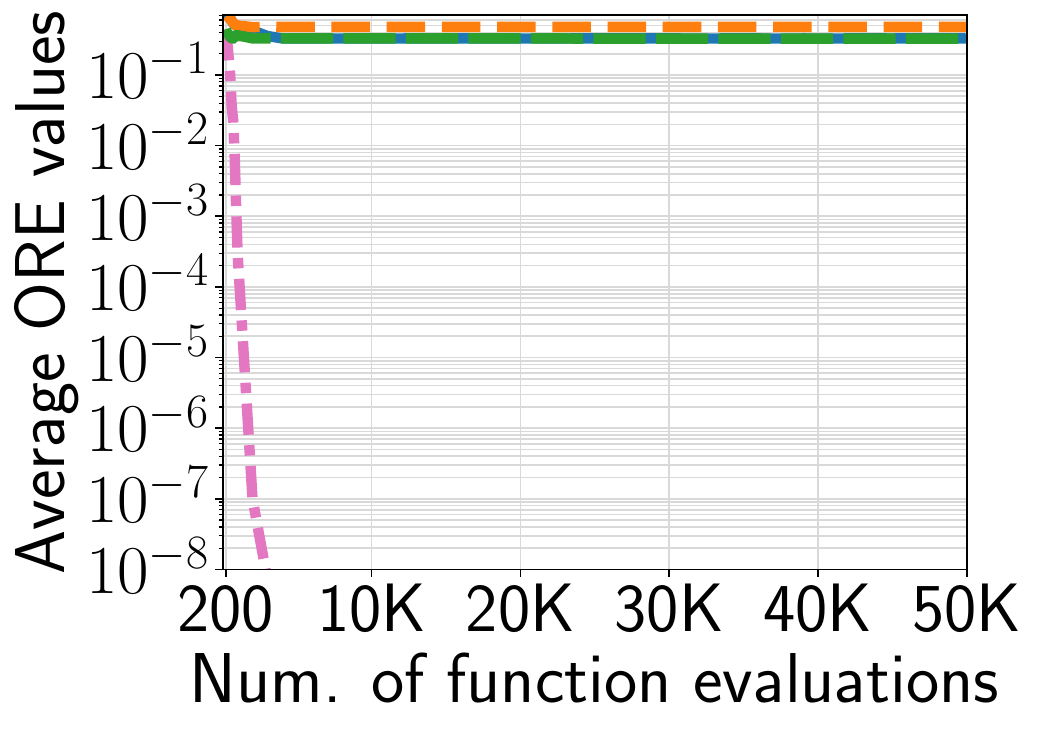}}
   \subfloat[ORE ($m=6$)]{\includegraphics[width=0.32\textwidth]{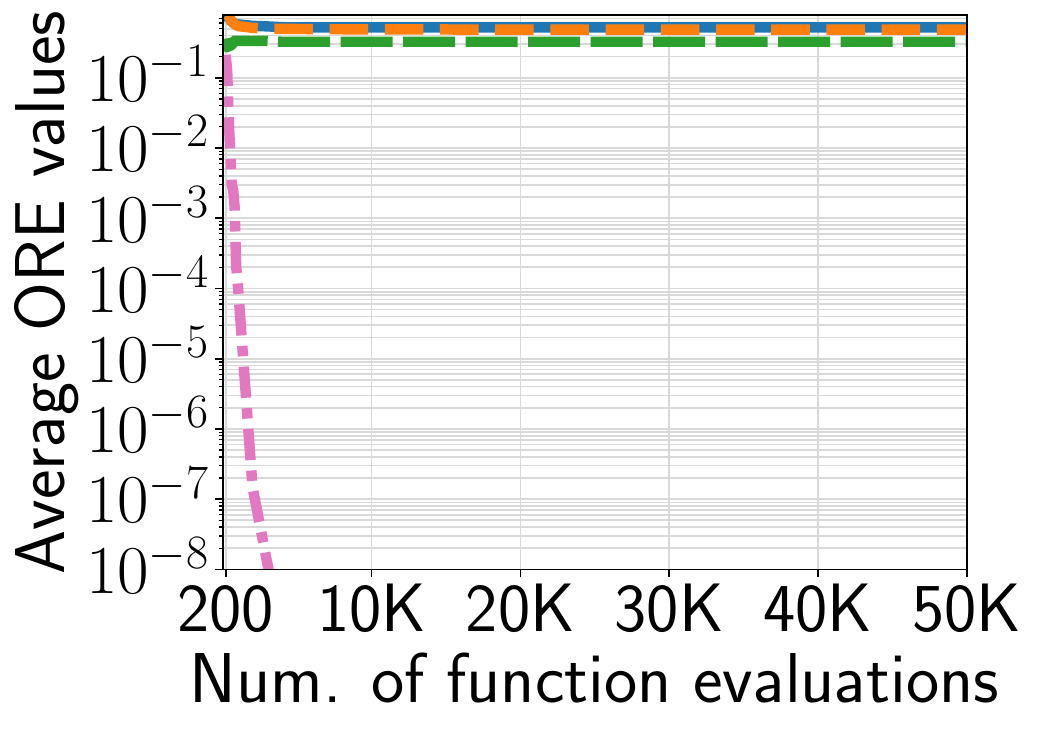}}
\\
\caption{Average $e^{\mathrm{ideal}}$, $e^{\mathrm{nadir}}$, and ORE values of the three normalization methods in MOEA/D-NUMS on DTLZ4.}
\label{supfig:3error_MOEADNUMS_DTLZ4}
\end{figure*}

\begin{figure*}[t]
\centering
  \subfloat{\includegraphics[width=0.7\textwidth]{./figs/legend/legend_3.pdf}}
\vspace{-3.9mm}
   \\
   \subfloat[$e^{\mathrm{ideal}}$ ($m=2$)]{\includegraphics[width=0.32\textwidth]{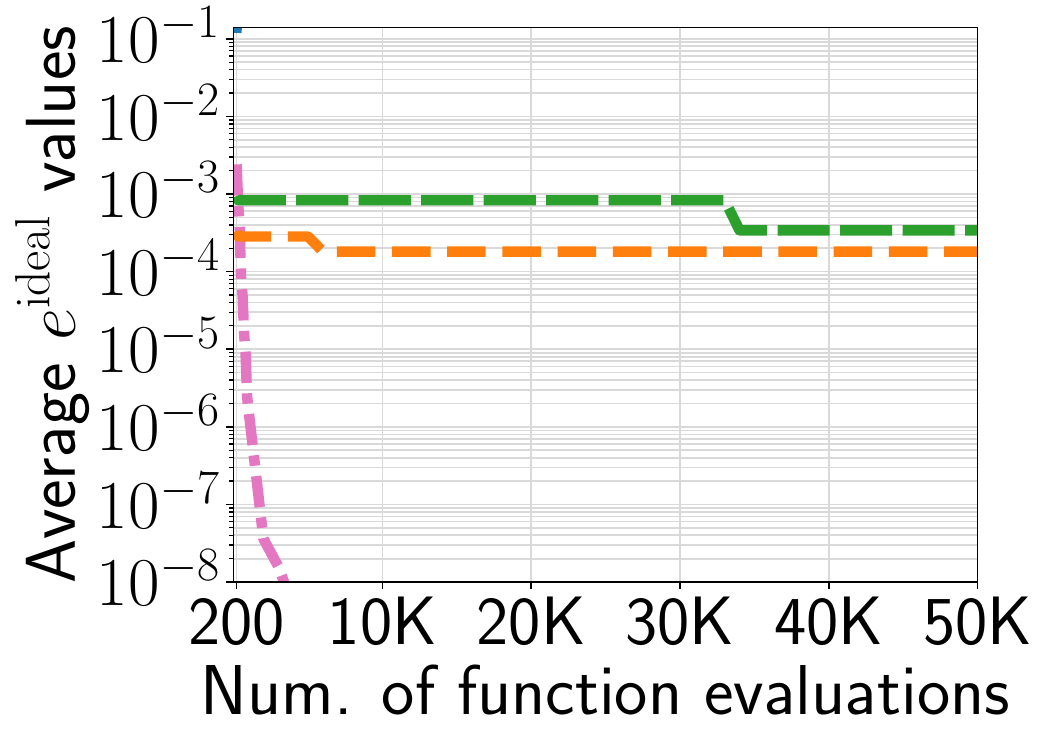}}
   \subfloat[$e^{\mathrm{ideal}}$ ($m=4$)]{\includegraphics[width=0.32\textwidth]{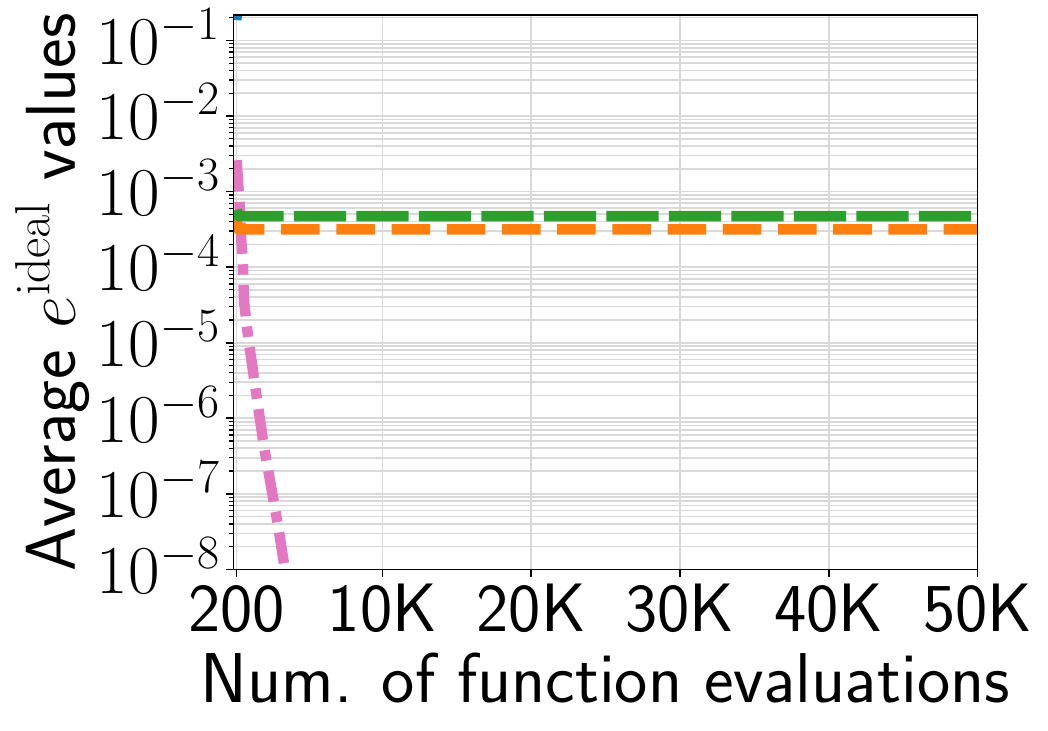}}
   \subfloat[$e^{\mathrm{ideal}}$ ($m=6$)]{\includegraphics[width=0.32\textwidth]{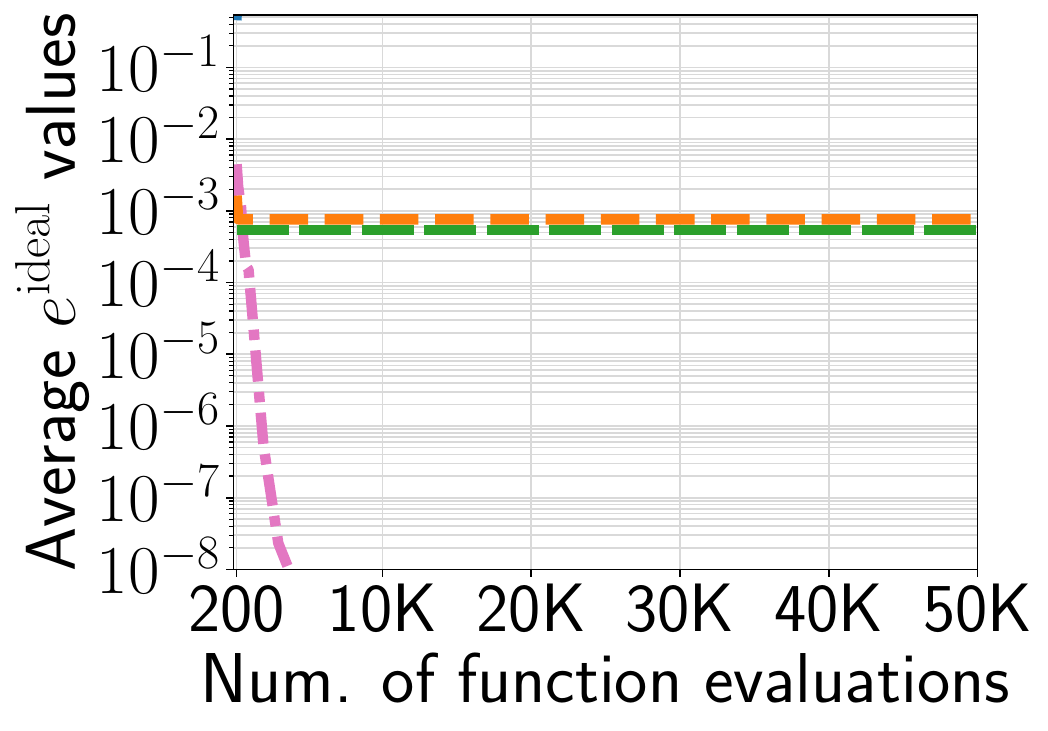}}
\\
   \subfloat[$e^{\mathrm{nadir}}$ ($m=2$)]{\includegraphics[width=0.32\textwidth]{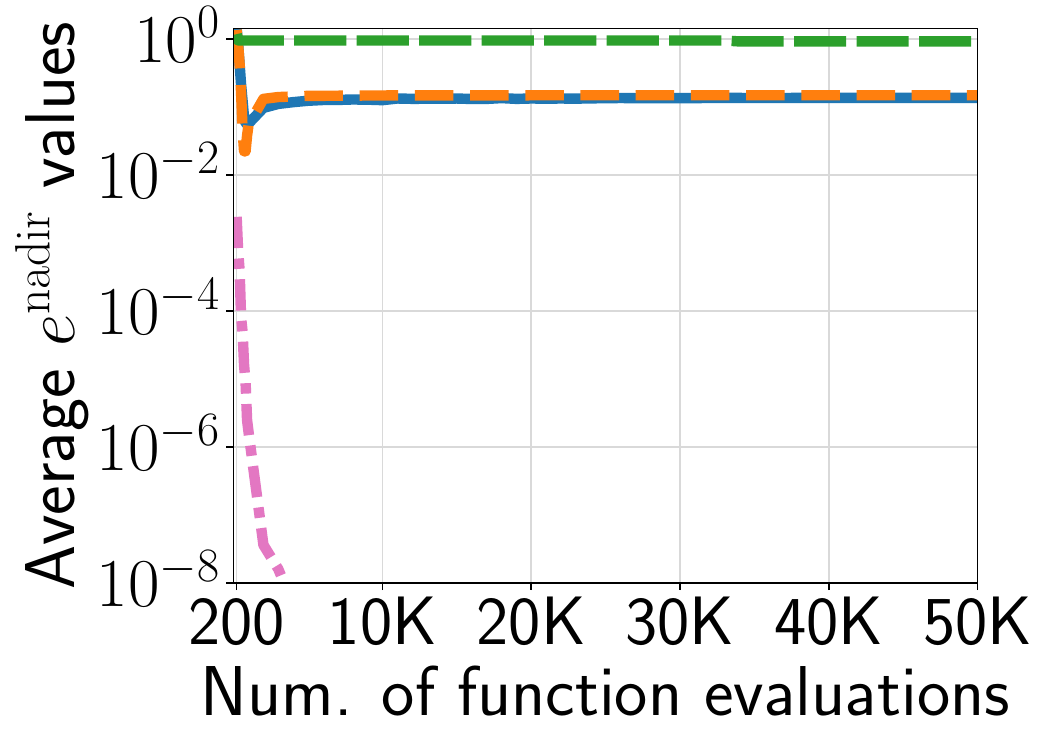}}
   \subfloat[$e^{\mathrm{nadir}}$ ($m=4$)]{\includegraphics[width=0.32\textwidth]{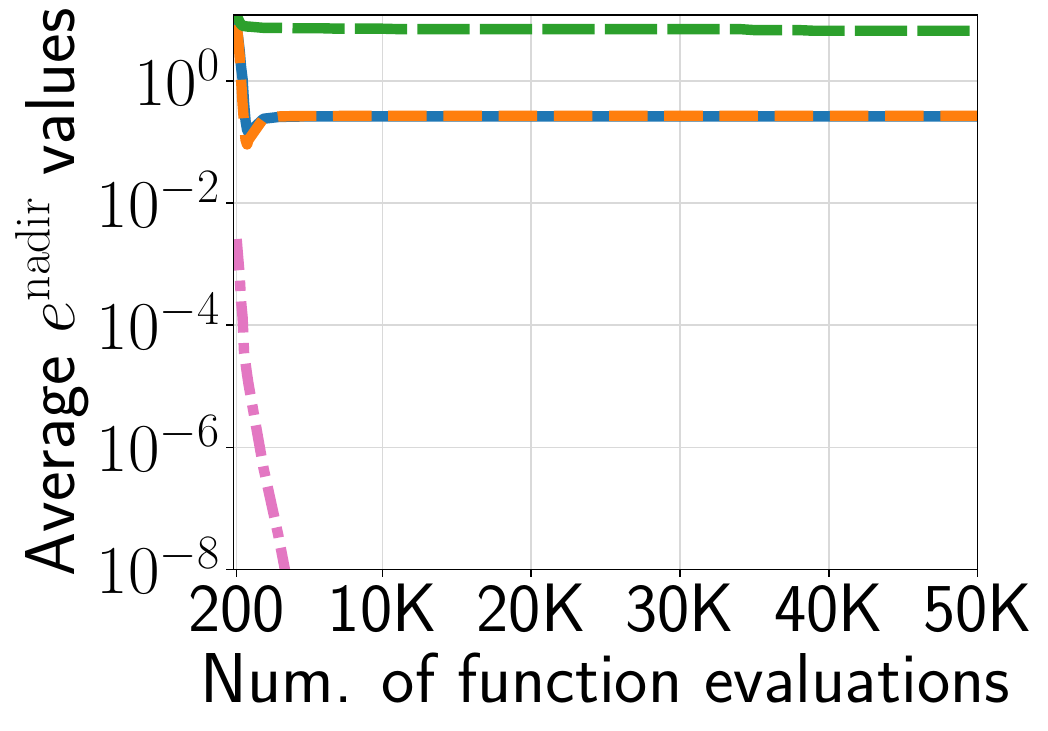}}
   \subfloat[$e^{\mathrm{nadir}}$ ($m=6$)]{\includegraphics[width=0.32\textwidth]{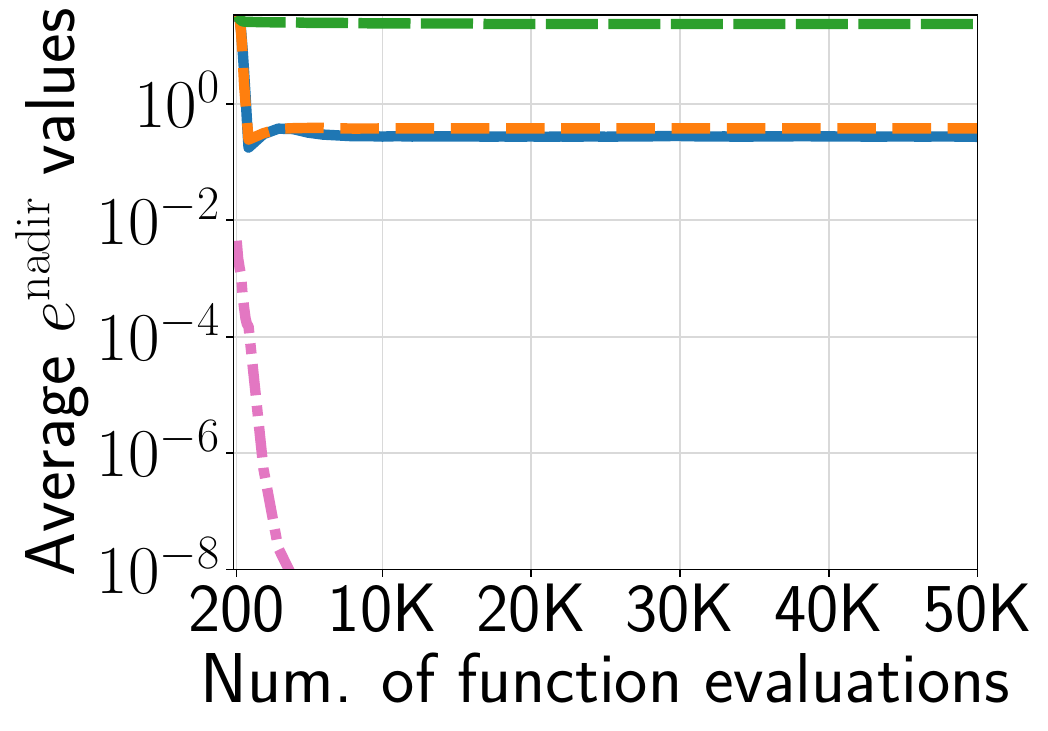}}
\\
   \subfloat[ORE ($m=2$)]{\includegraphics[width=0.32\textwidth]{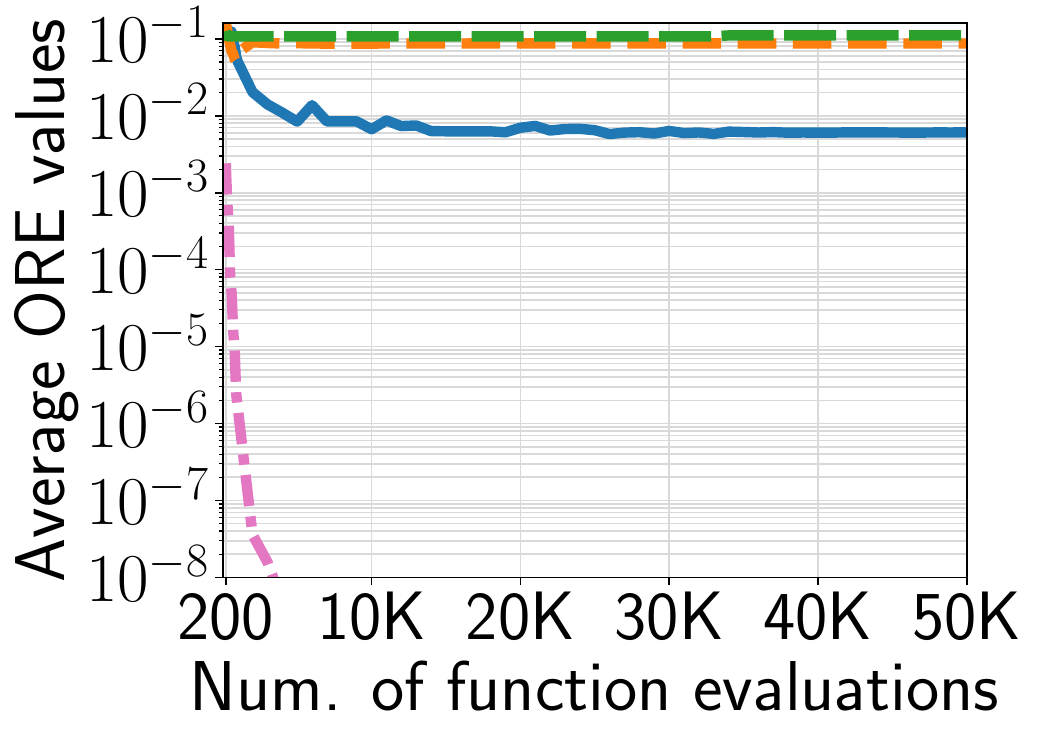}}
   \subfloat[ORE ($m=4$)]{\includegraphics[width=0.32\textwidth]{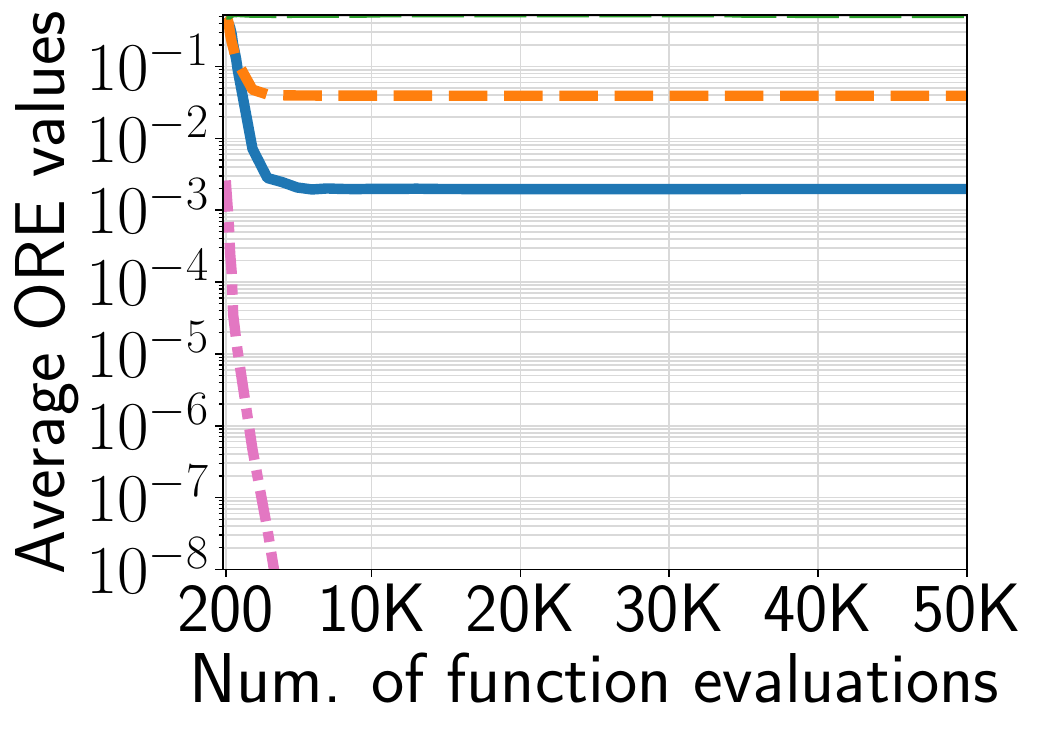}}
   \subfloat[ORE ($m=6$)]{\includegraphics[width=0.32\textwidth]{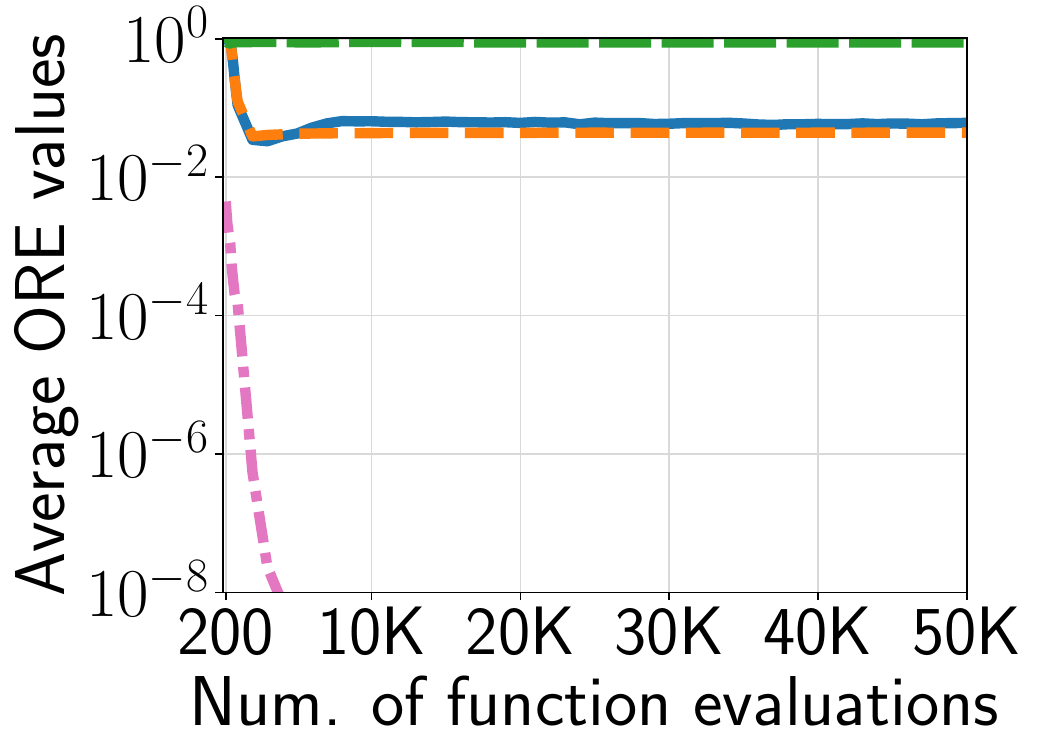}}
\\
\caption{Average $e^{\mathrm{ideal}}$, $e^{\mathrm{nadir}}$, and ORE values of the three normalization methods in MOEA/D-NUMS on DTLZ5.}
\label{supfig:3error_MOEADNUMS_DTLZ5}
\end{figure*}

\begin{figure*}[t]
\centering
  \subfloat{\includegraphics[width=0.7\textwidth]{./figs/legend/legend_3.pdf}}
\vspace{-3.9mm}
   \\
   \subfloat[$e^{\mathrm{ideal}}$ ($m=2$)]{\includegraphics[width=0.32\textwidth]{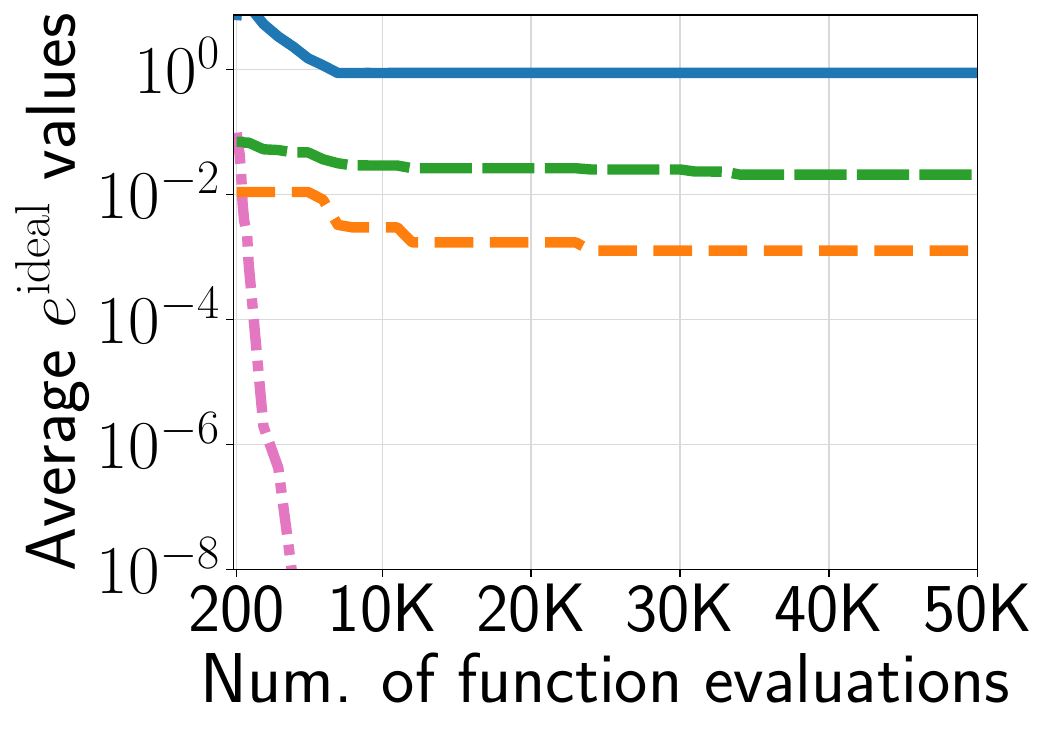}}
   \subfloat[$e^{\mathrm{ideal}}$ ($m=4$)]{\includegraphics[width=0.32\textwidth]{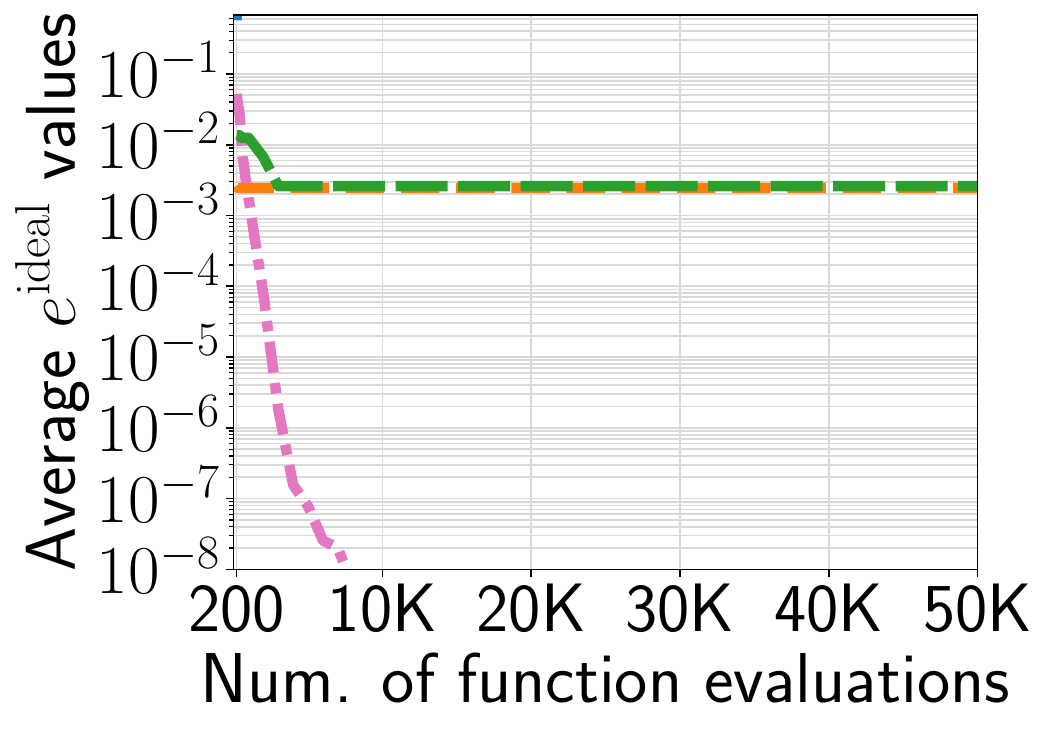}}
   \subfloat[$e^{\mathrm{ideal}}$ ($m=6$)]{\includegraphics[width=0.32\textwidth]{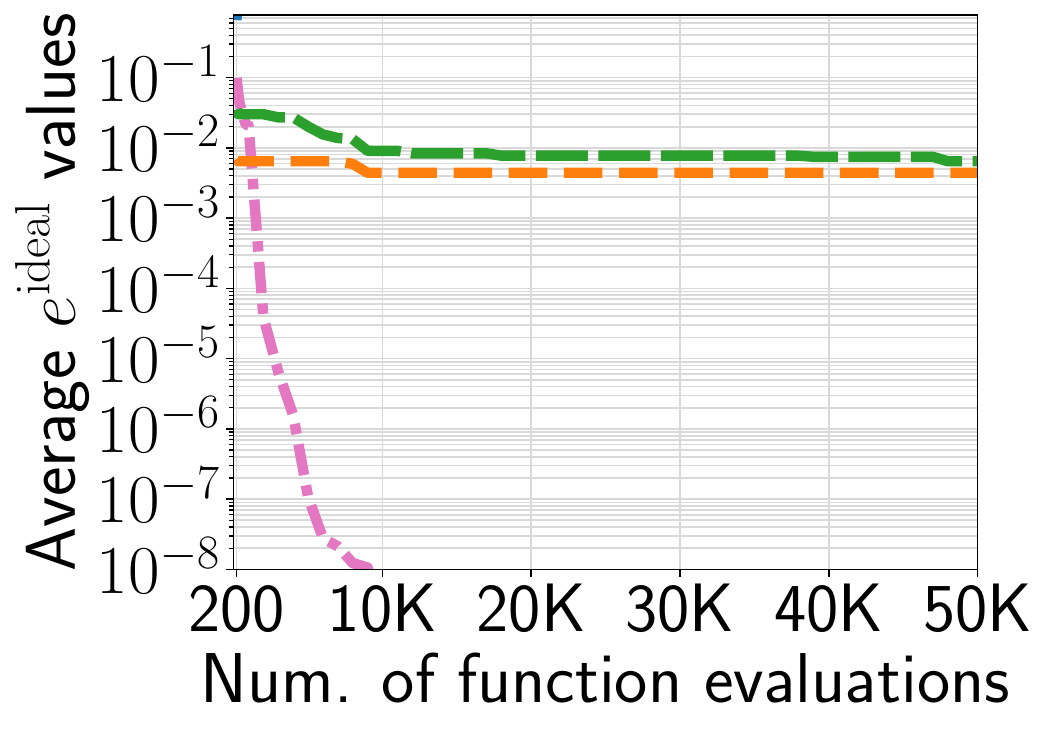}}
\\
   \subfloat[$e^{\mathrm{nadir}}$ ($m=2$)]{\includegraphics[width=0.32\textwidth]{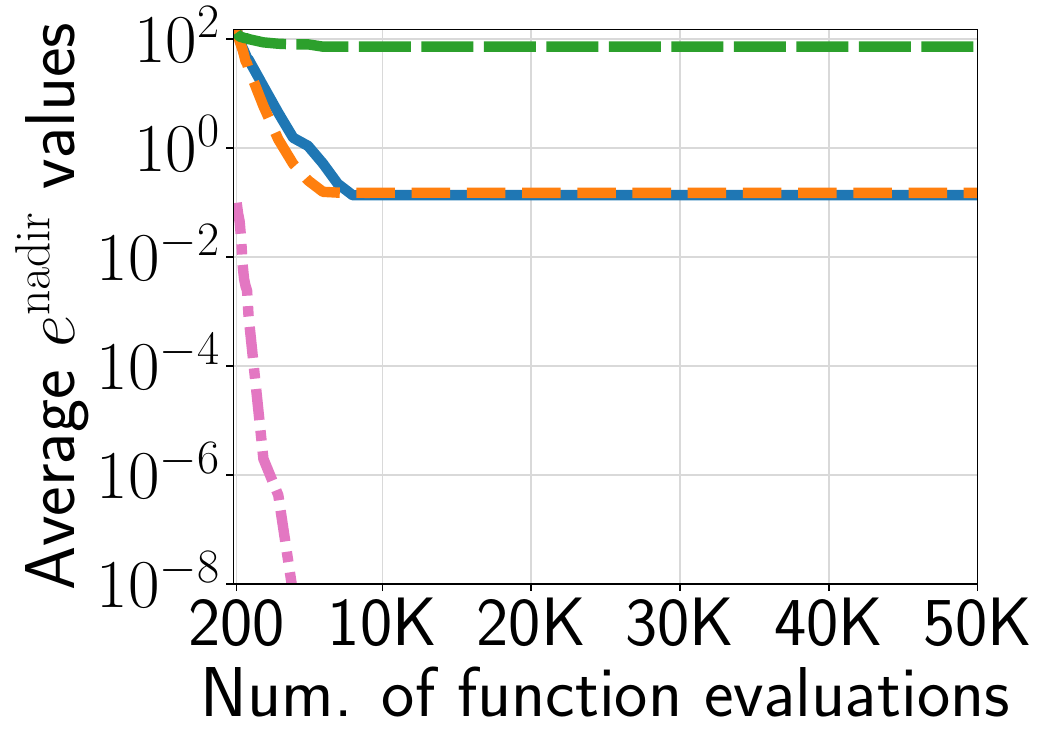}}
   \subfloat[$e^{\mathrm{nadir}}$ ($m=4$)]{\includegraphics[width=0.32\textwidth]{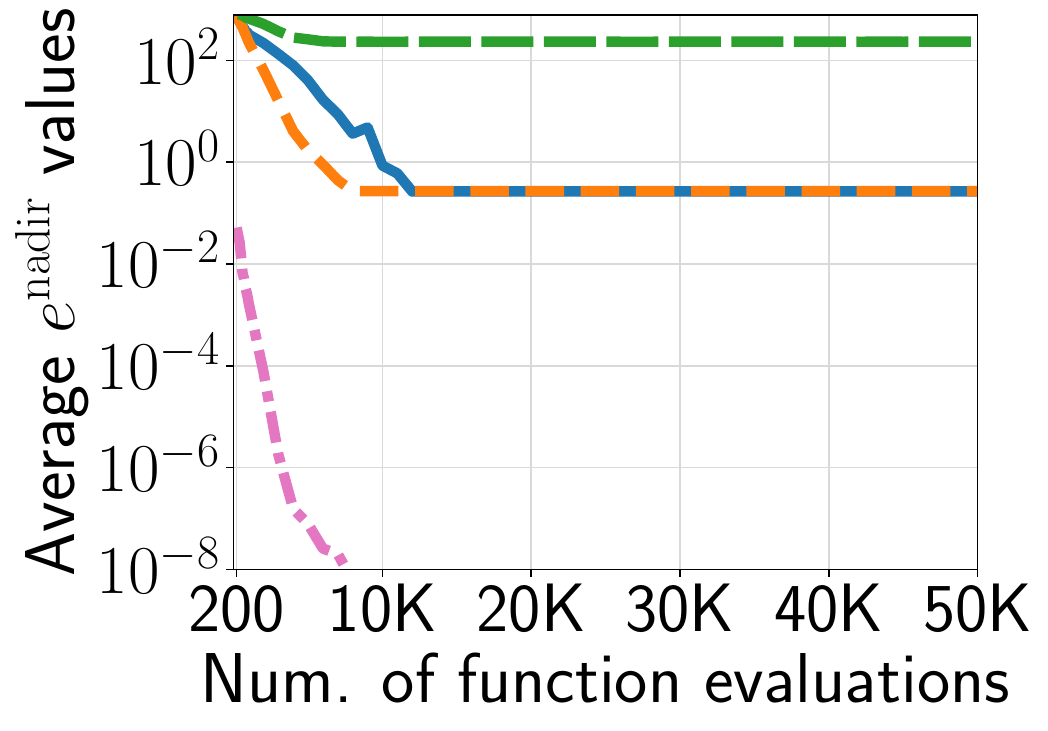}}
   \subfloat[$e^{\mathrm{nadir}}$ ($m=6$)]{\includegraphics[width=0.32\textwidth]{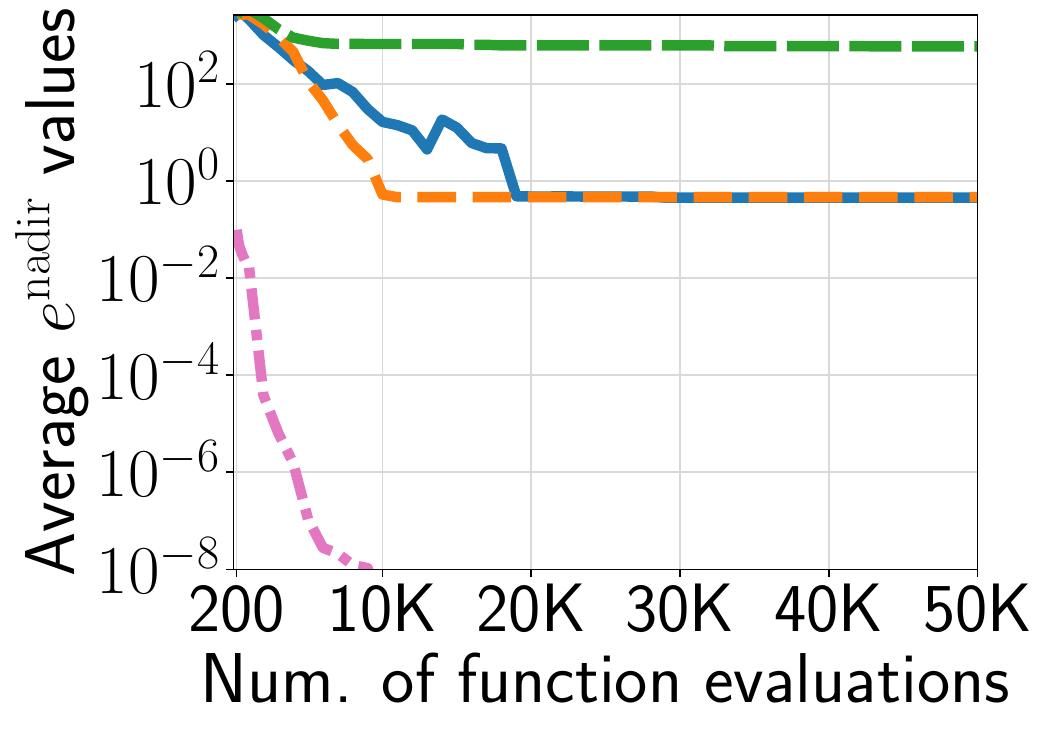}}
\\
   \subfloat[ORE ($m=2$)]{\includegraphics[width=0.32\textwidth]{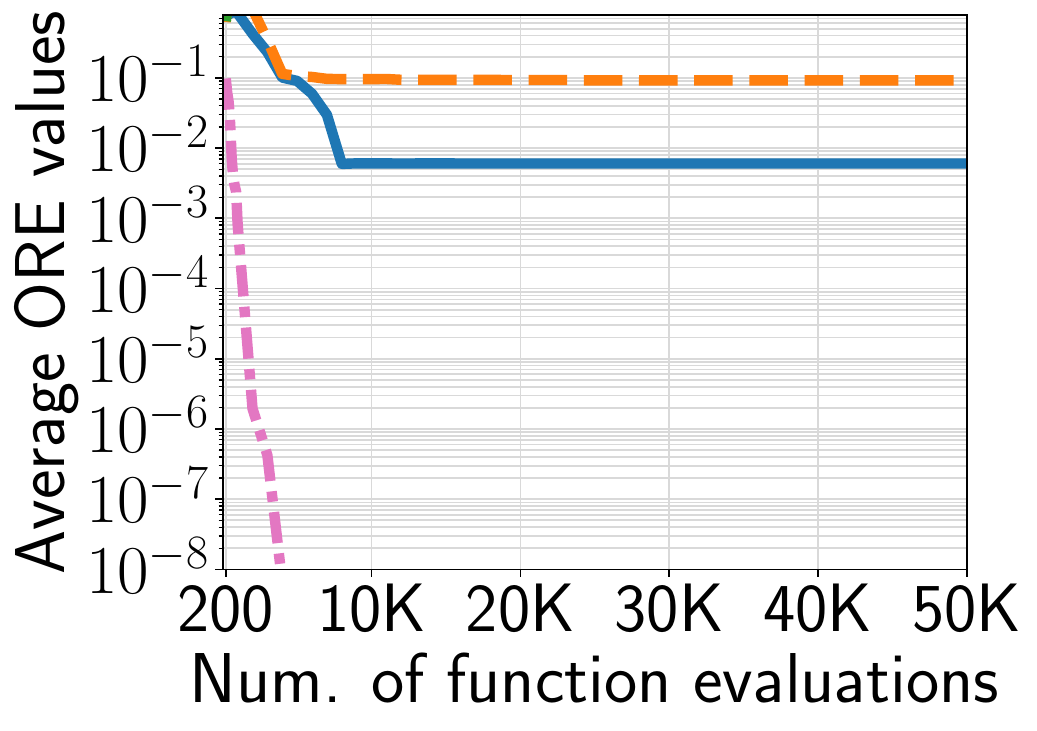}}
   \subfloat[ORE ($m=4$)]{\includegraphics[width=0.32\textwidth]{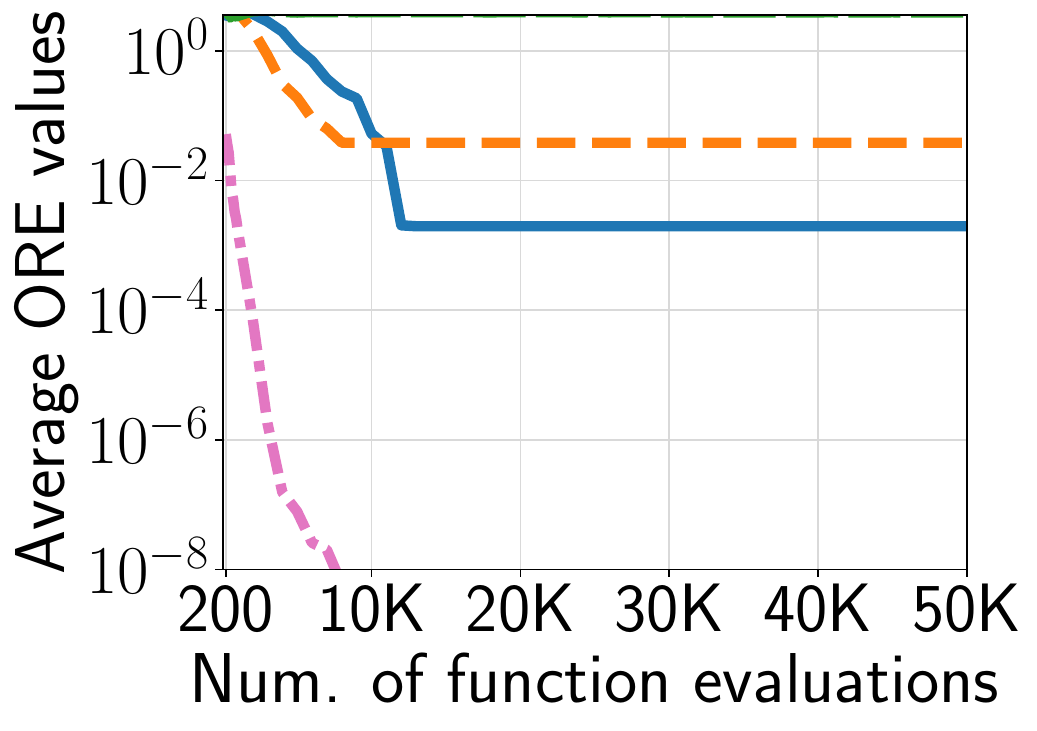}}
   \subfloat[ORE ($m=6$)]{\includegraphics[width=0.32\textwidth]{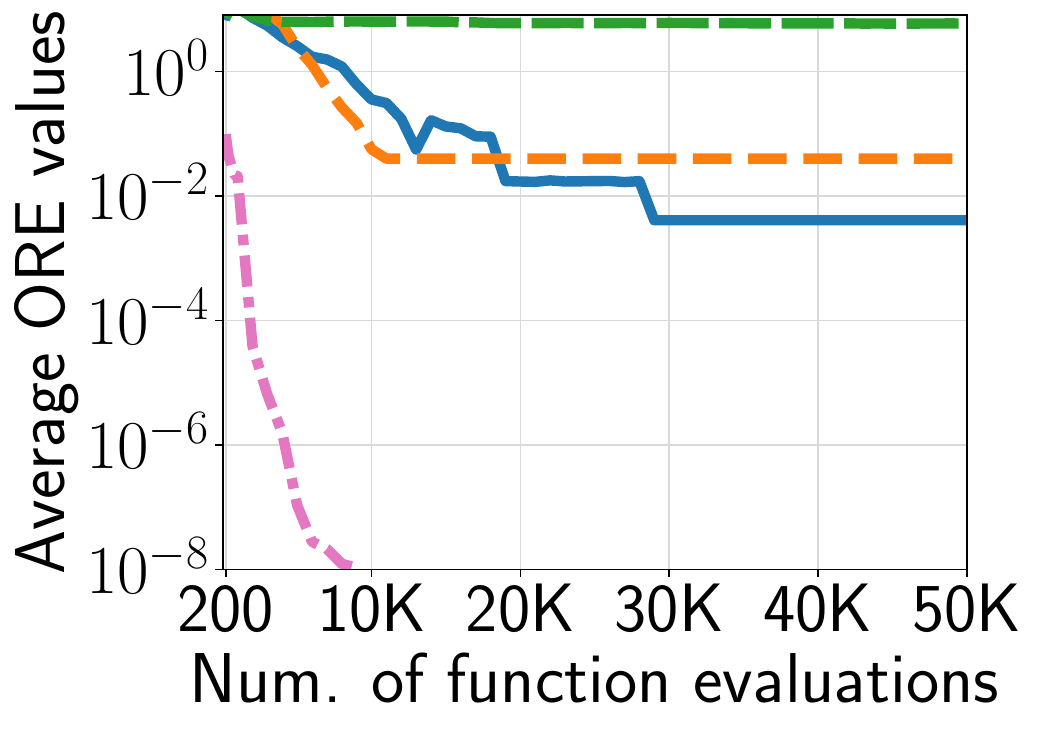}}
\\
\caption{Average $e^{\mathrm{ideal}}$, $e^{\mathrm{nadir}}$, and ORE values of the three normalization methods in MOEA/D-NUMS on DTLZ6.}
\label{supfig:3error_MOEADNUMS_DTLZ6}
\end{figure*}

\begin{figure*}[t]
\centering
  \subfloat{\includegraphics[width=0.7\textwidth]{./figs/legend/legend_3.pdf}}
\vspace{-3.9mm}
   \\
   \subfloat[$e^{\mathrm{ideal}}$ ($m=2$)]{\includegraphics[width=0.32\textwidth]{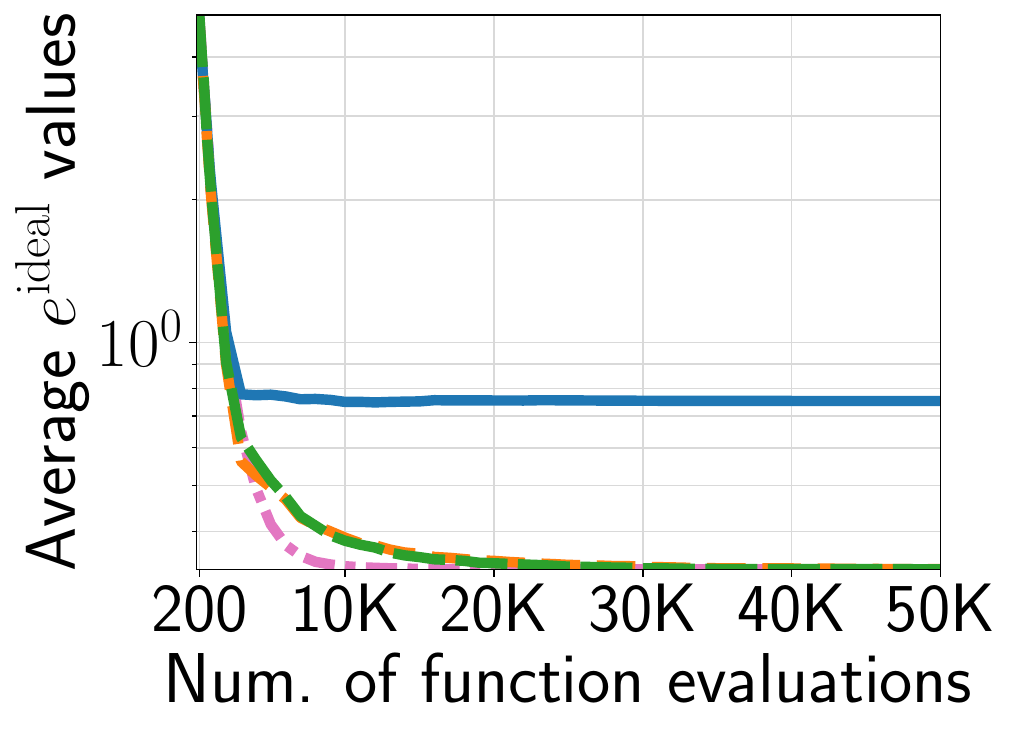}}
   \subfloat[$e^{\mathrm{ideal}}$ ($m=4$)]{\includegraphics[width=0.32\textwidth]{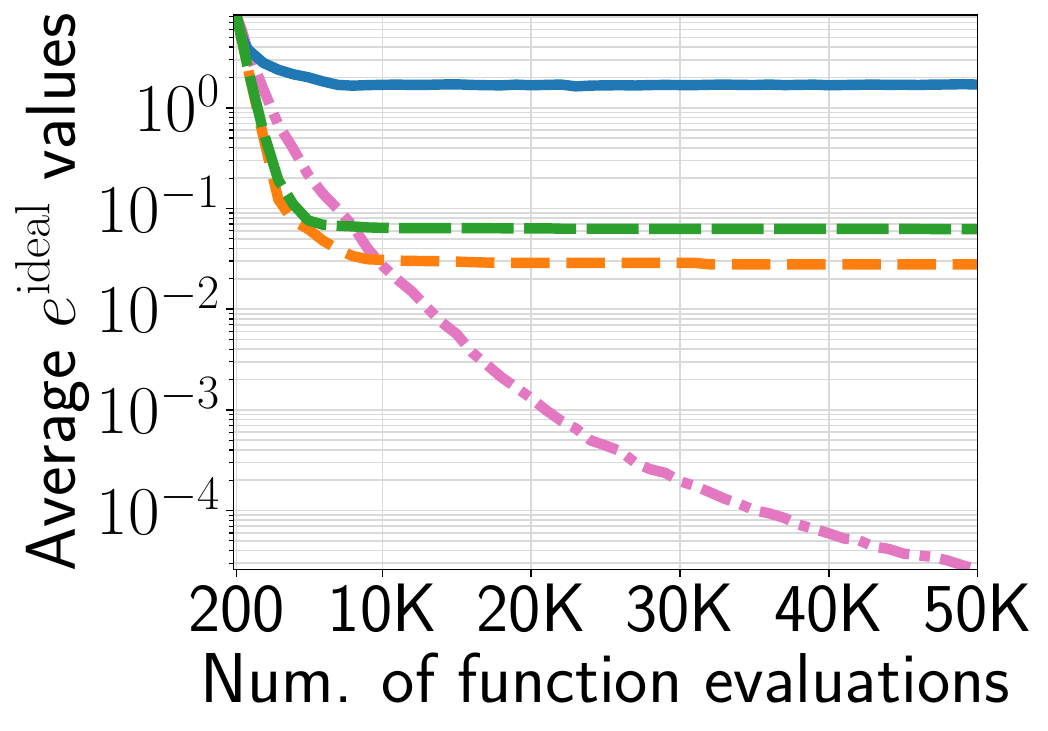}}
   \subfloat[$e^{\mathrm{ideal}}$ ($m=6$)]{\includegraphics[width=0.32\textwidth]{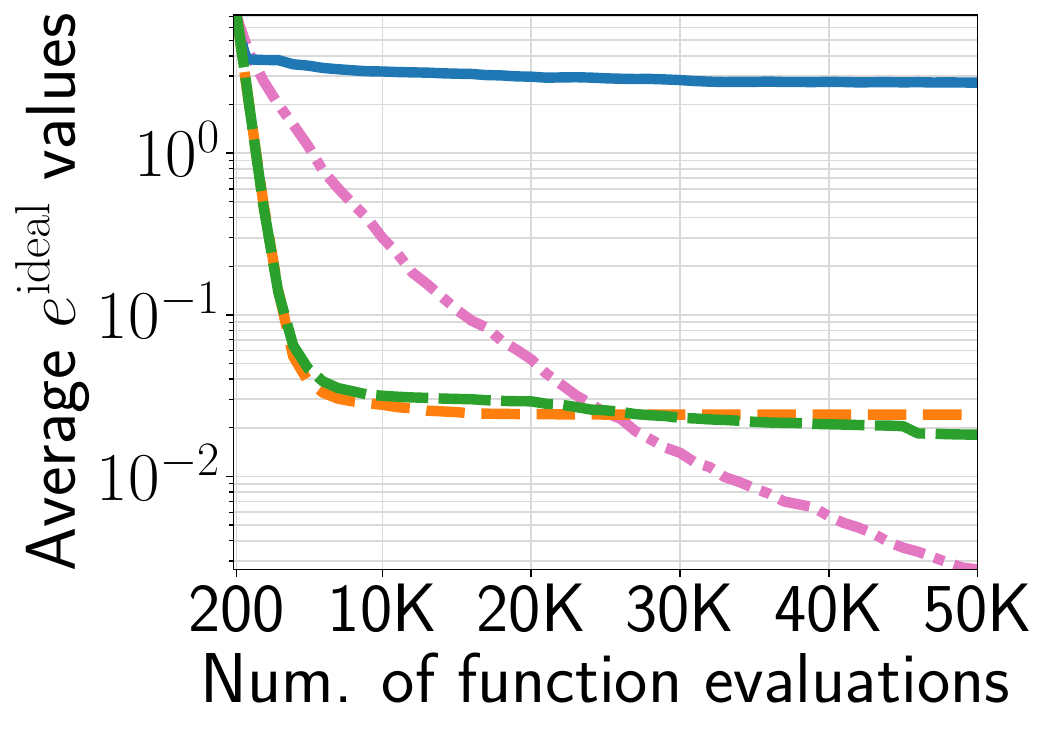}}
\\
   \subfloat[$e^{\mathrm{nadir}}$ ($m=2$)]{\includegraphics[width=0.32\textwidth]{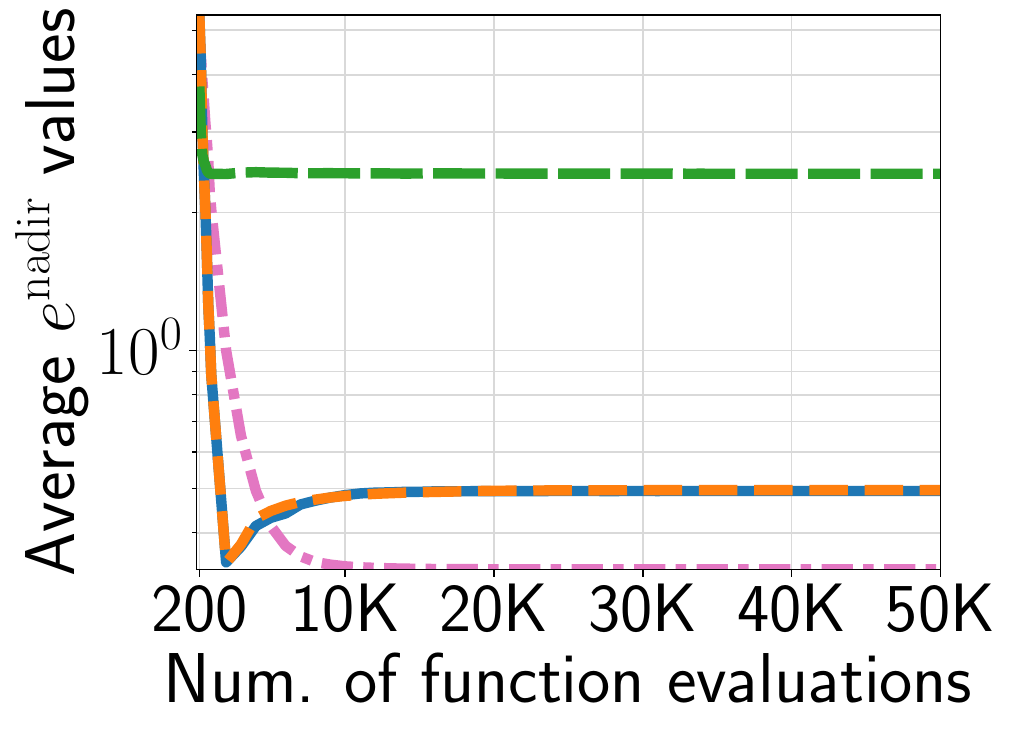}}
   \subfloat[$e^{\mathrm{nadir}}$ ($m=4$)]{\includegraphics[width=0.32\textwidth]{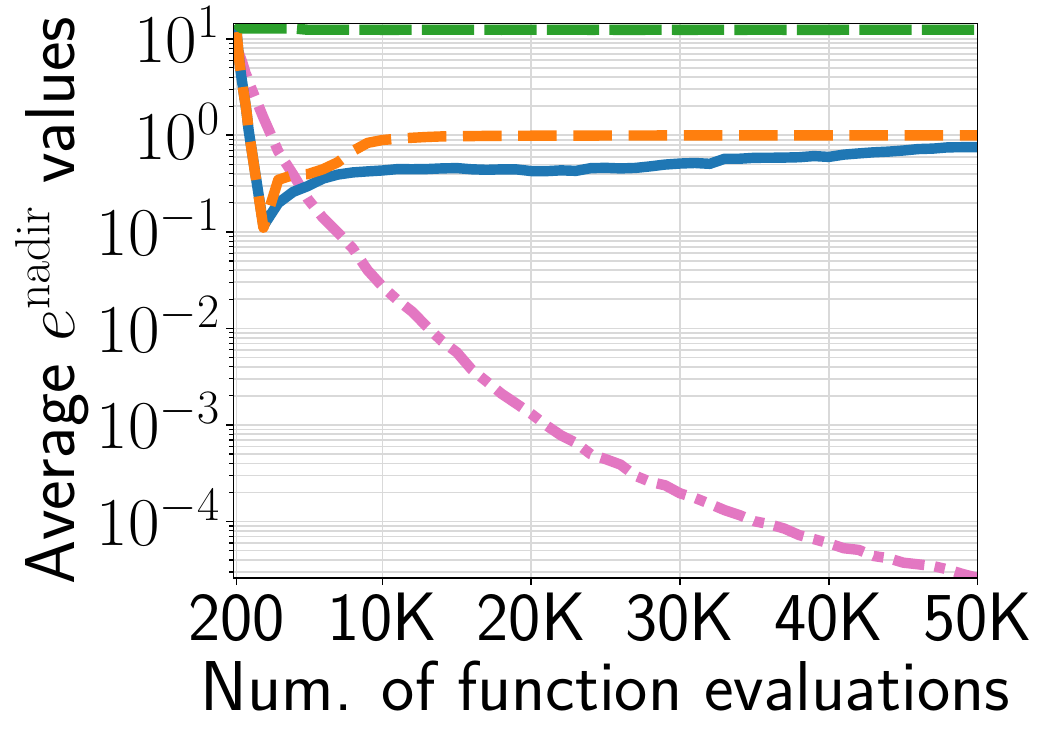}}
   \subfloat[$e^{\mathrm{nadir}}$ ($m=6$)]{\includegraphics[width=0.32\textwidth]{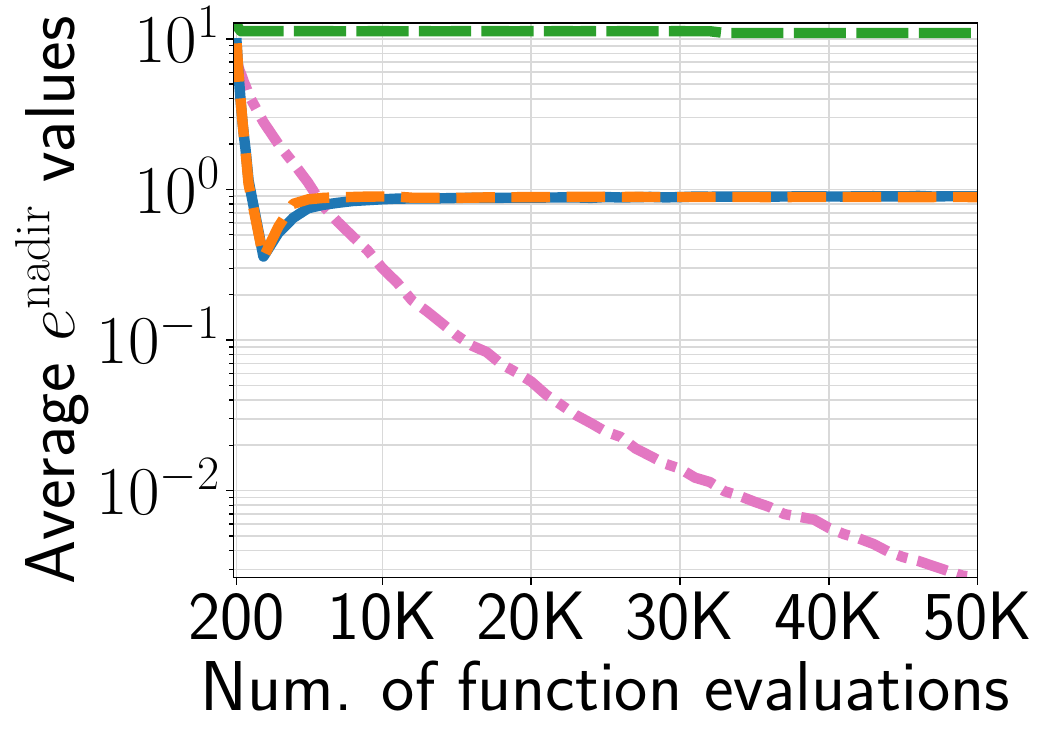}}
\\
   \subfloat[ORE ($m=2$)]{\includegraphics[width=0.32\textwidth]{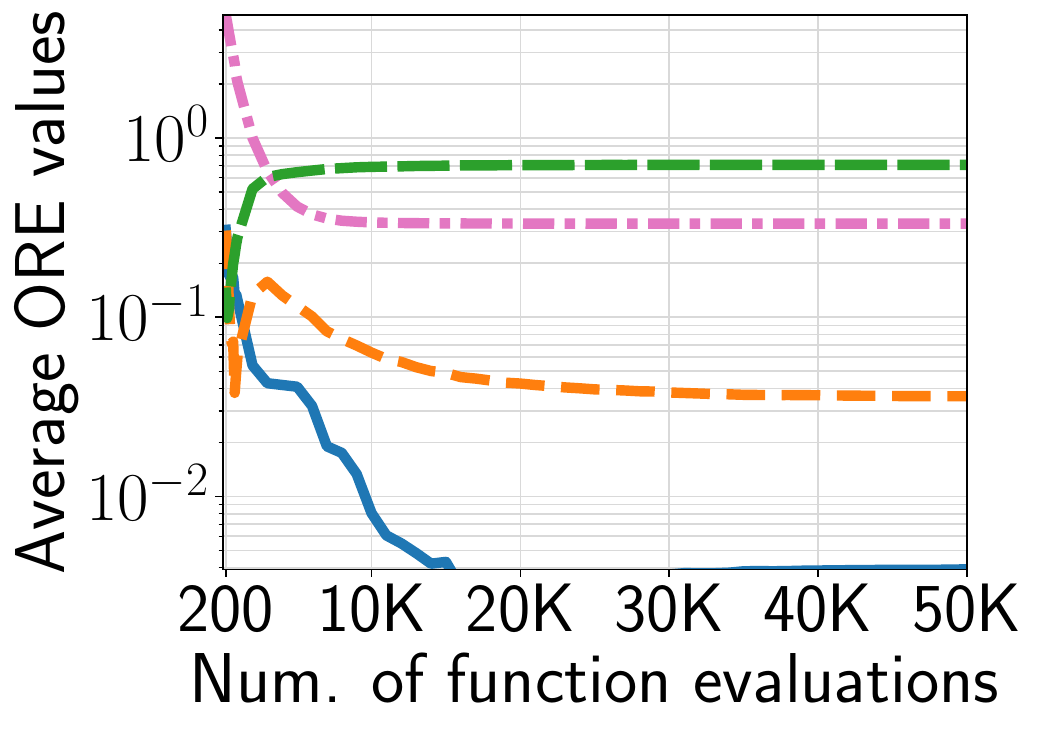}}
   \subfloat[ORE ($m=4$)]{\includegraphics[width=0.32\textwidth]{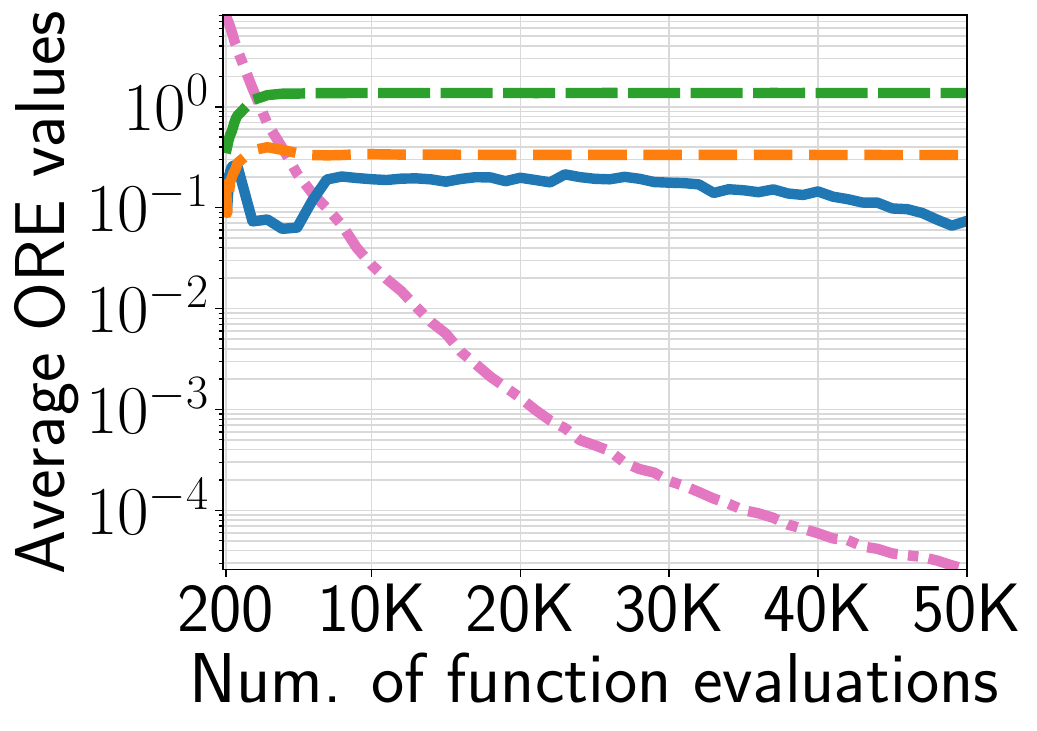}}
   \subfloat[ORE ($m=6$)]{\includegraphics[width=0.32\textwidth]{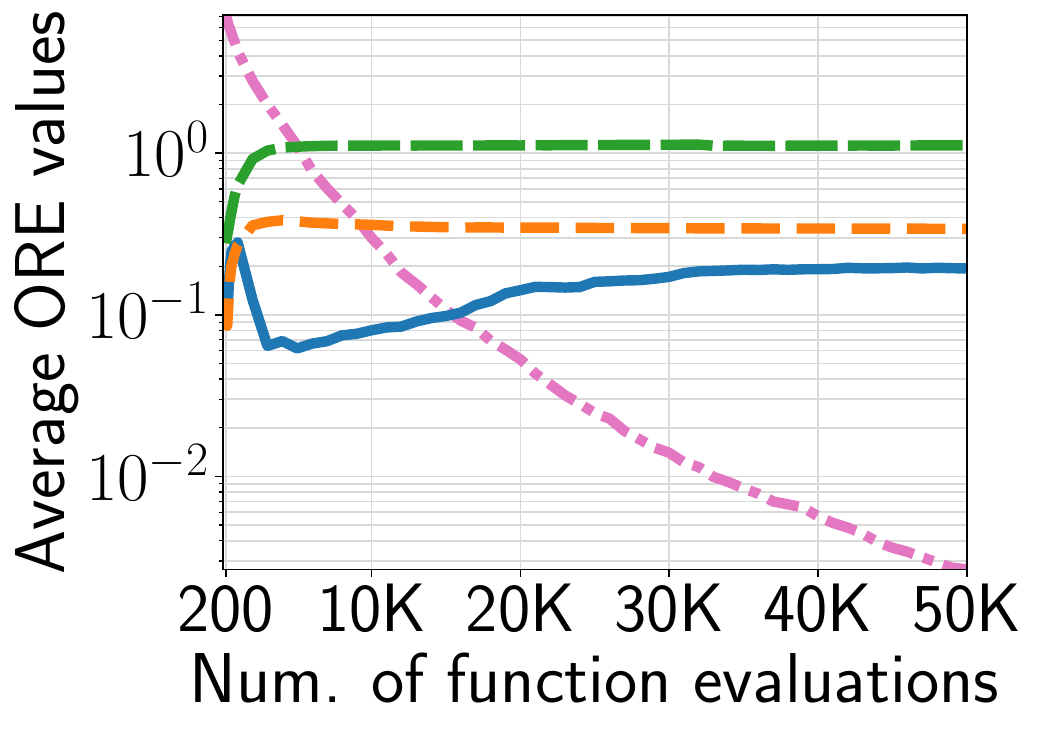}}
\\
\caption{Average $e^{\mathrm{ideal}}$, $e^{\mathrm{nadir}}$, and ORE values of the three normalization methods in MOEA/D-NUMS on DTLZ7.}
\label{supfig:3error_MOEADNUMS_DTLZ7}
\end{figure*}

\begin{figure*}[t]
\centering
  \subfloat{\includegraphics[width=0.7\textwidth]{./figs/legend/legend_3.pdf}}
\vspace{-3.9mm}
   \\
   \subfloat[$e^{\mathrm{ideal}}$ ($m=2$)]{\includegraphics[width=0.32\textwidth]{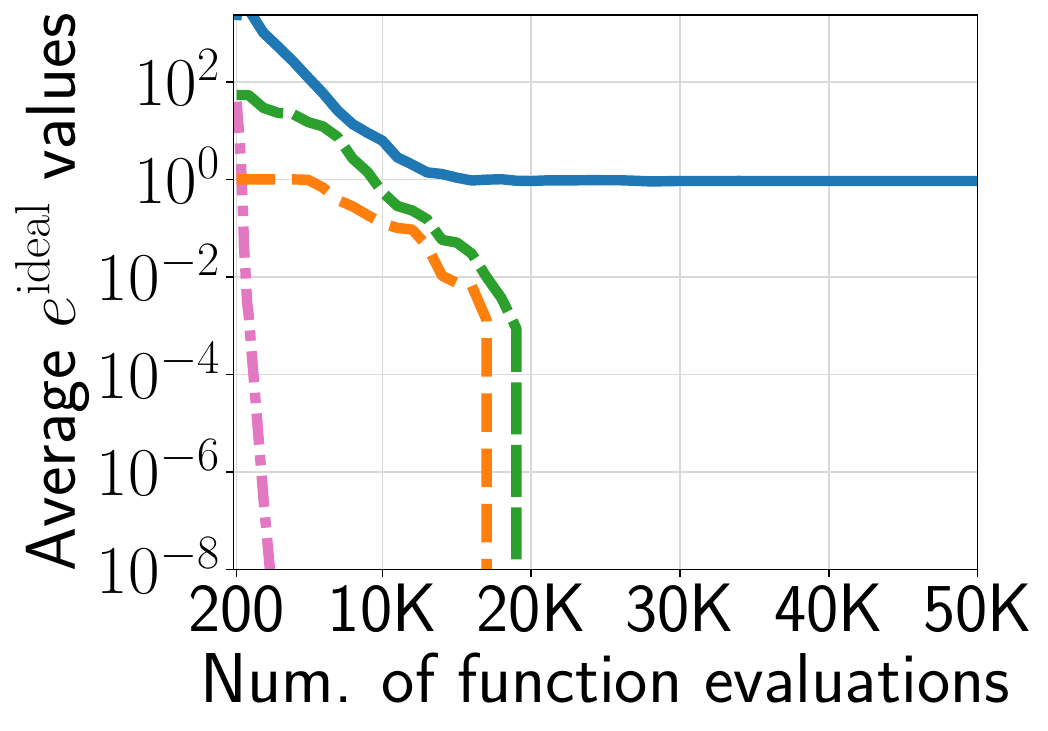}}
   \subfloat[$e^{\mathrm{ideal}}$ ($m=4$)]{\includegraphics[width=0.32\textwidth]{./figs/qi_error_ideal/MOEADNUMS_mu100/SDTLZ1_m4_r0.1_z-type1.pdf}}
   \subfloat[$e^{\mathrm{ideal}}$ ($m=6$)]{\includegraphics[width=0.32\textwidth]{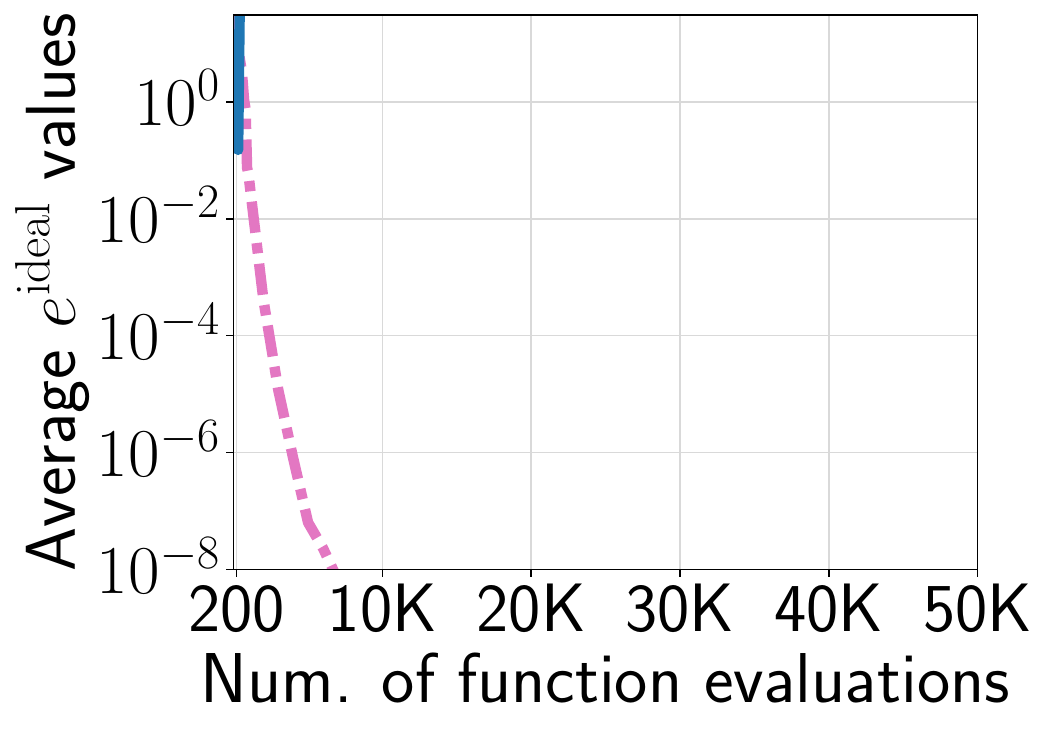}}
\\
   \subfloat[$e^{\mathrm{nadir}}$ ($m=2$)]{\includegraphics[width=0.32\textwidth]{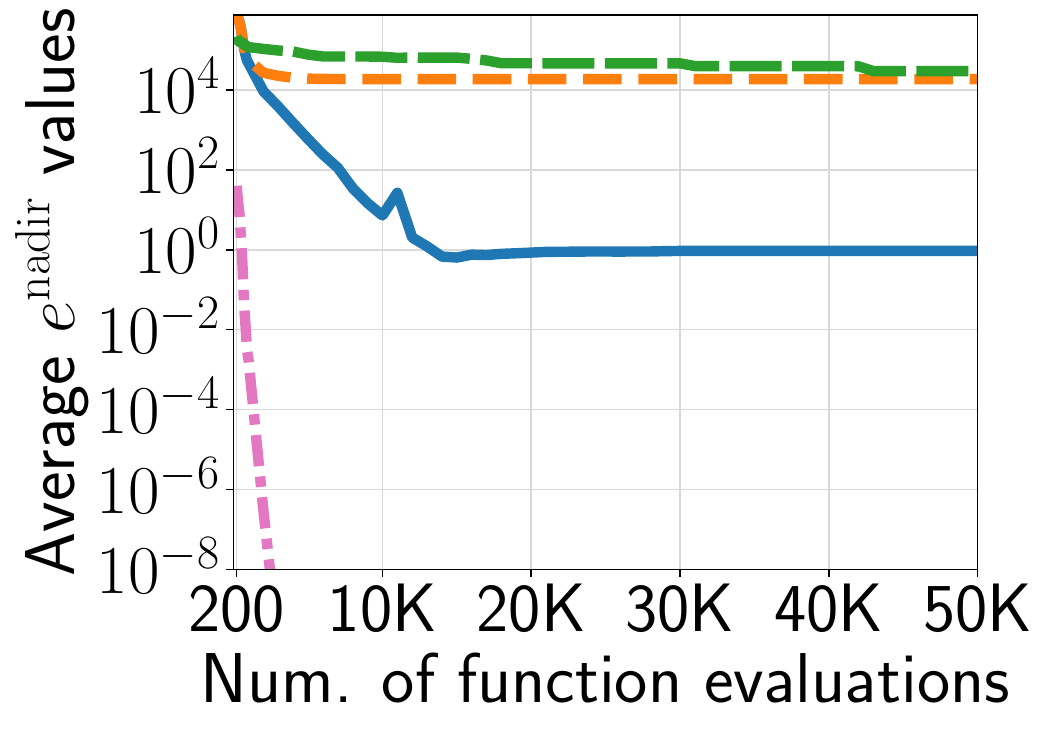}}
   \subfloat[$e^{\mathrm{nadir}}$ ($m=4$)]{\includegraphics[width=0.32\textwidth]{./figs/qi_error_nadir/MOEADNUMS_mu100/SDTLZ1_m4_r0.1_z-type1.pdf}}
   \subfloat[$e^{\mathrm{nadir}}$ ($m=6$)]{\includegraphics[width=0.32\textwidth]{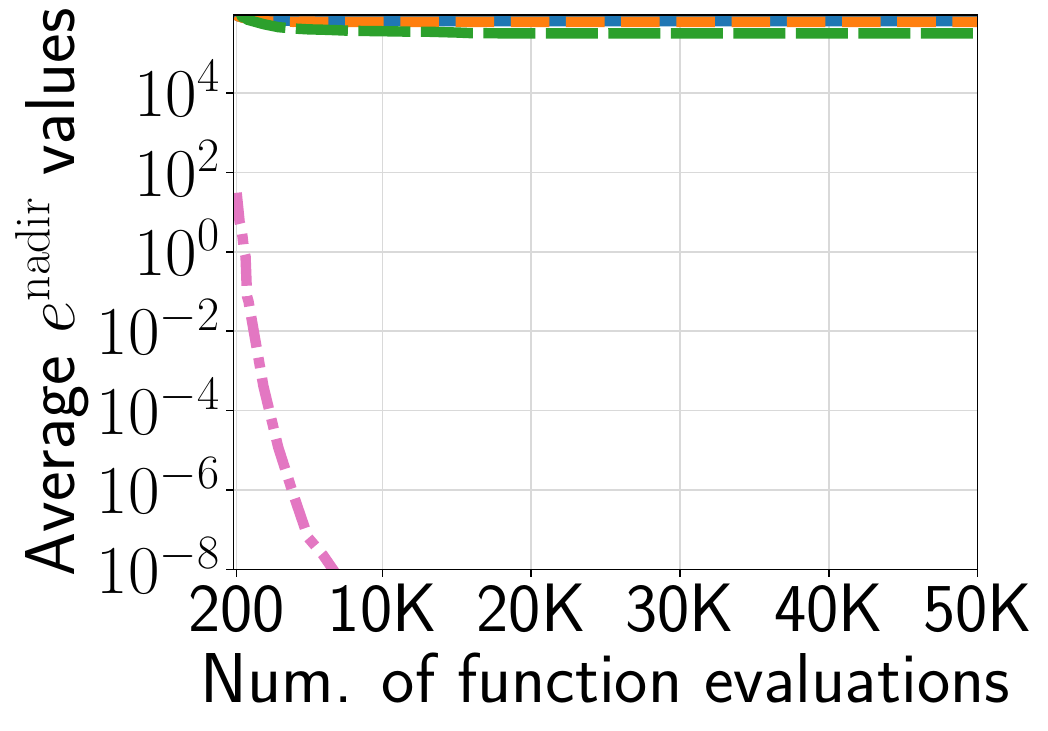}}
\\
   \subfloat[ORE ($m=2$)]{\includegraphics[width=0.32\textwidth]{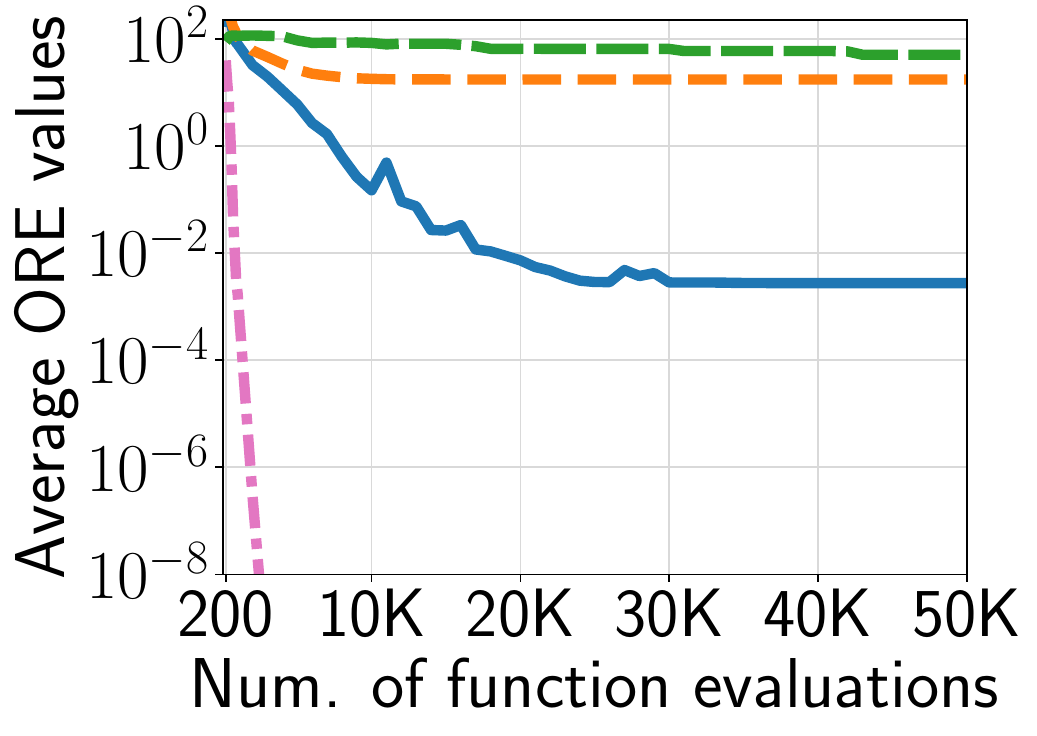}}
   \subfloat[ORE ($m=4$)]{\includegraphics[width=0.32\textwidth]{./figs/qi_ore/MOEADNUMS_mu100/SDTLZ1_m4_r0.1_z-type1.pdf}}
   \subfloat[ORE ($m=6$)]{\includegraphics[width=0.32\textwidth]{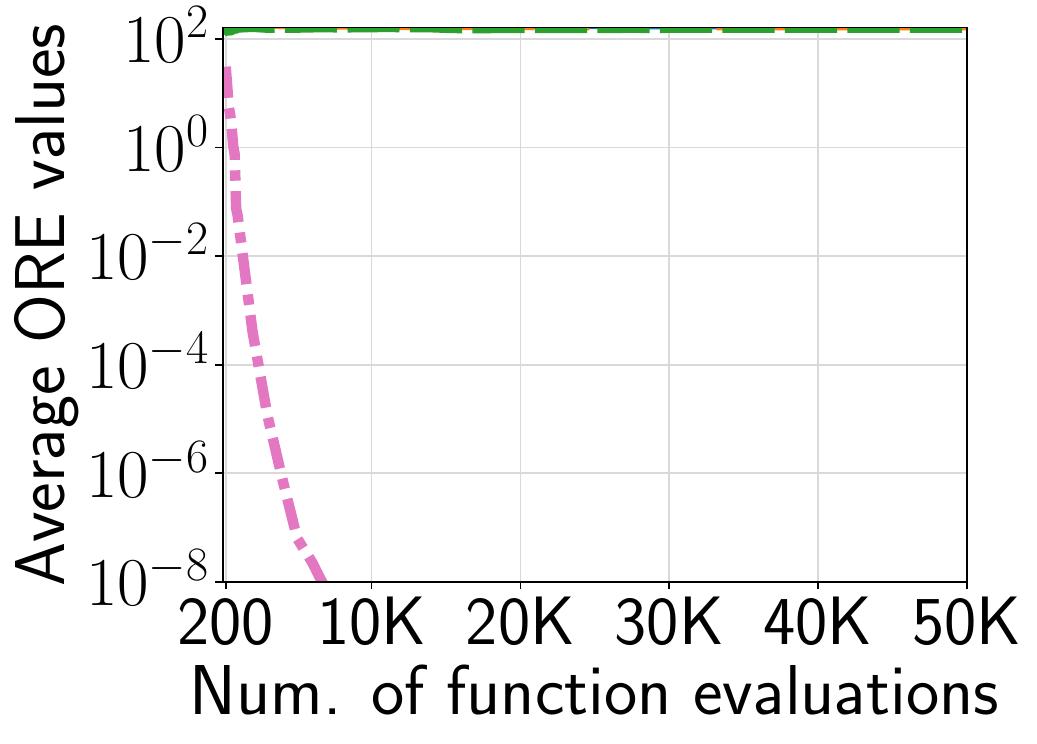}}
\\
\caption{Average $e^{\mathrm{ideal}}$, $e^{\mathrm{nadir}}$, and ORE values of the three normalization methods in MOEA/D-NUMS on SDTLZ1.}
\label{supfig:3error_MOEADNUMS_SDTLZ1}
\end{figure*}

\begin{figure*}[t]
\centering
  \subfloat{\includegraphics[width=0.7\textwidth]{./figs/legend/legend_3.pdf}}
\vspace{-3.9mm}
   \\
   \subfloat[$e^{\mathrm{ideal}}$ ($m=2$)]{\includegraphics[width=0.32\textwidth]{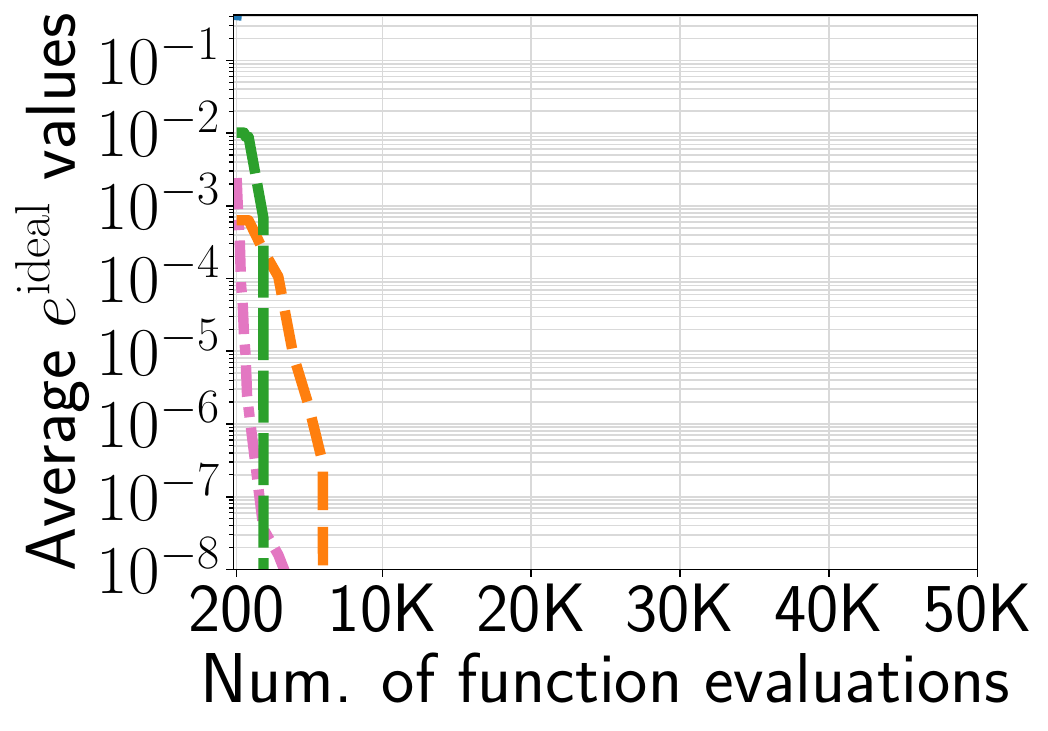}}
   \subfloat[$e^{\mathrm{ideal}}$ ($m=4$)]{\includegraphics[width=0.32\textwidth]{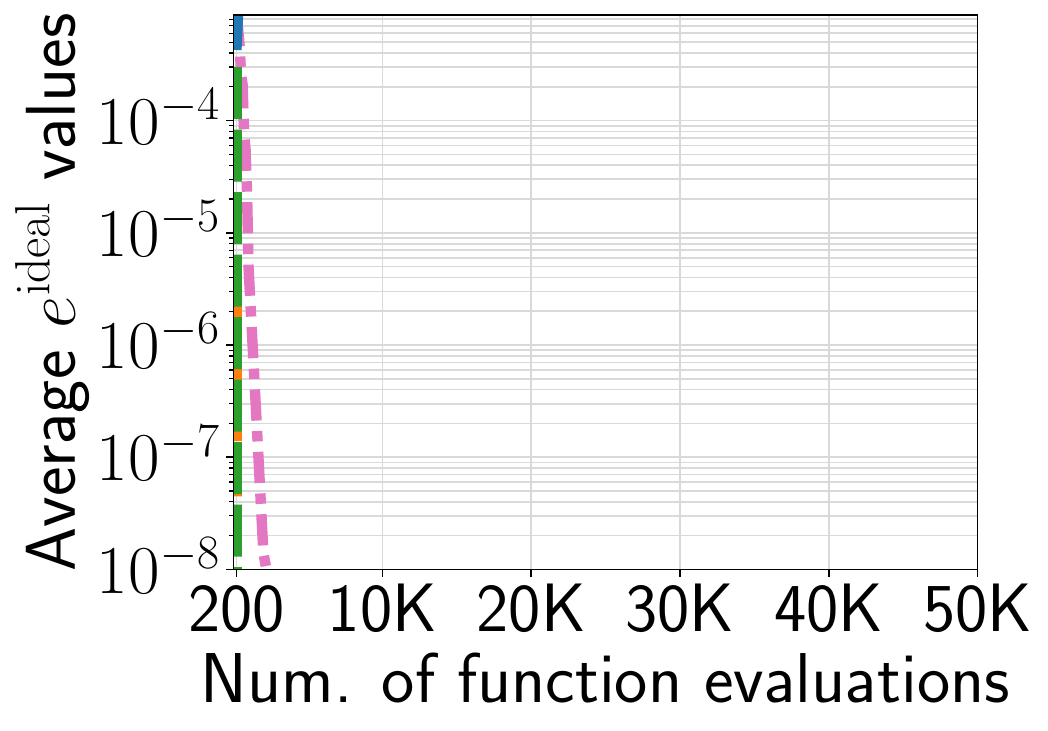}}
   \subfloat[$e^{\mathrm{ideal}}$ ($m=6$)]{\includegraphics[width=0.32\textwidth]{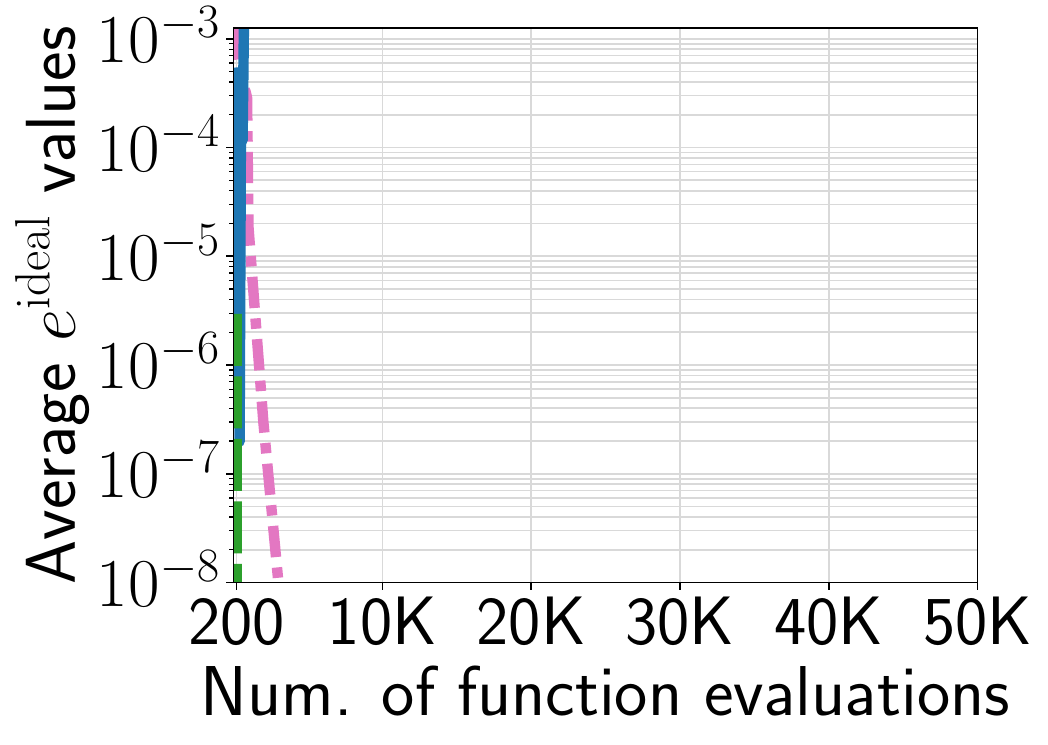}}
\\
   \subfloat[$e^{\mathrm{nadir}}$ ($m=2$)]{\includegraphics[width=0.32\textwidth]{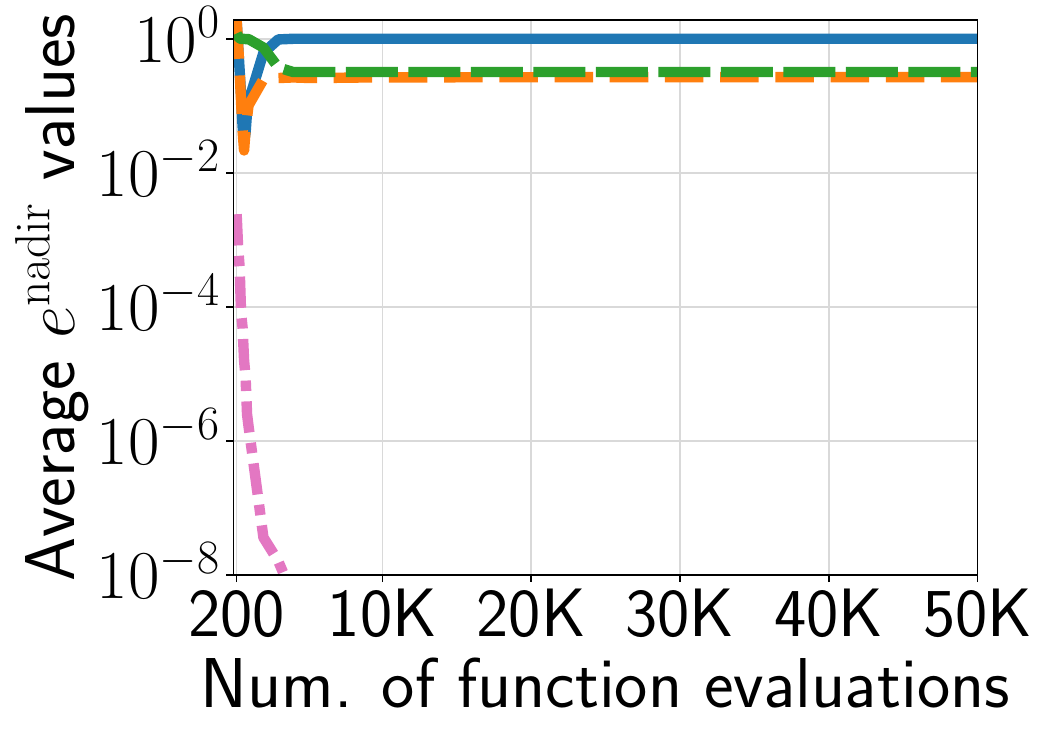}}
   \subfloat[$e^{\mathrm{nadir}}$ ($m=4$)]{\includegraphics[width=0.32\textwidth]{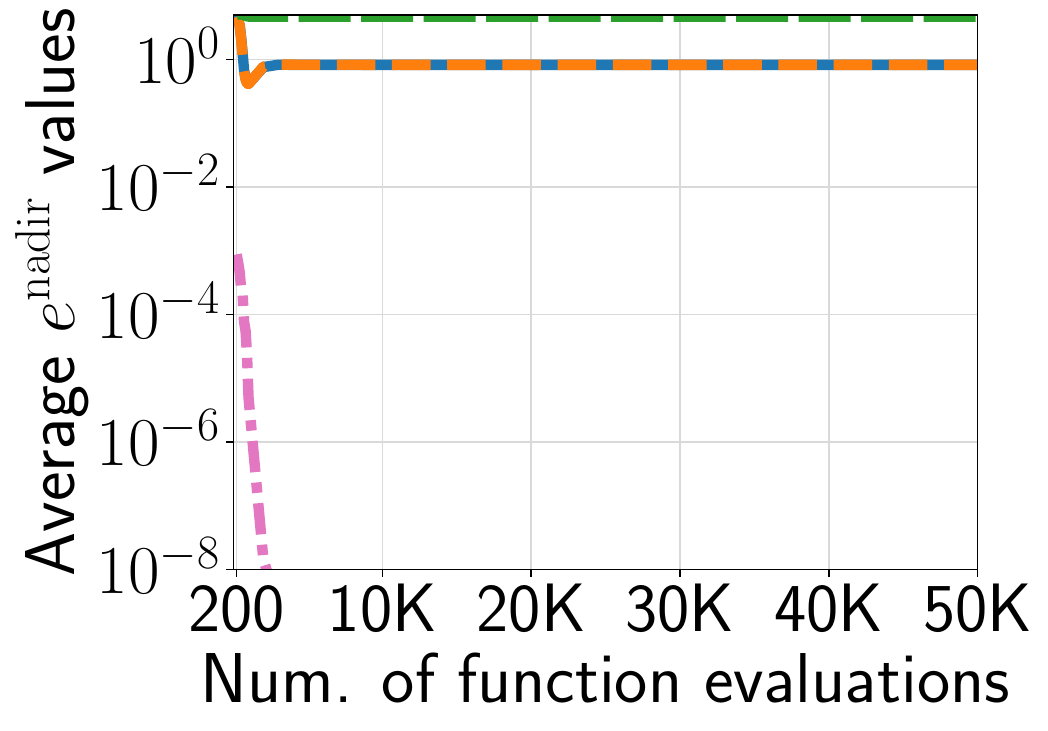}}
   \subfloat[$e^{\mathrm{nadir}}$ ($m=6$)]{\includegraphics[width=0.32\textwidth]{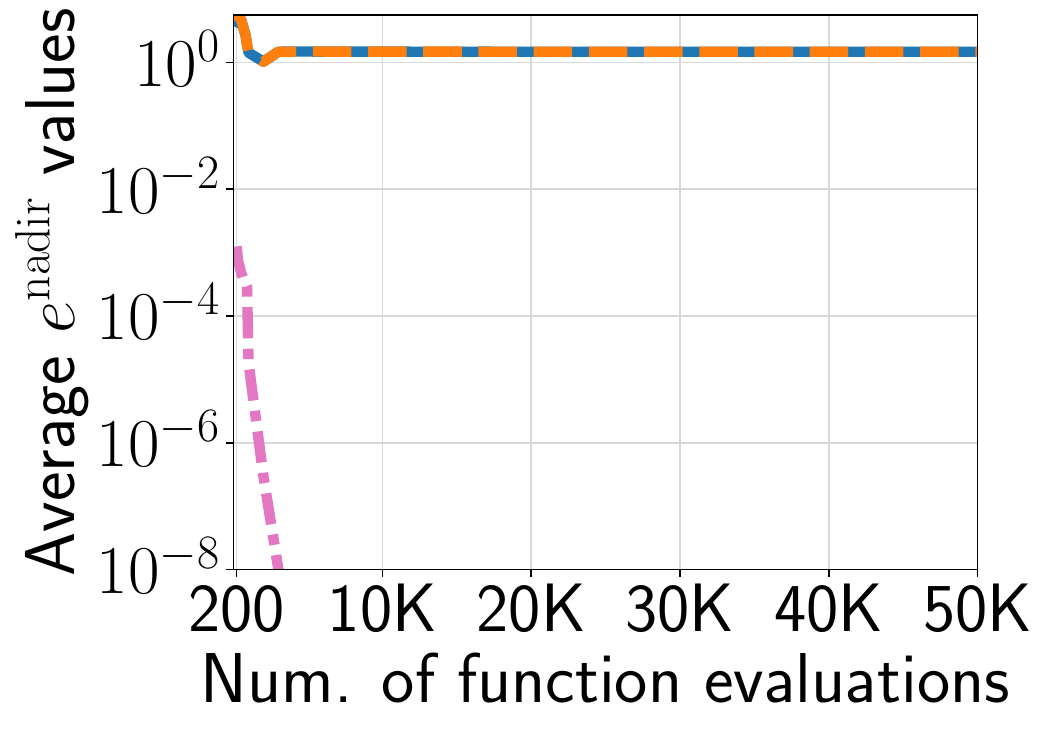}}
\\
   \subfloat[ORE ($m=2$)]{\includegraphics[width=0.32\textwidth]{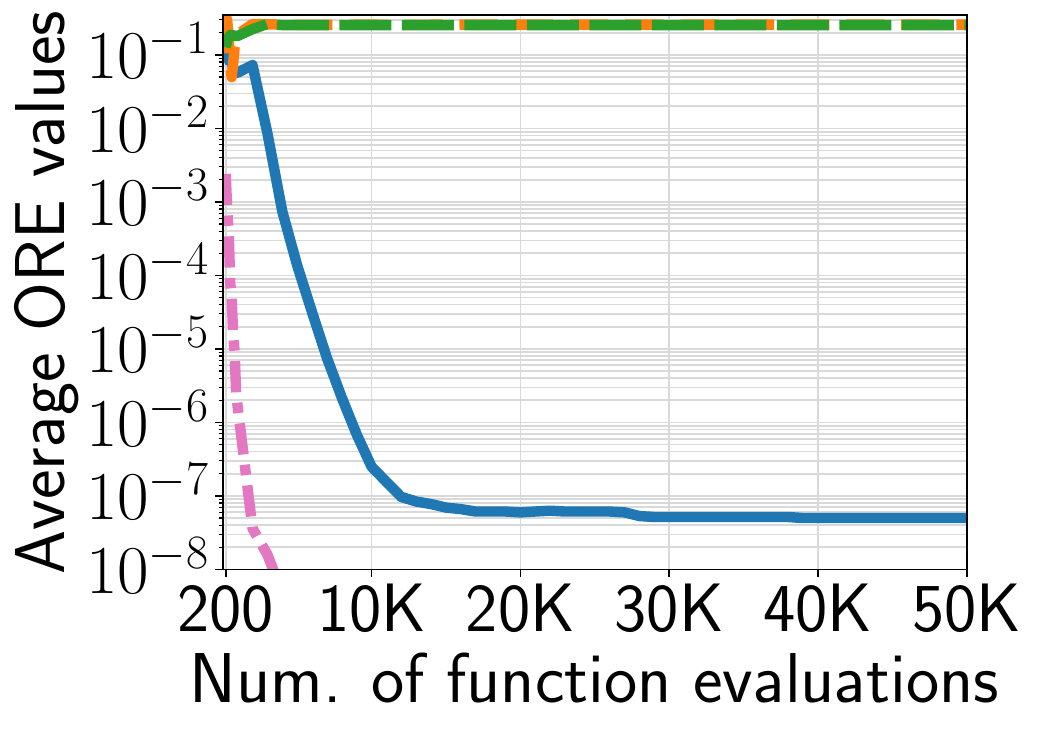}}
   \subfloat[ORE ($m=4$)]{\includegraphics[width=0.32\textwidth]{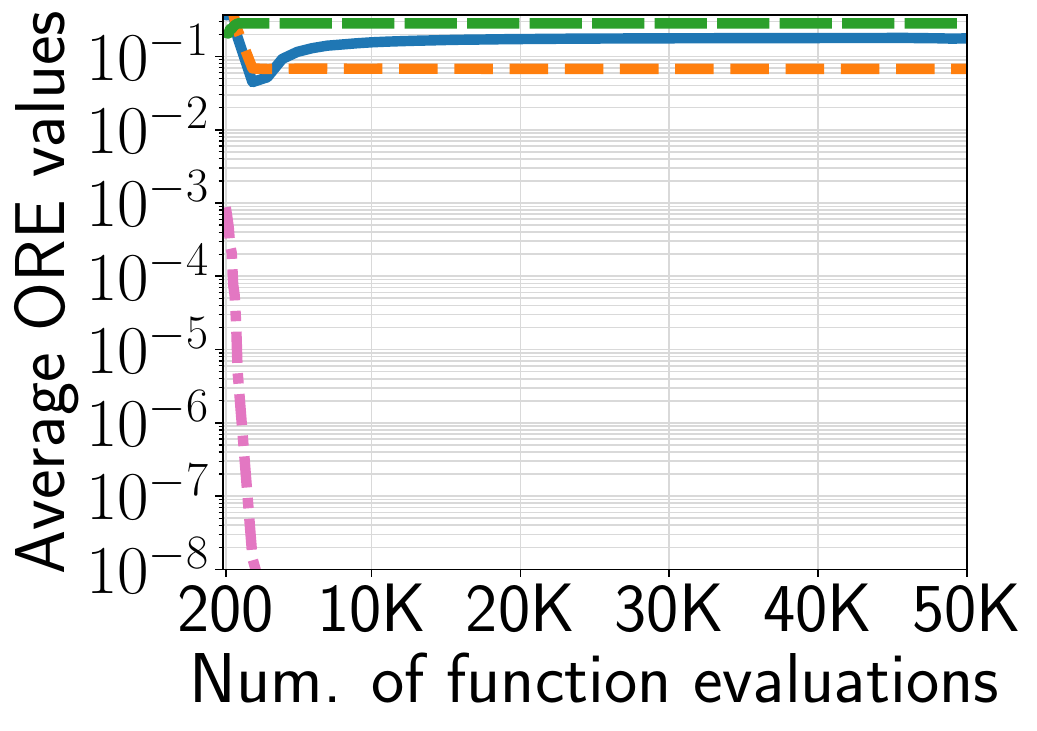}}
   \subfloat[ORE ($m=6$)]{\includegraphics[width=0.32\textwidth]{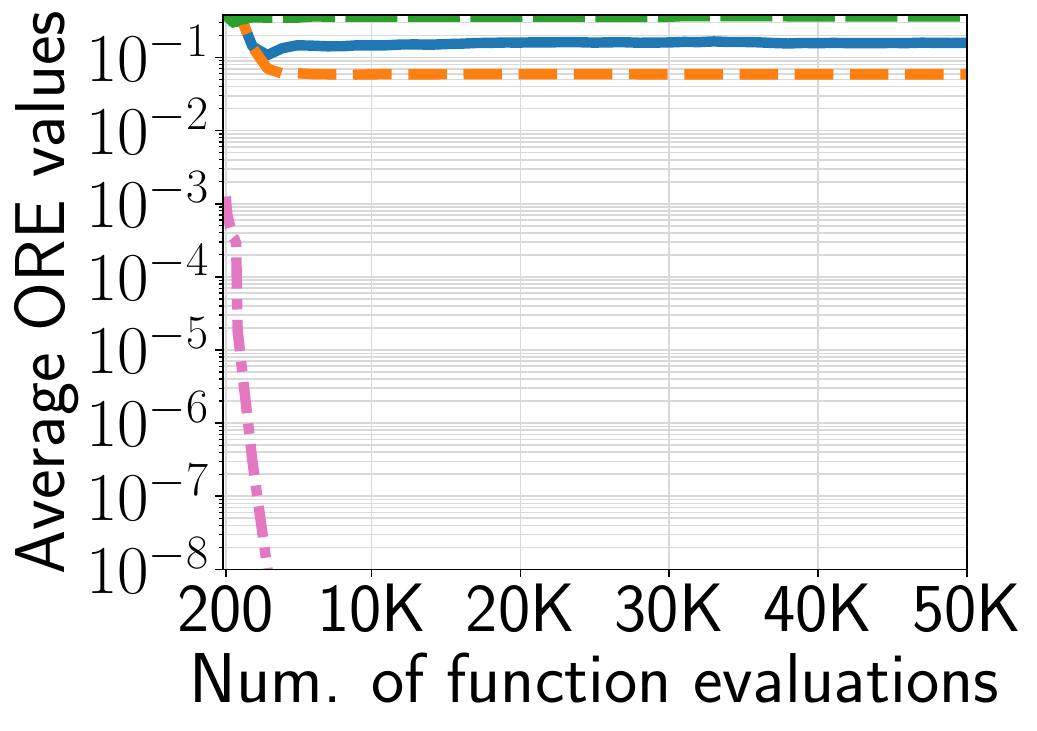}}
\\
\caption{Average $e^{\mathrm{ideal}}$, $e^{\mathrm{nadir}}$, and ORE values of the three normalization methods in MOEA/D-NUMS on SDTLZ2.}
\label{supfig:3error_MOEADNUMS_SDTLZ2}
\end{figure*}

\begin{figure*}[t]
\centering
  \subfloat{\includegraphics[width=0.7\textwidth]{./figs/legend/legend_3.pdf}}
\vspace{-3.9mm}
   \\
   \subfloat[$e^{\mathrm{ideal}}$ ($m=2$)]{\includegraphics[width=0.32\textwidth]{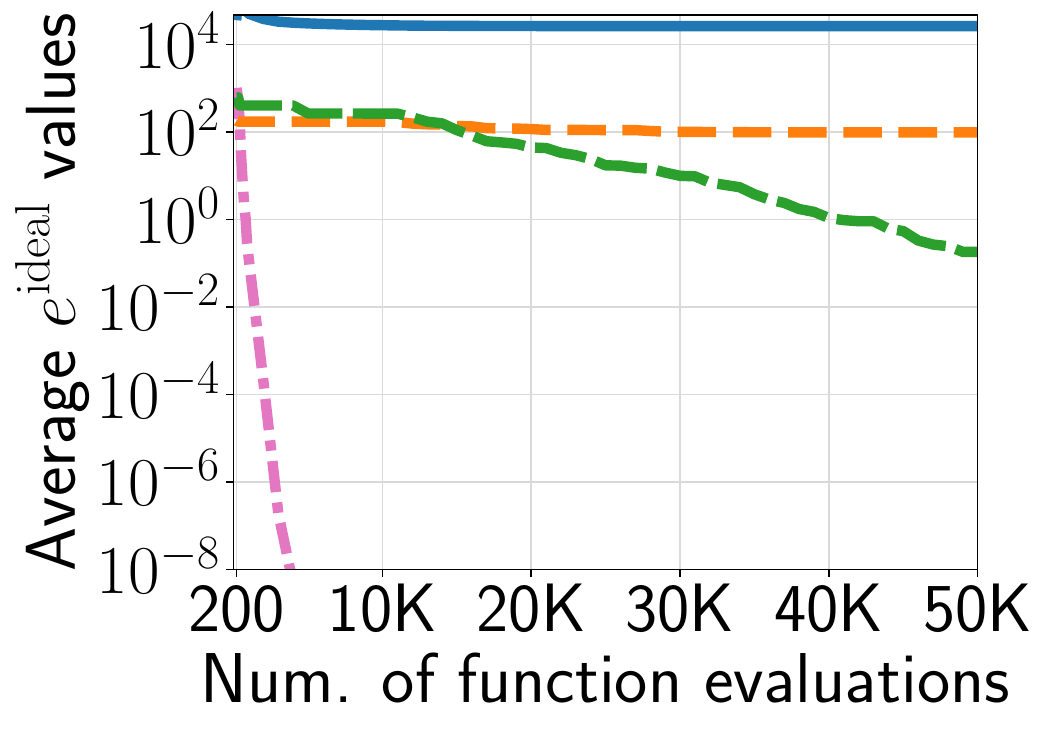}}
   \subfloat[$e^{\mathrm{ideal}}$ ($m=4$)]{\includegraphics[width=0.32\textwidth]{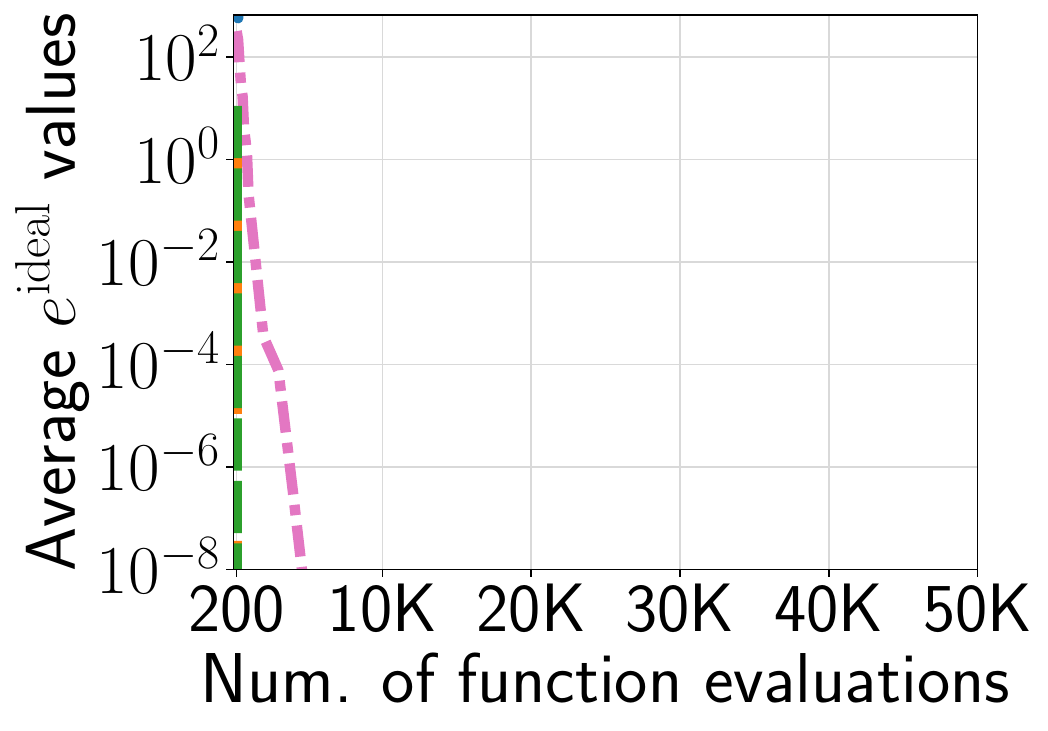}}
   \subfloat[$e^{\mathrm{ideal}}$ ($m=6$)]{\includegraphics[width=0.32\textwidth]{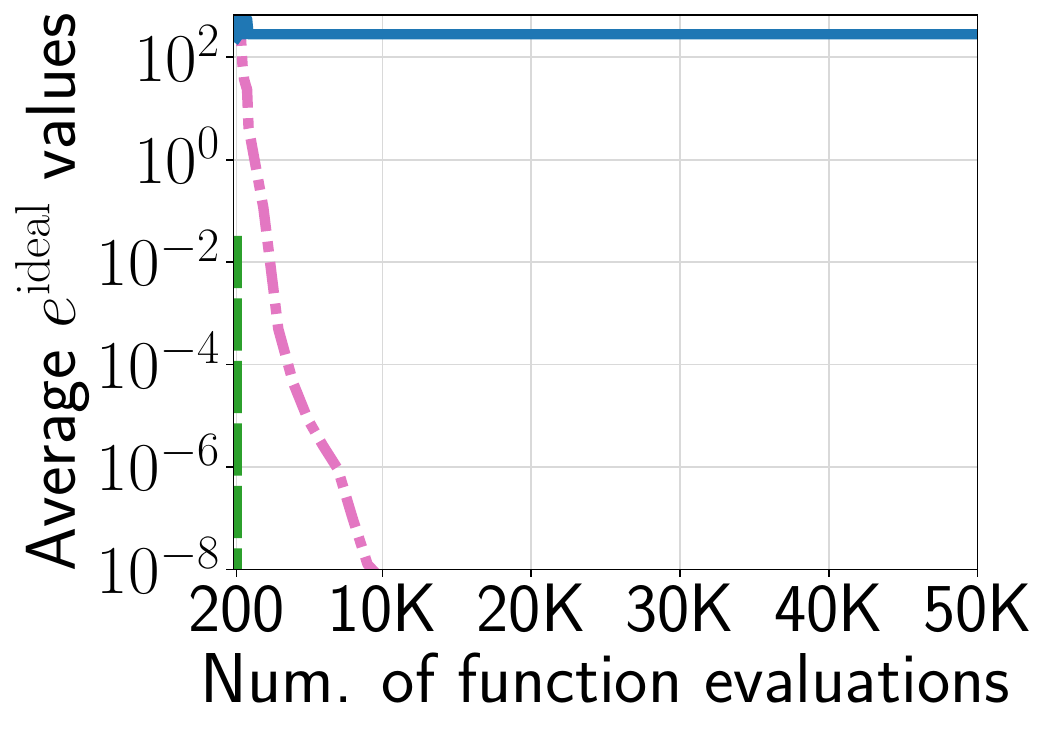}}
\\
   \subfloat[$e^{\mathrm{nadir}}$ ($m=2$)]{\includegraphics[width=0.32\textwidth]{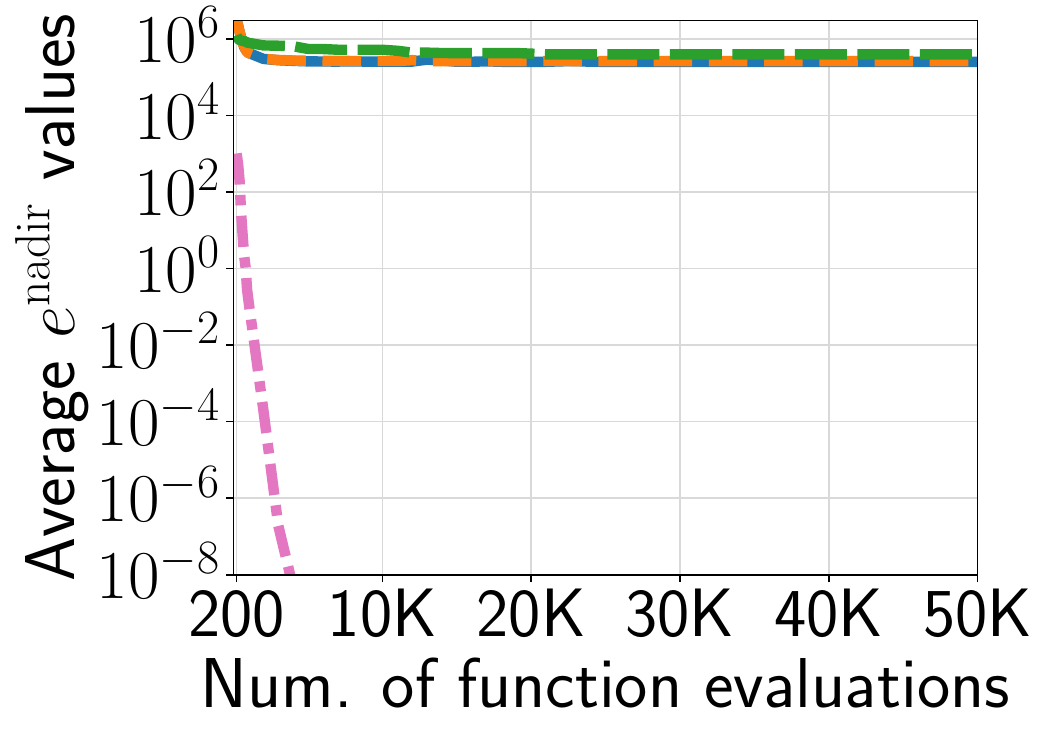}}
   \subfloat[$e^{\mathrm{nadir}}$ ($m=4$)]{\includegraphics[width=0.32\textwidth]{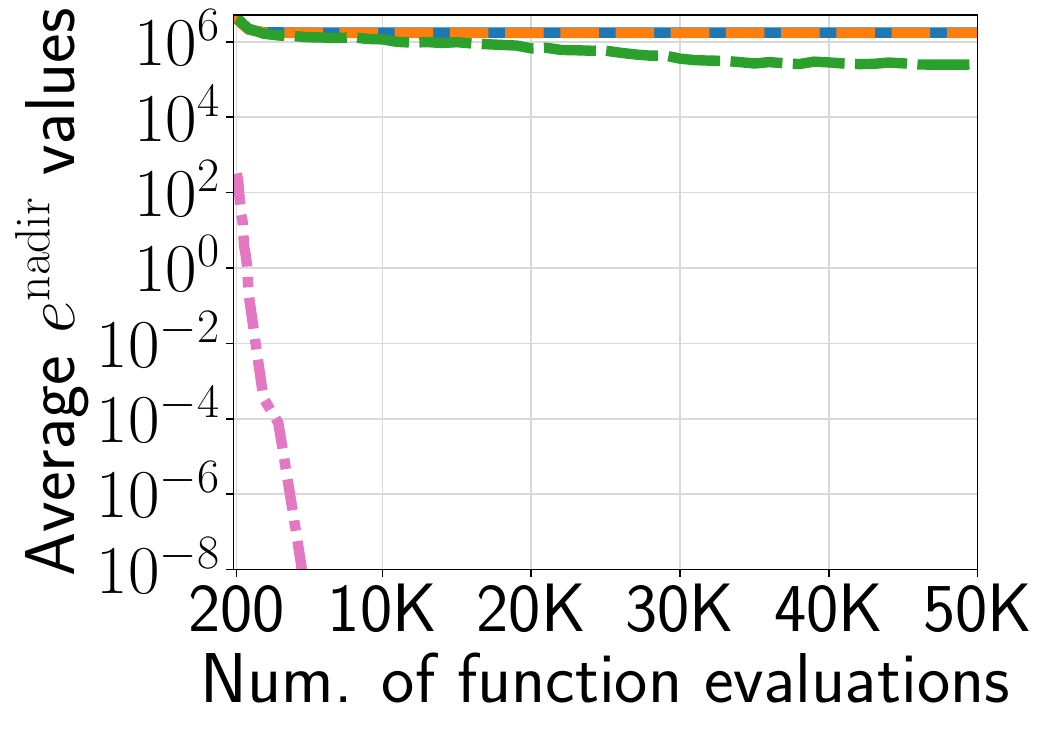}}
   \subfloat[$e^{\mathrm{nadir}}$ ($m=6$)]{\includegraphics[width=0.32\textwidth]{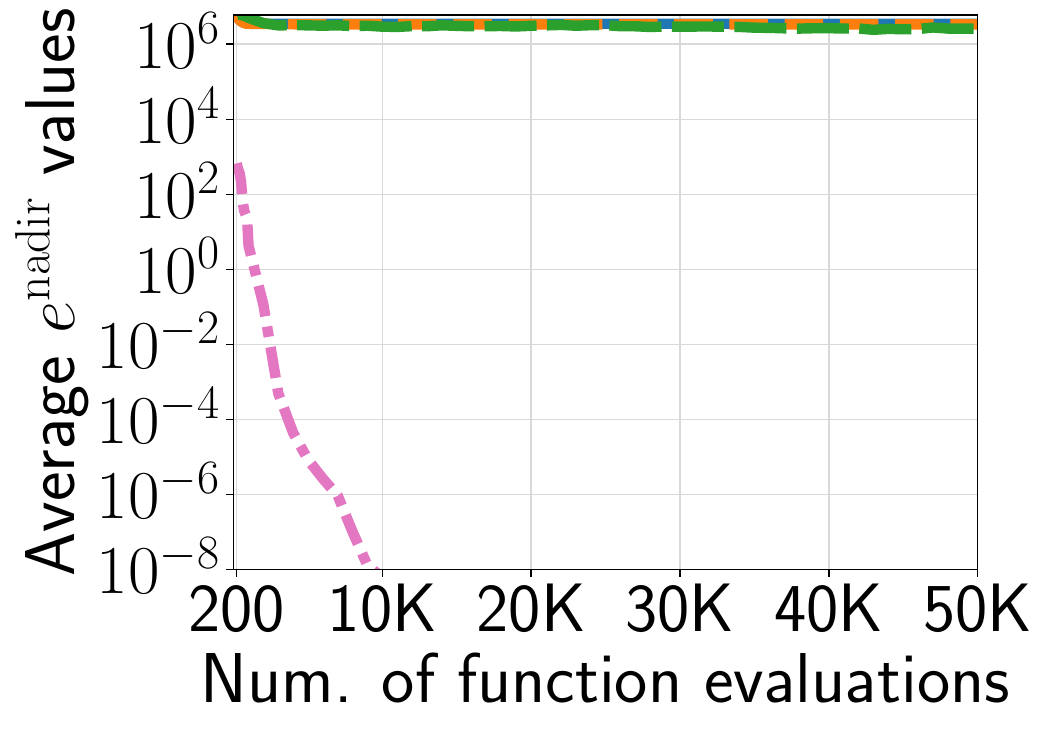}}
\\
   \subfloat[ORE ($m=2$)]{\includegraphics[width=0.32\textwidth]{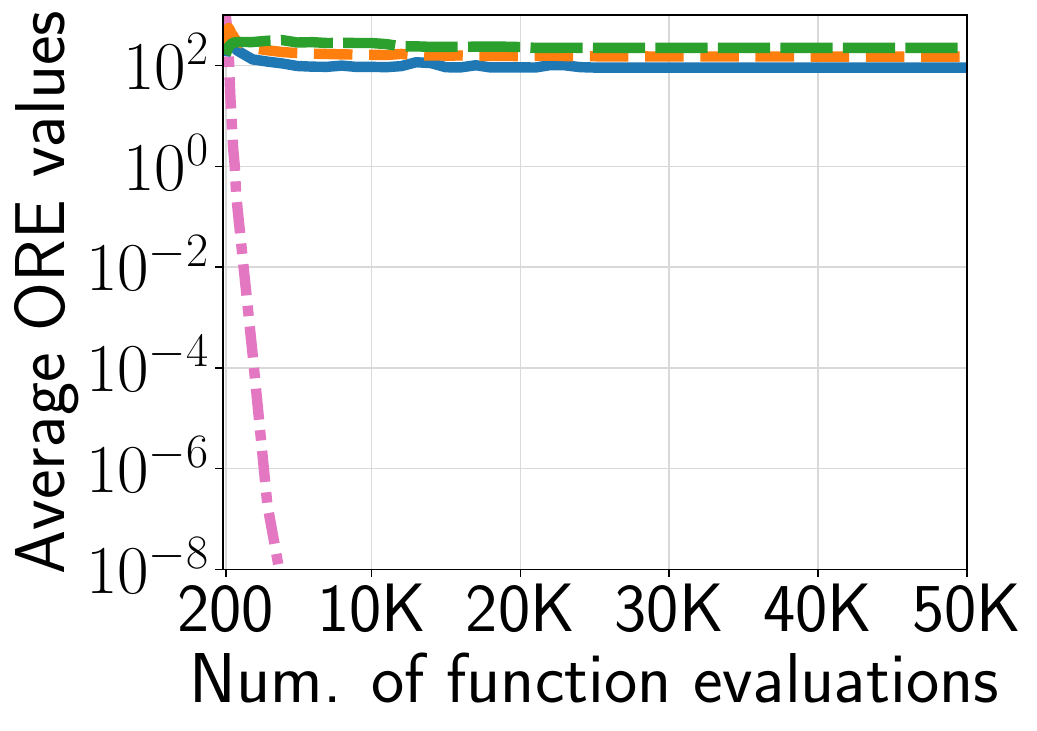}}
   \subfloat[ORE ($m=4$)]{\includegraphics[width=0.32\textwidth]{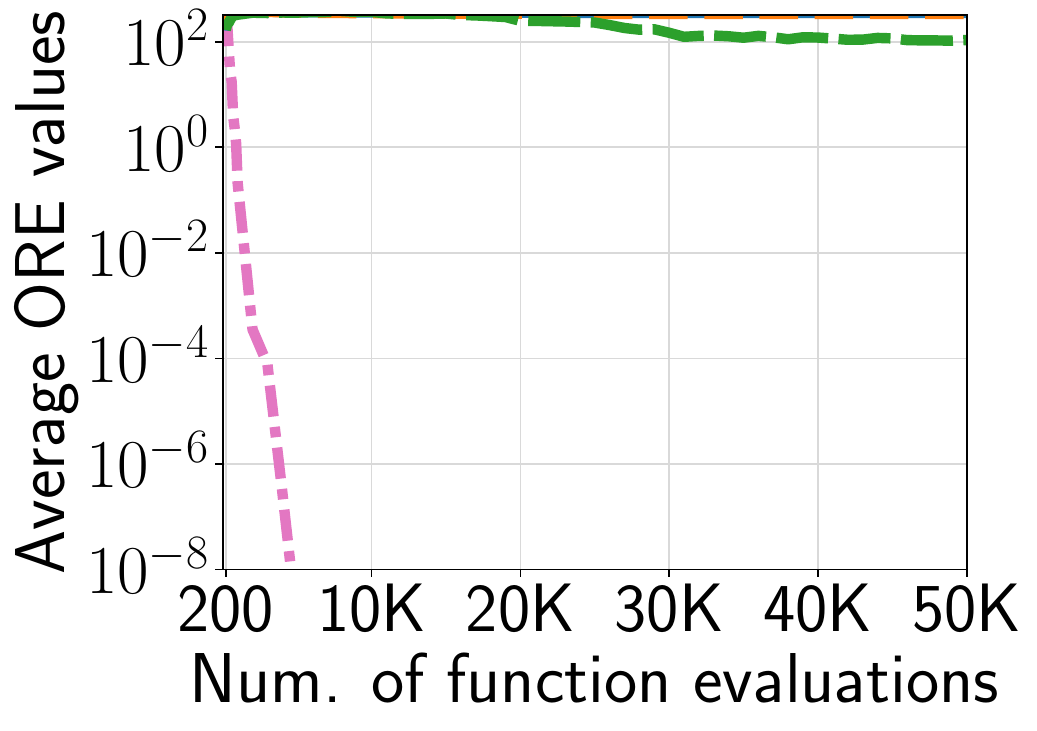}}
   \subfloat[ORE ($m=6$)]{\includegraphics[width=0.32\textwidth]{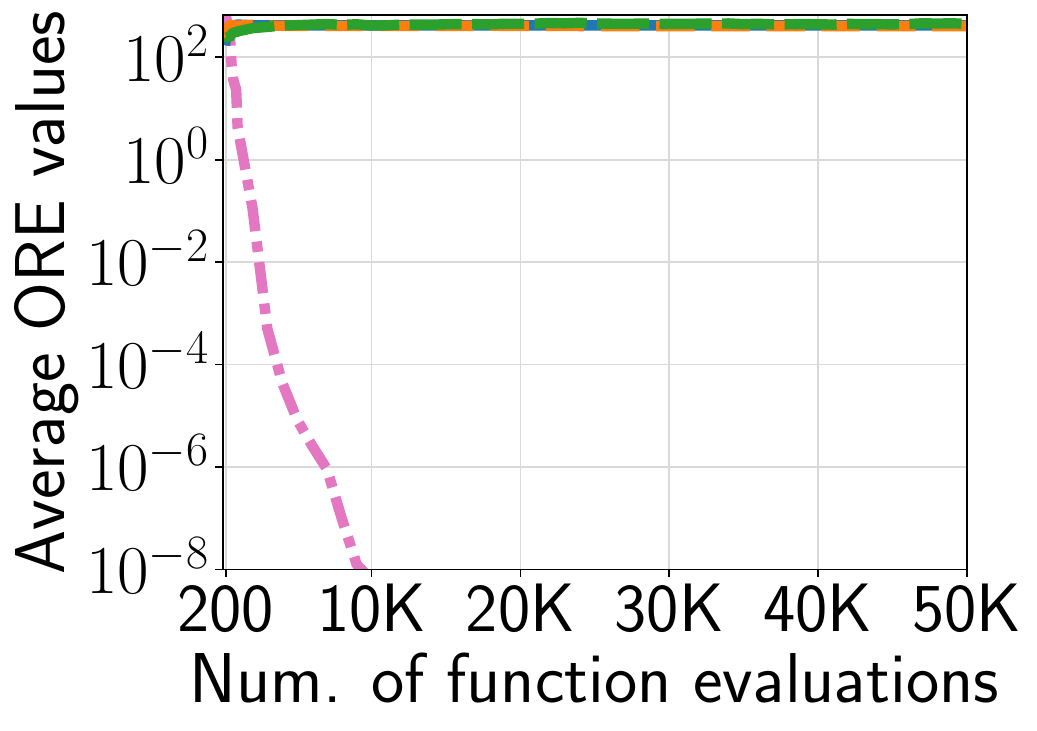}}
\\
\caption{Average $e^{\mathrm{ideal}}$, $e^{\mathrm{nadir}}$, and ORE values of the three normalization methods in MOEA/D-NUMS on SDTLZ3.}
\label{supfig:3error_MOEADNUMS_SDTLZ3}
\end{figure*}

\begin{figure*}[t]
\centering
  \subfloat{\includegraphics[width=0.7\textwidth]{./figs/legend/legend_3.pdf}}
\vspace{-3.9mm}
   \\
   \subfloat[$e^{\mathrm{ideal}}$ ($m=2$)]{\includegraphics[width=0.32\textwidth]{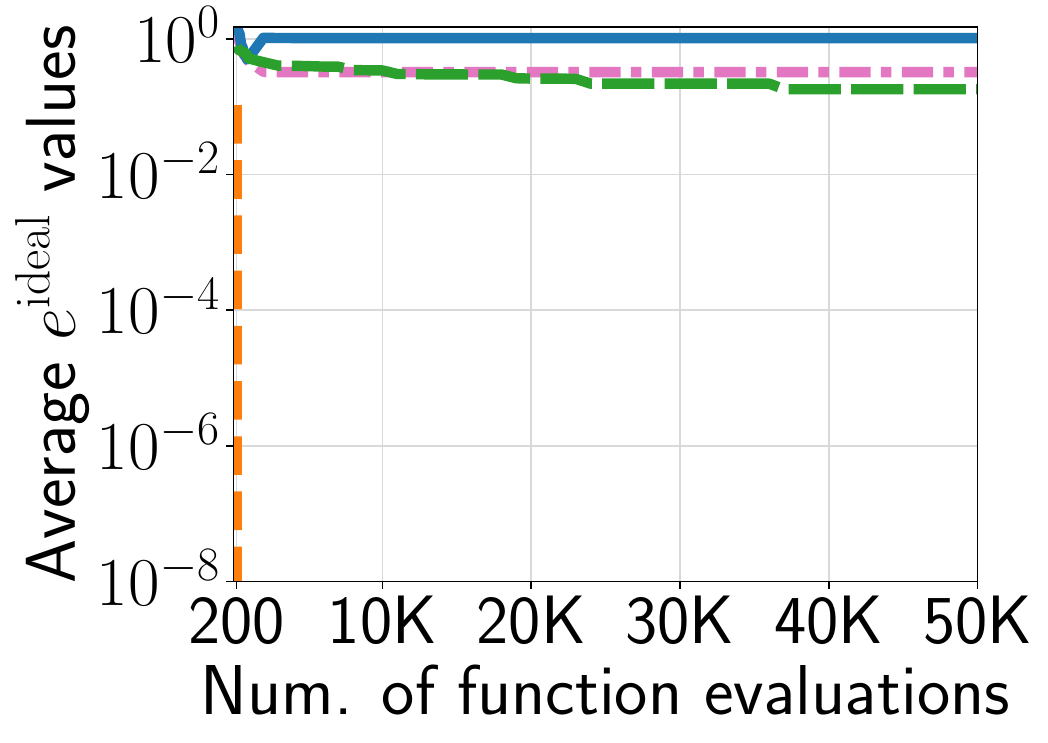}}
   \subfloat[$e^{\mathrm{ideal}}$ ($m=4$)]{\includegraphics[width=0.32\textwidth]{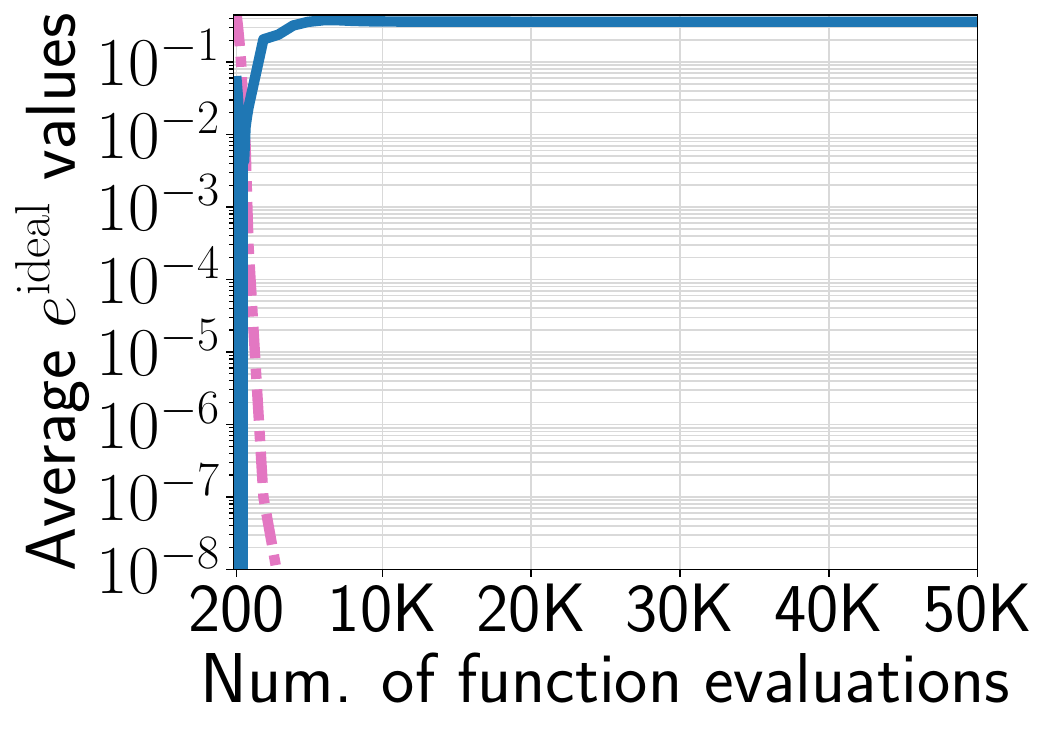}}
   \subfloat[$e^{\mathrm{ideal}}$ ($m=6$)]{\includegraphics[width=0.32\textwidth]{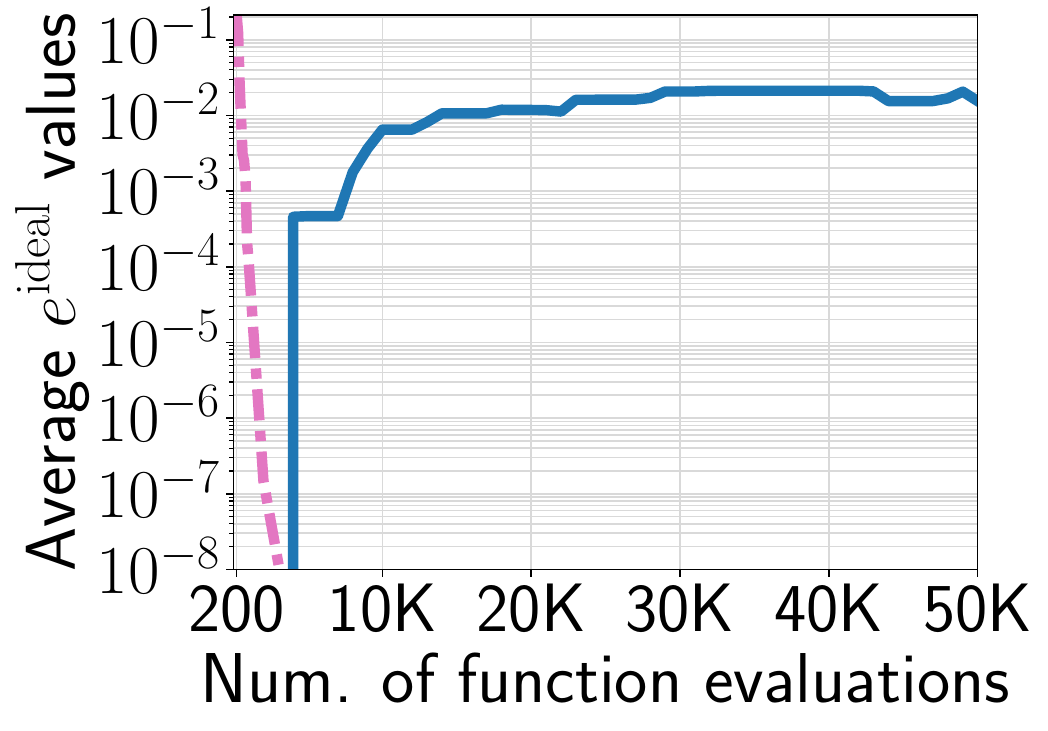}}
\\
   \subfloat[$e^{\mathrm{nadir}}$ ($m=2$)]{\includegraphics[width=0.32\textwidth]{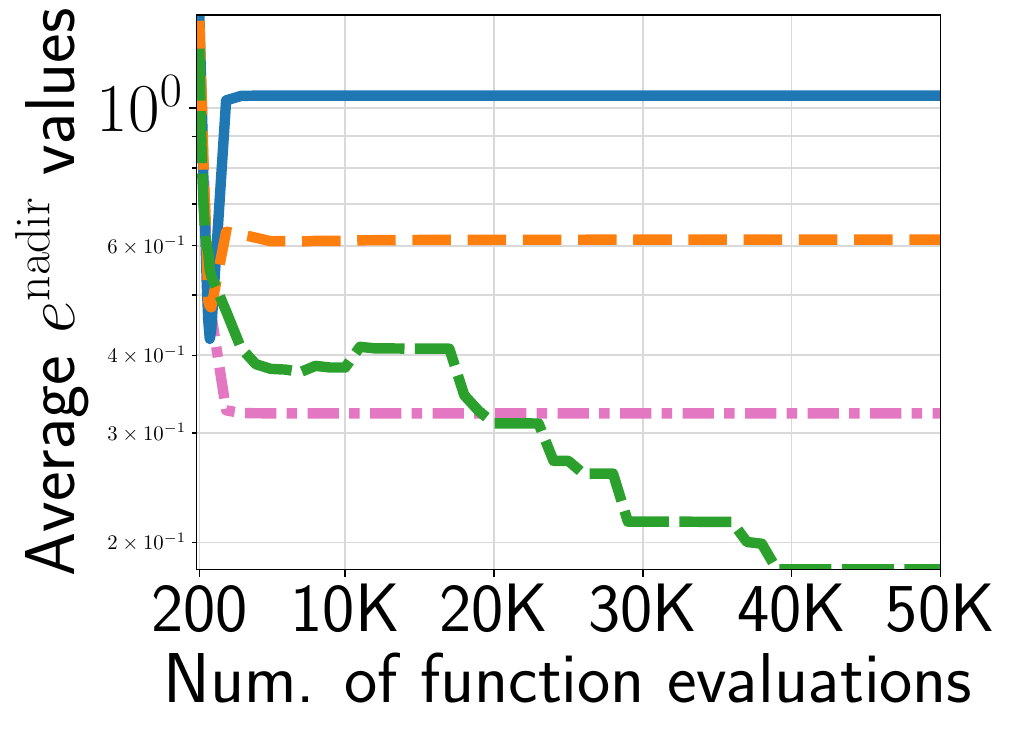}}
   \subfloat[$e^{\mathrm{nadir}}$ ($m=4$)]{\includegraphics[width=0.32\textwidth]{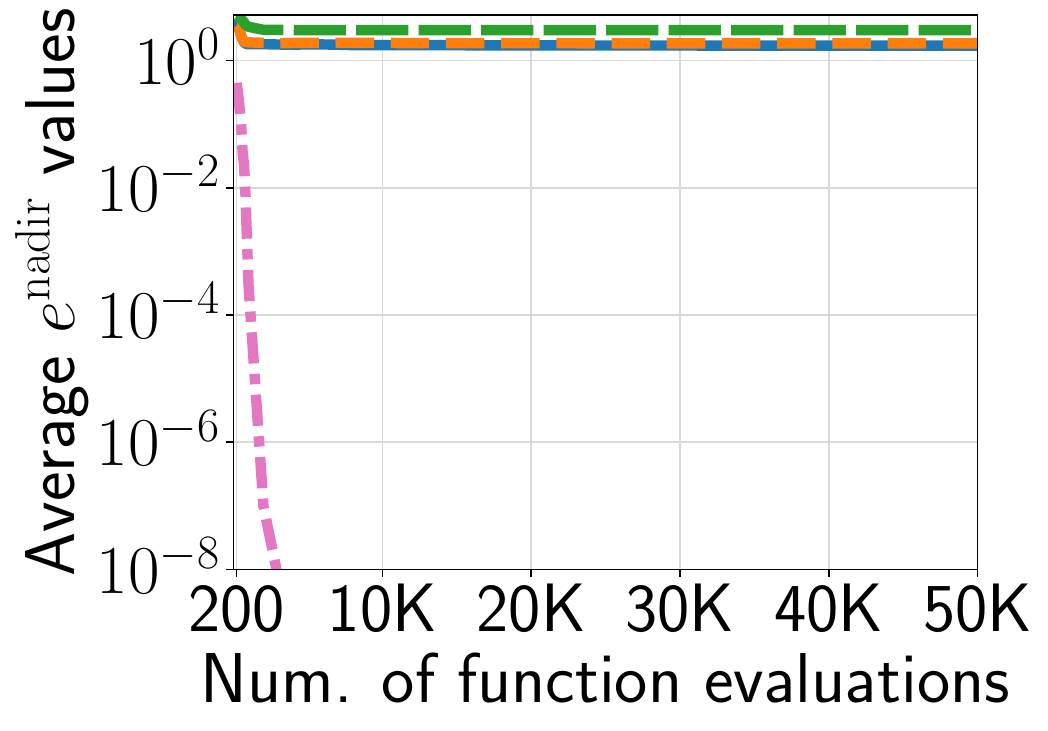}}
   \subfloat[$e^{\mathrm{nadir}}$ ($m=6$)]{\includegraphics[width=0.32\textwidth]{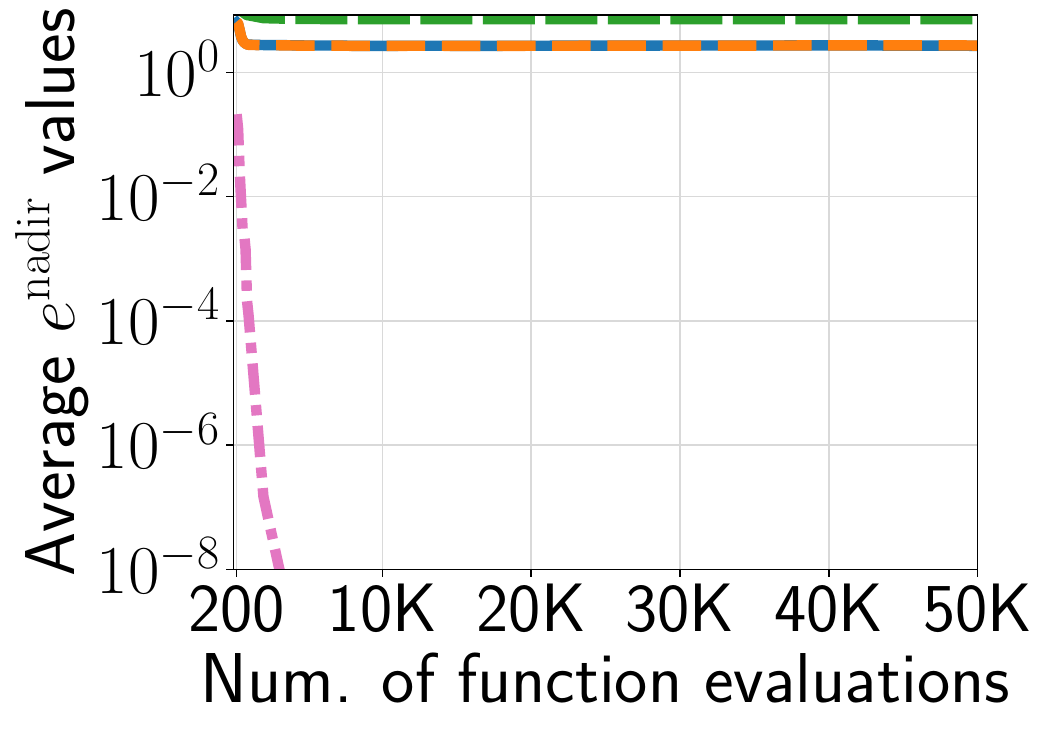}}
\\
   \subfloat[ORE ($m=2$)]{\includegraphics[width=0.32\textwidth]{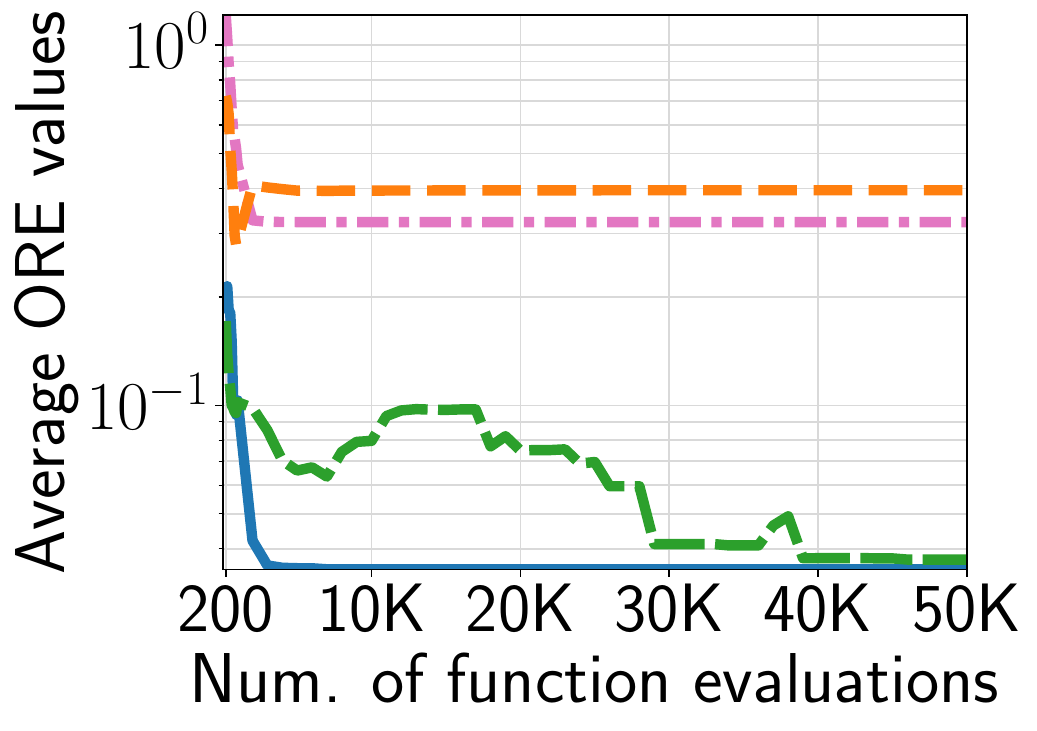}}
   \subfloat[ORE ($m=4$)]{\includegraphics[width=0.32\textwidth]{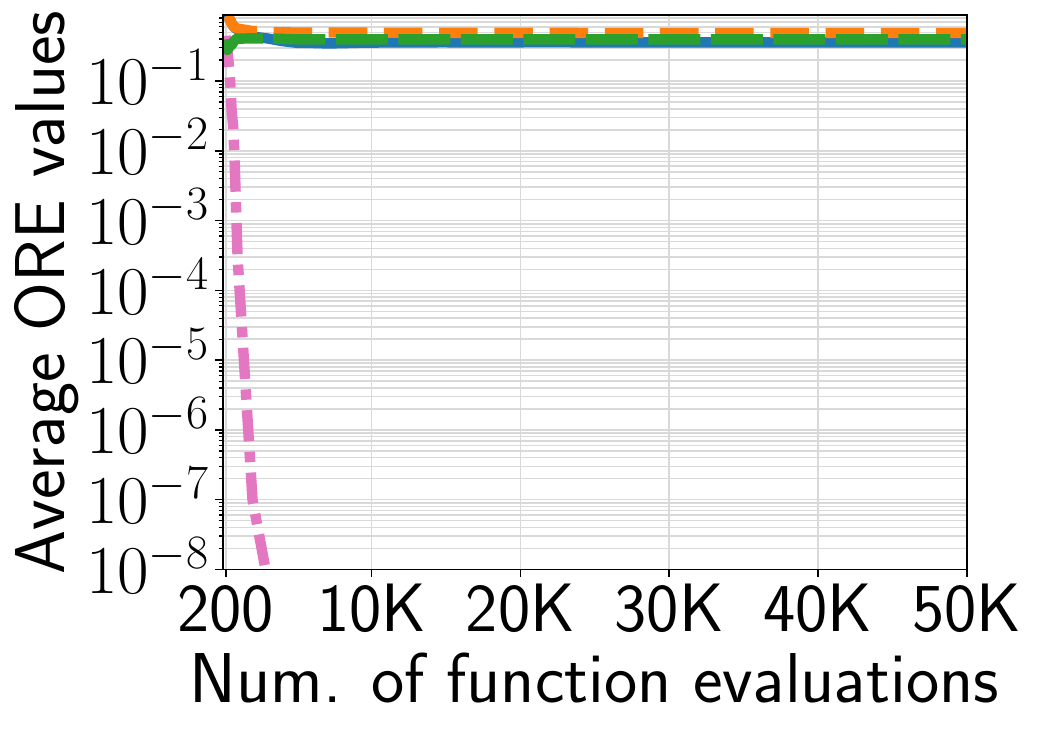}}
   \subfloat[ORE ($m=6$)]{\includegraphics[width=0.32\textwidth]{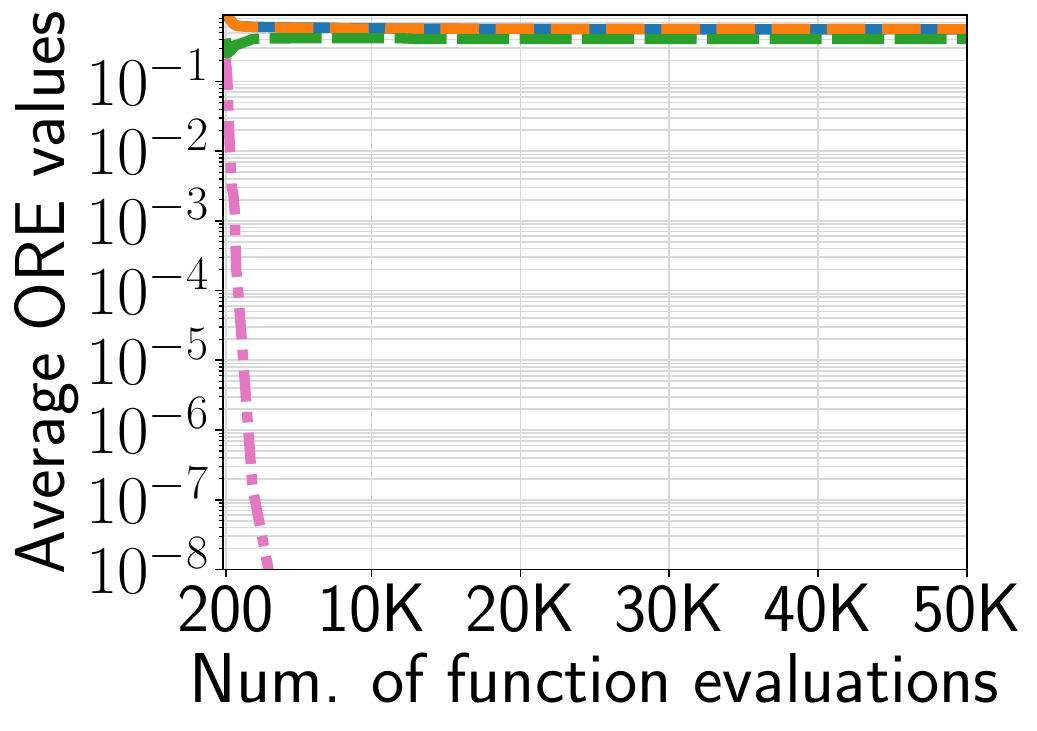}}
\\
\caption{Average $e^{\mathrm{ideal}}$, $e^{\mathrm{nadir}}$, and ORE values of the three normalization methods in MOEA/D-NUMS on SDTLZ4.}
\label{supfig:3error_MOEADNUMS_SDTLZ4}
\end{figure*}

\begin{figure*}[t]
\centering
  \subfloat{\includegraphics[width=0.7\textwidth]{./figs/legend/legend_3.pdf}}
\vspace{-3.9mm}
   \\
   \subfloat[$e^{\mathrm{ideal}}$ ($m=2$)]{\includegraphics[width=0.32\textwidth]{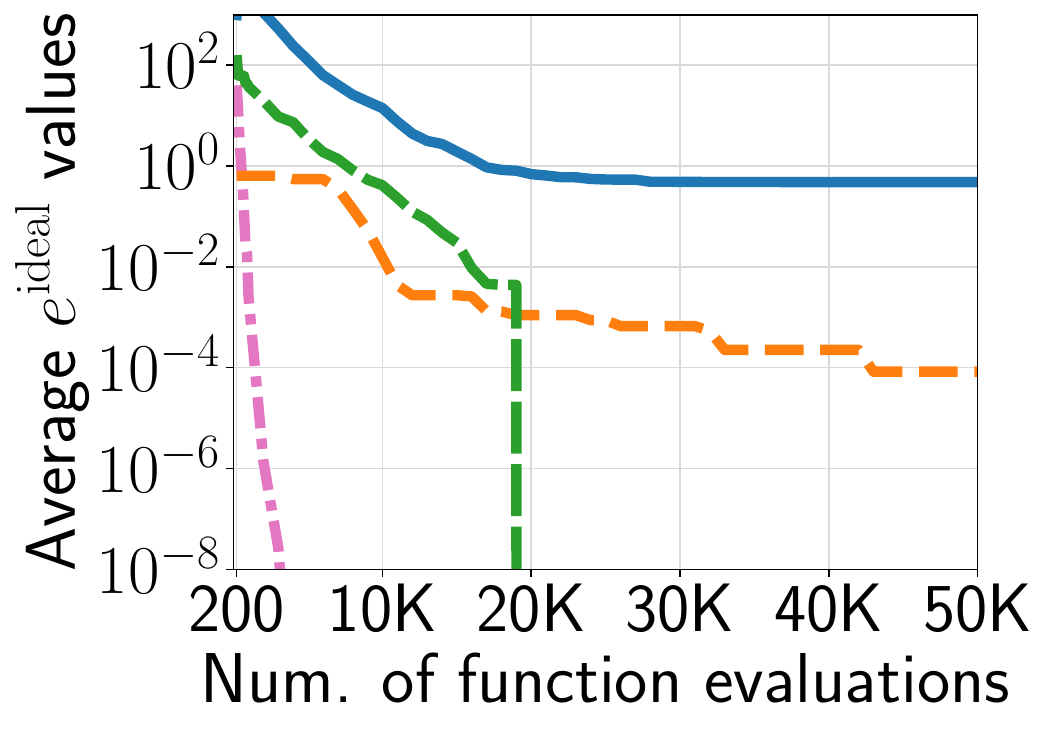}}
   \subfloat[$e^{\mathrm{ideal}}$ ($m=4$)]{\includegraphics[width=0.32\textwidth]{./figs/qi_error_ideal/MOEADNUMS_mu100/IDTLZ1_m4_r0.1_z-type1.pdf}}
   \subfloat[$e^{\mathrm{ideal}}$ ($m=6$)]{\includegraphics[width=0.32\textwidth]{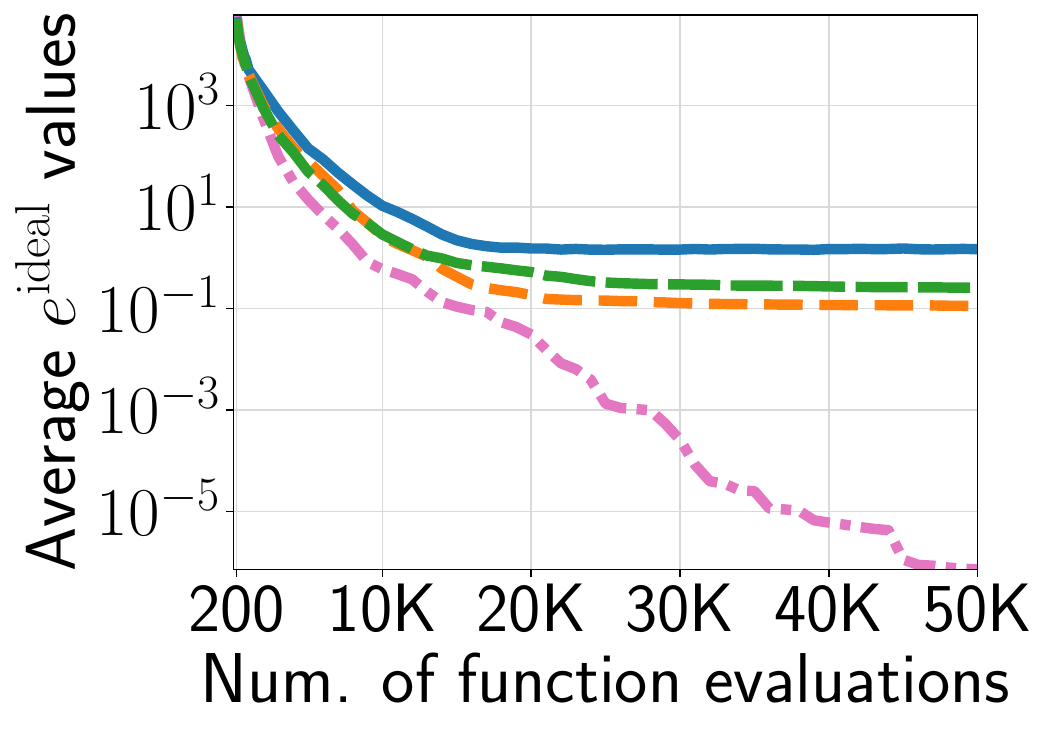}}
\\
   \subfloat[$e^{\mathrm{nadir}}$ ($m=2$)]{\includegraphics[width=0.32\textwidth]{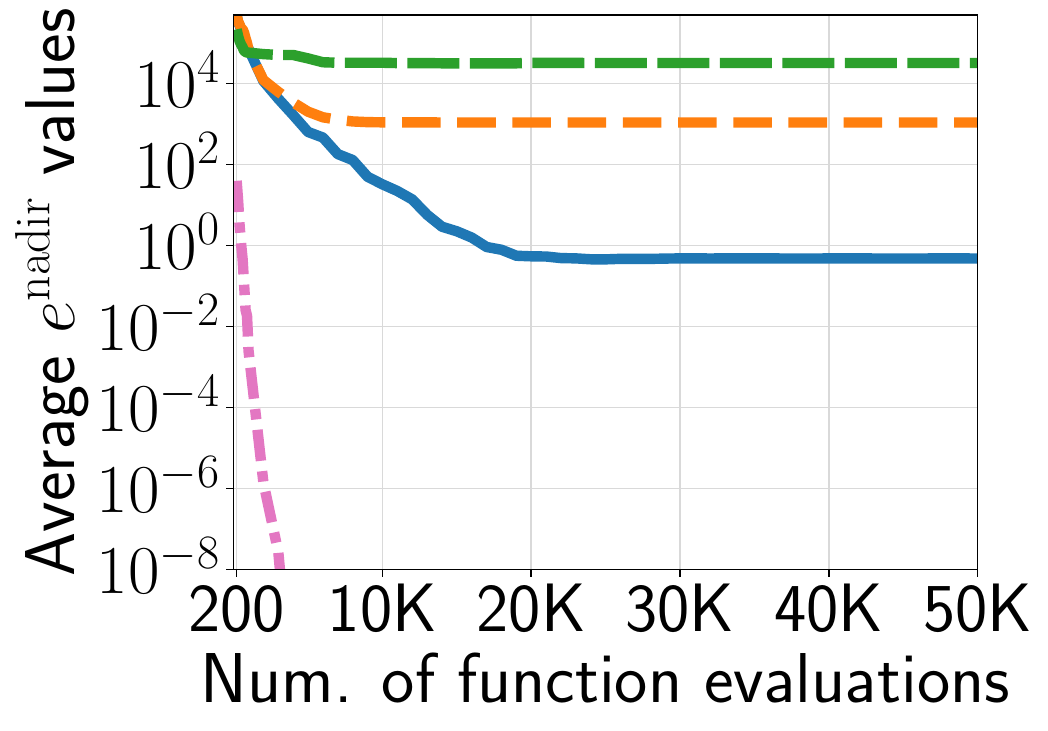}}
   \subfloat[$e^{\mathrm{nadir}}$ ($m=4$)]{\includegraphics[width=0.32\textwidth]{./figs/qi_error_nadir/MOEADNUMS_mu100/IDTLZ1_m4_r0.1_z-type1.pdf}}
   \subfloat[$e^{\mathrm{nadir}}$ ($m=6$)]{\includegraphics[width=0.32\textwidth]{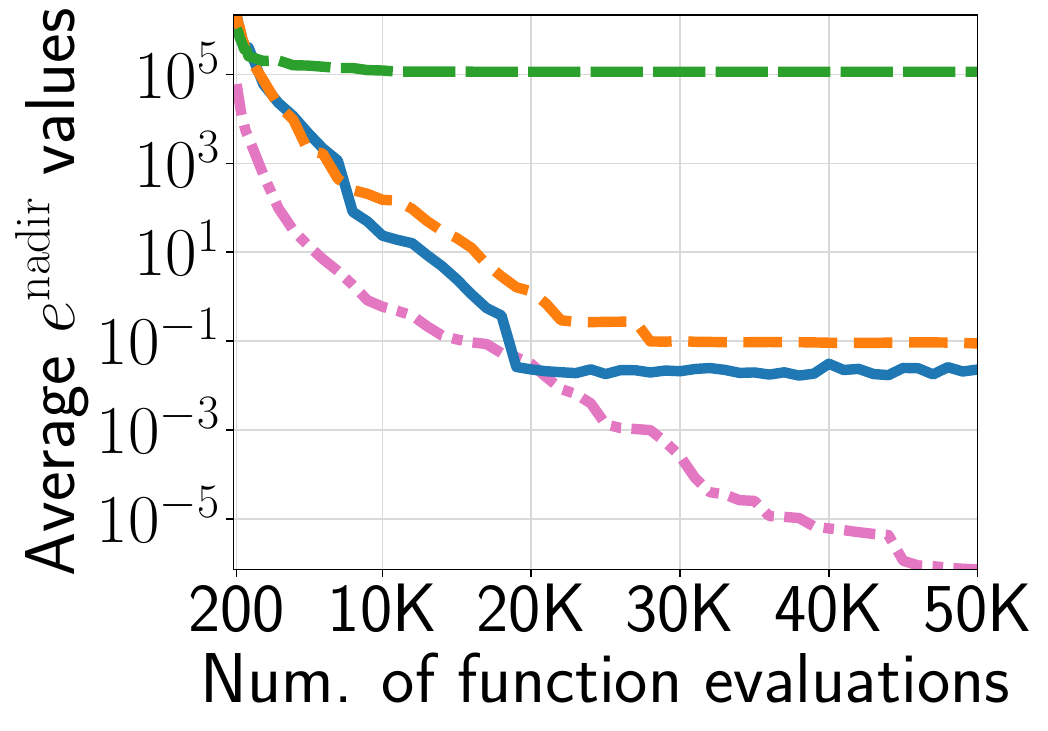}}
\\
   \subfloat[ORE ($m=2$)]{\includegraphics[width=0.32\textwidth]{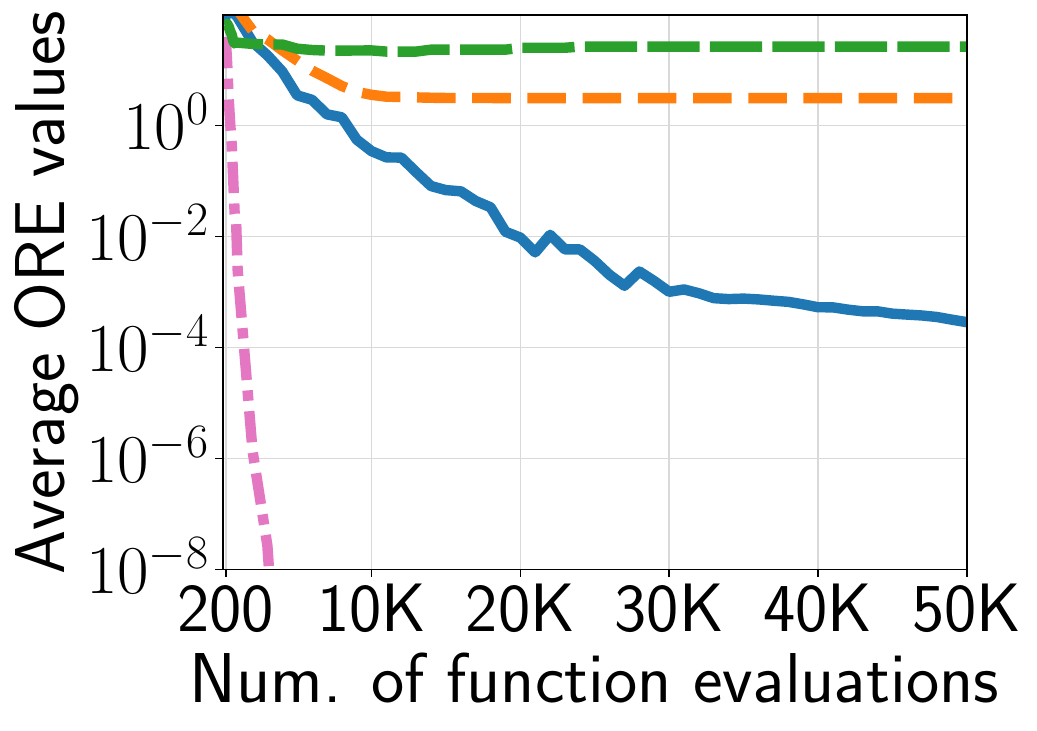}}
   \subfloat[ORE ($m=4$)]{\includegraphics[width=0.32\textwidth]{./figs/qi_ore/MOEADNUMS_mu100/IDTLZ1_m4_r0.1_z-type1.pdf}}
   \subfloat[ORE ($m=6$)]{\includegraphics[width=0.32\textwidth]{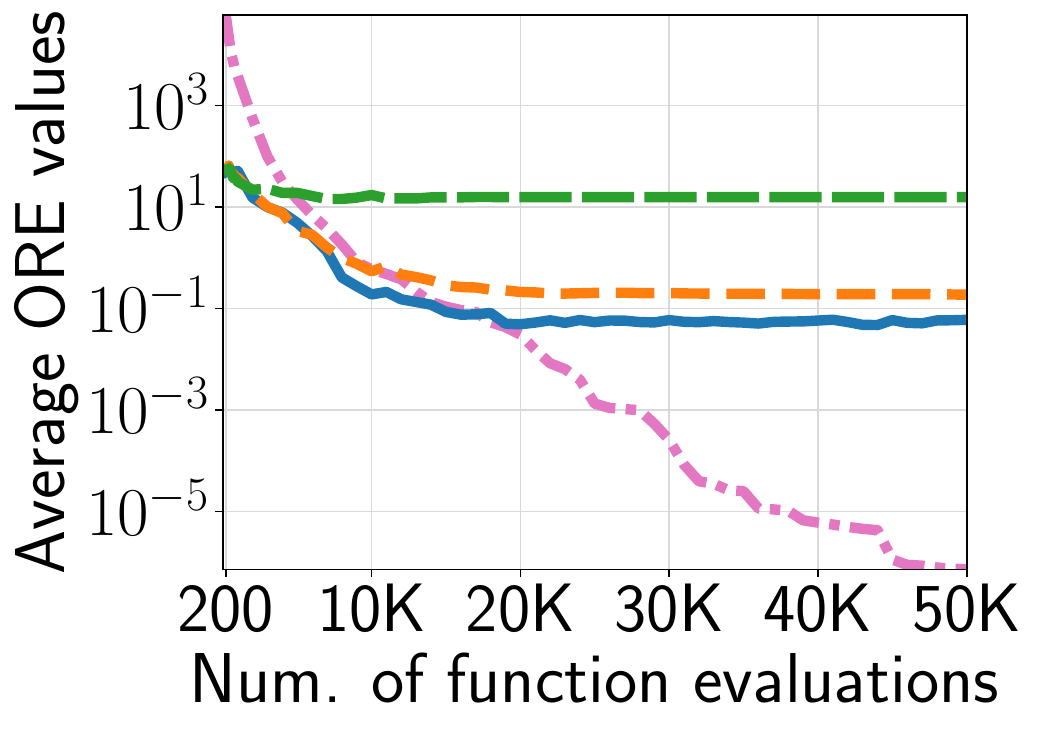}}
\\
\caption{Average $e^{\mathrm{ideal}}$, $e^{\mathrm{nadir}}$, and ORE values of the three normalization methods in MOEA/D-NUMS on IDTLZ1.}
\label{supfig:3error_MOEADNUMS_IDTLZ1}
\end{figure*}

\begin{figure*}[t]
\centering
  \subfloat{\includegraphics[width=0.7\textwidth]{./figs/legend/legend_3.pdf}}
\vspace{-3.9mm}
   \\
   \subfloat[$e^{\mathrm{ideal}}$ ($m=2$)]{\includegraphics[width=0.32\textwidth]{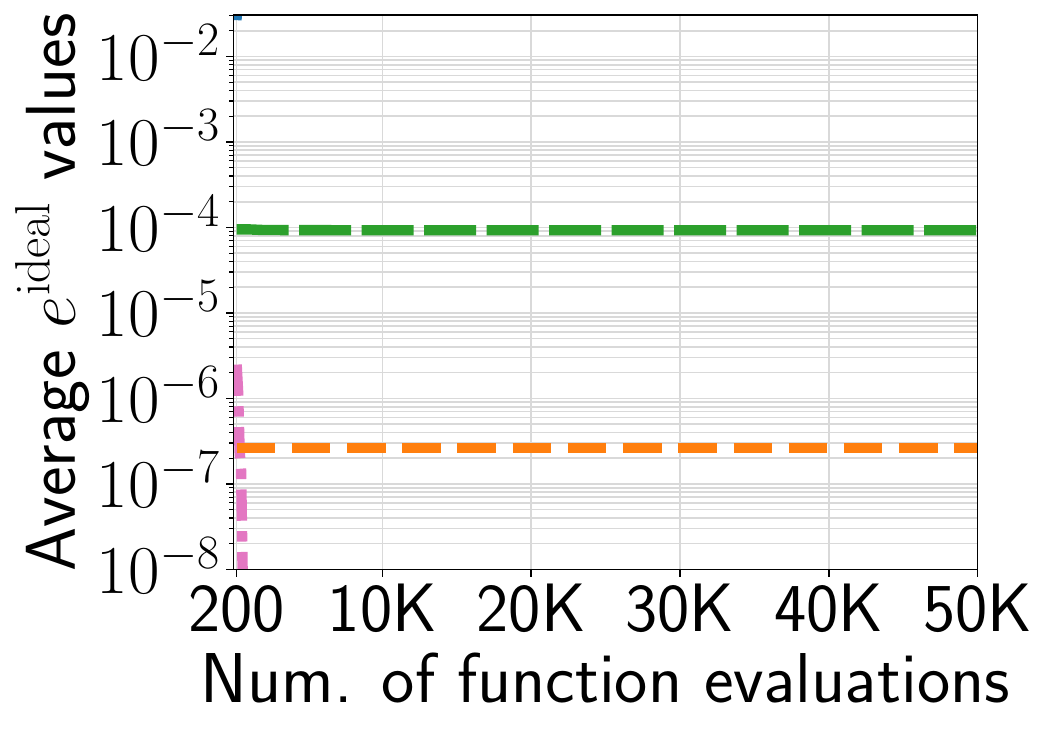}}
   \subfloat[$e^{\mathrm{ideal}}$ ($m=4$)]{\includegraphics[width=0.32\textwidth]{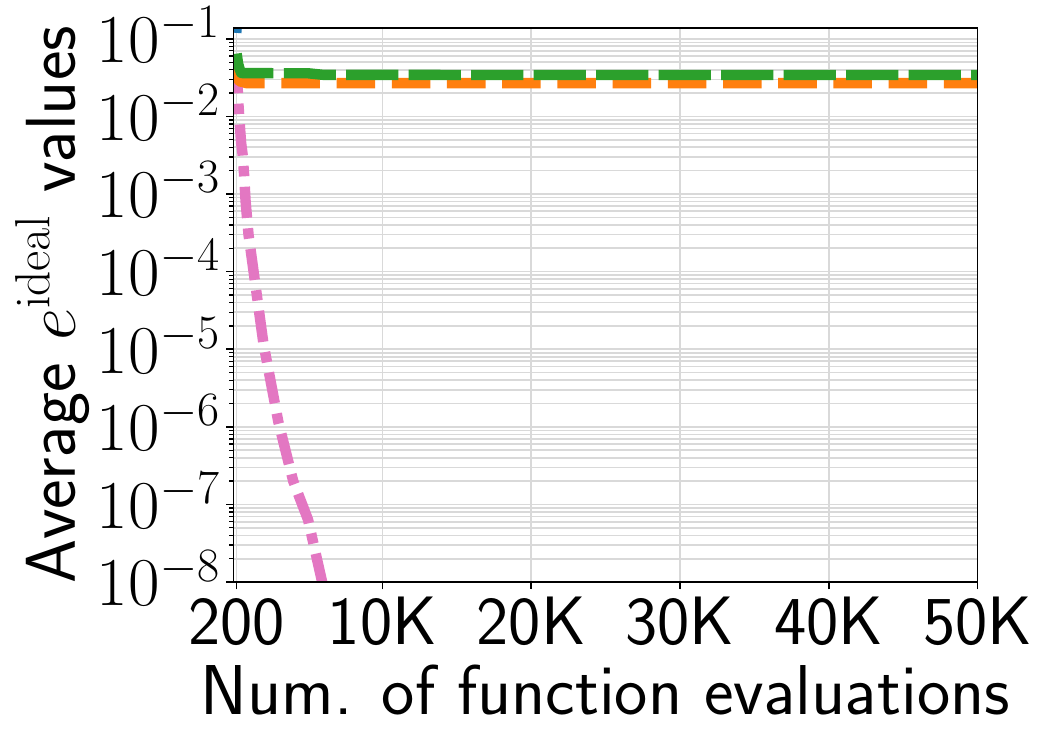}}
   \subfloat[$e^{\mathrm{ideal}}$ ($m=6$)]{\includegraphics[width=0.32\textwidth]{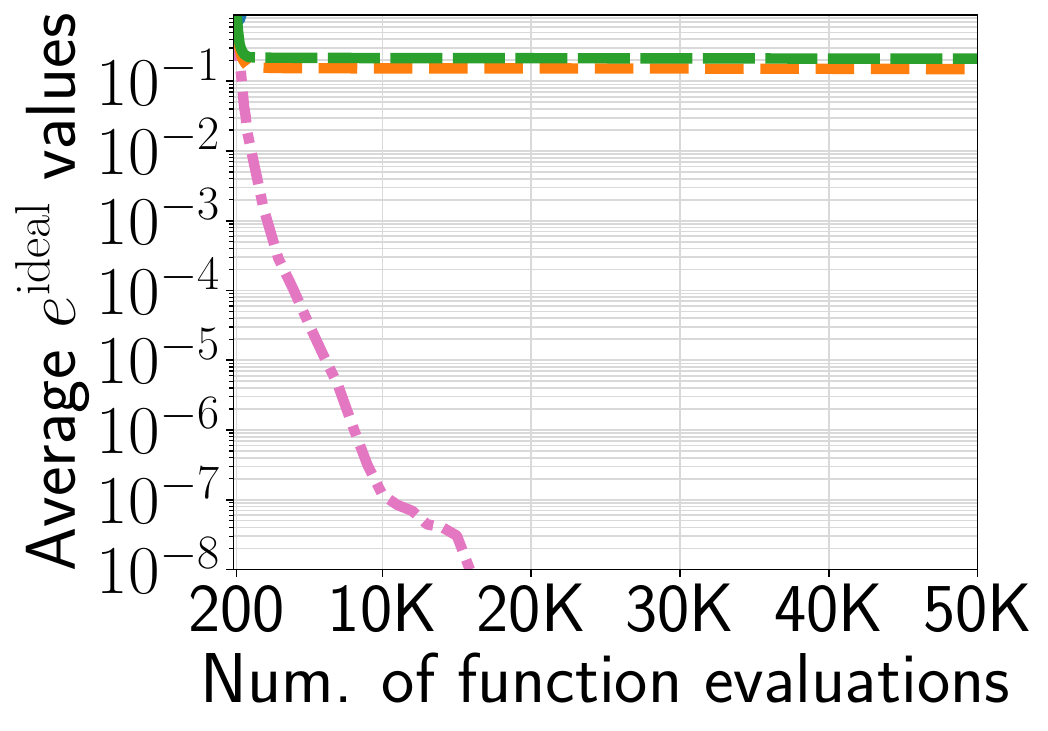}}
\\
   \subfloat[$e^{\mathrm{nadir}}$ ($m=2$)]{\includegraphics[width=0.32\textwidth]{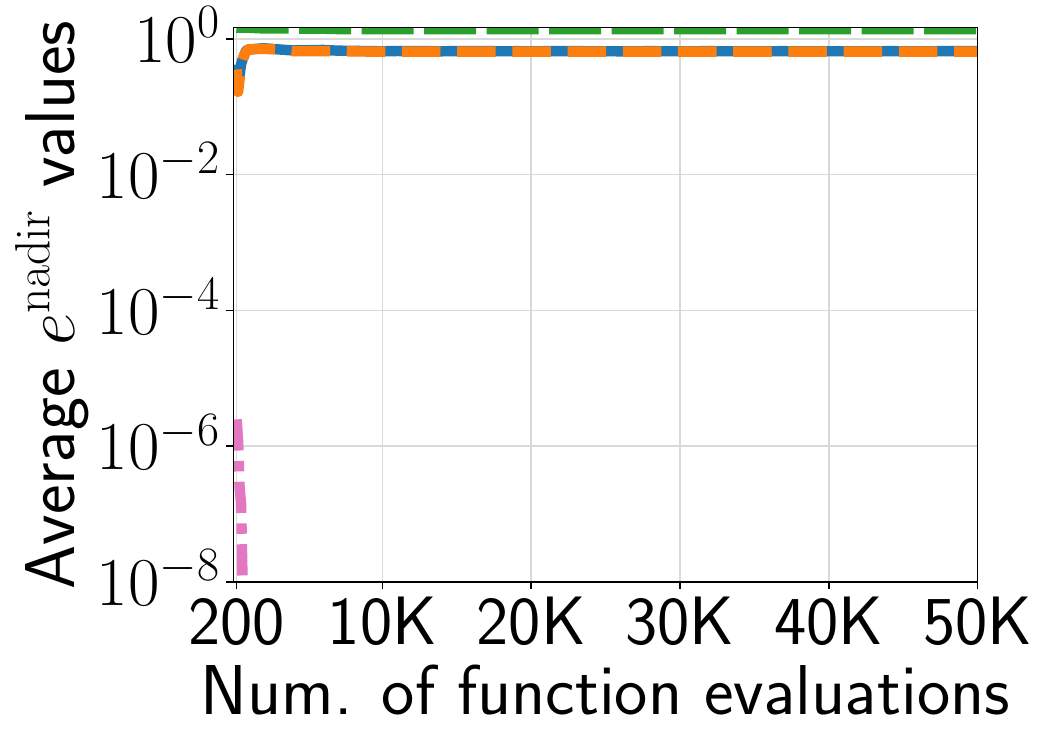}}
   \subfloat[$e^{\mathrm{nadir}}$ ($m=4$)]{\includegraphics[width=0.32\textwidth]{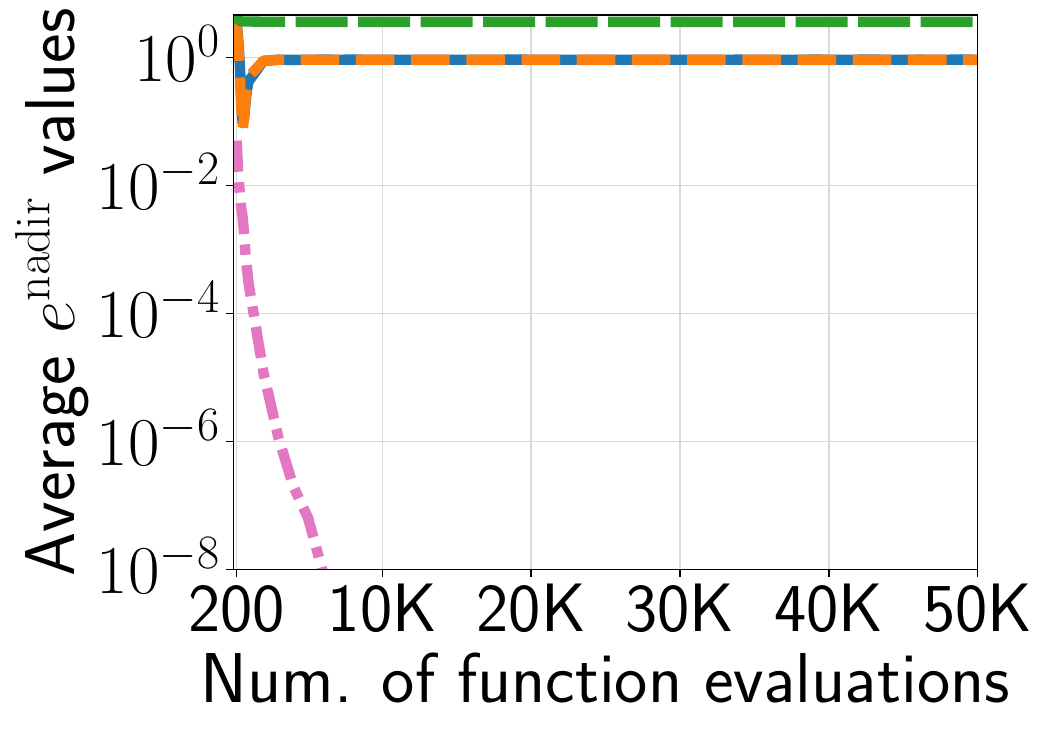}}
   \subfloat[$e^{\mathrm{nadir}}$ ($m=6$)]{\includegraphics[width=0.32\textwidth]{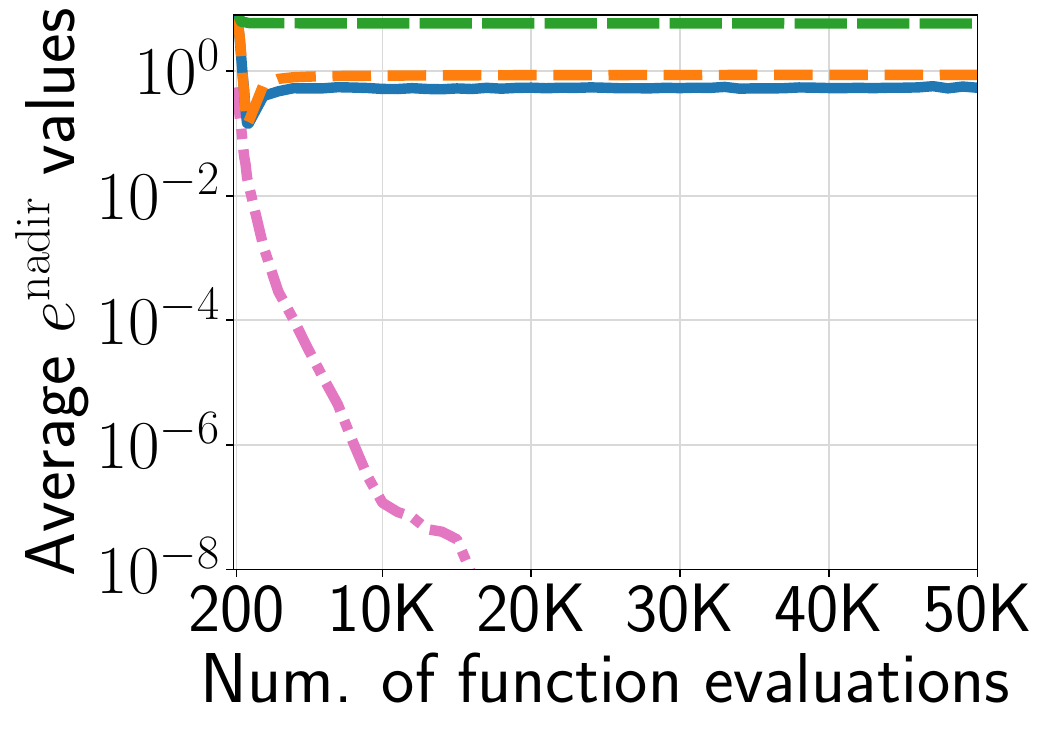}}
\\
   \subfloat[ORE ($m=2$)]{\includegraphics[width=0.32\textwidth]{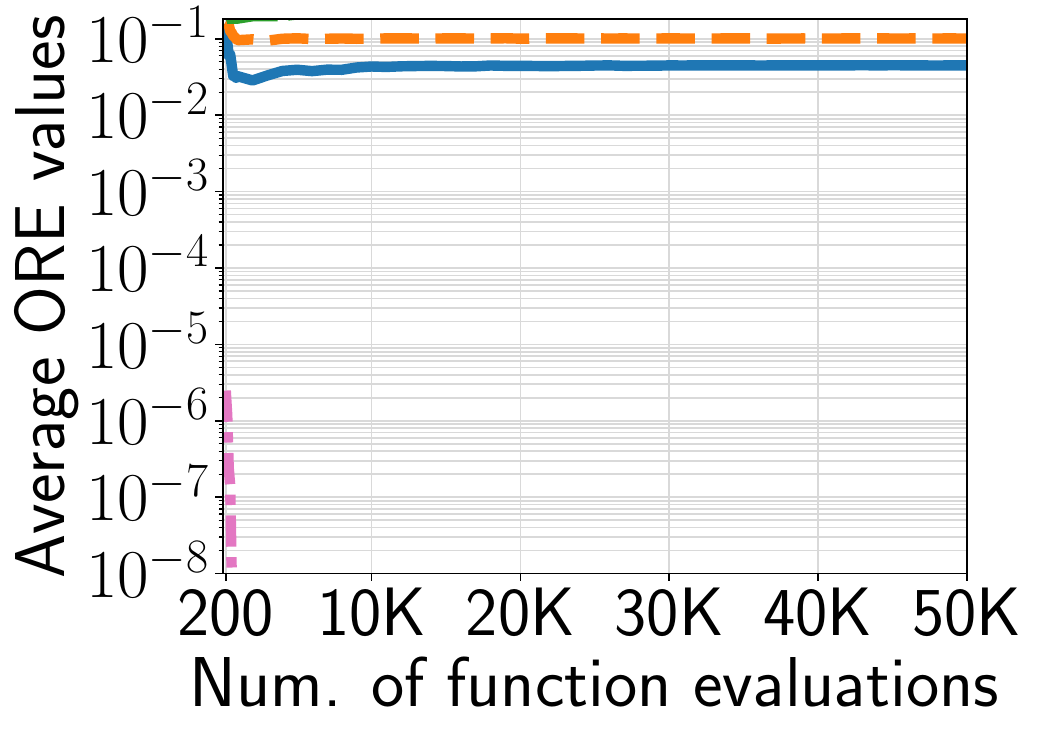}}
   \subfloat[ORE ($m=4$)]{\includegraphics[width=0.32\textwidth]{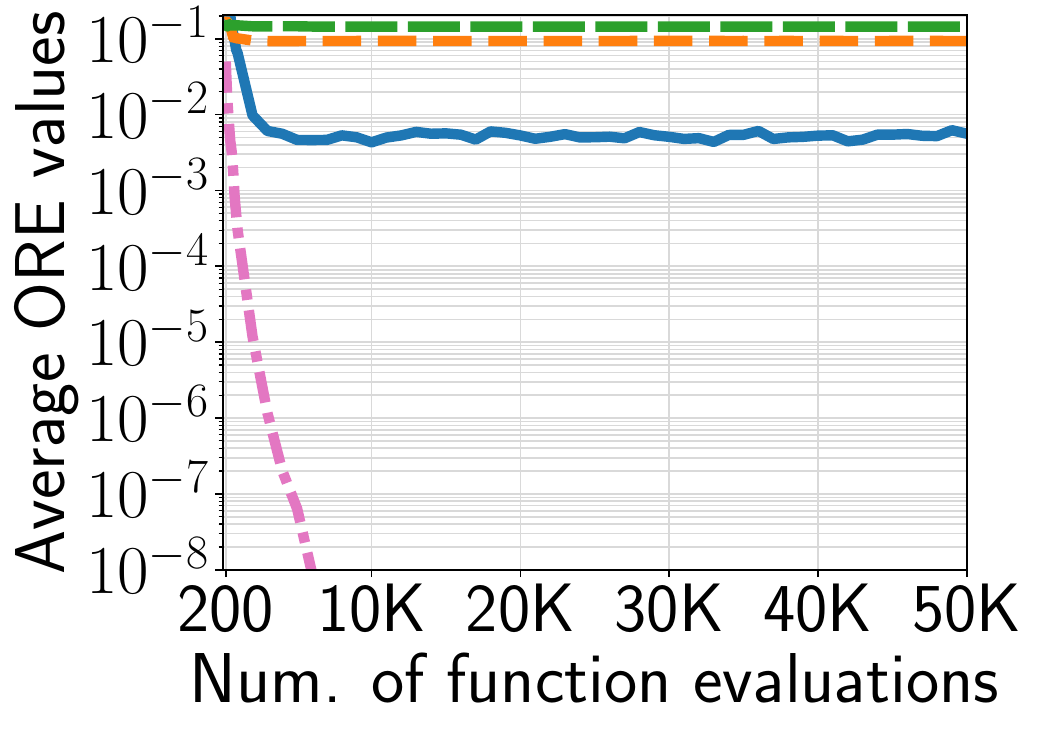}}
   \subfloat[ORE ($m=6$)]{\includegraphics[width=0.32\textwidth]{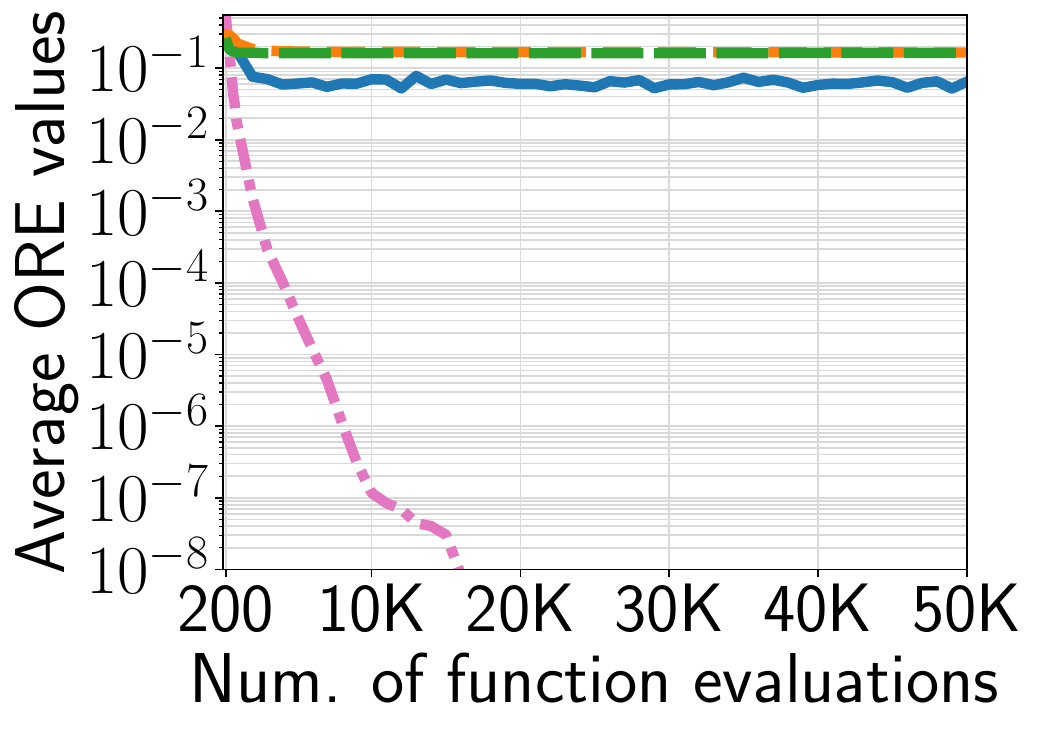}}
\\
\caption{Average $e^{\mathrm{ideal}}$, $e^{\mathrm{nadir}}$, and ORE values of the three normalization methods in MOEA/D-NUMS on IDTLZ2.}
\label{supfig:3error_MOEADNUMS_IDTLZ2}
\end{figure*}

\begin{figure*}[t]
\centering
  \subfloat{\includegraphics[width=0.7\textwidth]{./figs/legend/legend_3.pdf}}
\vspace{-3.9mm}
   \\
   \subfloat[$e^{\mathrm{ideal}}$ ($m=2$)]{\includegraphics[width=0.32\textwidth]{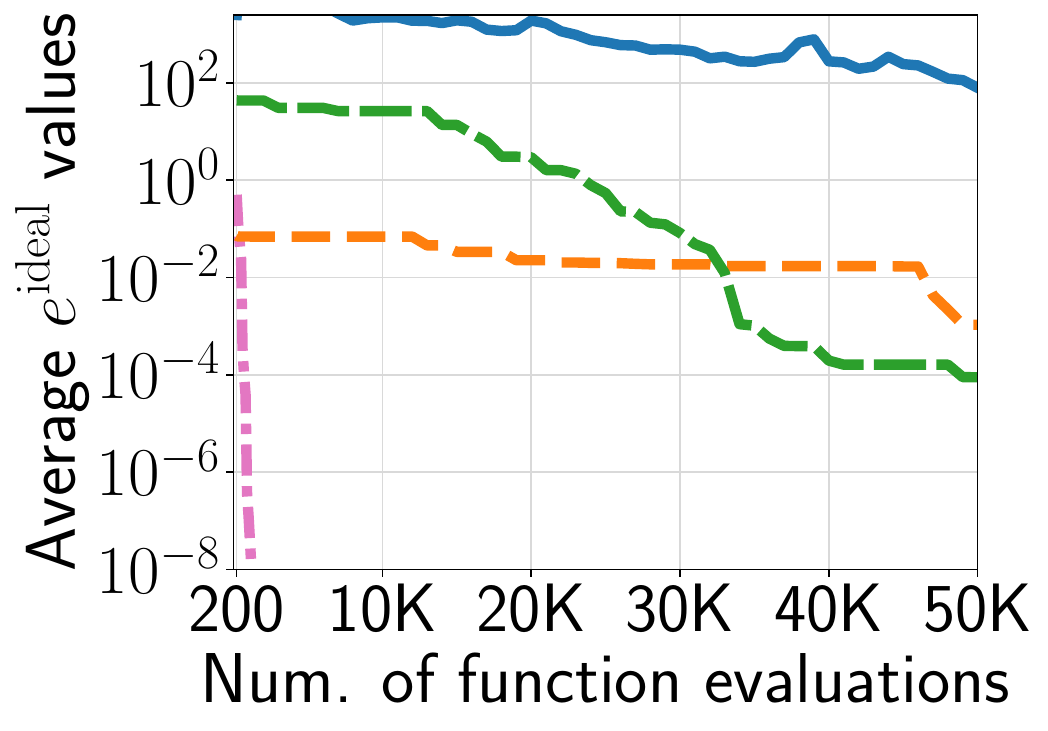}}
   \subfloat[$e^{\mathrm{ideal}}$ ($m=4$)]{\includegraphics[width=0.32\textwidth]{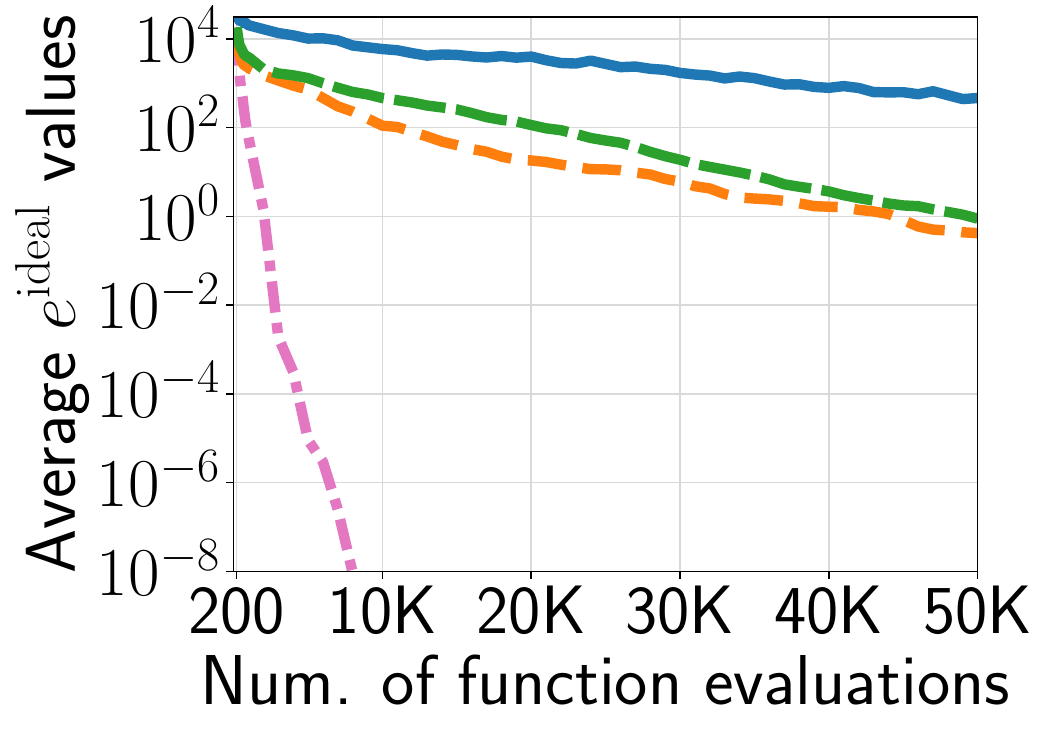}}
   \subfloat[$e^{\mathrm{ideal}}$ ($m=6$)]{\includegraphics[width=0.32\textwidth]{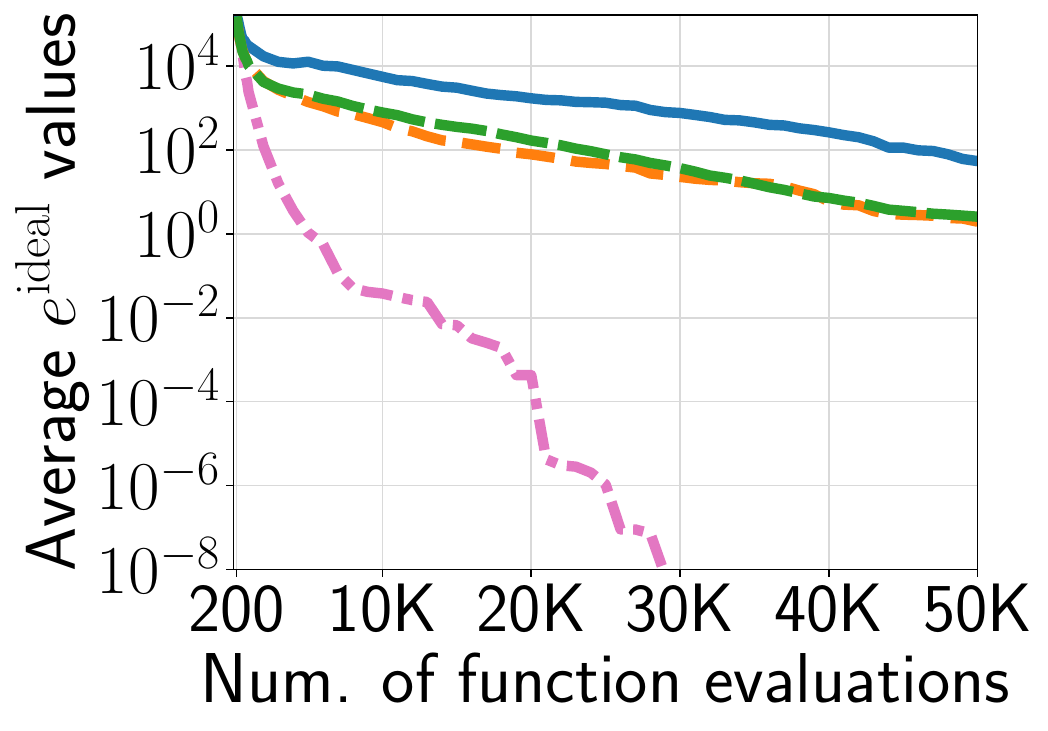}}
\\
   \subfloat[$e^{\mathrm{nadir}}$ ($m=2$)]{\includegraphics[width=0.32\textwidth]{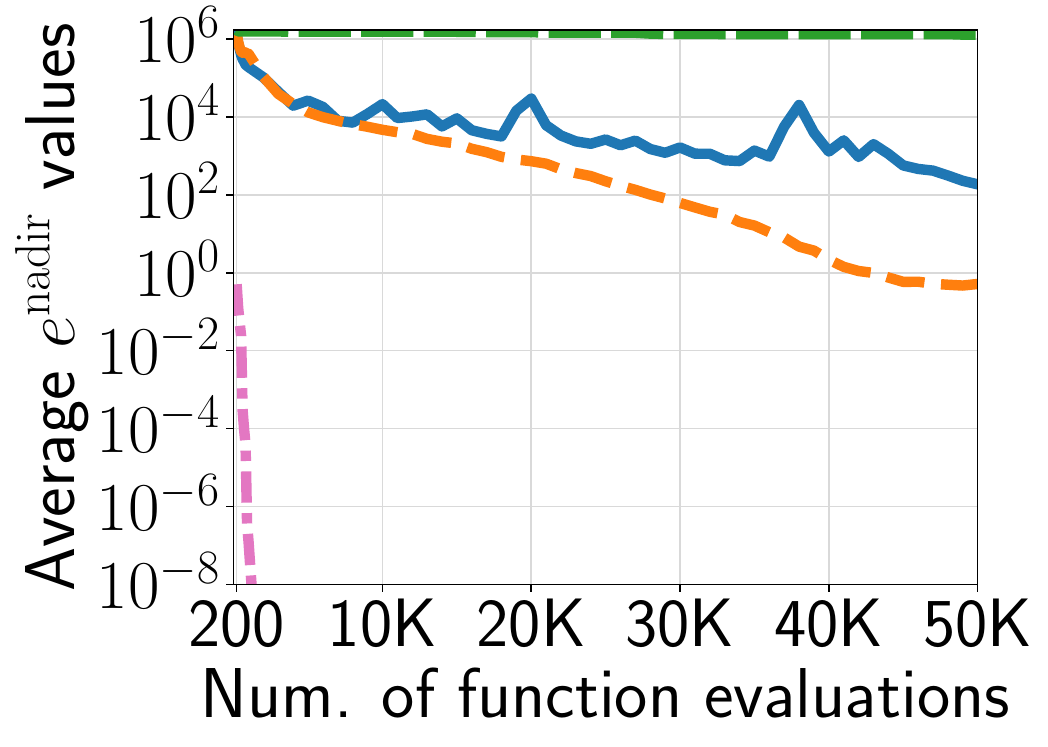}}
   \subfloat[$e^{\mathrm{nadir}}$ ($m=4$)]{\includegraphics[width=0.32\textwidth]{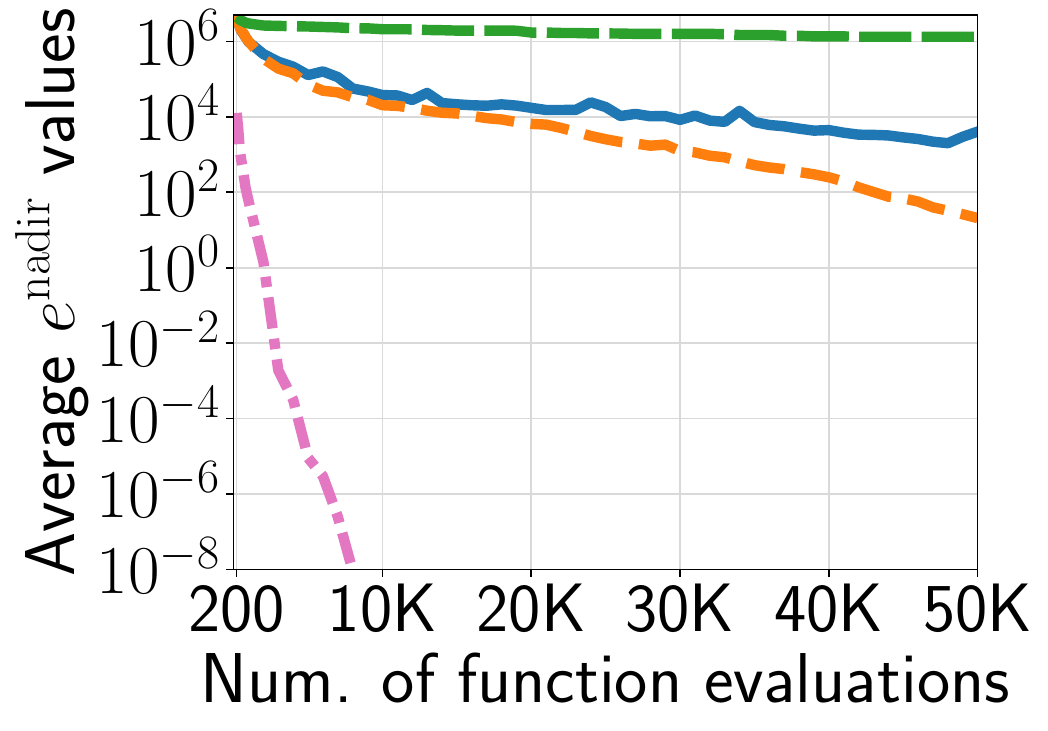}}
   \subfloat[$e^{\mathrm{nadir}}$ ($m=6$)]{\includegraphics[width=0.32\textwidth]{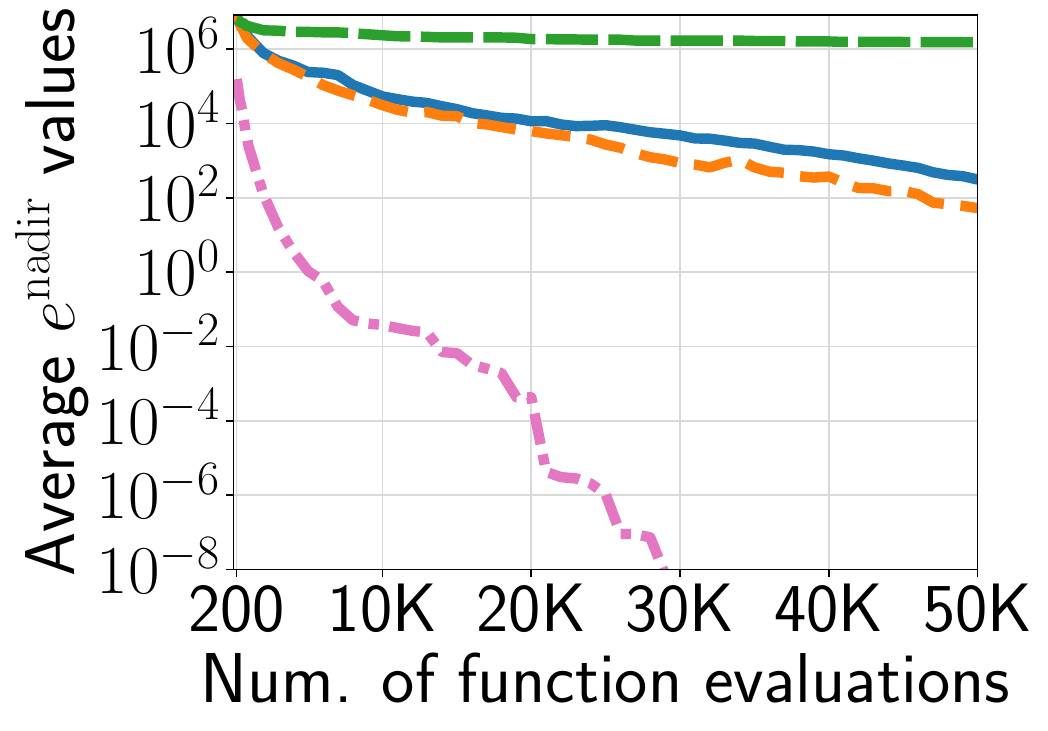}}
\\
   \subfloat[ORE ($m=2$)]{\includegraphics[width=0.32\textwidth]{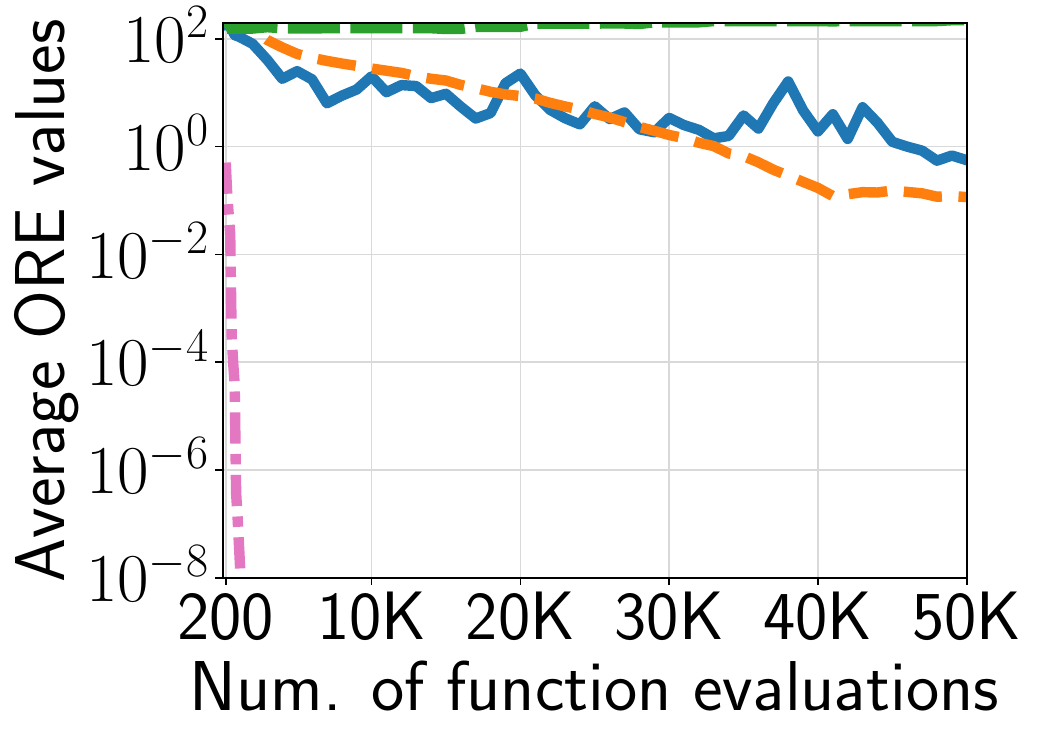}}
   \subfloat[ORE ($m=4$)]{\includegraphics[width=0.32\textwidth]{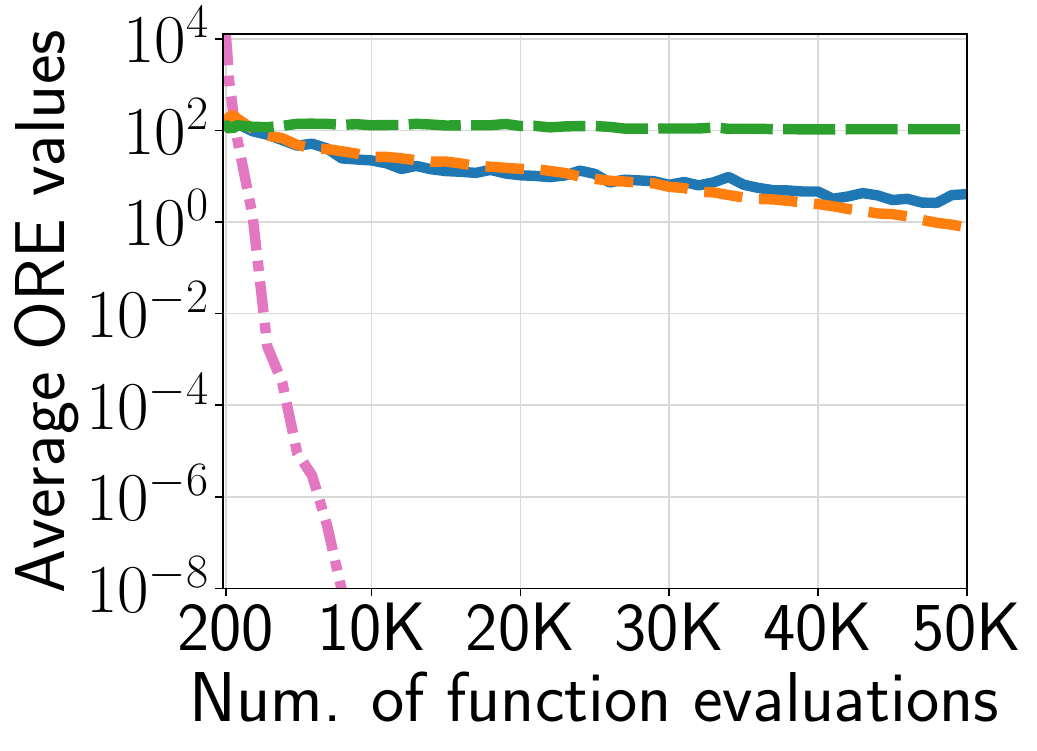}}
   \subfloat[ORE ($m=6$)]{\includegraphics[width=0.32\textwidth]{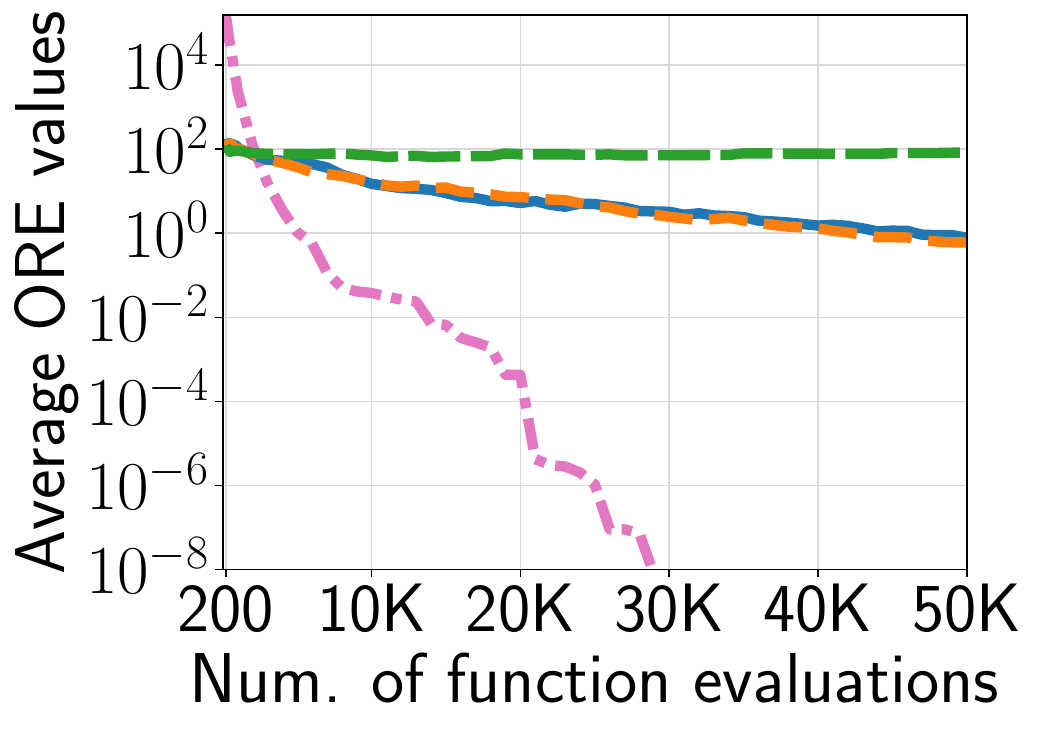}}
\\
\caption{Average $e^{\mathrm{ideal}}$, $e^{\mathrm{nadir}}$, and ORE values of the three normalization methods in MOEA/D-NUMS on IDTLZ3.}
\label{supfig:3error_MOEADNUMS_IDTLZ3}
\end{figure*}

\begin{figure*}[t]
\centering
  \subfloat{\includegraphics[width=0.7\textwidth]{./figs/legend/legend_3.pdf}}
\vspace{-3.9mm}
   \\
   \subfloat[$e^{\mathrm{ideal}}$ ($m=2$)]{\includegraphics[width=0.32\textwidth]{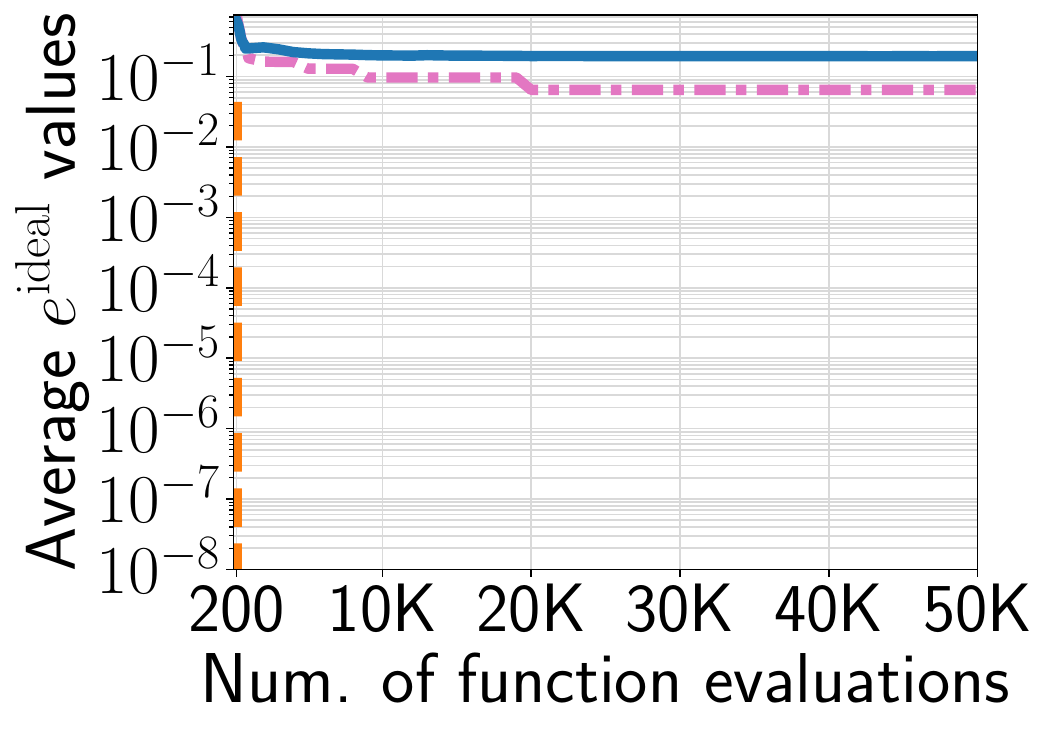}}
   \subfloat[$e^{\mathrm{ideal}}$ ($m=4$)]{\includegraphics[width=0.32\textwidth]{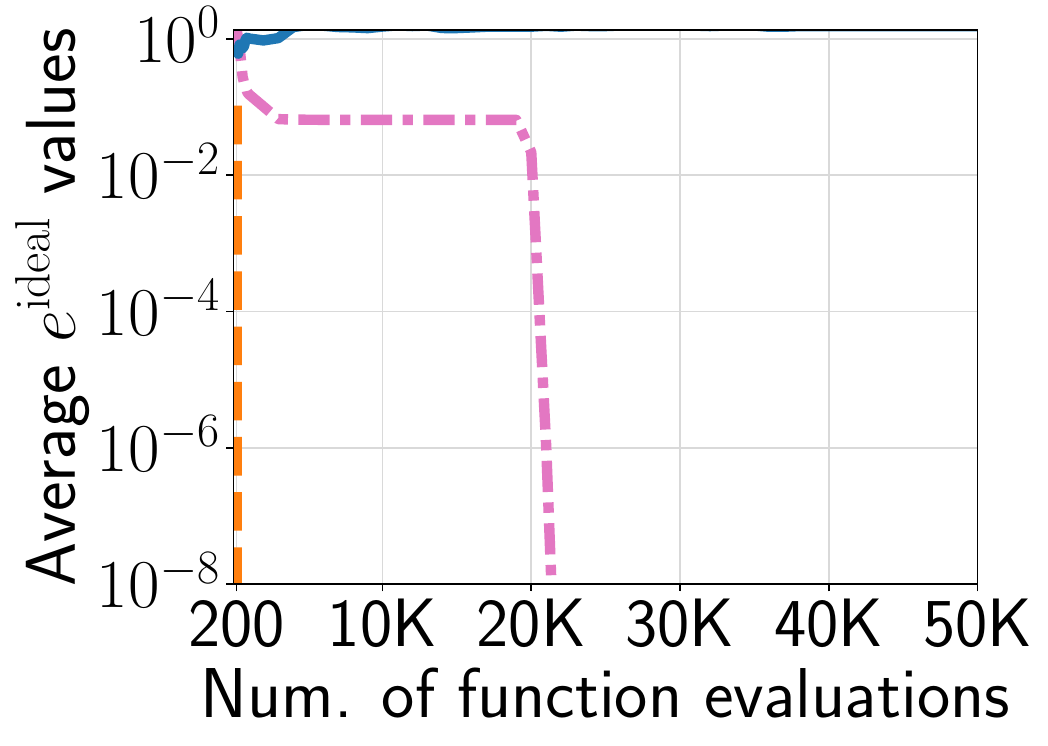}}
   \subfloat[$e^{\mathrm{ideal}}$ ($m=6$)]{\includegraphics[width=0.32\textwidth]{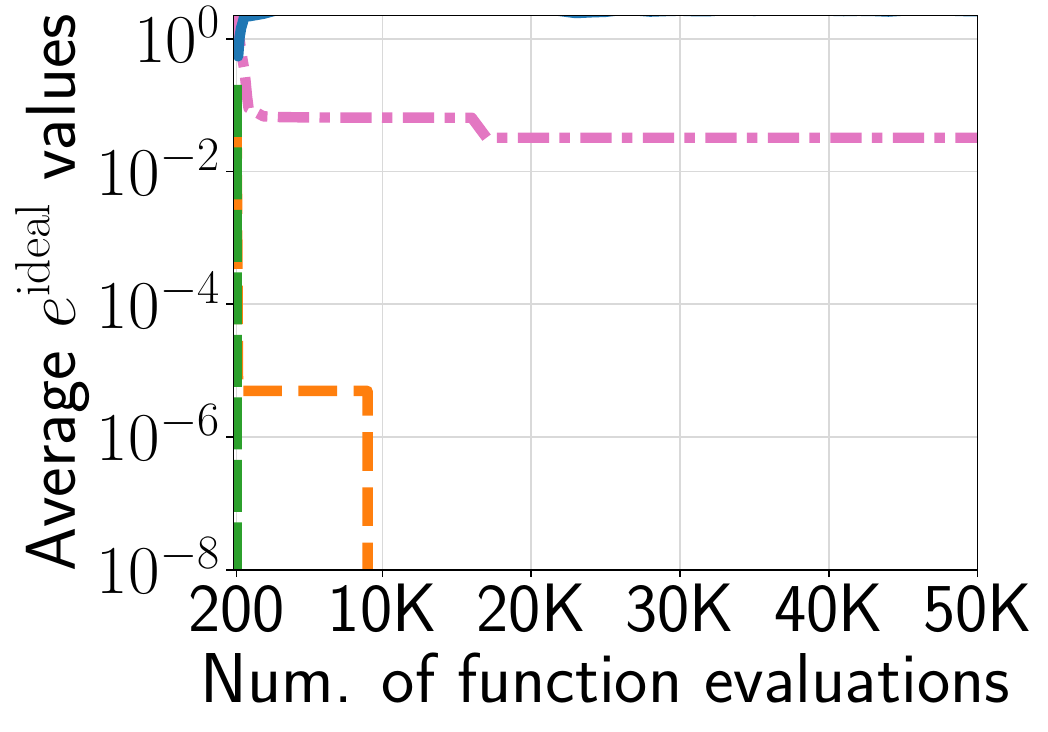}}
\\
   \subfloat[$e^{\mathrm{nadir}}$ ($m=2$)]{\includegraphics[width=0.32\textwidth]{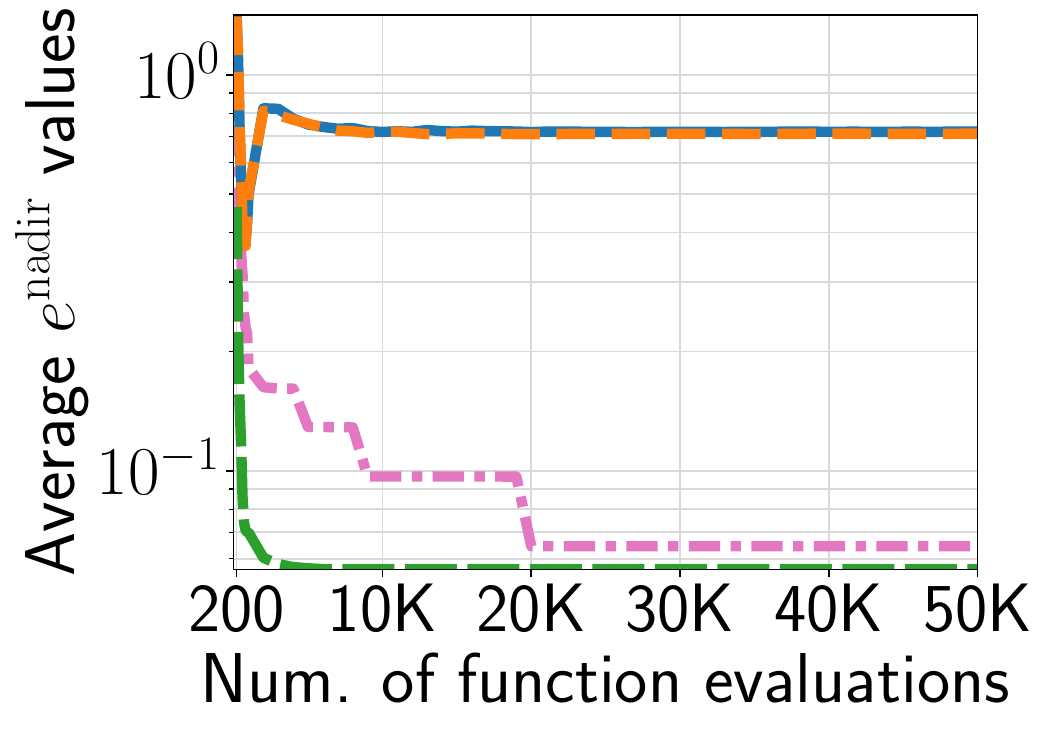}}
   \subfloat[$e^{\mathrm{nadir}}$ ($m=4$)]{\includegraphics[width=0.32\textwidth]{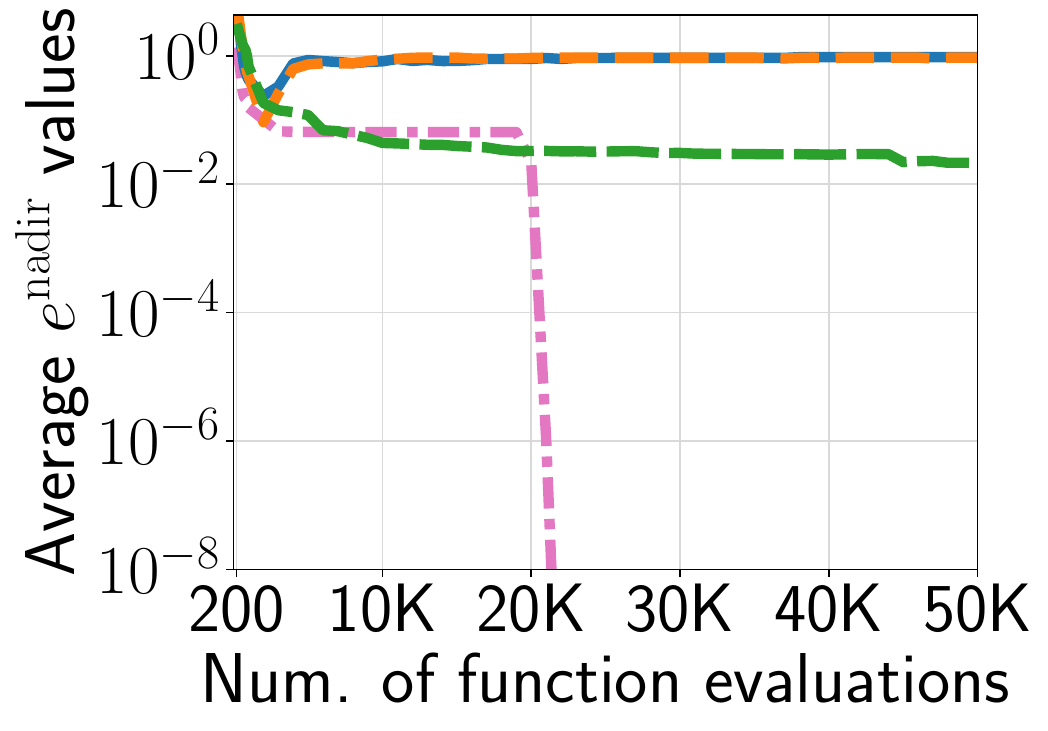}}
   \subfloat[$e^{\mathrm{nadir}}$ ($m=6$)]{\includegraphics[width=0.32\textwidth]{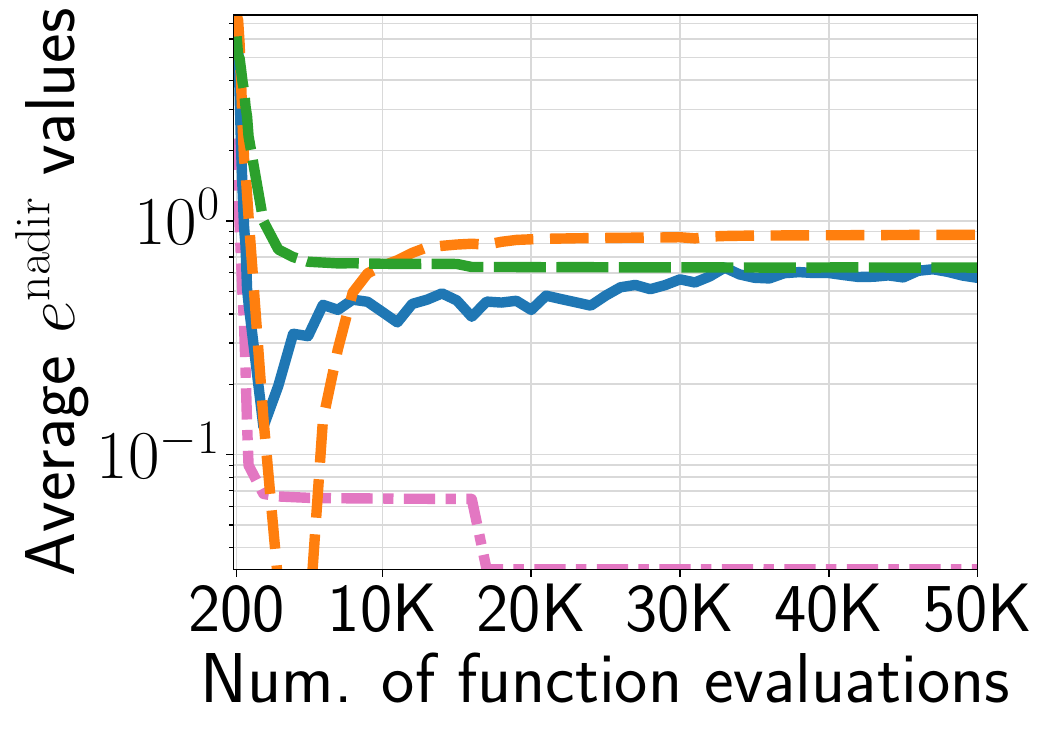}}
\\
   \subfloat[ORE ($m=2$)]{\includegraphics[width=0.32\textwidth]{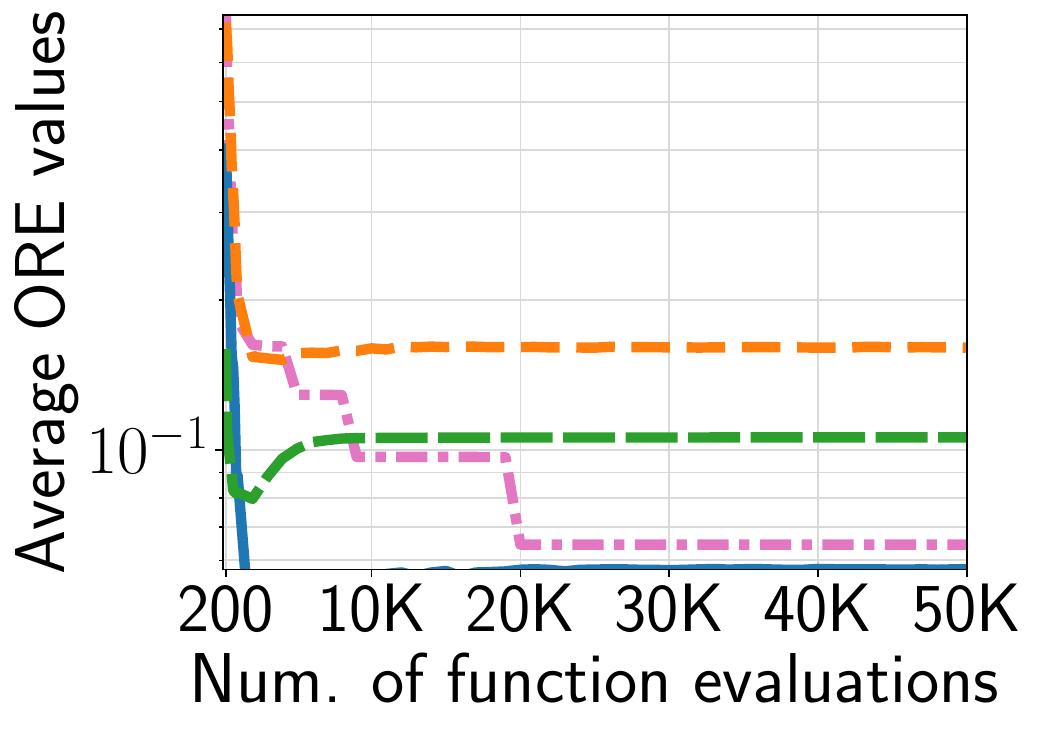}}
   \subfloat[ORE ($m=4$)]{\includegraphics[width=0.32\textwidth]{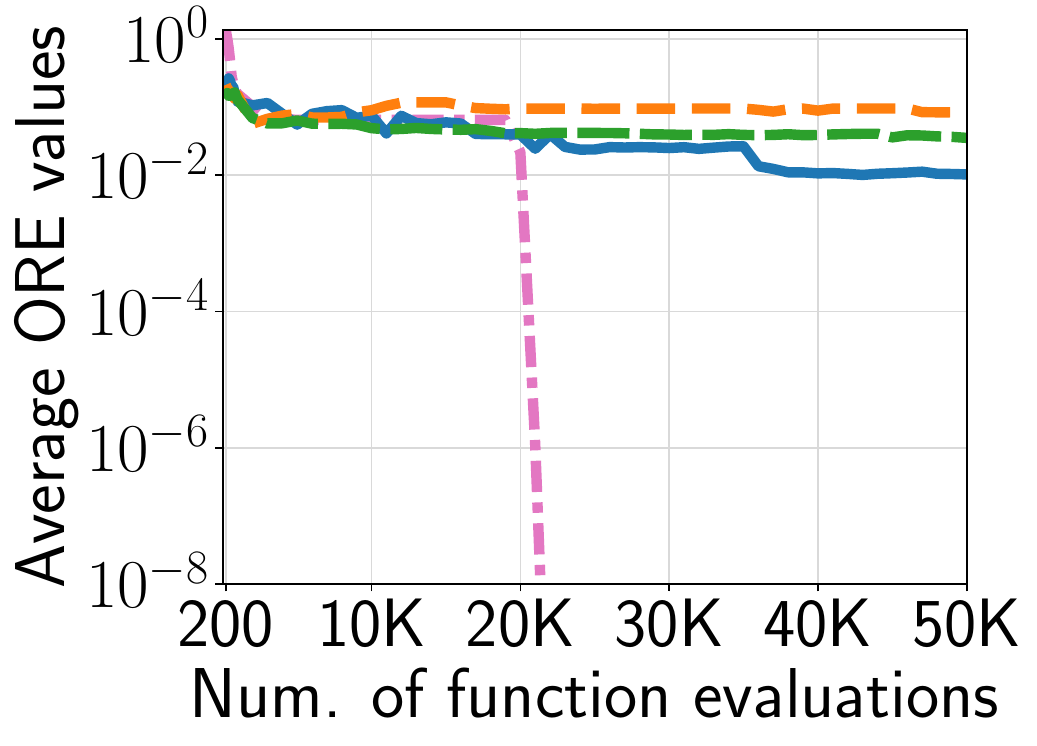}}
   \subfloat[ORE ($m=6$)]{\includegraphics[width=0.32\textwidth]{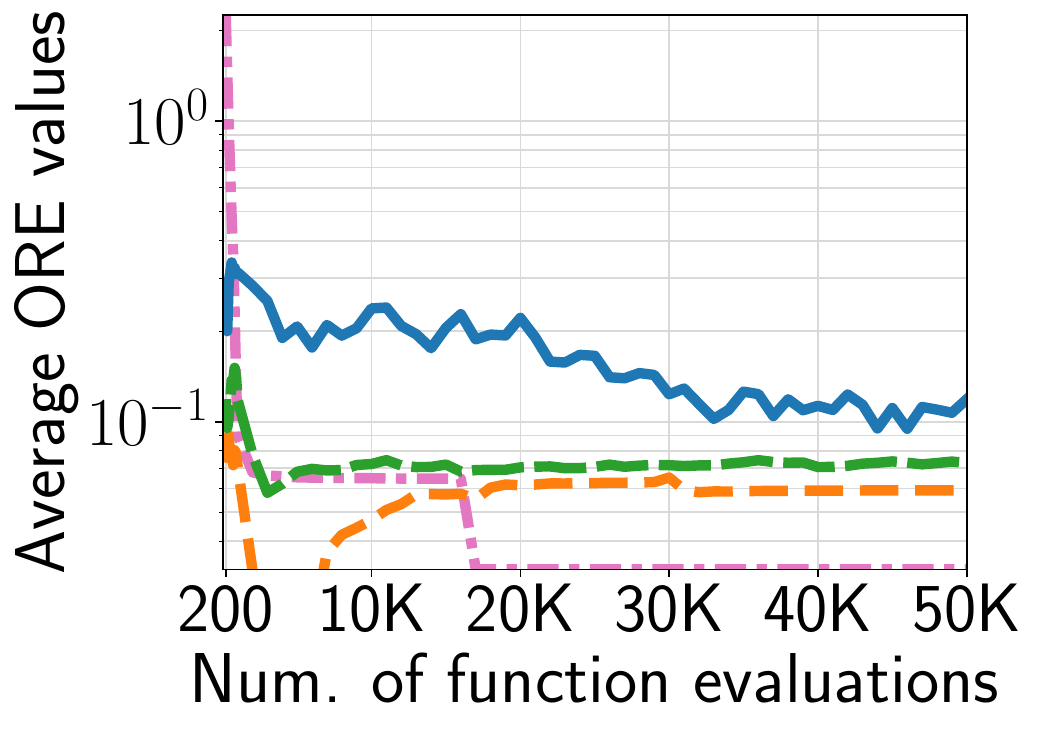}}
\\
\caption{Average $e^{\mathrm{ideal}}$, $e^{\mathrm{nadir}}$, and ORE values of the three normalization methods in MOEA/D-NUMS on IDTLZ4.}
\label{supfig:3error_MOEADNUMS_IDTLZ4}
\end{figure*}

\begin{table*}[t]
  \renewcommand{\arraystretch}{0.2} 
\centering
\caption{\small Average IGD$^+$-C values of R-NSGA-II with the three normalization methods (PP, BP, and BA) and with no normalization method (NO) on the DTLZ, SDTLZ, and IDTLZ problems with $m \in \{2, 3, 4, 5, 6\}$. The results at $50\,000$ function evaluations are shown. The numbers in parentheses indicate the rankings based on the Friedman test. The best and second best data are highlighted by dark gray and gray, respectively.}
  \label{tab:sup_rnsga2_type1}  
{\scriptsize
  \begin{tabular}{cccccccc}
\toprule
Problem & $m$ & PP & BP & BA & NO\\
\midrule
DTLZ1 & 2 & \cellcolor{c2}0.0191 (2) & 0.0624 (3) & 0.0954 (4) & \cellcolor{c1}0.0012 (1)\\
DTLZ1 & 3 & 0.0354 (4) & 0.0291 (3) & \cellcolor{c2}0.0274 (2) & \cellcolor{c1}0.0078 (1)\\
DTLZ1 & 4 & 0.0582 (3) & \cellcolor{c2}0.0570 (2) & 0.0932 (4) & \cellcolor{c1}0.0242 (1)\\
DTLZ1 & 5 & \cellcolor{c2}0.0791 (2) & 0.0808 (3) & 0.1219 (4) & \cellcolor{c1}0.0424 (1)\\
DTLZ1 & 6 & \cellcolor{c2}0.1441 (2) & 0.1512 (3) & 0.4296 (4) & \cellcolor{c1}0.0815 (1)\\
\midrule
DTLZ2 & 2 & 0.0753 (4) & 0.0134 (3) & \cellcolor{c2}0.0015 (2) & \cellcolor{c1}0.0011 (1)\\
DTLZ2 & 3 & 0.0330 (4) & 0.0174 (3) & \cellcolor{c1}0.0090 (1) & \cellcolor{c2}0.0138 (2)\\
DTLZ2 & 4 & 0.0445 (3) & 0.0458 (4) & \cellcolor{c1}0.0249 (1) & \cellcolor{c2}0.0373 (2)\\
DTLZ2 & 5 & 0.1194 (4) & 0.1130 (3) & \cellcolor{c1}0.0680 (1) & \cellcolor{c2}0.0789 (2)\\
DTLZ2 & 6 & 0.1710 (4) & 0.1657 (3) & \cellcolor{c1}0.1056 (1) & \cellcolor{c2}0.1300 (2)\\
\midrule
DTLZ3 & 2 & 0.0943 (4) & 0.0144 (3) & \cellcolor{c2}0.0116 (2) & \cellcolor{c1}0.0069 (1)\\
DTLZ3 & 3 & 0.0501 (3) & \cellcolor{c2}0.0379 (2) & 0.1468 (4) & \cellcolor{c1}0.0235 (1)\\
DTLZ3 & 4 & 0.1619 (3) & \cellcolor{c2}0.0617 (2) & 2.0146 (4) & \cellcolor{c1}0.0550 (1)\\
DTLZ3 & 5 & 0.1160 (3) & \cellcolor{c2}0.1011 (2) & 0.7749 (4) & \cellcolor{c1}0.0967 (1)\\
DTLZ3 & 6 & \cellcolor{c2}0.1436 (2) & 0.2422 (3) & 1.5421 (4) & \cellcolor{c1}0.1271 (1)\\
\midrule
DTLZ4 & 2 & 0.2244 (4) & 0.1928 (3) & \cellcolor{c2}0.1880 (2) & \cellcolor{c1}0.1745 (1)\\
DTLZ4 & 3 & \cellcolor{c2}0.0673 (2) & 0.0939 (4) & \cellcolor{c1}0.0455 (1) & 0.0782 (3)\\
DTLZ4 & 4 & \cellcolor{c1}0.0614 (1) & \cellcolor{c2}0.0637 (2) & 0.0657 (3) & 0.0759 (4)\\
DTLZ4 & 5 & \cellcolor{c2}0.0993 (2) & 0.1103 (4) & \cellcolor{c1}0.0654 (1) & 0.1062 (3)\\
DTLZ4 & 6 & 0.1520 (4) & 0.1403 (3) & \cellcolor{c1}0.0820 (1) & \cellcolor{c2}0.1063 (2)\\
\midrule
DTLZ5 & 2 & 0.0656 (4) & 0.0373 (3) & \cellcolor{c2}0.0018 (2) & \cellcolor{c1}0.0008 (1)\\
DTLZ5 & 3 & 0.0178 (4) & \cellcolor{c1}0.0024 (1) & \cellcolor{c2}0.0028 (2) & 0.0031 (3)\\
DTLZ5 & 4 & 0.0728 (4) & \cellcolor{c1}0.0422 (1) & \cellcolor{c2}0.0457 (2) & 0.0518 (3)\\
DTLZ5 & 5 & \cellcolor{c1}0.1749 (1) & \cellcolor{c2}0.2352 (2) & 0.2454 (3) & 0.2707 (4)\\
DTLZ5 & 6 & 0.0670 (3) & 0.0743 (4) & \cellcolor{c1}0.0385 (1) & \cellcolor{c2}0.0473 (2)\\
\midrule
DTLZ6 & 2 & 0.1566 (4) & 0.1520 (3) & \cellcolor{c2}0.1169 (2) & \cellcolor{c1}0.1056 (1)\\
DTLZ6 & 3 & 0.1398 (3) & 0.1412 (4) & \cellcolor{c2}0.1367 (2) & \cellcolor{c1}0.1283 (1)\\
DTLZ6 & 4 & 0.2047 (4) & 0.1975 (3) & \cellcolor{c1}0.1862 (1) & \cellcolor{c2}0.1904 (2)\\
DTLZ6 & 5 & 0.3447 (4) & 0.3345 (3) & \cellcolor{c1}0.2897 (1) & \cellcolor{c2}0.3284 (2)\\
DTLZ6 & 6 & \cellcolor{c2}0.1661 (2) & \cellcolor{c1}0.1371 (1) & 0.2909 (4) & 0.1780 (3)\\
\midrule
DTLZ7 & 2 & 0.0117 (3) & \cellcolor{c2}0.0093 (2) & \cellcolor{c1}0.0078 (1) & 0.0177 (4)\\
DTLZ7 & 3 & 0.0355 (3) & 0.0369 (4) & \cellcolor{c1}0.0112 (1) & \cellcolor{c2}0.0263 (2)\\
DTLZ7 & 4 & 0.0400 (3) & 0.0402 (4) & \cellcolor{c1}0.0234 (1) & \cellcolor{c2}0.0326 (2)\\
DTLZ7 & 5 & \cellcolor{c2}0.1431 (2) & 0.5936 (3) & \cellcolor{c1}0.0527 (1) & 0.7867 (4)\\
DTLZ7 & 6 & \cellcolor{c2}0.1558 (2) & 0.6623 (3) & \cellcolor{c1}0.0711 (1) & 0.8610 (4)\\
\midrule
SDTLZ1 & 2 & \cellcolor{c1}0.0191 (1) & \cellcolor{c2}0.0624 (2) & 0.0954 (3) & 0.2357 (4)\\
SDTLZ1 & 3 & 0.0354 (3) & \cellcolor{c2}0.0291 (2) & \cellcolor{c1}0.0274 (1) & 0.0930 (4)\\
SDTLZ1 & 4 & \cellcolor{c2}0.0582 (2) & \cellcolor{c1}0.0570 (1) & 0.0932 (3) & 0.1345 (4)\\
SDTLZ1 & 5 & \cellcolor{c1}0.0791 (1) & \cellcolor{c2}0.0808 (2) & 0.1219 (3) & 3.1722 (4)\\
SDTLZ1 & 6 & \cellcolor{c1}0.1441 (1) & \cellcolor{c2}0.1512 (2) & 0.4296 (3) & 20.5507 (4)\\
\midrule
SDTLZ2 & 2 & 0.0753 (3) & \cellcolor{c2}0.0134 (2) & \cellcolor{c1}0.0015 (1) & 0.1755 (4)\\
SDTLZ2 & 3 & 0.0330 (3) & \cellcolor{c2}0.0174 (2) & \cellcolor{c1}0.0090 (1) & 0.1253 (4)\\
SDTLZ2 & 4 & \cellcolor{c2}0.0445 (2) & 0.0458 (3) & \cellcolor{c1}0.0249 (1) & 0.1065 (4)\\
SDTLZ2 & 5 & 0.1178 (3) & \cellcolor{c2}0.1128 (2) & \cellcolor{c1}0.0723 (1) & 0.1839 (4)\\
SDTLZ2 & 6 & 0.1710 (3) & \cellcolor{c2}0.1657 (2) & \cellcolor{c1}0.1056 (1) & 0.2912 (4)\\
\midrule
SDTLZ3 & 2 & 0.0943 (3) & \cellcolor{c2}0.0144 (2) & \cellcolor{c1}0.0110 (1) & 0.1466 (4)\\
SDTLZ3 & 3 & \cellcolor{c2}0.0501 (2) & \cellcolor{c1}0.0377 (1) & 0.1464 (3) & 0.2045 (4)\\
SDTLZ3 & 4 & \cellcolor{c2}0.1619 (2) & \cellcolor{c1}0.0617 (1) & 1.8442 (4) & 0.5813 (3)\\
SDTLZ3 & 5 & \cellcolor{c2}0.1201 (2) & \cellcolor{c1}0.0999 (1) & 0.7759 (3) & 11.8076 (4)\\
SDTLZ3 & 6 & \cellcolor{c1}0.1436 (1) & \cellcolor{c2}0.2067 (2) & 1.5421 (3) & 51.2470 (4)\\
\midrule
SDTLZ4 & 2 & 0.2244 (3) & \cellcolor{c2}0.1928 (2) & \cellcolor{c1}0.1880 (1) & 0.2721 (4)\\
SDTLZ4 & 3 & \cellcolor{c2}0.0673 (2) & 0.0972 (3) & \cellcolor{c1}0.0536 (1) & 0.1307 (4)\\
SDTLZ4 & 4 & \cellcolor{c1}0.0614 (1) & 0.0637 (3) & \cellcolor{c2}0.0626 (2) & 0.1246 (4)\\
SDTLZ4 & 5 & \cellcolor{c2}0.0961 (2) & 0.1099 (3) & \cellcolor{c1}0.0654 (1) & 0.2168 (4)\\
SDTLZ4 & 6 & 0.1519 (3) & \cellcolor{c2}0.1403 (2) & \cellcolor{c1}0.0820 (1) & 0.3584 (4)\\
\midrule
IDTLZ1 & 2 & \cellcolor{c2}0.0224 (2) & 0.0634 (4) & 0.0634 (3) & \cellcolor{c1}0.0012 (1)\\
IDTLZ1 & 3 & 0.0316 (3) & 0.1146 (4) & \cellcolor{c2}0.0276 (2) & \cellcolor{c1}0.0128 (1)\\
IDTLZ1 & 4 & \cellcolor{c2}0.0685 (2) & 0.2203 (4) & 0.0692 (3) & \cellcolor{c1}0.0354 (1)\\
IDTLZ1 & 5 & \cellcolor{c2}0.0627 (2) & 0.2890 (4) & 0.1157 (3) & \cellcolor{c1}0.0466 (1)\\
IDTLZ1 & 6 & \cellcolor{c2}0.0742 (2) & 0.2843 (4) & 0.1261 (3) & \cellcolor{c1}0.0613 (1)\\
\midrule
IDTLZ2 & 2 & 0.1463 (3) & 0.1877 (4) & \cellcolor{c2}0.0001 (2) & \cellcolor{c1}0.0001 (1)\\
IDTLZ2 & 3 & 0.0327 (3) & 0.0423 (4) & \cellcolor{c2}0.0175 (2) & \cellcolor{c1}0.0117 (1)\\
IDTLZ2 & 4 & 0.0453 (4) & 0.0398 (3) & \cellcolor{c1}0.0210 (1) & \cellcolor{c2}0.0249 (2)\\
IDTLZ2 & 5 & 0.0660 (4) & 0.0487 (3) & \cellcolor{c1}0.0384 (1) & \cellcolor{c2}0.0450 (2)\\
IDTLZ2 & 6 & 0.0996 (4) & 0.0914 (3) & \cellcolor{c2}0.0451 (2) & \cellcolor{c1}0.0402 (1)\\
\midrule
IDTLZ3 & 2 & 0.1606 (3) & 0.1851 (4) & \cellcolor{c2}0.0922 (2) & \cellcolor{c1}0.0004 (1)\\
IDTLZ3 & 3 & 0.0780 (4) & 0.0742 (3) & \cellcolor{c2}0.0552 (2) & \cellcolor{c1}0.0307 (1)\\
IDTLZ3 & 4 & 0.0651 (3) & \cellcolor{c2}0.0590 (2) & 0.1652 (4) & \cellcolor{c1}0.0466 (1)\\
IDTLZ3 & 5 & 0.0542 (3) & \cellcolor{c2}0.0541 (2) & 0.2271 (4) & \cellcolor{c1}0.0499 (1)\\
IDTLZ3 & 6 & \cellcolor{c1}0.0909 (1) & 0.1537 (3) & 0.3270 (4) & \cellcolor{c2}0.1018 (2)\\
\midrule
IDTLZ4 & 2 & 0.1099 (4) & 0.0701 (3) & \cellcolor{c2}0.0016 (2) & \cellcolor{c1}0.0016 (1)\\
IDTLZ4 & 3 & 0.1502 (4) & 0.0898 (3) & \cellcolor{c2}0.0380 (2) & \cellcolor{c1}0.0320 (1)\\
IDTLZ4 & 4 & 0.0595 (4) & \cellcolor{c2}0.0329 (2) & \cellcolor{c1}0.0248 (1) & 0.0474 (3)\\
IDTLZ4 & 5 & 0.1464 (4) & \cellcolor{c1}0.0462 (1) & 0.0746 (3) & \cellcolor{c2}0.0543 (2)\\
IDTLZ4 & 6 & 0.2423 (4) & 0.0779 (3) & \cellcolor{c2}0.0774 (2) & \cellcolor{c1}0.0555 (1)\\
\bottomrule
\end{tabular}
}
\end{table*}

\begin{table*}[t]
  \renewcommand{\arraystretch}{0.2} 
\centering
\caption{\small Average IGD$^+$-C values of r-NSGA-II with the three normalization methods (PP, BP, and BA) and with no normalization method (NO) on the DTLZ, SDTLZ, and IDTLZ problems with $m \in \{2, 3, 4, 5, 6\}$. The results at $50\,000$ function evaluations are shown. The numbers in parentheses indicate the rankings based on the Friedman test. The best and second best data are highlighted by dark gray and gray, respectively.}
  \label{tab:sup_r2nsga2_type1}  
{\scriptsize
  \begin{tabular}{cccccccc}
\toprule
Problem & $m$ & PP & BP & BA & NO\\
\midrule
DTLZ1 & 2 & \cellcolor{c2}0.0036 (2) & 0.0036 (3) & 0.0036 (4) & \cellcolor{c1}0.0036 (1)\\
DTLZ1 & 3 & \cellcolor{c2}0.0428 (2) & 0.0446 (3) & 0.0456 (4) & \cellcolor{c1}0.0427 (1)\\
DTLZ1 & 4 & 2.3342 (3) & \cellcolor{c2}2.1585 (2) & \cellcolor{c1}1.7885 (1) & 6.4697 (4)\\
DTLZ1 & 5 & 14.6804 (3) & \cellcolor{c2}12.4316 (2) & \cellcolor{c1}5.9934 (1) & 48.1043 (4)\\
DTLZ1 & 6 & 17.0352 (3) & \cellcolor{c2}16.4495 (2) & \cellcolor{c1}8.4404 (1) & 58.7930 (4)\\
\midrule
DTLZ2 & 2 & 0.0443 (3) & \cellcolor{c1}0.0299 (1) & 0.0612 (4) & \cellcolor{c2}0.0440 (2)\\
DTLZ2 & 3 & \cellcolor{c1}0.0148 (1) & \cellcolor{c2}0.0487 (2) & 0.0642 (4) & 0.0575 (3)\\
DTLZ2 & 4 & \cellcolor{c1}0.0334 (1) & \cellcolor{c2}0.0714 (2) & 0.0857 (4) & 0.0788 (3)\\
DTLZ2 & 5 & \cellcolor{c1}0.1341 (1) & \cellcolor{c2}0.1730 (2) & 0.1960 (4) & 0.1874 (3)\\
DTLZ2 & 6 & \cellcolor{c1}0.1689 (1) & \cellcolor{c2}0.2135 (2) & 0.2226 (4) & 0.2166 (3)\\
\midrule
DTLZ3 & 2 & \cellcolor{c1}0.0068 (1) & 0.0078 (3) & \cellcolor{c2}0.0073 (2) & 0.0079 (4)\\
DTLZ3 & 3 & \cellcolor{c2}0.2113 (2) & \cellcolor{c1}0.1713 (1) & 0.2323 (3) & 0.3145 (4)\\
DTLZ3 & 4 & 12.8218 (3) & \cellcolor{c2}11.5672 (2) & \cellcolor{c1}6.4736 (1) & 38.1408 (4)\\
DTLZ3 & 5 & 28.3574 (3) & \cellcolor{c2}26.9617 (2) & \cellcolor{c1}12.2312 (1) & 132.2738 (4)\\
DTLZ3 & 6 & \cellcolor{c2}36.4303 (2) & 37.6427 (3) & \cellcolor{c1}14.9660 (1) & 194.1299 (4)\\
\midrule
DTLZ4 & 2 & \cellcolor{c2}0.0949 (2) & 0.1077 (3) & 0.1345 (4) & \cellcolor{c1}0.0815 (1)\\
DTLZ4 & 3 & 0.0246 (4) & 0.0162 (3) & \cellcolor{c1}0.0116 (1) & \cellcolor{c2}0.0130 (2)\\
DTLZ4 & 4 & 0.0568 (4) & 0.0425 (3) & \cellcolor{c1}0.0247 (1) & \cellcolor{c2}0.0301 (2)\\
DTLZ4 & 5 & 0.1071 (3) & \cellcolor{c2}0.0925 (2) & 0.1076 (4) & \cellcolor{c1}0.0894 (1)\\
DTLZ4 & 6 & 0.1554 (4) & \cellcolor{c2}0.1394 (2) & 0.1534 (3) & \cellcolor{c1}0.1335 (1)\\
\midrule
DTLZ5 & 2 & \cellcolor{c1}0.0171 (1) & \cellcolor{c2}0.0436 (2) & 0.0512 (3) & 0.0520 (4)\\
DTLZ5 & 3 & \cellcolor{c1}0.0117 (1) & \cellcolor{c2}0.0670 (2) & 0.0777 (4) & 0.0746 (3)\\
DTLZ5 & 4 & \cellcolor{c1}0.0108 (1) & \cellcolor{c2}0.1318 (2) & 0.1635 (4) & 0.1530 (3)\\
DTLZ5 & 5 & \cellcolor{c1}0.3122 (1) & 0.3452 (3) & 0.3478 (4) & \cellcolor{c2}0.3281 (2)\\
DTLZ5 & 6 & \cellcolor{c1}0.0186 (1) & \cellcolor{c2}0.0544 (2) & 0.0668 (4) & 0.0642 (3)\\
\midrule
DTLZ6 & 2 & 0.1154 (4) & \cellcolor{c1}0.1048 (1) & 0.1117 (3) & \cellcolor{c2}0.1116 (2)\\
DTLZ6 & 3 & \cellcolor{c2}0.1422 (2) & 0.1431 (3) & \cellcolor{c1}0.1301 (1) & 0.2122 (4)\\
DTLZ6 & 4 & 0.4483 (3) & \cellcolor{c2}0.3981 (2) & \cellcolor{c1}0.1850 (1) & 4.3222 (4)\\
DTLZ6 & 5 & 1.0532 (3) & \cellcolor{c2}0.9364 (2) & \cellcolor{c1}0.4336 (1) & 7.9007 (4)\\
DTLZ6 & 6 & 1.0786 (3) & \cellcolor{c2}1.0122 (2) & \cellcolor{c1}0.1500 (1) & 8.3415 (4)\\
\midrule
DTLZ7 & 2 & \cellcolor{c1}0.0043 (1) & 0.0124 (3) & 0.0194 (4) & \cellcolor{c2}0.0123 (2)\\
DTLZ7 & 3 & \cellcolor{c1}0.0201 (1) & 0.0383 (3) & 0.0659 (4) & \cellcolor{c2}0.0371 (2)\\
DTLZ7 & 4 & \cellcolor{c1}0.0219 (1) & \cellcolor{c2}0.0334 (2) & 0.0448 (4) & 0.0363 (3)\\
DTLZ7 & 5 & \cellcolor{c1}0.0252 (1) & \cellcolor{c2}0.0296 (2) & 0.0401 (3) & 0.2315 (4)\\
DTLZ7 & 6 & \cellcolor{c2}0.0342 (2) & \cellcolor{c1}0.0337 (1) & 0.0458 (3) & 0.3509 (4)\\
\midrule
SDTLZ1 & 2 & 0.0036 (3) & \cellcolor{c2}0.0036 (2) & 0.0036 (4) & \cellcolor{c1}0.0035 (1)\\
SDTLZ1 & 3 & \cellcolor{c1}0.0432 (1) & 0.0451 (3) & 0.0494 (4) & \cellcolor{c2}0.0432 (2)\\
SDTLZ1 & 4 & 2.3429 (3) & \cellcolor{c2}2.2283 (2) & \cellcolor{c1}1.7069 (1) & 6.6636 (4)\\
SDTLZ1 & 5 & 14.5382 (3) & \cellcolor{c2}12.4962 (2) & \cellcolor{c1}6.0407 (1) & 46.6699 (4)\\
SDTLZ1 & 6 & 17.0352 (3) & \cellcolor{c2}16.0408 (2) & \cellcolor{c1}8.3361 (1) & 59.7730 (4)\\
\midrule
SDTLZ2 & 2 & 0.0443 (3) & \cellcolor{c2}0.0299 (2) & 0.0612 (4) & \cellcolor{c1}0.0036 (1)\\
SDTLZ2 & 3 & \cellcolor{c1}0.0148 (1) & 0.0487 (3) & 0.0642 (4) & \cellcolor{c2}0.0436 (2)\\
SDTLZ2 & 4 & \cellcolor{c1}0.0334 (1) & \cellcolor{c2}0.0714 (2) & 0.0857 (3) & 0.1033 (4)\\
SDTLZ2 & 5 & \cellcolor{c1}0.1341 (1) & \cellcolor{c2}0.1743 (2) & 0.1968 (4) & 0.1960 (3)\\
SDTLZ2 & 6 & \cellcolor{c1}0.1689 (1) & \cellcolor{c2}0.2135 (2) & 0.2226 (3) & 0.7566 (4)\\
\midrule
SDTLZ3 & 2 & \cellcolor{c1}0.0068 (1) & 0.0078 (3) & \cellcolor{c2}0.0073 (2) & 0.0079 (4)\\
SDTLZ3 & 3 & \cellcolor{c1}0.1227 (1) & \cellcolor{c2}0.1724 (2) & 0.2327 (3) & 0.3881 (4)\\
SDTLZ3 & 4 & 12.8206 (3) & \cellcolor{c2}11.7217 (2) & \cellcolor{c1}6.4628 (1) & 37.0182 (4)\\
SDTLZ3 & 5 & 27.8065 (3) & \cellcolor{c2}26.6137 (2) & \cellcolor{c1}12.2967 (1) & 130.3618 (4)\\
SDTLZ3 & 6 & \cellcolor{c2}36.3027 (2) & 38.9605 (3) & \cellcolor{c1}15.6219 (1) & 198.1287 (4)\\
\midrule
SDTLZ4 & 2 & \cellcolor{c1}0.0949 (1) & \cellcolor{c2}0.1077 (2) & 0.1345 (3) & 0.1490 (4)\\
SDTLZ4 & 3 & 0.0246 (3) & \cellcolor{c2}0.0162 (2) & \cellcolor{c1}0.0116 (1) & 0.0445 (4)\\
SDTLZ4 & 4 & 0.0568 (3) & \cellcolor{c2}0.0425 (2) & \cellcolor{c1}0.0247 (1) & 0.1052 (4)\\
SDTLZ4 & 5 & \cellcolor{c2}0.1075 (2) & \cellcolor{c1}0.0929 (1) & 0.1077 (3) & 0.8364 (4)\\
SDTLZ4 & 6 & 0.1554 (3) & \cellcolor{c1}0.1394 (1) & \cellcolor{c2}0.1534 (2) & 1.7815 (4)\\
\midrule
IDTLZ1 & 2 & \cellcolor{c1}0.0035 (1) & \cellcolor{c2}0.0035 (2) & 0.0038 (4) & 0.0037 (3)\\
IDTLZ1 & 3 & 0.0449 (3) & \cellcolor{c2}0.0440 (2) & 0.0494 (4) & \cellcolor{c1}0.0433 (1)\\
IDTLZ1 & 4 & 0.0988 (3) & 0.1031 (4) & \cellcolor{c1}0.0967 (1) & \cellcolor{c2}0.0980 (2)\\
IDTLZ1 & 5 & \cellcolor{c1}0.1478 (1) & 0.1530 (4) & \cellcolor{c2}0.1497 (2) & 0.1512 (3)\\
IDTLZ1 & 6 & \cellcolor{c1}0.1719 (1) & 0.1756 (3) & 0.1763 (4) & \cellcolor{c2}0.1736 (2)\\
\midrule
IDTLZ2 & 2 & 0.0674 (3) & \cellcolor{c2}0.0666 (2) & 0.0975 (4) & \cellcolor{c1}0.0631 (1)\\
IDTLZ2 & 3 & \cellcolor{c1}0.1093 (1) & \cellcolor{c2}0.1478 (2) & 0.2093 (4) & 0.1784 (3)\\
IDTLZ2 & 4 & \cellcolor{c1}0.0292 (1) & \cellcolor{c2}0.0620 (2) & 0.0737 (4) & 0.0647 (3)\\
IDTLZ2 & 5 & \cellcolor{c1}0.0279 (1) & \cellcolor{c2}0.0449 (2) & 0.0522 (4) & 0.0501 (3)\\
IDTLZ2 & 6 & \cellcolor{c1}0.0432 (1) & \cellcolor{c2}0.0687 (2) & 0.0745 (4) & 0.0737 (3)\\
\midrule
IDTLZ3 & 2 & 0.0011 (3) & \cellcolor{c1}0.0011 (1) & 0.0093 (4) & \cellcolor{c2}0.0011 (2)\\
IDTLZ3 & 3 & \cellcolor{c2}0.0577 (2) & \cellcolor{c1}0.0546 (1) & 0.0580 (3) & 0.0597 (4)\\
IDTLZ3 & 4 & \cellcolor{c2}0.2107 (2) & 0.2845 (4) & 0.2732 (3) & \cellcolor{c1}0.1926 (1)\\
IDTLZ3 & 5 & 0.5555 (3) & \cellcolor{c2}0.5043 (2) & \cellcolor{c1}0.4351 (1) & 0.6714 (4)\\
IDTLZ3 & 6 & 1.3101 (4) & \cellcolor{c1}0.9059 (1) & \cellcolor{c2}1.0111 (2) & 1.1237 (3)\\
\midrule
IDTLZ4 & 2 & \cellcolor{c2}0.0153 (2) & \cellcolor{c1}0.0101 (1) & 0.0290 (3) & 0.0481 (4)\\
IDTLZ4 & 3 & 0.0682 (4) & \cellcolor{c2}0.0245 (2) & 0.0377 (3) & \cellcolor{c1}0.0185 (1)\\
IDTLZ4 & 4 & \cellcolor{c2}0.1120 (2) & 0.1407 (4) & 0.1157 (3) & \cellcolor{c1}0.0468 (1)\\
IDTLZ4 & 5 & \cellcolor{c2}0.1382 (2) & 0.1407 (3) & 0.1464 (4) & \cellcolor{c1}0.0903 (1)\\
IDTLZ4 & 6 & 0.1798 (3) & 0.2614 (4) & \cellcolor{c2}0.1635 (2) & \cellcolor{c1}0.1232 (1)\\
\bottomrule
\end{tabular}
}
\end{table*}

\begin{table*}[t]
  \renewcommand{\arraystretch}{0.2} 
\centering
\caption{\small Average IGD$^+$-C values of MOEA/D-NUMS with the three normalization methods (PP, BP, and BA) and with no normalization method (NO) on the DTLZ, SDTLZ, and IDTLZ problems with $m \in \{2, 3, 4, 5, 6\}$. The results at $50\,000$ function evaluations are shown. The numbers in parentheses indicate the rankings based on the Friedman test. The best and second best data are highlighted by dark gray and gray, respectively.}
  \label{tab:sup_nums_type1}  
{\scriptsize
  \begin{tabular}{cccccccc}
\toprule
Problem & $m$ & PP & BP & BA & NO\\
\midrule
DTLZ1 & 2 & 0.0160 (3) & \cellcolor{c1}0.0058 (1) & 0.0544 (4) & \cellcolor{c2}0.0157 (2)\\
DTLZ1 & 3 & \cellcolor{c2}0.0350 (2) & 2.0108 (4) & 0.0515 (3) & \cellcolor{c1}0.0330 (1)\\
DTLZ1 & 4 & \cellcolor{c2}0.0385 (2) & 1.0128 (4) & 0.0798 (3) & \cellcolor{c1}0.0356 (1)\\
DTLZ1 & 5 & 0.0406 (3) & \cellcolor{c1}0.0388 (1) & 0.0830 (4) & \cellcolor{c2}0.0393 (2)\\
DTLZ1 & 6 & 1.0592 (4) & \cellcolor{c1}0.0371 (1) & 0.0942 (3) & \cellcolor{c2}0.0380 (2)\\
\midrule
DTLZ2 & 2 & 0.1920 (4) & \cellcolor{c1}0.0096 (1) & 0.0335 (3) & \cellcolor{c2}0.0265 (2)\\
DTLZ2 & 3 & 0.0542 (4) & \cellcolor{c1}0.0278 (1) & 0.0356 (3) & \cellcolor{c2}0.0312 (2)\\
DTLZ2 & 4 & 0.0494 (4) & \cellcolor{c1}0.0396 (1) & 0.0433 (3) & \cellcolor{c2}0.0404 (2)\\
DTLZ2 & 5 & 0.0757 (4) & \cellcolor{c1}0.0335 (1) & 0.0460 (3) & \cellcolor{c2}0.0375 (2)\\
DTLZ2 & 6 & 0.0864 (4) & \cellcolor{c2}0.0415 (2) & 0.0518 (3) & \cellcolor{c1}0.0410 (1)\\
\midrule
DTLZ3 & 2 & 2.6618 (3) & 64.2259 (4) & \cellcolor{c1}0.0870 (1) & \cellcolor{c2}0.1046 (2)\\
DTLZ3 & 3 & 8.4386 (4) & 0.7566 (3) & \cellcolor{c2}0.3516 (2) & \cellcolor{c1}0.3133 (1)\\
DTLZ3 & 4 & 86.7709 (4) & 38.5490 (3) & \cellcolor{c2}0.5312 (2) & \cellcolor{c1}0.3010 (1)\\
DTLZ3 & 5 & 11.5435 (3) & 13.3031 (4) & \cellcolor{c1}0.3112 (1) & \cellcolor{c2}0.4147 (2)\\
DTLZ3 & 6 & 53.1476 (4) & 43.1581 (3) & \cellcolor{c1}0.3883 (1) & \cellcolor{c2}1.2338 (2)\\
\midrule
DTLZ4 & 2 & 0.2359 (4) & 0.0878 (3) & \cellcolor{c2}0.0591 (2) & \cellcolor{c1}0.0265 (1)\\
DTLZ4 & 3 & 0.1358 (3) & 0.2006 (4) & \cellcolor{c2}0.0671 (2) & \cellcolor{c1}0.0431 (1)\\
DTLZ4 & 4 & 0.2929 (3) & 0.3638 (4) & \cellcolor{c2}0.0713 (2) & \cellcolor{c1}0.0517 (1)\\
DTLZ4 & 5 & 0.4174 (3) & 0.4333 (4) & \cellcolor{c2}0.0785 (2) & \cellcolor{c1}0.0647 (1)\\
DTLZ4 & 6 & 0.5094 (4) & 0.4711 (3) & \cellcolor{c2}0.0868 (2) & \cellcolor{c1}0.0861 (1)\\
\midrule
DTLZ5 & 2 & 0.0577 (4) & \cellcolor{c1}0.0273 (1) & \cellcolor{c2}0.0439 (2) & 0.0443 (3)\\
DTLZ5 & 3 & \cellcolor{c1}0.0159 (1) & 0.0336 (3) & \cellcolor{c2}0.0276 (2) & 0.0429 (4)\\
DTLZ5 & 4 & \cellcolor{c1}0.1195 (1) & \cellcolor{c2}0.1266 (2) & 0.1269 (3) & 0.1323 (4)\\
DTLZ5 & 5 & \cellcolor{c1}0.0364 (1) & 0.1539 (3) & \cellcolor{c2}0.1117 (2) & 0.2067 (4)\\
DTLZ5 & 6 & 0.1664 (4) & \cellcolor{c1}0.1044 (1) & \cellcolor{c2}0.1083 (2) & 0.1183 (3)\\
\midrule
DTLZ6 & 2 & 0.0576 (4) & \cellcolor{c1}0.0266 (1) & 0.0558 (3) & \cellcolor{c2}0.0443 (2)\\
DTLZ6 & 3 & \cellcolor{c1}0.0158 (1) & 0.0332 (3) & \cellcolor{c2}0.0273 (2) & 0.0429 (4)\\
DTLZ6 & 4 & \cellcolor{c2}0.1195 (2) & 0.1263 (3) & \cellcolor{c1}0.1189 (1) & 0.1323 (4)\\
DTLZ6 & 5 & \cellcolor{c1}0.0154 (1) & 0.1504 (3) & \cellcolor{c2}0.0982 (2) & 0.2080 (4)\\
DTLZ6 & 6 & \cellcolor{c2}0.0022 (2) & \cellcolor{c1}0.0018 (1) & 0.0350 (3) & 0.0747 (4)\\
\midrule
DTLZ7 & 2 & \cellcolor{c2}0.0242 (2) & 0.0302 (3) & \cellcolor{c1}0.0236 (1) & 0.0309 (4)\\
DTLZ7 & 3 & 0.0317 (3) & 0.0366 (4) & \cellcolor{c1}0.0265 (1) & \cellcolor{c2}0.0280 (2)\\
DTLZ7 & 4 & 0.1162 (4) & \cellcolor{c2}0.0985 (2) & \cellcolor{c1}0.0533 (1) & 0.1008 (3)\\
DTLZ7 & 5 & 0.6958 (4) & \cellcolor{c2}0.6536 (2) & \cellcolor{c1}0.1386 (1) & 0.6646 (3)\\
DTLZ7 & 6 & 0.7683 (4) & \cellcolor{c2}0.7142 (2) & \cellcolor{c1}0.1810 (1) & 0.7297 (3)\\
\midrule
SDTLZ1 & 2 & \cellcolor{c2}0.2656 (2) & 4.9426 (4) & \cellcolor{c1}0.2065 (1) & 0.3000 (3)\\
SDTLZ1 & 3 & 23.2349 (3) & 24.7532 (4) & \cellcolor{c1}0.0338 (1) & \cellcolor{c2}0.0833 (2)\\
SDTLZ1 & 4 & 32.9092 (3) & 34.7875 (4) & \cellcolor{c1}0.0511 (1) & \cellcolor{c2}0.1138 (2)\\
SDTLZ1 & 5 & 42.9804 (3) & 48.2908 (4) & \cellcolor{c1}0.0722 (1) & \cellcolor{c2}0.0964 (2)\\
SDTLZ1 & 6 & 56.0657 (4) & 55.5353 (3) & \cellcolor{c1}0.1196 (1) & \cellcolor{c2}0.1362 (2)\\
\midrule
SDTLZ2 & 2 & 0.1920 (3) & \cellcolor{c1}0.0679 (1) & \cellcolor{c2}0.1630 (2) & 0.1920 (3)\\
SDTLZ2 & 3 & 0.0867 (3) & \cellcolor{c2}0.0648 (2) & \cellcolor{c1}0.0609 (1) & 0.1010 (4)\\
SDTLZ2 & 4 & 0.0999 (4) & \cellcolor{c1}0.0549 (1) & \cellcolor{c2}0.0554 (2) & 0.0919 (3)\\
SDTLZ2 & 5 & 0.1085 (3) & \cellcolor{c1}0.0981 (1) & \cellcolor{c2}0.0993 (2) & 0.1224 (4)\\
SDTLZ2 & 6 & 0.1174 (3) & \cellcolor{c2}0.1085 (2) & \cellcolor{c1}0.1079 (1) & 0.1744 (4)\\
\midrule
SDTLZ3 & 2 & 99.3743 (3) & 105.0844 (4) & \cellcolor{c1}0.1793 (1) & \cellcolor{c2}0.2262 (2)\\
SDTLZ3 & 3 & 170.2077 (3) & 173.2871 (4) & \cellcolor{c2}0.3696 (2) & \cellcolor{c1}0.2450 (1)\\
SDTLZ3 & 4 & 220.5080 (4) & 209.6023 (3) & \cellcolor{c2}1.7445 (2) & \cellcolor{c1}0.5201 (1)\\
SDTLZ3 & 5 & 296.4722 (4) & 292.6652 (3) & \cellcolor{c2}6.8895 (2) & \cellcolor{c1}2.3826 (1)\\
SDTLZ3 & 6 & 303.6419 (4) & 289.5561 (3) & \cellcolor{c2}8.1207 (2) & \cellcolor{c1}5.0503 (1)\\
\midrule
SDTLZ4 & 2 & 0.2827 (4) & \cellcolor{c2}0.2190 (2) & 0.2245 (3) & \cellcolor{c1}0.1920 (1)\\
SDTLZ4 & 3 & 0.2521 (3) & 0.2738 (4) & \cellcolor{c1}0.0713 (1) & \cellcolor{c2}0.1015 (2)\\
SDTLZ4 & 4 & 0.2995 (3) & 0.3414 (4) & \cellcolor{c1}0.0812 (1) & \cellcolor{c2}0.1035 (2)\\
SDTLZ4 & 5 & 0.3247 (3) & 0.3431 (4) & \cellcolor{c1}0.1076 (1) & \cellcolor{c2}0.1399 (2)\\
SDTLZ4 & 6 & 0.4435 (4) & 0.4410 (3) & \cellcolor{c1}0.1238 (1) & \cellcolor{c2}0.2016 (2)\\
\midrule
IDTLZ1 & 2 & \cellcolor{c2}0.0159 (2) & 1.5669 (4) & 0.0575 (3) & \cellcolor{c1}0.0157 (1)\\
IDTLZ1 & 3 & 0.0495 (4) & \cellcolor{c2}0.0226 (2) & 0.0382 (3) & \cellcolor{c1}0.0169 (1)\\
IDTLZ1 & 4 & 0.0984 (4) & \cellcolor{c2}0.0636 (2) & 0.0768 (3) & \cellcolor{c1}0.0320 (1)\\
IDTLZ1 & 5 & 0.1201 (4) & 0.1136 (3) & \cellcolor{c2}0.1081 (2) & \cellcolor{c1}0.0595 (1)\\
IDTLZ1 & 6 & 0.1350 (3) & 0.1468 (4) & \cellcolor{c2}0.1209 (2) & \cellcolor{c1}0.0966 (1)\\
\midrule
IDTLZ2 & 2 & 0.0885 (3) & 0.1104 (4) & \cellcolor{c1}0.0399 (1) & \cellcolor{c2}0.0401 (2)\\
IDTLZ2 & 3 & 0.0235 (3) & 0.0283 (4) & \cellcolor{c2}0.0175 (2) & \cellcolor{c1}0.0152 (1)\\
IDTLZ2 & 4 & 0.0401 (4) & 0.0395 (3) & \cellcolor{c2}0.0378 (2) & \cellcolor{c1}0.0375 (1)\\
IDTLZ2 & 5 & \cellcolor{c1}0.0532 (1) & \cellcolor{c2}0.0547 (2) & 0.0548 (3) & 0.0548 (4)\\
IDTLZ2 & 6 & 0.0676 (4) & 0.0643 (3) & \cellcolor{c2}0.0418 (2) & \cellcolor{c1}0.0389 (1)\\
\midrule
IDTLZ3 & 2 & 5.1163 (4) & 0.2589 (3) & \cellcolor{c2}0.1839 (2) & \cellcolor{c1}0.0546 (1)\\
IDTLZ3 & 3 & 17.4293 (4) & 1.3769 (3) & \cellcolor{c2}0.2064 (2) & \cellcolor{c1}0.0575 (1)\\
IDTLZ3 & 4 & 15.0517 (4) & 1.9969 (3) & \cellcolor{c1}0.4012 (1) & \cellcolor{c2}0.6838 (2)\\
IDTLZ3 & 5 & 7.6854 (4) & 3.6114 (3) & \cellcolor{c2}0.8347 (2) & \cellcolor{c1}0.5613 (1)\\
IDTLZ3 & 6 & 6.0567 (4) & 2.1248 (3) & \cellcolor{c2}1.0719 (2) & \cellcolor{c1}1.0214 (1)\\
\midrule
IDTLZ4 & 2 & 0.0927 (3) & 0.1131 (4) & \cellcolor{c1}0.0249 (1) & \cellcolor{c2}0.0401 (2)\\
IDTLZ4 & 3 & 0.2693 (4) & 0.1254 (3) & \cellcolor{c2}0.0163 (2) & \cellcolor{c1}0.0152 (1)\\
IDTLZ4 & 4 & 0.2599 (4) & 0.0670 (3) & \cellcolor{c2}0.0376 (2) & \cellcolor{c1}0.0375 (1)\\
IDTLZ4 & 5 & 0.2955 (4) & 0.0548 (3) & \cellcolor{c1}0.0548 (1) & \cellcolor{c2}0.0548 (2)\\
IDTLZ4 & 6 & 0.3763 (4) & \cellcolor{c1}0.0427 (1) & 0.0842 (3) & \cellcolor{c2}0.0524 (2)\\
\bottomrule
\end{tabular}
}
\end{table*}

\end{document}